\documentclass{alt2027}
\usepackage{times}
\usepackage{mathtools}
\usepackage{booktabs}
\usepackage{array}
\usepackage{longtable}
\usepackage{flafter}
\pdfinfoomitdate=1 \pdftrailerid{} \pdfsuppressptexinfo=-1 %
\newif\ifdraftnotes\draftnotesfalse %
\newcommand{\pend}[1]{\ifdraftnotes{\color{red}\footnotesize[#1]}\fi}
\newcommand{\KL}{\mathrm{KL}}
\newcommand{\E}{\mathbb{E}}
\newcommand{\Bin}{\mathrm{Bin}}
\newcommand{\LH}{L_{H}}
\newcommand{\LF}{L_{F}}
\newcommand{\Hc}{\mathcal{H}}
\providecommand{\qed}{\hfill\ensuremath{\square}}\providecommand{\qedsymbol}{\ensuremath{\square}}
\title[KL Before and After Mixing]{KL Before and After Mixing: A Round-Complexity Separation in Parallel Sampling}
\author{\Name{Tianyi Wu} \Email{wutianyi@amazon.com}\\ \addr Amazon}
\hypersetup{pdftitle={KL Before and After Mixing: A Round-Complexity Separation in Parallel Sampling},pdfauthor={Tianyi Wu}}
\makeatletter
\def\ps@jmlrtps{\let\@mkboth\@gobbletwo\def\@oddhead{}\let\@evenhead\@oddhead\def\@oddfoot{}\let\@evenfoot\@oddfoot}
\makeatother
\begin{document}
\maketitle
\begin{abstract}
A parallel sampler can choose at random which coordinates to generate together. For a target $p$ and the output law $q_U$
of a schedule $U$ drawn afresh for each output, we compare the schedule-averaged divergence $\mathbb{E}_U D_{\mathrm{KL}}(p\|q_U)$
with the output-law divergence $D_{\mathrm{KL}}(p\|\mathbb{E}_U q_U)$, in nats. Rounds commit independent draws from exact
conditional marginals frozen at the round's start, each output's schedule is drawn before sampling from a finite mixture of schedules, and every scheduled round is charged. For the uniform
distribution on binary strings of even length $n$ with $n/2$ ones, the least expected number of rounds at a fixed tolerance
$\varepsilon>0$ is $\Theta_\varepsilon((\log n)^2)$ under the first criterion but $(e^{-\varepsilon}+o(1))\log_2\ln n$ under the
second, and the double-logarithmic order is attainable with the same round count on every run. For this target, a count-state recursion gives
the exact least output divergence at every round cap over all causal randomized policies. The separation extends to weighted
fixed-count targets with bounded weight ratios and quota fractions bounded away from zero and one, without permutation
symmetry, and to the full-support Beta$(1,1)$--Bernoulli target, where the sharp expected cost under the output-law criterion
is $(e^{-\varepsilon}/2+o(1))\log_2\log_2n$, attained with rare $n$-round runs. Prior work implies the order of the schedule-averaged cost on the balanced target and its
$O_\varepsilon((\log n)^2)$ upper bounds on both extensions.
\end{abstract}

\section{Introduction}\label{sec:intro}

Masked diffusion and any-order autoregressive models are increasingly decoded in parallel: each round fills several masked
positions, each drawn independently from the model's conditional marginal given the committed ones, so dependence within a
round is lost even when the model is exact. A \emph{schedule} says which positions are filled in which round; fewer rounds
mean fewer sequential model calls but lose more dependence, and a growing theory designs schedules to control the error
\citep{chen2026optimal,lavenant2025error,secchi2026schedule,zhao2026adaptation,wainwright2026data,wen2026conditional}.

\paragraph{Two ways to judge a random schedule.} Let $p$ be the target, $q_U$ the output law under a schedule $U$, and
$\mu$ the law of a random schedule. The \emph{schedule-averaged} divergence is
$H(\mu)=\E_{U\sim\mu}\KL(p\|q_U)=\KL(\mu\otimes p\,\|\,J_\mu)$, where $J_\mu(U,x)=\mu(U)q_U(x)$ is the sampler's joint law
of schedule and output and the benchmark $\mu\otimes p$ draws the schedule independently of a target sample; $H$ averages
over schedules and is not a worst case. The \emph{output-law} divergence $F(\mu)=\KL(p\|\E_{U\sim\mu}q_U)$ is the error of
the law the sampler produces when each output receives a fresh schedule that is then discarded. By joint convexity $F\le H$,
with equality for a fixed schedule, so a bound on $H$ also bounds $F$; it does not say whether $F$ can be made small with
fewer rounds. We ask how many rounds each criterion needs.

\begin{figure}[t]
\centering\includegraphics[width=\textwidth]{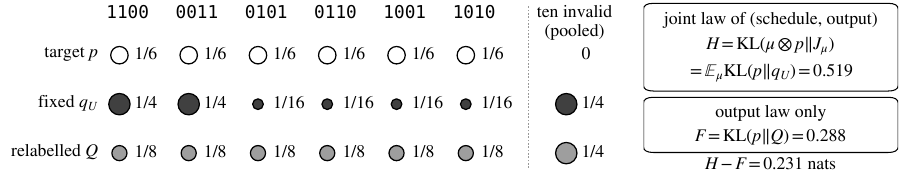}
\caption{The four-bit example. Columns: the six valid strings (two ones) and, pooled, the ten invalid ones; disc areas are
unconditional probabilities, and nothing is renormalized. The example implies no rate.}\label{fig:hf}
\end{figure}

\paragraph{A four-bit example.} Take the uniform law on the six strings of length four with two ones, sampled in two rounds
of two coordinates (Figure~\ref{fig:hf}). With fixed labels $U=(\{1,2\},\{3,4\})$, the output is invalid with probability
$\frac14$, and the valid strings $1100$ and $0011$ get $\frac14$ each against $\frac1{16}$ for the other four.
Permuting the labels uniformly before each run changes no labelled schedule's $\KL(p\|q_U)$, so $H\approx0.519$
is unchanged, but spreads the valid mass $\frac34$ equally: $F$ falls to $-\ln\Pr(\text{valid})=\ln\frac43\approx0.288$.
Relabelling removed the unequal weighting of valid strings; it repaired no invalid output.

\paragraph{The answer on the balanced slice.} In our model (Section~\ref{sec:setup}) rounds draw independently from exact
conditional marginals frozen at their start, each output's schedule is drawn before sampling from a finite mixture of
schedules, and every scheduled round is charged. On the uniform law on the \emph{balanced slice} of strings of even length $n$ with $n/2$ ones, a
target studied by \citet{wen2026conditional}, the least expected numbers of rounds under $H\le\varepsilon$ and under
$F\le\varepsilon$, at a fixed tolerance $\varepsilon>0$ in nats, are
\begin{equation*}
\LH(n,\varepsilon)=\Theta_\varepsilon\big((\log n)^2\big)\qquad\text{and}\qquad
\LF(n,\varepsilon)=\big(e^{-\varepsilon}+o(1)\big)\log_2\ln n
\end{equation*}
(Theorem~\ref{thm:sep}, Corollary~\ref{cor:t2}). The order of $\LH$ follows from \citet[Thm.~11]{wen2026conditional} with Markov's inequality; the
results for $\LF$ (its order, its leading coefficient and the converses behind them, which cover causal randomized policies
that adapt to the values already sampled) are, to our knowledge, new. The coefficient $e^{-\varepsilon}$ is a mean-cost effect: a
mixture can spend almost all of its error budget on a one-round schedule that usually fails (Corollary~\ref{cor:t2}(c)), which does not lower the hard
cap. Table~\ref{tab:main} adds the two extensions; on full support, randomized grid levels give a further factor $\frac12$ in the mean, beyond dilution (Section~\ref{sec:fs}).
The lower bounds hold for
this primitive (independent exact-marginal draws within each round) on known structured targets, not for unrestricted samplers; rounds, not computation,
are counted; the full-support $F$ upper-bound constants are astronomical, and the bounded-weight ones existence-only for weights without a linear-size equal-weight group; and small average prediction error need
not preserve the exact-oracle profile rate (Appendix~\ref{app:imperfect}; Section~\ref{sec:concl}).

\begin{table}[t]
\caption{Least number of rounds at a fixed tolerance $\varepsilon>0$, by design, criterion and cost; $\ell=\log_2\log_2n=\log_2\ln n-\log_2\ln2$.}\label{tab:main}
\centering\small
\begin{tabular}{@{}llll@{}}
\toprule
design, criterion, cost & balanced slice & full support & bounded weights\\
\midrule
fixed schedule, KL, length & $\Theta((\log n)^2)$ & $\Theta((\log n)^2)$ & $\Theta((\log n)^2)$\\
mixture, $H$, mean & $\Theta((\log n)^2)$ & $\Theta((\log n)^2)$ & $\Theta((\log n)^2)$\\
\midrule
relabelled profile, $F$, length & $\ell+O(1)$ & $\ell+O(1)$ & $\ell+O(1)^\ddagger$\\
causal policy, $F$, cap & $\ell+O(1)$ & $\Theta(\ell)^\dagger$ & $\ell+O(1)$\\
mixture, $F$, mean & $(e^{-\varepsilon}+o(1))\ell$ & $(\frac12e^{-\varepsilon}+o(1))\ell$ & $(e^{-\varepsilon}+o(1))\ell$\\
causal policy, $F$, $\E\tau$ & $(e^{-\varepsilon}+o(1))\ell$ & $(\frac12e^{-\varepsilon}+o(1))\ell$ & $(e^{-\varepsilon}+o(1))\ell$\\
\midrule
where & \S\ref{sec:bal}: Thm.~\ref{thm:sep}, Cor.~\ref{cor:hard}, \ref{cor:t2} & \S\ref{sec:fs}: Thm.~\ref{thm:fs}, \ref{thm:fsadapt}, Prop.~\ref{prop:hfs} & \S\ref{sec:tilt}: Thm.~\ref{thm:tilt}, App.~\ref{app:t1}, \ref{app:rf}\\
\bottomrule
\end{tabular}
\par\smallskip\raggedright\footnotesize
\emph{Length}: rounds on every run; \emph{mean}: scheduled mean $C(\mu)$, failed runs included; $\E\tau$: executed mean;
\emph{cap}: bound on the rounds of every run. Constants depend on $\varepsilon$ (for weights, also on the ratio and quota bounds, and for $^\ddagger$ on $\rho$). $^\dagger$Constant open: at least $(\frac12-o(1))\ell$, derived from the tail bound of Theorem~\ref{thm:fsadapt}(a),
and at most $\ell+O(1)$, from one relabelled profile. $^\ddagger$If at least a fixed fraction $\rho$ of the coordinates share one weight: one fixed labelled schedule, relabelled uniformly within
that group (Theorem~\ref{thm:tilt}(b)); otherwise open. Table~\ref{tab:classes} (Appendix~\ref{app:reference}) adds formal status and further classes.
\end{table}

\paragraph{Related work.} Our relabelled-profile constructions are the random-unmasking algorithm of
\citet{chen2026optimal} with particular block sizes, or finite mixtures of it. The schedule analyses cited above, other than
\citet{wen2026conditional}, bound $H$ or the KL between the laws of whole runs, which is at least $H$, and reach the output law through convexity or
data processing; \citet[Thm.~4.4]{feng2025theoretical} give an $\Omega(n)$ output-law lower bound for $F\le\ln2$ under random-position masking on a
hidden Markov target. For deterministic value-adaptive policies \citet[Thm.~11]{wen2026conditional} prove a
$\Theta_\varepsilon((\log n)^2)$ order on the balanced slice for the hard cap and for the mean number of rounds over draws from
the target (output-law KL in bits), and leave open whether their lower bounds extend to arbitrary randomized policies (their Remark~2).
\citet{li2026randomization} show in the same interface that such an extension can fail on a constructed family: a mixture of fixed
schedules beats every deterministic adaptive policy in accuracy at a cap of three rounds (their Thm.~4). On another family, a mixture attains
in three rounds an accuracy that fixed (nonadaptive) schedules need $\Omega(n/\log n)$ rounds to reach (their Cor.~9). On the
balanced slice randomization's advantage is in rounds, grows with $n$ and holds against deterministic adaptive policies: the least hard cap of
causal randomized policies is $\log_2\ln n+O_\varepsilon(1)$, against the deterministic $\Theta_\varepsilon((\log n)^2)$, and
the least executed mean is $(e^{-\varepsilon}+o(1))\log_2\ln n$, a count not compared with the mean of \citet{wen2026conditional} over draws from
the target (Appendix~\ref{app:related}).

\section{Model}\label{sec:setup}

\paragraph{One round.} A target $p$ is a law on $\{0,1\}^n$, and the sampler produces one output sample. Before each round
it has committed some coordinates; for each unrevealed coordinate $i$ an oracle gives the exact conditional probability under
$p$ that $x_i=1$ given the committed values. The round chooses a set $B$ of unrevealed coordinates (a nonempty block for
schedules; an adaptive policy may also choose an empty round, which is charged), draws each $x_i$, $i\in B$, independently with
its conditional probability frozen at the start of the round, and commits the draws. There is no joint draw, lookahead,
resampling or repair, so all error comes from the dependence that independent draws within a round ignore.

\paragraph{The balanced slice.} Our first target is the uniform law $p$ on $V=\{x\in\{0,1\}^n:\sum_ix_i=k\}$, $n=2k$. If
$M$ coordinates are unrevealed and $j$ of them must still equal one (the remaining quota), every unrevealed coordinate is one
with probability $j/M$. The \emph{count state} $(M,j)$ is \emph{legal} if $0\le j\le M$; a round that commits $b$ coordinates
at a legal state draws $X\sim\Bin(b,j/M)$ ones and moves to $(M-b,j-X)$; $\Bin(b,\theta)(x)$ denotes the probability of $x$
under $\Bin(b,\theta)$, zero outside $\{0,\dots,b\}$.

\paragraph{Schedules, mixtures and policies.} A labelled schedule $U=(B_1,\dots,B_r)$ is an ordered partition of $[n]$
into nonempty blocks, of length $|U|=r$ and \emph{profile} $b(U)=(|B_1|,\dots,|B_r|)$; $q_U$ is its output law. A schedule
mixture $\mu$ is a finite law on labelled schedules, drawn once before any coordinate is sampled and independently of all
oracle randomness, with output law $Q_\mu=\E_\mu q_U$; its support may grow with $n$, and describing or selecting it is not
charged. The \emph{uniform relabelling} of $U$ is the mixture of $\sigma(U)$ over a uniformly random permutation $\sigma$ of the
coordinates: it randomizes which coordinates fill each block, and every component keeps the profile and length of $U$. The
converses use the class $\mathcal A_r$ of causal randomized adaptive policies: a seed with arbitrary law, independent of all
oracle randomness; in round $t$ a set of unrevealed coordinates chosen measurably with respect to
$\mathcal F_{t-1}=\sigma(\text{seed},\text{committed labelled history})$; conditionally on $\mathcal F_{t-1}$, fresh
independent Bernoulli draws at the exact conditional marginals; outputs are exactly the committed draws, so the seed cannot
write values; and all $n$ coordinates committed within $r$ rounds almost surely, with an arbitrary completion at infeasible
states (on the slice, illegal count states). A predrawn seed is not a predrawn schedule: the seed may be consulted again in
every round, together with the values already drawn. A policy is deterministic if its choices are measurable in the committed
history alone; count-based rules are derived by the converse, not assumed.

\paragraph{Completion and charging.} A schedule always runs all its blocks, with fair bits after an infeasible prefix (on the
slice, an illegal count state); such a prefix never re-enters $V$, so this changes no divergence below, and $|U|$ is the cost of
every run, failed ones included. $\mathcal A_\infty$ drops the cap; $\tau$ counts executed rounds until all coordinates are
committed (empty rounds, rounds after a failure, and closing rounds whose values the quota already fixes all count; $\tau=\infty$,
output $\perp\notin V$, if never), under the joint law of seed, rounds and output (output law $Q_\pi$); a mixture is in
$\mathcal A_\infty$ with $\tau=|U|$.

\paragraph{The two criteria and three costs.} All divergences are forward KL in nats, $+\infty$ when $q$ gives zero
mass to a point that $p$ charges. For a mixture $\mu$ and $\varepsilon\ge0$,
\begin{gather*}
C(\mu)=\E_\mu|U|,\qquad H(\mu)=\E_\mu\KL(p\|q_U),\qquad F(\mu)=\KL(p\|Q_\mu),\\
\LH(n,\varepsilon)=\inf\{C(\mu):H(\mu)\le\varepsilon\},\qquad\LF(n,\varepsilon)=\inf\{C(\mu):F(\mu)\le\varepsilon\}.
\end{gather*}
We keep three costs apart: the \emph{scheduled mean} $C(\mu)$, a mixture's expected scheduled rounds, failures included; the
\emph{executed mean} $\E\tau$, a causal policy's expected executed rounds; and the \emph{hard cap} $r$ of a policy in
$\mathcal A_r$, at most $r$ rounds almost surely. A mixture has $\E\tau=C(\mu)$, and a policy in $\mathcal A_r$ has
$\E\tau\le r$, so a lower bound on the executed mean is also one on the cap.

\paragraph{Leakage and shape.} For a target $p$ with support $V$, part (a) of the next lemma splits a divergence into
\emph{leakage}, from mass outside $V$, and \emph{shape}, the divergence of $p$ from the output's conditional law on $V$; (b)
gives the gain $H-F$ from discarding the schedule. On the slice, (c) and (d) reduce a labelled schedule to its profile, and (e) shows that relabelling
removes the shape term but not the leakage, which a fixed labelled schedule, drawing block counts binomially, generally has
together with shape. With several symmetry orbits the chain rule adds a \emph{between-orbit} term: after
relabelling, the balanced slice keeps only leakage, the full-support target only this term, and a weighted slice both
(Appendix~\ref{app:alt}).

\begin{lemma}[standard tools; Appendix~\ref{app:hlow}]\label{lem:tools}
(a) $\KL(p\|q)=-\ln q(V)+\KL(p\|q(\cdot\mid V))$ if $q(V)>0$, and $\KL(p\|q)=+\infty$ if $q(V)=0$. (b) $F(\mu)\le H(\mu)$ and,
if $F(\mu)<\infty$, $H(\mu)-F(\mu)=\E_{x\sim p}\KL(\mu\|r_\mu(\cdot\mid x))$, where $r_\mu(U\mid x)\propto\mu(U)q_U(x)$ is the
law of the schedule given the output. (c) $q_U(V)$ depends on $U$ only through its profile $b(U)$, and equals the success
probability $s(b)$ of the count process $(M,j)\mapsto(M-b_t,j-X_t)$, $X_t\sim\Bin(b_t,j/M)$, started at $(n,k)$ and killed
when it leaves the legal states. (d) $\KL(p\|q_U)$ depends only on $b(U)$. (e) On the balanced slice, if an output law $Q$ is invariant on $V$ under
relabelling the coordinates (for example $Q_\mu$ for a relabelling-invariant $\mu$), then $\KL(p\|Q)=-\ln Q(V)$, and
$Q(\cdot\mid V)=p$ if $Q(V)>0$.
\end{lemma}

\section{The balanced slice: separation and an exact converse}\label{sec:bal}\label{sec:main}

\paragraph{From output divergence to survival.} The run outputs a valid string if and only if the count process ends at
$(0,0)$. A uniform relabelling leaves this process unchanged and makes the output law permutation-invariant, so
$F=-\ln\Pr(\text{the count process reaches }(0,0))$ (Lemma~\ref{lem:tools}(e)). Write $\delta=1-e^{-\varepsilon}$, the
failure probability that tolerance $\varepsilon$ allows a relabelled design, and
$K_\delta=\min\{\lceil1/\delta\rceil,\lceil2\ln(2/\delta)\rceil\}$.

\begin{theorem}\label{thm:sep}
Let $n=2k\ge2$, $\Hc_n=\sum_{m\le n}1/m$, $S_n=\sum_{m\le n}1/m^2$.
\begin{enumerate}
\item[(a)] For every $\varepsilon\ge0$, $\LH(n,\varepsilon)\ge\max\{1,[\Hc_n-2\varepsilon]_+,1+[\Hc_n-2\varepsilon-1]_+^2/
(4\varepsilon+S_n-1)\}\ge1+[\ln(n+1)-2\varepsilon-1]_+^2/(4\varepsilon+2/3)$.
\item[(b)] For every $\varepsilon>0$, $\LF(n,\varepsilon)\le\min\{n,\lceil\log_2\log_2n\rceil+8K_\delta\}$, witnessed
by the uniform relabelling of one deterministic size profile.
\item[(c)] Hence, for fixed $\varepsilon>0$, $\LH(n,\varepsilon)/\LF(n,\varepsilon)=\Omega_\varepsilon((\log n)^2/
\log\log n)$; and since one fixed geometric schedule gives $\LH(n,\varepsilon)=O_\varepsilon((\log n)^2)$
(Appendix~\ref{app:lhup}; see also \citealp[Thm.~11]{wen2026conditional}), $\LH(n,\varepsilon)=\Theta_\varepsilon((\log n)^2)$.
\end{enumerate}
\end{theorem}

\paragraph{Why averaged divergence needs $(\log n)^2$ rounds.} Let a schedule have block fractions $\rho_t=b_t/M_{t-1}$,
where $M_{t-1}$ coordinates are unrevealed before round $t$, so $\rho_t=1$ in the last round. By the chain rule, a coordinate drawn at the
marginal of the $M_{t-1}$ coordinates unrevealed at its round's start, rather than of the $m$ unrevealed at its turn, costs
$H_1(M_{t-1})-H_1(m)$, where $H_1(m')$ is the expected binary entropy of the fraction of ones among $m'$ unrevealed coordinates;
each one-coordinate reveal at $m'$ lowers it by at least $1/(2m'(m'-1))$ (Pinsker), and these telescope to at least
$\frac12(1/m-1/M_{t-1})$. Over all coordinates the terms $1/m$ sum to $\Hc_n$ and the
terms $1/M_{t-1}$ to $\sum_t\rho_t$, so a schedule $U$ with divergence $d_U$ has $\sum_t\rho_t\ge\Hc_n-2d_U$, about $\ln n$. Within a
block, $\sum_{M_t<m\le M_{t-1}}(1/m-1/M_{t-1})\ge b_t(b_t-1)/(2M_{t-1}^2)$, so $\rho_t^2$ is at most twice this sum plus $\sum_{M_t<m\le M_{t-1}}1/m^2$, and
$\sum_t\rho_t^2\le2(\Hc_n-\sum_t\rho_t)+S_n\le4d_U+S_n$. Both bounds are affine in $d_U$, so they survive averaging over a mixture
with $H\le\varepsilon$ even when some components have $d_U>\varepsilon$; Cauchy--Schwarz over the rounds before the last,
$(\E_\mu\sum_t\rho_t-1)^2\le(C(\mu)-1)(\E_\mu\sum_t\rho_t^2-1)$, then gives the last term of (a) (Appendix~\ref{app:hlow}). The matching upper bound uses one fixed schedule with
blocks of $\max(1,\lfloor\alpha M\rfloor)$ of the $M$ unrevealed coordinates, $\alpha=\min\{\varepsilon,\ln2\}/\ln n$
(Appendix~\ref{app:lhup}).

\paragraph{The square-root ladder.} For (b), the ladder leaves $M_j=K\,2^{j+2}\,2^{2^j}$ unrevealed coordinates after its
rungs, $j$ decreasing ($K=K_\delta$), and ends with $\min\{n,8K\}$ single-coordinate rounds, which never fail; the first bulk round goes from $n$ to the largest rung below $n$. The width of
each rung is chosen to control the failure accumulated before the singleton tail: widening rung $j$ by the factor $K2^{j+2}$
makes the conditional variances of the fraction of ones still required sum to less than $1/(4K)$, so the bulk of the
ladder fails with probability at most $\delta$ (Kolmogorov's inequality for $K=\lceil1/\delta\rceil$, the Hoeffding--Ville
inequality for $K=\lceil2\ln(2/\delta)\rceil$). Relabelled, the profile has $F=-\ln(\text{success})\le\varepsilon$
(Appendix~\ref{app:hlow}).

\paragraph{The exact recursion.} The ladder is not optimal; it must be compared with every causal policy, randomized and adaptive. Let $G_r(M,j)$ be the largest probability of reaching $(0,0)$ within $r$
rounds when each block size may depend on the current count state:
\begin{gather*}
G_r(0,j)=\mathbf 1\{j=0\},\qquad G_0(M,j)=0\ \ (M>0),\\
G_{r+1}(M,j)=\max_{1\le b\le M}\ \sum_{j'=0}^{M-b}\Bin(b,j/M)(j-j')\,G_r(M-b,j').
\end{gather*}
Here $j'$ is the number of ones still required after the block. $G_r$ is optimal among rules that see only the count state and
the rounds left; the next lemma shows that no causal policy gains from seeing more: whatever it has seen, its choice changes
the law of the next count state only through the block size, which the recursion already optimizes at every state; an empty
round only spends a round, as $G_r\le G_{r+1}$ (App.~\ref{app:conv}). The lemma also converts survival into executed rounds.

\begin{lemma}[survival and charging]\label{lem:surv}
Let $n=2k\ge2$, $\pi\in\mathcal A_\infty$ and $t\ge1$. (a) The count state reaches $(0,0)$ within $t$ rounds with
probability at most $G_t(n,k)$, and $G_s\le G_t$ for $s\le t$; in particular $q_U(V)\le G_{|U|}(n,k)$ for every
labelled schedule $U$. (b) If $Q_\pi(V)\ge\theta$, then $\E\tau\ge1+\sum_{s\ge1}(\theta-G_s(n,k))_+\ge1+t(\theta-G_t(n,k))_+$.
\end{lemma}
\emph{Proof} (Appendix~\ref{app:conv}). (a) Given the seed and the history, a round's draws are independent Bernoulli$(j/M)$
variables, so $G$ of the remaining horizon at the current count state ($0$ once illegal) is a supermartingale, from $G_t(n,k)$
to the indicator of success. (b) A valid output with $\tau\le s$ needs the count state to reach $(0,0)$ within $s$ rounds,
which by (a) has probability at most $G_s(n,k)$, so $\Pr(\tau>s)\ge Q_\pi(V)-G_s(n,k)\ge\theta-G_s(n,k)$. Every run executes
at least one round, so $\E\tau=1+\sum_{s\ge1}\Pr(\tau>s)$, and summing the tails, with $G_s\le G_t$ for $s\le t$ in the second
form, gives (b) (there is nothing to prove if $\Pr(\tau=\infty)>0$). \qed

For rates, Appendix~\ref{app:env} proves the envelope \eqref{eq:env}: for
$r\ge1$ and legal $(M,j)$, $G_r(M,j)\le36(d+1)^{-2^{-r}}$ with $d=\min(j,M-j)$. Its exponent halves once per
round. For $r=1$ it is the central binomial atom, at most $(d+1)^{-1/2}$. For one more round, the recursion and the bound for
$r$ rounds leave one estimate about a single block: the mean of $(d'+1)^{-2^{-r}}$, where $d'$ is the minority count after the
block and failed paths count zero. At exponent $\frac12$ this mean is at most $6(d+1)^{-1/4}$: a block that leaves more than
$M/8$ coordinates keeps $d'>d/16$ except with probability below $256/d$ (Chebyshev), and a larger block gives each count state
probability $O(d^{-1/2})$, so the states with $d'$ below about $\sqrt d/8$ contribute $O(d^{-1/4})$, and the others have
$(d'+1)^{-1/2}=O(d^{-1/4})$. As the block's law has total mass at most one, Cauchy--Schwarz bounds the mean at exponent
$\alpha$ by the square root of the mean at $2\alpha$; iterating, the estimate at $\frac12$ gives
$6^{2^{1-r}}(d+1)^{-2^{-(r+1)}}$ at exponent $2^{-r}$, and the factors $6^{2^{1-r}}$ multiply to less than $6^2=36$.

\begin{theorem}\label{thm:conv}
Let $n=2k\ge2$ and $r\ge1$.
(a) $\min_{\pi\in\mathcal A_r}\KL(p\|Q_\pi)=-\ln G_r(n,k)$, attained by a maximizing count-state rule applied to
the coordinates in a uniformly random order.
(b) $2^{-r}\ln(k+1)-\ln36<-\ln G_r(n,k)\le2^{-r}\ln k+\ln48$, uniformly in $r$ and $n$; the upper side is witnessed
by uniformly relabelling one fixed profile with at most $r$ blocks, within a factor $1728$ of the best adaptive policy in
success probability.
\end{theorem}
\emph{Proofs} (Appendices~\ref{app:env}--\ref{app:conv-window}). (a): every $\pi\in\mathcal A_r$ has
$\KL(p\|Q_\pi)\ge-\ln Q_\pi(V)\ge-\ln G_r(n,k)$ (Lemmas~\ref{lem:tools}(a) and~\ref{lem:surv}(a)), and the maximizing rule,
applied to the coordinates in a uniformly random order drawn before sampling, succeeds with probability $G_r(n,k)$ and has a relabelling-invariant output law, so Lemma~\ref{lem:tools}(e)
gives equality. In (b), the lower side is \eqref{eq:env} with $d=k$. The upper side uses a square-root ladder for a different task than Theorem~\ref{thm:sep}(b): with
$L^\ast\le r$ the largest $L$ whose rungs $M_0=n$, $M_t=\lceil\sqrt{2^{L-t+1}M_{t-1}}\,\rceil$ ($t<L$) give nonempty blocks, it ends with
one block of all remaining coordinates (Appendix~\ref{app:conv-window}). If $L^\ast=r$, a stopped-martingale Chebyshev bound
keeps it legal until that block with probability above $\frac12$, and the block succeeds with at least its central atom, so the
profile succeeds with probability at least $k^{-2^{-r}}/24$; if $L^\ast<r$, then $n^{2^{-L^\ast}}<2^{3/2}$ and it succeeds with
probability at least $\frac1{48}$. Either way the success is at least $k^{-2^{-r}}/48$, whence $1728=36\cdot48$. This witness estimates survival within a constant factor at every cap, but its final block fails with probability at
least $\frac12$ at a state that is not forced. The upper side of (b) certifies no tolerance below $\ln48$.

\begin{corollary}[matched hard length; Appendix~\ref{app:conv}]\label{cor:hard}
Fix $\varepsilon>0$. The least length of a single fixed schedule with $\KL(p\|q_U)\le\varepsilon$ is
$\Theta_\varepsilon((\log n)^2)$, while the least length $R_F^{\mathrm{PROF}}(n,\varepsilon)$ of a uniformly relabelled fixed
profile with output KL at most $\varepsilon$ is $\log_2\log_2n+O_\varepsilon(1)=\log_2\ln n+O_\varepsilon(1)$.
\end{corollary}
The relabelled lower bound is that of Corollary~\ref{cor:t2}(a), since a profile of length $L$ lies in $\mathcal A_L$.

\begin{corollary}[hard cap and executed rounds; Appendix~\ref{app:conv}]\label{cor:t2}
Let $n=2k\ge2$ and fix $\varepsilon>0$. (a) The least hard cap $R_F^{\mathrm{adapt}}(n,\varepsilon)$, the least $r$ with
$\min_{\pi\in\mathcal A_r}\KL(p\|Q_\pi)\le\varepsilon$, is the least $r$ with $G_r(n,k)\ge e^{-\varepsilon}$, and
$\log_2\ln(k+1)-\log_2(\varepsilon+\ln36)<R_F^{\mathrm{adapt}}(n,\varepsilon)\le R_F^{\mathrm{PROF}}(n,\varepsilon)\le
\lceil\log_2\log_2n\rceil+8K_\delta$, both $\log_2\ln n+O_\varepsilon(1)$.
(b) Every $\pi\in\mathcal A_\infty$ with $\KL(p\|Q_\pi)\le\varepsilon$ has $\E\tau\ge1+\sum_{t\ge1}(e^{-\varepsilon}-G_t(n,k))_+\ge
e^{-\varepsilon}\log_2\ln(k+1)-2.672\ge e^{-\varepsilon}\log_2\ln n-3.672$, as $k+1\ge\sqrt n$; the same bound holds for
$\LF(n,\varepsilon)$.
(c) $\LF(n,\varepsilon)\le(e^{-\varepsilon}+o(1))\log_2\ln n$, so $\LF(n,\varepsilon)=(e^{-\varepsilon}+o(1))\log_2\ln n$
and, with Theorem~\ref{thm:sep}, $\LH/\LF=\Theta_\varepsilon((\log n)^2/\log\log n)$.
\end{corollary}

Part (b) does not follow from the cap bound (a), since a policy without a cap may end some runs early, as the mixture in (c)
does; it sums the tails of Lemma~\ref{lem:surv}(b) with $\theta=e^{-\varepsilon}$, since $\KL(p\|Q_\pi)\le\varepsilon$ gives
$Q_\pi(V)\ge e^{-\varepsilon}$ by Lemma~\ref{lem:tools}(a). With $\ell_k=\log_2\ln(k+1)$, \eqref{eq:env} gives
$G_t(n,k)\le\min\{1,36\exp(-2^{\ell_k-t})\}$. Summing $\Pr(\tau>t)\ge e^{-\varepsilon}-G_t(n,k)$ over $t\le\lfloor\ell_k\rfloor$,
where these $G_t$ total less than $2.672$, gives (b). In (c), a mixture runs, with weight $w=e^{-\varepsilon}/(1-\delta_n)$,
a ladder failing with probability at most $\delta_n\to0$ slowly (App.~\ref{app:conv}), and otherwise a one-round schedule that usually fails;
since $Q\ge wQ'$ pointwise gives $\KL(p\|Q)\le\KL(p\|Q')+\ln(1/w)$, it has $F\le-\ln(1-\delta_n)+\ln(1/w)=\varepsilon$ at about
$e^{-\varepsilon}$ times the ladder's cost. The near-exact branch carries weight tending to $e^{-\varepsilon}$, so the cheap
branch need not be the majority; every failed run is charged. 

\begin{figure}[t]
\centering\includegraphics[width=\textwidth]{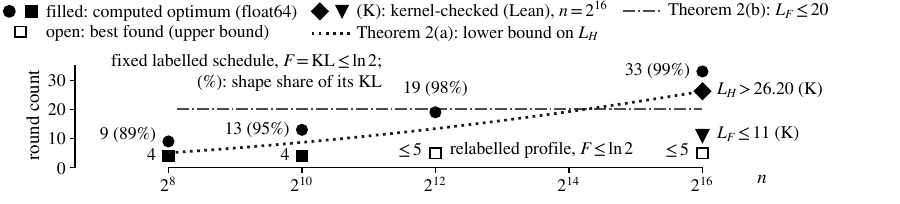}
\caption{Balanced slice, $\varepsilon=\ln2$: round counts of one fixed labelled schedule (circles) and one
relabelled profile (squares). Each has one length per run, so circles and squares bound $\LH$ and $\LF$ from above;
open squares (best-found witnesses) and the
kernel-checked $\LF\le11$ at $n=2^{16}$ are upper bounds not shown optimal.
No rate is fitted.}\label{fig:finite}
\end{figure}

\paragraph{At finite sizes.}\label{sec:finite} At $\varepsilon=\ln2$ and $n=65536$, Theorem~\ref{thm:sep}(a) gives the
kernel-checked bound $\LH>26.20$ (proved in Lean~4, Appendix~\ref{app:lean}), and one explicit $11$-round relabelled profile is
kernel-checked to have $F\le\ln2$, so $\LF\le11$; a float64 search finds a $5$-round profile with $F\le\ln2$, evaluated in exact arithmetic (not kernel-checked; minimality not shown). Theorem~\ref{thm:sep}(b) and the $\Hc_n$
form of~(a) alone give $\LH>\LF$ at $\varepsilon=\ln2$ for every even $n\ge2^{15}$ (kernel-checked). Figure~\ref{fig:finite} shows
Corollary~\ref{cor:hard} at four sizes. Evidence classes, full-support calculations and ten frozen exact-oracle predictions,
five of which failed without contradicting a theorem, are in Appendix~\ref{app:exhibits}.

\section{Beyond the slice: two extensions}\label{sec:beyond}

On the balanced slice a uniform relabelling does two jobs: it removes all shape error, and it leaves one event, survival,
whose probability alone determines the output divergence ($F=-\ln\Pr(\text{survival})$). Each extension removes the premise of one job. The full-support
Beta$(1,1)$--Bernoulli target is still exchangeable, so relabelling still removes the shape within each count class, but no
output is invalid: all the error sits in the law of the number of ones, the between-orbit term of Section~\ref{sec:setup}.
Weighted slices keep the hard constraint, so survival still bounds $F$ from below, but only permutations among equal weights
preserve the target, so survival no longer controls the shape. The upper-bound constructions behind Theorems~\ref{thm:fs}(c) and~\ref{thm:tilt}(a) are
analysed against a law used only in the analysis (an ideal count law in \S\ref{sec:fs}, a reference law in \S\ref{sec:tilt}). Neither extension
covers a full-support target without permutation symmetry.

\subsection{Full support: lattice geometry replaces survival}\label{sec:fs}

Let $p$ be the Beta$(1,1)$--Bernoulli law on $\{0,1\}^n$: the number of ones $W$ is uniform on $\{0,\dots,n\}$, the string
is uniform given $W$, and after $s$ reveals with $a$ ones each unrevealed coordinate is one with probability $(a+1)/(s+2)$.
Every string has positive target mass, so no output is a support violation; this does not make sampling accurate, as
independent draws in a round distort the count (one round of all $n$ coordinates concentrates $W$ within $O(\sqrt n)$ of $n/2$).
The target and every relabelled output law are uniform given $W$, so all error is in the count:
$F=\KL(p_W\|\bar Q_W)$, $\bar Q_W$ the output's law of $W$ averaged over the mixture (relabelling never increases $F$ or the
cost; Appendix~\ref{app:thmm}, X0).

\begin{theorem}[full-support frontier]\label{thm:fs}
For this target:
(a) $\LH(n,\varepsilon)=\Theta_\varepsilon((\ln n)^2)$ for every fixed $\varepsilon>0$ (Proposition~\ref{prop:hfs}; cf.\ \citealp[Thms.~4, 9]{chen2026optimal}).
(b) The least number of rounds of one uniformly relabelled profile with $F\le\varepsilon$ is
$R_F^{\mathrm{PROF}}(n,\varepsilon)=\log_2\log_2n+O_\varepsilon(1)$ for every fixed $\varepsilon>0$
(Corollary~\ref{cor:cf1}).
(c) Over finite predrawn schedule mixtures, under the scheduled mean $C(\mu)$,
$\lim_{n\to\infty}\LF(n,\varepsilon)/\log_2\log_2n=e^{-\varepsilon}/2$ for every fixed $\varepsilon>0$
(Corollary~\ref{cor:mixexact}).
(d) For fixed $R\ge1$ and every $\eta'>0$, for all large $n$ every finite predrawn mixture whose components all have at
most $R$ rounds has $F\ge n^{3/(4^R-1)-\eta'}$ (Theorem~\ref{thm:mixf}(c)); no matching upper bound on $F$
is claimed.
\end{theorem}

Parts (b) and (c) compare one grid with many. The upper bound in (c) takes four steps with separate jobs: grid geometry
decides which growth of the level is possible, an ideal law controls mass in total variation, Lemma~Z6 and an exact floor
turn this into forward KL, and a recursion fixes the depth, to which dilution adds the factor $e^{-\varepsilon}$. The mixture
lower bounds use one test set for all components.

\begin{figure}[t]
\centering\includegraphics[width=\textwidth]{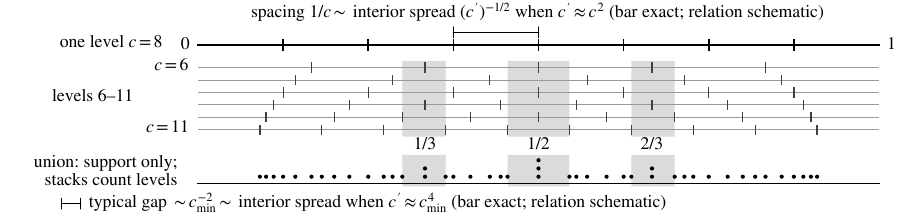}
\caption{Exact grid points and bars; relations schematic. Top: level $c=8$; middle: levels $6$--$11$; bottom: their union, with gaps of
order $c_{\min}^{-2}$ away from small denominators (shaded: larger gaps near $1/3$, $1/2$, $2/3$; not the converse's set $S^\star$). Dots: support, not probability.}\label{fig:grids}
\end{figure}

\paragraph{Geometry, one grid: squaring.} With the \emph{level} $c=2+(\text{coordinates revealed})$ and
$P=(1+\text{ones revealed})/c$, the target makes $P$ uniform on the grid $\{1/c,\dots,(c-1)/c\}$ at every level, and a
relabelled profile, with fixed levels, keeps $P$ on these grids. A round from $c$ to $c'$ moves an interior $P$ by a
spread of order $c'^{-1/2}$ when $c'\ge2c$ (smaller near the ends), and later rounds keep the final value within this order
(Lemma~X1), so neighbouring interior spreads overlap only if $c'\lesssim c^2$ (Figure~\ref{fig:grids}, top). Faster growth leaves gaps, first near the ends, where the target has mass and such a profile almost none: data processing
bounds $F$ below (Theorem~\ref{thm:thmm}). Conversely, a Jensen chain bounds $F$ by a sum of per-round terms (Lemma~U1: averaged
one-round log-likelihood ratios),
and Theorem~\ref{thm:cf1}'s profile ($\varepsilon\le1$; unit rounds throughout for $n$ below a size depending only on $\varepsilon$)
squares the level after a warm-up, up to
factors depending only on $\varepsilon$ that keep these terms summing to at most $\varepsilon$, in $\log_2\log_2n+O_\varepsilon(1)$
rounds (its exact schedule and final round: \S\ref{app:thmm-one-upper}).

\paragraph{Geometry, many grids: fourth powers.} A mixture can randomize the level. Theorem~\ref{thm:mixu}'s construction $\mu_n$
predraws a recursive schedule, relabelled uniformly: a call of size $n$ above a cutoff draws $d$ uniformly from the band
$\{c_{\min},\dots,2c_{\min}-1\}$, $c_{\min}$ a polylogarithmic factor above $(n+2)^{1/4}$, and schedules the first $d-2$
coordinates by the same rule and the rest in one final round, except on a rare floor branch (below). Grids of levels $d$ and $e$ share only the grid of
$\gcd(d,e)$, and band levels rarely share large factors, so away from small denominators and the ends their union has typical
spacing $c_{\min}^{-2}$, which the final round's spread can fill while $n+2\lesssim c_{\min}^4$ (Figure~\ref{fig:grids}). So one
round can raise the level from the band to $n+2$, a polylogarithmic factor below $c_{\min}^4$, where one grid allows about squaring. This
explains coverage, not mass: it shows which growth is possible, not how much mass the sampler puts anywhere; the next
two steps supply the mass and the forward-KL control.

\paragraph{Mass: an ideal comparison.} Mass is first controlled for an \emph{ideal} run, analysis only and never
executed: it has the same band draw and final round as $\mu_n$ but an exactly uniform prefix count. Its final count law
$\mathsf R_n$ is compared with the target's, $\mathsf U_n=\mathrm{Unif}\{0,\dots,n\}$, in total variation, which measures
the target mass that $\mathsf R_n$ misses. As a union of grids is discrete, the analysis (only) smooths the prefix value at
a width $\upsilon$. With each prefix grid completed to all $d$ points $j/d$ of the circle $[0,1)$, equally weighted (at
total-variation cost $O(1/c_{\min})$), two smoothed grids overlap like one grid of denominator $\mathrm{lcm}(d,e)$
(Lemma~Z3), and over the
band the reciprocals of these denominators sum to $O(\ln c_{\min})$, so the smoothed band average is within total
variation $O(\sqrt{\ln c_{\min}/(\upsilon c_{\min}^2)})$ of uniform on $[0,1]$ (Lemma~Z4; cf.\ randomized lattice rules,
\citealp{kritzer2019lattice}), a bound that is small only if $\upsilon\gg c_{\min}^{-2}\ln c_{\min}$. The width must also
lie a polylogarithmic factor below the final spread, so as barely to move the final count when the prefix value is away from the ends,
whose mass is charged separately (Lemma~Z5). Both hold because
$c_{\min}$ exceeds $(n+2)^{1/4}$ by a polylogarithmic factor: this is where the construction needs the fourth power. As a
binomial count with a continuous Uniform$[0,1]$ parameter is exactly uniform, $\mathsf R_n$ is within total variation
$\Delta_n=O((\ln n)^{-4})$ of $\mathsf U_n$, the leading terms being the smoothing's cost in the final count and the
values near the ends. This says nothing yet about the actual run (recursive prefix; final-count law $\mathsf Q_n$), and total variation alone is not a
forward-KL guarantee, as $\KL(\mathsf U_n\|\mathsf Q_n)$ is large wherever $\mathsf Q_n$ is tiny.

\paragraph{Forward KL: common mass and the floor.} Against this, with probability $m^{-2}$ a call of size $m$ above the
cutoff is the \emph{floor}, a predrawn branch executed when selected, not a repair, which keeps $\mathsf Q_m\ge m^{-2}\mathsf U_m$.
\emph{Lemma Z6} (common-mass lift) then compares records of ideal and actual runs (prefix size $s=d-2$, prefix count $a$,
final count $v$). On the mass $g:=\min(\mathsf U_n,\mathsf R_n)$ that the ideal law shares with the target, the lifted
target follows ideal records, whose band law $\mathsf B$ and final round $\mathsf K_s$ are shared with actual runs and
cancel, so the children's divergences $\KL(\mathsf U_s\|\mathsf Q_s)$ pass with coefficient one. The floor pays for the
target mass that the ideal law misses, $\Delta_n$, which goes on floor records at a cost of at most $2\ln n$ per unit;
dropping the ideal law's excess mass, also $\Delta_n$, costs at most $\ln(n+1)$ per unit, as no child log-ratio
$\ln(\mathsf U_s/\mathsf Q_s)$ is below $-\ln(n+1)$:
\[
\mathcal D_n\le\textstyle\sum_s\mathsf B(s)\,\mathcal D_s+\Delta_n[\ln(n+1)+2\ln n]-\ln(1-n^{-2}),\qquad\mathcal D_m:=\KL(\mathsf U_m\|\mathsf Q_m)=F(\mu_m).
\]
In the lift, each band record carries target weight $g(v)\mathsf B(s)\mathsf U_s(a)\mathsf K_s(a,v)/\mathsf R_n(v)$ against
the actual $(1-n^{-2})\mathsf B(s)\mathsf Q_s(a)\mathsf K_s(a,v)$, and floor records carry $\mathsf U_n-g$. Coefficient one
is what makes the construction work in forward KL: errors add along the recursion instead of compounding.

\paragraph{Cost: depth and dilution.} At or below the cutoff, which is fixed from the tolerance before $n$ varies, calls
commit single coordinates (exact sampling), so $\mathcal D_m=0$ there; above it, the display and the bound on $\Delta_n$
give $\mathcal D_n\le\E_d\mathcal D_{d-2}+O((\ln n)^{-3})$ ($\E_d$ the band average). The scheduled mean over all paths,
every floor charged in full, obeys $C(\mu_n)\le1+\E_dC(\mu_{d-2})+1/n$. Each call above the cutoff that takes the band branch
cuts $\ln n$ to about a quarter and adds one round, so along each band-only path the error terms, largest near the cutoff, sum below the tolerance, and the depth is about
$\log_4\ln n$: $C(\mu_n)\le\frac12\log_2\log_2(n+2)+O_\varepsilon(1)$ (Theorem~\ref{thm:mixu}). Dilution supplies the factor
$e^{-\varepsilon}$ separately: the construction, at a fixed tolerance $\varepsilon'\le\min(\varepsilon,1)$, gets weight
$w=e^{\varepsilon'-\varepsilon}$ and the one-round schedule the rest, so $F\le\varepsilon'+\ln(1/w)=\varepsilon$ by pointwise
domination, and $\varepsilon'\downarrow0$ after $n\to\infty$ (Corollary~\ref{cor:mixexact}). Both are mean costs: the rare
exact floors run when selected, and this construction's top floor is an $n$-round component, so it gives no hard cap
below $n$.

\paragraph{One obstruction for every fast component.} One test set must catch every fast component of every mixture, so small $F$ forces mass onto long components;
per-component gap sets fail, as components fill one another's gaps. A \emph{jump} takes the level from $c$ to at least $\xi c^\vartheta$ ($\vartheta>4$, $\xi\ge1$); every profile with at most $R_0(n)=\log_\vartheta\ln(n+2)+O(1)$ rounds has one (Lemma~X2). Just before a jump, $P$ is a reduced fraction $\ell/q$ with $q\le c$, and except with probability $2e^{-\omega^2}$ the run
ends within $\omega(\xi q^\vartheta)^{-1/2}$ of it (Lemma~X1), inside one fixed set $S^\star$ of windows around all reduced
fractions with $\xi q^\vartheta\le n+2$, of small target mass for large $\xi$ because $\sum_{q\ge2}\phi(q)q^{-\vartheta/2}<\infty$
($\phi$ the totient) exactly when $\vartheta>4$ (Lemma~X3). Data processing onto its complement forces mass $\eta_n$ onto
components longer than $R_0(n)$ rounds, so $C(\mu)\ge1+R_0(n)\eta_n$; taking $n\to\infty$, $\xi\to\infty$, $\omega\to\infty$,
$\vartheta\downarrow4$, in this order, sends $\eta_n$ to $e^{-\varepsilon}$, giving $e^{-\varepsilon}/2$ (Theorem~\ref{thm:mixf}; Remark~\ref{rem:mixf} for $1<\vartheta\le4$), while a cap $R$ fixed as $n$ grows lets $\ln\xi$ grow like $\ln n/(1+\vartheta+\dots+\vartheta^{R-1})$, giving (d) as $\vartheta\downarrow4$. Split at the first jump (decided before its draws), with the spread bound for adaptive batches given the committed
history, the argument covers causal policies, for which the limit of (c) holds under $\E\tau$
(Theorem~\ref{thm:fsadapt}).

The $\frac12$ comes from the grids (fourth powers; $\vartheta>4$), the $e^{-\varepsilon}$ from dilution
(Corollary~\ref{cor:mixexact}), and the least hard cap, for mixtures and causal policies, between
$(\frac12-o(1))\log_2\log_2n$ (\S\S\ref{app:thmm-mix-lower}, \ref{app:thmm-adapt}) and
$\log_2\log_2n+O_\varepsilon(1)$, is open. Theorem~\ref{thm:mixu}'s cutoff exceeds $e^{1024}$. The prior
is Beta$(1,1)$ throughout.

\makeatletter\@ifundefined{mainLemOneStep}{\newtheorem*{mainLemOneStep}{Lemma OS (one step)}}{}\makeatother
\subsection{Bounded weights: random retention replaces relabelling}\label{sec:tilt}

Let $p_w(x)\propto\prod_iw_i^{x_i}$ on the slice $S_k$ of strings with $k$ ones, with $w\in(0,\infty)^n$, $\max w/\min w\le\Lambda$ and
$\kappa_0n\le k\le(1-\kappa_0)n$; two weight values give the two-weight conditional Poisson (rejective) design of \citet{hajek1964asymptotic} (the \emph{two groups} of Appendices~\ref{app:tgenv} and~\ref{app:tgred}). The $F$ lower bounds still come from survival, by a bounded-weight version of \eqref{eq:env} (Appendix~\ref{app:t1}, \S10). Distinct weights cost only constants there: with weight ratio at most $\Lambda$ all conditional marginals stay within a factor $\Lambda$ of one another, so a one-round moment bound on the minority quota $d$ (the smaller of the numbers of ones and zeros still to be placed) holds uniformly over the residual weights, and H\"older's inequality halves its exponent at each round. So, from every legal state and for every $r\ge1$, every causal randomized policy with fresh exact-marginal draws reaches a legal completion within $r$ rounds with probability at most $A_r(d+1)^{-2^{-r}}$, $A_r\le288\Lambda^3$. The $\LH^w$ bounds come from a weighted entropy decrement and one fixed schedule (App.~\ref{app:t1}, \S\S2--4; upper order also \citealp[Thm.~9]{chen2026optimal}). But relabelling within equal-weight groups can leave the law of the group counts
wrong (Appendix~\ref{app:alt}), and distinct weights leave nothing to relabel. A run that lands on the slice can then still have the wrong within-slice law, $F=\KL(p_w\|Q(\cdot\mid S_k))-\ln Q(S_k)$ (App.~\ref{app:rf} (4.33)), so survival no longer determines $F$ and upper bounds must control the output law itself.

\begin{theorem}[bounded weights]\label{thm:tilt}
Fix $\varepsilon>0$, $\Lambda\ge1$ and $\kappa_0\in(0,\frac12]$, and let $\LH^w,\LF^w$ and $R_F^{\mathrm{adapt}}$ be the
costs of Section~\ref{sec:setup} and Corollary~\ref{cor:t2} for this target over their respective classes, with each
unrevealed coordinate's exact conditional marginal under $p_w$ in place of $j/M$.
(a) There are $c_H,C_H>0$ and $n_0$, depending only on $(\varepsilon,\Lambda,\kappa_0)$, such that for $n\ge n_0$,
uniformly in $w$ and $k$, $c_H(\ln n)^2\le\LH^w(n,\varepsilon)\le C_H(\ln n)^2$ and
$R_F^{\mathrm{adapt}}(n,\varepsilon)=\log_2\ln n+O(1)$, and both $\LF^w(n,\varepsilon)$ and the least expected executed
rounds of a causal policy with output KL at most $\varepsilon$ are $(e^{-\varepsilon}+o(1))\log_2\ln n$; the $O(1)$ and
$o(1)$ depend only on $(\varepsilon,\Lambda,\kappa_0)$.
(b) If at least $\rho n$ coordinates share one weight, for a fixed $\rho>0$ (for two weight values, the larger group gives
$\rho=\frac12$), then for large $n$ one fixed labelled schedule, relabelled uniformly within that group, has $F\le\varepsilon$ and
the same length $\log_2\ln n+O(1)$ on every path, with constants depending also on $\rho$.
\end{theorem}

\begin{figure}[b]
\centering\includegraphics[width=\textwidth]{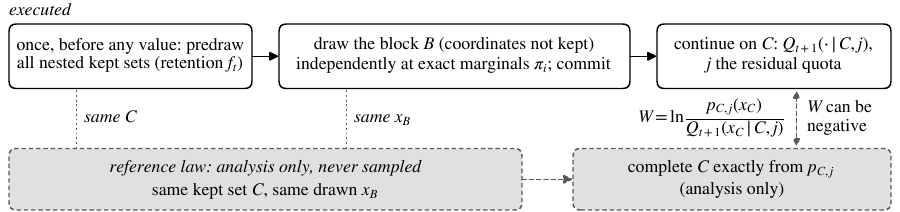}
\caption{One level of the bounded-weight sampler (top, executed) and of the reference law (bottom, analysis only). Both share
the kept set $C$ and the block values $x_B$; $W$ compares the two completions of $C$ ($W=0$ if $j$ is infeasible).}\label{fig:sampler}
\end{figure}

\paragraph{The sampler.} Part~(a)'s $F$ upper bounds (Theorem~A, App.~\ref{app:rf}) use a sampler that
randomizes which coordinates wait (Figure~\ref{fig:sampler}). For a parameter $A\ge1$, nominal sizes
$n=M_1>\dots>M_{R+1}$ depend on $(n,A)$ alone (below). Before any value, nested \emph{kept sets} $[n]=\mathcal C_1\supseteq\dots\supseteq\mathcal C_{R+1}$ ($F_t$ in App.~\ref{app:rf}) are drawn, each coordinate of $\mathcal C_t$ staying in $\mathcal C_{t+1}$ independently with
probability $f_t=M_{t+1}/M_t\le\nobreak\frac14$. Round $t$ commits $B=\mathcal C_t\setminus\mathcal C_{t+1}$, if nonempty, by fresh bits from
$\Pi_B=\nobreak\prod_{i\in B}\mathrm{Ber}(\pi_i)$ at the exact conditional marginals $\pi_i$ frozen at its start; $\mathcal C_{R+1}$ is then committed as singletons if $|\mathcal C_{R+1}|\le2m_0=2\lceil4A^2\rceil$, else as one block, and
for $n<4m_0$ the sampler is the exact $n$ singletons. This is a finite predrawn mixture, every block charged. For fixed $n$ and $A$ the schedule law is the same for every weight array and quota allowed by Theorem~\ref{thm:tilt}; only the probabilities of the scheduled draws, the oracle's exact marginals, depend on the target. (The choices of $A$ below, from $\varepsilon$ for the hard cap and growing with $n$ for the mean, use existence-only constants depending on $(\Lambda,\kappa_0)$; part~(b)'s schedule depends on the target's equal-weight group.) Below, $A\ge A_0(\Lambda,\kappa_0)$. As $N=|\mathcal C_t|$ is random around $M_t$, the analysis uses \emph{good
states}: $N/M_t\in[1-\delta_t,1+\delta_t]$ with $\delta_t\le\frac14$, and quota density in $[\kappa^{(t)},1-\kappa^{(t)}]$ with
$\kappa^{(t)}\ge\kappa_0/2$ (App.~\ref{app:rf} (4.9)--(4.11)). There $f_t\sqrt N$ is within constant factors of the nominal $L_t:=f_t\sqrt{M_t}$, which sets the scale of one round's error;
exits from good states are rare (4.14), and a good last-level state has $N<2m_0$, so its singletons are exact.

\paragraph{The reference law and the cancellation.} One round's error is isolated by comparing the executed sampler with a second
law, used only in the analysis: the \emph{reference law} $\mathcal R$ makes the same round and then finishes exactly. At a
good state, with $p$ the target's conditional law of the $N$ free coordinates and $J$ their quota, $\mathcal R$ shares $C=\mathcal C_{t+1}$ and
$x_B\sim\Pi_B$ and draws $x_C$ from $p_{C,j}=p(\cdot\mid x_B)$, the slice law on $C$ with quota $j=J-|x_B|$ (fair bits if $j$ is
infeasible); $\mathcal R^{\rm joint}$ and $\mathcal R$ are its laws of $(C,x)$ and $x$, and on the slice $\mathcal R(x)/p(x)=\E_C[\Pi_B(x_B)/p(x_B)]$. Averaging over $C$ removes, to first order, the quota-induced dependence that $\Pi_B$ ignores. To see what is averaged, write $p$, a conditional-Poisson law, as an independent logistic product law $\widetilde P$ conditioned on its total; its marginals $q_i$ are canonical logistic probabilities, a proof device with $|q_i-\pi_i|=O(1/N)$ (App.~\ref{app:rf} (2.7)). Then $p(x_B)=\widetilde P(x_B)\,\widetilde P(|x_C|=j)/\widetilde P(|x|=J)$ (App.~\ref{app:rf}, \S3.1) carries the probability that $C$ meets its residual quota. For nonempty $C$ its reciprocal is an exact integral over logistic tilts (the empty child is a separate term of (3.1)). If $\widetilde P_u$ tilts $\widetilde P$ on $C$ by $e^{u|x_C|}$, every $x_C$ with $|x_C|=j$ has the same likelihood ratio $\widetilde P_u(x_C)/\widetilde P(x_C)=\widetilde P_u(|x_C|=j)/\widetilde P(|x_C|=j)$, so
\[
\frac1{\widetilde P(|x_C|=j)}=\frac1{I_C(j)}\int_{\mathbb R}\frac{\widetilde P_u(x_C)}{\widetilde P(x_C)}\,\mathrm{Var}_{\widetilde P_u}(|x_C|)\,du,
\]
where $I_C(j)=\int_{\mathbb R}\widetilde P_u(|x_C|=j)\,\mathrm{Var}_{\widetilde P_u}(|x_C|)\,du$, and $I_C(j)=1+O(\sqrt{\log|C|}/|C|)$ at interior children (App.~\ref{app:rf} (2.9), (2.17), (3.1)). Coordinates are retained independently, so averaging over $C$ turns the integrand into a product of per-coordinate factors, whose logarithm is linear in the bits, times the variance factor $\Gamma_x(u)$ ((3.1)--(3.2)). Dropping the empty child, far tilts and the children in the exceptional event of (3.16), and lower-bounding $1/I_C$ on the rest, gives, at every slice point,
\[
\frac{\mathcal R(x)}{p(x)}\ \ge\ (1-\epsilon_1)\,N_0^b\int\rho^b(u)\,e^{D_x(u)}\,\frac{\Gamma_x(u)}{\bar\Gamma(u)}\,du,\qquad
D_x(u)=\sum_i\alpha_i(u)(x_i-\pi_i)
\]
(App.~\ref{app:rf} (3.17)): $\rho^b$ is an $x$-free probability density concentrated on $|u|=O(1/L_t)$, $\Gamma_x/\bar\Gamma\in[\frac12,\frac32]$ has $p$-mean one and $p$-variance $O(1/N)$, and $1-\epsilon_1,N_0^b\ge1-O(1/L_t)$ (3.13). Here $\alpha_i(u)=f_tu+r_i(u)$ with $r_i=O(1/N)+O(f_tu^2)$ for small $u$. On the slice $\sum_i(x_i-\pi_i)=|x|-J=0$ (as is $\sum_i(x_i-q_i)$), so the common term $f_tu\sum_i(x_i-\pi_i)$, otherwise of order one, cancels there: $D_x(u)=\sum_ir_i(u)(x_i-\pi_i)$. What remains is centered: $D_X(u)$ has mean zero under $p$ for every $u$, because the $\pi_i$ are the exact (not logistic) marginals. After the cancellation its coefficient vector is small ((3.20)), so it concentrates at scale $1/L_t$. Cancellation and centering are not yet a KL bound. They give three distinct one-step estimates, each with its own job below: (i) logarithmic Jensen applied to the display bounds the error of one round followed by an exact completion ((3.18)); (ii) a second moment of $\psi:=p/\mathcal R$ under $\mathcal R$ converts probabilities small under $\mathcal R$ into bounds on $\psi$-weighted terms; (iii) two-sided control of $\psi$ holds on a set $T'$ of strings $x$ that both laws almost exhaust. (ii) and (iii) let the sampler's recursive, only approximately exact continuation replace the exact completion.
\begin{mainLemOneStep}
There are $c_1,c>0$, depending only on $(\Lambda,\kappa_0)$, such that at every good state of a level $1\le t\le R$:
(i) $\KL(p\|\mathcal R)\le c_1/L_t$; (ii) $\E_{\mathcal R}[\psi^2]\le c_1$; (iii) $|\psi-1|\le\eta_t=O(\sqrt{\log M_t}/L_t)$ on a set $T'$ whose complement has $p$- and $\mathcal R$-mass $O(M_t^{-10}+e^{-cL_t^2})$ (on $T'$, finitely many linear statistics of $x$ are small and children that are not interior carry little execution mass); all uniformly in $w$ (App.~\ref{app:rf}: analytic estimates (3.18), (3.21), (3.34)--(3.35), instantiated level-wise by (4.3), (4.12), (4.16) and (4.22)--(4.24)).
\end{mainLemOneStep}

\paragraph{Composition along the levels.} The recursion replaces the exact completion by the sampler's own continuation, level by level. Let $Q_t$ be the sampler's law from level $t$ on and $\mathcal K_t$ the
largest $\KL(p\|Q_t)$ over good states at level $t$, with $\mathcal K_{R+1}=0$ because the last level's singletons are exact. As $Q_t$ and $\mathcal R$ make the same first
round, $Q_t/\mathcal R$ is a posterior average of $e^{-W}$, where $W=\ln[p_{C,j}(x_C)/Q_{t+1}(x_C\mid C,j)]$ compares the exact completion of the kept set with the sampler's continuation ($W=0$ at infeasible children), and Jensen gives $\KL(p\|Q_t)\le\KL(p\|\mathcal R)+\E_{\mathcal R^{\rm joint}}[\psi W]$ (App.~\ref{app:rf} (4.15)). Given $(C,x_B)$ at a good child, $W$ has mean equal to that child's divergence, at most $\mathcal K_{t+1}$. But $W$ can be negative, where the continuation puts more mass than the exact completion, and $\psi$ varies with $x_C$: a one-sided bound $\psi\le1+\eta_t$ controls $\psi W$ only through $W^+$, whose mean can exceed $\mathcal K_{t+1}$ by $\sqrt{\mathcal K_{t+1}/2}$ (Pinsker). This is the job of Lemma~OS(iii): two-sided control on $T'$ gives $\psi W\le(1+\eta_t)W+2\eta_tW^-$ there. Off $T'$ both exceptional masses are needed, the reference's and the target's, because the composition integrates under $\mathcal R^{\rm joint}$ while $\psi$ reweights to $p$: discarding $T'^c$ raises the mean of $W$ only through $W^-$, by at most $\sqrt{\mathcal R(T'^c)}$, and the discarded part of $\psi W$ costs at most $p(T'^c)N\ln(\Lambda N)$ (the atom bound on $W^+$). Lemma~OS(ii) prices, by Cauchy--Schwarz, the children that are not good, whose $\mathcal R^{\rm joint}$-probability is small (4.14). With Lemma~OS(i) for the round itself,
\[
\mathcal K_t\le(1+\eta_t)\mathcal K_{t+1}+2\eta_t\sqrt{\mathcal K_{t+1}/2}+c_1/L_t+\xi_t,
\]
with tails $\xi_t$ made of these masses (App.~\ref{app:rf} (4.17)--(4.21)). This is not a sum of per-level KLs: the square-root term couples adjacent levels. Unrolling from $\mathcal K_{R+1}=0$, with $\prod_t(1+\eta_t)\le2$, gives $\mathcal K_*:=\max_t\mathcal K_t\le2E+2\sqrt2\,(\sum_t\eta_t)\sqrt{\mathcal K_*}$ for the summed per-level errors $E=\sum_t(c_1/L_t+\xi_t)$, hence $\mathcal K_*\le4E+8(\sum_t\eta_t)^2$ (App.~\ref{app:rf} (4.29)); the start state is good.

\paragraph{Depth and outcome.} The sizes make these sums small and the kept-set ladder short: with $m_0=\lceil4A^2\rceil$ and $m_{j+1}=\lceil m_j^2/(A^22^{3j/2})\rceil$, the $R=\max\{r\ge1:4m_{r-1}\le n\}$ levels with a kept set have $M_1=n$ and $M_t=m_{R+1-t}$ for $t\ge2$, when $n\ge4m_0$ (App.~\ref{app:rf} (4.1), (4.4)). Then
$L_t\ge A2^{3(R-t)/4}/(2\sqrt2)$ (4.5) grows geometrically in $R-t$, faster than $\sqrt{\log M_t}$, so
$\sum_tL_t^{-1}=O(1/A)$, $\sum_t\eta_t=O(\sqrt{\log A}/A)$ and $\sum_t\xi_t=O(A^{-2})$ uniformly in $n$ (4.28), and $4E+8(\sum_t\eta_t)^2\le C^*/A$: Theorem~A's $F\le C^*/A$ for the mixture's output law, with an existence-only $C^*=C^*(\Lambda,\kappa_0)$ (4.29)--(4.30). The sizes grow doubly exponentially (their log-sizes exponentially, (4.2)), so $R\le\log_2\ln n+3$ (4.6). Every realization of the schedule, including a run that fails, executes and is charged all its blocks: for $n\ge4m_0$ at most $R$ kept-set rounds and $2m_0$ terminal rounds, so at most $\log_2\ln n+8A^2+5$ rounds (4.8), and for $n<4m_0$ its $n<16A^2+4$ exact singletons. Thus $A=\max(A_0,C^*/\varepsilon)$ gives the hard cap of part~(a): $F\le\varepsilon$ with at most
$\log_2\ln n+O_{\Lambda,\kappa_0}(1+\varepsilon^{-2})$ rounds per component, small $n$ included (Corollary~A$'$). Because these bounds hold for every $A\ge A_0$, uniformly in $n$, $w$ and $k$, $A$ may instead grow slowly with $n$, independently of $\varepsilon$; diluting with the cheap one-round schedule at weight $1-e^{-(\varepsilon-\varepsilon_n)}$, $\varepsilon_n=C^*/A\to0$, as in Corollary~\ref{cor:t2}(c), then gives the mean bounds of part~(a)
for the scheduled mean and expected executed rounds, with cap
$\log_2\ln n+O(\sqrt{\log\log n})$ (Corollary~A$''$); the dilution lowers only the mean, not the cap. These are caps on components of different lengths, not one common length; part~(b) supplies a common length when a linear-size equal-weight group exists.

\paragraph{A common length.} Part~(b) leaves an equal-weight group of $M\ge\rho n$ coordinates untouched, commits the others in at most $\lceil2\Lambda^3/(\varepsilon\rho^2)\rceil$ fixed blocks, and then runs the ladder of Theorem~\ref{thm:sep}(b) on the group, relabelled uniformly within it. Given the committed values, the output on the group is then uniform on its residual slice up to the ladder's success probability $s_J$ at the residual quota $J$ (balanced or not), so the chain rule splits $F$ exactly into the divergence of the committed values, which is not repaired later, and $\E_{p_w}[-\ln s_J]$ (App.~\ref{app:t1}, \S\S6.3 and~6.6). The first is at most $\varepsilon/4$, since beside the untouched group a block of $b$ committed coordinates costs at most $\Lambda^3b(b-1)/(2M^2)$. The second is small although $-\ln s_J$ is not small at every quota: the ladder rarely fails at typical residual quotas, $-\ln s_J\le\ln(M+1)$ at every legal one, and atypical quotas carry target probability $O(1/M)$ (\S\S6.4--6.8 and the Remark after the \S7 table).%

\section{Scope, limits and what the separation shows}\label{sec:concl}

This is theory for exact oracles in one interface, on structured targets with exact samplers outside it; relaxed interfaces (lookahead, joint draws) can beat the class at small $n$
(Appendix~\ref{app:alt}). Costs count rounds, not computation. Small average prediction error alone does not recover the exact-oracle profile rate (Theorem~IC2 in Appendix~\ref{app:imperfect}): on
the balanced slice, the head that is $\frac12$ at non-forced states and exact at forced ones has population masked-prediction excess
(mask size uniform on $\{1,\dots,n\}$, loss averaged over the masked coordinates) tending to~$0$, yet a fixed relabelled profile needs
$\Theta(\sqrt n)$ rounds for $F\le\ln2$; with the same head, a count-adaptive policy attains $F\le\ln2$ with hard cap $\lceil\log_2n\rceil+1$,
an upper bound, not the least adaptive cap or a mean cost. Beyond Appendix~\ref{app:imperfect}'s bounds no guarantee is claimed for
learned models, and no practical speedup; Appendix~\ref{app:controlled} reports a
controlled test with small trained transformers ($n\le16$), whose results are sample-level, with no population guarantee. Open are the constant of the
full-support hard cap (between $\frac12$ and $1$), a common length on weighted slices without a linear-size equal-weight group,
growing weight ratios, and sharp rates for vanishing tolerances. Schedule-averaged guarantees still upper-bound $F$ and claim no optimality for it, but the counts they
certify are not lower bounds for $F$: on the balanced slice $\LH/\LF=\Theta_\varepsilon((\log n)^2/\log\log n)$.

\bibliography{refs}

\section*{AI Disclosure}
In this work we used generative AI tools: Claude models by Anthropic (mainly Claude Opus~5 and Claude Opus~5.5) in
Claude Code, through an API and in a web-search agent; an OpenAI GPT-5.6 model through the Codex CLI; and ChatGPT~6
Pro. We used them to develop the theoretical framework and the mathematical claims, to derive and write the proofs, to
assist with the Lean~4 formalization, to design the experiments and implement the code, to interpret the results, and
to assist with the literature review and the preparation of the manuscript, figures and tables. Synthetic training
samples for the learned-conditional experiments were drawn from the specified target distributions by seeded code
written with AI assistance; no observation was generated by prompting a language model to supply data points, and no
data cleaning was performed. Translation and qualitative analysis do not apply. Human review, conducted with AI
assistance, covered the statements and assumptions of all main results, the proof-critical steps of the central
theorems, the correspondence between the formal declarations and the paper's claims, and the computations supporting
the principal numerical results, which were rerun or inspected. Additional checks included separate AI-based reviews,
Lean~4 kernel checking of the declarations listed in Appendix~\ref{app:lean}, and recomputation of selected numbers by
separate implementations, within the scope documented in the appendices (Appendix~\ref{app:exhibits}). We take
responsibility for the final content of this work, including text, claims and artifacts produced with the aid of
generative AI.
\pend{}\pend{}

\clearpage
\setcounter{tocdepth}{1}
\tableofcontents
\cleardoublepage
\appendix
\section*{Guide to the appendices}
Each lemma, theorem, corollary and proposition of the main text, and each further appendix result listed, has an ordinary proof in the place named below; the
last column names the step on which that proof turns.
Appendix~\ref{app:lean} lists the parts that are also kernel-checked; Appendix~\ref{app:exhibits} gives the evidence
class of each finite exhibit, and Appendix~\ref{app:repro} the scripts behind Figures~\ref{fig:hf}, \ref{fig:finite}, \ref{fig:grids}, \ref{fig:sampler}, \ref{fig:mech}, \ref{fig:bounds} and~\ref{fig:controlled} and Table~\ref{tab:bounds}. Table~\ref{tab:notation}, at the end of Appendix~\ref{app:reference}, indexes the notation.
{\small\renewcommand{\theHtable}{guide.\arabic{table}}\setlength{\LTpre}{\smallskipamount}\setlength{\LTpost}{\smallskipamount}
\begin{longtable}{@{}>{\raggedright\arraybackslash}p{0.30\textwidth}>{\raggedright\arraybackslash}p{0.32\textwidth}>{\raggedright\arraybackslash}p{0.31\textwidth}@{}}
\toprule
Statement & Proof & Key step \\
\midrule
\endfirsthead
\toprule
Statement & Proof & Key step \\
\midrule
\endhead
Lemma~\ref{lem:tools} & Appendix~\ref{app:hlow} & invariant laws are uniform on the single orbit $V$ \\
The survival envelope \eqref{eq:env} & Appendix~\ref{app:env}, Proposition~\ref{prop:env} (its weaker $\kappa_r$ form $\Gamma_r$, at the end of that appendix, is used in Appendices~\ref{app:conv} and~\ref{app:repro}) & one-round negative-moment bound, then Cauchy--Schwarz exponent halving \\
Lemma~\ref{lem:surv}, Theorem~\ref{thm:conv}, Corollary~\ref{cor:t2} & Appendix~\ref{app:conv} (total-variation versions in its paragraph \emph{Total variation}; the window in Appendix~\ref{app:conv-window}) & count-state supermartingale bounds success; tail sums charge rounds \\
Theorem~\ref{thm:sep}(a),(b) & Appendix~\ref{app:hlow} (sketches in Section~\ref{sec:main}) & (a) per-reveal entropy decrement; (b) square-root ladder \\
Theorem~\ref{thm:sep}(c), fixed-schedule upper bound on $\LH$ & Appendix~\ref{app:lhup} & per-block bound (Step~1) on a geometric schedule \\
Corollary~\ref{cor:hard} & Appendix~\ref{app:conv}, end, from Theorem~\ref{thm:sep}, Lemmas~\ref{lem:tools}(a),(c),(e) and~\ref{lem:surv}(a), and \eqref{eq:env} & point masses in Theorem~\ref{thm:sep}; \eqref{eq:env} caps a profile's success \\
Theorem~\ref{thm:tilt}(a), bounds on $\LH^w$ & Appendix~\ref{app:t1}, \S\S2--4 & weighted per-reveal entropy decrement (Lemma~W5) \\
Theorem~\ref{thm:tilt}(a), lower bounds on $\LF^w$, $\E\tau$ and the hard cap & Appendix~\ref{app:t1}, \S10 (two weight values: \S5 for $\LF^w$ and $\E\tau$, \S9(i) for the hard cap, from Appendix~\ref{app:tgenv}) & uniform one-step moment bound, then H\"older exponent halving \\
Theorem~\ref{thm:tilt}(a), upper bounds on $\LF^w$, $\E\tau$ and the hard cap & Appendix~\ref{app:rf}, Theorem~A and Corollaries~A$'$, A$''$ & randomly retained subsets; uniform lattice Edgeworth estimate \\
Theorem~\ref{thm:tilt}(b), the common-length schedule mixture & Appendix~\ref{app:t1}, \S\S6--7 (roadmap: opening paragraph of \S6) & equal-weight reservoir absorbs the quota; residual slices stay uniform \\
Least hard cap and sharp mean coefficient for two-value tilts & Appendix~\ref{app:t1}, \S9 & bounded-prefactor envelope meets the common length of the reservoir's schedule mixture \\
Hard-length analogue of Corollary~\ref{cor:hard} for tilts & Appendix~\ref{app:t1}, \S9 & point-mass $H$ bounds; cap bounds pin the profile length \\
Proposition~\ref{prop:perm} & Appendix~\ref{app:alt} & every fixed schedule is uniform on valid permutations \\
Two-group lower bounds and finite ratios (App.~\ref{app:alt}) & Appendices~\ref{app:tgenv}, \ref{app:tgred}, \ref{app:exhibits} & one-step bound (Lemma~TG6); count-policy hull and duality (Lemmas~TGR3, TGR4) \\
Theorem~\ref{thm:fs}(a), mean-cost $H$ rate, and the separation & Appendix~\ref{app:thmm}: Proposition~\ref{prop:hfs}, Corollary~\ref{cor:hfs} & increment bound (BB1): squared log-level steps cost KL \\
Theorem~\ref{thm:fs}(b), single relabelled profiles & Appendix~\ref{app:thmm}: Theorems~\ref{thm:thmm}, \ref{thm:cf1} and their corollaries & gap set past squaring; local limit (Lemma~U6) at squaring \\
Theorem~\ref{thm:fs}(c), lower half, and (d), mixtures & Appendix~\ref{app:thmm}: Theorem~\ref{thm:mixf} (X0 and Lemmas~X1--X3 in its proof) & a jump lands in the totient test set (Lemma~X3) \\
Theorem~\ref{thm:fs}(c), upper half & Appendix~\ref{app:thmm}: Theorem~\ref{thm:mixu} (Lemmas~Z1--Z6), Corollary~\ref{cor:mixexact} & lcm overlap of band grids (Lemmas~Z3, Z4); coefficient-one lift (Lemma~Z6) \\
Theorem~\ref{thm:fsadapt}, causal adaptive policies & Appendix~\ref{app:thmm}: Theorem~\ref{thm:fsadapt} & the first jump is fixed before its draws (Lemma~X3$'$) \\
Controlled learned-conditional test (Section~\ref{sec:concl}) & Appendix~\ref{app:controlled} & no theorem: frozen tests use float64 complete-law output quantities (KL, validity, shape, and the variance of the number of ones), and schedule-averaged $H$ where specified \\
Alternative explanations; selected related work & Appendices~\ref{app:alt}, \ref{app:related} & orbit chain rule: leakage plus a between-orbit term \\
Full results table and notation & Appendix~\ref{app:reference} & --- \\
Theorem~\ref{thm:sep}(a)'s rate from prior work & Appendix~\ref{app:hlow}, after the proof of (a) & Markov's inequality on \citet[Thm.~11]{wen2026conditional} \\
Finite balanced instance; total-variation figure & Table~\ref{tab:inst} (witness), Table~\ref{tab:finite} (comparison), Figure~\ref{fig:mech} & finite instance: Theorem~\ref{thm:sep}(a) against one explicit profile \\
Imperfect conditionals (App.~\ref{app:alt}; Prop.~IC1, Thm~IC2) & Appendix~\ref{app:imperfect} & first-exhaustion identity forces $\sqrt n$ rounds (Theorem~IC2) \\
\bottomrule
\end{longtable}\addtocounter{table}{-1}}

\section{Reference material for the main text}\label{app:reference}
This appendix collects reference material: the table of classes, criteria, costs and evidence status
(Table~\ref{tab:classes}), the one-round growth table (Table~\ref{tab:growth}) and the notation index (Table~\ref{tab:notation}).

\begin{table*}[p]
\caption{Classes, criteria and costs ($\varepsilon>0$ fixed; for tilts and bounded weights also $\Lambda,\kappa_0$; $\mathcal A_r$ is the class of
Section~\ref{sec:setup}, which for two-value tilts is $\mathrm{RAND}_r$ of Appendix~\ref{app:tgenv}). ``Mean'': scheduled mean $C(\mu)$; ``$\E\tau$'': executed mean; ``cap'':
hard cap; ``length'': rounds of one schedule or relabelled profile, the same on every path. K: kernel-checked (declarations in Appendix~\ref{app:lean}, whose table also marks the results whose class identification uses one ordinary step; for an asymptotic order, the asymptotic statement); P:
prior work (preprint). $^\ddagger$One fixed labelled schedule, relabelled uniformly within an equal-weight group of linear size; without one, a common length is open. Finite regime: on the balanced slice, explicit bounds (Table~\ref{tab:bounds}), kernel-checked bounds and finite computations at $n=2^{16}$ (Table~\ref{tab:inst}), and an exact least profile length at $n=2^4$, computed optima (float64), linear-program brackets, best-found upper bounds and one seeded Monte Carlo estimate (Table~\ref{tab:finite}, Figure~\ref{fig:finite}); on full support, explicit bounds on $\LH$ (Proposition~\ref{prop:hfs}) but astronomical constants for $F$, with computed $\LH$ optima and best-found upper bounds on $\LF$ (both float64; the latter interval-certified at $n\le2^{10}$; Table~\ref{tab:fs}); for two-value tilts, explicit but impractical constants (e.g., caps of hundreds to thousands of rounds; Appendix~\ref{app:t1}, \S6.9) and, at $n\le24$ and $r\le4$, rigorous enclosures of the optimal output KL, some kernel-checked, beside float64 values (Appendix~\ref{app:exhibits}); for bounded weights, existence-only constants in the reservoir-free $F$ upper bounds (Appendix~\ref{app:rf}).}\label{tab:classes}
\vspace{4pt}\centering\footnotesize\renewcommand{\arraystretch}{0.96}
\begin{tabular}{@{}l l l@{}}
\toprule
class, criterion, cost & order & status\\
\midrule
\multicolumn{3}{@{}l}{\emph{Balanced slice}}\\
fixed schedule, KL, length & $\Theta(\log^2 n)$ & K (Cor.~\ref{cor:hard})\\
one relabelled profile, $F$, length & $\log_2\ln n+O_\varepsilon(1)$ & K (Cor.~\ref{cor:hard})\\
mixture, $H$, mean & $\Theta(\log^2 n)$ & K (Thm~\ref{thm:sep}a,c)\\
mixture, $F$, mean & $(e^{-\varepsilon}+o(1))\log_2\ln n$ & K (Thm~\ref{thm:sep}b; Cor.~\ref{cor:t2}b,c)\\
$\mathcal A_r$, $F$, cap & $\log_2\ln n+O_\varepsilon(1)$ & K (Thm~\ref{thm:conv}a; Cor.~\ref{cor:t2}a)\\
causal adaptive, $F$, $\E\tau$ & $(e^{-\varepsilon}+o(1))\log_2\ln n$ & K (Cor.~\ref{cor:t2}b,c)\\
det.\ adaptive, KL, $\E_p$ rounds & $\Theta(\log^2n)$ & P \citep[Thm.~11]{wen2026conditional}\\
\multicolumn{3}{@{}l}{\emph{Full support, Beta$(1,1)$--Bernoulli}}\\
mixture, $H$, mean & $\Theta(\log^2n)$ & K (Thm~\ref{thm:fs}a)\\
one relabelled profile, $F$, length & $\log_2\log_2n+O_\varepsilon(1)$ & K (Thm~\ref{thm:fs}b)\\
mixture, $F$, mean & $(e^{-\varepsilon}/2+o(1))\log_2\log_2n$ & K (Thm~\ref{thm:fs}c)\\
causal adaptive, $F$, $\E\tau$ & $(e^{-\varepsilon}/2+o(1))\log_2\log_2n$ & K (Thm~\ref{thm:fsadapt}c)\\
mixture, $F$, fixed cap $R$ & $F\ge n^{3/(4^R-1)-o(1)}$ & K (Thm~\ref{thm:fs}d); no matching upper bound\\
\multicolumn{3}{@{}l}{\emph{Two-value tilts}}\\
fixed schedule, KL, length & $\Theta(\log^2n)$ & K (App.~\ref{app:t1}, \S\S3--4)\\
a relabelled schedule$^\ddagger$, $F$, length & $\log_2\ln n+O(1)$ & K (App.~\ref{app:t1}, \S9)\\
mixture, $H$ / $F$, mean & $\Theta(\log^2n)$ / $(e^{-\varepsilon}+o(1))\log_2\ln n$ & K (Thm~\ref{thm:tilt};\\
 & & App.~\ref{app:t1}, \S\S3--4, 9--10)\\
$\mathcal A_r$, $F$, cap & $\log_2\ln n+O_{\varepsilon,\Lambda,\kappa_0}(1)$ & K (App.~\ref{app:t1}, \S\S9--10)\\
causal adaptive, $F$, $\E\tau$ & $(e^{-\varepsilon}+o(1))\log_2\ln n$ & K (App.~\ref{app:t1}, \S\S9--10)\\
\multicolumn{3}{@{}l}{\emph{Bounded weights, interior quota}}\\
fixed schedule, KL, length & $\Theta(\log^2n)$ & K (App.~\ref{app:t1}, \S\S3--4)\\
a relabelled schedule$^\ddagger$, $F$, length & $\log_2\ln n+O(1)$ & K (Thm~\ref{thm:tilt}b)\\
mixture, $H$, mean & $\Theta(\log^2n)$ & K (App.~\ref{app:t1}, \S\S3--4)\\
$\mathcal A_r$, $F$, cap & $\log_2\ln n+O_{\varepsilon,\Lambda,\kappa_0}(1)$ & K (App.~\ref{app:rf}, Cor.~A$'$; App.~\ref{app:t1}, \S10)\\
mixture, $F$, mean & $(e^{-\varepsilon}+o(1))\log_2\ln n$ & K (App.~\ref{app:rf}, Cor.~A$''$; App.~\ref{app:t1}, \S10)\\
causal adaptive, $F$, $\E\tau$ & $(e^{-\varepsilon}+o(1))\log_2\ln n$ & K (App.~\ref{app:rf}, Cor.~A$''$; App.~\ref{app:t1}, \S10)\\
\midrule
outside: joint draw; rejection & exact in 1 round (resampling not costed) & standard; Lemma~\ref{lem:tools}(e)\\
\bottomrule
\end{tabular}
\end{table*}

\begin{table}[h]
\caption{One round's growth, up to lower-order factors ($M$: number of unrevealed coordinates; $c$: level; $c_{\min}$: start of a band of
levels), and the leading round count: the length of every path for the two profiles, and the scheduled mean $C(\mu)$ for the mixture,
which has rare paths of length $n$. Mixing in the one-round schedule multiplies the leading term of the scheduled mean by $e^{-\varepsilon}$.}\label{tab:growth}
\vspace{4pt}\centering\footnotesize
\begin{tabular}{@{}l l l@{}}
\toprule
design & one round & rounds\\
\midrule
balanced, ladder & unrevealed $M\to\sqrt M$ & $\log_2\log_2n$\\
full support, one profile & level $c\to c^2$ & $\log_2\log_2n$\\
full support, mixture & band $c_{\min}\to c_{\min}^4$ & $\frac12\log_2\log_2n$\\
\bottomrule
\end{tabular}
\end{table}
\clearpage
\begingroup\footnotesize
\setlength{\LTpre}{\smallskipamount}\setlength{\LTpost}{\smallskipamount}\setlength{\LTcapwidth}{\textwidth}
\begin{longtable}{@{}>{\raggedright\arraybackslash}p{0.17\textwidth}>{\raggedright\arraybackslash}p{0.50\textwidth}>{\raggedright\arraybackslash}p{0.27\textwidth}@{}}
\caption{Notation index. The first locator is the first place the paper defines the symbol; a second locator gives
the formal or per-round definition. ``(also: \dots)'' marks a symbol that has another meaning in another row.}\label{tab:notation}\\
\toprule
Symbol & Meaning & Defined in\\
\midrule
\endfirsthead
\toprule
Symbol & Meaning & Defined in\\
\midrule
\endhead
\bottomrule
\endlastfoot
\multicolumn{3}{@{}l}{\emph{Interface, criteria and costs}}\\
$p$, $n$ & target law on $\{0,1\}^n$; $n$ the number of coordinates & \S\ref{sec:intro}; \S\ref{sec:setup}\\
$V$, $k$ & balanced slice $\{x:\sum_ix_i=k\}$, $n=2k$; $k$ the required number of ones (also: $V(s)$) & \S\ref{sec:setup}\\
$(M,j)$ & count state: $M$ coordinates unrevealed, $j$ of them must still be one (also: rung index $j$); with a subscript, an unrevealed count at a stage: $M_{t-1}$ before round $t$ (\S\ref{sec:bal}), the rungs $M_j$ of Theorem~\ref{thm:sep}(b)'s ladder and $M_t$ of Theorem~\ref{thm:conv}(b)'s witness (rows $j$ and $L^\ast$). In \S\ref{sec:tilt} the nominal sizes $M_t$ are deterministic and the free count is $N=|\mathcal C_t|$, random around $M_t$; there the $M$ of $M\ge\rho n$ is the reservoir size of part~(b)'s construction (Theorem~\ref{thm:tilt}'s $j/M$ keeps the count-state meaning) & \S\ref{sec:setup}\\
$U$; $|U|$ (length); $b(U)$ (profile) & labelled schedule $(B_1,\dots,B_r)$, an ordered partition of $[n]$; its length (rounds); its profile (block sizes) & \S\ref{sec:setup}\\
$\sigma$ & a permutation of $[n]$, as in the uniform relabelling $\sigma(U)$; $\sigma(\cdot)$ is also the $\sigma$-algebra generated, as in $\mathcal F_{t-1}$. In the appendices $\sigma$ also denotes, each defined where used, a standard deviation ($\sigma(y)$, $\sigma_n$, $\sigma^2$ in Apps.~\ref{app:thmm}, \ref{app:exhibits}, \ref{app:tgenv}), the logistic function (App.~\ref{app:rf}) and a state $\sigma_t$ (App.~\ref{app:t1}) & \S\ref{sec:setup}\\
$q_U$ & output law when the schedule is $U$ & \S\ref{sec:intro}; \S\ref{sec:setup}\\
$\mu$, $Q_\mu$ & schedule mixture, a finite law on labelled schedules; its output law $\E_\mu q_U$ & \S\ref{sec:intro}; \S\ref{sec:setup}\\
$H(\mu)$ & schedule-averaged divergence $\E_\mu\KL(p\|q_U)$ (also: $\Hc_n$; entropy $H$) & \S\ref{sec:intro}; \S\ref{sec:setup}\\
$J_\mu$ & the sampler's joint law of schedule and output, $J_\mu(U,x)=\mu(U)q_U(x)$, so $H(\mu)=\KL(\mu\otimes p\|J_\mu)$ (also: quota $J$) & \S\ref{sec:intro}\\
$F(\mu)$ & output-law divergence $\KL(p\|Q_\mu)$ & \S\ref{sec:intro}; \S\ref{sec:setup}\\
$C(\mu)$ & scheduled mean $\E_\mu|U|$: expected scheduled rounds, failed runs included (also: kept set $C$) & \S\ref{sec:intro}; \S\ref{sec:setup}\\
$\LH(n,\varepsilon)$, $\LF(n,\varepsilon)$ & least $C(\mu)$ with $H(\mu)\le\varepsilon$, resp.\ $F(\mu)\le\varepsilon$ & \S\ref{sec:intro}; \S\ref{sec:setup}\\
$\mathcal A_r$, $\mathcal A_\infty$ & causal randomized adaptive policies finishing within $r$ rounds a.s.; $\mathcal A_\infty$: no cap (also: $A_r$) & \S\ref{sec:setup}\\
$\pi$, $Q_\pi$ & a policy in $\mathcal A_r$ or $\mathcal A_\infty$; its output law (also: $\pi_i$; $\pi$ in App.~\ref{app:thmm}) & \S\ref{sec:setup}; Lemma~\ref{lem:surv}\\
$\tau$, $\E\tau$ & executed rounds until all coordinates are committed; executed mean, its expectation & \S\ref{sec:intro}; \S\ref{sec:setup}\\
$\ell$, $\ell_k$ & $\ell=\log_2\log_2n$ in Table~\ref{tab:main}; $\ell_k=\log_2\ln(k+1)$ on the slice & Table~\ref{tab:main}; \S\ref{sec:bal}, after Corollary~\ref{cor:t2}\\
$r$, hard cap & of a policy in $\mathcal A_r$: almost every run has at most $r$ rounds (also: $r_\mu$; $r$ in App.~\ref{app:thmm}) & \S\ref{sec:intro}; \S\ref{sec:setup}\\
$r_\mu(U\mid x)$ & law of the schedule given the output (also: cap $r$) & Lemma~\ref{lem:tools}(b)\\
$s(b)$ & success probability of the count process under profile $b$ & Lemma~\ref{lem:tools}(c)\\
\multicolumn{3}{@{}l}{\emph{Balanced slice}}\\
$G_r(M,j)$ & largest probability of reaching $(0,0)$ within $r$ rounds by count-state rules & \S\ref{sec:bal}\\
$A_r$, $d$ & prefactor $6^{2-2^{2-r}}<36$ of \eqref{eq:env}; minority count $d=\min(j,M-j)$ (also: $\mathcal A_r$, parameter $A$; band level $d$, $d=R-t$ of \S\ref{sec:tilt}, $d_U$) & \S\ref{sec:bal}, \eqref{eq:env}\\
$\delta$, $K_\delta$ & $1-e^{-\varepsilon}$, the failure probability allowed; ladder parameter $\min\{\lceil1/\delta\rceil,\lceil2\ln(2/\delta)\rceil\}$ & \S\ref{sec:bal}, before Theorem~\ref{thm:sep}\\
$\Hc_n$, $S_n$ & the harmonic sums $\sum_{m\le n}1/m$ and $\sum_{m\le n}1/m^2$ (also: $H(\mu)$) & Theorem~\ref{thm:sep}\\
$d_U$ & divergence $\KL(p\|q_U)$ of one schedule (also: minority count $d$) & \S\ref{sec:bal}; App.~\ref{app:hlow}\\
$j$ (rung) & index of the ladder rungs $M_j=K\,2^{j+2}\,2^{2^j}$ (also: quota $j$) & \S\ref{sec:bal}, sketch of Theorem~\ref{thm:sep}(b); App.~\ref{app:hlow}\\
$L^\ast$, $M_t$ (witness) & the largest $L\le r$ whose rungs $M_0=n$, $M_t=\lceil\sqrt{2^{L-t+1}M_{t-1}}\,\rceil$ ($t<L$) give nonempty blocks; those rungs, which decrease from $M_0=n$; the ladder rungs $M_j$ instead increase with $j$ and are visited with $j$ decreasing, down to $M_0=8K$ & sketch of Theorem~\ref{thm:conv}(b); App.~\ref{app:conv-window}\\
$w$ (dilution) & weight of a component, as in $Q\ge wQ'$ (also: weights $w$) & \S\ref{sec:bal}, after Corollary~\ref{cor:t2}\\
$R_F^{\mathrm{PROF}}(n,\varepsilon)$ & least length of a uniformly relabelled fixed profile with output KL $\le\varepsilon$ & Corollary~\ref{cor:hard}\\
$R_F^{\mathrm{adapt}}(n,\varepsilon)$ & least hard cap: least $r$ such that some $\pi\in\mathcal A_r$ has $\KL(p\|Q_\pi)\le\varepsilon$ (a minimum on the balanced slice, Theorem~\ref{thm:conv}(a)) & Corollary~\ref{cor:t2}(a)\\
\multicolumn{3}{@{}l}{\emph{Full support}}\\
$W$; $p_W$, $\bar Q_W$ & number of ones; the target's and the output's laws of it (also: log ratio $W$) & \S\ref{sec:fs}\\
$c$, $c'$, $P$ & level $c=2+(\text{coordinates revealed})$; next level; $P=(1+\text{ones revealed})/c$ (also: kept-set level $t$) & \S\ref{sec:fs}; per round ($c_k$, $P_k$): App.~\ref{app:thmm}, Setting\\
band, $c_{\min}$, $d$ & band $\{c_{\min},\dots,2c_{\min}-1\}$, $c_{\min}=\lceil(n+2)^{1/4}(\ln(n+2))^8\rceil$; level $d$ drawn from it (also: $d$ of \eqref{eq:env}; $d=R-t$ of \S\ref{sec:tilt}; $d_U$) & \S\ref{sec:fs}; formula: App.~\ref{app:thmm-mix-upper}\\
$\mathcal N$ & cutoff size of the recursive construction & \S\ref{sec:fs}; as $\mathcal N_{\varepsilon_0}$: App.~\ref{app:thmm-mix-upper}\\
$\mathcal D_m$ & divergence $F(\mu_m)$ of the construction at size $m$ (also: $D_r$) & \S\ref{sec:fs}\\
$\eta_n$ & forced lower bound on the mass of components with more than $R_0(n)$ rounds (also: $\eta_t$) & \S\ref{sec:fs}; formula: Theorem~\ref{thm:mixf}(a)\\
$\vartheta$, $\xi$, $\omega$ & jump exponent ($\vartheta>4$ in the converse), jump factor and radius factor of the converse; a \emph{jump} takes the level from $c$ to at least $\xi c^\vartheta$ (also: $\xi_t$) & \S\ref{sec:fs}; App.~\ref{app:thmm-mix-lower}\\
$R_0(n)$, $S^\star$ & every profile with at most $R_0(n)$ rounds has a jump; the test set of windows around the reduced fractions whose denominators $q$ satisfy $\xi q^\vartheta\le n+2$ (also: $R$) & \S\ref{sec:fs}; App.~\ref{app:thmm-mix-lower}\\
$R$ (cap) & a cap $R\ge1$, fixed as $n$ grows, on the rounds of every component in Theorem~\ref{thm:fs}(d) (also: levels $R$) & Theorem~\ref{thm:fs}\\
$\mathsf U_n$, $\mathsf R_n$ & the target's count law $\mathrm{Unif}\{0,\dots,n\}$; the ideal final count law, used only in the analysis (also: $\mathcal R$) & \S\ref{sec:fs}; App.~\ref{app:thmm-mix-upper}\\
$\LF^{\mathrm{ad}}(n,\varepsilon)$ & least expected executed rounds, $\inf\{\E\tau:\pi\in\mathcal A_\infty,\ \KL(p\|Q_\pi)\le\varepsilon\}$ (full support) & App.~\ref{app:thmm-adapt}\\
$r$, $\pi$ (App.~\ref{app:thmm}) & $r$: ratio of consecutive levels, elsewhere a local index; $\pi$: a count law (also: cap $r$; policy $\pi$) & App.~\ref{app:thmm}, Setting (proofs of Theorems~\ref{thm:thmm}, \ref{thm:cf1})\\
\multicolumn{3}{@{}l}{\emph{Bounded weights}}\\
$w$, $p_w$ & weights in $(0,\infty)^n$; $p_w(x)\propto\prod_iw_i^{x_i}$ on strings with $k$ ones (also: dilution $w$) & \S\ref{sec:tilt}\\
$\Lambda$, $\kappa_0$ & bound on $\max w/\min w$; quota bound $\kappa_0n\le k\le(1-\kappa_0)n$ & \S\ref{sec:tilt}\\
$\LH^w$, $\LF^w$ & $\LH$ and $\LF$ for the target $p_w$ & Theorem~\ref{thm:tilt}\\
$\mathcal C_t$, $C$, $f_t$ & nested kept sets $[n]=\mathcal C_1\supseteq\dots\supseteq\mathcal C_{R+1}$, written $F_t$ in App.~\ref{app:rf}; in one round, the kept set $C=\mathcal C_{t+1}$, as in App.~\ref{app:rf}; retention probability $f_t=M_{t+1}/M_t$ (also: $C(\mu)$, constants $C^*$, $C_H$; level $c$; criterion $F$) & \S\ref{sec:tilt}\\
$A$, $m_0$ & the sampler's accuracy parameter, $A\ge1$, with $F\le C^*/A$ once $A\ge A_0(\Lambda,\kappa_0)$; $m_0=\lceil4A^2\rceil$, the smallest ladder size (also: $A_r$, $\mathcal A_r$) & \S\ref{sec:tilt}\\
$C^*$, $C_H$ & Theorem~A's existence-only constant $C^*=C^*(\Lambda,\kappa_0)$, with $F\le C^*/A$; the upper constant of $\LH^w$ in Theorem~\ref{thm:tilt}(a) (also: $C(\mu)$; kept set $C$) & \S\ref{sec:tilt}; App.~\ref{app:rf}, (4.30)\\
$t$, $R$, $M_t$, $m_j$, $L_t$, $d$ & level index ($\mathcal C_t$ is the free set at level $t$, which keeps $\mathcal C_{t+1}$); number of levels with a kept set, $R=\max\{r\ge1:4m_{r-1}\le n\}$ for $n\ge4m_0$; nominal size of level $t$; ladder sizes; one-step scale $f_t\sqrt{M_t}$; $d=R-t$ (also: cap $R$, $R_0(n)$, $\mathsf R_n$, $\mathcal R$; other $d$) & \S\ref{sec:tilt}; App.~\ref{app:rf}, (4.1)--(4.5)\\
$J$ & quota (ones still required) among the free coordinates; in Theorem~\ref{thm:tilt}(b), the reservoir's residual quota (also: $J_\mu$) & \S\ref{sec:tilt}\\
$\pi_i$ & exact conditional marginal of coordinate $i$, frozen at the round's start (also: policy $\pi$) & \S\ref{sec:tilt}\\
$q_i$ & independent logistic marginal at mean $J$ ($\sum_iq_i=J$), the exact marginals being $\pi_i$ & App.~\ref{app:rf}, Notation discipline\\
$\mathcal R^{\rm joint}$, $\mathcal R$ & reference law (block as sampled, $C$ completed exactly); its laws of $(C,x)$ and of $x$ (also: levels $R$; $\mathsf R_n$) & \S\ref{sec:tilt}\\
$\psi$ & $p/\mathcal R$ (also: local $\psi$ in Apps.~\ref{app:thmm}, \ref{app:tgenv}) & \S\ref{sec:tilt}; App.~\ref{app:rf}\\
$W$ & log ratio of the exact completion of $C$ to the sampler's continuation (also: number of ones $W$) & \S\ref{sec:tilt}\\
$\eta_t$ & bound on $|\psi-1|$ off the exceptional sets (also: $\eta_n$) & \S\ref{sec:tilt}; App.~\ref{app:rf}, (4.16)\\
$\mathcal K_t$ & supremum over good states of the divergence from level $t$ on & \S\ref{sec:tilt}; App.~\ref{app:rf}, (4.13)\\
$\xi_t$ & tail terms of the recursion for $\mathcal K_t$ (also: jump factor $\xi$) & \S\ref{sec:tilt}; App.~\ref{app:rf}, (4.21)\\
\multicolumn{3}{@{}l}{\emph{Two weight values}}\\
$\lambda$, $p_\lambda$ & tilt of group A; $p_\lambda(x)\propto\lambda^{|x_A|}$ on the slice & App.~\ref{app:tgenv}, Setting\\
$G^\lambda_r$ & largest probability that a causal policy with cap $r$ meets the quota & App.~\ref{app:exhibits}, Two groups; recursion: App.~\ref{app:tgenv}\\
$F^*_r$, $D_r$, $F^{\mathrm{NA}}_r$ & least output KL at cap $r$: causal randomized ($\mathrm{RAND}_r$); deterministic; predrawn mixtures (also: $\mathcal D_m$) & App.~\ref{app:exhibits}, Two groups; formal: App.~\ref{app:tgenv}, Corollary~(a) ($F^*_r$), App.~\ref{app:tgred}\\
$V(s)$, $V_{\mathrm{root}}$ & variance of the ones drawn by one round committing every free coordinate (also: slice $V$) & App.~\ref{app:tgenv}, opening paragraph; display: Marginals and variance\\
$H$ (entropy) & Shannon entropy, not the criterion $H(\mu)$ (also: $H(\mu)$) & App.~\ref{app:tgenv}, second remark after Corollary~(c)\\
\end{longtable}
\endgroup

\section{Proofs of Lemma~\ref{lem:tools} and Theorem~\ref{thm:sep}(a),(b)}\label{app:hlow}
\emph{Proof of Lemma~\ref{lem:tools}.} (a) $p$ is supported on $V$; if $q(V)>0$, then $\ln(p(x)/q(x))=\ln(p(x)/q(x\mid V))-\ln
q(V)$ for $x\in V$, and averaging over $p$ gives (a); if $q(V)=0$, both sides are infinite. (b) $F\le H$ is convexity of
$\KL(p\|\cdot)$. If $F(\mu)<\infty$, then $Q_\mu(x)>0$ on $V$, $r_\mu(U\mid x)=\mu(U)q_U(x)/Q_\mu(x)$, and
$\ln Q_\mu(x)-\ln q_U(x)=\ln\mu(U)-\ln r_\mu(U\mid x)$; averaging over $U\sim\mu$ and $x\sim p$ gives the identity (both
sides are infinite if some $q_U$ with $\mu(U)>0$ vanishes somewhere on $V$). This identity is the cross-entropy ambiguity decomposition for
arithmetic-mean ensembles \citep[Prop.~15]{wood2023unified}. (c) At a legal state the $b_t$ draws of
round $t$ are independent $\mathrm{Bernoulli}(j/M)$, so the number of ones is $X_t\sim\Bin(b_t,j/M)$ and the next state is
$(M-b_t,j-X_t)$. The output lies in $V$ exactly when no state is illegal, that is, when the count process reaches $(0,0)$
without being killed, and the law of the count process depends only on $b(U)$. (d) For $x\in V$, $q_U(x)=\prod_t(j_{t-1}/M_{t-1})^{c_t}(1-j_{t-1}/
M_{t-1})^{b_t-c_t}$, where $c_t$ counts the ones of $x$ in $B_t$ and $(M_t,j_t)$ is determined by $c_1,\dots,c_t$; under $p$
the vector $(c_t)_t$ is multivariate hypergeometric with parameters $b(U)$ and $k$, so $\KL(p\|q_U)=\E_p[\ln p(x)-\ln
q_U(x)]$ depends only on $b(U)$. (e) For $x\in V$ every state on the path is legal, so by the product formula in (d)
$q_{\sigma U}(\sigma x)=q_U(x)$ for every permutation $\sigma$ of $[n]$. If $\mu(\sigma U)=\mu(U)$ for all $\sigma$ and $U$, then
$Q_\mu(\sigma x)=Q_\mu(x)$ for $x\in V$; since $V$ is one orbit, $Q_\mu$ is constant on $V$, so $Q_\mu(\cdot\mid V)=p$ when
$Q_\mu(V)>0$, and (a) gives $F(\mu)=-\ln Q_\mu(V)$. For any law $Q$ with $Q(\sigma x)=Q(x)$ for all $\sigma$ and $x\in V$
the last two steps apply verbatim: $Q$ is constant on the orbit $V$, so $Q(\cdot\mid V)=p$ when $Q(V)>0$, and (a) gives $\KL(p\|Q)=-\ln Q(V)$ (both sides are infinite when $Q(V)=0$). \qed

\emph{A pointwise mixture bound.} If $Q$ and $Q'$ are laws on the same finite set and $Q\ge wQ'$ pointwise for some $w\in(0,1]$, then $\KL(p\|Q)\le\KL(p\|Q')+\ln(1/w)$. If $\KL(p\|Q')=+\infty$ there is nothing to prove; otherwise $Q'(x)>0$ wherever $p(x)>0$, so $Q(x)\ge wQ'(x)>0$ there and $\ln(p(x)/Q(x))\le\ln(p(x)/Q'(x))+\ln(1/w)$, and averaging over $x\sim p$ gives the bound. Corollary~\ref{cor:t2}(c) uses it for a mixture $Q=wQ'+(1-w)Q''$.

\emph{Proof of Theorem~\ref{thm:sep}(a).} Let $U$ have block sizes $b_1,\dots,b_R$, residual sizes $M_0=n>M_1>\dots>M_R=0$, and $\rho_t=b_t/M_{t-1}\in(0,1]$
($\rho_R=1$). Write $d_U=\KL(p\|q_U)$, $a_U=\sum_t\rho_t$, $\beta_U=\sum_t\rho_t^2$. Small divergence forces the fractions $\rho_t$ to add up to nearly $\Hc_n$ while their squares add up to $O(1)$:
Steps 1--4 prove $a_U\ge\Hc_n-2d_U$ and $\beta_U\le4d_U+S_n$, and the key step is the per-reveal entropy decrement, the increment bound of
Step~2, a Pinsker bound. Both bounds are affine in $d_U$, so Step~5 averages them over a feasible mixture, and
Cauchy--Schwarz over the rounds before the last gives $(\E_\mu a-1)^2\le(C(\mu)-1)(\E_\mu\beta-1)$, which forces $C(\mu)$
to be at least of order $(\ln n)^2$ at fixed $\varepsilon$; Step~6 bounds the two harmonic sums to give the last form of
Theorem~\ref{thm:sep}(a).

\emph{Step 1 (chain identity).} Let $h$ be the binary entropy in nats and $H_1(M)=\E\,h(J_M/M)$, where $J_M$ is the
number of ones among $M$ unrevealed coordinates under $p$ (hypergeometric; by exchangeability it depends only on $M$). Given
the state before round $t$, where $J$ ones remain among the $M=M_{t-1}$ unrevealed coordinates, each coordinate of $B_t$ is one with probability $J/M$ under $p$, and the round-$t$ factor of $q_U$ is $(J/M)^{c_t}(1-J/M)^{b_t-c_t}$, with $c_t$ the number of ones in $B_t$; so the target
expectation of minus its log, given the state, is $b_th(J/M)$. As $J$ has the law of $J_{M_{t-1}}$, and $-\ln p=\ln\binom nk$ on $V$,
$d_U=\sum_tb_tH_1(M_{t-1})-\ln\binom nk$. The all-singleton schedule is exact, so $\sum_{m=1}^nH_1(m)=\ln\binom nk$ and
$d_U=\sum_t\sum_{M_t<m\le M_{t-1}}[H_1(M_{t-1})-H_1(m)]$.

\emph{Step 2 (increment).} For $2\le m\le n$, $H_1(m)-H_1(m-1)\ge n/(2(n-1)m(m-1))$. Indeed, with $P=J_m/m$ and one
more revealed coordinate $X\sim\mathrm{Bernoulli}(P)$, $P'=(J_m-X)/(m-1)$ satisfies $\E[P'\mid P]=P$, and $\E h(P')=H_1(m-1)$ because $J_m-X$ has the law of $J_{m-1}$ under $p$; so
$h(P)-\E[h(P')\mid P]=\E[d(P'\|P)\mid P]\ge2\E[(P'-P)^2\mid P]=2P(1-P)/(m-1)^2$. The equality holds because $d(x\|y)=h(y)-h(x)+(y-x)\ln\frac{y}{1-y}$ for $y\in(0,1)$ and $\E[P'-P\mid P]=0$ (if $P\in\{0,1\}$
then $P'=P$ and both sides vanish); the inequality is Pinsker's inequality for the binary KL divergence $d(\cdot\|\cdot)$ in nats;
and $P'-P=(P-X)/(m-1)$ has conditional variance $P(1-P)/(m-1)^2$. Finally $J_m$ is hypergeometric with $\E J_m=m/2$ and
$\mathrm{Var}\,J_m=\frac m4\cdot\frac{n-m}{n-1}$ (as $k=n/2$), so
$\E P(1-P)=\frac14-\mathrm{Var}(J_m)/m^2=\frac14\big(1-\frac{n-m}{m(n-1)}\big)=n(m-1)/(4m(n-1))$.

\emph{Step 3.} For $M_t<m\le M_{t-1}$, summing Step 2 over the indices $m+1,\dots,M_{t-1}$, where
$1/(m'(m'-1))=1/(m'-1)-1/m'$ telescopes, gives $H_1(M_{t-1})-H_1(m)\ge c_n(1/m-1/M_{t-1})$ with
$c_n=n/(2(n-1))\ge\frac12$; summing over the block and over $t$, $d_U\ge c_n(\Hc_n-a_U)$, where
$\Hc_n-a_U=\sum_t\sum_{M_t<m\le M_{t-1}}(1/m-1/M_{t-1})\ge0$ termwise; so $a_U\ge\Hc_n-2d_U$.

\emph{Step 4.} For one block of size $b$ taken from $M$, $1/m\ge1/M+(M-m)/M^2$, so
$\sum_{m=M-b+1}^M(1/m-1/M)\ge b(b-1)/(2M^2)$ and $(b/M)^2\le2\sum(1/m-1/M)+\sum_{m=M-b+1}^M1/m^2$. Summing over blocks and
using Step 3, $\beta_U\le2(\Hc_n-a_U)+S_n\le4d_U+S_n$.

\emph{Step 5 (mixtures).} The bounds of Steps 3--4 are affine in $d_U$, so averaging them over a mixture with $\E_\mu d_U\le\varepsilon$ gives
$\E_\mu a\ge\Hc_n-2\varepsilon$ and $\E_\mu\beta\le4\varepsilon+S_n$. Dropping the last block of each schedule
($\rho_R=1$) and applying Cauchy--Schwarz to the measure $\mu(U)\times$(counting over $t<|U|$),
$(\E_\mu a-1)^2\le(C(\mu)-1)(\E_\mu\beta-1)$. If $\Hc_n-2\varepsilon-1>0$ this gives
$C(\mu)\ge1+(\Hc_n-2\varepsilon-1)^2/(4\varepsilon+S_n-1)$; otherwise $C(\mu)\ge1$ trivially. The bound
$C(\mu)\ge\E_\mu a\ge\Hc_n-2\varepsilon$ gives the middle term of the maximum in Theorem~\ref{thm:sep}(a).

\emph{Step 6.} $\Hc_n\ge\ln(n+1)$ and $S_n-1\le\sum_{m\ge2}(1/(m-\frac12)-1/(m+\frac12))=2/3$. \qed

\pend{}

\paragraph{Theorem~\ref{thm:sep}(a)'s rate from a prior depth bound.} Theorem~\ref{thm:sep}(a)'s rate also follows from \citet[Theorem~11]{wen2026conditional}, the deterministic depth bound described in Section~\ref{sec:intro} (depth: their least mean number of rounds over draws from the target; for a fixed schedule, its length), by a Markov argument. If $H(\mu)\le\varepsilon$, the components with $\KL(p\|q_U)\le2\varepsilon$ carry
$\mu$-mass at least $\frac12$. Each is a fixed schedule, a deterministic policy whose depth is $|U|$ on every path, so
their Theorem~11 at tolerance $2\varepsilon$ ($2\varepsilon/\ln2$ bits) gives $|U|=\Omega_\varepsilon(\log^2n)$ for each of them, and
$C(\mu)\ge\frac12\min|U|$. Appendix~\ref{app:related} compares the paper with selected related work.

\emph{Proof of Theorem~\ref{thm:sep}(b).} For a parameter $K\ge1$, the witness relabels uniformly one profile that,
when $n>8K$, brings the number of unrevealed coordinates down to $8K$ through the rungs of a ladder, each rung about the square root
of the next one up (the first bulk round goes from $n$ to the highest rung below it), and then commits the last $8K$
coordinates as singletons. By Lemma~\ref{lem:tools}(e) its $F$ is $-\ln$ of its
success probability, and the key step is that, before failure, the fraction of ones still required is a martingale whose conditional
variances over the bulk rounds sum to less than $1/(4K)$. In this proof $j$ indexes the rungs $M_j$ of the ladder, which
are visited in decreasing order of $j$, and is not the quota of a count state. Let $M_j=K\,2^{j+2}\,2^{2^j}$, so $M_0=8K$,
$M_j=M_{j-1}^2/(K2^j)$ and
$M_j\ge2^{2^j}$. Let $T$ be the least index with $n\le M_T$ ($T\le\lceil\log_2\log_2n\rceil$). Use blocks that reduce the
number of unrevealed coordinates from $n$ to $M_{T-1}$, then $M_{T-2}$, \dots, $M_0$, then $M_0$ singletons (if
$n\le M_0$, use $n$ singletons), so at most $\min\{n,T+M_0\}\le\min\{n,\lceil\log_2\log_2n\rceil+8K\}$ rounds. Write $N_t$ and $J_t$ for the numbers of unrevealed coordinates and of ones still required after round $t$. Before
failure, $P_t=J_t/N_t$ satisfies $P_{t+1}-P_t=(b_tP_t-X_t)/N_{t+1}$, where $b_t=N_t-N_{t+1}$ is the size of round $t+1$'s block and, given the past,
$X_t\sim\mathrm{Bin}(b_t,P_t)$ is the number of ones it draws; this is a martingale difference
with conditional variance at most $b_t/(4N_{t+1}^2)$ that is a sum of $b_t$ independent terms of range $1/N_{t+1}$. Freeze $P$
at its first exit from $(0,1)$ and let $S=\sum_{\text{bulk }t}b_t/N_{t+1}^2$. A bulk round that ends at $M_{j-1}$ starts
from at most $M_j$, so $b_t/N_{t+1}^2\le M_j/M_{j-1}^2=2^{-j}/K$, and $S\le\sum_{j=1}^{T}2^{-j}/K<1/K$. Kolmogorov's maximal inequality for martingales (Doob's) gives $\Pr(\max_t|P_t-\tfrac12|\ge\tfrac12)\le
4\cdot S/4<1/K$. For the second choice of $K$: given the past, $P_{t+1}-P_t=\sum_{i\le b_t}(P_t-\xi_i)/N_{t+1}$ with $\xi_i$
independent Bernoulli$(P_t)$, a sum of $b_t$ independent mean-zero terms of range $1/N_{t+1}$, so Hoeffding's lemma gives
$\E[e^{\lambda(P_{t+1}-P_t)}\mid\mathcal F_t]\le\exp(\lambda^2b_t/(8N_{t+1}^2))$ for every real $\lambda$. As the block sizes are fixed and
the frozen process has zero increments, $\exp(\lambda(P_t-\frac12)-\lambda^2S_t/8)$, with $S_t$ the partial sums of $S$, is a
nonnegative supermartingale started at $1$, and Ville's inequality gives $\Pr(\max_t(P_t-\frac12)\ge\frac12)\le\exp(-\lambda/2+\lambda^2S/8)$
for $\lambda>0$, that is $e^{-1/(2S)}$ at $\lambda=2/S$. With the same bound for $\frac12-P_t$,
$\Pr(\max_t|P_t-\tfrac12|\ge\tfrac12)\le2e^{-1/(2S)}<2e^{-K/2}$. A bulk round fails only if $P$ leaves $(0,1)$; boundary values are safe and a singleton
round from a legal state never fails, so success is at least $1-\delta$ with $K=\lceil1/\delta\rceil$ or
$K=\lceil2\ln(2/\delta)\rceil$. Relabelling uniformly makes $Q_\mu$ exchangeable, and Lemma~\ref{lem:tools}(e) gives
$F=-\ln(\text{success})\le\varepsilon$. \qed

\section{Proof of the fixed-schedule upper bound in Theorem~\ref{thm:sep}(c)}\label{app:lhup}
We prove the bound of Theorem~\ref{thm:sep}(c) in explicit form: for $n\ge16$ and $0<\varepsilon\le\ln2$, one fixed geometric
schedule gives $\LH(n,\varepsilon)\le(\ln n)(\ln n+2)/\varepsilon+1$; for $\varepsilon>\ln2$, $\LH(n,\varepsilon)\le\LH(n,\ln2)$,
as $\LH$ is nonincreasing in $\varepsilon$. As in Appendix~\ref{app:hlow}, a labelled
schedule $U$ has block sizes $b_1,\dots,b_R$ ($|U|=R$) and residual sizes $M_0=n>M_1>\dots>M_R=0$, so block $t$ is
drawn when $M_{t-1}$ coordinates are unrevealed. Step~1 bounds the divergence of the target's law of one block from the product law that a round draws, Step~2 adds these
bounds over a fixed schedule, Step~3 bounds the sum for a geometric profile, and Step~4 chooses its fraction
$\alpha=\varepsilon/\ln n$.

\emph{Step 1 (per-block bound).} At a legal state $(M,j)$ let $\mathrm{WOR}(M,j,b)$ be the law of the bits of
$b\le M$ unrevealed coordinates under $p$ (sampling $b$ balls without replacement from an urn with $j$ ones among
$M$), and let $\mathrm{Bernoulli}(j/M)^{\otimes b}$ be the law of the independent bits that one round commits there. Then
\begin{equation*}
\KL\bigl(\mathrm{WOR}(M,j,b)\,\big\|\,\mathrm{Bernoulli}(j/M)^{\otimes b}\bigr)\le\frac{b(b-1)}{2(M-1)(M-b+1)}
\qquad(\text{read as }0\text{ for }b=1).
\end{equation*}
This is the two-colour case of a bound of \citet{stam1978distance}: both laws are uniform given their number of ones, so their divergence is that of the hypergeometric and binomial counts. We include a short proof.
\emph{Proof.} If $j/M\in\{0,1\}$ both laws are the same point mass. Otherwise, by the chain rule with a product
reference, $\KL=\sum_{i=1}^{b}\E\,\KL(\mathrm{Bernoulli}(P_i)\,\|\,\mathrm{Bernoulli}(P_1))$, where
$P_i=(j-c_{<i})/(M-i+1)$ is the fraction of ones left after $i-1$ draws, $c_{<i}$ being the number of ones among
those draws (so $P_1=j/M$). Moreover $\KL(\mathrm{Bernoulli}(a)\,\|\,\mathrm{Bernoulli}(P_1))\le\chi^2
=(a-P_1)^2/(P_1(1-P_1))$. Under sampling without replacement $\E P_i=P_1$ and
$\mathrm{Var}\,P_i=P_1(1-P_1)(i-1)/((M-1)(M-i+1))$, because $c_{<i}$ is hypergeometric with mean $(i-1)P_1$ and variance
$(i-1)P_1(1-P_1)(M-i+1)/(M-1)$, and $P_i=(j-c_{<i})/(M-i+1)$. So the $i$-th term is at most
$(i-1)/((M-1)(M-i+1))\le(i-1)/((M-1)(M-b+1))$; summing over $i$ gives the claim. \qed
\emph{Step 2 (fixed schedules).} For a fixed schedule $U$ with size profile $b=(b_1,\dots,b_R)$,
\begin{equation*}
\KL(p\|q_U)\le\Phi(b):=\sum_{t=1}^{R}\frac{b_t(b_t-1)}{2(M_{t-1}-1)(M_{t-1}-b_t+1)},
\end{equation*}
by the chain rule over blocks (every $x\in V$ passes only through legal states; given the committed prefix, the
bits of block $t$ have law $\mathrm{WOR}(M_{t-1},j,b_t)$ under $p$ and $\mathrm{Bernoulli}(j/M_{t-1})^{\otimes b_t}$
under $q_U$) and Step~1; $M_{t-1}$ is deterministic for a fixed schedule, so no expectation over $j$ remains.
\emph{Step 3 (geometric profile).} For $\alpha\in(0,\frac12]$ let $b_t=\max(1,\lfloor\alpha M_{t-1}\rfloor)$
(geometric, then singletons). Then:
(i) each block with $b_t\ge2$ contributes at most $\alpha^2/(2(1-\alpha))$ to $\Phi(b)$, because
$b(b-1)\le\alpha^2M(M-1)$ and $M-b+1>(1-\alpha)M$ (here $b=b_t$, $M=M_{t-1}$);
(ii) the blocks with $b_t\ge2$ come first, as $b_t$ is nonincreasing; while $b_t\ge2$,
$M_t-1/\alpha\le(1-\alpha)(M_{t-1}-1/\alpha)$, as $b_t=\lfloor\alpha M_{t-1}\rfloor>\alpha M_{t-1}-1$, and $b_t\ge2$ needs $M_{t-1}-1/\alpha\ge1/\alpha$, so if there are $N_2\ge1$ such
blocks then $1/\alpha\le M_{N_2-1}-1/\alpha\le(1-\alpha)^{N_2-1}(n-1/\alpha)<ne^{-\alpha(N_2-1)}$; hence
$N_2\le\max\{0,\ln(n\alpha)/\alpha+1\}$, and these blocks are followed by fewer than $2/\alpha$ singletons (which contribute $0$);
(iii) $\Phi(b)\le\max\{0,(\alpha\ln(n\alpha)+\alpha^2)/(2(1-\alpha))\}$ and
$|U|\le\max\{0,\ln(n\alpha)/\alpha+1\}+2/\alpha$.
\emph{Step 4 (the choice $\alpha=\varepsilon/\ln n$).} Take $\alpha=\varepsilon/\ln n$. For $n\ge16$ and
$0<\varepsilon\le\ln2$ (so $\alpha\le\frac14$ and $\varepsilon/\ln^2n\le0.091$), $\Phi(b)\le\varepsilon(1.091)/1.5
<\varepsilon$.
Hence, by Step~2 and (iii), this geometric schedule $U$ has $\KL(p\|q_U)<\varepsilon$ and
$|U|\le(\ln n)(\ln n+2)/\varepsilon+1$. The point mass at $U$ has $H=\KL(p\|q_U)$ and $C=|U|$, so
\begin{equation*}
\LH(n,\varepsilon)\le(\ln n)(\ln n+2)/\varepsilon+1,
\end{equation*}
and the same bound holds for the least length of a single fixed schedule with $\KL(p\|q_U)\le\varepsilon$
(Corollary~\ref{cor:hard}). \qed
\emph{Remark (asymptotic constant).} With $\alpha=\varepsilon/\ln n$, only $|U|\le(1+o(1))(\ln n)^2/\varepsilon$
follows: this choice uses only about half of the KL budget, as the chain's bound on $\Phi$ is then $(1+o(1))\varepsilon/2$. With $\alpha=2\varepsilon/\ln n$, the same chain gives
$\Phi(b)\le\max\{0,\alpha\ln(n\alpha)+\alpha^2\}/(2(1-\alpha))$, which is at most $\varepsilon$ once
$\ln(\ln n/(2\varepsilon))\ge2\varepsilon(1+1/\ln n)$ ($n\gtrsim10^3$ at $\varepsilon=\ln2$; $n\ge16$ for
$\varepsilon\le0.3$). This gives a single fixed schedule with $\KL(p\|q_U)\le\varepsilon$ and
$|U|\le(1+o(1))(\ln n)^2/(2\varepsilon)$ (for fixed $\varepsilon$, as $n\to\infty$).
\pend{}

\section{Proof of the survival envelope}\label{app:env}
\begin{proposition}\label{prop:env}
For $r\ge1$ and legal $(M,j)$ with $d=\min(j,M-j)$,
\begin{equation}\label{eq:env}
G_r(M,j)\le\min\{1,\,A_r(d+1)^{-\alpha_r}\},\qquad A_r=6^{2-2^{2-r}}<36,\qquad\alpha_r=2^{-r}.\tag{Env}
\end{equation}
\end{proposition}
This is the envelope of Section~\ref{sec:bal}; Theorem~\ref{thm:conv}(b) and Corollaries~\ref{cor:hard} and~\ref{cor:t2} use it (Appendix~\ref{app:conv}). The prefactors satisfy $A_1=1$ and $A_{r+1}=A_r\,6^{2^{1-r}}$, so
$A_r=6^{\sum_{s=1}^{r-1}2^{1-s}}<6^2$; for example $A_2=6$ and $A_3=6^{3/2}$.

\emph{Notation.} For legal $(M,j)$ with $M\ge1$ write $\theta=j/M$, so that $d=M\min(\theta,1-\theta)$, and write
$d(M',j')=\min(j',M'-j')$. For a block of size $b$ at $(M,j)$ let $m=M-b$. The surviving successor quotas are
$j'\in[0,m]$, with kernel $K_b(j,j')=\Bin(b,\theta)(j-j')$ (killed mass dropped, never renormalized, so
$\sum_{j'}K_b(j,j')\le1$). Put $d'=d(m,j')$ and, for $\alpha\ge0$,
$T_b(\alpha)=\sum_{j'=0}^{m}K_b(j,j')\,(d'+1)^{-\alpha}$.

\emph{Plan.} Only two estimates are analytic: the $r=1$ success is a central binomial atom, (A1), and one round bounds
the expected potential of the next minority count, $T_b(\frac12)\le6(d+1)^{-1/4}$, (A3), with the uniform atom bound
(A2) as a tool. The exponent halves once per round: (A3) takes the exponent $\frac12$ to $\frac14$, and the Cauchy--Schwarz halving (A4),
iterated, gives $T_b(2^{-r})\le T_b(\frac12)^{2^{1-r}}$, so the same step applies at every exponent $2^{-r}$; the
induction collects one factor $6^{2^{1-r}}$ per round into $A_r$.
\emph{(A1) Central atom.} $\Bin(M,j/M)(j)\le(d+1)^{-1/2}$; for $d\ge1$, even $\Bin(M,j/M)(j)\le(\pi d)^{-1/2}$.
\emph{Proof.} If $d=0$ the atom is $1$. For $1\le j\le M-1$ let $\varsigma(t)=t!/(\sqrt t\,(t/e)^t)$, which decreases
to $\sqrt{2\pi}$: $\ln(\varsigma(t+1)/\varsigma(t))=1-(t+\frac12)\ln(1+\frac1t)<0$ because $\ln(1+x)>2x/(2+x)$ for $x>0$, and the
limit is Stirling's formula. Then
\begin{equation*}
\Bin(M,j/M)(j)=\binom Mj\frac{j^j(M-j)^{M-j}}{M^M}=\frac{\varsigma(M)}{\varsigma(j)\,\varsigma(M-j)}\sqrt{\frac{M}{j(M-j)}} .
\end{equation*}
Since $\varsigma(M)\le\varsigma(M-j)$ and $\varsigma(j)\ge\sqrt{2\pi}$, the fraction is at most $1/\sqrt{2\pi}$. And
$M/(j(M-j))=1/j+1/(M-j)\le2/d$. So the atom is at most $(\pi d)^{-1/2}\le(d+1)^{-1/2}$, because $\pi d\ge d+1$ for
$d\ge1$. \qed

At $(M,j)=(n,k)$, the central binomial local limit gives $-\ln G_1(n,k)=\tfrac12\ln(\pi n/2)+o(1)$.
\emph{(A2) Uniform atom.} For $b\ge1$, $\theta\in[0,1]$, $\nu:=b\min(\theta,1-\theta)$ and every integer $z$,
$\Bin(b,\theta)(z)^2\,\nu\le49/16$.
\emph{Proof.} If $\nu\le49/16$ use $\Bin(b,\theta)(z)\le1$. Otherwise let $d_z=\min(z,b-z)$ (for $z\in[0,b]$; for
other $z$, $\Bin(b,\theta)(z)=0$).
\emph{Bulk:} if $d_z+1\ge(16/49)\nu$, the map $\theta'\mapsto\Bin(b,\theta')(z)$ is maximised at $\theta'=z/b$ (the
maximum-likelihood value), so by (A1) with $(M,j)=(b,z)$, $\Bin(b,\theta)(z)^2\le1/(d_z+1)\le49/(16\nu)$.
\emph{Edge:} if $d_z+1<(16/49)\nu$, then $z$ (or $b-z$) is $<(16/49)\nu$, while $b\theta\ge\nu$ and
$b(1-\theta)\ge\nu$, so $|z-b\theta|>(33/49)\nu$. Chebyshev's inequality with variance $b\theta(1-\theta)\le\nu$ gives
$\Bin(b,\theta)(z)\le(49/33)^2/\nu$, hence
$\Bin(b,\theta)(z)^2\,\nu\le(49/33)^4/\nu<(49/33)^4\cdot16/49<1.6<49/16$. \qed
\emph{(A3) One step at exponent $\frac12$.} For $1\le b\le M$ and $0\le j\le M$,
$T_b(\frac12)\le\min\{1,6(d+1)^{-1/4}\}$.

The proof of (A3) below splits by block size. A block with $8b<7M$ leaves more than $M/8$ coordinates unrevealed,
enough to absorb the binomial fluctuation of its draws, so by Chebyshev's inequality the next minority count $d'$ exceeds
$d/16$ except with probability below $256/d$. A larger block may leave too few coordinates for that, but then every atom
of the kernel is at most $(7/4)\nu^{-1/2}$ by (A2), and a cutoff estimate bounds the contribution of small $d'$. We first
record that estimate. \emph{Cutoff.} If $K_b(j,j')\le a$ for all $j'$, then for every integer $v_0\ge1$,
$T_b(\frac12)\le4a\sqrt{v_0}+(v_0+1)^{-1/2}$. Indeed, split the $j'$ with $d'<v_0$ from the rest. The rest contribute
at most $(v_0+1)^{-1/2}\sum_{j'}K_b(j,j')\le(v_0+1)^{-1/2}$. For the $j'$ with $d'<v_0$, split $[0,m]$ at $m/2$: for
$j'\le m/2$, $d'=j'$; for $j'>m/2$, $d'=m-j'$. So on each half every value $d'\in\{0,\dots,v_0-1\}$ occurs at most
once, the weights $(d'+1)^{-1/2}$ run over $1^{-1/2},\dots,v_0^{-1/2}$ at most once per half, and
$\sum_{v=1}^{v_0}v^{-1/2}\le2\sqrt{v_0}$; with $K_b(j,j')\le a$ this part is at most $4a\sqrt{v_0}$.
\emph{Proof of (A3).} $T_b(\frac12)\le\sum_{j'}K_b(j,j')\le1$ always; since $6(d+1)^{-1/4}\ge1$ if and only if
$d\le1295$ ($6^4=1296$), assume $d\ge1295$.
\emph{Small blocks, $8b<7M$.} Let $\gamma=m/M>1/8$ and $X=j-j'\sim\Bin(b,\theta)$. Write $j'=\theta m+(\theta b-X)$. If
$|X-\theta b|<\gamma d/2$, then $j'>\theta m-\gamma d/2$ and $m-j'>(1-\theta)m-\gamma d/2$, and
$\min(\theta m,(1-\theta)m)=\gamma d$, so $d'>\gamma d/2>d/16$ and those $j'$ contribute at most
$(d/16)^{-1/2}=4/\sqrt d$. The complementary event has probability at most
$b\theta(1-\theta)/(\gamma d/2)^2\le d/(\gamma^2d^2/4)<256/d$ (using $b\theta(1-\theta)\le M\min(\theta,1-\theta)=d$).
So $T_b(\frac12)\le4/\sqrt d+256/d\le6(d+1)^{-1/4}$ for $d\ge1295$ (the ratio $(4d^{-1/2}+256d^{-1})(d+1)^{1/4}/6$
decreases in $d$ (its logarithmic derivative is at most $-\frac1{2d}+\frac1{4(d+1)}<0$) and is $0.309$ at $d=1295$; the inequality is also kernel-checked as \texttt{AGF.agfv\_c08s\_S\_numeric}).
\emph{Large blocks, $8b\ge7M$.} Here $\nu=b\min(\theta,1-\theta)\ge(7/8)d$. By (A2), $K_b(j,j')\le a:=(7/4)\nu^{-1/2}$.
Take $v_0=\lfloor\lfloor\sqrt d\rfloor/8\rfloor$ (this cutoff balances the two terms of the estimate below: with $56v_0^2\le\nu$ the
small-$d'$ term is at most $\sqrt{7/8}\,v_0^{-1/2}$, like the cutoff term, and both are of order $d^{-1/4}$); for $d\ge1295$: $v_0\ge4$, $56v_0^2\le7d/8\le\nu$, and
$\sqrt{d+1}\le8(v_0+1)$ (since $\lfloor\sqrt d\rfloor\ge35$ and $8v_0\le\lfloor\sqrt d\rfloor\le8v_0+7$). By the cutoff estimate,
\begin{equation*}
T_b(\tfrac12)\le\frac{7\sqrt{v_0}}{\sqrt\nu}+(v_0+1)^{-1/2}\le\sqrt{7/8}\;v_0^{-1/2}+(v_0+1)^{-1/2}
\le\frac{6}{\sqrt{8(v_0+1)}}\le6(d+1)^{-1/4},
\end{equation*}
where the third inequality holds for $v_0\ge4$ (it is equivalent to $1+1/v_0\le c^2$ with $c=(3/\sqrt2-1)\sqrt{8/7}$,
i.e.\ to $v_0\ge2.29$; the three facts and this inequality are also kernel-checked inside
\texttt{AGF.agfv\_c08s\_step\_regimeF} and \texttt{AGF.agfv\_c08s\_F\_numeric}). \qed
\emph{(A4) Cauchy--Schwarz halving.} $T_b(\alpha)^2\le\bigl(\sum_{j'}K_b(j,j')\bigr)\,T_b(2\alpha)\le T_b(2\alpha)$,
by Cauchy--Schwarz on $\sum_{j'}K_b(j,j')^{1/2}\cdot K_b(j,j')^{1/2}(d'+1)^{-\alpha}$. Iterating,
$T_b(2^{-(t+1)})^{2^t}\le T_b(\frac12)$ for all $t\ge0$. \qed
\emph{Proof of the proposition.} Induction on $r$. $G_r\le1$ for all $r$ (each $b$-term is at most
$\sum_{j'}K_b(j,j')\cdot1$), so it suffices to show $G_r(M,j)\le A_r(d+1)^{-\alpha_r}$. If $M=0$, the only legal state
is $(0,0)$, where $d=0$ and $G_r(0,0)=1\le A_r$, as $A_r\ge1$. Let $M\ge1$. For $r=1$ only $b=M$ survives the horizon,
so $G_1(M,j)=\Bin(M,\theta)(j)\le(d+1)^{-1/2}=A_1(d+1)^{-\alpha_1}$ by (A1). Step $r\to r+1$ ($r\ge1$): by the induction
hypothesis at every legal successor (including $m=0$), for every $b$
\begin{align*}
\sum_{j'}K_b(j,j')\,G_r(m,j')&\le A_r\,T_b(\alpha_r)\le A_r\,T_b(\tfrac12)^{2^{1-r}}
\quad\text{((A4) with }t=r-1\text{)}\\
&\le A_r\bigl[6(d+1)^{-1/4}\bigr]^{2^{1-r}}=A_r\,6^{2^{1-r}}(d+1)^{-2^{-(r+1)}}
\quad\text{((A3))};
\end{align*}
maximise over $b$. Since $A_r\,6^{2^{1-r}}=A_{r+1}$ and $2^{-(r+1)}=\alpha_{r+1}$, this is the claim for $r+1$. \qed

\emph{Where substochasticity is used.} In the final induction, after (A3) is established, substochasticity enters
through (A4) and $G_r\le1$; (A3)'s cutoff and trivial-range arguments also use $\sum_{j'}K_b(j,j')\le1$.
\pend{}
\emph{The $\kappa_r$ form.} Put $\kappa_r=2\cdot6^{r-1}$. Then $A_1<\kappa_1$, and $A_{r+1}\le6A_r$ while
$\kappa_{r+1}=6\kappa_r$, so $A_r\le\kappa_r$ for every $r\ge1$ and the proposition implies
$G_r(M,j)\le\min\{1,\kappa_r(d+1)^{-\alpha_r}\}$. At the balanced start $(M,j)=(n,k)$, $d=k$, this reads
$G_r(n,k)\le\min\{1,2\cdot6^{r-1}(k+1)^{-2^{-r}}\}=:\Gamma_r(n)$. Step~6 of the single-horizon bound in Appendix~\ref{app:conv} is stated with this form; Appendix~\ref{app:lean}
lists the kernel-checked declarations for both forms.

At the balanced start the proposition's bound is non-vacuous only for $n>2590$ at $r=2$, only for $n>4.3\cdot10^9$
at $r=3$, and only for larger $n$ as $r$ grows (the $\kappa_r$ form needs $n>4.1\cdot10^4$ and $n>1.4\cdot10^{15}$): although it holds at every $n$, it is informative only at large $n$, and
finite exhibits must use the exact $G_r$ from the recursion. The proposition is a statement about the recursion only. It
reaches samplers through the tower argument of Appendix~\ref{app:conv} (Lemma~\ref{lem:surv}, on which
Theorem~\ref{thm:conv}(a) and Corollary~\ref{cor:t2}(b) rest) or, for fixed schedules, through that argument's elementary
special case $s(b)\le G_{|b|}(n,k)$ (Lemma~\ref{lem:surv}(a)). At $(M,j)=(n,k)$ it gives the lower side of
Theorem~\ref{thm:conv}(b), the lower bound for relabelled profiles in Corollary~\ref{cor:hard} and the lower bounds of
Corollary~\ref{cor:t2}(a),(b); Theorem~\ref{thm:conv}(a) itself is exact and does not use it.
\pend{}

\section{Proofs of Lemma~\ref{lem:surv}, Theorem~\ref{thm:conv} and Corollary~\ref{cor:t2}}\label{app:conv}
Every converse bound in this appendix starts from one fact: given the seed and the committed history, a round's draws are
$\mathrm{Bernoulli}(j/M)$ products whose law depends only on the count state, so $G$ of the remaining horizon is a
supermartingale and no policy in $\mathcal A_\infty$ reaches $(0,0)$ within $t$ rounds with probability above $G_t(n,k)$
(Lemma~\ref{lem:surv}(a)). With Lemma~\ref{lem:tools}(a), Lemma~\ref{lem:surv}(a) gives $\KL(p\|Q_\pi)\ge-\ln G_r(n,k)$
for every $\pi\in\mathcal A_r$, and a maximizing count-state rule applied in a uniformly random order attains equality by
Lemma~\ref{lem:tools}(e) (Theorem~\ref{thm:conv}(a)). A valid output with $\tau\le s$ needs the state to reach $(0,0)$
within $s$ rounds, so $\Pr(\tau>s)\ge Q_\pi(V)-G_s(n,k)$, and summing these tails bounds $\E\tau$ from below
(Lemma~\ref{lem:surv}(b)); Corollary~\ref{cor:t2} combines these facts with \eqref{eq:env}, the ladder of
Theorem~\ref{thm:sep}(b) and, in part (c), its dilution by the one-round schedule. The next paragraph proves
Lemma~\ref{lem:surv}(b), Theorem~\ref{thm:conv}(a) and Corollary~\ref{cor:t2}(c) and outlines Lemma~\ref{lem:surv}(a);
the paragraphs after it give the details of Lemma~\ref{lem:surv}(a), the proof of Corollary~\ref{cor:t2}(a),(b) and the
total-variation forms, and Appendix~\ref{app:conv-window} proves Theorem~\ref{thm:conv}(b).

\emph{Proofs of Lemma~\ref{lem:surv}(b), Theorem~\ref{thm:conv}(a) and Corollary~\ref{cor:t2}(c), with an outline of
the rest.} Lemma~\ref{lem:surv}(a): for a fixed profile the success satisfies the recursion with the maximum
replaced by the prescribed block size, so $s(b)\le G_{|b|}(n,k)$, with $|b|$ the number of blocks, by induction, and $G_r$ is nondecreasing in $r$; for a
general policy, conditionally on the seed and history the next draws are $\mathrm{Bernoulli}(j/M)$ products whose law
depends only on $(M,j)$, so induction on the remaining horizon with the tower property bounds the conditional success by
$G$ at the current state. Lemma~\ref{lem:surv}(b): if $\tau=\infty$ with positive probability there is nothing to prove;
otherwise, since every path executes at least one round (there are $n\ge2$ coordinates to commit),
$\E\tau=1+\sum_{s\ge1}\Pr(\tau>s)$ and
$\Pr(\tau>s)\ge\Pr(\text{valid output})-\Pr(\text{valid output},\,\tau\le s)\ge Q_\pi(V)-G_s(n,k)$: a valid output with
$\tau\le s$ means that the state reaches $(0,0)$ within $s$ rounds, which has probability at most $G_s(n,k)$ by
Lemma~\ref{lem:surv}(a), so the rest of the valid mass lies on $\{\tau>s\}$. Summing over $s\ge1$ gives the first form,
and keeping only the terms with $s\le t$, where $G_s\le G_t$, gives the second.
Theorem~\ref{thm:conv}(a): ($\ge$) $\KL\ge-\ln Q_\pi(V)$ by Lemma~\ref{lem:tools}(a), and $Q_\pi(V)\le G_r(n,k)$ by
Lemma~\ref{lem:surv}(a), since every $\pi\in\mathcal A_r$ is complete after $r$ rounds almost surely. ($\le$) The optimal
count-state rule commits, at each legal state, a block size that attains the maximum in the recursion for $G$ at that
state and the remaining horizon, so its success obeys the recursion with the maximum attained; applied in the order of
an independent uniform permutation of the coordinates, it therefore succeeds with probability $G_r(n,k)$, by induction on
the horizon, and it lies in $\mathcal A_r$: at a remaining horizon of one round only $b=M$ has positive value,
and after any nonfinal illegal transition the rule commits all remaining coordinates in the next round, so every path is complete
after $r$ rounds. The lower bounds hold for every policy complete within $r$ rounds almost surely, and the optimal rule is complete after $r$ rounds for every seed and every word (sequence of drawn values), including words of probability zero; so the stronger reading of the cap, on every seed and every word (\texttt{CapEvery} in Appendix~\ref{app:lean}), gives the
same optimum as the almost-sure cap of Section~\ref{sec:setup}. On $V$ the rule's output law is invariant under every permutation of the coordinates, because the order is
uniform and on legal paths the block sizes and draws depend only on counts; so Lemma~\ref{lem:tools}(e) gives
$\KL=-\ln G_r(n,k)$. Theorem~\ref{thm:conv}(b): the lower side is \eqref{eq:env} at $(n,k)$; the upper side is proved at
the end of this appendix. Corollary~\ref{cor:t2}(c): build the construction of Theorem~\ref{thm:sep}(b) for a failure probability $\delta_n\to0$
(slowly, so that its length stays $\log_2\ln n+o(\log\log n)$); it succeeds with probability at least $1-\delta_n$, and $\delta_n\le\delta$ ensures that the weight
$w=e^{-\varepsilon}/(1-\delta_n)$ is at most $1$; mix it with weight $w$ against the one-round schedule, a cheap schedule
that usually fails, and relabel. The ladder alone gives the mixture success at least $w(1-\delta_n)=e^{-\varepsilon}$, and
the mixture is relabelling-invariant, so $F=-\ln(\text{success})\le\varepsilon$ (equivalently, the pointwise mixture bound
of Appendix~\ref{app:hlow} gives $F\le-\ln(1-\delta_n)+\ln(1/w)=\varepsilon$); meanwhile the ladder's rounds are charged
with weight only $w\to e^{-\varepsilon}$, which gives the factor $e^{-\varepsilon}$ in the leading coefficient of the upper
bound; the matching lower bound, and hence the equality in (c), is (b), proved below. The cost is at most
$e^{-\varepsilon}(\lceil\log_2\log_2n\rceil+8K)/(1-\delta_n)+1$ for the ladder built with $K$. With $\ell=\log_2\log_2n$, the
branch $K=\lceil2\ln(2/\delta_n)\rceil$ (Hoeffding--Ville) at $\delta_n=16/\ell$ makes the $o(1)$ of order
$\log\ell/\ell$, that is $O(\log\log\log n/\log\log n)$, and the branch $K=\lceil1/\delta_n\rceil$ (Kolmogorov) at
$\delta_n\asymp\ell^{-1/2}$ makes it $O((\log\log n)^{-1/2})$. The second is the kernel-checked rate (Appendix~\ref{app:lean}):
$|\LF(n,\varepsilon)/\log_2\ln n-e^{-\varepsilon}|\le30/\sqrt{\log_2\ln n}$ whenever $\log_2\ln n\ge\max\{4,(1-e^{-\varepsilon})^{-2}\}$, and the
same for the least $\E\tau$. To see the constants, write $\nu=\log_2\ln n$ and $\chi=\sqrt \nu\ge2$, and take $K=\lceil \chi\rceil$ and
$\delta_n=1/K$. Then $\delta_n\le1/\chi\le1-e^{-\varepsilon}$, so $w\le1$; $\delta_n\le\frac12$, so $1/(1-\delta_n)\le1+2\delta_n$; $8K<n$, as $\ln n\ge16$;
and $\lceil\log_2\log_2n\rceil\le \nu+2$ and $K\le \chi+1$. The cost bound above is therefore at most
$e^{-\varepsilon}(1+2\delta_n)(\nu+8\chi+10)+1\le e^{-\varepsilon}\nu+(8\chi+10)+2(\chi+13)+1\le e^{-\varepsilon}\nu+30\chi$, using
$\delta_n(\nu+8\chi+10)\le \chi+13$ and $\chi\ge2$; and Corollary~\ref{cor:t2}(b) gives $\LF\ge e^{-\varepsilon}\nu-3.672\ge e^{-\varepsilon}\nu-30\chi$.
Dividing by $\nu$ gives the bound, and the same two estimates hold for the least $\E\tau$, since the mixture is a causal policy with
$\tau=|U|$.

\emph{Fixed schedules and the single-horizon bound} (Lemma~\ref{lem:surv}(a) for fixed schedules and the
single-horizon form of Lemma~\ref{lem:surv}(b) for their mixtures, by direct counting; Step~4 is also used for general
policies below, and Appendix~\ref{app:lean} lists the formal counterparts). Step 1: $F(\mu)\le\varepsilon$ implies $Q_\mu(V)\ge e^{-\varepsilon}$ by Lemma~\ref{lem:tools}(a). Step 2: $q_U(V)=s(b(U))=S(n,k;b(U))$ with
$S(M,j;b_1,\dots)=\sum_{j'}\Bin(b_1,j/M)(j-j')S(M-b_1,j';b_2,\dots)$ and $S(M,j;\varnothing)=\mathbf 1\{M=j=0\}$. Step 3:
$S\le G_{|U|}$ by induction, since $G$ maximizes over the block size. Step 4: $G_r\le G_{r+1}$, by induction on $r$: $G_0\le G_1$ pointwise (they agree at $M=0$, and $G_0=0$ for $M>0$), and the right side of the recursion is nondecreasing in $G_r$. Step 5: for each
component, $1+t(q_U(V)-G_t)\le|U|$ (if $|U|\le t$ the left side is at most $1$; otherwise at most $1+t$); average
over $\mu$. Step 6: take the infimum over feasible $\mu$ (the feasible set is nonempty: it contains the all-singleton schedule), so
$\LF(n,\varepsilon)\ge1+t(e^{-\varepsilon}-G_t(n,k))$ for every $t\ge1$, and hence, as $\LF\ge1$, $\LF(n,\varepsilon)\ge1+t(e^{-\varepsilon}-\Gamma_t(n))_+$
with $\Gamma_t$ of Appendix~\ref{app:env} (the $\kappa_r$ form of the envelope). Summing over all horizons instead
(Lemma~\ref{lem:surv}(b)) and using \eqref{eq:env} gives the explicit rate bound
$\LF(n,\varepsilon)\ge e^{-\varepsilon}\log_2\ln(k+1)-2.672$ of Corollary~\ref{cor:t2}(b), proved below.

\emph{Lemma~\ref{lem:surv}(a) for general policies.} Fix a horizon $r\ge1$ and let
$Y_t=G_{r-t}(M_t,j_t)\mathbf 1\{\text{legal}\}$ for $t\le r$. Conditionally on the seed and the history after $t$ rounds, the draws of round $t+1$
are independent $\mathrm{Bernoulli}(j_t/M_t)$ on a set chosen from that information, so
$\E[Y_{t+1}\mid\mathcal F_t]\le Y_t$ by the definition of $G$ (for an executed empty round, by $G_{r-t-1}\le G_{r-t}$, Step~4); the probability of succeeding within $r$ rounds is at most
$Y_0=G_r(n,k)$, whether or not the number of rounds is capped. An illegal state means that more than $k$ ones or more than $n-k$ zeros are already committed, so no completion of it, arbitrary or not, outputs a point of $V$. \qed

\emph{Proof of Corollary~\ref{cor:t2}.} By the bounded-prefactor envelope \eqref{eq:env} (Proposition~\ref{prop:env}, Appendix~\ref{app:env}) at $(M,j)=(n,k)$,
$G_r(n,k)\le\min\{1,A_r(k+1)^{-2^{-r}}\}$ with $A_r<36$. (a) By Theorem~\ref{thm:conv}(a), $\min_{\pi\in\mathcal A_r}\KL(p\|Q_\pi)\le\varepsilon$ if and only if
$G_r(n,k)\ge e^{-\varepsilon}$, which is the characterization. A policy in $\mathcal A_r$ with output KL at most
$\varepsilon$ has $\varepsilon\ge-\ln G_r(n,k)\ge2^{-r}\ln(k+1)-\ln A_r$, so
$2^{-r}\ln(k+1)<\varepsilon+\ln36$, which is the lower bound. A
uniformly relabelled fixed profile of length $L$ lies in $\mathcal A_L$, so $R_F^{\mathrm{adapt}}\le R_F^{\mathrm{PROF}}$, and the relabelled profile of Theorem~\ref{thm:sep}(b) gives the upper bound; since $\ln(k+1)\ge\frac12\ln n$, the lower bound is $\log_2\ln n-O_\varepsilon(1)$. (b) Lemma~\ref{lem:surv}(b) with $\theta=e^{-\varepsilon}$, which is at most $Q_\pi(V)$ by Lemma~\ref{lem:tools}(a),
gives the first inequality; a mixture is in $\mathcal A_\infty$ with $\tau=|U|$, which gives the bound for $\LF$. Let $\ell_k=\log_2\ln(k+1)$ (so $2^{\ell_k}=\ln(k+1)$) and $c_*:=\sum_{i\ge0}\min\{1,36e^{-2^i}\}<2.672$.
If $\ell_k<1$ (including $\ell_k<0$ at $n=2$), then $\E\tau\ge1>e^{-\varepsilon}\ell_k-c_*$. Otherwise
$m=\lfloor\ell_k\rfloor\ge1$; the envelope gives $G_t(n,k)\le\min\{1,36\exp(-2^{\ell_k-t})\}$ and $\ell_k-t\ge m-t$, so
$\sum_{t=1}^mG_t(n,k)\le c_*$, and $\E\tau\ge1+me^{-\varepsilon}-c_*\ge e^{-\varepsilon}\ell_k-c_*$ because
$m\ge\ell_k-1$. \qed

\emph{Total variation.} The randomized-side results transfer to total variation:
$\mathrm{TV}(p,Q)\ge p(V)-Q(V)=1-Q(V)$ for every law $Q$, with equality when $Q$ is relabelling-invariant. Hence the
success-threshold statements (Theorem~\ref{thm:sep}(b) and Corollary~\ref{cor:t2}) hold for
$\mathrm{TV}\le\delta$ with $\varepsilon=-\ln(1-\delta)$, that is, $1-\delta$ in place of $e^{-\varepsilon}$, and in Theorem~\ref{thm:conv}(a) the least TV at hard cap $r$ is
$1-G_r(n,k)$.\pend{}

\paragraph{Proof of Corollary~\ref{cor:hard}.} For a single fixed schedule, the lower bound follows from Theorem~\ref{thm:sep}(a) for point masses, and the upper bound from the fixed geometric schedule of
Theorem~\ref{thm:sep}(c) (for $\varepsilon>\ln2$, the schedule built for $\ln2$). For a relabelled profile, the upper bound follows from Theorem~\ref{thm:sep}(b), whose relabelled mixture has all components of equal length. For the lower bound, if a
relabelled profile $b$ of length $L$ has output KL at most $\varepsilon$, then $e^{-\varepsilon}\le s(b)\le G_L(n,k)<36(k+1)^{-2^{-L}}$ by Lemmas~\ref{lem:tools}(a),(c),(e) and~\ref{lem:surv}(a) and \eqref{eq:env}, so $L>\log_2\ln(k+1)-\log_2(\varepsilon+\ln36)$. \qed

\subsection{Uniform window in Theorem~\ref{thm:conv}(b)}\label{app:conv-window}
For $n=2k\ge2$ and $r\ge1$,
\[
2^{-r}\ln(k+1)-\ln36\;<\;-\ln G_r(n,k)\;\le\;2^{-r}\ln k+\ln48,
\]
and the upper inequality is witnessed by uniformly relabelling one explicit profile $b^\ast$ with at most $r$ blocks.
The profile is a square-root ladder whose last block commits all remaining coordinates at once. Before that block a
Chebyshev bound keeps the run legal with probability above $\frac12$; the block itself succeeds with probability at least
$2/(3\sqrt m)$, a central atom, where $m\le32\,n^{2^{1-L^\ast}}$ is its size after $L^\ast-1$ square-root rounds and $L^\ast\le r$ is the largest
admissible length (defined below). So the profile, although it ends with one block, loses at most a constant factor in success
against $G_r(n,k)$ (the factor $1728$ below), that is, at most an additive constant in $-\ln G_r$. As the right side of the upper
inequality is at least $\ln48$, this window cannot certify $-\ln G_r(n,k)\le\varepsilon$ when $\varepsilon<\ln48$; the upper
bound of Corollary~\ref{cor:t2}(a) uses instead the ladder of Theorem~\ref{thm:sep}(b), which stops its bulk rounds at
$8K_\delta$ unrevealed coordinates and commits these as singletons, which never fail; that ladder fails with probability at
most $\delta$, at the cost of at most $8K_\delta$ rounds.

\emph{Lower side.} The bounded-prefactor envelope \eqref{eq:env} (Appendix~\ref{app:env}) gives
$G_r(n,k)\le\min\{1,A_r(k+1)^{-2^{-r}}\}$ with $A_r=6^{2-2^{2-r}}<36$; taking logarithms gives the lower inequality.

\emph{The profile.} For $L\ge1$ put $M_0=n$ and $M_t=\lceil\sqrt{2^{L-t+1}M_{t-1}}\,\rceil$ for $1\le t\le L-1$, and let $b^{(L)}$ be
the profile with blocks $b_t=M_{t-1}-M_t$ for $t\le L-1$ and a final block $b_L=M_{L-1}$; $s(b^{(L)})$ is its terminal success
under uniform relabelling (Lemma~\ref{lem:tools}(c)). Call $L$ \emph{admissible} if $M_{t-1}\ge2^{L-t+1}+2$ for all
$1\le t\le L-1$ (vacuous for $L=1$). An admissible profile has every $b_t\ge1$: with $c=2^{L-t+1}$,
$M_t\le\sqrt{cM_{t-1}}+1<M_{t-1}$ whenever $M_{t-1}>(c+2+\sqrt{c^2+4c})/2$, and this threshold is below $c+2$. Conversely, at the first $t$ with $M_{t-1}\le c+1$, $\lceil\sqrt{cM_{t-1}}\,\rceil\ge M_{t-1}$, so $b_t\le0$;
hence $L$ is admissible exactly when its blocks $b_1,\dots,b_{L-1}$ are all nonempty, as in Section~\ref{sec:bal}. So for admissible $L$ the counts decrease from $M_0=n$,
while the rungs $M_0=8K<M_1<\cdots$ of Theorem~\ref{thm:sep}(b) are numbered upward and visited downward.

\emph{Stopped martingale, keeping the exit value.} Run the exact oracle on the relabelled profile, and let $T$ be the
first $t\le L-1$ at which the state is illegal ($j_t\notin[0,M_t]$). Put $P_t=j_t/M_t$ for $t<T$ and $P_t=P_T$ for
$t\ge T$: the first illegal value, which lies outside $[0,1]$, is kept and frozen. Given the committed history, the
number of ones drawn in round~$t$ has mean $b_tP_{t-1}$, so $(P_t)_{t\le L-1}$ is a martingale with $P_0=\frac12$, and
$\E[(P_t-P_{t-1})^2\mid\mathcal F_{t-1}]\le b_t/(4M_t^2)$ on $\{t\le T\}$ and $0$ afterwards. On $\{T\le L-1\}$ the kept
exit value gives $|P_{L-1}-\frac12|>\frac12$, so by Chebyshev's inequality
$\Pr(T\le L-1)\le4\,\mathrm{Var}(P_{L-1})\le\sum_{t\le L-1}b_t/M_t^2$. Since $M_t^2\ge2^{L-t+1}M_{t-1}\ge2^{L-t+1}b_t$, this is at
most $\sum_{u=2}^L2^{-u}<\frac12$.

\emph{Final block.} On $\{T>L-1\}$ the state $(m,j)$, with $m=M_{L-1}\ge1$, is legal. At a forced state ($j\in\{0,m\}$) the
final block is exact and succeeds with probability $1$. At an interior state, it succeeds with probability equal to the central atom
$\Bin(m,j/m)(j)\ge e^{-1/6}\sqrt{2/(\pi m)}>2/(3\sqrt m)$, by Robbins' form of Stirling's formula
\citep{robbins1955remark} and $j(m-j)\le m^2/4$: writing $t!=\sqrt{2\pi t}\,(t/e)^te^{\rho_t}$ with $0<\rho_t<1/(12t)$,
$\Bin(m,j/m)(j)=\binom mj\frac{j^j(m-j)^{m-j}}{m^m}=\sqrt{\frac{m}{2\pi j(m-j)}}\,e^{\rho_m-\rho_j-\rho_{m-j}}\ge\sqrt{\frac2{\pi m}}\,e^{-1/6}$, as $\rho_m>0$ and
$\rho_j,\rho_{m-j}<\frac1{12}$; and $e^{-1/6}\sqrt{2/\pi}=0.675\ldots>\frac23$. In both cases the success is at least $2/(3\sqrt m)$, so, with the martingale bound above,
$s(b^{(L)})\ge\frac12\cdot2/(3\sqrt{M_{L-1}})$.

\emph{Size of the last block.} Since $\lceil x\rceil\le2x$ for $x\ge1$, $\log_2M_t\le1+(L-t+1)/2+\frac12\log_2M_{t-1}$, and unrolling
gives $\log_2M_{L-1}\le\sum_{u\ge0}(2+u/2)2^{-u}+2^{1-L}\log_2n=5+2^{1-L}\log_2n$. So $M_{L-1}\le32\,n^{2^{1-L}}$ and
$s(b^{(L)})\ge n^{-2^{-L}}/(3\sqrt{32})$.

\emph{Lengths that are not admissible} (used in the Conclusion when $L^\ast<r$). If some $t\le L-1$ has
$M_{t-1}<2^{L-t+1}+2\le2^{L-t+2}$, then, since $\log_2M_{t-1}\ge2^{1-t}\log_2n$ (as $M_s\ge\sqrt{M_{s-1}}$ for $s<t$), $\log_2n<(L-t+2)2^{t-1}$; with $u=L-t\ge1$ the right side is $2^L(u+2)/2^{u+1}\le\frac342^L$, so
$n^{2^{-L}}<2^{3/4}$.

\emph{Conclusion.} Let $L^\ast\le r$ be the largest admissible length ($L=1$ always is) and $b^\ast=b^{(L^\ast)}$. If $L^\ast=r$,
$s(b^\ast)\ge(2k)^{-2^{-r}}/(3\sqrt{32})\ge k^{-2^{-r}}/24$, because $2^{-2^{-r}}\ge2^{-1/2}$. If $L^\ast<r$, then $L^\ast+1$ is not
admissible, so $n^{2^{-L^\ast}}<2^{3/2}$ and $s(b^\ast)\ge2^{-3/2}/(3\sqrt{32})=\frac1{48}\ge k^{-2^{-r}}/48$. In both cases
$G_r(n,k)\ge s(b^\ast)\ge k^{-2^{-r}}/48$, since $b^\ast$ has at most $r$ blocks and its uniform relabelling lies in $\mathcal A_r$; so
$-\ln G_r(n,k)\le2^{-r}\ln k+\ln48$. By Lemma~\ref{lem:tools}(e) the relabelled profile has $F=-\ln s(b^\ast)$, which witnesses the
upper inequality. Finally, $G_r(n,k)\le36(k+1)^{-2^{-r}}\le36\,k^{-2^{-r}}\le36\cdot48\,s(b^\ast)=1728\,s(b^\ast)$. \qed

The result is an absolute additive window, uniform in $r$ and $n$, so it gives a multiplicative asymptotic only while
$2^{-r}\ln k\to\infty$: at every hard cap, one uniformly relabelled fixed
profile is within a factor $1728$ of the best causal randomized adaptive policy in terminal success. It concerns output
KL under a hard cap in the balanced exact-oracle class $\mathcal A_r$; it says nothing about operational mean cost, $H$,
tilted targets or approximate heads.
\pend{}

\section{Rates on the full-support target}\label{app:thmm}

\pend{}

\emph{Roadmap.} The subsections follow Section~\ref{sec:fs}: one grid, many grids, and what stops faster growth. A single
relabelled profile can square the level in each round and no more: Theorem~\ref{thm:thmm} and Corollary~\ref{cor:thmm} give the lower
bound (gaps past squaring, \S\ref{app:thmm-one-lower}), and Theorem~\ref{thm:cf1} and Corollary~\ref{cor:cf1} the matching upper bound, whose hard step is
the local-limit estimate of Lemma~U6 (\S\ref{app:thmm-one-upper}). Proposition~\ref{prop:hfs} gives the rate of the schedule-averaged criterion
(\S\ref{app:thmm-hfs}). An auxiliary last-round bound for all policies, Proposition~L (\S\ref{app:thmm-lastround}), is used only in \S\ref{app:thmm-sep}, as one of the two lower bounds there that cover all policies on this target. For mixtures, Theorem~\ref{thm:mixf} gives the common lower bound and the fixed-cap divergence, and Remark~\ref{rem:mixf}
the limit of its test set (\S\ref{app:thmm-mix-lower}); Theorem~\ref{thm:mixu} gives the fourth-power construction, through the coefficient-one lift
of Lemma~Z6, and Corollary~\ref{cor:mixexact} the exact constant (\S\ref{app:thmm-mix-upper}). Theorem~\ref{thm:fsadapt} extends the mean-cost $F$ frontier of Corollary~\ref{cor:mixexact} to the expected
executed rounds of causal adaptive policies (\S\ref{app:thmm-adapt}), and Corollary~\ref{cor:hfs} records the separation (\S\ref{app:thmm-sep}).

\emph{Setting.} Throughout this appendix $\alpha=\beta=1$: $p$ is the Beta$(1,1)$--Bernoulli law of
Section~\ref{sec:fs}. Fix a profile $(b_1,\dots,b_R)$, all $b_k\ge1$, $\sum_kb_k=n$, used with a uniform
relabelling (one profile, relabelling drawn before sampling, independently of the sampling randomness): the schedule mixture $\mu$ is the law of $\sigma U$ for
a labelled schedule $U$ with this profile and a uniform permutation $\sigma$ of the coordinates. Write $F=F(\mu)$
(Section~\ref{sec:setup}) for its output KL; by T1 below (the identity of Section~\ref{sec:fs} on this
target), $F=\KL(p_W\|Q_W)$, a divergence between laws of the number of ones $W$. Let $m_k=b_1+\dots+b_k$ and
$c_k=2+m_k$ (so $c_0=2$, $c_R=n+2$). Sampler: $A_0=0$, $A_k=A_{k-1}+\Bin(b_k,P_{k-1})$, $P_k:=(1+A_k)/c_k$ (the
exact conditional marginal after the first $k$ rounds). Target: $A_k\sim\mathrm{Unif}\{0,\dots,m_k\}$, i.e.\ $P_k$
is uniform on $\{x/c_k:x=1,\dots,c_k-1\}$; terminally $W=A_R\sim\mathrm{Unif}\{0,\dots,n\}$ and
$P_R=(1+W)/(n+2)$. These target facts hold because $p$ is the law of the coordinates when $\theta\sim\mathrm{Unif}(0,1)$ is drawn first and then all coordinates independently Bernoulli$(\theta)$, which is equivalent to its description in Section~\ref{sec:fs}: so $p(x)=1/\big((n+1)\binom n{|x|}\big)$, the number of ones among any $m$ coordinates is uniform on $\{0,\dots,m\}$, and given the first $k$ rounds $\theta$ has the law $\mathrm{Beta}(1+A_k,1+m_k-A_k)$; hence $P_k$ is the exact conditional marginal, and the target's next block carries $\mathrm{BetaBin}(b_{k+1},c_kP_k,c_k(1-P_k))$ ones, so that $P_{k+1}$ has the law $T(P_k,\cdot)$ of Lemma L1 with $c=c_k$, $b=b_{k+1}$. Every string has positive mass, so, unlike on the balanced slice (Section~\ref{sec:bal}), no output is a support violation and $F$ is not $-\ln$ of a success probability: it compares two whole count laws. Subscripts $s$ and $t$ ($\Pr_s,\E_s,\mathrm{Var}_s,\mathrm{Law}_s$ and
$\Pr_t,\E_t,\mathrm{Law}_t$) refer to the sampler and to the target; $\mathcal F_k$ is the history through round
$k$; for laws $\nu,\nu'$ on $[0,1]$, $\nu\le_{\mathrm{cx}}\nu'$ (convex order) means
$\int\varphi\,d\nu\le\int\varphi\,d\nu'$ for every convex $\varphi$; $\mathrm{kl}(a\|b)$ is the binary KL
divergence. In the arguments for Theorems~\ref{thm:thmm} and~\ref{thm:cf1}, $r$, with or without an index, is the ratio of two
consecutive levels, such as $c_{k+1}/c_k$, not a cap on the number of rounds, and $\tilde r$ is defined where it is used;
elsewhere $r$ is a local index. In local formulas $p$ is also a Bernoulli parameter or the numerator of a fraction $p/q$;
otherwise it is the target law. In the argument for Theorem~\ref{thm:cf1}, $\pi$ with or without an index, tilde or
prime is a count law, but bare $\pi$ in factors such as $\sqrt{2\pi}$ is the circle constant; in the argument for
Theorem~\ref{thm:fsadapt}, $\pi$ is a policy. Like $r$ and $p$, the letters $N$, $\ell$, $D$, $h$, $T$ and $Q$ have more than one
meaning in this appendix; each meaning is fixed where the letter is introduced.

\emph{Prerequisites.} Two facts are used; they are restated here with proofs.

\emph{T1 (relabelling reduces $F$ to count KL).} The target $p$ is exchangeable and the oracle is exact. Relabelling a
profile by a uniform permutation $\sigma$ drawn before sampling gives the output $\bar q=\E_\sigma q_U\circ\sigma^{-1}$
(that is, $Q_\mu$). This law is exchangeable, so $\KL(p\|\bar q)=\KL(p_W\|Q_W)$.

\emph{Proof.} $p\circ\sigma=p$, so the oracle answers on relabelled positions are the relabelled answers. The average
over $\sigma$ is therefore $\mathfrak{S}_n$-invariant. For exchangeable $P$ and $Q$, $P(x)=P_W(|x|)/\binom{n}{|x|}$, and likewise
for $Q$; so $\ln(P(x)/Q(x))=\ln(P_W(|x|)/Q_W(|x|))$, and summing over the $\binom nw$ strings with $|x|=w$ gives
$\KL(P\|Q)=\KL(P_W\|Q_W)$. \qed

\emph{E1 (moment identities, used in Lemma L5 and in round~1).} Take a state with $t$ revealed coordinates, $a$ of
them ones, so that $c=2+t$ and $p=(1+a)/c$; commit a block of size $b$ under the sampler, and let
$P'=(cp+H)/(c+b)$ with $H\sim\Bin(b,p)$. Then $\E[P'\mid p]=p$, and $\E[P'(1-P')\mid p]=p(1-p)\big(1-b/(c+b)^2\big)$.
Hence, for a fixed profile, $\E[P_R(1-P_R)\mid\mathcal F_k]=P_k(1-P_k)\prod_{j>k}(1-b_j/c_j^2)$, and
$\mathrm{Var}(P_R\mid\mathcal F_k)=P_k(1-P_k)\big(1-\prod_{j>k}(1-b_j/c_j^2)\big)$.

\emph{Proof.} $\E H=bp$ and $\E H^2=bp(1-p)+b^2p^2$. Expand $\E[P']$ and $\E[P'^2]$; the tower property gives the
product form. \qed

\pend{}

\subsection{One relabelled profile: the lower bound}\label{app:thmm-one-lower}

\begin{theorem}[multi-level edge lower bound for relabelled fixed profiles]\label{thm:thmm}
There are absolute constants $K_0,K_1,K_2<\infty$ such that for every $k\in\{1,\dots,R-1\}$ with $c_k\ge4$ and
\[
r:=c_{k+1}/c_k\ge K_1\sqrt{1+\ln c_k},
\]
the output KL satisfies
\begin{align*}
F&\ge\frac{c_{k+1}}{K_2c_k^2}&&\text{if }r\le K_0(c_k-1),\\
F&\ge\frac18\ln\Big(\frac{c_{k+1}}{2eK_0\,c_k^2}\Big)&&\text{if }r>K_0(c_k-1)\quad\text{(gap set saturates; log growth).}
\end{align*}
Round~1 ($c_0=2$ has a single atom): $F\ge0.45\ln(0.45\,c_1/32)-0.45$ ($n\ge40$).
Rounds from $c_k=3$ (possible only for $k=1$, when $b_1=1$): $F\ge\frac1{12}\big(\ln(c_{k+1}/(3\cdot4813))-1\big)$.
\end{theorem}

In the first regime $c_{k+1}\le K_0c_k(c_k-1)$ already, and $F\le\varepsilon$ gives $c_{k+1}\le K_2\varepsilon c_k^2$; in the
second, $F\le\varepsilon$ gives $c_{k+1}\le2eK_0e^{8\varepsilon}c_k^2$.

\begin{corollary}[depth]\label{cor:thmm}
If $n\ge1$ and $F\le\varepsilon$ then $c_1\le72e^{1+2.3\varepsilon}$ (for $n\ge40$ by the round-1 bound; for $n<40$,
$c_1\le n+2\le41<72e$ trivially), $c_{k+1}\le3\cdot4813\cdot e^{1+12\varepsilon}$ when $c_k=3$, and
$c_{k+1}\le K_3(\varepsilon)c_k^2$ for every $k\ge1$ with $c_k\ge4$, where
$\ln K_3(\varepsilon)=O(1+\varepsilon)$; hence, for $n\ge2$,
\[
R\ge\log_2\ln n-O(1+\log(1+\varepsilon)).
\]
\end{corollary}
So single relabelled profiles need at least $\log_2\ln n-O(1)$ rounds at every fixed tolerance. The constants are
useless at practical $n$; the statement is a rate ($\Omega(\log\log n)$), not a finite bound.

\emph{Plan of the lower bound.} The proof of Theorem~\ref{thm:thmm} tests the terminal value $P_R$ on a gap set $G$: the middle halves of the first $X$ gaps between neighbouring atoms $x/c_k$ of $P_k$, counted from the edge $0$, which carry target mass at least $X/(4c_k)$. Given $P_k=x'/c_k$, the sampler's terminal value has conditional mean $P_k$ and conditional variance at most $2P_k/c_{k+1}$ (Lemma L5), so by Chebyshev it rarely reaches $G$ when $c_{k+1}/c_k$ is large; Lemma L4 bounds the sampler's mass on the first $y$ atoms by $4y/c_k$, a constant multiple of the target's, so these conditional bounds average to a sampler mass of $G$ that is a small fraction of its target mass, and data processing onto $\{P_R\in G\}$ bounds $F$ from below. Lemmas L1--L3 serve only to prove Lemma L4: one sampler round is below the target's round (the kernel $T$) in convex order (Lemma L1), the sampler kernel preserves convex order (Lemma L2), and hence every sampler marginal is below the target's (Lemma L3). Round~1 and rounds from $c_k=3$ are treated separately, and Corollary~\ref{cor:thmm} chains the resulting bounds $c_{k+1}\le K_3(\varepsilon)c_k^2$ over the rounds; it is the lower half of Theorem~\ref{thm:fs}(b).

\emph{Lemma L1 (per-state convex order).} For $p\in(0,1)$, $c>0$, $b\ge1$ let $S(p,\cdot)$ be the law of
$(cp+H)/(c+b)$ with $H\sim\Bin(b,p)$, and $T(p,\cdot)$ the same with $H\sim\mathrm{BetaBin}(b,cp,c(1-p))$. Then
$S(p,\cdot)\le_{\mathrm{cx}}T(p,\cdot)$. (At $p\in\{0,1\}$ both kernels are the point mass at $p$, so the claim is
trivial; Lemma L1 is applied only at interior grid points $p=x/c_k$.)

\emph{Proof.} $\mathrm{BetaBin}(b,cp,c(1-p))$ is the mixture of $\Bin(b,\theta)$ over
$\theta\sim\mathrm{Beta}(cp,c(1-p))$, $\E\theta=p$. For convex $\varphi$,
$\psi(\theta):=\sum_h\varphi\big(\frac{cp+h}{c+b}\big)\binom bh\theta^h(1-\theta)^{b-h}$ is convex in $\theta$:
\[
\psi''(\theta)=b(b-1)\sum_j\Delta_h^2\varphi\Big(\frac{cp+h}{c+b}\Big)\Big|_{h=j}\binom{b-2}{j}\theta^j(1-\theta)^{b-2-j}\ge0
\]
(apply twice the derivative formula
$\frac d{d\theta}\sum_hf(h)\binom bh\theta^h(1-\theta)^{b-h}=b\sum_j\big(f(j+1)-f(j)\big)\binom{b-1}j\theta^j(1-\theta)^{b-1-j}$
for Bernstein polynomials; second differences of a convex function of $h$ are $\ge0$). For $b=1$, $\psi$ is affine in $\theta$, so $\psi''=0$ and the sum is empty. Jensen:
$\E_T\varphi=\E_\theta\psi(\theta)\ge\psi(p)=\E_S\varphi$. \qed

\emph{Lemma L2 (the sampler kernel preserves convex order).} For fixed $c$, $b$ and convex $\varphi$ on $[0,1]$,
$f(p):=\E\varphi\big((cp+\Bin(b,p))/(c+b)\big)$ is convex on $[0,1]$.

\emph{Proof.} Write $a=c/(c+b)$, $a'=1/(c+b)$, $\varphi_h:=\varphi(ap+a' h)$ and
$B_h:=\binom bh p^h(1-p)^{b-h}$ (and $B^{(m)}_h$ the same with $m$ in place of $b$). Then
\[
f''=a^2\sum_h\varphi''(ap+a' h)B_h+2a\sum_h\varphi'(ap+a' h)B_h'+\sum_h\varphi_hB_h''.
\]
With $B_h'=b\big(B^{(b-1)}_{h-1}-B^{(b-1)}_h\big)$ and
$B_h''=b(b-1)\big(B^{(b-2)}_{h-2}-2B^{(b-2)}_{h-1}+B^{(b-2)}_h\big)$, where $B^{(m)}_h:=0$ for $m<0$ or $h\notin\{0,\dots,m\}$ (for
$b=1$ the $B''$ term vanishes, because $b(b-1)=0$):
\begin{align*}
\sum_h\varphi'(ap+a' h)B_h'&=b\sum_j\big[\varphi'(ap+a'(j+1))-\varphi'(ap+a' j)\big]B^{(b-1)}_j\\
&\ge0
\quad(\varphi'\text{ nondecreasing}),\\
\sum_h\varphi_hB_h''&=b(b-1)\sum_j\big[\varphi_{j+2}-2\varphi_{j+1}+\varphi_j\big]B^{(b-2)}_j\\
&\ge0
\quad(h\mapsto\varphi(ap+a' h)\text{ convex}).
\end{align*}
All three terms are $\ge0$. General convex $\varphi$ on $[0,1]$: a convex $\varphi$ is continuous on $(0,1)$ and may
jump up at the endpoints, so write $\varphi=\varphi_c+j_0\cdot\mathbf 1\{v=0\}+j_1\cdot\mathbf 1\{v=1\}$ with
$\varphi_c$ continuous convex and $j_0,j_1\ge0$. Approximate $\varphi_c$ uniformly by $C^2$ convex functions, for example its Bernstein polynomials $B_N\varphi_c$: they are
polynomials, convex by the second-derivative formula of L1's proof, and converge uniformly on $[0,1]$ ($f$ is a finite sum, so uniform convergence passes
to $f$). For the jump parts, $\E_{S(p)}\mathbf 1\{v=0\}=\mathbf 1\{p=0\}$
and $\E_{S(p)}\mathbf 1\{v=1\}=\mathbf 1\{p=1\}$ are convex on $[0,1]$. \qed
\emph{Lemma L3 (convex order of the marginals).} For every $k$, $\mathrm{Law}_s(P_k)\le_{\mathrm{cx}}
\mathrm{Law}_t(P_k)$.

\emph{Proof.} Let $S_{k+1}$ and $T_{k+1}$ be the kernels $p\mapsto S(p,\cdot)$ and $p\mapsto T(p,\cdot)$ of Lemma L1
with $c=c_k$, $b=b_{k+1}$. Induction:
\[
\mathrm{Law}_s(P_{k+1})=S_{k+1}\mathrm{Law}_s(P_k)\le_{\mathrm{cx}}S_{k+1}\mathrm{Law}_t(P_k)
\le_{\mathrm{cx}}T_{k+1}\mathrm{Law}_t(P_k)=\mathrm{Law}_t(P_{k+1}),
\]
the first $\le_{\mathrm{cx}}$ by Lemma L2 (with $c=c_k$, $b=b_{k+1}$ fixed since the profile is fixed) and the second
by Lemma L1 integrated. \qed

\emph{Lemma L4 (edge CDF).} $\Pr_s(P_k\le u)\le2uc_k/(c_k-1)\le4u$ for $u>0$. (At $u=0$ the probability is $0$,
since $P_k\ge1/c_k$.)

\emph{Proof.} Markov's inequality with the convex $\varphi(v)=(2u-v)_+$ and Lemma L3:
$\Pr_s(P_k\le u)\le\E_s\varphi(P_k)/u\le\E_t\varphi(P_k)/u$, and
$\E_t\varphi(P_k)=(c_k-1)^{-1}\sum_{x<2uc_k}(2u-x/c_k)\le(c_k-1)^{-1}\cdot(2uc_k)\cdot(2u)/2=2u^2c_k/(c_k-1)$. \qed

\emph{Lemma L5 (conditional spread).} For $0\le k<R$, $\mathrm{Var}_s(P_R\mid\mathcal F_k)\le2P_k/c_{k+1}$.

\emph{Proof.} By E1, $\E[P_R(1-P_R)\mid\mathcal F_k]=P_k(1-P_k)\Pi$ with $\Pi:=\prod_{j>k}(1-b_j/c_j^2)$, so
$\mathrm{Var}(P_R\mid\mathcal F_k)=P_k(1-P_k)(1-\Pi)\le P_k\sum_{j>k}b_j/c_j^2$; $b_{k+1}/c_{k+1}^2\le1/c_{k+1}$ and,
for $j\ge k+2$, $b_j/c_j^2\le b_j/(c_jc_{j-1})=1/c_{j-1}-1/c_j$, which telescopes to $\le1/c_{k+1}$. \qed

\emph{Proof of Theorem~\ref{thm:thmm}.} Fix $k\ge1$ with $c:=c_k\ge4$ and $r:=c_{k+1}/c\ge8$ (implied by $K_1\ge8$; the constants are fixed below: $K_0:=4608e^2$, $K_1:=2K_0$ and
$K_2:=8K_0/0.3$).
Take an integer $X$ with $1\le X\le\min(r/K_0,c-1)$ ($X=c-1$ is allowed: $G$ stays inside $(0,1)$ and case~(b) below is then
empty).
Gap set:
\[
G:=\bigcup_{x=1}^{X}\Big[\frac{x+1/4}{c},\frac{x+3/4}{c}\Big]\subset(0,1),
\]
and write $\Pr_s(G):=\Pr_s(P_R\in G)$, $\Pr_t(G):=\Pr_t(P_R\in G)$.
Every point of $G$ is at distance $\ge1/(4c)$
from every atom $x'/c$ ($x'=1,\dots,c-1$) of $P_k$.

\emph{Target mass.} Each interval has length $1/(2c)$ and contains $\ge(n+2)/(2c)-1$ values $(1+w)/(n+2)$ of $P_R$,
each of target mass $1/(n+1)$. Since $n+2\ge c_{k+1}=rc\ge8c$, each interval has mass $\ge1/(2c)-1/(8c-1)\ge1/(4c)$:
$\Pr_t(G)\ge X/(4c)$.

\emph{Sampler mass.} Condition on $P_k=x'/c$.
\begin{enumerate}
\item[(a)] $x'\le2X+1$: by Lemma L5 and Chebyshev, $\Pr_s(P_R\in G\mid P_k=x'/c)\le16c^2\cdot2x'/(c\cdot c_{k+1})
=32x'/r$.
\item[(b)] $x'\ge2X+2$: $G\subset[0,(X+1)/c]$ and $x'/c-(X+1)/c\ge x'/(2c)$, so the same probability is
$\le(4c^2/x'^2)\cdot2x'/(c\cdot c_{k+1})=8/(rx')$.
\end{enumerate}
With $Y:=cP_k\in\{1,\dots,c-1\}$ and Lemma L4 ($\Pr_s(Y\le y)\le4y/c$):
\begin{align*}
\E_s\big[32Y/r;\,Y\le2X+1\big]&\le(32/r)(2X+1)\cdot4(2X+1)/c\le1152X^2/(rc);\\
\E_s\big[8/(rY)\big]&\le\frac8r\Big[\frac1{c-1}+\sum_{y=1}^{c-2}\Big(\frac1y-\frac1{y+1}\Big)\frac{4y}{c}\Big]\\
&\le\frac8r\cdot\frac{4\ln c+2}{c}\qquad\text{(summation by parts).}
\end{align*}
So
\[
\rho_G:=\frac{\Pr_s(G)}{\Pr_t(G)}\le\frac{4608X}{r}+\frac{32(4\ln c+2)}{rX}.
\]
Put $K_0:=4608e^2$.

\emph{Case $r\le K_0(c-1)$:} $X:=\lfloor r/K_0\rfloor\le c-1$ (allowed). With $r\ge2K_0$, $X\ge r/(2K_0)$, so the second
term is $\le64K_0(4\ln c+2)/r^2\le e^{-2}$ when $r\ge8e\sqrt{K_0(4\ln c+2)}$. Then $\rho_G\le2e^{-2}$.

\emph{Case $r>K_0(c-1)$:} $X:=c-2\ge c/2$ (any $X\le c-1$ works; $c-2$ keeps the displayed constants). Then
$\rho_G\le4608c/r+64(4\ln c+2)/(rc)\le2\cdot4608c/r$ ($c\ge4\Rightarrow64(4\ln c+2)/c\le4608c$).

Data processing onto the event $\{P_R\in G\}$ and $\mathrm{kl}(a\|b)\ge a\ln(a/b)-a$ (valid for all $a,b\in(0,1)$
since $(1-a)\ln((1-a)/(1-b))\ge b-a$) give $F\ge\Pr_t(G)(\ln(1/\rho_G)-1)$:

\emph{Case 1:} $F\ge(1-\ln2)\cdot X/(4c)\ge0.3\cdot r/(8K_0c)=c_{k+1}/(K_2c^2)$, $K_2:=8K_0/0.3$.

\emph{Case 2:} $F\ge\frac{c-2}{4c}\big(\ln(r/(9216c))-1\big)\ge\frac18\ln\big(c_{k+1}/(9216e\cdot c^2)\big)$
($c\ge4$; the theorem states it with the larger constant $2eK_0\ge9216e$). If the logarithm is negative the right side is $<0\le F$ and the claim is
trivial.
\emph{Case $c=3$.} Take $X=1$, which is allowed because the $\rho_G$ display above used only $r\ge8$ and
$1\le X\le c-1$, never $X\le r/K_0$ or $c\ge4$. With $c=3$: (a) covers both atoms $x'\in\{1,2\}$ and (b) is empty. The
target mass is $\Pr_t(G)\ge X/(4c)=1/12$, and $\rho_G\le4608/r+32(4\ln3+2)/r=4812.6/r\le4813/r$. Hence
$F\ge\frac1{12}(\ln(r/4813)-1)$, and $F\le\varepsilon$ forces $c_{k+1}\le3\cdot4813\cdot e^{1+12\varepsilon}$. The
bound is positive only for $r>4813e\approx13{,}083$, where $r\ge8$ holds automatically; otherwise it is trivial.

Both Case-1 conditions on $r$ hold when $r\ge K_1\sqrt{1+\ln c}$ with $K_1:=2K_0$, because
$8e\sqrt{K_0(4\ln c+2)}\le8e\sqrt{5K_0}\sqrt{1+\ln c}\le2K_0\sqrt{1+\ln c}$ ($K_0\ge80e^2$).

\emph{Round 1 (complete calculation).} Here $c_0=2$, so $P_0=\frac12$, and $b_1=c_1-2$.
\begin{enumerate}
\item[(a)] E1 at $k=0$ ($\mathcal F_0$ trivial, $P_0=\frac12$) gives $\E_sP_R=\frac12$ and
$\mathrm{Var}_s(P_R)=\frac14\big(1-\prod_{j=1}^R(1-b_j/c_j^2)\big)$; since each $b_j/c_j^2\in[0,1]$,
$1-\prod_{j=1}^R(1-b_j/c_j^2)\le\sum_{j=1}^Rb_j/c_j^2$.
\item[(b)] $b_1/c_1^2\le1/c_1$, and for $2\le j\le R$, $b_j/c_j^2\le b_j/(c_jc_{j-1})=1/c_{j-1}-1/c_j$; so
$\sum_{j=1}^Rb_j/c_j^2\le2/c_1-1/c_R$.
\item[(c)] Hence $1-\prod_{j=1}^R(1-b_j/c_j^2)\le2/c_1$, and $\mathrm{Var}_s(P_R)\le1/(2c_1)$.
\item[(d)] Chebyshev gives $\beta_1:=\Pr_s(|P_R-\frac12|\ge\frac14)\le16\cdot\mathrm{Var}_s(P_R)\le8/c_1$.
\item[(e)] Target mass. $P_R=(1+W)/(n+2)$ with $W\sim\mathrm{Unif}\{0,\dots,n\}$. The event holds for
$(1+W)\le(n+2)/4$ or $(1+W)\ge3(n+2)/4$, so $\alpha_1:=\Pr_t(|P_R-\frac12|\ge\frac14)\ge((n+2)/2-2)/(n+1)
=(n-2)/(2(n+1))\ge0.45$ for $n\ge40$ (the exact minimum of $\alpha_1$ over $n\in[40,2000]$ is $10/21\approx0.4762$, at $n=41$).
\item[(f)] Data processing onto the event gives $F\ge\mathrm{kl}(\alpha_1\|\beta_1)\ge\alpha_1\ln(\alpha_1/\beta_1)
-\alpha_1$. Split on $c_1$:
\begin{itemize}
\item If $c_1\ge18$ then $\beta_1\le8/18<0.45\le\alpha_1$ ($n\ge40$). The map $a'\mapsto a'\ln(a'/\beta_1)-a'$ is
increasing on $[\beta_1,1]$, and the bound decreases in $\beta_1$. Hence
$F\ge0.45\ln(0.45\,c_1/8)-0.45\ge0.45\ln(0.45\,c_1/32)-0.45$ (the statement's form).
\item If $c_1<18$ the right side is negative, so the claim is trivial.
\end{itemize}
\end{enumerate}
Consequence: $c_1\le(8/0.45)\cdot e^{1+F/0.45}\le17.8\cdot e^{1+2.23\varepsilon}$, which implies the Corollary's
$c_1\le72e^{1+2.3\varepsilon}$. \qed

\emph{Proof of Corollary~\ref{cor:thmm}.} Let $\varepsilon\ge F$. Round 1 gives $\ln c_1\le O(1+\varepsilon)$. If
$c_1=3$ ($b_1=1$), the Case $c=3$ bound gives $\ln c_2\le O(1+\varepsilon)$, so the chain may start from whichever of
$c_1,c_2$ is $\ge4$ (without this step the profile $(1,n-1)$ would escape). For $k$ with $c_k\ge4$: either
$r<K_1\sqrt{1+\ln c_k}\le K_1c_k$, or Case 1 gives $c_{k+1}\le K_2\varepsilon c_k^2$, or Case 2 gives
$c_{k+1}\le2eK_0e^{8\varepsilon}c_k^2$. So $c_{k+1}\le K_3(\varepsilon)c_k^2$ with
\[
K_3=\max\big(K_1,\;K_2\varepsilon,\;2eK_0e^{8\varepsilon},\;(4813/3)\cdot e^{1+12\varepsilon}\big).
\]
The last entry covers the $c=3$ step exactly: $c_2\le3\cdot4813e^{1+12\varepsilon}=(4813/3)e^{1+12\varepsilon}\cdot
c_1^2$ with $c_1=3$, so the same doubling applies from $k=1$. With $u_k:=\ln c_k+\ln K_3$: $u_{k+1}\le2u_k$, so
$\ln(n+2)\le u_R\le2^{R-1}u_1$, i.e.\
$R\ge1+\log_2\ln(n+2)-\log_2(\ln c_1+\ln K_3)=\log_2\ln n-O(1+\log(1+\varepsilon))$. \qed

\subsection{An auxiliary last-round bound for general policies}\label{app:thmm-lastround}

\emph{The restricted class.} A policy reveals coordinates in rounds, as in Section~\ref{sec:setup} but without a cap. Before
each round it chooses a set of unrevealed coordinates, measurably in an exogenous seed (drawn once, before any
coordinate is sampled) and the values revealed so far; conditionally on that information, the round's coordinates are
drawn independently from their exact conditional marginals given all revealed values. Every path commits all $n$
coordinates, and there is no joint draw and no revision.

\emph{Proposition L (last-round edge bound, all policies).} Take any policy in the restricted class: labelled or
relabelled, value-adaptive, seeded; write $q$ for its output law. Let $\tilde m$ be the number of coordinates committed
before the round that commits the last coordinate, and $\tilde a$ the number of ones among them. For every $j\ge0$,
\[
\Pr\nolimits_q(W\le j)\le\E[\varphi_j(\tilde m)],\qquad\varphi_j(m):=\Pr\big(\Bin(n-m,1/(2+m))\le j\big),
\]
and hence $\KL(p\|q)\ge\mathrm{kl}\big(\Pr_t(W\le j)\,\big\|\,\E\varphi_j(\tilde m)\big)$ whenever
$\E\varphi_j(\tilde m)\le\Pr_t(W\le j)$.

\emph{Proof.} The final round is identifiable when it starts: it is the round whose block is all $n-\tilde m$ remaining
coordinates, chosen from the past. So $\tilde m$, and $\tilde a$ (the ones among the first $\tilde m$ committed coordinates), are determined by the past
at its start. Given that past, every unrevealed coordinate has the exact marginal
$(1+\tilde a)/(2+\tilde m)\ge1/(2+\tilde m)$, because the target is exchangeable. The final round draws
them independently, so its count stochastically dominates $\Bin(n-\tilde m,1/(2+\tilde m))$. Since $W$ is at least
that count, take expectations. Then apply data processing onto $\{W\le j\}$; $\mathrm{kl}(a\|\cdot)$ is decreasing on
$[0,a]$. \qed

It constrains only the last round.

\pend{}

\subsection{One relabelled profile: the matching upper bound}\label{app:thmm-one-upper}

The lower bound of Corollary~\ref{cor:thmm} is attained, up to an additive $O_\varepsilon(1)$, by
one explicit relabelled profile. Write $R_F^{\mathrm{PROF}}(n,\varepsilon)$, as in Corollary~\ref{cor:hard} but for this target, for the least number of rounds of a profile
$(b_1,\dots,b_R)$ (all $b_k\ge1$, $\sum_kb_k=n$, one uniform relabelling drawn before sampling, as in the Setting) whose
output KL satisfies $F\le\varepsilon$.

\begin{theorem}[single-profile upper bound]\label{thm:cf1}
For every $\varepsilon\in(0,1]$ and every $n\ge2$, the profile constructed below (Schedule) satisfies $F\le\varepsilon$
and has
\[
R\le\log_2\log_2n+8c_\varepsilon\ \text{rounds},\qquad c_\varepsilon:=\big\lceil4.680\cdot10^{20}\varepsilon^{-2}\big\rceil.
\]
\end{theorem}

\begin{corollary}[single-profile rate]\label{cor:cf1}
With Corollary~\ref{cor:thmm}, $R_F^{\mathrm{PROF}}(n,\varepsilon)=\log_2\log_2n+O_\varepsilon(1)$ for every fixed
$\varepsilon>0$.
\end{corollary}

The constant is astronomical, so Theorem~\ref{thm:cf1}, though an explicit finite bound, is meaningful only as an asymptotic rate, not as a practical finite-$n$ guarantee. It concerns the same class as
Theorem~\ref{thm:thmm} (one profile, one predrawn uniform relabelling, the exact oracle, forward KL, the number of
rounds, the target with $\alpha=\beta=1$). No matching bound is claimed for adaptive schedules or other $\alpha,\beta$.
For mixtures of profiles the mean-length quantity $\LF$ is matched: Theorem~\ref{thm:cf1} gives an upper bound (a
relabelled profile being one feasible design), Theorem~\ref{thm:mixf} below gives the order, and
Corollary~\ref{cor:mixexact} the exact constant, $\LF(n,\varepsilon)/\log_2\log_2n\to e^{-\varepsilon}/2$; so the
separation of Corollary~\ref{cor:hfs} below is two-sided.
An exact sampler exists outside the interface, in this sense: draw $\theta\sim\mathrm{Unif}(0,1)$, then all
coordinates independently Bernoulli$(\theta)$; for $n\ge2$ this one-shot draw is not a round of the interface, since
a first round makes the committed values independent Bernoulli$(\frac12)$, so two of them are both $1$ with
probability $\frac14$, against $\frac13$ for the one-shot sampler. The theorem therefore concerns the round complexity of
this interface, not a new sampler.

\emph{Plan of the upper bound.} The proof first bounds $F$ by a sum over rounds of one-round log-likelihood ratios ($d_k$ below), each averaged under a law pulled back from the terminal target through the posterior kernels of the later rounds (Lemma U1). It then replaces each pulled-back law by the target's count law at that round, at the cost of the variation of an envelope of $d_k$ times the later rounds' Kolmogorov distances (Lemmas U2--U4). What remains is one round at a time: an envelope with small mean and variation at most $4r+O(1)$ (Lemmas U5 and U6 and the per-round bound; the hard step is the local-limit lower bound of Lemma U6 in the bulk), and the Kolmogorov bound of Lemma U7, by which, since the schedule's levels grow geometrically, the later rounds' distances sum to order $1/(rc)$, so that the replacement costs order $1/c$. In the schedule each non-unit round goes from a level $c$ to a level $c'$ with $c'/c^2$ at most a design slot $\bar\kappa_j$ that depends only on $\varepsilon$ (at most $8\bar\kappa_J$ for a merged final round), and the slots and the warm-up level $c_\varepsilon$ are chosen so that the per-round terms add up to at most $\varepsilon$; Theorem~\ref{thm:cf1}, applied at $\min(\varepsilon,1)$ (proof of Corollary~\ref{cor:cf1}), is thus the upper half of Theorem~\ref{thm:fs}(b).

\emph{Notation for the upper bound.} Write $\pi_k:=\mathrm{Unif}\{0,\dots,m_k\}=\mathrm{Law}_t(A_k)$ and
$Q_k:=\mathrm{Law}_s(A_k)$, so that $F=\KL(\pi_R\|Q_R)$ by T1. Round $k$ uses the kernel
\[
\mathcal S_k(a,y):=\Pr\big(a+\Bin(b_k,(1+a)/c_{k-1})=y\big),\qquad a\in\{0,\dots,m_{k-1}\},
\]
so $Q_k=Q_{k-1}\mathcal S_k$. Put $\tilde\pi_k:=\pi_{k-1}\mathcal S_k$ (one sampler round applied to the target law),
$d_k:=\ln(\pi_k/\tilde\pi_k)$ and $\ell_k:=\ln(\pi_k/Q_k)$ on $\{0,\dots,m_k\}$, so $F=\E_{\pi_R}\ell_R$. The
posterior kernel is $\mathcal B_k(a\mid y):=\pi_{k-1}(a)\mathcal S_k(a,y)/\tilde\pi_k(y)$; for a function $f$ on level
$k-1$, $(\mathcal B_kf)(y):=\sum_a\mathcal B_k(a\mid y)f(a)$, and for a measure $\lambda$ on level $k$,
$(\lambda\mathcal B_k)(a):=\sum_y\lambda(y)\mathcal B_k(a\mid y)$. Write $F_\nu$ for the distribution function of a law
$\nu$ on the integers, $\mathrm{Kol}_k:=\max_y|F_{\pi_k}(y)-F_{\tilde\pi_k}(y)|$, and
$V(g):=\sum_y|g(y+1)-g(y)|$ for the variation of a function. For one round from level $c$ to $c'=c+b$ (so $r=c'/c$,
$b=(r-1)c$, and $N:=c'-1$ output values), put, for $0\le y\le N-1$,
\[
\chi(y):=\tfrac{c}{c'}\min(y+1,\,c'-1-y),\qquad \chi_y:=\big\lfloor\chi(y)+\tfrac12\big\rfloor,\qquad
\sigma(y)^2:=(r-1)\chi(y)\big(1-\chi(y)/c\big).
\]
What $\chi(y)$ and $\sigma(y)^2$ measure is said after the symmetry below.
In one-round statements the index $k$ is dropped: $\pi:=\mathrm{Unif}\{0,\dots,c-2\}$, $\pi':=\mathrm{Unif}\{0,\dots,N-1\}$,
$p_a:=(1+a)/c$, $\mathcal S(a,y):=\Pr(a+\Bin(b,p_a)=y)$, $\tilde\pi:=\pi\mathcal S$, $d:=\ln(\pi'/\tilde\pi)$ and
$\mathrm{Kol}:=\max_y|F_{\pi'}(y)-F_{\tilde\pi}(y)|$ (for round $k$: $\pi_{k-1},\pi_k,(1+a)/c_{k-1},\mathcal S_k,\tilde\pi_k,d_k,\mathrm{Kol}_k$).
The map $a\mapsto m_{k-1}-a$, $y\mapsto m_k-y$ leaves $\mathcal S_k$, $\pi_{k-1}$, $\pi_k$ and $\tilde\pi_k$ invariant,
so every bound below is stated on the lower half $y\le(c'-2)/2$ and transfers by symmetry. On the lower half $\chi(y)=(y+1)/r$, which is $1+a$ for the real number $a$ (non-integer in general, and negative when $y+1<r$) whose one-round mean $a+bp_a=r(1+a)-1$ equals $y$; for $y+1\ge r/2$, $\chi_y-1\ge0$ is the nearest integer state, and $\sigma(y)^2$ is the binomial variance $bp_a(1-p_a)$ of the round at $a=\chi(y)-1$. Lemma U5(ii) bounds $d(y)$ through the state $\chi_y-1$, Lemma U6 through a window of integer states around $\chi(y)-1$, and Lemma U5(i) through the edge state $a=0$.

\emph{Lemma U1 (pointwise Jensen chain).} For every $k$ and $y$, $\ell_k(y)\le d_k(y)+(\mathcal B_k\ell_{k-1})(y)$.
Hence, with $\lambda_R:=\pi_R$ and $\lambda_{k-1}:=\lambda_k\mathcal B_k$ (probability measures),
$F\le\sum_{k=1}^R\E_{\lambda_k}[d_k]$.

\emph{Proof.} $Q_k(y)/\tilde\pi_k(y)=\sum_a\mathcal B_k(a\mid y)e^{-\ell_{k-1}(a)}$, and $Q_{k-1}>0$ on
$\{0,\dots,m_{k-1}\}$ because every $(1+a)/c_{k-1}\in(0,1)$. Since $-\ln$ is convex,
$\ln(\tilde\pi_k/Q_k)(y)\le(\mathcal B_k\ell_{k-1})(y)$; add $d_k(y)$. Iterate down to $\ell_0=0$ and take
$\E_{\pi_R}$. \qed

\emph{Lemma U2 (monotone posteriors).} $\mathcal S_k$ is totally positive of order two: for $a<a'$ and $y<y'$,
$\mathcal S_k(a,y)\mathcal S_k(a',y')\ge\mathcal S_k(a,y')\mathcal S_k(a',y)$. Hence $\mathcal B_k(\cdot\mid y')$
dominates $\mathcal B_k(\cdot\mid y)$ stochastically for $y'>y$, and $V(\mathcal B_kf)\le V(f)$ for every $f$.

\emph{Proof.} Adjacent case: with $h=y-a$, $p=(1+a)/c_{k-1}$, $p'=(2+a)/c_{k-1}$ and all four terms positive,
\[
\frac{\mathcal S_k(a,y)\mathcal S_k(a+1,y+1)}{\mathcal S_k(a,y+1)\mathcal S_k(a+1,y)}
=\frac{\binom{b}{h}^2}{\binom{b}{h-1}\binom{b}{h+1}}\cdot\frac{p'(1-p)}{p(1-p')}>1
\]
by log-concavity of binomial coefficients and $p'>p$; if a right-hand term vanishes ($h\notin\{1,\dots,b-1\}$) the
inequality is trivial. The support $\{0\le y-a\le b\}$ is a band, so the adjacent inequalities chain to all
$a<a'$, $y<y'$: if $\mathcal S_k(a,y')\mathcal S_k(a',y)=0$ there is nothing to prove; otherwise $y-a'\ge0$ and $y'-a\le b$,
so every entry of the rectangle $\{a,\dots,a'\}\times\{y,\dots,y'\}$ is positive, and the logarithm of the claimed ratio is
the sum of the logarithms of the adjacent ratios over the rectangle's unit squares. The prior factor $\pi_{k-1}(a)$ does not affect total positivity, so the posteriors increase in
likelihood ratio, hence stochastically. Write $f=f^+-f^-$ with $f^\pm$ non-decreasing and $V(f)=\mathrm{range}(f^+)+
\mathrm{range}(f^-)$; $\mathcal B_k$ maps non-decreasing functions to non-decreasing ones and does not enlarge
ranges. \qed

\emph{Lemma U3 (pull-back error).} $\tilde\pi_j\mathcal B_j=\pi_{j-1}$ (Bayes), hence
$\lambda_k-\pi_k=\sum_{j>k}(\pi_j-\tilde\pi_j)\mathcal B_j\mathcal B_{j-1}\cdots\mathcal B_{k+1}$. For every $g$ on level
$j$, $|\sum_y(\pi_j-\tilde\pi_j)(y)g(y)|\le\mathrm{Kol}_j\,V(g)$.

\emph{Proof.} $(\tilde\pi_j\mathcal B_j)(a)=\sum_y\pi_{j-1}(a)\mathcal S_j(a,y)=\pi_{j-1}(a)$; telescope
$\lambda_{k}-\pi_{k}=(\lambda_{k+1}-\pi_{k+1})\mathcal B_{k+1}+(\pi_{k+1}-\tilde\pi_{k+1})\mathcal B_{k+1}$ from
$\lambda_R=\pi_R$. For the second claim sum by parts: with $\Delta:=F_{\pi_j}-F_{\tilde\pi_j}$ ($\Delta(m_j)=0$),
$\sum_y(\pi_j-\tilde\pi_j)(y)g(y)=-\sum_{y<m_j}\Delta(y)\,(g(y+1)-g(y))$. \qed

\emph{Lemma U4 (envelope bound).} For every family of functions $\mathrm{env}_k\ge d_k$ on level $k$,
\[
F\le\sum_{k=1}^R\Big[\E_{\pi_k}\mathrm{env}_k+V(\mathrm{env}_k)\sum_{j>k}\mathrm{Kol}_j\Big].
\]

\emph{Proof.} By Lemma U1, $F\le\sum_k\E_{\lambda_k}d_k\le\sum_k\E_{\lambda_k}\mathrm{env}_k$. Summation by parts gives
$\E_{\lambda_k}\mathrm{env}_k-\E_{\pi_k}\mathrm{env}_k\le V(\mathrm{env}_k)\max_y|F_{\lambda_k}(y)-F_{\pi_k}(y)|$. By Lemma
U3, $F_{\lambda_k}(y)-F_{\pi_k}(y)=\sum_{j>k}\sum_z(\pi_j-\tilde\pi_j)(z)\,g_{j,y}(z)$ with
$g_{j,y}:=\mathcal B_j\cdots\mathcal B_{k+1}\mathbf 1_{[0,y]}$, which by Lemma U2 is monotone with values in $[0,1]$, so
$V(g_{j,y})\le1$ and each term is at most $\mathrm{Kol}_j$. \qed

\emph{Unit rounds.} If $b_k=1$ then $\Bin(1,(1+a)/c_{k-1})$ is the exact conditional of the target, so
$\tilde\pi_k=\pi_k$, $d_k\equiv0$ and $\mathrm{Kol}_k=0$; such rounds contribute nothing (take $\mathrm{env}_k\equiv0$).

\emph{Lemma U5 (edge pointwise bounds).} Fix one round with $r\ge2$ and $c\ge8r+4$, and $y\le(c'-2)/2$. Then
(i) $d(y)\le2r$ if $y+1<r/2$; (ii) $d(y)\le r/\chi_y+\frac12\ln(16\chi_y/r)$ if $y+1\ge r/2$ and $\chi_y\le c/2$.

\emph{Proof.} Tools: (T-a) $\Bin(b,p)(h)\ge e^{-b\,\mathrm{kl}(h/b\|p)}/\sqrt{8h(1-h/b)}$ for $1\le h\le b-1$, and
$\Bin(b,p)(0)=(1-p)^b\ge e^{-bp/(1-p)}$; (T-b) $\mathrm{kl}(q\|p)\le(q-p)^2/(p(1-p))$. For the first bound in (T-a), put
$u:=h/b$. Then $\Bin(b,p)(h)=\Bin(b,u)(h)\,e^{-b\,\mathrm{kl}(u\|p)}$, and Robbins' bounds \citep{robbins1955remark},
$\sqrt{2\pi m}\,(m/e)^m\le m!\le\sqrt{2\pi m}\,(m/e)^me^{1/(12m)}$ for $m\ge1$, give
$\Bin(b,u)(h)\ge e^{-1/(12h)-1/(12(b-h))}/\sqrt{2\pi h(1-u)}$. For $b\ge4$ the exponent is at least $-1/9$, and
$e^{-1/9}/\sqrt{2\pi}>1/\sqrt8$; for $b\le3$ directly, $\Bin(2,\frac12)(1)=\frac12$ and
$\Bin(3,\frac13)(1)=\Bin(3,\frac23)(2)=\frac49>\sqrt3/4$. The second bound in (T-a) holds because
$\ln(1-p)\ge-p/(1-p)$, and (T-b) because KL is at most $\chi^2$. For every
state $a$, $\tilde\pi(y)\ge\pi(a)\mathcal S(a,y)$, so $d(y)\le-\ln\tilde r-\ln\mathcal S(a,y)$ with
$\tilde r:=(c'-1)/(c-1)\ge r$. The mean of $a+\Bin(b,(1+a)/c)$ is exactly $r(1+a)-1$.
(i) Take $a=0$: $h=y$, $bp=r-1$, $|h-bp|\le r-1$ and $bp(1-p)\ge(r-1)/2$. For $h\ge1$,
$-\ln\mathcal S(0,y)\le2(r-1)+\frac12\ln(4r)$; for $h=0$ (the point $y=0$), $-\ln\mathcal S(0,0)=-b\ln(1-1/c)\le
(r-1)c/(c-1)\le2(r-1)$. Hence $d(y)\le-\ln r+2(r-1)+\frac12\ln(4r)\le2r$.
(ii) Take $a=\chi_y-1\le c/2-1$: then $\delta:=h-bp_a=y+1-r\chi_y$ satisfies $|\delta|\le r/2$,
$bp_a(1-p_a)=(r-1)\chi_y(1-\chi_y/c)\ge(r-1)\chi_y/2$ and $h\le(r-1)\chi_y+r/2\le2r\chi_y$, and $h\le b-1$ since
$\chi_y\le c/2$ and $b\ge c\ge8r+4$. If $h\ge1$, (T-a) and (T-b) give $-\ln\mathcal S(a,y)\le
r^2/(2(r-1)\chi_y)+\frac12\ln(16r\chi_y)\le r/\chi_y+\frac12\ln(16r\chi_y)$, and $-\ln\tilde r\le-\ln r$ gives the
claim. In (ii) the only point with $h=0$ is $r=2$, $y=0$ ($\chi_y=1$, $a=0$), where $d(0)\le2-\ln2$, below the claimed
$2+\frac12\ln8$. \qed

\emph{Lemma U6 (bulk local limit, lower bound).} Fix one round with $c\ge64$ and $8\le r\le c/4$, and let
$y\le(c'-2)/2$ with $\chi(y)\ge4r$. Then
\[
\frac{\tilde\pi(y)}{\pi'(y)}\ge1-C_\star\Big(\frac r{\sigma(y)}+\frac1c\Big),\qquad C_\star:=1{,}600{,}100,
\]
where $\pi'$ is the uniform law on $\{0,\dots,N-1\}$ and $\tilde\pi$ is the one-round law. Only this lower bound is
used, in the third row of the per-round envelope below.

\emph{Proof.} Write $\chi:=\chi(y)=(y+1)/r$ and $\sigma:=\sigma(y)$. On the lower half $\chi\le c/2$, so
$(r-1)\chi/2\le\sigma^2\le(r-1)\chi$; also $\chi\ge4r\ge32$. If $\sigma<8r$, the right side is below $1-C_\star/8<0$; so
assume $\sigma\ge8r$, whence $\sigma\ge64$ and $1/\sigma\le r/(8\sigma)$. Since
$\tilde\pi(y)/\pi'(y)=\frac{c'-1}{c-1}\sum_{a=0}^{c-2}\mathcal S(a,y)$ and $\frac{c'-1}{c-1}\ge r$, it suffices to show
$r\sum_{a\in\mathcal W}\mathcal S(a,y)\ge1-72\,r/\sigma$, where $\mathcal W$ is the set of integers $a$ with $|g_a|\le\chi/4$ and
$g_a:=a-a_0$ with $a_0:=\chi-1$; as $72\le C_\star$, this gives the lemma. (The proof needs only $C_\star\ge72$, and not the term
$C_\star/c$; the larger stated $C_\star$ is the one with which the side conditions and constants of the per-round bound and of
the schedule below are computed.)

\emph{Pointwise bound.} Fix $a\in\mathcal W$, write $g:=g_a$, and let $h:=y-a$ be the number of ones drawn in the round, as in
Lemma U5. Then $bp_a=(r-1)(\chi+g)$ and $b(1-p_a)=(r-1)(c-\chi-g)$ are both at least $\frac34(r-1)\chi$, and
$h-bp_a=-rg$ with $r|g|\le\frac38(r-1)\chi$. So $\mathcal W\subseteq\{0,\dots,c-2\}$; and $h=(r-1)\chi-g$ and
$b-h=(r-1)(c-\chi)+g$ are at least $(r-\frac54)\chi\ge\sigma^2/6$. Robbins' bounds
$\sqrt{2\pi m}\,(m/e)^m\le m!\le\sqrt{2\pi m}\,(m/e)^me^{1/(12m)}$ ($m\ge1$) give
\[
\ln\mathcal S(a,y)\ge-\Big[h\ln\frac h{bp_a}+(b-h)\ln\frac{b-h}{b(1-p_a)}\Big]-\frac12\ln\frac{2\pi h(b-h)}b
-\frac1{12h}-\frac1{12(b-h)}.
\]
The two terms in the bracket are $bp_a(1+u)\ln(1+u)$ and $b(1-p_a)(1+u')\ln(1+u')$ with $u=-rg/(bp_a)$ and
$u'=rg/(b(1-p_a))$, both in $[-\frac12,\frac12]$; use $(1+u)\ln(1+u)\le u+u^2/2+3|u|^3$ there. The linear terms cancel.
The quadratic terms sum to $r^2g^2/(2v_g)$, where $v_g:=bp_a(1-p_a)$; since $v_g-\sigma^2=(r-1)g\,(1-(2\chi+g)/c)$, we
have $|v_g-\sigma^2|\le(r-1)|g|\le\sigma^2/2$ and so $1/v_g\le1/\sigma^2+2(r-1)|g|/\sigma^4$. The cubic terms are at most
$6r^3|g|^3/(\frac34(r-1)\chi)^2\le\frac{32}3r^3|g|^3/\sigma^4$. So the bracket is at most
$r^2g^2/(2\sigma^2)+12r^3|g|^3/\sigma^4$. Next, $h(b-h)/b=v_g+rg(2p_a-1)-r^2g^2/b\le\sigma^2+2r|g|$, so the middle term is
at least $-\frac12\ln(2\pi\sigma^2)-r|g|/\sigma^2$; and the last two terms are at least $-1/\sigma^2$. With $e^{-x}\ge1-x$,
\[
\mathcal S(a,y)\ge\frac{e^{-r^2g^2/(2\sigma^2)}}{\sqrt{2\pi}\,\sigma}\big(1-\mathcal E(g)\big),\qquad
\mathcal E(g):=\frac{12r^3|g|^3}{\sigma^4}+\frac{r|g|}{\sigma^2}+\frac1{\sigma^2}.
\]

\emph{Summation.} Put $\psi:=r^2/(2\sigma^2)$ and $w:=r|g|/\sigma$, so that $e^{-\psi g^2}=e^{-w^2/2}$ and
$\mathcal E(g)\le(12w^3+w+1)/\sigma$. As $w^3e^{-w^2/2}\le32e^{-w^2/4}$ and $we^{-w^2/2}\le e^{-w^2/4}$,
$e^{-\psi g^2}\mathcal E(g)\le\frac{386}\sigma e^{-\psi g^2/2}$. For $f(t)=e^{-\psi'(t-a_0)^2}$ with $\psi'>0$, comparing each
monotone piece of $f$ with its integral gives $\sum_{a\in\mathbb Z\cap I}f(a)\ge\int_If-1$ for every interval $I$ and
$\sum_{a\in\mathbb Z}f(a)\le\int_{\mathbb R}f+1$; also $\int_{|t-a_0|\ge\chi/4}f\le e^{-\psi'\chi^2/32}\sqrt{2\pi/\psi'}$. Apply
these with $\psi'\in\{\psi,\psi/2\}$. As
$\sqrt{\pi/\psi}=\sqrt{2\pi}\,\sigma/r$,
\begin{align*}
r\sum_{a\in\mathcal W}\mathcal S(a,y)&\ge\frac r{\sqrt{2\pi}\,\sigma}\Big[\sum_{a\in\mathcal W}e^{-\psi g_a^2}
-\frac{386}\sigma\sum_{a\in\mathcal W}e^{-\psi g_a^2/2}\Big]\\
&\ge1-\sqrt2\,e^{-\psi\chi^2/32}-\frac r{\sqrt{2\pi}\,\sigma}-\frac{386}\sigma\Big(\sqrt2+\frac r{\sqrt{2\pi}\,\sigma}\Big).
\end{align*}
Finally $\psi\chi^2/32=r^2\chi^2/(64\sigma^2)\ge r\chi/64\ge\sigma$, because $\sigma^2\le r\chi$ and $\sigma\ge64$; so
$e^{-\psi\chi^2/32}\le e^{-\sigma}\le1/\sigma$. With $r/\sigma\le\frac18$ and $1/\sigma\le r/(8\sigma)$, the three subtracted
terms total at most $(0.18+0.4+70.7)\,r/\sigma<72\,r/\sigma$. \qed

\emph{Lemma U7 (one-round Kolmogorov distance).} For one round from $c\ge2$ to $c'=c+b$, $b\ge1$,
\[
(c-1)\max_y|F_{\pi'}(y)-F_{\tilde\pi}(y)|\le1-\frac b{c'(c'-1)}<1.
\]

\emph{Proof.} At a fixed $y$, $(c-1)F_{\tilde\pi}(y)$ is a sum of binomial distribution functions evaluated at the grid parameters $j/c$, while $(c-1)F_{\pi'}(y)$ is linear in $y$. The proof writes each grid value as an integral of polynomials in the parameter $t$ over the interval $[(j-1)/c,j/c]$ just below it, by the fundamental theorem of calculus, and sums over the intervals; the scaled difference (written $D(y)$ below) is then an explicit term minus a nonnegative term that is at most $\rho\le1$. Write $G_K(t):=\Pr(\Bin(b,t)\le K)$ and $q_K(t):=\Pr(\Bin(b-1,t)=K)$, polynomials in $t$ that vanish
identically for $K<0$. From $\frac d{dt}\Pr(\Bin(b,t)=k)=b(q_{k-1}-q_k)$, telescoping gives $G_K'=-b\,q_K$. For real $\bar y$
let $Q_K(t):=\sum_{k\le K}\Pr(\Bin(b,t)=k)(\bar y-ct-k)$; the same derivative and
$G_K=\Pr(\Bin(b-1,t)\le K-1)+(1-t)q_K$ give $-Q_K'=c'G_K-b\big((c-1)t-(\bar y-K-1)\big)q_K$. Both distribution functions
equal $1$ from $y=N-1$ on, so fix $0\le y\le N-2$ and put $\bar y:=y+1$. For $0\le i\le c-1$ let $I_i:=[i/c,(i+1)/c]$ and
write $G^{(i)},q^{(i)},Q^{(i)}$ for $G_K,q_K,Q_K$ at $K=y-i$, so that $\bar y-K-1=i$. With $j=a+1$,
$(c-1)F_{\tilde\pi}(y)=\sum_{j=1}^{c-1}G^{(j-1)}(j/c)$, while $(c-1)F_{\pi'}(y)=(c-1)\bar y/N$. On each $I_i$ the fundamental
theorem of calculus gives
\[
Q^{(i)}\big(\tfrac ic\big)-Q^{(i)}\big(\tfrac{i+1}c\big)=\int_{I_i}\big[c'G^{(i)}-b((c-1)t-i)q^{(i)}\big],\qquad
G^{(i)}\big(\tfrac{i+1}c\big)=\int_{I_i}\big[cG^{(i)}-b(ct-i)q^{(i)}\big],
\]
the second by differentiating $(ct-i)G^{(i)}$, and $G^{(i)}(\frac ic)-G^{(i)}(\frac{i+1}c)=\int_{I_i}b\,q^{(i)}$. Sum over
$i$. The first identities telescope to $\bar y$: $Q^{(0)}(0)=\bar y$, $Q^{(c-1)}(1)=0$ because $\bar y-c<b$, and $Q^{(i)}$ and
$Q^{(i+1)}$ agree at $(i+1)/c$, where the extra term $k=y-i$ of $Q^{(i)}$ has the factor $0$. The second sum to
$(c-1)F_{\tilde\pi}(y)$, since $G^{(c-1)}(1)=0$. The third sum to $\rho:=\sum_i\int_{I_i}b\,q^{(i)}=1-\sum_{j=1}^{c-1}
\Pr(\Bin(b,j/c)=\bar y-j)$, since $G^{(0)}(0)=1$, so $0\le\rho\le1$. With $\Sigma_G:=\sum_i\int_{I_i}G^{(i)}$, the first sums give
$c'\Sigma_G=\bar y+\sum_i\int_{I_i}b((c-1)t-i)q^{(i)}$ and the second $c\Sigma_G=(c-1)F_{\tilde\pi}(y)+\sum_i\int_{I_i}b(ct-i)q^{(i)}$.
Eliminating $\Sigma_G$, and using $c((c-1)t-i)-c'(ct-i)=-((b+1)ct-bi)$ and $\frac c{c'}-\frac{c-1}N=\frac b{c'N}$,
\[
D(y):=(c-1)\big(F_{\tilde\pi}(y)-F_{\pi'}(y)\big)=\frac{\bar yb}{c'N}
-\frac1{c'}\sum_{i=0}^{c-1}\int_{I_i}\big((b+1)ct-bi\big)\,b\,q^{(i)}(t)\,dt.
\]
On $I_i$ the weight $(b+1)ct-bi=b(ct-i)+ct$ lies in $[0,c']$, and $q^{(i)}\ge0$, so $\bar yb/(c'N)-\rho\le D(y)\le\bar yb/(c'N)$.
As $1\le\bar y\le N$ and $0\le\rho\le1$, $b/(c'N)-1\le D(y)\le b/c'=1-c/c'$, and $c/c'\ge b/(c'N)$ because $c(c'-1)\ge b$.
Hence $|D(y)|\le1-b/(c'N)$; for $y<0$ and $y\ge N-1$, $D(y)=0$. \qed

\emph{Per-round bound.} Consider a non-unit round with $8\le r\le c/4$, $c\ge\max(64,8C_\star,8r+4)$ and
$\kappa:=c'/c^2\le1/(128C_\star^2)$. Put $\Xi:=64C_\star^2$ (so $\Xi r\le c/2$) and $L_\Xi:=\frac12\ln(16\Xi)+\frac14=
18.0013\ldots$. On the lower half let
\[
\mathrm{env}(y):=\begin{cases}\max(2r,\,r+L_\Xi),&y+1<r/2,\\ r/\chi_y+L_\Xi,&y+1\ge r/2\text{ and }\chi(y)<\Xi r,\\
4C_\star\big(\sqrt{r/\chi(y)}+1/c\big),&\chi(y)\ge\Xi r,\end{cases}
\]
and extend it by symmetry. It dominates $d$: the first row by Lemma U5(i); the second by Lemma U5(ii), because
$\chi(y)<\Xi r\le c/2$ gives $\chi_y=\lfloor\chi(y)+\frac12\rfloor\le\lfloor c/2\rfloor\le c/2$ and
$\frac12\ln(16\chi_y/r)\le L_\Xi$; the third by Lemma U6, because there $x:=2C_\star(\sqrt{r/\chi(y)}+1/c)\le\frac12$
and $d\le-\ln(1-x)\le2x$ (using $r/\sigma\le2\sqrt{r/\chi}$, as $\sigma^2\ge(r-1)\chi/2\ge r\chi/4$). It is non-increasing from each end towards the middle
(each row is non-increasing, the first row is $\ge r+L_\Xi\ge$ the second, and the second row is
$\ge L_\Xi>1\ge\frac12+4C_\star/c$, the largest value of the third), so
$V(\mathrm{env})\le2\max\mathrm{env}\le4r+\ln(16\Xi)+1$.
For the mean, sum over the $N=rc-1$ output values by halves ($\mathrm{env}\ge0$ is symmetric) and index the lower half
by $t:=y+1\le T:=\lfloor(N-1)/2\rfloor+1\le rc/2$, so that $\chi(y)=t/r$. The first row has fewer than $r/2$ values of $t$;
in the second, $\chi_y$ takes the values $1,\dots,\chi_{\max}$, $\chi_{\max}:=\lfloor\Xi r+\frac12\rfloor$, each for at most
$r+1$ values of $t$; in the third, $\sqrt{r/\chi(y)}=r/\sqrt t$, and $\sum_{t\le T}t^{-1/2}\le2\sqrt T$. Hence
\[
\sum_y\mathrm{env}(y)\le2\Big[\frac r2\max(2r,r+L_\Xi)+(r+1)\big(r(1+\ln\chi_{\max})+L_\Xi\chi_{\max}\big)
+4C_\star\big(2r\sqrt T+T/c\big)\Big].
\]
Divide by $N=c'-1\ge0.9999\,rc$ (as $rc\ge64C_\star$) and use $\kappa=r/c$, $(r+1)/r\le1.125$, $\max(2r,r+L_\Xi)\le2r+L_\Xi-8$, $\chi_{\max}\le\Xi r+\frac12$,
$\ln\chi_{\max}\le\ln\Xi+\ln r+1$ and $2L_\Xi=\ln(16\Xi)+\frac12$: the first row contributes at most
$1.001\kappa(2+(L_\Xi-8)/8)$, the second at most $1.127\kappa(4+2\ln\Xi+2\ln r+\Xi(\ln(16\Xi)+\frac12)+L_\Xi/8)$, and the
third at most $11.33C_\star\sqrt\kappa+4.01C_\star/c$. Since $1.001(2+(L_\Xi-8)/8)+1.127L_\Xi/8<5.8<1.13(\frac14\ln(16\Xi)+5)$
and $1.127\le1.13$, the first two rows together contribute at most $1.13\kappa(2\ln r+A_\Xi)$, with $A_\Xi$ as below, and
$11.33\le1.13\cdot11.32$ and $4.01\le1.13\cdot4$ bound the third, so
\begin{gather*}
\E_{\pi'}\mathrm{env}\le1.13\big[\kappa(2\ln r+A_\Xi)+11.32C_\star\sqrt\kappa+4C_\star/c\big],\\
A_\Xi:=4+2\ln\Xi+\Xi(\ln(16\Xi)+\tfrac12)+\tfrac14\ln(16\Xi)+5.
\end{gather*}

\emph{Schedule.} Put $\varepsilon':=\varepsilon/16$ and let $c_\varepsilon$ be as in Theorem~\ref{thm:cf1}. If
$n+2\le8c_\varepsilon$, use $n$ unit rounds ($F=0$, $R=n$). Otherwise use $c_\varepsilon-2$ unit rounds (levels
$2,3,\dots,c_\varepsilon$), set $\gamma_0:=c_\varepsilon$, and define the design slots, for every $j\ge1$,
\[
\bar\kappa_j:=\min\Big\{\frac{\varepsilon'2^{-j}}{8(2u_{j-1}+A_\Xi)},\ \Big(\frac{\varepsilon'2^{-j}}{90.6\,C_\star}\Big)^2\Big\},\qquad
\gamma_j:=\lfloor\bar\kappa_j\gamma_{j-1}^2\rfloor,
\]
with $u_j:=\ln\gamma_j$ ($j\ge0$), $x_j:=\bar\kappa_j\gamma_{j-1}$ (the squaring ratio $\gamma_j/\gamma_{j-1}$ before the floor) and $\hat\kappa_j:=\gamma_j/\gamma_{j-1}^2\le\bar\kappa_j$; the
slots depend on $\varepsilon$ only, not on $n$. Stop at $J:=\min\{j\ge0:\bar\kappa_{j+1}\gamma_j^2\ge n+2\}$. If $J=0$ or
$(n+2)/\gamma_J\ge8$, the realised squaring rounds are $\gamma_{j-1}\to\gamma_j$ for $1\le j\le J$, followed by a final
round $\gamma_J\to n+2$ with $\kappa_f:=(n+2)/\gamma_J^2\le\bar\kappa_{J+1}$. Otherwise (merge, as the per-round bound needs $r\ge8$) level $\gamma_J$ is dropped:
the squaring rounds are those with $1\le j\le J-1$, followed by a final round $\gamma_{J-1}\to n+2$ with
$\kappa_f:=(n+2)/\gamma_{J-1}^2<8\hat\kappa_J\le8\bar\kappa_J$ and $r_f=(n+2)/\gamma_{J-1}>\gamma_J/\gamma_{J-1}>20$ (by the minimality of $J$,
$n+2>\bar\kappa_J\gamma_{J-1}^2\ge\gamma_J$), and slot $J$ is not realised. ($J=0$ never needs the merge,
since $n+2>8c_\varepsilon$.)

\emph{Growth and side conditions.} $c_\varepsilon$ is an explicit sufficient choice: $c_\varepsilon\ge64$,
$c_\varepsilon\ge8C_\star$, $c_\varepsilon\ge8.5B_\Xi/\varepsilon$ with $B_\Xi:=4.6C_\star+4.3+1.1(\ln(16\Xi)+1)$, and
$x_1\ge8e$: the second term of $\bar\kappa_1$ gives $x_1\ge4.680\cdot10^{20}/(2899.2\,C_\star)^2=21.74679\ldots>8e=21.74625\ldots$, and the
first gives $x_1\ge309$. Since $\varepsilon'2^{-j}\le1/32$, $\bar\kappa_j\le(32\cdot90.6C_\star)^{-2}<1/(128C_\star^2)$.
For $j\ge2$ each term of the minimum shrinks by at most a factor $4$ per step ($2u_{j-1}+A_\Xi\le2(2u_{j-2}+A_\Xi)$, as
$\gamma_{j-1}\le\gamma_{j-2}^2$), so $\bar\kappa_j\ge\bar\kappa_{j-1}/4$; the floor loses at most a factor $2$
($\gamma_j\ge\bar\kappa_j\gamma_{j-1}^2/2$, as $\bar\kappa_j\gamma_{j-1}^2=x_j\gamma_{j-1}\ge2$). Hence $x_{j+1}\ge x_j^2/8$ for
$j\ge1$, so $x_j\ge8e^{2^{j-1}}$, every squaring ratio $\gamma_j/\gamma_{j-1}\ge x_j-1/\gamma_{j-1}$ exceeds $20$,
$u_j\ge u_{j-1}+2^{j-1}$, and $J\le\lceil\log_2\ln(n+2)\rceil+1$ (as $u_j\ge u_0+2^j-1\ge2^j$ and
$\bar\kappa_{j+1}\gamma_j^2=x_{j+1}\gamma_j\ge\gamma_j$, the stopping condition holds once $2^j\ge\ln(n+2)$).
The block count is $R=c_\varepsilon-1+J$ without the merge
and $R=c_\varepsilon-2+J$ with it, so $R\le c_\varepsilon-1+J\le\log_2\log_2n+8c_\varepsilon$. Every non-unit round,
including the final round in both cases, has $8\le r\le c/4$, $c\ge\max(64,8C_\star,8r+4)$ and $\kappa\le1/(128C_\star^2)$.

\emph{Proof of Theorem~\ref{thm:cf1}.} Apply Lemma U4 with $\mathrm{env}_k\equiv0$ on unit rounds and the per-round
envelope on the others. By Lemma U7, $(c_{j-1}-1)\mathrm{Kol}_j<1$ for every round $j$, and consecutive squaring levels
grow by factors $>20$, so $\sum_{j>k}1/(c_{j-1}-1)\le1.02\cdot\frac{20}{19}/c_k<1.074/c_k$, as every starting level is at
least $64$ and so $1/(c-1)\le1.02/c$ (the final level $n+2$ never appears as a denominator). Hence, for a round from level $c$ to $c_k=r_kc$,
\[
V(\mathrm{env}_k)\sum_{j>k}\mathrm{Kol}_j\le\big(4r_k+\ln(16\Xi)+1\big)\frac{1.074}{r_kc}\le\big[4.3+1.1(\ln(16\Xi)+1)/r_k\big]/c.
\]
This estimate uses the schedule and is not a consequence of the per-round hypotheses alone; it holds with and without
the merge, since the realised starting levels grow by factors $>20$. The design budgets run over all slots $j\ge1$:
$\sum_{j\ge1}\bar\kappa_j(2u_{j-1}+A_\Xi)\le\sum_{j\ge1}\varepsilon'2^{-j}/8=\varepsilon'/8$ and
$\sum_{j\ge1}\sqrt{\bar\kappa_j}\le\varepsilon'/(90.6C_\star)$. A realised squaring round $j$ ($1\le j\le J$ without the
merge, $1\le j\le J-1$ with it) has $\hat\kappa_j\le\bar\kappa_j$ and $\ln r_j\le u_{j-1}$, so it is dominated by slot $j$;
an unmerged final round has $\kappa_f\le\bar\kappa_{J+1}$ and $\ln r\le u_J$ (as $n+2\le\gamma_J^2$), so it is dominated by slot $J+1$. These
rounds occupy distinct slots, so together they contribute at most $1.13\varepsilon'/8$ through the terms
$1.13\kappa(2\ln r+A_\Xi)$ and at most $1.13\cdot11.32\,\varepsilon'/90.6<0.1413\varepsilon'$ through the terms
$1.13\cdot11.32C_\star\sqrt\kappa$. A merged final round ($\kappa_f<8\bar\kappa_J$, and $\ln r\le u_{J-1}$ as $r=\kappa_f\gamma_{J-1}<8\bar\kappa_J\gamma_{J-1}\le\gamma_{J-1}$) is charged
separately and adds at most $1.13\,(\varepsilon'2^{-J}+11.32\sqrt8\,\varepsilon'2^{-J}/90.6)\le1.53\varepsilon'$. A realised
non-unit round from level $c$ contributes at most $(4.52C_\star+4.3+1.1(\ln(16\Xi)+1))/c\le B_\Xi/c$ through its $1/c$
terms; the starting levels grow by factors $>20$ from $c_\varepsilon$, so these total at most
$\frac{20}{19}B_\Xi/c_\varepsilon\le1.053B_\Xi/c_\varepsilon\le\varepsilon/8$ in both cases. So
$F\le(0.1413+0.1413+1.53)\varepsilon'+\varepsilon/8<0.2384\,\varepsilon\le\varepsilon$. \qed

\emph{Proof of Corollary~\ref{cor:cf1}.} Theorem~\ref{thm:cf1} at $\min(\varepsilon,1)$ gives $R_F^{\mathrm{PROF}}(n,\varepsilon)\le\log_2\log_2n+
8c_{\min(\varepsilon,1)}$ (for $\varepsilon>1$, a profile with $F\le1$ also has $F\le\varepsilon$), and Corollary~\ref{cor:thmm} gives $R_F^{\mathrm{PROF}}(n,\varepsilon)\ge\log_2\ln n-O(1+\log(1+\varepsilon))$; the
two differ by $O_\varepsilon(1)$, since $\log_2\ln n=\log_2\log_2n+\log_2\ln2$. \qed

\pend{}

\subsection{The schedule-averaged criterion}\label{app:thmm-hfs}

In the mean-cost class of Section~\ref{sec:setup} (finite predrawn mixtures
$\mu$ of full labelled schedules, $C(\mu)=\E_\mu|U|$, $H(\mu)=\E_\mu\KL(p\|q_U)$, $\LH(n,\varepsilon)=\inf\{C(\mu):
H(\mu)\le\varepsilon\}$), put $T_n:=\ln\frac{n+2}2$ and $D_n:=\mathcal H_{n+1}-1-T_n$, where $\mathcal H_m:=\sum_{i\le m}1/i$.

\begin{proposition}[schedule-averaged depth on this target]\label{prop:hfs}
For every $n\ge1$ and every finite predrawn mixture $\mu$,
\[
C(\mu)\,\big(6H(\mu)+2D_n\big)\ge T_n^2,\qquad 0<D_n<\tfrac12;
\]
hence $\LH(n,\varepsilon)\ge\max\{1,T_n^2/(6\varepsilon+1)\}$ for every $\varepsilon\ge0$. For $\varepsilon>0$ put
$\alpha_H:=\min\{\frac12,\varepsilon/(4(T_n+1))\}$; the labelled schedule that, while $c<n+2$, commits
$b=\min\{n+2-c,\max(1,\lfloor\alpha_H c\rfloor)\}$ coordinates at level $c$ has $\KL(p\|q_U)<\varepsilon$ and at most
$\min\{n,2+(3T_n+2)/\alpha_H\}$ rounds. So $\LH(n,\varepsilon)=\Theta_\varepsilon((\ln n)^2)$ as $n\to\infty$ at every fixed
$\varepsilon>0$, while $\LH(n,0)=n$.
\end{proposition}

\emph{Proof.} After $s$ reveals put $c=s+2$ and $\hat\theta_s:=(1+A_s)/c$, the posterior mean; under $p$ the count $A_s$ is
uniform on $\{0,\dots,s\}$. Put $h_s:=\E\,h(\hat\theta_s)$, with $h$ the binary entropy in nats. The next reveal gives
$\hat\theta_{s+1}=\hat\theta_s+(X-\hat\theta_s)/(c+1)$, where the revealed value $X$ is Bernoulli$(\hat\theta_s)$ given the
past. So $\hat\theta_{s+1}$ has conditional mean $\hat\theta_s$ and conditional variance $\hat\theta_s(1-\hat\theta_s)/(c+1)^2$,
and $h_s-h_{s+1}=\E\,\mathrm{kl}(\hat\theta_{s+1}\|\hat\theta_s)$. Pinsker's inequality $\mathrm{kl}(x\|y)\ge2(x-y)^2$ and
$\E[\hat\theta_s(1-\hat\theta_s)]=(c+1)/(6c)$ give the lower bound $2(c+1)/(6c(c+1)^2)=1/(3c(c+1))$; the Bernoulli bound
$\mathrm{kl}(x\|y)\le(x-y)^2/(y(1-y))$ gives the upper bound $1/(c+1)^2$ conditionally. Hence
\[
\frac1{3c(c+1)}\le h_s-h_{s+1}\le\frac1{(c+1)^2}.\tag{BB1}
\]
By the chain rule, $\KL(p\|q_U)$ is the sum over rounds of the mean, over the committed values, of the
divergence $\KL\big(p(x_{B_t}\mid\cdot)\,\|\,\bigotimes_{i\in B_t}p(x_i\mid\cdot)\big)$ of the block's conditional law from the product of
its conditional marginals, which is the block's conditional total correlation: the sum of its coordinates' conditional
entropies minus their joint conditional entropy. After $s$ committed coordinates each unrevealed one is
Bernoulli$(\hat\theta_s)$, so each coordinate's conditional entropy has mean $h_s$, and expanding the joint entropy coordinate
by coordinate inside the block gives mean $\sum_{i<b}h_{s+i}$. So a block of $b$ coordinates entered at level $c$ costs exactly
$d(c,b):=b\,h_s-\sum_{i<b}h_{s+i}$ in expectation under $p$, and $\KL(p\|q_U)=\sum_td(c_{t-1},b_t)$; this is the
block total-correlation decomposition of fixed-schedule KL \citep{chen2026optimal,secchi2026schedule,wen2026conditional}. For fixed schedules the exact optimum is characterized by \citet[Thm.~4]{chen2026optimal}; the proposition bounds it explicitly on this target through the two-sided increment bound (BB1), and extends the lower bound to predrawn mixtures with expected cost by the linearity of $H$ and $C$ in the schedule law together with Cauchy--Schwarz over the mixture and its rounds; the upper-order conclusion also follows from \citet[Thm.~9]{chen2026optimal} with the mutual-information bound on the dual total correlation in \citet[App.~D]{secchi2026schedule}. The consequence specific to this paper is the same-class join with Theorems~\ref{thm:cf1} and~\ref{thm:mixf} in Corollary~\ref{cor:hfs}. At fixed dimension, as the number of steps grows, \citet[Thm.~3]{wainwright2026data} gives sharp asymptotics of the optimal path divergence; the proposition instead holds at fixed $\varepsilon$ as $n$ grows.
Telescoping (BB1) gives $\frac13\Psi\le d\le\Psi$ with $\Psi(c,b):=\sum_{i<b}i/(c(c+i))\le b(b-1)/(2c^2)$. Over a
schedule the terms $1/(c_{t-1}+i)$, $0\le i<b_t$, cover the levels $2,\dots,n+1$ once each, so
$\sum_t\Psi(c_{t-1},b_t)=\sum_tb_t/c_{t-1}-(\mathcal H_{n+1}-1)$. With $z_t:=\ln(c_t/c_{t-1})$, which sum to $T_n$, and
$e^z-1-z\ge z^2/2$, this yields
\[
\sum_tz_t^2\le6\KL(p\|q_U)+2D_n,\qquad D_n=\sum_{c=2}^{n+1}\Big[\frac1c-\ln\Big(1+\frac1c\Big)\Big]\in\big(0,\tfrac12\big).
\]
Cauchy--Schwarz over $\mu$ and rounds gives $T_n^2=(\E_\mu\sum_tz_t)^2\le C(\mu)\,\E_\mu\sum_tz_t^2\le
C(\mu)(6H(\mu)+2D_n)$ for components of any lengths. For the schedule: singletons cost $0$; every other block has
$b\le\alpha_H c$ and so costs at most $\alpha_H^2/2$; there are fewer than $2/\alpha_H$ initial singletons, at most one final
truncated block, and at most $3T_n/\alpha_H$ other blocks, since each multiplies $c$ by at least $1+\alpha_H/2$ and
$\ln(1+\alpha_H/2)\ge\alpha_H/3$; so $\KL(p\|q_U)\le\alpha_H^2/2+\frac32\alpha_H T_n<\varepsilon$. Finally $d(c,b)>0$ for $b\ge2$ by
(BB1), so $H(\mu)=0$ forces all-singleton schedules. \qed

\subsection{Mixtures: the common lower bound}\label{app:thmm-mix-lower}

The gap sets of Theorem~\ref{thm:thmm} depend on the profile, and the
atoms of one component fill another's gaps (\S\ref{app:thmm-sep}), so Corollary~\ref{cor:thmm} gives no lower bound on $\LF$.
A single test set, the same for every component, does: a finite union of thin neighbourhoods of the reduced fractions in $(0,1)$ with small denominators ($S^\star$ below). In the mean-cost class of
Section~\ref{sec:setup}, with $\LF(n,\varepsilon)=\inf\{C(\mu):F(\mu)\le\varepsilon\}$ evaluated for the present Beta$(1,1)$--Bernoulli target, write $b(U)=(|B_1|,\dots,|B_{|U|}|)$ for the profile of a labelled schedule $U$, $Q_W^{(b)}$
for the count law of the relabelled profile $b$ (T1), and $\bar Q_W:=\sum_U\mu(U)\,Q_W^{(b(U))}$; under a profile the
sampler is that of the Setting, with $P_k=(1+A_k)/c_k$ and $\mathcal F_k$. Let $\phi$ be Euler's totient,
$Z(\vartheta):=\sum_{q\ge2}\phi(q)q^{-\vartheta/2}=\zeta(\vartheta/2-1)/\zeta(\vartheta/2)-1$ ($\zeta$ the Riemann zeta function), which is finite iff
$\vartheta>4$, and $h$ the binary entropy in nats. For $\vartheta>1$, $\xi\ge1$ and $\omega>0$ put
$\mathcal Q_n:=\{q\ge2:\xi q^\vartheta\le n+2\}$, $N_n:=\sum_{q\in\mathcal Q_n}\phi(q)$,
$\omega_q:=\omega(\xi q^\vartheta)^{-1/2}$,
\[
S^\star:=\bigcup_{q\in\mathcal Q_n}\ \bigcup_{1\le p<q,\ \gcd(p,q)=1}[p/q-\omega_q,\,p/q+\omega_q],\qquad
\delta_\omega:=2e^{-\omega^2},
\]
$\lambda_\xi:=\ln\xi/(\vartheta-1)$ and $R_0(n):=\lfloor\log_\vartheta((\ln(n+2)+\lambda_\xi)/(\ln2+\lambda_\xi))\rfloor$;
for $\vartheta>4$ also put $a_n:=[2\omega(n+2)\xi^{-1/2}Z(\vartheta)+N_n]/(n+1)$. Here $\vartheta$, $\xi$ and $\omega$ are the jump exponent, the jump factor and the radius factor; $\mathcal Q_n$ contains every reduced denominator that the run's value can have just before a jump, $N_n$ counts the reduced fractions in $(0,1)$ with these denominators, and $a_n$ bounds the target mass of $S^\star$ (Lemma~X3(ii)).

\emph{The obstruction.} Just before a jump, a round from level $c$ to at least $\xi c^\vartheta$, the run's value is a reduced fraction $p/q$ with $q\le c$, so $q\in\mathcal Q_n$, and the run then ends within $\omega_q$ of it except with probability $\delta_\omega$ (Lemmas~X1 and~X3(i) below). One set must serve every component, so $S^\star$ surrounds all $\phi(q)$ reduced fractions of each $q\in\mathcal Q_n$; its target mass is at most about $2\omega\xi^{-1/2}Z(\vartheta)$ (Lemma~X3(ii)), small for large $\xi$ only because $Z(\vartheta)<\infty$, that is, $\vartheta>4$. Every component with at most $R_0(n)\approx\log_\vartheta\ln n$ rounds jumps (Lemma~X2), so $F\le\varepsilon$ forces mass about $e^{-\varepsilon}$ onto longer components. The resulting coefficient $(\ln2/\ln\vartheta)\,e^{-\varepsilon}$ of $\log_2\log_2n$ tends to $e^{-\varepsilon}/2$ as $\vartheta\downarrow4$, and Remark~\ref{rem:mixf} shows why $\vartheta\le4$ gives nothing.

\begin{theorem}[finite predrawn mixtures on this target]\label{thm:mixf}
(a) Let $n\ge1$. Take $\vartheta>4$, $\xi\ge1$, $\omega>0$ and $\varepsilon\ge0$ with $a_n<1$ and $\delta_\omega<1$. Every
finite predrawn mixture $\mu$ with $F(\mu)\le\varepsilon$ satisfies
\[
\mu(|U|>R_0(n))\ge\eta_n:=\frac{e^{-(\varepsilon+h(a_n))/(1-a_n)}-\delta_\omega}{1-\delta_\omega},
\]
and hence $C(\mu)\ge1+R_0(n)\max(\eta_n,0)$. So $\LF(n,\varepsilon)\ge1+R_0(n)\max(\eta_n,0)$.

(b) For every $\varepsilon\ge0$, $\liminf_n\LF(n,\varepsilon)/\log_2\log_2n\ge e^{-\varepsilon}/2$, and for every
$\varepsilon>0$, $\limsup_n\LF(n,\varepsilon)/\log_2\log_2n\le e^{-\varepsilon}$. Hence, for every $\varepsilon>0$,
$\LF(n,\varepsilon)=\Theta_\varepsilon(\log\log n)$.

(c) Take $R\ge1$, $\vartheta>4$, $n+2\ge2^{\vartheta^R}$ and $\bar\omega>0$. Set
$\xi_n:=((n+2)/2^{\vartheta^R})^{(\vartheta-1)/(\vartheta^R-1)}$ and $\omega:=\bar\omega\sqrt{\xi_n}$, and write
$a_n=[2\bar\omega(n+2)Z(\vartheta)+N_n]/(n+1)$, with $N_n$ computed at $\xi=\xi_n$. Then every finite predrawn mixture $\mu$
whose components all have at most $R$ rounds satisfies
\[
F(\mu)\ge(1-a_n)_+\,(\bar\omega^2\xi_n-\ln2)_+-\ln2.
\]
With $\bar\omega=1/(3Z(\vartheta))$, $a_n\le0.85$ for all $n\ge14$ with $n+2\ge2^{\vartheta^R}$, and the infimum $F^{\mathrm{mix}}_R(n)$ of $F$ over these
mixtures satisfies $F^{\mathrm{mix}}_R(n)\ge(1/(27Z(\vartheta)^2)-o(1))\,\xi_n$. Since $\xi_n\asymp n^{1/(1+\vartheta+\dots+\vartheta^{R-1})}$,
for every $\eta'>0$ a fixed $\vartheta=\vartheta(\eta')>4$ gives $F^{\mathrm{mix}}_R(n)\ge n^{3/(4^R-1)-\eta'}$ for all large $n$; for $R=2$
that is $n^{1/5-\eta'}$.
\end{theorem}
The upper bound in (b) is not sharp: for every $\varepsilon>0$, Corollary~\ref{cor:mixexact}, through
Theorem~\ref{thm:mixu}, lowers its coefficient from $e^{-\varepsilon}$ to $e^{-\varepsilon}/2$. The proof of the lower half of
(b) also shows that, for every $\varepsilon\ge0$, the least hard cap of finite predrawn mixtures with $F\le\varepsilon$ is at least
$(\frac12-o(1))\log_2\log_2n$. The finite-$n$ reach of
Theorem~\ref{thm:mixf} is discussed after Theorem~\ref{thm:fsadapt} and in \S\ref{app:thmm-sep}.

\begin{remark}[the limit of this test set]\label{rem:mixf}
Let $2\le\vartheta<4$, $\xi\ge1$ and $\omega>0$, assume $\mathcal Q_n\ne\varnothing$, put $q_n:=\max\mathcal Q_n$, and suppose
$q_n^{2-\vartheta/2}\ge\sqrt\xi/\omega$. Then $S^\star\supseteq[1/q_n,1-1/q_n]$. For $1<\vartheta<2$ the same conclusion
holds, now under the hypothesis $q_n\ge\sqrt\xi/\omega$. At $\vartheta=4$ with $\xi$ and $\omega$ fixed,
$\Pr_t(P_R\in S^\star)\to1$. So no test set of this form, with $\vartheta$, $\xi$ and $\omega$ fixed, gives a coefficient of $\log_2\log_2n$ above
$\frac12$ in the lower bound on $\LF$: for $1<\vartheta\le4$ it captures almost all target mass for large $n$, so the divergence step of Part~(a) yields nothing, and for $\vartheta>4$ the coefficient $(\ln2/\ln\vartheta)\,\eta_\infty$ of the proof of~(b) is below~$\frac12$.
\end{remark}

\emph{Proof of Theorem~\ref{thm:mixf}.} Four facts are used.

\emph{X0 (count reduction for mixtures).} $F(\mu)\ge\KL(p_W\|\bar Q_W)$. Indeed $x\mapsto|x|$ pushes $p$ to $p_W$ and
$Q_\mu$ to $\bar Q_W$: given the past, round $t$ adds $\Bin(|B_t|,(1+A_{t-1})/(2+|B_1|+\dots+|B_{t-1}|))$ ones, so a
labelled schedule's count law depends only on its block sizes. KL does not increase under a measurable map. Equality
holds for the uniform relabelling of $\mu$, which has the same cost.

\emph{Lemma X1 (tail after round $k-1$; any single profile).} For every $1\le k\le R$ and $t>0$, almost surely,
$\Pr_s(|P_R-P_{k-1}|\ge t\mid\mathcal F_{k-1})\le2\exp(-2t^2/V_k)\le2\exp(-c_kt^2)$, with
$V_k:=\sum_{j=k}^Rb_j/c_j^2<2/c_k$.

\emph{Proof.} Order the coordinates within each round. Let $\mathcal G_{j,i}$ be $\mathcal F_{j-1}$ together with the
first $i$ coordinates of round $j$, and $Y_{j,i}:=\E_s[P_R\mid\mathcal G_{j,i}]$. $P$ is a martingale, and round $j$ is
i.i.d.\ Bernoulli$(P_{j-1})$ given $\mathcal F_{j-1}$. Hence $Y_{j,i}=(1+A_{j-1}+H_{j,i}+(b_j-i)P_{j-1})/c_j$, with
$H_{j,i}$ the ones among those first $i$ coordinates. So $Y_{k,0}=P_{k-1}$, $Y_{R,b_R}=P_R$ and $Y_{j,b_j}=Y_{j+1,0}$.
Each increment $(x-P_{j-1})/c_j$, with $x\in\{0,1\}$ the new coordinate, lies in $[-P_{j-1}/c_j,(1-P_{j-1})/c_j]$, an
interval of deterministic length $1/c_j$ with a predictable left end. Iterating conditional Hoeffding over all
coordinates of rounds $k,\dots,R$ gives $\E_s[e^{\lambda(P_R-P_{k-1})}\mid\mathcal F_{k-1}]\le e^{\lambda^2V_k/8}$.
Chernoff with $\lambda=4t/V_k$ bounds each tail by $e^{-2t^2/V_k}$; the same argument applies to $-(P_R-P_{k-1})$.
Finally $V_k<2/c_k$: first $b_k/c_k^2<1/c_k$; then for $j>k$, $b_j/c_j^2\le(c_j-c_{j-1})/(c_jc_{j-1})=1/c_{j-1}-1/c_j$,
which telescopes to $<1/c_k$. \qed

\emph{Lemma X2 (forced jumps).} Fix $\vartheta>1$ and $\xi\ge1$, and call round $k$ a \emph{jump} if
$c_k\ge\xi c_{k-1}^\vartheta$. Let $\Gamma_R:=\xi^{(\vartheta^R-1)/(\vartheta-1)}2^{\vartheta^R}$. If $n\ge1$ and $\Gamma_R\le n+2$,
every profile of $n$ with at most $R$ parts has a jump. Moreover
$R_0(n)=\max\{R\ge0:\Gamma_R\le n+2\}=\lfloor\log_\vartheta((\ln(n+2)+\lambda_\xi)/(\ln2+\lambda_\xi))\rfloor$.

\emph{Proof.} If there is no jump, then $c_k<\xi c_{k-1}^\vartheta$ for every $k$. Since $x\mapsto\xi x^\vartheta$ is
increasing, induction gives $c_k<\Gamma_k$ for $k\ge1$, where $\Gamma_0=2=c_0$ and $\Gamma_k=\xi\Gamma_{k-1}^\vartheta$. $\Gamma_k$
is nondecreasing because $\xi\ge1$ and $\Gamma\ge2$. For a profile with $R'\le R$ parts ($R'\ge1$, since $n\ge1$) this gives
$n+2=c_{R'}<\Gamma_{R'}\le\Gamma_R\le n+2$, a contradiction. The closed form is
$\ln\Gamma_R=\vartheta^R(\lambda_\xi+\ln2)-\lambda_\xi$. \qed

\emph{Lemma X3 (one test set).} Fix $\vartheta>4$, $\xi\ge1$ and $\omega>0$. (i) If the profile $b$ has a jump, then
$\Pr_s(P_R\notin S^\star)\le\delta_\omega$ under $b$. (ii) $\Pr_t(P_R\in S^\star)\le a_n$, and
$N_n\le((n+2)/\xi)^{2/\vartheta}/2$.

\emph{Proof.} (i) Let $c=c_{k-1}$ at the jump round $k$, so $\xi c^\vartheta\le c_k\le n+2$. $P_{k-1}=j/c$ with
$1\le j\le c-1$ reduces to $p/q$ with $2\le q\le c$ and $\gcd(p,q)=1$. Then $\xi q^\vartheta\le\xi c^\vartheta\le n+2$,
so $q\in\mathcal Q_n$, and $\omega_q\ge\omega(\xi c^\vartheta)^{-1/2}\ge\omega c_k^{-1/2}$. So $P_R\notin S^\star$
forces $|P_R-P_{k-1}|\ge\omega c_k^{-1/2}$. By Lemma X1, conditionally on $\mathcal F_{k-1}$ this has probability at
most $2e^{-\omega^2}$. Take expectations. (ii) The target's atoms are the $n+1$ points $1/(n+2),\dots,(n+1)/(n+2)$, each with mass
$1/(n+1)$, and a closed interval of length $\ell$ contains at most $(n+2)\ell+1$ atoms. Sum over the $\phi(q)$
intervals of length $2\omega_q$ for each $q\in\mathcal Q_n$, and use $\sum_{q\ge1}\phi(q)q^{-s}=\zeta(s-1)/\zeta(s)$
at $s=\vartheta/2$, which is finite iff $\vartheta>4$. Finally $N_n\le\sum_{q=2}^{Q}(q-1)\le Q^2/2$, with
$Q=((n+2)/\xi)^{1/\vartheta}$. \qed

\emph{Part (a).} By X0 and data processing onto the indicator of $\{P_R\notin S^\star\}$,
$\varepsilon\ge\mathrm{kl}(u\|v)$, where $u:=\Pr_t(P_R\notin S^\star)\ge1-a_n$ and $v:=\bar Q_W(P_R\notin S^\star)$.
We claim $v\ge e^{-(\varepsilon+h(a_n))/(1-a_n)}$. If $v\ge1-a_n$, this holds because
$1-a_n\ge e^{-h(a_n)/(1-a_n)}$, which is equivalent to $-a_n\ln a_n\ge0$. Otherwise $v<1-a_n\le u$. $\mathrm{kl}(\cdot\|v)$ is
increasing on $[v,1]$, so $\varepsilon\ge\mathrm{kl}(1-a_n\|v)\ge(1-a_n)\ln(1/v)-h(a_n)$; the dropped term
$-a_n\ln(1-v)$ is $\ge0$. Rearrange. By Lemma X2, every $U$ with $|U|\le R_0(n)$ has a jump. By Lemma X3(i), such a
component puts mass at most $\delta_\omega$ on $\{P_R\notin S^\star\}$. With $m:=\mu(|U|>R_0(n))$ this gives
$v\le\delta_\omega(1-m)+m$, hence $m\ge(v-\delta_\omega)/(1-\delta_\omega)\ge\eta_n$. Finally $|U|\ge1$ always, as $n\ge1$, and
$|U|\ge R_0(n)+1$ on $\{|U|>R_0(n)\}$. So $C(\mu)\ge1+R_0(n)\,m$.

\emph{Part (b), lower bound.} Fix $\vartheta>4$, $\xi\ge1$ and $\omega$ with $a_\infty:=2\omega Z(\vartheta)/\sqrt\xi<1$
and $\omega^2>\ln2$ (so $\delta_\omega<1$); part (a) needs both premises. As $n\to\infty$, $N_n/(n+1)\to0$ because
$\vartheta>2$, so $a_n\to a_\infty$. Also $R_0(n)\ge\log_\vartheta\ln(n+2)-\log_\vartheta(\ln2+\lambda_\xi)-1$. Part (a)
gives $\liminf_n\LF(n,\varepsilon)/\log_\vartheta\ln n\ge\eta_\infty:=[e^{-(\varepsilon+h(a_\infty))/(1-a_\infty)}-
\delta_\omega]/(1-\delta_\omega)$. Since $\log_\vartheta\ln n/\log_2\log_2n\to\ln2/\ln\vartheta$,
$\liminf_n\LF(n,\varepsilon)/\log_2\log_2n\ge(\ln2/\ln\vartheta)\,\eta_\infty$. Every admissible fixed
$(\vartheta,\xi,\omega)$ bounds the same liminf, so take the supremum: $\xi\to\infty$ gives $a_\infty\to0$ and
$h(a_\infty)\to0$; then $\omega\to\infty$ gives $\delta_\omega\to0$; then $\vartheta\downarrow4$ gives
$\ln2/\ln\vartheta\to\frac12$. The bound is $e^{-\varepsilon}/2$. For a lower bound on the least hard cap only $\eta_\infty>0$ is needed, and for every $\varepsilon\ge0$ and $\vartheta>4$ some admissible $(\xi,\omega)$ (with $\xi\ge1$, $\omega>0$, $\omega^2>\ln2$ and $a_\infty<1$) gives it. Then $\eta_n>0$ for all large $n$, so by part (a) every finite predrawn mixture with $F(\mu)\le\varepsilon$ gives positive weight to a component with more than $R_0(n)$ rounds; as $R_0(n)/\log_2\log_2n\to\ln2/\ln\vartheta$, letting $\vartheta\downarrow4$ shows that the least hard cap of these mixtures is at least $(\frac12-o(1))\log_2\log_2n$.

\emph{Part (b), upper bound.} Fix $\varepsilon'\in(0,\min(\varepsilon,1)]$. Theorem~\ref{thm:cf1} holds for every
$n\ge2$ and $\varepsilon'\in(0,1]$; let $\mu'$ be the uniform relabelling of its profile, with $F(\mu')\le\varepsilon'$
and cost at most $\log_2\log_2n+8c_{\varepsilon'}$. Put $w:=e^{-(\varepsilon-\varepsilon')}$ and
$\mu:=w\mu'+(1-w)\cdot$(the one-round schedule), a finite predrawn mixture. Pointwise $Q_\mu\ge w\,Q_{\mu'}$, so
$\ln(p/Q_\mu)\le\ln(p/Q_{\mu'})+\ln(1/w)$, and $F(\mu)\le\varepsilon'+\ln(1/w)=\varepsilon$. The cost is
$C(\mu)\le w(\log_2\log_2n+8c_{\varepsilon'})+(1-w)$. With $\varepsilon'$ fixed, $c_{\varepsilon'}$ is a constant, so
$\limsup_nC(\mu)/\log_2\log_2n\le w$. Let $\varepsilon'\downarrow0$.

\emph{Part (c).} Part (a), and through it the lower bound of part (b), keeps $\xi$ fixed and bounds from below the mass of components longer than $R_0(n)$; part (c) instead caps every component at $R$ rounds and lets the jump factor grow with $n$. Here $\xi_n$ is the largest jump factor allowed by the hypothesis $\Gamma_R\le n+2$ of Lemma X2, so every component jumps, and $\omega=\bar\omega\sqrt{\xi_n}$ keeps the bound $a_n$ on the target mass of $S^\star$ free of $\xi_n$ except through $N_n$, while each component's mass outside $S^\star$ is at most $2e^{-\bar\omega^2\xi_n}$. With $\xi=\xi_n$, $\Gamma_R=n+2$ exactly, and $\xi_n\ge1$ iff $n+2\ge2^{\vartheta^R}$, which also gives $n\ge1$ (as $2^{\vartheta^R}>16$). So by Lemma X2
every component of $\mu$ has a jump. By Lemma X3 with $\omega=\bar\omega\sqrt{\xi_n}$:
$v:=\bar Q_W(P_R\notin S^\star)\le2e^{-\bar\omega^2\xi_n}$, and $u:=\Pr_t(P_R\notin S^\star)\ge1-a_n$. Then
$\mathrm{kl}(u\|v)\ge u\ln(1/v)-\ln2$, since $u\ln u+(1-u)\ln(1-u)\ge-\ln2$ and $-(1-u)\ln(1-v)\ge0$. If
$\bar\omega^2\xi_n\le\ln2$, the claimed bound is $-\ln2$, which is trivial. Otherwise $\ln(1/v)\ge\bar\omega^2\xi_n-\ln2>0$ and
$u\ge(1-a_n)_+\ge0$, so $u\ln(1/v)\ge(1-a_n)_+(\bar\omega^2\xi_n-\ln2)$. With $\bar\omega=1/(3Z(\vartheta))$:
$a_n=\frac23+(\frac23+N_n)/(n+1)\le0.85$ for $n\ge14$
(Lemma X3(ii) and $\xi_n\ge1$ give $N_n\le\sqrt{n+2}/2$, since $2/\vartheta<\frac12$; and $(\frac23+\sqrt{n+2}/2)/(n+1)$ is
decreasing in $n$ and equals $8/45<11/60=0.85-\frac23$ at $n=14$). The bound is then
$(\frac13-o(1))(\xi_n/(9Z(\vartheta)^2)-\ln2)-\ln2=(1/(27Z(\vartheta)^2)-o(1))\,\xi_n-O(1)$. Finally
$\xi_n=((n+2)/2^{\vartheta^R})^{1/(1+\vartheta+\dots+\vartheta^{R-1})}$, and the exponent increases to $3/(4^R-1)$ as
$\vartheta\downarrow4$. Given $\eta'$, choose a fixed $\vartheta$ with
$1/(1+\vartheta+\dots+\vartheta^{R-1})\ge3/(4^R-1)-\eta'/2$. \qed

\emph{Proof of Remark~\ref{rem:mixf}.} By Dirichlet's approximation theorem, for $y\in[1/q_n,1-1/q_n]$ there are $p$
and $q$ with $1\le q\le q_n$ and $|y-p/q|<1/(qq_n)$. Here $q=1$ would give $|y-p|<1/q_n$ with $p\in\{0,1\}$, which the
range of $y$ excludes; so $q\ge2$ and $p/q\in(0,1)$. Reducing $p/q$ to $p'/q'$ with $q'\le q$ keeps
$q'\in\mathcal Q_n$ and only enlarges the radius, since $\omega_q$ is decreasing in $q$. It remains to check
$1/(qq_n)\le\omega_q$, i.e.\ $q^{\vartheta/2-1}\le\omega q_n/\sqrt\xi$. For $\vartheta\ge2$,
$q^{\vartheta/2-1}\le q_n^{\vartheta/2-1}$, and $q_n^{\vartheta/2-1}\le\omega q_n/\sqrt\xi$ is equivalent to
$q_n^{2-\vartheta/2}\ge\sqrt\xi/\omega$. For $1<\vartheta<2$ and $q\ge2$, $q^{\vartheta/2-1}<1\le\omega q_n/\sqrt\xi$,
using the hypothesis $q_n\ge\sqrt\xi/\omega$. At $\vartheta=4$: $\psi(q)=\omega/(\sqrt\xi\,q)$ is
non-increasing and $\sum\psi(q)=\infty$; for such $\psi$, by Khintchine's theorem \citep{khintchine1924einige}, in the form stated by \citet[Thm.~2.3, arXiv v1 numbering]{beresnevich2016metric} (here $q\psi(q)$ is constant, so Khintchine's original monotonicity hypothesis on $q\psi(q)$ also holds), almost every $y$ has infinitely many $p/q$ with
$|y-p/q|<\psi(q)/q=\omega_q$. These can be taken reduced: $\psi(q)/q$ is decreasing, so reducing $p/q$ to $p'/q'$
keeps $|y-p'/q'|<\psi(q')/q'$, and for irrational $y$ infinitely many solutions converging to $y$ give infinitely many
distinct reduced fractions. The set $S^\star=S^\star_n$ increases with $n$, and for fixed $m$,
$\Pr_t(P_R\in S^\star_m)\to\mathrm{Leb}(S^\star_m\cap[0,1])$ as $n\to\infty$, since $S^\star_m$ is a finite union of intervals and
the target's atoms are equally weighted at spacing $1/(n+2)$. So $\liminf_n\Pr_t(P_R\in S^\star_n)\ge\mathrm{Leb}(S^\star_m\cap[0,1])$ for every $m$, and the right
side tends to $1$ as $m\to\infty$: a reduced solution $p/q$ puts $y$ in $S^\star_m$ once $q\in\mathcal Q_m$, and almost
every $y\in[0,1]$ has one. Hence $\Pr_t(P_R\in S^\star_n)\to1$. \qed

\pend{}

\subsection{Mixtures: the fourth-power construction and the exact constant}\label{app:thmm-mix-upper}

Theorem~\ref{thm:mixf} leaves the $\liminf$ and $\limsup$ of
$\LF(n,\varepsilon)/\log_2\log_2n$ between $e^{-\varepsilon}/2$ and $e^{-\varepsilon}$; a recursive mixture reaches the
lower end. Below a cutoff chosen from the tolerance, its schedules commit one coordinate per round. Above it, the
schedule at size $n$ is, with probability $n^{-2}$, a \emph{floor} of $n$ unit rounds, which sample exactly; otherwise
it draws a prefix level $d$ uniformly from the band $\{c_{\min}(n),\dots,2c_{\min}(n)-1\}$, where
$c_{\min}(n)=\lceil(n+2)^{1/4}(\ln(n+2))^8\rceil$, schedules the first $d-2$ coordinates by the same recursion and commits the
other $n-d+2$ in one round. If the prefix count $a$ were exactly uniform, the prefix value $(1+a)/d$ would be a uniform
point of the grid $\{1/d,\dots,(d-1)/d\}$, and the band would average $c_{\min}(n)$ such grids. Two grids, smoothed by the same
window, overlap exactly as one grid, whose denominator is the least common multiple of theirs, overlaps with itself
(Lemma~Z3), and a band has few small common multiples: over its pairs of denominators the reciprocals of the least
common multiples sum to $O(\ln c_{\min}(n))$. So the band average, smoothed at a width a polylogarithmic factor below the
binomial spread $(n+2)^{-1/2}$, is close to uniform on $[0,1]$ (Lemma~Z4); the bound of Lemma~Z4 is small only when this width exceeds the band's typical gap $c_{\min}(n)^{-2}$ by a factor large compared with $\ln c_{\min}(n)$, which holds once $c_{\min}(n)$ exceeds $(n+2)^{1/4}$ by a polylogarithmic factor, and this is where the fourth power comes from. A binomial count with a uniform parameter is
exactly uniform, so one round from this ideal prefix gives almost the target's count law (Lemma~Z5, with Lemmas~Z1
and~Z2). The actual prefix comes from the recursion and is only approximately uniform. The common-mass lift (Lemma~Z6)
compares the two on records of prefix size, prefix count and final count: the lifted target follows the ideal law on the
mass that the ideal law shares with the target, and the rest of the target is matched with the floor. Since the ideal and the actual law
share the final round, the prefix's error passes to size $n$ with coefficient one, while the one-step distance enters
multiplied by $O(\ln n)$ and the floor's weight adds $O(n^{-2})$. Errors therefore add along the recursion instead of
compounding, and their sum is at most the tolerance: the logarithm of the size at least halves at each recursion step, and the
cutoff is chosen from the tolerance. Each floor is charged in full, $1/n$ in expectation at size $n$, and along any
branch these charges sum to less than one. As $\ln c_{\min}(n)\approx\frac14\ln n$, a potential, $\ln(n+2)$ less a logarithmic
correction, shrinks by a factor of at least four per recursion step, so there are at most $1+\log_4\ln(n+2)$ recursion steps of one round
each, and the scheduled mean is at most $\frac12\log_2\log_2(n+2)+O_\varepsilon(1)$; dilution
(Corollary~\ref{cor:mixexact}) then gives the factor $e^{-\varepsilon}$. Averaging a lattice denominator over a dyadic
band also underlies randomized lattice rules \citep{kritzer2019lattice}. In this part, the auxiliary count laws and
kernels introduced for the construction are set in sans serif: $\mathsf U_s$ is the uniform law on $\{0,\dots,s\}$, the target's count law at
size $s$; $\mathsf Q_s$ is the count law that the construction produces at size $s$; $\mathsf R_n$ and $\mathsf A_n$ are
the ideal one-step law and the fast law, the count law of the non-floor branch, at size $n$; and $\mathsf K$ denotes a kernel. The band start is written $c_{\min}$, or $c_{\min}(s)$ at size $s$,
and letters introduced inside a proof are local to it.

\emph{Construction.} Fix $\varepsilon_0>0$, put $Y_{\varepsilon_0}:=\max\{1024,(10^4/\varepsilon_0)^{1/3}\}$ and
$\mathcal{N}_{\varepsilon_0}:=\lceil e^{Y_{\varepsilon_0}}\rceil$. The following recursion draws a random profile
$\mathrm{Schedule}(n)$ of $n$ (a sequence of block sizes); it draws no output coordinate. If $n\le \mathcal{N}_{\varepsilon_0}$, return $n$
singleton blocks. Otherwise put $N:=n+2$, $c_{\min}(n):=\lceil N^{1/4}(\ln N)^8\rceil$ and $\varrho_n:=n^{-2}$; with probability
$\varrho_n$ return $n$ singleton blocks (the \emph{floor}); otherwise draw $d$ uniformly from $\{c_{\min}(n),\dots,2c_{\min}(n)-1\}$, draw
$b'$ from $\mathrm{Schedule}(d-2)$, independently, and return $b'$ followed by one block of size $n-(d-2)$. After the
profile is drawn, draw one independent uniform permutation of the $n$ labels and assign consecutive labels to the
blocks; only then run the independent oracle draws. Multiplication by the finite uniform label law preserves
normalization and finite support; call the result $\mu_n$. \emph{Legality.} Display~(Z7.5) in the proof of
Theorem~\ref{thm:mixu} below gives $1\le d-2<n$ at every nonbase call. By strong induction in $n$, the result is then a
finite, normalized law on profiles of $n$: the weights $n^{-2}$ and $1-n^{-2}$ sum to one, the band has exactly $c_{\min}(n)$
elements, and each child law is normalized. The final block $n-(d-2)\ge1$ is nonempty, and each component has at most
$n$ nonempty rounds. \emph{Exactness of singletons.} If the count after $s$ singleton draws is uniform on
$\{0,\dots,s\}$, then an interior next count $a$ has probability $\frac1{s+1}\big(\frac{s+1-a}{s+2}+\frac
a{s+2}\big)=\frac1{s+2}$; the endpoints give the same value. \emph{Prefix embedding.} The oracle conditional
$(a+1)/(s+2)$ depends only on the committed counts, so the count dynamics of a prefix inside the larger problem are
exactly the child's count law, even with a single global permutation. \emph{Count reduction.} Let $\mathsf{Q}_n$ be the count law
of $\mu_n$ (the law of the number of ones of its output), and $\mathsf{U}_n:=\mathrm{Unif}\{0,\dots,n\}=p_W$. Then
$Q_{\mu_n}(x)=\mathsf{Q}_n(|x|)/\binom n{|x|}$, so $F(\mu_n)=\KL(p\|Q_{\mu_n})=\KL(\mathsf{U}_n\|\mathsf{Q}_n)$ exactly (X0 gives equality for a
uniform relabelling). \emph{Floor decomposition.} For $n>\mathcal{N}_{\varepsilon_0}$, $\mathsf{Q}_n=\varrho_n\mathsf{U}_n+(1-\varrho_n)\mathsf{A}_n$, where the fast law $\mathsf{A}_n$
draws $d$, takes the recursive prefix count $a$ at size $s=d-2$, and moves to the new count $W=a+\Bin(n-s,(a+1)/(s+2))$,
a sum of fresh independent bit draws. Hence $\mathsf{Q}_n\ge n^{-2}\mathsf{U}_n$ and $Q_{\mu_n}\ge n^{-2}p$ pointwise. The floor is an
alternative schedule chosen in advance; it repairs no executed trajectory.

\emph{Lemma Z1 (translation smoothing).} (a) Let $Y\sim\Bin(b,\theta)$, with $b\ge0$ and $0<\theta<1$, and let $v$ be an
integer. Then $\mathrm{TV}(Y,Y+v)\le|v|/(2\sqrt{(b+1)\theta(1-\theta)})$. (b) For integers $N>d\ge2$, $j\in\{1,\dots,d-1\}$ and
$\theta=j/d$, let $Y\sim\Bin(N-d,\theta)$ and, independently, $Z\sim\Bin(d,\theta)$, so $\E Z=j$. Then
\[
\mathrm{TV}(Y+j,\Bin(N,\theta))\le\frac{\E|Z-j|}{2\sqrt{(N-d+1)\theta(1-\theta)}}\le\frac12\sqrt{\frac d{N-d+1}}\le\sqrt{\frac d{N-d}},
\]
uniformly in $j$, including the edge values $j=1$ and $j=d-1$.

\emph{Proof.} (a) Extend $u_b(k):=\Pr(Y=k)$ by $0$ outside $\{0,\dots,b\}$. For $0\le k\le b+1$,
$u_b(k-1)-u_b(k)=\frac{k-(b+1)\theta}{(b+1)\theta(1-\theta)}\,u_{b+1}(k)$, since
$u_b(k-1)=\frac k{(b+1)\theta}u_{b+1}(k)$ and $u_b(k)=\frac{b+1-k}{(b+1)(1-\theta)}u_{b+1}(k)$. So $\mathrm{TV}(Y+1,Y)=\E|Z-(b+1)\theta|/(2(b+1)\theta(1-\theta))$ for
$Z\sim\Bin(b+1,\theta)$, which is at most $1/(2\sqrt{(b+1)\theta(1-\theta)})$ by Cauchy--Schwarz and the binomial variance.
Translation invariance and the triangle inequality give the claim. (b) $\Bin(N,\theta)$ is the law of $Y+Z$: average (a)
over $Z$, with $b=N-d$ and $v=Z-j$, and use $\E|Z-j|\le\sqrt{d\theta(1-\theta)}$. \qed

\emph{Lemma Z2 (parameter perturbation).} For an integer $N\ge1$, $0<\theta_0\le\frac14$ and $\theta,\theta'\in[\theta_0,1-\theta_0]$,
$\mathrm{TV}(\Bin(N,\theta),\Bin(N,\theta'))\le\sqrt{N/\theta_0}\,|\theta-\theta'|$.

\emph{Proof.} The binomial masses $u_N(k;t)$ satisfy $\frac d{dt}u_N(k;t)=\frac{k-Nt}{t(1-t)}u_N(k;t)$, so
$\sum_k|u_N'(k;t)|=\E|\Bin(N,t)-Nt|/(t(1-t))\le\sqrt{N/(t(1-t))}$. Integrate from $\theta$ to $\theta'$ and halve, using
$t(1-t)\ge\theta_0(1-\theta_0)\ge\theta_0/2$. \qed

\emph{Grids.} Work on the circle $[0,1)$ modulo one, with Lebesgue measure $\mathrm{Leb}$. For an integer $d\ge1$ let
$G_d:=\frac1d\sum_{j=0}^{d-1}\delta_{j/d}$. For $0<\upsilon\le\frac12$ let $\kappa_\upsilon$ be the density of the uniform law on
$[-\upsilon/2,\upsilon/2)$, wrapped modulo one, and put $f_d(t):=\frac1d\sum_{j=0}^{d-1}\kappa_\upsilon(t-j/d)$, a step density with
integral one. These objects keep the denominator, the grid point and the smoothing window apart. The denominator $d$ is the prefix's level, drawn by the schedule before any coordinate is sampled; the grid point is the prefix value $(1+a)/d$, set by the prefix count $a$, so it takes only the $d-1$ points $j/d$ with $1\le j\le d-1$, with equal masses under the ideal law of Lemma~Z5 and with the recursive count law under the actual law. $G_d$ adds the point $0$ and gives each of its $d$ points the mass $1/d$, which moves that law by $1/d$ in total variation (step~(2) of Lemma~Z5). The window $\kappa_\upsilon$ is never sampled: at the width $\upsilon$ chosen in Lemma~Z5, a polylogarithmic factor below the final round's binomial spread $N^{-1/2}$, smoothing the parameter changes the final round's law only slightly (step~(4) of Lemma~Z5, through Lemma~Z2), and the smoothed band average is close to Lebesgue measure (Lemma~Z4).

\emph{Lemma Z3 (exact overlap identity).} For positive integers $d$ and $e$ put $L:=\mathrm{lcm}(d,e)$, $z:=L\upsilon$ and
$\{z\}:=z-\lfloor z\rfloor$. Then
\[
\int_0^1f_d(t)f_e(t)\,dt=1+\frac{\{z\}(1-\{z\})}{(L\upsilon)^2},\qquad\text{hence}\qquad
0\le\int_0^1f_df_e-1\le\frac1{\upsilon\,\mathrm{lcm}(d,e)}.
\]

\emph{Proof.} The map $(i\bmod d,\,j\bmod e)\mapsto(L/d)i-(L/e)j\bmod L$ is onto, since $\gcd(L/d,L/e)=1$, and every
fibre has $\gcd(d,e)=de/L$ elements. So the difference of independent uniform points of $G_d$ and $G_e$ is uniform on
$G_L$. With $\mathcal{O}(u):=\int_0^1\kappa_\upsilon(t)\kappa_\upsilon(t+u)\,dt$ this gives
$\int f_df_e=\frac1L\sum_{r=0}^{L-1}\mathcal{O}(r/L)=\int f_L^2$; the second equality is the case $d=e=L$. For almost every $t$,
$f_L(t)=N_L(t)/(L\upsilon)$, where $N_L(t)$ counts the points of $G_L$ in a circular window of length $\upsilon$. That count is
$m:=\lfloor L\upsilon\rfloor$ or $m+1$, and since $\int N_L=L\upsilon=m+\{z\}$, it is $m+1$ on a set of measure $\{z\}$. So
$\int(f_L-1)^2=\mathrm{Var}(N_L)/(L\upsilon)^2=\{z\}(1-\{z\})/(L\upsilon)^2$, which is the identity, since $\int f_L=1$. Finally
$\{z\}(1-\{z\})\le L\upsilon$: if $L\upsilon<1$ then $\{z\}=L\upsilon$, and otherwise $\{z\}(1-\{z\})\le\frac14\le L\upsilon$. \qed

\emph{Lemma Z4 (band smoothing).} Let $c_{\min}\ge2$ be an integer with $\ln c_{\min}\ge2$, put
$\bar\nu_{c_{\min}}:=\frac1{c_{\min}}\sum_{d=c_{\min}}^{2c_{\min}-1}G_d$, and let $0<\upsilon\le\frac12$. Then
$\mathrm{TV}(\bar\nu_{c_{\min}}*\kappa_\upsilon,\mathrm{Leb})\le3\sqrt{\ln c_{\min}/(\upsilon c_{\min}^2)}$.

\emph{Proof.} The density of $\bar\nu_{c_{\min}}*\kappa_\upsilon$ is $f:=\frac1{c_{\min}}\sum_{d=c_{\min}}^{2c_{\min}-1}f_d$, and Lemma~Z3 gives
$\int(f-1)^2\le\frac1{\upsilon c_{\min}^2}\sum_{d,e\in[c_{\min},2c_{\min})}1/\mathrm{lcm}(d,e)$. That sum equals $\sum_{d,e}\gcd(d,e)/(de)$, which
is at most $\frac1{c_{\min}^2}\sum_{r\le2c_{\min}}r\,\#\{d\in[c_{\min},2c_{\min}):r\mid d\}^2$, since $\gcd(d,e)$ is at most the sum of its
divisors and $de\ge c_{\min}^2$. A band of $c_{\min}$ consecutive integers has at most $c_{\min}/r+1\le3c_{\min}/r$ multiples of $r\le2c_{\min}$, so the
sum is at most $9\mathcal H_{2c_{\min}}\le9(1+\ln2c_{\min})\le18\ln c_{\min}$, as $\ln c_{\min}\ge2$; here $\mathcal H_m:=\sum_{i\le m}1/i$.
Cauchy--Schwarz then gives $\mathrm{TV}=\frac12\int|f-1|\le\frac12\sqrt{18\ln c_{\min}/(\upsilon c_{\min}^2)}\le3\sqrt{\ln c_{\min}/(\upsilon c_{\min}^2)}$. \qed

\emph{Lemma Z5 (one-step comparison).} Let $N=n+2$, and let $c_{\min}\ge3$ be an integer with $4c_{\min}\le N$. The \emph{ideal}
one-step law $\mathsf{R}_n$ draws $d$ uniformly from $\{c_{\min},\dots,2c_{\min}-1\}$, the prefix count $a$ uniformly from $\{0,\dots,d-2\}$,
and then $W=a+\Bin(n-(d-2),(a+1)/d)$; it is an analysis reference only. If $\ln c_{\min}\ge2$ and $0<\upsilon\le\theta_0\le\frac14$,
then
\[
\mathrm{TV}(\mathsf{R}_n,\mathsf{U}_n)\le2\sqrt{c_{\min}/N}+\frac3{c_{\min}}+3\sqrt{\frac{\ln c_{\min}}{\upsilon c_{\min}^2}}+4\theta_0+\sqrt{N/\theta_0}\,\upsilon+\frac2{N+1}.
\]
In particular, if $\ln c_{\min}\ge2$ and the choice $\theta_0:=(\ln c_{\min})^{-4}$, $\upsilon:=N^{-1/2}(\ln c_{\min})^{-6}$ satisfies
$\upsilon\le\theta_0\le\frac14$, then
\[
\mathrm{TV}(\mathsf{R}_n,\mathsf{U}_n)\le2\sqrt{c_{\min}/N}+\frac3{c_{\min}}+5(\ln c_{\min})^{-4}+\frac{3N^{1/4}(\ln c_{\min})^{7/2}}{c_{\min}}+\frac2{N+1}.
\]

\emph{Proof.} In shifted counts, with $j=a+1$, $W+1=j+\Bin(N-d,j/d)$; a shift preserves TV. (1) By Lemma~Z1(b),
replacing this by $\Bin(N,j/d)$ costs averaged TV at most $\sqrt{2c_{\min}/(N-2c_{\min})}\le2\sqrt{c_{\min}/N}$. (2) The law of $j/d$,
$\nu_{c_{\min}}:=\frac1{c_{\min}}\sum_{d=c_{\min}}^{2c_{\min}-1}\frac1{d-1}\sum_{j=1}^{d-1}\delta_{j/d}$, differs from $\bar\nu_{c_{\min}}$ by TV at most
$1/c_{\min}$: each incomplete grid is at TV $1/d$ from $G_d$. (3) For $P\sim\bar\nu_{c_{\min}}$,
$\Pr\{P\notin[2\theta_0,1-2\theta_0]\}\le4\theta_0+2/c_{\min}$, since at most $2\theta_0d+1$ of the $d$ points of each $G_d$ lie in
$[0,2\theta_0)$, at most $2\theta_0d+1$ lie in $(1-2\theta_0,1)$, and $d\ge c_{\min}$. (4) Let $P\sim\bar\nu_{c_{\min}}$ and, independently,
$Z\sim\mathrm{Unif}[-\upsilon/2,\upsilon/2)$, and put $P':=(P+Z)\bmod1$, so $P'\sim\bar\nu_{c_{\min}}*\kappa_\upsilon$. Since $\upsilon\le\theta_0$, off the
event in (3) there is no wrapping and $P,P'\in[\theta_0,1-\theta_0]$, so Lemma~Z2 bounds the conditional TV there by
$\sqrt{N/\theta_0}\,|Z|$; bounding TV by one on that event and using $\E|Z|\le \upsilon$ gives
$\mathrm{TV}(\bar\nu_{c_{\min}}\mathsf{K}_N,(\bar\nu_{c_{\min}}*\kappa_\upsilon)\mathsf{K}_N)\le4\theta_0+2/c_{\min}+\sqrt{N/\theta_0}\,\upsilon$, where $\mathsf{K}_N$ is the kernel
$\theta\mapsto\Bin(N,\theta)$ and $\nu \mathsf{K}_N$ is the law of $\Bin(N,P)$ for $P\sim\nu$. (5) Kernels contract TV, so Lemma~Z4 gives
$\mathrm{TV}(\bar\nu_{c_{\min}}\mathsf{K}_N,\mathrm{Leb}\,\mathsf{K}_N)\le3\sqrt{\ln c_{\min}/(\upsilon c_{\min}^2)}+4\theta_0+2/c_{\min}+\sqrt{N/\theta_0}\,\upsilon$. (6)
$\mathrm{Leb}\,\mathsf{K}_N=\mathrm{Unif}\{0,\dots,N\}$, since $\int_0^1\binom Nk\theta^k(1-\theta)^{N-k}\,d\theta=1/(N+1)$; this is integration by
parts at natural exponents, and no real Gamma function is needed. (7) The target's shifted count is uniform on
$\{1,\dots,N-1\}$, at TV exactly $2/(N+1)$ from $\mathrm{Unif}\{0,\dots,N\}$. Steps (1), (2) (through the kernel $\mathsf{K}_N$)
and (5)--(7) give the first bound. (8) Substituting $\theta_0=(\ln c_{\min})^{-4}$ and $\upsilon=N^{-1/2}(\ln c_{\min})^{-6}$ gives the
second. \qed

\emph{How the next lemma is used.} In the notation of Lemma~Z6, at a nonbase size $n$ in the proof of Theorem~\ref{thm:mixu}, $\mathsf{B}$ is the uniform law
of the prefix size $s=d-2$ over the band, $\mathsf{U}_s=\mathrm{Unif}\{0,\dots,s\}$ is the target's count law at size $s$ and
$\mathsf{Q}_s$ the recursive one, and $\mathsf{K}_s(a,\cdot)$ is the final round, the law of $a+\Bin(n-s,(a+1)/(s+2))$. Then $\mathsf{P}=\mathsf{U}_n$,
$\mathsf{R}_K$ is the ideal law $\mathsf{R}_n$ of Lemma~Z5, $\mathsf{A}_K$ is the fast law $\mathsf{A}_n$, and with the floor weight $\varrho=n^{-2}$ the law $\mathsf{Q}$
is the actual count law $\mathsf{Q}_n$; $\Delta$ is the one-step distance $\mathrm{TV}(\mathsf{U}_n,\mathsf{R}_n)$. In the proof of Lemma~Z6 a fast record $(s,a,v)$ is a run of the band branch with prefix size $s$, prefix count $a$ and final count $v$, and an exact record $v$ is a floor run, recorded by its final count. The floor enters in three ways: together with the exactness of singletons at base sizes it makes every $\mathsf{Q}_s$ strictly positive; its term $\varrho\mathsf{P}$ in $\mathsf{Q}$ absorbs the target mass that the ideal law misses, at a cost of at most $\ln(1/\varrho)=2\ln n$ per unit; and it leaves the band branch the weight $1-\varrho$, which costs at most $-\ln(1-\varrho)$. The ideal law's excess mass, also $\Delta$, is dropped from the lifted target at a cost of at most $\ln(s_{\max}+1)$ per unit.

\emph{Lemma Z6 (common-mass lift).} Let $\mathsf{B}$ be a law on finitely many positive integer prefix sizes $s\le s_{\max}$.
For each $s$ let $\mathsf{U}_s:=\mathrm{Unif}\{0,\dots,s\}$, let $\mathsf{Q}_s$ be a strictly positive law on $\{0,\dots,s\}$, and let
$\mathsf{K}_s(a,\cdot)$ be a Markov kernel into a finite terminal space; let $\mathsf{P}$ be a strictly positive target law on that
space. Put $\mathsf{R}(s,a,v):=\mathsf{B}(s)\mathsf{U}_s(a)\mathsf{K}_s(a,v)$ and $\mathsf{A}(s,a,v):=\mathsf{B}(s)\mathsf{Q}_s(a)\mathsf{K}_s(a,v)$, with terminal marginals $\mathsf{R}_K$ and $\mathsf{A}_K$,
and, for $0<\varrho<1$, $\mathsf{Q}:=(1-\varrho)\mathsf{A}_K+\varrho \mathsf{P}$ and $\Delta:=\mathrm{TV}(\mathsf{P},\mathsf{R}_K)$. Then
\[
\KL(\mathsf{P}\|\mathsf{Q})\le\sum_s\mathsf{B}(s)\KL(\mathsf{U}_s\|\mathsf{Q}_s)+\Delta\,[\ln(s_{\max}+1)+\ln(1/\varrho)]-\ln(1-\varrho).
\]

\emph{Proof.} With $g:=\min(\mathsf{P},\mathsf{R}_K)$, define the lifted target $\mathsf{T}$ on fast records $(s,a,v)$ and exact records $v$ by
$\mathsf{T}_F(s,a,v):=g(v)\mathsf{R}(s,a,v)/\mathsf{R}_K(v)$ ($0$ if $\mathsf{R}_K(v)=0$) and $\mathsf{T}_E(v):=\mathsf{P}(v)-g(v)$. It is normalized, with terminal
marginal $\mathsf{P}$. The actual lift is $(1-\varrho)\mathsf{A}$ on fast records and $\varrho \mathsf{P}$ on exact ones, with terminal marginal $\mathsf{Q}$.
Finite data processing gives $\KL(\mathsf{P}\|\mathsf{Q})\le\KL(\mathsf{T}\|\text{lift})$. \emph{Fast records.} On those with positive target
mass (there $\mathsf{A}>0$, since $\mathsf{Q}_s>0$), $\ln(\mathsf{T}_F/((1-\varrho)\mathsf{A}))=\ln(g/\mathsf{R}_K)+\ell(s,a)-\ln(1-\varrho)$, with $\ell:=\ln(\mathsf{U}_s/\mathsf{Q}_s)$, because $\mathsf{B}(s)$ and the kernel $\mathsf{K}_s$, common to $\mathsf{R}$ and $\mathsf{A}$, cancel; this is why the prefix's KL passes on with coefficient one.
Records with $\mathsf{T}_F=0$ contribute zero. The fast part has total mass $1-\Delta$, so the $-\ln(1-\varrho)$ term contributes
exactly $(1-\Delta)(-\ln(1-\varrho))\le-\ln(1-\varrho)$. The first term is at most $0$. Since $\mathsf{T}_F\le \mathsf{R}$, the removed mass
is $\Delta$ and $\ell\ge-\ln(s_{\max}+1)$ (because $\mathsf{Q}_s\le1$),
$\sum \mathsf{T}_F\ell\le\sum \mathsf{R}\ell+\Delta\ln(s_{\max}+1)=\sum_s\mathsf{B}(s)\KL(\mathsf{U}_s\|\mathsf{Q}_s)+\Delta\ln(s_{\max}+1)$. \emph{Exact
records.} $\mathsf{T}_E\le \mathsf{P}$ gives $\sum \mathsf{T}_E\ln(\mathsf{T}_E/(\varrho \mathsf{P}))\le\Delta\ln(1/\varrho)$. \qed

No upper bound on the log-ratio $\ell$, that is, no lower bound on the input laws $\mathsf{Q}_s$ beyond their strict positivity, is used.

\begin{theorem}[finite predrawn mixtures on this target: upper bound]\label{thm:mixu}
For every $\varepsilon\in(0,1]$ there is $K_\star(\varepsilon)<\infty$ such that
$\LF(n,\varepsilon)\le\frac12\log_2\log_2(n+2)+K_\star(\varepsilon)$ for all $n\ge1$.
\end{theorem}

\begin{corollary}[the exact asymptotic constant]\label{cor:mixexact}
For every $\varepsilon>0$, $\lim_{n\to\infty}\LF(n,\varepsilon)/\log_2\log_2n=e^{-\varepsilon}/2$.
\end{corollary}

\emph{Proof of Theorem~\ref{thm:mixu}.} For $\varepsilon_0>0$ let $\mu_n=\mu_n^{(\varepsilon_0)}$ be the mixture of the construction
above, a finite predrawn mixture. We show that, for every fixed $\varepsilon_0>0$ and every integer $n\ge1$,
\begin{gather*}
F(\mu_n)=\KL(p\|Q_{\mu_n})\le\varepsilon_0\qquad\text{and}\\
C(\mu_n)\le\log_4\ln(n+2)+\mathcal{N}_{\varepsilon_0}+2\le\tfrac12\log_2\log_2(n+2)+\mathcal{N}_{\varepsilon_0}+2;
\end{gather*}
the second inequality holds because $\log_4\log_2(n+2)=\log_4\ln(n+2)-\log_4\ln2$. The divergence bound comes from a recursion for the count-law divergence: \emph{Scales} checks the premises of Lemma~Z5 at $c_{\min}=c_{\min}(n)$, which gives the one-step bound (Z7.4), and records (Z7.5), by which every child has at most half its parent's logarithm; \emph{Error induction} feeds (Z7.4) into Lemma~Z6, so that the divergence at size $n$ is at most the band average of the children's divergences plus $7000/(\ln(n+2))^3$, and (Z7.5) makes these terms summable along every branch. The cost bound unrolls the exact cost recurrence along the same branches (\emph{Cost}), with a potential that (Z7.5) shrinks by a factor four per step (\emph{Depth}) and a bound on the expected charge of the floors (\emph{Floors}).

\emph{Scales.} At a nonbase size $n>\mathcal{N}_{\varepsilon_0}$ put $y:=\ln N\ge Y_{\varepsilon_0}\ge1024$, so $c_{\min}(n)=\lceil e^{y/4}y^8\rceil$. Then
\[
e^{y/4}y^8\le c_{\min}(n)\le2e^{y/4}y^8,\qquad y/4\le\ln c_{\min}(n)\le y/2,\qquad4c_{\min}(n)\le N.\tag{Z7.1}
\]
The bounds $\ln c_{\min}(n)\le y/2$ and $4c_{\min}(n)\le N$ use $y/4+8\ln y+\ln4\le y/2$ (Z7.2), which holds at $1024$, and the derivative
$1/4+8/y<1/2$ of its left side thereafter. Also, by (Z7.1), $2\sqrt{c_{\min}(n)/N}\le y^{-4}$, $c_{\min}(n)^{-1}\le y^{-4}$,
$N^{1/4}(\ln c_{\min}(n))^{7/2}/c_{\min}(n)\le y^{-9/2}$ and $(N+1)^{-1}\le y^{-4}$ (Z7.3); the first reduces to
$2\sqrt2e^{-3y/8}y^8\le1$, which holds at $1024$ and decreases thereafter. Lemma~Z5 applies at $c_{\min}=c_{\min}(n)$: $c_{\min}(n)\ge3$,
$4c_{\min}(n)\le N$ and $\ln c_{\min}(n)\ge y/4\ge2$, and its parameters satisfy $\theta_0\le\frac14$ and $\upsilon\le\theta_0$. Using
$5(\ln c_{\min}(n))^{-4}\le1280y^{-4}$, it gives
\[
\Delta_n:=\mathrm{TV}(\mathsf{R}_n,\mathsf{U}_n)\le2000\,y^{-4}.\tag{Z7.4}
\]
Finally $c_{\min}(n)\ge3$ and $2c_{\min}(n)-1<N$, so every $d$ in the band, with child logarithm $y':=\ln d$, satisfies
\[
1\le d-2<n,\qquad y'\le y/4+8\ln y+\ln4\le y/2.\tag{Z7.5}
\]

\emph{Error induction.} Let $\mathcal{D}_n:=\KL(\mathsf{U}_n\|\mathsf{Q}_n)$; then $\mathcal{D}_n=0$ for $n\le \mathcal{N}_{\varepsilon_0}$. For $n>\mathcal{N}_{\varepsilon_0}$ apply Lemma~Z6
with the actual child laws as $\mathsf{Q}_s$ (strictly positive, by exactness of singletons and the floor decomposition), the
real binomial transition, $\varrho=n^{-2}$ and $s_{\max}<n$; here $\mathsf{B}$ is uniform on the prefix sizes $s=d-2$ of the band,
$\mathsf{P}=\mathsf{U}_n$, $\mathsf{R}_K=\mathsf{R}_n$ and $\mathsf{Q}=\mathsf{Q}_n$. With (Z7.4),
\begin{align*}
\mathcal{D}_n&\le\frac1{c_{\min}(n)}\sum_{d=c_{\min}(n)}^{2c_{\min}(n)-1}\mathcal{D}_{d-2}+\Delta_n[\ln(n+1)+2\ln n]-\ln(1-n^{-2})\\
&\le\frac1{c_{\min}(n)}\sum_{d=c_{\min}(n)}^{2c_{\min}(n)-1}\mathcal{D}_{d-2}+\frac{7000}{y^3},
\end{align*}
since the $\Delta_n$ term is at most $6000/y^3$ and $-\ln(1-n^{-2})\le2/n^2\le y^{-3}$. Expand this recursion along
auxiliary paths which, at each nonbase size, always take the band branch, to the child of size $d-2$ with $d$ uniform on the band, and never the floor, whose contribution the recursion already counts; this
is analysis only, and $\E$ below averages over these random paths. Along such a path write $y_i:=\ln(n_i+2)$ for its nonbase sizes $n_0=n,\dots,n_{\ell-1}$, so
$y_0=\ln(n+2)$ and $n_i=e^{y_i}-2$. Iterating the recursion along a path, which stops at a base size where $\mathcal D=0$,
gives $\mathcal D_n\le\E\sum_{i<\ell}7000/y_i^3$. By (Z7.5), read backward they satisfy $y_{\ell-1-r}\ge2^rY_{\varepsilon_0}$, so
\[
\sum_{i<\ell}\frac{7000}{y_i^3}\le\frac{7000}{Y_{\varepsilon_0}^3}\sum_{r\ge0}2^{-3r}=\frac{8000}{Y_{\varepsilon_0}^3}\le\varepsilon_0.
\]
Averaging gives $\mathcal{D}_n\le\varepsilon_0$, that is, $F(\mu_n)\le\varepsilon_0$ by the count reduction. The cutoff was fixed from $\varepsilon_0$
before $n$ varied.

\emph{Cost.} For $n\le \mathcal{N}_{\varepsilon_0}$, $C(\mu_n)=n$. Otherwise, exactly,
\[
C(\mu_n)=n^{-2}\cdot n+(1-n^{-2})\Big(1+\frac1{c_{\min}(n)}\sum_{d=c_{\min}(n)}^{2c_{\min}(n)-1}C(\mu_{d-2})\Big),
\]
hence $C(\mu_n)\le1+c_{\min}(n)^{-1}\sum_dC(\mu_{d-2})+1/n$. Unrolling along the same auxiliary paths gives
$C(\mu_n)\le \mathcal{N}_{\varepsilon_0}+\E\ell+\E\sum_{i<\ell}1/n_i$; every floor is charged all $n$ of its singleton rounds.
\emph{Depth.} Put $f(y):=y-16\ln y$. For $y\ge1024$,
\[
f(y/4+8\ln y+\ln4)\le y/4+8\ln y+\ln4-16\ln(y/4)=y/4-8\ln y+17\ln4\le f(y)/4;
\]
the last step is $4\ln y\ge17\ln4$. Since $f$ is increasing on that range and $y'\le y/4+8\ln y+\ln4$ by (Z7.5), a nonbase
child has $f(y')\le f(y)/4$. Hence, for $\ell\ge1$,
$1<f(\ln(\mathcal{N}_{\varepsilon_0}+2))<f(y_{\ell-1})\le4^{-(\ell-1)}f(y_0)\le4^{-(\ell-1)}y_0$, so $\ell\le1+\log_4y_0$.
\emph{Floors.} Since $n_i=e^{y_i}-2\ge e^{y_i}/2$,
\[
\sum_{i<\ell}\frac1{n_i}\le2\sum_{r\ge0}e^{-2^r\ln(\mathcal{N}_{\varepsilon_0}+2)}\le\frac4{\mathcal{N}_{\varepsilon_0}+2}<1,
\]
using $2^r\ge r+1$ and a geometric series. Therefore $C(\mu_n)\le\log_4\ln(n+2)+\mathcal{N}_{\varepsilon_0}+2$ for every $n\ge1$; for
$n\le \mathcal{N}_{\varepsilon_0}$ this holds because $\log_4\ln(n+2)>0$.

For $\varepsilon\in(0,1]$ take $\varepsilon_0=\varepsilon$: the finite predrawn mixture $\mu_n^{(\varepsilon)}$ has
$F(\mu_n^{(\varepsilon)})\le\varepsilon$ and $C(\mu_n^{(\varepsilon)})\le\frac12\log_2\log_2(n+2)+\mathcal{N}_\varepsilon+2$, so
$\LF(n,\varepsilon)\le\frac12\log_2\log_2(n+2)+K_\star(\varepsilon)$ for all $n\ge1$, with
$K_\star(\varepsilon):=\mathcal{N}_\varepsilon+2$. \qed

\emph{Proof of Corollary~\ref{cor:mixexact}.} Fix $\varepsilon>0$ and $\varepsilon'\in(0,\min(\varepsilon,1)]$. Let
$\mu'$ be the mixture of Theorem~\ref{thm:mixu} at tolerance $\varepsilon'$, and put $w:=e^{\varepsilon'-\varepsilon}$
and $\mu:=w\mu'+(1-w)\cdot$(the one-round schedule), a finite predrawn mixture. Pointwise $Q_\mu\ge wQ_{\mu'}$, so
$F(\mu)\le\varepsilon'+\ln(1/w)=\varepsilon$, and $C(\mu)\le w(\frac12\log_2\log_2(n+2)+K_\star(\varepsilon'))+1$. Since
$K_\star(\varepsilon')$ does not depend on $n$, $\limsup_n\LF(n,\varepsilon)/\log_2\log_2n\le e^{\varepsilon'-\varepsilon}/2$
for each fixed $\varepsilon'$; taking the infimum over $\varepsilon'\in(0,\min(\varepsilon,1)]$ gives
$\limsup_n\LF(n,\varepsilon)/\log_2\log_2n\le e^{-\varepsilon}/2$. Theorem~\ref{thm:mixf}(b) gives
$\liminf_n\LF(n,\varepsilon)/\log_2\log_2n\ge e^{-\varepsilon}/2$ for every $\varepsilon\ge0$. \qed

\emph{Scope of the construction.} Let $\mu'$ be the mixture $\mu_n$ that proves Theorem~\ref{thm:mixu}, at a tolerance $\varepsilon'\in(0,\min(\varepsilon,1)]$, as in the
proof of Corollary~\ref{cor:mixexact}: the construction above at $\varepsilon_0=\varepsilon'$, so with cutoff
$\mathcal{N}_{\varepsilon'}$. At every nonbase size $m$ of its recursion a floor of $m$ unit rounds has probability $m^{-2}$,
and $C(\mu')$ charges it in full; the top floor alone is a branch of $n$ unit rounds with positive probability (for
$n\le \mathcal{N}_{\varepsilon'}$, every component is $n$ unit rounds). Every component of $\mu'$ begins with at least $L_0:=\min(n,c_{\min}(\mathcal{N}_{\varepsilon'}+1)-2)$
unit rounds, where $c_{\min}(\mathcal{N}_{\varepsilon'}+1)$ is the band start at the smallest nonbase size: for $n>\mathcal{N}_{\varepsilon'}$ the innermost prefix is either a
floor at a nonbase size $m\ge \mathcal{N}_{\varepsilon'}+1$, of $m$ unit rounds, or a base schedule at size $d-2$ with $d\ge c_{\min}(n')\ge c_{\min}(\mathcal{N}_{\varepsilon'}+1)$
for its nonbase parent $n'$, since $c_{\min}(m)$ is nondecreasing in $m$; and a floor is no shorter, as $c_{\min}(\mathcal{N}_{\varepsilon'}+1)-2<\mathcal{N}_{\varepsilon'}+1$, since $4c_{\min}(\mathcal{N}_{\varepsilon'}+1)\le \mathcal{N}_{\varepsilon'}+3$ by (Z7.1).
The constants are astronomical: $\mathcal{N}_{\varepsilon'}\ge e^{1024}$, and (Z7.1) at size $\mathcal{N}_{\varepsilon'}+1$ gives $\ln c_{\min}(\mathcal{N}_{\varepsilon'}+1)\ge Y/4+8\ln Y>311$, with
$Y:=Y_{\varepsilon'}\ge1024$. The mixture of Corollary~\ref{cor:mixexact} gives $\mu'$ the weight $w>0$ and adds the
one-round schedule with weight $1-w$. For fixed $\varepsilon\in(0,1]$, the saving of Corollary~\ref{cor:mixexact} over
single relabelled profiles, whose constant is $1$ (Corollary~\ref{cor:cf1}), is therefore asymptotic only: these
constructions do not certify a practical finite-$n$ speedup. They are explicit: a component of $\mu'$ is drawn by running the recursion, which makes at most $1+\log_4\ln(n+2)$ nonbase calls on every branch, and then one uniform relabelling (for the mixture of Corollary~\ref{cor:mixexact}, after choosing $\mu'$ with probability $w$ and otherwise the one-round schedule), so the support need not be listed. Neither the description of a mixture nor the computation and randomness used to draw a component are charged in $C(\mu)$, and no running-time or random-bit bound is given for this construction. Adding a nonnegative charge for them, with the classes of mixtures and policies and the accuracy requirement unchanged, can only raise costs, so the lower-bound parts of Theorems~\ref{thm:mixf} and~\ref{thm:fsadapt} still hold under their hypotheses, while no upper bound under such a charge follows. Also, $\mu'$ has a positive-weight component with $n$ unit
rounds (its top floor, or for $n\le \mathcal{N}_{\varepsilon'}$ every component), so for $R<n$ neither mixture has all its components of at most
$R$ rounds (and for $R<L_0$ no component of $\mu'$ has); for such $R$ they supply no hard-cap upper bound. For a fixed cap
of $R$ rounds, Theorem~\ref{thm:mixf}(c) gives, for every $\eta'>0$, $F^{\mathrm{mix}}_R(n)\ge n^{3/(4^R-1)-\eta'}$ for
large $n$, and no matching hard-cap upper bound is claimed.

\pend{}

\subsection{Causal adaptive policies}\label{app:thmm-adapt}

Theorems~\ref{thm:mixf} and~\ref{thm:mixu} concern finite predrawn
mixtures. Here $\pi\in\mathcal A_\infty$ is a causal randomized adaptive policy of Section~\ref{sec:setup}, run with the
present oracle: in round $j$ it commits a set $B_j$ of unrevealed coordinates, chosen measurably with respect to
$\mathcal F_{j-1}=\sigma(\text{seed},\text{committed labelled history})$, and given $\mathcal F_{j-1}$ the values on $B_j$
are fresh independent Bernoulli$(P_{j-1})$ draws, where $c_j:=2+(\text{number committed after round }j)$, $A_j$ is the
number of ones among them and $P_j:=(1+A_j)/c_j$. So $b_j:=|B_j|$ and $c_j=c_{j-1}+b_j$ are fixed before round $j$'s
draws. $\tau$ counts executed rounds (empty rounds included; $\tau=\infty$ if the policy never completes) and
$\tau_+\le\tau$ the nonempty ones. Put $\LF^{\mathrm{ad}}(n,\varepsilon):=\inf\{\E\tau:\pi\in\mathcal A_\infty,\
\KL(p\|Q_\pi)\le\varepsilon\}$, the least expected number of executed rounds. A finite predrawn mixture $\mu$ lies in
$\mathcal A_\infty$ with $\tau=\tau_+=|U|$ and $Q_\pi=Q_\mu$, so $\E\tau=C(\mu)$; for the mixture of Theorem~\ref{thm:mixu} this charges every floor branch
in full. Hence $\LF^{\mathrm{ad}}\le\LF$, and part (a) below, applied to a mixture (for which $\tau_+=|U|$), contains
Theorem~\ref{thm:mixf}(a); the component split in that theorem's proof is the simpler route for mixtures.

\begin{theorem}[causal adaptive policies on this target]\label{thm:fsadapt}
(a) Let $n\ge1$. Take $\vartheta>4$, $\xi\ge1$, $\omega>0$ and $\varepsilon\ge0$ with $a_n<1$ and $\delta_\omega<1$. Every
$\pi\in\mathcal A_\infty$ with $\KL(p\|Q_\pi)\le\varepsilon$ that completes almost surely satisfies
$\Pr(\tau_+>R_0(n))\ge\eta_n$, with $\eta_n$ as in Theorem~\ref{thm:mixf}(a) (and $a_n$, $\delta_\omega$ and $R_0(n)$ as defined before that
theorem), and hence
$\E\tau\ge\E\tau_+\ge1+R_0(n)\max(\eta_n,0)$. So $\LF^{\mathrm{ad}}(n,\varepsilon)\ge1+R_0(n)\max(\eta_n,0)$.

(b) For every $\varepsilon\ge0$, $\liminf_n\LF^{\mathrm{ad}}(n,\varepsilon)/\log_2\log_2n\ge e^{-\varepsilon}/2$.

(c) For every $\varepsilon>0$, $\lim_n\LF^{\mathrm{ad}}(n,\varepsilon)/\log_2\log_2n=e^{-\varepsilon}/2$.
\end{theorem}

\emph{Proof.} Part (a) runs the test of Theorem~\ref{thm:mixf}(a) on the adaptive output law itself: data processing onto the indicator that the output's value $(1+\text{ones})/(n+2)$ lies outside $S^\star$, and then a lower bound, from the binary divergence, on the probability of that event under the output law. Because block sizes are now chosen from the seed and the past, two steps change: the spread bound of Lemma X1 is proved again for such batches (Lemma X1$'$), and the split of a mixture by component is replaced by a split by the first jump, which is decided before its draws (Lemma X3$'$); Lemma X2's induction, run over the nonempty rounds, and the target bound of Lemma X3(ii) carry over. Part (b) follows from (a) as for mixtures, and part (c) adds the class inclusion $\LF^{\mathrm{ad}}\le\LF$ and Corollary~\ref{cor:mixexact}. Let $\pi$ complete almost surely; otherwise $\E\tau=\infty$. Since $B_j$ is a function of the seed and
the history, $\sigma(\mathcal F_{j-1},B_j)=\mathcal F_{j-1}$, and the values on $B_j$ are i.i.d.\ Bernoulli$(P_{j-1})$
given $\mathcal F_{j-1}$ whatever labels $B_j$ contains, because every unrevealed coordinate has the same exact marginal.
At most $n$ rounds are nonempty, so $P_\infty:=\lim_jP_j$ exists on every path and, on $\{\tau<\infty\}$, equals $P_\tau$, the value
$(1+\text{number of ones})/(n+2)$ of the output.

\emph{Lemma X1$'$ (adaptive tail).} Let $\pi$ complete almost surely. For every $k\ge1$ and every $\mathcal F_{k-1}$-measurable $t>0$, almost surely,
$\Pr(|P_\infty-P_{k-1}|\ge t\mid\mathcal F_{k-1})\le2e^{-c_kt^2}$.

\emph{Proof.} Given $\mathcal F_{j-1}$, $P_j-P_{j-1}$ is the sum of the $b_j$ independent terms $(x-P_{j-1})/c_j$, with
$x$ the new values, each of mean $0$ in an interval of length $1/c_j$. Hoeffding's lemma for each term gives
$\E[e^{\lambda(P_j-P_{j-1})}\mid\mathcal F_{j-1}]\le e^{\lambda^2b_j/(8c_j^2)}$ for bounded $\mathcal F_{j-1}$-measurable
$\lambda$; bounding the round as one increment of range $b_j/c_j$ would not suffice. With
$\Lambda_m:=\sum_{i=k}^mb_i/c_i^2$, whose increments are predictable although later batch sizes are random,
$Z_m:=\exp(\lambda(P_m-P_{k-1})-\lambda^2\Lambda_m/8)$ is, for bounded $\mathcal F_{k-1}$-measurable $\lambda$, a supermartingale for $m\ge k-1$, with $Z_{k-1}=1$: for $j\ge k$ such a $\lambda$ is $\mathcal F_{j-1}$-measurable, so the bound above applies to it. On every
path $\Lambda_\infty\le b_k/c_k^2+\sum_{i>k}(1/c_{i-1}-1/c_i)<2/c_k$, as $b_i/c_i^2\le b_i/(c_{i-1}c_i)=1/c_{i-1}-1/c_i$, as in Lemma X1; an empty round adds $0$. Since
$c_k=c_{k-1}+b_k$ is $\mathcal F_{k-1}$-measurable ($B_k$ is chosen before round $k$'s draws), $\E[e^{\lambda(P_m-P_{k-1})}\mid\mathcal F_{k-1}]\le e^{\lambda^2/(4c_k)}$ for every
$m$. The event is empty where $t>1$, since every $P_j$ lies in $(0,1)$. So take $\lambda:=\pm2c_kt\,\mathbf 1\{t\le1\}$,
which is $\mathcal F_{k-1}$-measurable with $|\lambda|\le2(n+2)$. Then $e^{\lambda(P_m-P_{k-1})}\le e^{2(n+2)}$ and the
variables are eventually constant, so conditional dominated convergence gives the same bound for $P_\infty$, with no
optional stopping, and the conditional Markov inequality on $\{t\le1\}$ gives the claim. \qed

\emph{Lemma X3$'$ (first jump).} Let $\pi$ complete almost surely. Call round $j$ a jump if $b_j\ge1$ and $c_j\ge\xi c_{j-1}^\vartheta$, and let $J$ be
the first jump ($J=\infty$ if there is none). Then $\{J=k\}\in\mathcal F_{k-1}$ for every $k$, and
$\Pr(P_\infty\notin S^\star,\,J=k)\le\delta_\omega\Pr(J=k)$.

\emph{Proof.} $\{J=k\}$ is determined by $c_0,\dots,c_k$. On it, $P_{k-1}$ reduces to $p/q$ with $2\le q\le c_{k-1}$,
$\xi q^\vartheta\le c_k\le n+2$ and $\omega_q\ge\omega c_k^{-1/2}$, exactly as in Lemma X3(i), so $P_\infty\notin S^\star$
forces $|P_\infty-P_{k-1}|\ge\omega c_k^{-1/2}$. Apply Lemma X1$'$ with this $\mathcal F_{k-1}$-measurable $t$, and take
expectations. \qed

A similar partition, by the first crossing of a threshold sequence, handles adaptive grids in batched bandits
\citep[proof of Thm.~3]{gao2019batched}.

\emph{Part (a).} Data processing onto the indicator that the output's value $(1+\text{ones})/(n+2)$ lies outside
$S^\star$ is applied to the adaptive output law itself. The realized profile of an adaptive policy is not an independent
predrawn mixture, so X0's mixture form is not used. Data processing gives $\varepsilon\ge\mathrm{kl}(u\|v)$, with $u:=\Pr_t((1+W)/(n+2)\notin S^\star)\ge1-a_n$ by Lemma
X3(ii) and $v:=\Pr(P_\infty\notin S^\star)$; as in Theorem~\ref{thm:mixf}(a), $v\ge e^{-(\varepsilon+h(a_n))/(1-a_n)}$.
Summing Lemma X3$'$ over the disjoint events $\{J=k\}$ gives
$v\le\Pr(J=\infty)+\delta_\omega\Pr(J<\infty)=\delta_\omega+(1-\delta_\omega)\Pr(J=\infty)$. An empty round never
jumps, since $\xi c^\vartheta>c$. On a completed path, deleting the empty rounds leaves a profile of $n$, the nonempty block sizes, with the same levels and the same jumps; if the path has no jump, Lemma X2 therefore gives it more than $R_0(n)$ nonempty rounds:
$\{J=\infty\}\cap\{\tau<\infty\}\subseteq\{\tau_+>R_0(n)\}$. Hence $\Pr(\tau_+>R_0(n))\ge\eta_n$. Finally, as $n\ge1$, $\tau\ge\tau_+\ge1$, and
$\tau_+\ge R_0(n)+1$ on $\{\tau_+>R_0(n)\}$. The first jump is the right event because it is fixed before its draws; the
event $\{\tau_+\le R_0(n)\}$, on which Theorem~\ref{thm:mixf}(a) splits a mixture component by component, depends on later
draws.

\emph{Parts (b) and (c).} Part (b) follows from (a) exactly as the lower half of Theorem~\ref{thm:mixf}(b): for fixed
admissible $(\vartheta,\xi,\omega)$ let $n\to\infty$, then take $\xi\to\infty$, $\omega\to\infty$ and
$\vartheta\downarrow4$. For (c), $\LF^{\mathrm{ad}}\le\LF$ and Corollary~\ref{cor:mixexact} give the matching
$\limsup$; this uses only the statement of Corollary~\ref{cor:mixexact} and the class inclusion. \qed

Theorem~\ref{thm:fsadapt} concerns expected executed rounds, but its part (a), a tail bound, also bounds hard caps. A
policy in $\mathcal A_r$ completes, and has $\tau_+\le\tau\le r$, almost surely, so $\Pr(\tau_+>R_0(n))=0$ when $r\le R_0(n)$. For
every $\varepsilon\ge0$ and $\vartheta>4$, the admissible $(\xi,\omega)$ with $\eta_\infty>0$ of the proof of
Theorem~\ref{thm:mixf}(b) give $a_n<1$ and $\eta_n>0$ for all large $n$, so by (a) every $\pi\in\mathcal A_r$ with
$\KL(p\|Q_\pi)\le\varepsilon$ has $r>R_0(n)$; as $R_0(n)/\log_2\log_2n\to\ln2/\ln\vartheta$, letting $\vartheta\downarrow4$ shows
that the least hard cap of causal policies with $F\le\varepsilon$ is at least $(\frac12-o(1))\log_2\log_2n$, as for mixtures.
For fixed $\varepsilon>0$, one uniformly relabelled profile of length $L$, a finite predrawn mixture and so a policy in
$\mathcal A_L$, gives at most $\log_2\log_2n+O_\varepsilon(1)$ (Theorem~\ref{thm:fs}(b)); the sharp constant between these
bounds is open. Theorem~\ref{thm:fsadapt} says nothing about upper bounds on tail latency, $H$ for
adaptive policies, other Beta parameters or practical finite $n$ (the bound in (a) is at most $2$ while
$n+2<\xi^{1+\vartheta}2^{\vartheta^2}=\Gamma_2$, since then $R_0(n)\le1$ by Lemma X2 and $\eta_n\le1$, and, like Theorem~\ref{thm:mixf}, it bites only at astronomical $n$), and its upper half is the predrawn construction, with the qualifications
stated after Corollary~\ref{cor:mixexact}.

\pend{}

\subsection{Separation and scope}\label{app:thmm-sep}

\begin{corollary}[same-class separation on this target]\label{cor:hfs}
For $0<\varepsilon\le1$ and $n\ge2$, the upper construction of Theorem~\ref{thm:cf1} (not the optimality statement of Corollary~\ref{cor:cf1}
among single relabelled profiles) is a finite predrawn mixture with cost $R$, so
$\LF(n,\varepsilon)\le\log_2\log_2n+8c_\varepsilon$, and with Proposition~\ref{prop:hfs}
\[
\frac{\LH(n,\varepsilon)}{\LF(n,\varepsilon)}\ge\frac{\max\{1,T_n^2/(6\varepsilon+2D_n)\}}{\log_2\log_2n+8c_\varepsilon}.
\]
With Theorem~\ref{thm:mixf}(b) and the schedule of Proposition~\ref{prop:hfs}, which gives $\LH(n,\varepsilon)\le2+4(T_n+1)(3T_n+2)/\varepsilon$, for large $n$, $\LH(n,\varepsilon)/\LF(n,\varepsilon)\le(1+o(1))\,(24e^{\varepsilon}/\varepsilon)\,(\ln n)^2/\log_2\log_2n$, so
\[
\frac{\LH(n,\varepsilon)}{\LF(n,\varepsilon)}=\Theta_\varepsilon\Big(\frac{(\ln n)^2}{\log\log n}\Big).
\]
A single labelled schedule has $F=H$, so the best fixed labelled schedule needs $\Theta_\varepsilon((\ln n)^2)$ rounds for
$F\le\varepsilon$, whereas among single relabelled profiles Corollary~\ref{cor:cf1} gives
$R_F^{\mathrm{PROF}}(n,\varepsilon)=\log_2\log_2n+O_\varepsilon(1)$.
\end{corollary}

The ratio statement is two-sided over finite predrawn mixtures, because Theorem~\ref{thm:mixf} bounds $\LF$ from
below. For adaptive policies the lower rate for $F$ in expected executed rounds is Theorem~\ref{thm:fsadapt}, and no adaptive-policy lower rate for $H$ is claimed; $c_\varepsilon$ is astronomical and Theorem~\ref{thm:mixf} bites only at astronomical $n$ (likewise $K_\star(\varepsilon)$, Theorem~\ref{thm:mixu} and Theorem~\ref{thm:fsadapt}), so the separation is
qualitative (at practical $n$ the comparison is Table~\ref{tab:fs}), and since $F\le H$ a lower bound on $H$ gives none
on $F$. The explicit constructions above, the schedule of Proposition~\ref{prop:hfs} and the profile of Theorem~\ref{thm:cf1}, prove
the asymptotic statements only; at the sizes of Table~\ref{tab:fs} the least expected numbers of rounds are far below their lengths (computed $\LH$ optima, best-found upper bounds on $\LF$; float64).

\pend{}

\emph{Scope.} Theorem~\ref{thm:thmm} and Corollary~\ref{cor:thmm} concern one profile with a predrawn uniform
relabelling only: the gap set $G$ and Lemmas L4 and L5 all use a deterministic round $k$ with the deterministic grid
$\{x/c_k\}$ and a deterministic $c_{k+1}$. They do not cover mixtures of different profiles, where one component's
atoms fill another's gaps (for the profiles $(2,n-2)$ and $(6,n-6)$ and $n=5000$, the ratio $\rho_G$ of the proof of Theorem~\ref{thm:thmm} on the gap set of the first is
$1.7\cdot10^{-18}$ ($X\in\{1,2\}$) under it but $0.94$ under the equal mixture of the two). Also, at a fixed number $R$ of rounds the gain of mixing profiles over the best single profile, $1-F_{\mathrm{mix}}/F_{\mathrm{prof}}$,
grows with $n$ on the computed cells; here $F_{\mathrm{prof}}$ is the least $F$ of one uniformly relabelled profile with $R$ rounds
(exhaustive search over profiles) and $F_{\mathrm{mix}}$ the least $F$ of a finite predrawn mixture of such profiles (conditional
gradient, with a duality lower bound within $10^{-4}$ relative), computed in float64: at
$R=2$ from $0.7\%$ at $n=64$ to $27.9\%$ at $n=1024$; at $R=3$ from $0.04\%$ at $n=64$ to $7.7\%$ at $n=512$. Hence they give no lower bound on the mean-length
quantity $\LF$, whose feasible laws are mixtures that may combine profiles of different lengths; Theorem~\ref{thm:mixf}
gives one with a test set that does not depend on the component. They do not cover
value-adaptive policies: there is no deterministic level grid, and the forward induction of Lemma L3 fails for
state-dependent block sizes ($c=4$, $\nu=(\frac14,\frac12,\frac14)$ on $\{\frac14,\frac12,\frac34\}$, $\nu\le_{\mathrm{cx}}\nu'=\mathrm{Unif}\{\frac14,
\frac12,\frac34\}$, $b(\frac12)=7$, $b(\frac14)=b(\frac34)=1$ and $\varphi(v)=(\frac15-v)_+$ give
$\E_{\nu S}\varphi=1/14080>\E_{\nu' S}\varphi=1/21120$, with $S$ the resulting state-dependent sampler kernel). On
this target the lower bounds covering all policies are the last-round edge bound of Proposition L, which
constrains only the last round, and, for expected executed rounds and, through its part (a), hard caps,
Theorem~\ref{thm:fsadapt}.
Nor do they cover $\alpha,\beta\ne1$, where Lemma L4 and the target mass of $G$ change; no extension to them is attempted. The matching
upper bound of Theorem~\ref{thm:cf1} holds in the same class. These single-profile arguments do not themselves extend
to mixtures of profiles or to general seeded or value-adaptive policies. Separately, for every fixed $\varepsilon>0$, Theorem~\ref{thm:mixf} proves the order $\Theta_\varepsilon(\log\log n)$ for finite predrawn mixtures under expected cost, and Corollary~\ref{cor:mixexact} the exact constant, $\LF(n,\varepsilon)/\log_2\log_2n\to e^{-\varepsilon}/2$; the value-adaptive case is settled for the constant in expected executed rounds by Theorem~\ref{thm:fsadapt}, whose part (a), with the relabelled-profile upper bound of Theorem~\ref{thm:fs}(b), also gives the least hard cap the order $\Theta_\varepsilon(\log\log n)$, between $(\frac12-o(1))\log_2\log_2n$ and $\log_2\log_2n+O_\varepsilon(1)$ (after Theorem~\ref{thm:fsadapt}); the sharp hard-cap constant and $H$ for adaptive policies remain open.
The constants of Theorem~\ref{thm:thmm} are deliberately crude (Chebyshev everywhere); only their existence matters (see after
Corollary~\ref{cor:thmm}).

\section{Two-group survival envelope}\label{app:tgenv}
\pend{}
This appendix bounds, for the two-group tilted fixed-weight target, the probability that a causal randomized policy meets the quota within $r$ rounds (Lemma~TG9 and Theorem~\ref{thm:tgenv}), and derives lower bounds on the least hard cap, the mean cost and the expected executed rounds (Corollaries~(a)--(d)). The survival recursion $G^\lambda_r$ bounds that probability for every causal randomized policy (Lemma~TG9) and is attained by a deterministic count policy (Lemma~TG11); by induction on $r$, the one-step estimate of Lemma~TG6 and the Cauchy--Schwarz halving of Lemma~TG7 bound $G^\lambda_r$ by a negative power of $V(s)+1$ (Theorem~\ref{thm:tgenv}), where $V(s)$ is the variance of the number of ones drawn by one round that commits every free coordinate at the exact marginals. Since $\KL(p_\lambda\|Q)\ge-\ln Q(S_k)$, an upper bound on survival yields a lower bound on output KL; this appendix proves no upper bound matching its lower bounds, and its only achievability results are $F^*_n=0$ in~(a), the one-round counterexample in the second of Corollary~(c)'s two remarks, and Lemma~TG11 (see the Scope at the end). The constants do not depend on $\lambda$, which enters only through $V(s)$; at the root state $(n_A,n_B,k)$ this variance can be $o(1)$ even as $n\to\infty$, and the bounds give doubly logarithmic rates in $n$ when it is of order $n$ (Corollary~(c)). Its setting and results are used in Appendices~\ref{app:tgred} and~\ref{app:alt}, and in Appendix~\ref{app:t1}, whose \S\S5 and~9 derive from them, for two weight values, the lower bounds of Theorem~\ref{thm:tilt}(a) on $\LF^w$, on the expected executed rounds and on the hard cap (Section~\ref{sec:tilt}).

\emph{Setting.} \emph{Target.} There are two groups, A of size $n_A$ and B of size $n_B$, with $n=n_A+n_B\ge1$, a
total quota $0<k<n$, and a tilt $\lambda>0$. The slice is $S_k=\{x\in\{0,1\}^n:|x|=k\}$ (the set
written $V$ in the main text; in this appendix $V(s)$ is the variance defined below), where $|x|=\sum_ix_i$, and the target is $p_\lambda(x)\propto\lambda^{|x_A|}\cdot\mathbf 1\{x\in S_k\}$,
where $|x_A|$ is the number of ones of $x$ in group~A.

\emph{Prefixes and states.} A (labelled) prefix is a set of committed coordinates together with their values. Its
count state is $s=(m_A,m_B,j)$: the numbers of free (uncommitted) coordinates in each group, and the residual quota
$j=k-(\text{number of committed ones})$. The state $s$ is \emph{legal} iff $0\le j\le M:=m_A+m_B$. A legal state
always has positive target mass. Once a prefix is illegal, no completion lies in $S_k$.

\emph{Interface.} A round commits a batch $\beta$ of free coordinates. Empty batches are allowed and change nothing.
\emph{Legal prefix:} the values of $\beta$ are drawn as fresh Bernoulli bits at the exact conditional marginals of the
current prefix (Lemma~TG0). \emph{Fresh} means: conditionally on the complete pre-draw information $\mathcal F_t$ (defined under Policies below) and
on the selected batch, the bits are mutually independent with those marginals. Equivalently, the bits are realized
from fresh auxiliary randomness, e.g.\ $\mathrm{bit}_i=\mathbf 1\{\zeta_i\le a_i(s_t)\}$ with $\zeta_i$ uniform on
$[0,1]$, where $s_t$ is the current count state and $a_i(s_t)$ the marginal of the group of coordinate~$i$. This
auxiliary randomness is independent of the seed $\omega$ and of all earlier auxiliary randomness; the seed is drawn
before, and independently of, all of it. The bits themselves are generally \emph{not} independent of the past, because
their parameters $a_i(s_t)$ depend on the revealed prefix. Only the auxiliary randomness is independent.
\emph{Illegal prefix:} the values are drawn arbitrarily, e.g.\ as fair bits.

\emph{Policies.} A \emph{causal randomized policy} $\pi$ draws a seed $\omega$ (any law) and chooses each batch
$\beta_{t+1}$ as a measurable function of $\omega$ and the values committed through round~$t$. Write
$\mathcal F_t:=\sigma(\omega,\text{values through round }t)$. If $\pi$ stops before round~$r$, with all coordinates
committed or no further rounds, set $s_t:=s_{\mathrm{final}}$ for every later $t$, where $s_t$ is the count state after
$t$ rounds.
This convention is covered by the $M=0$ and empty-batch cases of Lemma~TG9. Stopping at an illegal prefix is covered too, since
$Y_t=0$ there ($Y_t$ as in the proof of Lemma~TG9). The policy $\pi$ has an \emph{every-path hard cap~$r$} if, on every
execution path, all $n$ coordinates are committed within $r$ rounds; $\mathrm{RAND}_r$ denotes this class. Here an
execution path is a path of positive probability under the seed's finite pushforward onto depth-$r$ policy tables and the
oracle's conditional kernels; a depth-$r$ policy table assigns a batch to each committed value history of fewer than $r$
rounds, there are finitely many, and the seed's law pushes forward to a finite law on them (Appendix~\ref{app:tgred}, proof of TGR3, step~(4)). So the cap means that all coordinates
are committed within $r$ rounds with probability one ($\Pr(M_r=0)=1$, with $M_r$ the number of free coordinates
after $r$ rounds), as in Section~\ref{sec:setup}: this is Appendix~\ref{app:lean}'s \texttt{CapAS}, not its stronger \texttt{CapEvery}, which also constrains paths of probability zero. So $\mathrm{RAND}_r$ is the class $\mathcal A_r$ of Section~\ref{sec:setup} on this target, with the same filtration: $\mathcal F_t$, the information after round $t$, chooses the batch of round $t+1$. A \emph{full
labelled schedule} $U=(\beta_1,\dots,\beta_L)$ is an ordered partition of $[n]$ into nonempty blocks, executed in order.
It is a seedless member of $\mathrm{RAND}_L$, and $L=|U|\ge1$. A \emph{predrawn mixture} $\mu=\sum_iw_i\delta_{U_i}$ draws
the index $i$ before any bit. The output laws are denoted $Q_\pi$, $q_U$ and $Q_\mu$; the success probability of an output law $Q$ is $Q(S_k)$.

\emph{Marginals and variance.} At a legal state, $a_A(s)$ and $a_B(s)$ denote the exact conditional marginals of a free coordinate in group~A and in group~B (Lemma~TG0). They
satisfy $m_Aa_A+m_Ba_B=j$. Define
\[
V(s):=m_Aa_A(1-a_A)+m_Ba_B(1-a_B).
\]
If $m_g=0$, then $a_g$ is undefined and every expression carries it with weight $m_g=0$.

\emph{Count kernel.} At a legal $s$ with $M\ge1$, a batch composition $b=(b_A,b_B)\ne0$ (the numbers of batch coordinates in groups A and~B) with $b_g\le m_g$ gives
independent committed counts $h_A\sim\Bin(b_A,a_A)$ and $h_B\sim\Bin(b_B,a_B)$. The child is
$s'=(m'_A,m'_B,j')=(m_A-b_A,m_B-b_B,j-h_A-h_B)$. It is \emph{alive} iff it is legal.
The variance of $X:=h_A+h_B$ is $v_b:=b_Aa_A(1-a_A)+b_Ba_B(1-a_B)$, and $\E_{\mathrm{prod}}$ denotes expectation under
these product draws.

\emph{Survival recursion (count DP).} On legal states, $G^\lambda_0(s)=\mathbf 1\{M=0\}$; for $r\ge1$,
$G^\lambda_r(s)=1$ if $M=0$, and otherwise
\[
G^\lambda_r(s)=\max_{b\ne0}\E_{\mathrm{prod}}\big[G^\lambda_{r-1}(s');\,s'\text{ alive}\big],
\]
where $\E[Y;\mathcal E]:=\E[Y\mathbf 1_{\mathcal E}]$ for an event $\mathcal E$.

\emph{Link.} Lemma~TG9 shows that the probability of success within $r$ rounds is $\le G^\lambda_r(\mathrm{root})$ for every causal
randomized policy, where $\mathrm{root}:=(n_A,n_B,k)$. Lemma~TG11 shows that equality is attained in $\mathrm{RAND}_r$ by
a deterministic count policy. At $\lambda=1$ the marginals are $j/M$, only $b_A+b_B$ matters,
and $G^\lambda_r$ is the balanced $G_r(M,j)$ of Section~\ref{sec:main} (paragraph ``The exact recursion'').

\begin{theorem}[two-group envelope]\label{thm:tgenv}
For every $\lambda>0$, all group sizes, every legal state $s$ and every $r\ge1$,
\[
G^\lambda_r(s)\le\min\big\{1,\ C_r\,(V(s)+1)^{-2^{-r}}\big\},\qquad C_r=\sqrt2\cdot18^{2-2^{2-r}}<324\sqrt2 .
\]
\end{theorem}
The constants do not depend on $\lambda$, the group sizes or the state. They are $C_1=\sqrt2$, $C_2=18\sqrt2$ and
$C_3=108$, and since $C_r/C_{r-1}=18^{2^{2-r}}\le18$ for $r\ge2$, the theorem also holds with the geometric constant
$\sqrt2\cdot18^{r-1}\ge C_r$, the form used in Corollary~(a), in the asymptotics of~(b) and in Appendix~\ref{app:t1}.
\emph{The bound is $\lambda$-uniform in form
only.} Its content enters through $V(s)$, which can be $o(1)$ even as $n\to\infty$: for example $n_A=n_B=k$ with
$\lambda=k^4$, where $V_{\mathrm{root}}=2/k^2+O(k^{-4})$ (Corollary~(c) below). So it is not a $\lambda$-uniform doubly-logarithmic lower bound.
\emph{Comparison with the balanced envelope.} At $\lambda=1$ and the balanced root ($n=2k$, $V=k/2$), the bound has the
shape of the balanced envelope \eqref{eq:env}, $A_r(k+1)^{-2^{-r}}$ with $A_r=6^{2-2^{2-r}}$ (Appendix~\ref{app:env}). The
constant is larger, $C_r/A_r=\sqrt2\cdot3^{2-2^{2-r}}\in[\sqrt2,9\sqrt2)$, and the base is $(k/2+1)$ in place of $(k+1)$, a factor $\le2^{2^{-r}}$. \emph{The
constants are very loose.} The largest \emph{observed} values of the effective constant $\max_sG^\lambda_r(s)\,(V(s)+1)^{2^{-r}}$, over the scanned
states $s$, are $1.1404$ at $r=3$ and $1.1176$ at $r=4$ ($n\le80$, extreme $\lambda$, still rising with size), and $1.487$ for the
one-step $T(s,b)\,(1+V(s))^{1/4}$ of Lemma~TG6 ($n=600$, $\lambda=10^{\pm4}$), with a $\lambda=1$ Gaussian-limit value of
$1.4055$. These are finite-grid observations, not suprema. So $C_r$ appears very loose: about
$22\times$ at $r=2$ ($C_2=\sqrt2\cdot18\approx25.46$ against the observed $1.136$, the largest $r=2$ value in the same scan ($n\le80$, extreme $\lambda$),
attained at a small state). That is an
empirical regularity, not a theorem.
\emph{Plan of the lemmas.} Lemma~TG0 gives the target's conditional law and the sampler's count law, the count kernel above. Lemmas~TG1--TG7 serve the proof of Theorem~\ref{thm:tgenv}, given after Lemma~TG11: the case $r=1$ is the atom bound of Lemma~TG1, and the step to $r\ge2$ combines the one-step bound of Lemma~TG6 with the halving of Lemma~TG7. Case~F of Lemma~TG6 uses Lemmas~TG1 and~TG3, and Case~S uses Lemma~TG5, which rests on Lemmas~TG2 and~TG4. Lemma~TG8 is needed only for Corollary~(c). Lemma~TG9, with Lemma~TG10, bounds the probability of success within $r$ rounds of every causal randomized policy by $G^\lambda_r(\mathrm{root})$; this is how the Corollaries use the Theorem. Lemma~TG11 (attainment) is needed by neither.

\emph{Lemma TG0 (conditional law).} Let a prefix have legal count state $(m_A,m_B,j)$. Such a prefix has positive
$p_\lambda$-mass, so conditioning on it is legitimate. \emph{Target facts} (properties of $p_\lambda$): (i)~under
$p_\lambda$, the conditional law of the free coordinates $y$ is $\propto\lambda^{|y_A|}\cdot\mathbf 1\{|y|=j\}$, where
$|y_A|$ is the number of ones of $y$ in group~A; (ii)~every free coordinate of group $g$ has the same conditional
marginal $a_g$; (iii)~given any batch values of positive target probability (equivalently, values leaving a legal
child), the remaining free coordinates have the law (i) at the child state. Under $p_\lambda$ the free coordinates are
in general \emph{not} independent. They are dependent whenever $M\ge2$ and $0<j<M$, and always nonpositively correlated
(Lemma~TG4(ii)). At $j\in\{0,M\}$, or with a single free coordinate, the law is a point mass or a single Bernoulli,
hence trivially independent. \emph{Interface fact} (a property of the sampler, not of $p_\lambda$): (iv)~conditionally on
$\mathcal F_t$ and the selected batch of composition $(b_A,b_B)$, the interface's committed counts are
$\Bin(b_A,a_A)\otimes\Bin(b_B,a_B)$, whichever coordinates the batch contains. This follows from (ii) and the
definition of fresh independent draws.

\emph{Proof.} (1) Let $u_A$ be the number of committed A-ones and $c$ the number of committed ones. For a
completion $y$,
\[
p_\lambda(\text{prefix},y)\propto\lambda^{u_A+|y_A|}\cdot\mathbf 1\{c+|y|=k\}\propto\lambda^{|y_A|}\cdot\mathbf 1\{|y|=j\}.
\]
This gives (i). (2) The law (i) is invariant under permutations within each group, which gives (ii). (3)
Conditioning on batch values of positive probability is the same computation with a longer, still legal prefix,
which gives (iii). (4) Property (iv) is the interface definition combined with (ii). \qed

\emph{Lemma TG1 (Poisson-binomial anti-concentration).} Let $X=\sum_i\xi_i$ with independent
$\xi_i\sim\mathrm{Bernoulli}(q_i)$, and let $v=\sum_iq_i(1-q_i)$. Then
\[
\max_x\Pr(X=x)\le\sqrt2\cdot(1+v)^{-1/2},\qquad\text{and for $v>0$ also}\qquad\max_x\Pr(X=x)\le\sqrt{\pi/(8v)}.
\]

\emph{Proof.} If $v=0$, $X$ is deterministic and $\max_x\Pr(X=x)=1\le\sqrt2$. Now let $v>0$. (1) By Fourier
inversion, $\Pr(X=x)\le\frac1{2\pi}\int_{-\pi}^{\pi}|\varphi_X(t)|\,dt$, where $\varphi_X(t)=\E e^{itX}$. (2) For each
factor, $|1-q+qe^{it}|^2=1-2q(1-q)(1-\cos t)\le\exp(-2q(1-q)(1-\cos t))$. Hence $|\varphi_X(t)|\le\exp(-v(1-\cos t))$.
(3) For $|t|\le\pi$, $1-\cos t=2\sin^2(t/2)\ge2t^2/\pi^2$ (Jordan). So
$\Pr(X=x)\le\frac1{2\pi}\int_{\mathbb R}e^{-2vt^2/\pi^2}\,dt=\sqrt{\pi/(8v)}$. (4) For the first inequality: if
$v\ge1$, then $\sqrt{\pi/(8v)}\le\sqrt2(1+v)^{-1/2}$, since $\pi(1+v)/(8v)\le\pi/4\le2$; if $0<v<1$, then
$\Pr\le1\le\sqrt2(1+v)^{-1/2}$. \qed

\emph{Lemma TG2 (monotone marginals).} Fix a composition $(m_A,m_B)$ and write $a_g(x)$ for the marginal at quota $x$.
Then $a_A(x)$ and $a_B(x)$ are nondecreasing in $x\in\{0,\dots,M\}$. Consequently, for $x\le y$,
\[
m_A\big(a_A(y)-a_A(x)\big)+m_B\big(a_B(y)-a_B(x)\big)=y-x,\qquad\text{with both summands $\ge0$.}
\]

\emph{Proof.} (1) Let $U_A$ be the number of ones among the free A-coordinates. Given quota $x$,
$\Pr(U_A=u)\propto\binom{m_A}{u}\binom{m_B}{x-u}\lambda^u$. (2) The zero-extended row $\binom{m_B}{\cdot}$ is log-concave
with interval support, so $\binom{m_B}{x-u}$ is TP2 in $(x,u)$. Hence the law of $U_A$ is, in the likelihood-ratio order,
increasing in $x$, and $\E[U_A\mid x]=m_Aa_A(x)$ is nondecreasing. (3) The same argument applies to $W=x-U_A$, whose law
is $\propto\binom{m_B}{w}\binom{m_A}{x-w}\lambda^{x-w}$. The factor $\lambda^{x-w}=\lambda^x\lambda^{-w}$ preserves
TP2. This gives the claim for $a_B$. (4) The monotonicity is the two-block case of \citet{efron1965increasing}. The displayed consequence follows from it and the identity
$m_Aa_A(x)+m_Ba_B(x)=x$. \qed

\emph{Lemma TG3 (level set).} At every legal state, $\mathrm{dist}(j,D)\le2V(s)$, where $D=\{0,m_A,m_B,m_A+m_B\}$.

\emph{Proof.} (1) For $a\in[0,1]$, $a(1-a)\ge\frac12\min(a,1-a)$. (2) Let $e_g=\mathbf 1\{a_g>\frac12\}$. Then
$|j-\sum_gm_ge_g|=|\sum_gm_g(a_g-e_g)|\le\sum_gm_g\min(a_g,1-a_g)\le2V$. (3) Finally, $\sum_gm_ge_g\in D$. \qed

Lemmas~TG4 and~TG5 serve Case~S of Lemma~TG6, in which the batch carries less than half of $V(s)$. Together they bound the variance of a child from below by the remaining variance $V(s)-v_b$, up to the distance of the child quota from its mean $\mu'$ and a $\sqrt{v_b}$ correction. Lemma~TG4 is stated under the target law $p_s$ of the batch values, not under the product law of the count kernel, because its part~(i), that the child marginals average back to the parent marginals, holds under $p_s$; part~(ii) shows that under $p_s$ the child quota varies no more than under the product law, where its variance is $v_b$. Lemma~TG5 turns these facts into a deterministic bound at every legal child quota, which Case~S applies under the product law.

\emph{Lemma TG4 (martingale; negative correlation).} Fix a legal parent $s$ with $M\ge1$ and a batch $b$, and let $p_s$
be the \emph{target} law of the batch values at $s$. Write $a'_g(x)$ for the marginal of group $g$ at the child
composition $(m'_A,m'_B)$ and quota $x$. Then: (i)~$\E_{p_s}[a'_g(j')]=a_g$ whenever $m'_g\ge1$, and
$\E_{p_s}[j']=\mu':=m'_Aa_A+m'_Ba_B$; (ii)~$\mathrm{Var}_{p_s}(j')\le v_b$.

\emph{Proof.} (i) Take a free coordinate $i$ of group $g$ outside the batch. Then
$a_g=\Pr(y_i=1)=\E_{p_s}[\Pr(y_i=1\mid\text{batch values})]$, and by Lemma~TG0 the inner probability is $a'_g(j')$.
Also $\E_{p_s}[j']=j-b_Aa_A-b_Ba_B=\mu'$.

(ii) The target on the free coordinates is $\Pr(y)\propto\prod_iw_i^{y_i}\cdot\mathbf 1\{|y|=j\}$, with $w_i=\lambda$ on
A and $1$ on B. (1) For $i\ne l$, let $R$ be the other free coordinates, and let $e_t$ denote elementary symmetric
polynomials of the weights. Use the convention $e_t=0$ for $t<0$ or $t>|R|$. Then
\[
\mathrm{Cov}(y_i,y_l)=w_iw_l\big[e_{j-2}(R)\,e_j(R\cup\{i,l\})-e_{j-1}(R\cup\{l\})\,e_{j-1}(R\cup\{i\})\big]\big/e_j(R\cup\{i,l\})^2.
\]
(2) Since $e_t(R\cup\{l\})=e_t(R)+w_le_{t-1}(R)$ and $e_t(R\cup\{i,l\})=e_t(R)+(w_i+w_l)e_{t-1}(R)+w_iw_le_{t-2}(R)$, the
bracket is $e_{j-2}[e_j+(w_i+w_l)e_{j-1}+w_iw_le_{j-2}]-[e_{j-1}+w_le_{j-2}][e_{j-1}+w_ie_{j-2}]=e_{j-2}e_j-e_{j-1}^2$, with
every $e$ taken over $R$: the cross terms cancel. This is $\le0$: for $1\le j-1\le N-1$, $N=|R|$, by Newton's inequality
$e_{j-1}^2\ge e_{j-2}e_j\binom N{j-1}^2/(\binom N{j-2}\binom Nj)$ together with $\binom N{j-1}^2\ge\binom N{j-2}\binom Nj$; if
$j-1\in\{0,N\}$, one of $e_{j-2},e_j$ vanishes and the bracket is $-e_{j-1}^2\le0$; outside $0\le j-1\le N$ every term is $0$. (3) Hence
$\mathrm{Var}_{p_s}(\sum_{i\in\text{batch}}y_i)\le\sum_{i\in\text{batch}}a_i(1-a_i)=v_b$, with $a_i:=a_g$ for $i$ in group
$g$. (4) At the boundary quotas $j\in\{0,M\}$, all marginals are $0$ or $1$ and the target is a point mass, so (i) and
(ii) hold trivially, with $\mathrm{Var}_{p_s}=v_b=0$. (5) Other routes to the nonpositive covariances: negative association of independent PF$_2$ variables
conditioned on their sum \citep[Thm.~2.8]{joagdev1983negative}, or the strong Rayleigh property \citep{borcea2009negative}.
\qed

\emph{Lemma TG5 (child variance near the mean point).} In the setting of Lemma~TG4, let
$V_{\mathrm{rem}}:=m'_Aa_A(1-a_A)+m'_Ba_B(1-a_B)=V(s)-v_b$. Then for every legal child quota $x$,
\[
V(m'_A,m'_B,x)\ge V_{\mathrm{rem}}-|x-\mu'|-\sqrt{v_b}.
\]

\emph{Proof.} (1) Extend each $a'_g$ piecewise-linearly to real $y\in[0,M']$, $M':=m'_A+m'_B$. By Lemma~TG2 at the child composition $(m'_A,m'_B)$, each $a'_g$ is nondecreasing on $\{0,\dots,M'\}$ and $m'_Aa'_A(x)+m'_Ba'_B(x)=x$ (the identity used in its proof); so the extensions are
nondecreasing and satisfy $m'_Aa'_A(y)+m'_Ba'_B(y)=y$. So for real $y$ and $z$ both differences have the sign of
$z-y$, and
\[
\sum_gm'_g|a'_g(z)-a'_g(y)|=|z-y|.
\]
Note that $\mu'\in[0,M']$. (2) By Lemma~TG4(i) and Jensen,
\[
\sum_gm'_g|a_g-a'_g(\mu')|\le\E_{p_s}\sum_gm'_g|a'_g(j')-a'_g(\mu')|=\E_{p_s}|j'-\mu'|\le\sqrt{v_b}.
\]
The last step is Lemma~TG4(ii) and Cauchy--Schwarz. Groups with $m'_g=0$ carry weight $0$. Under $p_s$, $j'\in[0,M']$
always. (3) $f(a)=a(1-a)$ is $1$-Lipschitz on $[0,1]$. So
\[
|V(m'_A,m'_B,x)-V_{\mathrm{rem}}|\le\sum_gm'_g|a'_g(x)-a_g|\le|x-\mu'|+\sqrt{v_b}.\qquad\text{\qedsymbol}
\]

\emph{Note on the $\sqrt{v_b}$ term.} Some additive correction is needed: without one, the bound fails. The exact
witnesses are below, where the gap $V_{\mathrm{rem}}-|x-\mu'|-V(m'_A,m'_B,x)$ is the amount by which the bound without
$\sqrt{v_b}$ fails.
\begin{center}\small
\begin{tabular}{@{}cccccl@{}}
\toprule
witness & $\lambda$ & $s$ & $b$ & $x$ & gap\\
\midrule
1 & $1/4$ & $(2,2,2)$ & $(1,1)$ & $1$ & $232/3025$\\
2 & $1/50$ & $(5,5,5)$ & $(2,2)$ & $3$ & $0.2262520345\cdot\sqrt{v_b}$\\
3 & $1/100$ & $(25,4,5)$ & $(18,0)$ & $4$ & $0.2738280224\cdot\sqrt{v_b}$\\
\bottomrule
\end{tabular}
\end{center}
In witness~1, $(a_A,a_B)=(3/11,8/11)$, $V_{\mathrm{rem}}=v_b=48/121$, $\mu'=1$ and $V(1,1,1)=8/25$. None of the witnesses shows
that $\sqrt{v_b}$ is the necessary \emph{order} of the correction.

\emph{Lemma TG6 (one step).} For every legal $s$ with $M\ge1$ and every batch $b$,
\[
T(s,b):=\E_{\mathrm{prod}}\big[(V(s')+1)^{-1/2};\,s'\text{ alive}\big]\le\min\big\{1,\ 18\cdot(V(s)+1)^{-1/4}\big\}.
\]

\emph{Proof.} Write $V=V(s)$. Under the product law, $\E_{\mathrm{prod}}j'=\mu'$ and $\mathrm{Var}_{\mathrm{prod}}j'=v_b$,
and $X=j-j'$ is Poisson-binomial. The bound $T\le1$ is trivial. For the second bound, the proof splits on whether the batch carries at least half of the variance $V$ (Case~F) or less (Case~S). Write
$C_F:=(1+\sqrt{3/2})\cdot\sqrt{32}=4(\sqrt2+\sqrt3)\approx12.585$.

\emph{Case F: $v_b\ge V/2$.} (1) By Lemma~TG3 at the child, $V(s')\ge\mathrm{dist}(j',D')/2$, with
$D'=\{0,m'_A,m'_B,M'\}$. For each $t\ge0$, at most $8$ integers $x$ have $\mathrm{dist}(x,D')=t$. (2) By Lemma~TG1,
$\Pr_{\mathrm{prod}}(j'=x)\le \chi:=\sqrt2(1+v_b)^{-1/2}$. Split the alive children according to whether the distance of $j'$ from $D'$ is below an integer $t_0\ge0$, and use
$\sum_{t<t_0}(1+t/2)^{-1/2}\le4\sqrt{1+t_0/2}$:
\[
T\le\sum_{t<t_0}8\chi(1+t/2)^{-1/2}+(1+t_0/2)^{-1/2}\le32\chi\cdot\sqrt{1+t_0/2}+(1+t_0/2)^{-1/2}.
\]
(3) We show $T\le C_F\cdot\sqrt \chi$. If $32\chi\ge1$, then $C_F\sqrt \chi=(1+\sqrt{3/2})\sqrt{32\chi}\ge1+\sqrt{3/2}>1\ge T$.
Otherwise, let $z^*=1/(32\chi)\ge1$ and pick $t_0$ with $1+t_0/2\in[z^*,z^*+\frac12]\subset[z^*,1.5z^*]$. This gives
$T\le(1+\sqrt{3/2})\cdot\sqrt{32\chi}=C_F\cdot\sqrt \chi$. (4) Since $1+v_b\ge(1+V)/2$, $\sqrt \chi\le\sqrt2\cdot(1+V)^{-1/4}$.
Hence
\[
T\le\sqrt2\cdot C_F\cdot(1+V)^{-1/4}=(8+4\sqrt6)\cdot(1+V)^{-1/4}<18\cdot(1+V)^{-1/4},
\]
where $8+4\sqrt6\approx17.798$.

\emph{Case S: $v_b<V/2$.} (1) Here $V_{\mathrm{rem}}>V/2$ and $\sqrt{v_b}<\sqrt{V/2}$. (2) On
$E=\{|j'-\mu'|\le V_{\mathrm{rem}}/2\}$, Lemma~TG5 gives $V(s')\ge V_{\mathrm{rem}}/2-\sqrt{v_b}>V/4-\sqrt{V/2}$ at every
alive child. (Lemma~TG5 is a pointwise statement about the child quota, so applying it under the product law is
valid.) (3) By Chebyshev under the product law, $\Pr(E^c)\le4v_b/V_{\mathrm{rem}}^2<8/V$. (4) If $V\ge32$, then
$V/4-\sqrt{V/2}\ge V/8$, so
\[
T\le(1+V/8)^{-1/2}+8/V=g(V)\cdot(1+V)^{-1/4}\le1.68\cdot(1+V)^{-1/4},
\]
where $g(V):=(1+V)^{1/4}\big[(1+V/8)^{-1/2}+8/V\big]=g_1+g_2$, with $g_1:=(1+V)^{1/4}(1+V/8)^{-1/2}$ and
$g_2:=8(1+V)^{1/4}/V$. Here $(\ln g_1)'=1/(4(1+V))-1/(2(8+V))=-(V-6)/(4(V+1)(V+8))$, which is $<0$ for $V>6$, and
$(\ln g_2)'=-(3V+4)/(4V(V+1))<0$. So $g$ decreases on $V>6$, and
$\max_{V\ge32}g=g(32)=33^{1/4}(5^{-1/2}+\frac14)=1.6710688\ldots<1.68$. (5) If $V<32$, then
$18(1+V)^{-1/4}>18\cdot33^{-1/4}>7.5>1\ge T$. \qed

\emph{Lemma TG7 (Cauchy--Schwarz halving).} Let $T_\alpha(s,b):=\E_{\mathrm{prod}}[(V(s')+1)^{-\alpha};\,\text{alive}]$.
The alive mass is $\le1$, so Cauchy--Schwarz gives $T_\alpha^2\le T_{2\alpha}$. Iterating,
$T_{2^{-(r-1)}}\le T_{1/2}^{2^{2-r}}$ for $r\ge2$.

\emph{Lemma TG8 (variance vs.\ the tilted product).} Let $s$ be legal with $0<j<M$. Let
$(\theta_A,\theta_B)\in(0,1)^2$ be the unique pair with $\theta_A(1-\theta_B)/((1-\theta_A)\theta_B)=\lambda$ and
$m_A\theta_A+m_B\theta_B=j$. It exists because $t\mapsto m_A\cdot\lambda t/(1-t+\lambda t)+m_B\cdot t$ increases
continuously and strictly from $0$ to $M$. Let $\sigma^2:=m_A\theta_A(1-\theta_A)+m_B\theta_B(1-\theta_B)$. Then
\[
\sum_gm_g|a_g-\theta_g|\le\sigma\qquad\text{and}\qquad V(s)\ge\sigma^2-\sigma.
\]

\emph{Proof.} (1) Take independent group counts with law $\Bin(m_A,\theta_A)\otimes\Bin(m_B,\theta_B)$, and let $S$ be their total. Given $S=x$, the law
of the A-count is $\propto\binom{m_A}{u}\binom{m_B}{x-u}\lambda^u$, the law of $U_A$ in the proof of Lemma~TG2. So
$g_g(x):=\E[\text{count in }g\mid S=x]=m_ga_g(x)$. (2) $\E S=j$ and $\mathrm{Var}\,S=\sigma^2$. (3) By Lemma~TG2,
$\sum_g|g_g(S)-g_g(j)|=|S-j|$. Hence
\[
\sum_gm_g|\theta_g-a_g|=\sum_g|\E g_g(S)-g_g(j)|\le\E|S-j|\le\sigma.
\]
(4) $f(a)=a(1-a)$ is $1$-Lipschitz on $[0,1]$, since $|f'|=|1-2a|\le1$. So
$V\ge\sum_gm_gf(\theta_g)-\sum_gm_g|a_g-\theta_g|\ge\sigma^2-\sigma$. \qed

\emph{Lemma TG9 (success bound for causal policies).} Let $\pi$ be any causal randomized policy, and let $s_t$ be the
count state after $t$ rounds, with $M_t$ its number of free coordinates. Then for every $r\ge0$,
\[
\Pr(s_r\text{ is legal and }M_r=0)\le G^\lambda_r(\mathrm{root}).
\]
In particular, if $\pi\in\mathrm{RAND}_r$, then $Q_\pi(S_k)\le G^\lambda_r(\mathrm{root})$. This holds because, under an
every-path hard cap $r$, the output lies in $S_k$ iff $s_r$ is legal with $M_r=0$.

\emph{Proof.} Let $Y_t:=G^\lambda_{r-t}(s_t)\cdot\mathbf 1[s_t\text{ legal}]$ for $0\le t\le r$. Then
$Y_0=G^\lambda_r(\mathrm{root})$ and $Y_r=\mathbf 1[s_r\text{ legal},\,M_r=0]$. We show $\E[Y_{t+1}\mid\mathcal F_t]\le
Y_t$, by cases; that is, $Y_t$ is a supermartingale. If $s_t$ is illegal, then $s_{t+1}$ is illegal, so $Y_{t+1}=0=Y_t$. Illegality is absorbing whatever
the draws, because the batch's one-counts satisfy $0\le h_g\le b_g$ (here $b:=b_A+b_B$ and $h:=h_A+h_B$): if $j<0$,
then $j'=j-h\le j<0$; if $j>M$, then $j'-M'=(j-M)+(b-h)\ge j-M>0$. If $s_t$ is legal with $M_t=0$, then $s_{t+1}=s_t$
and $Y_{t+1}=1=Y_t$. If $s_t$ is legal with $M_t\ge1$ and the $\mathcal F_t$-measurable batch is empty, then
$Y_{t+1}=G^\lambda_{r-t-1}(s_t)\le G^\lambda_{r-t}(s_t)=Y_t$ by Lemma~TG10. If $s_t$ is legal with $M_t\ge1$ and the
batch has composition $b\ne0$ ($\mathcal F_t$-measurable), then given $\mathcal F_t$ the counts are
$\Bin(b_A,a_A(s_t))\otimes\Bin(b_B,a_B(s_t))$ by Lemma~TG0(iv). So
\[
\E[Y_{t+1}\mid\mathcal F_t]=\E_{\mathrm{prod}}\big[G^\lambda_{r-t-1}(s');\,\text{alive}\big]\le G^\lambda_{r-t}(s_t)=Y_t,
\]
by the definition of $G^\lambda$ as a maximum over $b$. Hence
$\Pr(s_r\text{ legal},\,M_r=0)=\E Y_r\le\E Y_0=G^\lambda_r(\mathrm{root})$. Only the batch composition enters, never
which coordinates are chosen. \qed

(At $\lambda=1$ and $n=2k$, Lemma~TG9 is the supermartingale argument for Lemma~\ref{lem:surv}(a) in
Appendix~\ref{app:conv}.)

\emph{Lemma TG10 (monotone horizon).} $G^\lambda_r\le G^\lambda_{r+1}$ at every legal state, for every $r\ge0$.

\emph{Proof.} \emph{$r=0$.} If $M=0$, both sides equal $1$. If $M\ge1$, $G^\lambda_0=0\le G^\lambda_1$. \emph{Induction step.}
Suppose $G^\lambda_{r-1}\le G^\lambda_r$ pointwise. For $M\ge1$,
\[
G^\lambda_r(s)=\max_b\E_{\mathrm{prod}}\big[G^\lambda_{r-1}(s');\,\text{alive}\big]\le\max_b\E_{\mathrm{prod}}\big[G^\lambda_r(s');\,\text{alive}\big]=G^\lambda_{r+1}(s).
\]
For $M=0$, both sides equal $1$. \qed

\emph{Lemma TG11 (attainment).} For every $r\ge1$ there is a deterministic, seedless policy $\pi^\circ\in\mathrm{RAND}_r$
with $Q_{\pi^\circ}(S_k)=G^\lambda_r(\mathrm{root})$. Hence
\[
\max_{\pi\in\mathrm{RAND}_r}Q_\pi(S_k)=G^\lambda_r(\mathrm{root}).
\]
The maximum is attained. Together with Lemma~TG9, the supremum over all causal randomized policies of the probability
of success within $r$ rounds also equals $G^\lambda_r(\mathrm{root})$.

\emph{Proof.} (1)~\emph{The policy.} For $1\le\varrho\le r$ and every legal $s$ with $M\ge1$, fix a maximizer
$b^*(\varrho,s)$ in the recursion for $G^\lambda_\varrho(s)$; the maximum is over a finite set. At round $t+1$, for
$0\le t<r$, the policy $\pi^\circ$ commits $b^*(r-t,s_t)$ if $s_t$ is legal with $M_t\ge1$ (the lowest-indexed free
coordinates: $b^*_A$ of group~A and $b^*_B$ of group~B); it commits all remaining coordinates if $s_t$ is illegal; and it
does nothing if $M_t=0$. (2)~\emph{Success.} Under $\pi^\circ$, $\E[Y_{t+1}\mid\mathcal F_t]=Y_t$ with $Y_t$ as in the
proof of Lemma~TG9; that is, the supermartingale is a martingale. On legal $s_t$ with $M_t\ge1$ the counts are
$\Bin(b^*_A,a_A(s_t))\otimes\Bin(b^*_B,a_B(s_t))$ (Lemma~TG0(iv)), and $b^*$ attains the maximum, so
$\E_{\mathrm{prod}}[G^\lambda_{r-t-1}(s');\,\text{alive}]=G^\lambda_{r-t}(s_t)$. If $s_t$ is illegal, $\pi^\circ$ commits all remaining coordinates. Illegality is absorbing (Lemma~TG9), so
$s_{t+1}$ is illegal and $Y_{t+1}=0=Y_t$. If $s_t$ is legal with $M_t=0$, then $s_{t+1}=s_t$ and $Y_{t+1}=1=Y_t$. The
policy $\pi^\circ$ never uses an empty batch at a legal state with $M_t\ge1$, so Lemma~TG9's empty-batch case does not
arise. Hence $\Pr(s_r\text{ legal},\,M_r=0)=\E Y_r=Y_0=G^\lambda_r(\mathrm{root})$; under the hard cap of
step~(3), this is $Q_{\pi^\circ}(S_k)$. (3)~\emph{Every-path hard cap~$r$.} Let $s$ be legal with $M\ge1$. If $0<j<M$,
every group with $m_g\ge1$ has $a_g(s)\in(0,1)$, since a free coordinate can be $1$ ($j\ge1$) or $0$ ($j\le M-1$); so
$X_{\mathrm{full}}$, the count $X$ of the full batch, has full support on $\{0,\dots,M\}$. If $j\in\{0,M\}$, then $X_{\mathrm{full}}=j$ almost surely.
Either way $G^\lambda_1(s)=\Pr_{\mathrm{prod}}(X_{\mathrm{full}}=j)>0$. Every batch $b$ other than the full one leaves
$M'\ge1$, where $G^\lambda_0=0$, so its value is $0$. Hence $b^*(1,s)$ is the full batch, and every path that is legal
before round $r$ completes in round $r$. A path that first becomes illegal after round $t<r$ completes in round
$t+1\le r$, and a legal path with $M_t=0$ is already complete. So every path commits all $n$ coordinates within $r$
rounds, and $\pi^\circ\in\mathrm{RAND}_r$. \qed

(This is the two-group analogue of the attainment in Theorem~\ref{thm:conv}(a).)
\emph{Proof of Theorem~\ref{thm:tgenv}.} Induction on $r$, with $C_1=\sqrt2$ and $C_r=18^{2^{2-r}}C_{r-1}$ for $r\ge2$, so
$C_r=\sqrt2\cdot18^{\sum_{i=2}^r2^{2-i}}=\sqrt2\cdot18^{2-2^{2-r}}$, which is at least $1$ and less than $\sqrt2\cdot18^2=324\sqrt2$.

\emph{$r=1$.} If $M\ge1$, only the full batch reaches $M'=0$, and $G^\lambda_0=0$ elsewhere. So
$G^\lambda_1(s)=\Pr_{\mathrm{prod}}(X_{\mathrm{full}}=j)$, where $X_{\mathrm{full}}$, the count $X$ of the full batch, has
variance $V(s)$. Lemma~TG1 gives $G^\lambda_1(s)\le\sqrt2(1+V)^{-1/2}$. If $M=0$, then $V=0$, and the bound $\sqrt2$ is
$\ge1=G^\lambda_1$.

\emph{$r\ge2$.} The induction hypothesis holds at every alive child. This includes children with $M'=0$, where $V=0$
and $G^\lambda_{r-1}\le1\le C_{r-1}$. Applying the induction hypothesis inside the recursion gives the first line below, and Lemmas~TG7 and~TG6 give the second:
\begin{align*}
G^\lambda_r(s)&\le C_{r-1}\cdot\max_bT_{2^{-(r-1)}}(s,b)\\
&\le C_{r-1}\cdot\big(18(V+1)^{-1/4}\big)^{2^{2-r}}\\
&=18^{2^{2-r}}C_{r-1}\cdot(V+1)^{-2^{-r}}=C_r\,(V+1)^{-2^{-r}}.
\end{align*}
The last step uses $\frac14\cdot2^{2-r}=2^{-r}$. For $M=0$, $G^\lambda_r=1\le C_r$. \qed

\emph{Corollaries (two-group terminal lower bounds).} Write $V_{\mathrm{root}}:=V(n_A,n_B,k)$. Throughout,
$\KL(p_\lambda\|Q)\ge-\ln Q(S_k)$ for every sub-probability $Q$, by Jensen on $S_k$, since $p_\lambda(S_k)=1$.

\emph{(a) Hard cap, class $\mathrm{RAND}_r$.} Assume $V_{\mathrm{root}}\ge2$, so that $L_V$ below is $>0$. Basic facts
about the class and about $F^*_r$ (defined below): $\mathrm{RAND}_0=\varnothing$ for $n\ge1$, so $F^*_0=+\infty$; $\mathrm{RAND}_r\subseteq\mathrm{RAND}_{r+1}$,
so $F^*_r$ is nonincreasing; and $F^*_n=0$, because committing one coordinate per round at the exact marginals samples
$p_\lambda$ exactly. Hence, for every $\varepsilon\ge0$, $R^\lambda_F(\varepsilon)$ below is a well-defined integer in $[1,n]$, and only $r\ge1$ is
needed, where Theorem~\ref{thm:tgenv} applies. The quantifier is over $\mathrm{RAND}_r$ only: causal randomized policies
with fresh exact-marginal draws and an every-path hard cap $r$. This is not a statement about arbitrary $r$-round
algorithms. By the preamble and Lemma~TG9, every $\pi\in\mathrm{RAND}_r$ has
$\KL(p_\lambda\|Q_\pi)\ge-\ln G^\lambda_r(\mathrm{root})$. Hence, by Theorem~\ref{thm:tgenv} with the geometric constant $\sqrt2\cdot18^{r-1}$,
$F^*_r:=\inf_{\pi\in\mathrm{RAND}_r}\KL(p_\lambda\|Q_\pi)$ satisfies
\[
F^*_r\ge2^{-r}\ln(V_{\mathrm{root}}+1)-\tfrac12\ln2-(r-1)\ln18.
\]
Let $L_V:=\log_2\ln(V_{\mathrm{root}}+1)$. If $F^*_r\le\varepsilon$, then either $r>L_V$ or
$r\ge L_V-\log_2(\varepsilon+\frac12\ln2+L_V\ln18)$. So
\[
R^\lambda_F(\varepsilon):=\min\{r:F^*_r\le\varepsilon\}\ge L_V-\log_2L_V-O_\varepsilon(1).
\]
Here $O_\varepsilon(1)$ is a constant depending only on $\varepsilon$, given the standing hypothesis
$V_{\mathrm{root}}\ge2$. That hypothesis gives $L_V\ge\log_2\ln3\approx0.136$, which controls both the branch $r>L_V$ and
the term $\log_2L_V$. Corollary~(d) below gives the sharper $R^\lambda_F(\varepsilon)>L_V-\log_2(\varepsilon+\ln(324\sqrt2))$ without this
hypothesis, from the bounded prefactor $C_r<324\sqrt2$; (a) records what the geometric form $C_r\le\sqrt2\cdot18^{r-1}$ alone gives.

\emph{(b) Mean cost, predrawn mixtures of full labelled schedules.} Define
$\LF^\lambda(n,\varepsilon):=\inf\{C(\mu):\KL(p_\lambda\|Q_\mu)\le\varepsilon\}$, where $\mu=\sum_iw_i\delta_{U_i}$ ranges
over predrawn mixtures of full labelled schedules and $C(\mu)=\sum_iw_i|U_i|$. This is the two-group analogue of the
least scheduled mean under the $F$ criterion, $\LF(n,\varepsilon)$ of Section~\ref{sec:setup}. Under mean cost, the class excludes adaptive policies and adaptive stopping, so (b) does not cover them; Corollary~(d) below extends its lower bound to every causal policy, adaptive stopping included, under expected executed rounds. Adaptive stopping under a hard cap that holds almost surely is covered by~(a).
\emph{Markov split.} Fix $r\ge0$. For each component $U$ of $\mu$: if $|U|\le r$, then
$s_U:=q_U(S_k)\le G^\lambda_{|U|}(\mathrm{root})\le G^\lambda_r(\mathrm{root})$, by Lemma~TG9 ($U\in\mathrm{RAND}_{|U|}$)
and Lemma~TG10, so $1+r(s_U-G^\lambda_r(\mathrm{root}))\le1\le|U|$; if $|U|\ge r+1$, then
$1+r(s_U-G^\lambda_r(\mathrm{root}))\le1+r\le|U|$. The first case uses $|U|\ge1$, which holds because full schedules are
nonempty when $n\ge1$. Averaging over $i$ gives $1+r(Q_\mu(S_k)-G^\lambda_r(\mathrm{root}))\le C(\mu)$. Since
$\KL\le\varepsilon$ forces $Q_\mu(S_k)\ge e^{-\varepsilon}$,
\[
\LF^\lambda(n,\varepsilon)\ge1+r\big(e^{-\varepsilon}-G^\lambda_r(\mathrm{root})\big)\qquad\text{for every integer $r\ge0$.}
\]
(At $\lambda=1$ and $n=2k$ this is Step~5 of the single-horizon bound in Appendix~\ref{app:conv}.)

\emph{Asymptotics.} Let $\Lambda_V:=\ln(V_{\mathrm{root}}+1)$, take
$r:=\lfloor\log_2\Lambda_V-\log_2\log_2\Lambda_V-c_0\rfloor$ with a constant $c_0>\log_2\ln18\approx1.53$ (e.g.\ $c_0=2$),
and apply Theorem~\ref{thm:tgenv}. Then $G^\lambda_r(\mathrm{root})\to0$: the choice of $r$ gives $2^{-r}\Lambda_V\ge2^{c_0}\log_2\Lambda_V$, and with the geometric
constant $\sqrt2\cdot18^{r-1}$ of Theorem~\ref{thm:tgenv} and $r\le\log_2\Lambda_V$,
$\ln G^\lambda_r\le\ln\sqrt2+(r-1)\ln18-2^{-r}\Lambda_V\le\ln\sqrt2+(\ln18-2^{c_0})\log_2\Lambda_V\to-\infty$, as
$2^{c_0}>\ln18$. Also $r\ge\log_2\Lambda_V-\log_2\log_2\Lambda_V-c_0-1=(1-o(1))\log_2\Lambda_V$, so the Markov split gives
\[
\LF^\lambda(n,\varepsilon)\ge\big(e^{-\varepsilon}-o(1)\big)\cdot\log_2\ln V_{\mathrm{root}}\qquad\text{as }V_{\mathrm{root}}\to\infty.
\]

\emph{(c) When $V_{\mathrm{root}}$ is of order $n$.} This corollary is conditional.

\emph{1.~$\lambda$-uniform, away from $D$.} By Lemma~TG3, $V_{\mathrm{root}}\ge\mathrm{dist}(k,D)/2$, with
$D=\{0,n_A,n_B,n\}$. The lower bound $\mathrm{dist}(k,D)/2$ is of order $n$ if and only if
$\mathrm{dist}(k,D)\ge\delta n$ for a fixed $\delta>0$ (a constant, not the $\delta=1-e^{-\varepsilon}$ of
Section~\ref{sec:bal}); then $\delta n/2\le V_{\mathrm{root}}\le n/4$. The ``if and only if'' concerns this bound, not
$V_{\mathrm{root}}$ itself, which can be of order $n$ without the hypothesis (item~2). Without that hypothesis, no
$\lambda$-uniform lower bound of order $n$ exists.
Counterexample: for every $k\ge1$, $n_A=n_B=k$ gives $\mathrm{dist}(k,D)=0$ and $V_{\mathrm{root}}\to0$ as $\lambda\to\infty$; for instance $k=5$ and $\lambda=10^6$ give $V_{\mathrm{root}}=5.0\cdot10^{-5}$.

\emph{2.~$\lambda$ in a compact subset of $(0,\infty)$, with interior proportions.} \emph{Setup.} Let $n\to\infty$ with
$n_A/n\to\rho\in(0,1)$ and $k/n\to\kappa\in(0,1)$. Let $\lambda$ range over a fixed nonempty compact $K\subset(0,\infty)$, so
that both $\lambda$ and $1/\lambda$ are bounded. Let $\theta^{(n)}$ be Lemma~TG8's pair at the root, $\sigma_n^2$ its
variance, and $\theta=\theta(\lambda,\rho,\kappa)$ the solution of the limiting equations
$\theta_A(1-\theta_B)/((1-\theta_A)\theta_B)=\lambda$ and $\rho\theta_A+(1-\rho)\theta_B=\kappa$. \emph{Convergence.}
Define $\psi_n(t):=(n_A/n)\cdot\lambda t/(1-t+\lambda t)+(n_B/n)\cdot t$ and
$\psi(t):=\rho\cdot\lambda t/(1-t+\lambda t)+(1-\rho)\cdot t$. (1) $|\psi_n-\psi|\le|n_A/n-\rho|+|n_B/n-(1-\rho)|$ on
$[0,1]$, uniformly in all $\lambda>0$. (2) $\psi'\ge1-\rho>0$, and $\psi(0)=0<\kappa<1=\psi(1)$. (3) So
$\theta_B^{(n)}\to\theta_B\in(0,1)$, with error at most $(\|\psi_n-\psi\|_\infty+|k/n-\kappa|)/(1-\rho)$. This is
uniform in all $\lambda>0$. (4) For $\theta_A=\chi_\lambda(\theta_B)$, where $\chi_\lambda(t):=\lambda t/(1-t+\lambda t)$,
use $\chi_\lambda'(t)=\lambda/(1+(\lambda-1)t)^2\le\max(\lambda,1/\lambda)$ on $[0,1]$. This is where compactness of $K$
enters: $|\theta_A^{(n)}-\theta_A|\le\max_K\max(\lambda,1/\lambda)\cdot|\theta_B^{(n)}-\theta_B|\to0$, uniformly over
$\lambda\in K$. (5) Hence $\sigma_n^2/n\to c(\lambda,\rho,\kappa):=\rho\theta_A(1-\theta_A)+(1-\rho)\theta_B(1-\theta_B)$
uniformly on $K$, and $\min_Kc>0$ by continuity and compactness. \emph{Conclusion.} By Lemma~TG8,
$V_{\mathrm{root}}\ge\sigma_n^2-\sigma_n=(c(\lambda,\rho,\kappa)-o(1))\cdot n$, uniformly over $\lambda\in K$, since $c$ is
continuous and positive on $K$. The $o(1)$ can be as slow as the convergence of the proportions, so
``$cn-O(\sqrt n)$'' is not claimed. \emph{Example.} $n_A=n_B=k$ gives $\theta_A=\sqrt\lambda/(1+\sqrt\lambda)$ and
$\sigma^2=n\sqrt\lambda/(1+\sqrt\lambda)^2$.

\emph{Where $c$ is small.} For every $\lambda$, $c(\lambda,\rho,\kappa)\ge\mathrm{dist}(\kappa,\{0,\rho,1-\rho,1\})/2$.
This is the product analogue of Lemma~TG3: $\sigma^2\ge\frac12\sum_gn_g\min(\theta_g,1-\theta_g)\ge\frac12\mathrm{dist}(k,D)$.
This bound proves only necessity: $c\to0$ requires $\kappa\in\{\rho,1-\rho\}$. The exact limits come from the odds
equation $\theta_A/(1-\theta_A)=\lambda\cdot\theta_B/(1-\theta_B)$ together with $\rho\theta_A+(1-\rho)\theta_B=\kappa$.
\emph{As $\lambda\to\infty$:} if $\kappa>\rho$, then $\theta_B\ge(\kappa-\rho)/(1-\rho)>0$ forces $\theta_A\to1$, so
$\theta_B\to(\kappa-\rho)/(1-\rho)$ and $c\to(\kappa-\rho)(1-\kappa)/(1-\rho)>0$; if $\kappa<\rho$, then
$\theta_A\le\kappa/\rho<1$ forces $\theta_B\to0$, so $\theta_A\to\kappa/\rho$ and $c\to\kappa(\rho-\kappa)/\rho>0$; if
$\kappa=\rho$, then $\theta_A\to1$ and $\theta_B\to0$, so $c\to0$. Indeed $\rho\theta_A+(1-\rho)\theta_B=\rho$ gives
$\theta_B=\rho(1-\theta_A)/(1-\rho)$. If $\theta_A\le1-\delta$ for some $\delta>0$ along a subsequence, then $\theta_B\ge\rho\delta/(1-\rho)>0$,
and the odds equation forces $\theta_A\to1$, a contradiction. \emph{As $\lambda\to0$}, by the $A\leftrightarrow B$
symmetry (swap the groups and replace $\lambda$ by $1/\lambda$): if $\kappa\ne1-\rho$ then $c$ has a positive limit,
and if $\kappa=1-\rho$ then $c\to0$. Hence $c\to0$ at extreme tilt exactly on two diagonals: as $\lambda\to\infty$ iff
$\kappa=\rho$, and as $\lambda\to0$ iff $\kappa=1-\rho$. Off these diagonals, item~1 gives a $\lambda$-uniform rate even
at finite $n$. Item~2's own uniformity is only over compact $K$. For example, $(\rho,\kappa)=(\frac12,\frac14)$ gives
$c\to1/8$ as $\lambda\to0$ or $\infty$, meeting the floor with equality, and $(0.3,0.5)$ gives $c\to1/7$. In particular,
$c\to0$ in both limits only at $\rho=\kappa=\frac12$.
\emph{One-sided bounds on $\lambda$ are not enough.} The family $n_A=n_B=k=N$ with $\lambda=N^{-4}$ has
$V_{\mathrm{root}}=2/N^2+O(N^{-4})\to0$. Rounded to the digits shown, the values are $0.1179$, $0.0309$, $0.00779$ and $0.00195$ at
$N=4$, $8$, $16$ and $32$. The same holds for $\lambda=N^4$, by the $A\leftrightarrow B$ symmetry.

\emph{3.~Consequence (conditional).} Suppose either (1)~$\mathrm{dist}(k,D)\ge\delta n$, or (2)~$\lambda$ lies in a fixed
compact subset of $(0,\infty)$, with $n_A/n\to\rho\in(0,1)$ and $k/n\to\kappa\in(0,1)$. Then
$\log_2\ln V_{\mathrm{root}}=\log_2\ln n+o(1)$. So, for every fixed $\varepsilon>0$:
$R^\lambda_F(\varepsilon)\ge\log_2\log_2n-O(\log\log\log n)$ for $\mathrm{RAND}_r$; and
$\LF^\lambda(n,\varepsilon)\ge(e^{-\varepsilon}-o(1))\cdot\log_2\ln n$ for predrawn mixtures of full labelled schedules.
With~(d) below, the first bound is $\log_2\log_2n-O_\varepsilon(1)$, and the second holds for every causal policy in
expected executed rounds, with $-O(1)$ in place of $-o(1)\cdot\log_2\ln n$.

Two remarks. First, the two hypotheses are \emph{sufficient, not necessary}: (a) and (b) hold in terms of $V_{\mathrm{root}}$
whenever $V_{\mathrm{root}}\to\infty$. For example, $n_A=n_B=k=N$ with $\lambda=(\ln N)^2$ violates both, yet
the Example of Corollary~(c)2 and Lemma~TG8 give $V_{\mathrm{root}}\ge N/(4\ln N)$ for large $N$, so
$\log_2\ln V_{\mathrm{root}}=\log_2\ln n+o(1)$. Second, \emph{no $\lambda$-uniform rate in $n$ holds without a condition like (1).}
We prove it for the family $n_A=n_B=k=N$ with $\lambda=N^4$ ($\lambda=N^{-4}$ follows by the $A\leftrightarrow B$ symmetry).
Step~1. With $Z$ the number of B-ones under $p_\lambda$, $\Pr(Z=z)/\Pr(Z=0)=\binom Nz^2N^{-4z}\le N^{-2z}$. Hence
$\E Z\le\sum_{z\ge1}zN^{-2z}=N^{-2}/(1-N^{-2})^2$. Step~2. The marginals are $a_A=1-\E Z/N$ and $a_B=\E Z/N$. So
$\nu_N:=\min(a_g,1-a_g)=\E Z/N\le N^{-3}/(1-N^{-2})^2$, and $V_{\mathrm{root}}=O(N^{-2})$. Step~3. One full round at the
exact marginals has output law $Q=\bigotimes_i\mathrm{Bernoulli}(a_i)$. For $p=p_\lambda$, whose coordinate marginals are the $a_i$, and
$H$ the Shannon entropy (not the criterion $H(\mu)$ of Section~\ref{sec:setup}), $\KL(p\|Q)$ is the total correlation $\sum_iH_b(a_i)-H(p)\le\sum_iH_b(a_i)$, where
$H_b$ is the binary entropy in nats. Step~4. Here
$\sum_iH_b(a_i)=2N\cdot H_b(\nu_N)\le2N\nu_N(1+\ln(1/\nu_N))=O(N^{-2}\ln N)\to0$, using $H_b(x)\le x(1+\ln(1/x))$ for
$x\le\frac12$ and that $x(1+\ln(1/x))$ increases on $(0,1]$. Step~5. So for every fixed $\varepsilon>0$ and all large
$N$, one round suffices: $R^\lambda_F(\varepsilon)=1$ and $\LF^\lambda(n,\varepsilon)=1$. No $\lambda$-uniform
doubly-logarithmic lower bound in $n$ can hold. Note that success probability alone would not give this:
$\KL(p_\lambda\|Q)=-\ln Q(S_k)+\KL(p_\lambda\|p_{\lambda'})$, with $\lambda'=a_A(1-a_B)/((1-a_A)a_B)$. Step~3 bounds both
terms at once. Numerically, at $N=4$, $8$, $16$, $32$, $64$ and $128$, the one-round KL is $0.229$, $0.080$,
$0.0255$, $0.0077$, $0.0023$ and $0.00065$, and the bound $2N\,H_b(\nu_N)$ is $0.622$, $0.224$, $0.073$, $0.022$,
$0.0066$ and $0.0019$ ($60$-digit arithmetic, not interval-certified).

\emph{(d) Bounded prefactor: hard cap and executed mean.} Here Theorem~\ref{thm:tgenv} is used with its bounded constant $C_r<324\sqrt2$ instead of the geometric form of~(a)
and of the asymptotics of~(b). So, without the hypothesis $V_{\mathrm{root}}\ge2$ of~(a), $F^*_r\ge2^{-r}\ln(V_{\mathrm{root}}+1)-\ln C_r$ and
$R^\lambda_F(\varepsilon)>L_V-\log_2(\varepsilon+\ln(324\sqrt2))$. Now let $\pi$ be a causal randomized policy with
fresh exact-marginal draws and no cap, whose output has KL at most $\varepsilon$, and let $\tau$ be its executed-round count (Section~\ref{sec:setup}).
By Lemma~TG9 (empty rounds via Lemma~TG10),
the probability that the state is legal with $M_t=0$ after round $t$, which is the probability of a valid output with
$\tau\le t$, is at most $G^\lambda_t(\mathrm{root})$. As in Lemma~\ref{lem:surv}(b) and the proof of Corollary~\ref{cor:t2}(b): $\E\tau=\sum_{t\ge0}\Pr(\tau>t)$ with $\Pr(\tau>0)=1$, and for $t\ge1$,
$\Pr(\tau>t)\ge Q_\pi(S_k)-\Pr(\text{valid output},\ \tau\le t)\ge e^{-\varepsilon}-G^\lambda_t(\mathrm{root})$, by Lemma~\ref{lem:tools}(a); so
\[
\E\tau\ge1+\sum_{t\ge1}\big(e^{-\varepsilon}-G^\lambda_t(\mathrm{root})\big)_+\ge e^{-\varepsilon}L_V-c_{\mathrm{TG}},\qquad
c_{\mathrm{TG}}=\sum_{i\ge0}\min\{1,324\sqrt2\,e^{-2^i}\}<3.154;
\]
for $L_V\ge1$ put $m=\lfloor L_V\rfloor$: for $1\le t\le m$, Theorem~\ref{thm:tgenv} and $(V_{\mathrm{root}}+1)^{-2^{-t}}=e^{-2^{L_V-t}}\le e^{-2^{m-t}}$ give $G^\lambda_t(\mathrm{root})\le\min\{1,324\sqrt2\,e^{-2^{m-t}}\}$, so $\sum_{t\le m}G^\lambda_t(\mathrm{root})\le c_{\mathrm{TG}}$ and $\E\tau\ge1+me^{-\varepsilon}-c_{\mathrm{TG}}\ge e^{-\varepsilon}L_V-c_{\mathrm{TG}}$, since $e^{-\varepsilon}\le1$ and $m+1>L_V$. In $c_{\mathrm{TG}}$ the terms $i\le2$ equal $1$, the term $i=3$ is below $0.15372$, and the terms $i\ge4$ sum to less than $10^{-4}$. For $L_V<1$, including $L_V<0$, the second inequality follows from $\tau\ge1$. A predrawn mixture is such a policy with
$\tau=|U|$, so the same bound holds for $\LF^\lambda(n,\varepsilon)$. These are lower bounds only: no two-group upper
construction of matching order is proved here.

\emph{Numerical checks (not used in the proofs).} All of the following is finite falsification evidence; the proof carries the general
claim. \emph{Checks that cannot fail at small sizes.} The full envelope at $r\ge2$ exceeds $1$ while $V<419{,}903$
($r=2$) or $V<1.8\cdot10^{16}$ ($r=3$; $V<1.9\cdot10^{21}$ for the geometric form), and Lemma~TG6's bound $18(1+V)^{-1/4}$ is at least $1$ for $V\le104{,}975$, with equality at $V=104{,}975$
since $18^4=104{,}976$. So
``$0$ violations'' of those two inequalities at $n\le16$ ($V\le4$), or of Case~F at $n=200$, carries no information;
likewise the survival-DP envelope check ($n\le80$ with two groups, $n\le400$ at $\lambda=1$) carries information only at
$r=1$. \emph{Informative checks.} These are the single-lemma inequalities (Lemmas~TG1--TG5 and~TG8), the Case-S chain at
$V\ge32$, whose bound $1.68(1+V)^{-1/4}$ is $<1$, and Lemma~TG6 at $V=113{,}750$ and $125{,}000$. None was violated, beyond floating-point
rounding. The single-lemma inequalities were checked by independent small-size sweeps (float64 with exact-rational
re-checks of the tightest cases, and a separate recomputation in rational and high-precision decimal arithmetic), by a further test at $m_A,m_B\le6$ that also covers the
endpoint quotas $j\in\{0,M\}$, $v_b=0$, every batch and every legal child quota (exact rationals except for square
roots, powers, Lemma~TG6's product probabilities and bisection), and on larger instances at tilts between $10^{-12}$ and $10^{12}$. The larger instances covered Lemmas~TG2, TG3
and~TG4(ii) in exact big-integer arithmetic for group sizes up to $100$, Lemma~TG5 exhaustively for group sizes up to
$60$, and Lemma~TG8 for group sizes up to $200$ (and up to $40$ in the small-size sweeps). At $m_A=m_B=100$ (float64, $j\in\{60,80,100,120,140\}$,
$V\in[33.4,50]$, $\lambda\in\{1/3,1,3,10\}$, all $10{,}200$ batches) the largest Case-S effective constant
$T(s,b)\,(1+V)^{1/4}$ is $0.579$, against the bound $1.68$; the Case-F maximum, $1.236$, is against a bound that is
vacuous at these $V$. Lemma~TG6's final inequality holds at the two tested states, $V=113{,}750$ and $V=125{,}000$ ($n=5\cdot10^5$, every
batch), with a $13\times$ margin, but both have one nonempty group, so this does not test extreme $\lambda$.
The constants of
Lemmas~TG1 and~TG6, including $g(32)=1.6710688<1.68$ and the monotonicity of $g$, are also checked numerically.
\emph{Other coverage.} Lemma~TG9's inequality: exact labelled enumeration over deterministic policies at $n\le8$,
which suffice for randomized ones because success is linear in the mixture over deterministic policies; there the
supremum equals $G^\lambda_r$. The inequality was also checked by exact labelled brute force for fixed schedules ($s_U\le G^\lambda_{|U|}$).
Lemma~TG10: no monotonicity violation for $n\le80$. Lemma~TG0 is elementary; it is implicit in every exact marginal
computation above. Corollary~(c)2 is an analytic argument; only its limit values are spot-checked
($(\frac12,\frac14)\to1/8$ and $(0.3,0.5)\to1/7$), together with the floor
$c\ge\mathrm{dist}(\kappa,\{0,\rho,1-\rho,1\})/2$ at $20{,}000$ random $(\lambda,\rho,\kappa)$.

\emph{Scope.} \emph{Narrow scope.} This is an ordinary, lower-bound-only result for the two-group tilted fixed-weight
target under the exact-marginal interface. It covers $\mathrm{RAND}_r$ (the almost-sure hard cap) and, through Corollary~(d), every causal policy with fresh
exact-marginal draws under expected executed rounds; part~(b) states the mean-cost bound for predrawn mixtures of full
labelled schedules. Its bounds are in terms of $V_{\mathrm{root}}$ and give doubly logarithmic rates in $n$ whenever $V_{\mathrm{root}}$ is of order $n$: for example under either sufficient condition of~(c), or with a bounded weight ratio and a quota fraction bounded away from $0$ and $1$, for any split (Appendix~\ref{app:t1}, \S5). \emph{Evidence labels.} ``Exact'' means exact rational or big-integer arithmetic; $60$-digit results are
high-precision but not interval-certified, and are labelled as such.

\emph{Lower bounds only.} This appendix proves no upper bound matching its lower bounds; the only
achievability results proved here are $F^*_n=0$ in (a), the one-round counterexample in the second of Corollary~(c)'s two remarks, and Lemma~TG11. Lemma~TG11 is
an attainment result for the survival optimum $G^\lambda_r(\mathrm{root})$, not a KL upper bound. The survival analysis of this appendix does not control the between-group shape term, the between-orbit divergence of the chain-rule identity in Appendix~\ref{app:alt}. For fixed-quota weighted targets, a rate-matching upper bound for $F$ off the balanced orbit is proved elsewhere in the paper only under bounded weight ratio and a quota fraction bounded away from $0$ and $1$,
by different constructions: with a homogeneous group of linear size in Appendix~\ref{app:t1} (Theorem~\ref{thm:tilt} and its
hard-cap corollary for two weight values; the Remark after its \S7 table for more weight values), and without one in
Appendix~\ref{app:rf}. The other rate-matching upper bounds are for the balanced uniform case $\lambda=1$, $n=2k$ (Theorem~\ref{thm:conv}(b), Theorem~\ref{thm:sep}(b) and Corollary~\ref{cor:t2}(c)) or for a target without a quota, the full-support Beta$(1,1)$--Bernoulli law (Theorem~\ref{thm:fs}(b),(c), Appendix~\ref{app:thmm}); Proposition~\ref{prop:perm} is an exact hard-cap identity for a different target, uniform permutations. Nothing in this appendix addresses the $H$ side. For bounded $\lambda$, the bounded-weight envelope of
Appendix~\ref{app:t1}, \S10 gives this appendix's adaptive lower-bound rates for any weight array of ratio at most
$\Lambda$; this appendix's $V$-based envelope, whose constants do not depend on $\lambda$, remains the form for
unbounded tilts, where $V$ can collapse (Corollary~(c)).

\emph{Loose constants.} The envelope is below $1$ only if $V+1>C_r^{2^r}$: $V+1>419{,}904$ at $r=2$, and
$V+1>16\cdot18^{12}>1.8\cdot10^{16}$ at every $r\ge3$, as $C_r^{2^r}$ increases with $r$. At $\lambda=1$ and the balanced
root, where $V=n/4$, these need $n>1{,}679{,}612$ and $n>7.4\cdot10^{16}$; the balanced envelope \eqref{eq:env} of
Appendix~\ref{app:env} is below $1$ at $r=2$ once $n>2590$. Finite certificates should use the exact DP. \emph{Classes.} Part~(a) covers
$\mathrm{RAND}_r$: causal adaptive policies, including adaptive stopping, under a hard cap $r$ that holds almost surely.
Part~(b) treats predrawn mixtures of full labelled schedules under mean cost; Corollary~(d) extends this lower bound
to every causal policy, adaptive stopping included, under expected executed rounds (empty, failed-path and forced final
rounds counted). Neither covers repair or restart, or samplers outside the exact-marginal
interface.
 \emph{Citations.} Lemmas~TG2 and~TG4(ii) have self-contained proofs; the citations
\citep{efron1965increasing,joagdev1983negative,borcea2009negative} are attribution, not load-bearing. Newton's
inequality, Jordan's inequality and Chebyshev's inequality are classical. The proof is ordinary;
Appendix~\ref{app:lean} lists the parts that are also kernel-checked.

\section{Proof of Theorem~\ref{thm:tilt} (bounded weights)}\label{app:t1}
\pend{}

This appendix proves Theorem~\ref{thm:tilt} with explicit constants, except the $F$ upper bounds of part~(a) for weights without an equal-weight group of linear size, which are Appendix~\ref{app:rf}'s. The Result below states the four bounds (HL), (HU), (FL) and (FU) for two weight values. \S1 fixes the setting and \S2 proves the weighted-slice lemmas; \S\S3--6 prove the four bounds in turn: (HL) from the entropy decrement of Lemma~W5, (HU) from the per-block bound of Lemma~W4, (FL) from the two-group survival envelope of Appendix~\ref{app:tgenv}, and (FU) from the chain rule of \S6.3 and the atom lemma of \S6.6. \S7 treats edge cases and more weight values; \S8 explains why survival alone is not enough; \S9 derives the least hard cap, a hard-length analogue of Corollary~\ref{cor:hard} and the sharp mean coefficient; and \S10 proves an adaptive converse for any bounded weights, the lower bound that the upper bounds of Appendix~\ref{app:rf} match.

\paragraph{Assembly of Theorem~\ref{thm:tilt}.} For any weights with ratio at most $\Lambda$ and $\kappa_0n\le k\le(1-\kappa_0)n$,
part~(a) is assembled as follows. The bounds on $\LH^w$ are (HL) and (HU) below. The lower bounds on $R_F^{\mathrm{adapt}}$ (written
$R^w_F$ in \S9), on $\LF^w$ and on the least expected executed rounds are \S10(c), which covers every causal policy and hence
every predrawn mixture ($\tau=|U|$). The matching upper bounds are Corollary~A$'$ of Appendix~\ref{app:rf} for the hard cap and
Corollary~A$''$ for the scheduled mean, and so for $\E\tau$, since a predrawn mixture executes every block; with an equal-weight
group of linear size, (FU), its extension in the Remark after the \S7 table and their \S9 dilution give these upper bounds with
explicit constants. For part~(b), the schedule mixture is (FU)'s $\mu_{\mathrm{ff}}$ (\S6) for two weight values and the Remark's schedule mixture
for a group of at least $\rho n$ equal weights; each has $F\le\varepsilon$ for large $n$ and one length at most
$\log_2\ln n+O(1)$ on every path, and \S10(c) bounds that length below by $\log_2\ln n-O(1)$.

\paragraph{Result (explicit constants).} Fix $\varepsilon>0$, $\Lambda\ge1$ and $\kappa_0\in(0,1/2]$. Write $\Hc_n=\sum_{m\le n}1/m$ and $S_n=\sum_{m\le n}1/m^2$, as in Theorem~\ref{thm:sep}. Set
\[
c:=2\kappa_0^2/\Lambda^6,\qquad\delta_4:=1-e^{-\varepsilon/4},\qquad\eta:=\kappa_0/(2\Lambda),\qquad
K_4:=\max\{1,\lceil\ln(2/\delta_4)/(2\eta^2)\rceil\}.
\]
Take any $n\ge2$, any weights with at most two values and $\max/\min\le\Lambda$ (any split, one group possibly empty), and
any $k\in[\kappa_0n,(1-\kappa_0)n]$. ((HL) and (HU) hold more generally, as each states; for arbitrary bounded weights, the Assembly above gives Theorem~\ref{thm:tilt}.) Then:
\begin{list}{}{\setlength{\leftmargin}{3.2em}\setlength{\labelwidth}{2.7em}\setlength{\labelsep}{0.5em}}
\item[(HL)] $\LH^w(n,\varepsilon)\ge\max\{1,\Hc_n-\varepsilon/c,1+[\Hc_n-\varepsilon/c-1]_+^2/(2\varepsilon/c+S_n-1)\}
\ge1+[\ln(n+1)-\varepsilon/c-1]_+^2/(2\varepsilon/c+2/3)$. This holds for \emph{any} number of weight values with ratio
$\le\Lambda$.
\item[(HU)] $\LH^w(n,\varepsilon)\le1+(2/\alpha)(\ln n+1)$, with $\alpha:=\min\{1/4,\varepsilon/(8\Lambda^3(\ln n+1))\}$. So
$\LH^w\le1+\max\{8(\ln n+1),16\Lambda^3(\ln n+1)^2/\varepsilon\}$. The bound is attained by \emph{one} fixed schedule and
holds for any weights with ratio $\le\Lambda$ and any $k$.
\item[(FL)] For every integer $r\ge0$,
$\LF^w(n,\varepsilon)\ge1+r(e^{-\varepsilon}-\min\{1,\sqrt2\cdot18^{r-1}(n\kappa_0/(2\Lambda)+1)^{-2^{-r}}\})$. Hence
$\LF^w\ge(e^{-\varepsilon}-o(1))\log_2\ln n$, where the $o(1)$ is a function of $n$ that depends only on $(\varepsilon,\Lambda,\kappa_0)$, uniformly in the split, $\lambda$ and $k$.
\item[(FU)] Let $n_0^{\mathrm{ff}}:=2M^*-1$, where $M^*$ is the least integer $M\ge1$ with
$\Lambda^2\ln(M+1)/(\kappa_0^2M)\le\varepsilon/2$. For $n\ge n_0^{\mathrm{ff}}$
the single group-aware schedule mixture $\mu_{\mathrm{ff}}$ of \S6 below has $F\le\varepsilon$, and \emph{every} component has length
$\le\max\{1,\lceil2\Lambda^2/\varepsilon\rceil\}+8K_4+\lceil\log_2\log_2n\rceil\le\log_2\ln n+C_{\mathrm{FU}}$, with
$C_{\mathrm{FU}}:=\lceil2\Lambda^2/\varepsilon\rceil+8K_4+2$. For $n<n_0^{\mathrm{ff}}$, the all-singleton schedule is exact, so
$\LF^w\le n<n_0^{\mathrm{ff}}$.
\end{list}
Hence $\LH^w=\Theta_{\varepsilon,\Lambda,\kappa_0}((\log n)^2)$ and $\LF^w=\Theta_{\varepsilon,\Lambda,\kappa_0}(\log\log n)$,
uniformly in the split, $\lambda\in[1/\Lambda,\Lambda]$ and $k$; \S9 sharpens the latter to
$\LF^w=(e^{-\varepsilon}+o(1))\log_2\ln n$.

\paragraph{1. Setting (the class of Section~\ref{sec:setup}, restated).} \emph{Target.}
$p_w(x)\propto\prod_iw_i^{x_i}\cdot\mathbf 1\{|x|=k\}$. The weights take the value $w_A$ on group~A (size $n_A$) and $w_B$
on group~B (size $n_B$), with $\lambda=w_A/w_B\in[1/\Lambda,\Lambda]$. This is the $p_\lambda$ of Appendix~\ref{app:tgenv}.

\emph{States.} At a committed prefix, let $m_A$ and $m_B$ be the free counts, $M=m_A+m_B$, and let the residual quota be
$J=k-(\text{committed ones})$. The prefix is legal iff $0\le J\le M$. A legal prefix has positive target mass. Given a
legal prefix, the free coordinates follow the weighted slice on the free set with quota $J$ (Lemma~TG0(i) of
Appendix~\ref{app:tgenv}). Their marginals depend only on the group and the count state (Lemma~TG0(ii)).

\emph{Interface.} A round commits a nonempty block by fresh independent Bernoulli draws at the exact conditional marginals
of the current prefix. Fresh means the draws are realised from auxiliary randomness independent of everything before. At
an infeasible prefix the draws are fair bits. An infeasible prefix never re-enters the slice.

\emph{Schedules, mixtures, costs.} A full labelled schedule $U$ is an ordered partition of $[n]$ into nonempty blocks. Every
block is executed on every path, and $|U|$ is the cost. A predrawn mixture $\mu$ is a finite distribution over full
labelled schedules, drawn once before any bit and independent of the auxiliary randomness. Write $Q_\mu=\E_\mu q_U$ and
$C(\mu)=\E_\mu|U|$. The criteria are $H(\mu)=\E_\mu\KL(p\|q_U)$ and $F(\mu)=\KL(p\|Q_\mu)$, forward KL in nats. The least costs
are $\LH^w(n,\varepsilon)=\inf\{C(\mu):H(\mu)\le\varepsilon\}$ and $\LF^w(n,\varepsilon)=\inf\{C(\mu):F(\mu)\le\varepsilon\}$.

\emph{Chain identity (used throughout).} For a fixed $U$ with blocks $B_1,\dots,B_R$, let $E_t:=B_1\cup\dots\cup B_{t-1}$.
Every prefix of an $x$ in the support of $p$ is legal, and there the sampler's marginals are the target's. Hence
\[
\KL(p\|q_U)=\sum_t\E_p[\mathrm{TC}_t],\qquad\mathrm{TC}_t:=\sum_{i\in B_t}H(X_i\mid X_{E_t})-H(X_{B_t}\mid X_{E_t})\quad
\text{(target law)}.\tag{C}
\]
Infeasible paths carry zero target mass and do not enter (C). In (C), $H(\cdot\mid X_{E_t})$ is the Shannon entropy of the target's conditional law at the realised prefix, so $\mathrm{TC}_t$ is a function of the prefix and $\E_p$ averages it; this $H$ is not the criterion $H(\mu)$.

\paragraph{2. Weighted-slice lemmas.} Throughout this part, $S$ is a free set of size $M$ with residual quota $J$, and the
weights on $S$ are positive with ratio $\le\Lambda$. The quota is nondegenerate iff $1\le J\le M-1$. Write $p_i=\Pr(X_i=1)$,
$v_i=p_i(1-p_i)$, $Z=e_J(S)$, and let $e_m(\cdot)$ be the elementary symmetric polynomials, with $e_m=0$ for $m<0$ or
$m>\text{size}$.

\emph{Lemma W1 (ratios and floors).} For $i\ne j$, let $R=S\setminus\{i,j\}$. Then
\[
p_i=w_i[e_{J-1}(R)+w_je_{J-2}(R)]/Z\qquad\text{and}\qquad1-p_i=[e_J(R)+w_je_{J-1}(R)]/Z.
\]
If $0<J<M$ and $w_i\ge w_j$:
\begin{itemize}
\item $1\le p_i/p_j\le w_i/w_j$. After cancelling $Z$, the numerator and denominator of $p_i/p_j$ are $w_ie_{J-1}+w_iw_je_{J-2}$ and $w_je_{J-1}+w_iw_je_{J-2}$.
\item $w_j/w_i\le(1-p_i)/(1-p_j)\le1$.
\end{itemize}
Hence (a) $v_i/v_j\in[w_j/w_i,w_i/w_j]\subset[1/\Lambda,\Lambda]$. (b) $p_i\ge p_j/\Lambda$ for every $j$. Summing over $j$,
$M\cdot p_i\ge J/\Lambda$, so $p_i\ge J/(\Lambda M)$. Likewise $1-p_i\ge(M-J)/(\Lambda M)$. At $J\in\{0,M\}$ both hold
trivially. At the root ($M=n$, $J=k$) this gives $\min(p_i,1-p_i)\ge\kappa_0/\Lambda$, so
$a_0:=\min_i\min(p_i,1-p_i)\ge\kappa_0/\Lambda$.

\emph{Lemma W2 (influences).} Assume a nondegenerate quota and $M\ge2$.

(i) $\mathrm{Cov}(X_i,X_j)=w_iw_j[e_J(R)e_{J-2}(R)-e_{J-1}(R)^2]/Z^2$. To see this, expand $Ze_{J-2}$ and the product of
the two marginal numerators; the terms in $(w_i+w_j)e_{J-1}e_{J-2}$ and $w_iw_je_{J-2}^2$ cancel.

(ii) Strict negativity. For positive reals and $0\le m\le N:=|R|=M-2$, we have $e_m^2>e_{m-1}e_{m+1}$:
\begin{itemize}
\item for $1\le m\le N-1$ this is Newton's inequality, with factor $(m+1)(N-m+1)/(m(N-m))>1$;
\item at $m\in\{0,N\}$ the right side is $0$ and $e_m>0$.
\end{itemize}
Take $m=J-1\in[0,M-2]$. Then $\mathrm{Cov}<0$ at every nondegenerate quota.

(iii) Define the influences $d_{ij}:=-\mathrm{Cov}(X_i,X_j)/v_i$. Then $\sum_{j\ne i}d_{ij}=1$, because $\E[\sum_{j\ne i}X_j\mid X_i=x]=J-x$; both conditionings are defined, since
$p_i\in(0,1)$.

(iv) Ratio. Fix $i$ and take $j\ne\ell$ (so $M\ge3$). Let $S'=S\setminus\{i,j,\ell\}$. The numerator
$N_j:=e_{J-1}(R)^2-e_{J-2}(R)e_J(R)$ with $R=S'\cup\{\ell\}$ equals $f(w_\ell)$, where $f(t)=f_0+f_1t+f_2t^2$ with
\[
f_0=e_{J-1}^2-e_{J-2}e_J,\qquad f_1=e_{J-1}e_{J-2}-e_{J-3}e_J,\qquad f_2=e_{J-2}^2-e_{J-3}e_{J-1}\qquad\text{(over $S'$)}.
\]
$f_0\ge0$ and $f_2\ge0$ by log-concavity, with zero extension. For $f_1\ge0$: if $e_{J-3}e_J>0$, multiply
$e_{J-2}^2\ge e_{J-3}e_{J-1}$ by $e_{J-1}^2\ge e_{J-2}e_J$ and divide by $e_{J-1}e_{J-2}>0$; otherwise
$f_1=e_{J-1}e_{J-2}\ge0$. Now $d_{ij}/d_{i\ell}=w_jf(w_\ell)/(w_\ell f(w_j))$, where $f(w_j)=N_\ell>0$ by (ii). If
$w_\ell\ge w_j$, then $f(w_j)\le f(w_\ell)\le(w_\ell/w_j)^2f(w_j)$. So the ratio lies in
$[w_j/w_\ell,w_\ell/w_j]\subset[1/\Lambda,\Lambda]$.

(v) By (iv), $\max_jd_{ij}\le\Lambda\cdot\min_jd_{ij}$, and by (iii), $(M-1)\cdot\min\le1\le(M-1)\cdot\max$. Hence
\[
1/(\Lambda(M-1))\le d_{ij}\le\Lambda/(M-1).\tag{I}
\]
For $M=2$, $d_{ij}=1$. With two weight values, $d_{ij}$ depends only on the groups of $i$ and $j$.

(vi) Same-group refinement. Let $i,j$ both lie in group~A, with $m_A$ free A-coordinates and $m_B$ free B-coordinates.
Then $(m_A-1)d_{AA}+m_Bd_{AB}=1$, where the second term is absent when $m_B=0$; if $m_B\ge1$, also $d_{AB}\ge d_{AA}/\Lambda$. In both cases $d_{AA}\le1/(m_A-1+m_B/\Lambda)$, which is at most $\Lambda/m_B$ when $m_B\ge1$.

\emph{Lemma W3 (one-step variance).} At a legal state with $0<J<M$, let $P_i'=\Pr(X_i=1\mid X_j)$ under the target's conditional law given that state. Since $X_j$ is
binary, $P_i'=p_i+(\mathrm{Cov}/v_j)(X_j-p_j)$, so $\mathrm{Var}(P_i')=\mathrm{Cov}^2/v_j$. Using
$\mathrm{Cov}=-d_{ij}v_i=-d_{ji}v_j$:
\[
\mathrm{Var}(P_i')=d_{ij}^2v_i(v_i/v_j)=d_{ji}^2v_j.
\]
By (I) and W1(a):
\[
v_i/(\Lambda^3(M-1)^2)\le\mathrm{Var}(P_i')\le\Lambda^3v_i/(M-1)^2,\qquad\text{and}\qquad
\mathrm{Var}(P_i')\ge v_j/(\Lambda^2(M-1)^2).\tag{II}
\]
If $w_i=w_j$, then $\mathrm{Var}(P_i')=d_{ij}^2v_i$. At a deterministic state ($J\in\{0,M\}$), every term of (II) is $0$.

\emph{Lemma W4 (per-block total correlation).} Take a block $B=(i_1,\dots,i_b)$ at any legal state with $M$ free
coordinates. Condition on the past, and reveal the block one coordinate at a time, under the target (a proof device). Let
$P^{(u)}$ be the conditional probability of $X_{i_s}$ after $i_1,\dots,i_{u-1}$; it is a target martingale. Then:
\begin{itemize}
\item $\mathrm{TC}(B)=\sum_sI(X_{i_s};X_{i_1\dots i_{s-1}})$ and $I_s=\E[d(P^{(s)}\|P^{(1)})]$.
\item The $\chi^2$ bound $d(a\|p)\le(a-p)^2/(p(1-p))$ comes from $\ln x\le x-1$; the term is $0$ if $p\in\{0,1\}$.
\item Orthogonal increments give $\E(P^{(s)}-P^{(1)})^2=\sum_{u<s}\E\,\mathrm{Var}_u$, where $\mathrm{Var}_u$ is the conditional variance of the increment $P^{(u+1)}-P^{(u)}$ given $X_{i_1},\dots,X_{i_{u-1}}$.
\item By (II) at free size $M-u+1$, with $v^{(u)}:=P^{(u)}(1-P^{(u)})$, $\E\,\mathrm{Var}_u\le\Lambda^3\E v^{(u)}/(M-u)^2\le\Lambda^3v^{(1)}/(M-u)^2$. The last
step is Jensen: $x(1-x)$ is concave and $P$ is a martingale.
\end{itemize}
Hence $I_s\le\Lambda^3\sum_{u<s}(M-u)^{-2}$, and
\[
\mathrm{TC}(B\mid\text{past})\le\Lambda^3\sum_{u=1}^{b-1}(b-u)/(M-u)^2\le\Lambda^3b(b-1)/(2(M-b+1)^2),\qquad
\text{for \emph{every} legal past.}\tag{III}
\]
\emph{(IV) Same-group block with a reservoir.} Let the block lie in group~A while group~B has at least $M$ free
coordinates, untouched during the block (here $M$ is this lower bound, not the total free count). W2(vi) gives $d_{AA}\le\Lambda/M$, and $v$ is equal within a group. So
$\mathrm{Var}_u\le\Lambda^2v^{(u)}/M^2$, and
\[
\mathrm{TC}(B\mid\text{past})\le\Lambda^2b(b-1)/(2M^2)\qquad\text{for every legal past.}\tag{IV}
\]

\emph{Lemma W5 (entropy decrement at deterministic labelled sets).} Let $E\subset[n]$ be a \emph{fixed} set (it must not
depend on the values), $S=[n]\setminus E$ with $|S|=M\ge2$, and $i\ne j\in S$. Let $J=\sum_{l\in S}X_l$; it is random under $p$. Write $P_i=\Pr(X_i=1\mid X_E)$, $P_i'=\Pr(X_i=1\mid X_E,X_j)$ and $v_j^E=\mathrm{Var}(X_j\mid X_E)$.
Then:
\begin{itemize}
\item By Pinsker,
$\Delta:=\E_p[H(X_i\mid X_E)-H(X_i\mid X_E,X_j)]=\E_p\E[d(P_i'\|P_i)\mid X_E]\ge2\E_p\mathrm{Var}(P_i'\mid X_E)$.
\item By (II), $\mathrm{Var}(P_i'\mid X_E)\ge v_j^E/(\Lambda^2(M-1)^2)$.
\item By W1(b), $v_j^E\ge J(M-J)/(\Lambda^2M^2)$.
\item By pairwise negative correlation at the \emph{root},
$\E_p[J(M-J)]=\sum_{l\ne m\in S}\Pr(X_l=1,X_m=0)\ge\sum_{l\ne m}p_l(1-p_m)\ge a_0^2M(M-1)$.
\end{itemize}
Therefore
\[
\Delta\ge2a_0^2/(\Lambda^4M(M-1))\ge c/(M(M-1)),\qquad c:=2\kappa_0^2/\Lambda^6.\tag{V}
\]
Every expectation is over target values at a fixed labelled $E$. The root bound on $\E[J(M-J)]$ is a statement about a
fixed $S$. If $S$ were chosen by looking at values, for example ``the free set when $J$ first equals $0$'', then $J(M-J)$
could vanish identically on the selected event, and (V) would not follow. Exact finite checks illustrate this (not used in the proofs). At $n=6$, $k=3$, $\Lambda=1$, revealing the coordinates in a fixed order and stopping once the residual quota is determined ($J\in\{0,M\}$) gives decrement
$0$ on $2/5$ of the mass. For weights $(100,100,100,1,1,1)$ with $k=3$ and $M=3$, a causal adaptive pair choice reaches
$0.60\cdot2a_0^2/(M(M-1))$, while every fixed $E$ of the same size and every fixed pair give at least $55$ times that value: adaptivity lowers the decrement by a factor of about $92$ here, although not below (V), whose factor $\Lambda^{-4}$ is $10^{-8}$.

\paragraph{3. $H$ lower bound (HL).} Fix $U$. At round $t$, $E_t$ is deterministic. Order $B_t$ as $i_1,\dots,i_b$, and write
$M=M_{t-1}$ for the free count before round $t$. By the chain rule for mutual information, each $I(X_{i_s};X_{i_{<s}}\mid X_{E_t})$ is a sum over $u<s$ of
decrements for the pair $(i_s,i_u)$ at the deterministic set $E_t\cup\{i_1,\dots,i_{u-1}\}$, of free size $M-u+1$. By (V),
telescoping:
\[
\E_p\mathrm{TC}_t\ge c\sum_{s=1}^{b}\sum_{u=1}^{s-1}1/((M-u+1)(M-u))=c\sum_{m=M-b+1}^{M}(1/m-1/M).
\]
Summing over $t$ via (C) gives $d_U:=\KL(p\|q_U)\ge c(\Hc_n-a_U)$, where $a_U=\sum_tb_t/M_{t-1}$. Step~4 of
Appendix~\ref{app:hlow} is pure arithmetic, valid for any target:
\[
\beta_U=\sum_t(b_t/M_{t-1})^2\le2(\Hc_n-a_U)+S_n\le2d_U/c+S_n.
\]
Both bounds are affine in $d_U$, so they average under $\mu$. If $H(\mu)=\E_\mu d_U\le\varepsilon$, then
$\E_\mu a\ge\Hc_n-\varepsilon/c$ and $\E_\mu\beta\le2\varepsilon/c+S_n$. Drop each schedule's last block ($\rho_R=1$, in the notation of Appendix~\ref{app:hlow}) and
apply Cauchy--Schwarz on $\mu\times(\text{rounds }t<R_U)$:
\[
(\E_\mu a-1)^2\le(C(\mu)-1)(\E_\mu\beta-1).
\]
This is (HL), including the middle term $C(\mu)\ge\E_\mu a$. Finally $\Hc_n\ge\ln(n+1)$ and $S_n-1\le2/3$ (Step~6 of
Appendix~\ref{app:hlow}). Nothing in this part uses two weight values or the group split.

\emph{Informativeness.} HL is an order statement. At finite $n$ with $\Lambda\ge3$, (HL) can be vacuous. On $363$ targets with one to three weight values ($n$ from $8$ to $40$, $\Lambda\in\{1,\frac32,2,3,4,6,8,9\}$) and seven values of $\varepsilon$ from $10^{-3}$ to $1$ ($2541$ cells), the Result's (HL), at the largest admissible $\kappa_0=\min(k,n-k)/n$, was $1$ in all $1197$ cells with $\Lambda\ge3$.
At $\Lambda=2$ it exceeded $1$ only for $\varepsilon\le0.01$. With (V)'s sharper constant $2a_0^2/\Lambda^4$ in place of $c$, $19$ of the $1197$ cells exceed $1$, all at $\varepsilon=10^{-3}$. The $(\ln n)^2$ regime can begin at astronomically large
$n$. The exact $\LH^w$ itself barely depends on $\lambda$ at small $n$; for example $9.30$, $9.28$ and $9.21$ at
$\lambda=1,2,4$ with $n=24$, $k=12$, $\varepsilon=0.1$ (two groups of size 12; exact computation).

\paragraph{4. $H$ upper bound (HU).} From the current free count $M$, take a block of size $b=\max(1,\lfloor\alpha M\rfloor)$,
with any labels. This is one fixed schedule. A nonsingleton block has $\alpha M\ge2$ and $b\le M/4$, so (III) gives
$\mathrm{TC}\le\Lambda^3\alpha^2M^2/(2(3M/4)^2)=(8/9)\Lambda^3\alpha^2$ for every past. Each nonsingleton block removes
$\lfloor\alpha M\rfloor\ge\alpha M/2$ coordinates, and such blocks occur only while $M\ge2/\alpha$. They form an initial
segment, since $\lfloor\alpha M\rfloor$ is nonincreasing as $M$ decreases; let $N_{\mathrm{ns}}$ be their number. If $\alpha n<2$ there are none
($N_{\mathrm{ns}}=0$), and the schedule is all singletons and exact. Otherwise
$N_{\mathrm{ns}}\le1+(2/\alpha)\ln(\alpha n/2)$. In every case $N_{\mathrm{ns}}\le1+(2/\alpha)\ln n$. After them come fewer
than $2/\alpha$ singletons, which are exact. By (C):
\[
H\le(8/9)\Lambda^3\alpha^2(1+(2/\alpha)\ln n)\le(16/9)\Lambda^3\alpha(\ln n+1)\le2\varepsilon/9,
\]
and the number of rounds is $R\le1+(2/\alpha)(\ln n+1)=1+\max\{8(\ln n+1),16\Lambda^3(\ln n+1)^2/\varepsilon\}$. Uniformly in
$\varepsilon$ this is $O(\ln n+\Lambda^3(\ln n)^2/\varepsilon)$.

\paragraph{5. $F$ lower bound (FL).} This section gives the two-weight form, through Appendix~\ref{app:tgenv}; for arbitrary bounded weights, the $F$ lower bounds of Theorem~\ref{thm:tilt}(a) are \S10(c). Corollary~(b) of Appendix~\ref{app:tgenv} gives
$\LF^w(n,\varepsilon)\ge1+r(e^{-\varepsilon}-G^\lambda_r(\mathrm{root}))$ for every integer $r\ge0$, where $G^\lambda_r(\mathrm{root})$ is the largest probability of meeting the quota within $r$ rounds (the recursion of Appendix~\ref{app:tgenv}). The inputs are that corollary's
Markov split (a component with at most $r$ blocks meets the quota with probability at most $G^\lambda_r(\mathrm{root})$, and any other component costs at least $r+1$) and $\KL\ge-\ln Q(S_k)$, with $S_k=\{x:|x|=k\}$. Theorem~\ref{thm:tgenv} of Appendix~\ref{app:tgenv}, with its geometric constant, gives
$G^\lambda_r(\mathrm{root})\le\min\{1,\sqrt2\cdot18^{r-1}(V_{\mathrm{root}}+1)^{-2^{-r}}\}$. Let $a_g$ be the root marginal of a coordinate in group~$g$. By W1(b), each nonempty
group has $a_g(1-a_g)\ge\min\cdot\max\ge(\kappa_0/\Lambda)\cdot(1/2)$, where $\min$ and $\max$ denote the smaller and the larger of $a_g$ and $1-a_g$. An empty group has $n_g=0$, so it contributes nothing to $V_{\mathrm{root}}$. So
\[
V_{\mathrm{root}}=n_Aa_A(1-a_A)+n_Ba_B(1-a_B)\ge n\kappa_0/(2\Lambda),
\]
uniformly in the split. The ``Asymptotics'' of Appendix~\ref{app:tgenv} then applies with $V_{\mathrm{root}}\to\infty$ at a
rate fixed by $(n,\kappa_0,\Lambda)$. For example, take $r=\max\{0,\lfloor\log_2\Lambda'-\log_2\log_2\Lambda'-2\rfloor\}$ with
$\Lambda'=\ln(n\kappa_0/(2\Lambda)+1)$, a lower bound for the $\Lambda_V=\ln(V_{\mathrm{root}}+1)$ of those Asymptotics (not a weight ratio). Then $G^\lambda_r(\mathrm{root})\le\sqrt2\,\Lambda'^{\log_218-4/\ln2}\to0$ and
$r=\log_2\ln n-O(\log\log\log n)$. This supplies a sufficient condition that Corollary~(c) of Appendix~\ref{app:tgenv} does
not list: bounded $\Lambda$ with an interior quota, and \emph{any} split. It is consistent with Corollary~(c)1, whose
``no $\lambda$-uniform bound'' concerns unbounded $\lambda$. The bound on $V_{\mathrm{root}}$ is attained at
$(n,n_A,k,\lambda)=(2,0,1,1)$ with $\kappa_0=\frac12$ and $\Lambda=1$.

\emph{Remark (an explicit two-weight constant, from Corollary~(d) of Appendix~\ref{app:tgenv}).} Corollary~(d) of
Appendix~\ref{app:tgenv} bounds the expected executed rounds of every causal policy with output KL at most $\varepsilon$, and so
gives $\LF^w\ge1+\sum_t(e^{-\varepsilon}-G^\lambda_t)_+\ge e^{-\varepsilon}\log_2\ln(V_{\mathrm{root}}+1)-c_{\mathrm{TG}}$ with
$c_{\mathrm{TG}}<3.154$. Hence $\LF^w\ge e^{-\varepsilon}\log_2\ln n-4.154$ for $n\ge(2\Lambda/\kappa_0)^2$. The undiluted schedule mixture of (FU) alone gives no two-sided
mean law, since its leading coefficient is $1$; mixing it with a cheap one-round schedule gives the matching mean law
$\LF^w=(e^{-\varepsilon}+o(1))\log_2\ln n$ (\S9, ``Sharp mean coefficient'').

\paragraph{6. $F$ upper bound: the reservoir construction (FU).} Group~A, a smallest group, is committed first, in fixed blocks, while group~B is untouched; then a ladder like that of Theorem~\ref{thm:sep}(b), with the constant $K_4$ of the Result, runs on~B, relabelled uniformly within~B. Because B is homogeneous, this relabelling makes the conditional output law of $x_B$ given $x_A$ uniform on each residual slice, so the chain rule splits the output KL $F$ exactly into the divergence $\KL(p_A\|Q_A)$ of the values committed in Stage~A and $\E_p[-\ln s_J]$, where $s_J$ is the ladder's success probability from the residual quota $J=k-|x_A|$ (\S\S6.3, 6.6, 6.8). The first term is at most $\varepsilon/4$ because B is untouched during Stage~A, so (IV) applies to every Stage~A block, and Stage~A splits A into $\bar L$ blocks (into singletons if $N\le\bar L$), enough to make the summed (IV) bounds at most $\varepsilon/4$ (\S6.4; by \S6.9 a single block does not suffice); the second is at most $\varepsilon/4+\Lambda^2\ln(M+1)/(\kappa_0^2M)$ (\S6.8) because the ladder succeeds with probability at least $1-\delta_4$ at typical quotas (a supermartingale bound with Ville's inequality, valid at every quota fraction in $[\eta,1-\eta]$ because $K_4\ge\ln(2/\delta_4)/(2\eta^2)$; \S6.5) and at least $1/(n_B+1)$ at every legal quota ($J$ is a mode of $\Bin(n_B,J/n_B)$; \S6.6), while atypical quotas have target probability at most $1/(4\eta^2n_B)$ (\S6.7). \S6.1 defines the schedule mixture, \S\S6.2--6.8 prove (FU), and \S6.9 gives explicit constants.

\emph{6.1 Construction.} Name the groups so that A is a smallest group: $N:=n_A\le M:=n_B$, hence $M\ge n/2$. If $n_A=n_B$,
use the group~B of Appendix~\ref{app:tgenv} as the reservoir. Parameters: $\delta_4$, $\eta$ and $K_4$ as in the Result,
and $\bar L:=\max\{1,\lceil2\Lambda^2N^2/(\varepsilon M^2)\rceil\}$.

\emph{Stage~A}, the prefix $U_A$, identical in every component:
\begin{itemize}
\item if $N=0$, there are no rounds;
\item if $1\le N\le\bar L$, there are $N$ singletons;
\item otherwise there are $\bar L$ consecutive blocks in a fixed label order, with sizes $\lceil N/\bar L\rceil$ for the
first $(N\bmod\bar L)$ blocks and $\lfloor N/\bar L\rfloor$ for the rest.
\end{itemize}
Let $L_A$ be the number of Stage~A rounds. Then $L_A\le\max\{1,\lceil2\Lambda^2/\varepsilon\rceil\}$, since $N\le M$.

\emph{Ladder profile on B}, the same for every residual quota:
\begin{itemize}
\item if $M\le8K_4$, $M$ singletons;
\item otherwise $M_j:=K_4\cdot2^{2^j+j+2}$ ($M_0=8K_4$) and $T:=\min\{j:M\le M_j\}$. The blocks reduce the free B-count
$M\to M_{T-1}\to\dots\to M_0$, followed by $M_0$ singletons.
\end{itemize}
The length is $\ell_{\mathrm{lad}}=M\le8K_4$ if $M\le8K_4$, and otherwise $\ell_{\mathrm{lad}}=T+8K_4\le\lceil\log_2\log_2M\rceil+8K_4$,
since $M_j\ge2^{2^j}$.

\emph{The schedule mixture $\mu_{\mathrm{ff}}$}, drawn before any bit: $U_A$ followed by $\sigma(\text{ladder})$, where $\sigma$ is a
uniformly random assignment of B's labels to the ladder's positions. All $M!$ components share the group-composition
sequence and the labels of~A, and B is permuted only within~B; this is not a symmetrization over all $n$ coordinates. The
support is finite, every block is nonempty and every block is charged, so every component has length
$L_A+\ell_{\mathrm{lad}}$ and the hard cap equals the mean cost.

\emph{6.2 Class membership (FC).} Every component begins with the same $U_A$. Stage~A's committed values have law
$q_{U_A}(x_A)$ whatever follows, since the marginals at A-prefixes are the target's, with B unrevealed. Hence, for every
$x\in\{0,1\}^n$,
\[
Q(x)=Q_A(x_A)\cdot Q_{B|A}(x_B\mid x_A),\qquad Q_A=q_{U_A},\quad Q_{B|A}=\E_\sigma q_{\sigma(\mathrm{lad})}(\cdot\mid x_A).
\]
The relabelling $\sigma$ is drawn before any bit and never depends on committed values; only the conditional marginals
do, which the interface permits. Two facts about the class, not used below: choosing any part of the future schedule
after seeing committed values is outside the predrawn class, while drawing it later but independently of them is
law-equivalent to drawing it first; and components may use different predetermined prefixes, in which case conditional
weights given $x_A$ must be computed rather than assumed.

\emph{6.3 Chain rule.} Here and below, $D(\cdot\|\cdot)$ is $\KL(\cdot\|\cdot)$. Since $p=p_A\cdot p_{B|A}$ and $Q=Q_A\cdot Q_{B|A}$, and $D(p_A\|Q_A)<\infty$ (\S6.4),
\[
D(p\|Q)=D(p_A\|Q_A)+\E_{x_A\sim p_A}D(p_{B|A}(\cdot\mid x_A)\|Q_{B|A}(\cdot\mid x_A)).
\]
Here $p_A$-a.e.\ $x_A$ is legal, with $J:=k-|x_A|\in[0,M]$.

\emph{6.4 Stage~A (FA).} By (C) restricted to Stage~A, and by (IV), which applies at every legal past because B is
untouched with $M$ free coordinates:
\begin{align*}
D(p_A\|Q_A)&\le(\Lambda^2/(2M^2))\sum_tb_t(b_t-1)\le(\Lambda^2/(2M^2))\cdot N\cdot(\lceil N/\bar L\rceil-1)\\
&<\Lambda^2N^2/(2M^2\bar L)\le\varepsilon/4.
\end{align*}
The singleton and $N=0$ cases give $0$.

\emph{6.5 Ladder success at typical quotas (FB1).} Let $J/M=:p_0\in[\eta,1-\eta]$ and $M>8K_4$; for $M\le8K_4$ the ladder is
all singletons and succeeds surely. Let $P_t=J_t/M_t$ be the fraction at the start of bulk round $t$, where $M_t$ and $J_t$ are the free B-count and the residual quota at that time, the bulk rounds being the ladder's blocks before its final singletons. Here a subscript on $M$ counts bulk rounds, the first starting from $(M,J)$, whereas the ladder values $M_j$ of \S6.1 are indexed by level; the sentence after the definition of $S$ below translates between the two. The increment is
\[
P_{t+1}-P_t=(b_tP_t-h_t)/M_{t+1},\qquad h_t\sim\Bin(b_t,P_t),
\]
a sum of $b_t$ independent mean-zero terms, each of range $1/M_{t+1}$. Freeze $P$ at a kill, a round after which the state is illegal. By Hoeffding's lemma,
\[
Z_t=\exp\Big(\theta(P_t-p_0)-(\theta^2/8)\sum_{s<t}b_s/M_{s+1}^2\Big)
\]
is a nonnegative supermartingale. Let $S:=\sum_{\mathrm{bulk}}b_s/M_{s+1}^2$. A bulk round that starts from the ladder value $M_j$ (from $M\le M_T$, for the
first) leaves $M_{j-1}$ free coordinates, so the $M_{s+1}$ of its term is $M_{j-1}$. The first bulk block has $b\le M\le M_T$ and the
others $b\le M_j$, so each term is $\le M_j/M_{j-1}^2=1/(K_4\,2^j)$, and $S<1/K_4$. A kill means $P_{t+1}\notin[0,1]$, which
from $p_0\in[\eta,1-\eta]$ forces $|P_{t+1}-p_0|>\eta$. By Ville's inequality at $\theta=4\eta/S$, on both sides,
\[
\Pr(\text{kill})\le2\exp(-2\eta^2/S)<2\exp(-2\eta^2K_4)\le\delta_4.
\]
Singletons from a legal state never fail. So $s_J\ge1-\delta_4$.

\emph{6.6 The relabelled ladder is uniform on each residual slice (FB): an exact identity.} Fix a legal $J$ and let
$S_J=\{x_B:|x_B|=J\}$. Given $x_A$, every free B-coordinate has marginal $J_t/M_t$, so the ladder's law depends on $x_A$
only through $J$. Relabelling the schedule and the configuration together shows that $\E_\sigma q_{\sigma(\mathrm{lad})}(\cdot\mid J)$
is invariant under $\mathrm{Sym}(B)$. Indeed, at a legal state $(m,j)$ of the ladder every free B-coordinate has the same
exact marginal $j/m$, because B is homogeneous (Lemma~TG0(ii)). So the probability that a labelled ladder $\mathrm{lad}$
outputs $x_B$ depends on $\mathrm{lad}$ only through which positions its blocks occupy, and
$q_{\sigma_0(\mathrm{lad})}(\sigma_0x_B\mid J)=q_{\mathrm{lad}}(x_B\mid J)$ for every $\sigma_0\in\mathrm{Sym}(B)$. Hence, for $\sigma$
uniform, $\E_\sigma q_{\sigma(\mathrm{lad})}(\sigma_0x_B\mid J)=\E_\sigma q_{\sigma_0^{-1}\sigma(\mathrm{lad})}(x_B\mid J)=\E_\sigma q_{\sigma(\mathrm{lad})}(x_B\mid J)$, since $\sigma_0^{-1}\sigma$ is again uniform. $S_J$ is one orbit, so this law equals $s_J/\binom MJ$ at every point of $S_J$, where
$s_J$ is the count-process success probability from $(M,J)$. Since $p_{B|A}=\mathrm{Unif}(S_J)$ (Lemma~TG0(i) with a
homogeneous B),
\[
D(p_{B|A}\|Q_{B|A})=-\ln s_J\qquad\text{exactly.}\tag{FB}
\]
Typical $J$ gives $D\le-\ln(1-\delta_4)=\varepsilon/4$ (the typical-quota bound, \S6.5).

\emph{Lemma (one-block atom).} On a homogeneous slice with $m$ free coordinates and residual $j$, $0\le j\le m$, every count
policy drawing at the exact marginals and committing all $m$ coordinates in nonempty blocks succeeds with probability at
least $A(j,m)\ge1/(m+1)$. Here $A(j,m):=\Pr(\Bin(m,j/m)=j)$ for $m\ge1$ and $0\le j\le m$, $A(0,0):=1$, and $A(j,m):=0$ for
every other integer $j$.
\emph{Proof.} Induction on $m$; $m=0$ is trivial. Let the first block have size $b\ge1$ and put $p=j/m$. The child residual is
$j-u$ with $u\sim\Bin(b,p)$, and by induction the continuation succeeds with probability at least
$A(j-u,m-b)=\max_q\Pr(\Bin(m-b,q)=j-u)\ge\Pr(\Bin(m-b,p)=j-u)$ (for a legal child with $m-b\ge1$, the probability of $j-u$ successes in $m-b$ trials is largest at $q=(j-u)/(m-b)$, and for $m-b=0$ the residual $0$ has probability $1$ for every $q$); both sides are $0$ at illegal children. Summing over $u$,
the success probability is at least $\sum_u\Pr(\Bin(b,p)=u)\Pr(\Bin(m-b,p)=j-u)=\Pr(\Bin(m,p)=j)$, with equality for
$b=m$. Finally $A(j,m)\ge1/(m+1)$ because $j$ is a mode of $\Bin(m,j/m)$. \qed

Hence every legal $J$ gives $D\le\ln(M+1)$ (the all-quota bound). For $J\in\{0,M\}$, $s_J=1$.

\emph{6.7 Target tail (FB2).} By W1(b) at the root, $a_B\in[\kappa_0/\Lambda,1-\kappa_0/\Lambda]=[2\eta,1-2\eta]$. Under $p$,
$J=|x_B|$ has mean $Ma_B$. By negative correlation (Lemma~TG4(ii)), $\mathrm{Var}\,J\le\sum v\le M/4$. By Chebyshev,
$\Pr_p(J/M\notin[\eta,1-\eta])\le\Pr(|J-Ma_B|>\eta M)\le1/(4\eta^2M)$.

\emph{6.8 Assembly and cost.} If $M\le8K_4$, the ladder is all singletons and exact, so $s_J=1$ at every legal $J$ and
$F=D(p_A\|Q_A)\le\varepsilon/4$ with no condition on $n$. Otherwise, by \S\S6.3--6.7, with the typical-quota bound at typical quotas, the all-quota bound at the
atypical ones, and the tail bound of \S6.7 for their mass,
\[
F(\mu_{\mathrm{ff}})=D(p_A\|Q_A)+\E_p[-\ln s_J]\le\varepsilon/2+\Lambda^2\ln(M+1)/(\kappa_0^2M),
\]
using $1/(4\eta^2M)=\Lambda^2/(\kappa_0^2M)$. The last term decreases in $M\ge1$, and $M\ge\lceil n/2\rceil$, so $F\le\varepsilon$
for $n\ge n_0^{\mathrm{ff}}:=2M^*-1$, where $M^*$ is the least $M$ with $\Lambda^2\ln(M+1)/(\kappa_0^2M)\le\varepsilon/2$. Every
component has length $L_A+\ell_{\mathrm{lad}}$, which is at most
\[
\max\{1,\lceil2\Lambda^2/\varepsilon\rceil\}+8K_4+\lceil\log_2\log_2n\rceil\le\log_2\ln n+C_{\mathrm{FU}},
\]
using $\lceil\log_2\log_2n\rceil\le\log_2\ln n+\log_2(1/\ln2)+1<\log_2\ln n+1.53$. This is a hard cap, and the mean cost
equals it.

\emph{6.9 Explicit constants and finite behaviour.} The table evaluates the constants of the Result at a few parameter choices.
\begin{center}\small
\begin{tabular}{@{}lrrr@{}}
\toprule
$(\varepsilon,\Lambda,\kappa_0)$ & $K_4$ & $n_0^{\mathrm{ff}}$ & $C_{\mathrm{FU}}=\lceil2\Lambda^2/\varepsilon\rceil+8K_4+2$\\
\midrule
$(1,1,\tfrac12)$ & $18$ & $53$ & $148$\\
$(1,2,\tfrac12)$ & $71$ & $327$ & $578$\\
$(1,2,\tfrac14)$ & $282$ & $1731$ & $2266$\\
$(\tfrac12,3/2,2/5)$ & $80$ & $651$ & $651$\\
$(1/100,2,\tfrac12)$ & $214$ & $66{,}651$ & $2514$\\
\bottomrule
\end{tabular}
\end{center}
An independent implementation evaluated the exact law of $\mu_{\mathrm{ff}}$, with exact rational component laws: on every residual
slice the law is constant (spread $0$), (FB) holds to float64 rounding, the atom lemma holds at every quota of the actual
ladder profiles, and at theorem constants $F$ exceeds $D(p_A\|Q_A)$ by at most $1.4\cdot10^{-8}$ (19 instances with $\Lambda\in\{1,2,3\}$, $\varepsilon\in\{\frac14,1\}$, $(n_A,n_B)$ of the forms $(1,M)$, $(M,M)$ and $(1250,5000)$ with $M$ from $200$ to $20{,}000$, and $k/n$ near $\frac14$, $\frac25$ or $\frac12$; $n$ up to
$20{,}001$), with $F/\varepsilon\le0.061$ there: the ladder essentially never fails at the quotas that carry mass.
\pend{}
The caps are hundreds to thousands of rounds, against $\log_2\ln n\approx3.2$ at $n=10^4$. \emph{The constants are
impractical; Theorem~\ref{thm:tilt} is an asymptotic order statement, and no finite-size advantage is claimed.} Two in-class
variants that fail show which elements are load-bearing:
\begin{itemize}
\item without uniform B-relabelling, (FB) can fail: take $n=4$, $k=2$, weights $(2,1,1,1)$, group A $=\{0\}$, Stage~A
$\{0\}$ and then the blocks $\{1,2\}$, $\{3\}$ on B, and write words as $x_0x_1x_2x_3$. Given $x_0=1$, the residual slice
$\{1100,1010,1001\}$ has mass $s_J=8/9$ under $Q_{B|A}$ (the count-process success probability from $(M,J)=(3,1)$), and
$Q_{B|A}$ restricted to that slice and normalized by $s_J$ is $(1/4,1/4,1/2)$, not uniform. So
$D(p_{B|A}\|Q_{B|A})>-\ln s_J$, and this within-slice shape term makes $F$ strictly larger than for the schedule mixture that draws
$\sigma\sim\mathrm{Unif}(\mathrm{Sym}(B))$ and runs Stage~A followed by the $\sigma$-relabelled blocks, as in \S6.1;
\item Stage~A in one block ($\bar L=1$) at $N=M$ leaves a $D_A:=D(p_A\|Q_A)$ that does not shrink with $n$: at
$\lambda=1$ and $k=n/2$, admissible for every row of the table above, $D_A>0.096$ for every even $n\ge2000$, so there
$F\ge D_A>0.01$ (\S6.3) and the variant cannot meet $\varepsilon=0.01$ uniformly. Indeed, there $p$ is uniform on
the slice and $Q_A$ is the product of $\mathrm{Ber}(\frac12)$ on~A; both are exchangeable within~A, so $D_A$ is the
divergence between the laws of $w=|x_A|$, hypergeometric under $p$ and $\Bin(N,\frac12)$ under $Q_A$. Apply the
Donsker--Varadhan inequality $D(P\|R)\ge\E_Pf-\ln\E_Re^f$ with $f(w)=-2(w-N/2)^2/N$. The hypergeometric variance gives
$\E_pf=-n/(4(n-1))$. For a standard normal $g$ and every real $z$, $e^{-2z^2/N}=\E_g\cos(2gz/\sqrt N)$; with $z=w-N/2$, and since $\E_{\Bin}\cos(s(w-N/2))=\cos(s/2)^N$ for every real $s$, this gives
$\E_{\Bin}e^f=\E_g[\cos(g/\sqrt N)^N]\le2^{-1/2}+\Pr(|g|\ge\frac\pi2\sqrt N)$, since $\cos x\le e^{-x^2/2}$ for
$|x|\le\pi/2$ and $|\cos x|\le1$. Hence
\[
D_A\ge\tfrac12\ln2-\frac{n}{4(n-1)}-\ln\big(1+\sqrt2\,\Pr(|g|\ge\tfrac\pi2\sqrt N)\big),
\]
which increases in $n$, is $0.09644\ldots$ at $n=2000$ and tends to $\frac12\ln2-\frac14\approx0.0966$. In float64, on the grid
$\lambda=\Lambda^{j/10}$ ($|j|\le10$) and $k=\mathrm{round}(n(\kappa_0+(1-2\kappa_0)i/20))$ ($0\le i\le20$) at $n=2000$ and
$n=20{,}000$, the largest value of $D_A$ is $0.0965$--$0.0966$ for the three rows with $\kappa_0=\frac12$ (at $\lambda=1$, where
$k=n/2$), $0.1487$ for $(1,2,\frac14)$ and $0.1071$--$0.1072$ for $(\frac12,\frac32,\frac25)$.
\end{itemize}

\paragraph{7. Edge cases and more weight values.}
\begin{center}\small
\begin{tabular}{@{}p{0.25\textwidth}p{0.71\textwidth}@{}}
\toprule
case & what happens\\
\midrule
$N=0$ (one group) & Stage~A has zero rounds. $J=k$ is deterministic and typical, since $\kappa_0\ge2\eta$, so
$F\le\varepsilon/4$ with no tail term.\\
$N=1$ & Stage~A is one singleton, so $D_A=0$.\\
$\lambda=1$ & The weights are equal. Either treat the target as one group ($N=0$), or run the construction; both bounds
hold with any $\Lambda\ge1$.\\
$N\ll M$, including $N=o(n)$ & $\bar L=1$ when $2\Lambda^2N^2\le\varepsilon M^2$. No positive limiting proportion is
assumed anywhere.\\
$N=M=n/2$ & $\bar L=\lceil2\Lambda^2/\varepsilon\rceil$; $M\ge n/2$ still holds.\\
$J\in\{0,M\}$ & $s_J=1$, so $D=0$. It counts in the atypical set, harmlessly.\\
$J\notin[0,M]$ (infeasible Stage~A) & $p$-null, so it does not enter any KL. The component still runs to completion with
fair bits and is charged in full.\\
$M\le8K_4$ & The ladder is all singletons, exact given $x_A$.\\
small fixed $\varepsilon$ & $K_4\approx(2\Lambda^2/\kappa_0^2)\cdot\ln(8/\varepsilon)$, $L_A\le\lceil2\Lambda^2/\varepsilon\rceil$,
and $n_0^{\mathrm{ff}}\approx(4\Lambda^2/(\varepsilon\kappa_0^2))\cdot\ln n_0^{\mathrm{ff}}$. All are finite for fixed $\varepsilon$.\\
$n<n_0^{\mathrm{ff}}$ & Use $n$ singletons: $F=0$ and length $n<n_0^{\mathrm{ff}}$, absorbed into constants.\\
$H$ side & No group structure is used. HL and HU hold for any number of weight values; for the $F$ side beyond two
values, see the Remark below.\\
\bottomrule
\end{tabular}
\end{center}

\emph{Remark (more weight values: a homogeneous group suffices).} The proof of (FU) needs the two-value structure only
through a homogeneous group of size at least $n/2$; its two other two-group inputs are replaced below. Let $B$ be any set of $M\ge\rho n$ coordinates, $\rho\in(0,1]$, on which
the weights are equal, let the other $N=n-M$ weights be arbitrary, all $n$ weights having ratio at most $\Lambda$, and let
$\kappa_0n\le k\le(1-\kappa_0)n$. In \S6.4 use (III) in place of (IV): while B is untouched the free count satisfies
$M_t-b+1>M$, so a Stage~A block costs at most $\Lambda^3b(b-1)/(2M^2)$, and $\bar L$ becomes
$\max\{1,\lceil2\Lambda^3N^2/(\varepsilon M^2)\rceil\}$. In \S6.7 use W2(ii) in place of Lemma~TG4(ii), and in \S6.8 use
$M\ge\lceil\rho n\rceil$. The schedule mixture then has $F\le\varepsilon$, with every component of length at most $\log_2\ln n+C_R$,
for $n\ge\lfloor(M^*-1)/\rho\rfloor+1$, where $C_R=\max\{1,\lceil2\Lambda^3(1-\rho)^2/(\varepsilon\rho^2)\rceil\}+8K_4+2$.
With $\ell=\ln(n/M)$ and $\alpha=\min\{1/4,9\varepsilon/(64\Lambda^3(\ell+1))\}$, a geometric Stage~A, whose blocks are the
next $\min\{R,\lfloor\alpha M_t\rfloor\}$ A-coordinates ($R$ of them left, $M_t$ free), has at most
$1+\max\{8\ell,128\Lambda^3\ell(\ell+1)/(9\varepsilon)\}=O(1+\ell+\Lambda^3\ell(\ell+1)/\varepsilon)$ rounds once
$M\ge\max\{M^*,2/\alpha\}$; so a homogeneous group with $M\ge n/(\ln n)^C$, for a fixed $C$ and large $n$, still gives
$\LF^w=O((\log\log n)^2)$. Write $M_t=R_t+M$ for the free count at the start of round $t$ ($R_t$ the $R$ above at round $t$);
B is untouched, so $M_t>M$
while $R_t>0$. \emph{Cost.} Each block has $b_t\le\alpha M_t\le M_t/4$, so $M_t-b_t+1>\frac34M_t$, and by (III) at free
size $M_t$ its cost is at most $\Lambda^3b_t^2/(2(\frac34M_t)^2)\le\frac89\Lambda^3\alpha^2$. \emph{Rounds.} Every round
before the last has $b_t=\lfloor\alpha M_t\rfloor\ge\alpha M_t/2$, because $\alpha M_t\ge\alpha M\ge2$; hence
$M_{t+1}\le(1-\alpha/2)M_t$, and after $K$ such rounds $M<M_{K+1}\le ne^{-\alpha K/2}$. So $K<2\ell/\alpha$, and there are at
most $1+2\ell/\alpha=1+\max\{8\ell,128\Lambda^3\ell(\ell+1)/(9\varepsilon)\}$ rounds. \emph{Total.} By (C),
$D(p_A\|Q_A)\le(1+2\ell/\alpha)\frac89\Lambda^3\alpha^2\le\frac{16}9\Lambda^3\alpha(\ell+1)\le\varepsilon/4$. 

The lower bounds need no homogeneous group. For any weights with ratio at most $\Lambda$, part~(c) of the bounded-weight envelope theorem in \S10 gives $\LF^w\ge(e^{-\varepsilon}-o(1))\log_2\ln n$ and a least common length of at least
$\log_2\ln n-O_{\varepsilon,\Lambda,\kappa_0}(1)$, since it covers every causal policy and so every predrawn mixture ($\tau=|U|$); and (HL) and (HU) hold for any weights. So every target with a homogeneous group of size $M\ge\rho n$ has, in
the predrawn class, $\LH^w=\Theta((\log n)^2)$, $\LF^w=(e^{-\varepsilon}+o(1))\log_2\ln n$ (mixing as in \S9) and least
common length $\log_2\ln n+O_{\varepsilon,\Lambda,\kappa_0,\rho}(1)$; this includes weights with any fixed number $q$ of values, since a largest class has
$\rho\ge1/q$. The adaptive $F$ converse holds for any weights with ratio at most $\Lambda$, at hard cap and in expected executed
rounds (\S10); with a homogeneous group of size at least $\rho n$, the predrawn $F$ constructions above are
therefore optimal over all causal policies (the schedule mixture to an additive constant in the hard cap, and its \S9
dilution to leading order in expected executed rounds; \S10).
Without a homogeneous group, Appendix~\ref{app:rf} gives matching $F$ upper bounds for any weights with ratio at most
$\Lambda$ and interior quota: a finite predrawn mixture with $F\le\varepsilon$ each of whose components has at most
$\log_2\ln n+O_{\Lambda,\kappa_0}(1+\varepsilon^{-2})$ rounds (Corollary~A$'$) and a mixture with $F\le\varepsilon$ and scheduled mean at most
$(e^{-\varepsilon}+o(1))\log_2\ln n$ (Corollary~A$''$), with existence-only constants. Their components can have different lengths,
so they do not guarantee a common length, and the least-common-length statement above is still proved only with a homogeneous group.

\paragraph{8. Why survival alone is not enough.}
\begin{itemize}
\item \emph{Warning example.} Here $G^\lambda_r$ is the largest probability of meeting the quota within $r$ rounds and
$F^*_r$ the least output KL over $\mathrm{RAND}_r$ (Appendix~\ref{app:tgenv}, Corollary~(a)). The instance $(n_A,n_B,k)=(2,2,2)$, $\lambda=2$, $r=3$ has $G^\lambda_3=1$ but $F^*_3\approx5.01857\cdot10^{-4}$;
the instance $(3,3,3)$, $\lambda=2$, $r=3$ has $-\ln G^\lambda_3=0.05518<F^*_3=0.05881$. A policy can survive surely and still have the
wrong within-support law, so a survival-optimal two-group planner gives no $F$ bound by itself.
\item This construction controls the law directly:
\begin{itemize}
\item the between-orbit law (the law of $|x_A|$) is controlled by $D(p_A\|Q_A)\le\varepsilon/4$ in \S6.4;
\item the law on each residual slice is exactly uniform (\S6.6);
\item what remains is only $\E_p[-\ln s_J]$.
\end{itemize}
\item The one-block atom lemma (\S6.6) makes the atypical-quota term $O(\ln M/M)$: the conditional KL there is
$-\ln s_J\le\ln(M+1)$, because the ladder succeeds with probability at least $1/(M+1)$ by that lemma, and the atypical mass is
$O(1/M)$ by Chebyshev (\S6.7). No extra mixture component is needed for the atypical quotas.
\end{itemize}

\paragraph{9. The least hard cap, common length and the sharp mean coefficient.} Let $R^w_F(n,\varepsilon)$ be the least $r$ such that some policy in the
class $\mathrm{RAND}_r$ of Appendix~\ref{app:tgenv} (the class $\mathcal A_r$ of Section~\ref{sec:setup} for this target: causal randomized policies with fresh exact-marginal draws that
commit all $n$ coordinates within $r$ rounds almost surely) has output KL at most $\varepsilon$. For two weight values under the hypotheses of the Result, uniformly in
$\lambda\in[1/\Lambda,\Lambda]$, in the split (an empty group included) and in $k$:
\begin{itemize}
\item[(i)] for every $n\ge2$,
$R^w_F(n,\varepsilon)\ge\max\{1,\lceil\log_2((\ln(n\kappa_0/(2\Lambda)+1)+4\ln18)/(\varepsilon+\ln(324\sqrt2)))\rceil\}
\ge\log_2\ln n-c_L$, with $c_L:=1+\max\{\log_2(\varepsilon+\ln(324\sqrt2)),\log_2\ln(2\Lambda/\kappa_0)\}$;
\item[(ii)] for $n\ge n_0^{\mathrm{ff}}$, $R^w_F(n,\varepsilon)\le\max\{1,\lceil2\Lambda^2/\varepsilon\rceil\}+8K_4+\lceil\log_2\log_2n\rceil
\le\log_2\ln n+C_{\mathrm{FU}}$, and $R^w_F(n,\varepsilon)\le n$ for every $n\ge2$.
\end{itemize}
Hence $R^w_F(n,\varepsilon)=\log_2\ln n+O_{\varepsilon,\Lambda,\kappa_0}(1)$, at fixed $\varepsilon$; the statement is not uniform
as $\varepsilon\to0$, $\Lambda\to\infty$ or $\kappa_0\to0$.

\emph{Proof.} (i) Corollary~(d) of Appendix~\ref{app:tgenv} gives, for every $\pi\in\mathrm{RAND}_r$,
$\KL(p_w\|Q_\pi)\ge2^{-r}\ln(V_{\mathrm{root}}+1)-\ln C_r$ with $C_r=\sqrt2\cdot18^{2-2^{2-r}}$, the constant of Theorem~\ref{thm:tgenv} ($C_r\le\sqrt2\cdot18^{r-1}$, with equality at $r=1,2$); since
$\ln C_r=\ln(324\sqrt2)-2^{2-r}\ln18$, this is $(\ln(V_{\mathrm{root}}+1)+4\ln18)/2^r-\ln(324\sqrt2)$. By Lemma~W1, at the
root every coordinate has $a_g(1-a_g)\ge\kappa_0/(2\Lambda)$, because $\min(a_g,1-a_g)\ge\kappa_0/\Lambda$ and
$\max(a_g,1-a_g)\ge\frac12$; so $V_{\mathrm{root}}\ge n\kappa_0/(2\Lambda)$. The right side increases in $V_{\mathrm{root}}$, so
$\KL\le\varepsilon$ forces $2^r\ge(\ln(n\kappa_0/(2\Lambda)+1)+4\ln18)/(\varepsilon+\ln(324\sqrt2))$, which with $R^w_F\ge1$ is
the first bound. For $n\ge(2\Lambda/\kappa_0)^2$, $\ln(n\kappa_0/(2\Lambda)+1)\ge\ln n-\ln(2\Lambda/\kappa_0)\ge\frac12\ln n$, so
the bound is at least $\log_2\ln n-1-\log_2(\varepsilon+\ln(324\sqrt2))\ge\log_2\ln n-c_L$; for $2\le n<(2\Lambda/\kappa_0)^2$,
$\log_2\ln n-c_L<0<1\le R^w_F$. (ii) For $n\ge n_0^{\mathrm{ff}}$ the schedule mixture $\mu_{\mathrm{ff}}$ of \S6 has $F\le\varepsilon$ (\S6.8). It is a predrawn
finite mixture of full labelled schedules, and every block is executed on every path, so every path runs for the
common component length $L$ of \S6.8, and $\mu_{\mathrm{ff}}$ lies in $\mathrm{RAND}_L$. The all-singleton schedule is exact
and lies in $\mathrm{RAND}_n$. \qed

The mean cost of $\mu_{\mathrm{ff}}$ equals its cap, whose coefficient on $\log_2\ln n$ is $1$; the lower bound's coefficient
is $e^{-\varepsilon}$, which a different mixture attains (``Sharp mean coefficient'' below).

\emph{Hard-length analogue of Corollary~\ref{cor:hard}.} For a point mass $\mu=\delta_U$ on one fixed full labelled
schedule, $H(\mu)=F(\mu)=\KL(p_w\|q_U)$ and $C(\mu)=|U|$. So (HL) bounds the length of every fixed schedule with
$\KL(p_w\|q_U)\le\varepsilon$ from below by $\Omega_{\varepsilon,\Lambda,\kappa_0}((\ln n)^2)$, and (HU) constructs one with
$\KL(p_w\|q_U)\le2\varepsilon/9$ and at most $1+\max\{8(\ln n+1),16\Lambda^3(\ln n+1)^2/\varepsilon\}$ rounds. Both bounds are explicit for every
$n\ge2$, so the least length of a fixed labelled schedule with KL at most $\varepsilon$ is $\Theta_{\varepsilon,\Lambda,\kappa_0}((\ln n)^2)$. The schedule mixture
$\mu_{\mathrm{ff}}$ of \S6 draws $\sigma\sim\mathrm{Unif}(\mathrm{Sym}(B))$ before any bit, and every component executes all its blocks, so
it is a policy in $\mathrm{RAND}_L$, with $L$ the common component length. For $n\ge n_0^{\mathrm{ff}}$, \S6.8 gives
$F\le\varepsilon$ with $L\le\log_2\ln n+C_{\mathrm{FU}}$ (for $n<n_0^{\mathrm{ff}}$ the uniformly relabelled all-singleton
schedule is an exact member of this class, as in (ii)), and (i) gives $L\ge\log_2\ln n-c_L$ for every such schedule mixture. So the
least common component length of a single group-aware labelled schedule relabelled uniformly within $B$, with output KL at most
$\varepsilon$, is $\log_2\ln n+O_{\varepsilon,\Lambda,\kappa_0}(1)$. Both bounds are uniform in $\lambda$, the split and $k$.
This is a hard-length statement: for $\mu_{\mathrm{ff}}$ the mean equals the cap, but no new mean-cost law follows from it by itself; the sharp mean coefficient is proved next.
\pend{}
\pend{}

\emph{Sharp mean coefficient.} Uniformly in $\lambda$, the split and $k$, $\LF^w(n,\varepsilon)=(e^{-\varepsilon}+o(1))\log_2\ln n$,
with the $o(1)$ depending only on $(\varepsilon,\Lambda,\kappa_0)$ and $n$. \emph{Proof.} (FL) gives the lower bound. For any
target and $\beta'\in[0,1)$, $(1-\beta')Q_1+\beta'Q_2\ge(1-\beta')Q_1$ pointwise, so
$\KL(p\|(1-\beta')Q_1+\beta'Q_2)\le-\ln(1-\beta')+\KL(p\|Q_1)$. Fix $\nu\in(0,\varepsilon)$, and let $\mu_\nu$ be the
schedule mixture $\mu_{\mathrm{ff}}$ of \S6 built at tolerance $\nu$ ($\delta_4$, $K_4$ and $\bar L$ computed with $\nu$ in place of $\varepsilon$). For
$n\ge n_0^{\mathrm{ff}}(\nu)$, $F(\mu_\nu)\le\nu$ and every component has length at most $\log_2\ln n+C_{\mathrm{FU}}(\nu)$. Mix
$\mu_\nu$ (weight $e^{-(\varepsilon-\nu)}$) with the one-round schedule that commits all $n$ coordinates in one block.
This is a finite predrawn mixture with every block charged. Its output KL is at most $(\varepsilon-\nu)+\nu=\varepsilon$, and
its mean cost is at most $e^{-\varepsilon+\nu}(\log_2\ln n+C_{\mathrm{FU}}(\nu))+1$. Hence
$\limsup_n\sup\LF^w/\log_2\ln n\le e^{-\varepsilon+\nu}$, the supremum being over $\lambda$, the split and $k$; let
$\nu\downarrow0$. \qed

This is the device of Corollary~\ref{cor:t2}(c), with the mixing inequality in place of the relabelling identity. As on the
balanced slice, the hard cap ($\log_2\ln n+O(1)$, above) and the mean cost differ by the factor $e^{-\varepsilon}$ in the
leading coefficient, and the mean-cost bound mixes in a schedule that usually fails. The limit is not approached at any
practical $n$: at $(\varepsilon,\Lambda,\kappa_0)=(1,2,\frac12)$ the bound is within $10\%$ of $e^{-\varepsilon}\log_2\ln n$ only
once $\log_2\ln n$ exceeds about $2.5\cdot10^4$, even with $\nu$ chosen optimally.
\pend{}

\paragraph{10. Adaptive converse for any bounded weights.} Let the weights be arbitrary with ratio at most $\Lambda$; no
homogeneous group is assumed. A state is a free set $S$ of size $M$ with its weights and a residual quota $J$; it is legal
iff $0\le J\le M$, and its minority quota is $d:=\min(J,M-J)$. At a legal state the free coordinates have law
$\propto\prod_{i\in S}w_i^{y_i}\mathbf 1\{\sum_{i\in S}y_i=J\}$ with marginals $p_i$, and a batch $B\subseteq S$ (possibly empty)
is drawn as independent Bernoulli$(p_i)$, $i\in B$; write $X_B$ for its sum. Let $G^w_0=1$ on legal completed states
($S=\emptyset$) and $0$ on the other legal states, and
\[
G^w_{r+1}(S,J)=\max_{B\subseteq S}\E\big[G^w_r(S\setminus B,J-X_B);\ \text{child legal}\big].
\]
\emph{Theorem (bounded-weight envelope).} For every legal state and every $r\ge1$,
\[
G^w_r(S,J)\le\min\big\{1,\ A_r(d+1)^{-2^{-r}}\big\},\qquad A_r=2\Lambda(12\Lambda)^{2-2^{2-r}}\le K_\Lambda:=288\Lambda^3 .
\]
This $A_r$ depends on $\Lambda$; it is not the constant $A_r<36$ of~\eqref{eq:env}. (At $r=0$ the formula fails at the completed state.) Every causal randomized policy with fresh exact-marginal draws (the
class of Section~\ref{sec:setup} and Appendix~\ref{app:tgenv}, with the seed, the committed history, the full residual
weight array and any batch identities available) reaches a legal completion within $r$ rounds with probability at most
$G^w_r$ of the starting state. Hence, with $d_0=\min(k,n-k)$:
\begin{itemize}
\item[(a)] every $\pi\in\mathrm{RAND}_r$ has $\KL(p_w\|Q_\pi)\ge2^{-r}\ln(d_0+1)-\ln K_\Lambda$; so for $d_0\ge1$, output KL at
most $\varepsilon$ needs $r\ge\log_2\ln(d_0+1)-\log_2(\varepsilon+\ln K_\Lambda)$;
\item[(b)] every causal policy without a cap whose output KL is at most $\varepsilon$ has
$\E\tau\ge1+\sum_{t\ge1}(e^{-\varepsilon}-G^w_t(\mathrm{root}))_+$; for $d_0\ge1$ this is at least
$e^{-\varepsilon}\log_2\ln(d_0+1)-c(K_\Lambda)$, with $c(K)=\sum_{i\ge0}\min\{1,Ke^{-2^i}\}\le\log_2\ln K+2.42$ for $K\ge e$. For $d_0=0$
the target is a point mass, and one round committing every coordinate has output KL $0$;
\item[(c)] if $\kappa_0n\le k\le(1-\kappa_0)n$ and $n\ge\kappa_0^{-2}$, then, uniformly in the weights,
$R^w_F(n,\varepsilon)\ge\log_2\ln n-1-\log_2(\varepsilon+\ln K_\Lambda)$ and $\E\tau\ge e^{-\varepsilon}\log_2\ln n-e^{-\varepsilon}-c(K_\Lambda)$.
\end{itemize}
Under the hypotheses of the Remark after the \S7 table, its schedule mixture (in $\mathrm{RAND}_{\lfloor\log_2\ln n+C_R\rfloor}$, for
$n\ge\lfloor(M^*-1)/\rho\rfloor+1$) and its \S9 dilution (a predrawn mixture, so $\tau=|U|$) attain these orders. So the
least adaptive hard cap is $\log_2\ln n+O_{\varepsilon,\Lambda,\kappa_0,\rho}(1)$, and the least expected executed rounds are
$(e^{-\varepsilon}+o(1))\log_2\ln n$, over all causal policies. Without a homogeneous group, the predrawn mixtures of Appendix~\ref{app:rf} (Corollaries~A$'$ and~A$''$) attain
the same orders under (c)'s hypotheses: the least adaptive hard cap is $\log_2\ln n+O_{\varepsilon,\Lambda,\kappa_0}(1)$, the lower
constant in (c) explicit and the upper one existence-only and $O_{\Lambda,\kappa_0}(1+\varepsilon^{-2})$, and the least expected executed rounds
are $(e^{-\varepsilon}+o(1))\log_2\ln n$. Nothing here concerns $H$ under adaptive selection. $K_\Lambda$ grows with $\Lambda$, so
Appendix~\ref{app:tgenv}'s $V$-based two-group envelope, whose constants are free of $\lambda$, remains the form for
unbounded tilts.

\emph{Proof.} \emph{Policies.} For a causal policy, let $\sigma_t$ be the state after round $t$ and $\mathcal F_t$ the
$\sigma$-field of the seed and the committed history; put $Y_t=G^w_{r-t}(\sigma_t)\mathbf 1\{\sigma_t\text{ legal}\}$.
\begin{itemize}
\item An illegal state stays illegal, and a completed state keeps $G^w_s=1$ (empty batches).
\item On a legal $\sigma_t$, the next batch is $\mathcal F_t$-measurable, with any identities, and its draws are
conditionally independent at the marginals of $\sigma_t$. So $\E[Y_{t+1}\mid\mathcal F_t]\le Y_t$ by the definition of $G^w$.
\item Hence the success probability within $r$ rounds is $\E Y_r\le Y_0$. The quantity $d$ enters only as a bound on $G^w$
that is uniform over residual arrays of ratio at most $\Lambda$; counts are never assumed to be a sufficient state.
\end{itemize}
\emph{Inputs.} The weights of every residual array have ratio at most $\Lambda$. By W1, at every legal state
$p_i\le\Lambda p_\ell$, $1-p_i\le\Lambda(1-p_\ell)$ and $v_i\ge v_\ell/\Lambda$, where $v=p(1-p)$. With $\sum_Sp_i=J$ this gives
$p_i\le\Lambda J/M$, $1-p_i\le\Lambda(M-J)/M$, $v_i\ge V/(\Lambda M)$ with $V:=\sum_Sv_\ell$, and
$v_i\ge\min(p_i,1-p_i)/2\ge d/(2\Lambda M)$, so $V\ge d/(2\Lambda)$. At $d=0$ every bound below is trivial.

\emph{Base.} A completion in one round needs $B=S$ and $X_S=J$. By Lemma~TG1 of Appendix~\ref{app:tgenv} (a sum of independent Bernoulli variables whose variances sum to $v>0$ has every atom at most $\sqrt{\pi/(8v)}$), for
$d\ge1$ this has probability at most $\sqrt{\pi/8}/\sqrt V\le\sqrt{\pi\Lambda/(4d)}\le2\Lambda(d+1)^{-1/2}$.

\emph{One step.} Fix a legal state with $d\ge1$ and a nonempty batch of size $b$, with child $(M-b,J-X_B)$ and $d'$ its
minority quota. Let $T:=\E[(d'+1)^{-1/2};\ \text{child legal}]$. We show $T\le12\Lambda(d+1)^{-1/4}$.
\begin{itemize}
\item If $b\le M/(4\Lambda)$, then $\E X_B\le J/4$ and $\E(b-X_B)\le(M-J)/4$, and each has variance at most its mean.
Off $\{X_B>J/2\}\cup\{b-X_B>(M-J)/2\}$ the child is legal with $d'\ge d/2$, and Chebyshev bounds that union by
$4/J+4/(M-J)\le8/d$. So $T\le(d/2+1)^{-1/2}+8/d\le10.49(d+1)^{-1/4}$.
\item If $b>M/(4\Lambda)$, then $\mathrm{Var}\,X_B>V/(4\Lambda^2)\ge d/(8\Lambda^3)$, so every atom of $X_B$ is at most
$A:=\min\{1,\sqrt{\pi\Lambda^3/d}\}$ (Lemma~TG1). Each value of $d'$ comes from at most two legal child quotas. Splitting at
$d'<u=\lceil1/A\rceil$ gives $T\le4A\sqrt u+(u+1)^{-1/2}\le(4\sqrt2+1)\sqrt A\le10.54\,\Lambda^{3/4}(d+1)^{-1/4}$. Both bounds are at most $12\Lambda(d+1)^{-1/4}$, as $1\le\Lambda^{3/4}\le\Lambda$.
\end{itemize}
The last round, $B=S$, is a large batch.

\emph{Induction.} Let $\alpha=2^{-r}\le\frac12$. On the killed (substochastic) child kernel, Hölder gives
$\E[(d'+1)^{-\alpha};\text{legal}]\le T^{2\alpha}$.
\begin{itemize}
\item For a nonempty batch, $\E[G^w_r(\text{child});\text{legal}]\le A_rT^{2\alpha}\le A_r(12\Lambda)^{2^{1-r}}(d+1)^{-2^{-(r+1)}}$.
\item The empty batch gives $G^w_r$ of the current state, which is at most the same bound.
\item So $A_{r+1}=A_r(12\Lambda)^{2^{1-r}}$ with $A_1=2\Lambda$, which gives $A_r=2\Lambda(12\Lambda)^{2-2^{2-r}}$.
\end{itemize}
\emph{Costs.}
\begin{itemize}
\item (a): $\KL\ge-\ln Q_\pi(S_k)\ge-\ln G^w_r(\mathrm{root})$.
\item (b): $\Pr(\tau>t)\ge(e^{-\varepsilon}-G^w_t(\mathrm{root}))_+$ for $t\ge1$, as in Corollary~(d) of Appendix~\ref{app:tgenv}. Keeping
$t\le\lfloor L\rfloor$ with $L=\log_2\ln(d_0+1)$ and using $(d_0+1)^{-2^{-t}}=e^{-2^{L-t}}$ gives the bound for $L\ge1$ (for $t\le\lfloor L\rfloor$, $e^{-2^{L-t}}\le e^{-2^i}$ with $i=\lfloor L\rfloor-t$, and $1+\lfloor L\rfloor e^{-\varepsilon}\ge e^{-\varepsilon}L$); for $L<1$
it follows from $\tau\ge1$. The terms $i\le\log_2\ln K$ of $c(K)$ are at most $1$ each, and the rest sum to less than $1.42$.
\item (c): $\ln(d_0+1)\ge\ln(\kappa_0n)\ge\frac12\ln n$.
\item The attainment statements import the Remark, \S9 and Appendix~\ref{app:rf} (Corollaries~A$'$, A$''$) unchanged. \qed
\end{itemize}
\pend{}

\clearpage
\section{Reservoir-free upper bound for bounded weights}\label{app:rf}\pend{}

\makeatletter
\@ifundefined{rfThmA}{\newtheorem*{rfThmA}{Theorem A}}{}%
\@ifundefined{rfCorAp}{\newtheorem*{rfCorAp}{Corollary A\('\)}}{}%
\@ifundefined{rfCorApp}{\newtheorem*{rfCorApp}{Corollary A\(''\)}}{}%
\@ifundefined{rfLemTone}{\newtheorem*{rfLemTone}{Lemma T1}}{}%
\@ifundefined{rfLemNCD}{\newtheorem*{rfLemNCD}{Lemma NCD}}{}%
\@ifundefined{rfproof}{\newenvironment{rfproof}[1]{\par\normalfont
  \topsep6\p@\@plus6\p@\relax\trivlist\item[]\textbf{#1}\hspace\labelsep\ignorespaces}%
  {\qed\endtrivlist\@endpefalse}}{}%
\makeatother

\pend{}
\pend{}
This appendix proves Theorem~A and Corollaries~A$'$ and~A$''$ below: for any weights with ratio at most $\Lambda$ and an interior quota, a finite predrawn mixture of full labelled schedules, built from randomly kept sets, has output KL $O(1/A)$, and each of its components has at most $\log_2\ln n+O(A^2)$ rounds. Here $A\ge A_0$ is a free parameter: Corollary~A$'$ fixes it by the target error $\varepsilon$ and bounds the hard cap, and Corollary~A$''$ lets it grow slowly with $n$, so that its error bound $C^*/A_n$ vanishes, and bounds the scheduled mean of a mixture with the one-block schedule. These corollaries supply the upper bounds of Theorem~\ref{thm:tilt}(a) on $R_F^{\mathrm{adapt}}$, on $\LF^w$ and on the least expected executed rounds for arbitrary bounded weights (Appendix~\ref{app:t1}, Assembly of Theorem~\ref{thm:tilt}); with an equal-weight group of linear size, Appendix~\ref{app:t1}, \S\S6, 7 and~9 give these bounds with explicit constants, and also the common length of Theorem~\ref{thm:tilt}(b), which the mixtures here do not give. No homogeneous group is used. Survival is still necessary, since $F\ge-\ln Q(S_k)$ by (4.33), where $S_k=\{x\in\{0,1\}^n:\sum_ix_i=k\}$ is the slice, but it does not control the output law of a weighted target (Appendix~\ref{app:t1}, \S8); so the proof bounds $F$ itself, level by level, and then reads off survival, which a first-exit argument also bounds directly ((4.31)--(4.32)), and the within-slice shape, which (4.33) bounds by $F$ (\S4.8). It compares each round with the reference law $\mathcal R$ below: averaged over the random kept set, $\mathcal R$ is close to the target from both sides, off exceptional sets of small target and reference-law mass ((3.34)--(3.35)), and this averaging replaces the orbit averaging of Section~\ref{sec:bal}.

\S1 collects tools for weighted slices, among them Lemma~T1 (sub-Gaussian linear statistics); \S2 proves a uniform lattice estimate and the integrated-normalizer bound (2.17); \S3 compares one step of the sampler with the reference law $\mathcal R$ of Section~\ref{sec:tilt}, which completes the kept set exactly: the identities (3.1)--(3.2) write $\mathcal R/p$, up to quota normalizers, as an empty-child term plus an integral of product tilts, (3.20) bounds their coefficients once the common one is removed on the slice, (3.18) and (3.21) bound the one-step divergence and the second moment of $p/\mathcal R$, and (3.35) gives separate target-law and execution-law exceptional sets; and \S4 constructs the kept-set ladder (\S\S4.1--4.2), composes the one-step bounds through the signed recursion (4.15)--(4.20), sums it in (4.28)--(4.30), bounds survival and the within-slice shape in (4.31)--(4.33), and derives Corollaries~A$'$ and~A$''$ in (4.34)--(4.36). \S5 fixes the constants in a noncircular order, and \S6 gives the dependency graph. The matching lower bounds, over all causal policies, are those of Appendix~\ref{app:t1}, \S10.

\subsection*{0. Statement}

\pend{}

\textbf{Hypotheses.}
\begin{itemize}
\item \(\Lambda\ge1\) and \(\kappa_0\in(0,1/2]\) are fixed.
\item \(n\ge1\) and \(w\in(0,\infty)^n\) with \(\max w/\min w\le\Lambda\). Rescale so \(\min w=1\). No homogeneity is assumed.
\item \(k\in[\kappa_0n,(1-\kappa_0)n]\), and \(p=p_w\) on \(S_k\).
\end{itemize}

\textbf{Constants.} \(C\) and \(c\) are positive constants depending only on \((\Lambda,\kappa_0)\); their values change from line to line. Each is shown to exist, and none is evaluated. ``\(A\ge A_0\)'' means \(A\ge A_0(\Lambda,\kappa_0)\), with \(A_0\) chosen in \S5 to meet the simultaneous conditions listed there.

\begin{rfThmA}
There exist \(A_0\) and \(C^*\) such that for all \(A\ge A_0\) and all admissible \(n\), \(w\), \(k\), the predrawn mixture \(U(A)\) of \S\S4.1--4.2 satisfies:
\begin{enumerate}
\item It is a finite mixture of full labelled schedules, with output law \(Q\).
\item Its hard cap is \(\le\log_2\ln n+8A^2+5\) when \(n\ge4m_0=4\lceil4A^2\rceil\), and is \(n<16A^2+4\) otherwise (the exact sampler).\pend{}
\item \(F=\KL(p\|Q)\le C^*/A\).
\item The survival term satisfies \(-\ln Q(S_k)\le C\cdot e^{-cA^{2/3}}\).
\item The shape term satisfies \(\KL(p\|Q(\cdot\mid S_k))=F+\ln Q(S_k)\le F\).
\end{enumerate}
The schedule law depends only on \((n,A)\), not on \(w\).
\end{rfThmA}

\begin{rfCorAp}[fixed \(\varepsilon\)]
Take \(A=\max(A_0,C^*/\varepsilon)\). Then \(F\le\varepsilon\), and the hard cap is \(\le\log_2\ln n+O_{\Lambda,\kappa_0}(1+\varepsilon^{-2})\), uniformly over arrays and quotas.\pend{}
\end{rfCorAp}

\begin{rfCorApp}[scheduled mean]\leavevmode
\begin{itemize}
\item Let \(A_n=\max(A_0,(\log\log n)^{1/4})\) and \(\varepsilon_n=C^*/A_n\to0\). For large \(n\) set \(1-\beta=e^{-(\varepsilon-\varepsilon_n)}\).
\item Let \(\mu'=(1-\beta)\cdot U(A_n)+\beta\cdot(\text{the 1-block schedule})\).
\item By the pointwise mixture bound (4.35), \(F\le\varepsilon\).
\item With \(\tau\) the number of rounds of the drawn schedule, as in (4.8), the scheduled mean is \(C(\mu')=\E\tau\le e^{-\varepsilon+\varepsilon_n}(\log_2\ln n+8A_n^2+5)+\beta=(e^{-\varepsilon}+o(1))\cdot\log_2\ln n\).
\item The hard cap of \(\mu'\) is \(\log_2\ln n+O(\sqrt{\log\log n})\).
\end{itemize}
\end{rfCorApp}

\subsection*{Scope and notation of the proof}

\pend{} This proof establishes Theorem A, its survival estimate, and Corollaries A\('\) and A\(''\) of \S{}0, with the stated parameter dependence.\pend{}

The target is the positive weighted fixed-quota law $p_w$ of \S{}0 (the bounded-ratio weighted slice of Appendix~\ref{app:t1}, \S10), not a full-support target.\pend{} The schedule is exactly the kept-set mixture \(U(A)\).\pend{} No homogeneous reservoir is used. All parameters called constants below depend at most on the fixed pair $(\Lambda, \kappa_0)$, unless explicitly stated otherwise. The appendix does not establish an $A^{-2}$ rate, an $O(1+\varepsilon^{-1})$ additive cap, a numerical $A_0$, practical speed, robustness to learned conditionals, a full-support extension, or any statement about $H$. Theorem~A, Corollaries~A\('\) and~A\(''\) and Lemma~T1 are also kernel-checked; Appendix~\ref{app:lean} lists the declarations and the parts proved only by ordinary (paper) arguments, which are not kernel-checked.

\pend{}

\textbf{Notation discipline.} In a one-step lemma, $N$ is the actual number of free coordinates, $J$ the actual quota, $\pi_i$ the exact target marginal, $q_i$ the independent logistic marginal at mean $J$, and $V=\sum q_i(1-q_i)$. A parent $(F,J)$ is \emph{interior} if $\kappa N\le J\le(1-\kappa)N$ for a fixed $\kappa\in(0,1/2]$; the proof of Theorem~A takes $\kappa=\kappa_0/2$, the density bound of every good state (\S4.3), and child interiority $\theta=\kappa/2$ (\S2.4). In \S\S2--3, $F$, $C$ and $B$ denote coordinate sets and $U$ an integer quota, and $A(u)$ of \S3.1 is a function of $u$; none of them is the criterion $F$, the schedule law $U(A)$ or its parameter $A$. Locally, $w=\sqrt{m\log m}$ in \S2.4 is a window half-width, not a weight; $\rho=J/N$ in \S3.6 is the parent density, not the ratios $\rho_i$ of (3.1) or the density $\rho^b$ of \S3.3; the coefficient $D$ of (4.22) is not a divergence; and $B$ in \S4.7 is a number, not a block. The letter $C$ also denotes generic constants, as in (3.25); a set $C$ appears as an index set or under $|\cdot|$, $\in$ or $\subseteq$. In the ladder, $M_t$ remains a nominal size. Write $\ell=f \sqrt{N}$ for the actual one-step scale and $L_t=f_t \sqrt{M_t}$ for its nominal counterpart. Also: $B_N=N\log(\Lambda N)$ in \S4.5 is the atom bound, not a block; $B_t$ in \S4.9 is the large block of level $t$; $N_0^b$ of \S3.3 is the bulk normalization, not the dimension threshold $N_0$; $E(u)$ in \S3.3 is the error term of (3.9), not the sum $E$ of \S4.7; and $e$ in \S3.3 is a generic nonnegative error. The reference law of \S3 is denoted $\mathcal R$;\pend{} it is not the schedule-averaged KL criterion $H$ of the main text.\pend{}

Throughout, logarithms are natural except $\log_2$ and the explicitly base-two ladder variable. A statement with an analytic threshold applies only above that threshold; the schedule's small-dimension branch deals with all remaining admissible dimensions.

\subsection*{1. Weighted slices and elementary probability tools}

\paragraph{1.1 Weighted slices, conditional laws, floors, and pairwise covariance}

For a finite set $F$, positive weights $w$, and $0\le J\le|F|$, put
\[
p_{F,J}(x)=\frac{\prod_{i\in F}w_i^{x_i}}{e_J(w_F)}\mathbf1_{\{|x|=J\}}.
\]
Every feasible coordinate conditioning leaves another such law. Rescaling all weights does not change the law. The two-coordinate numerator comparisons in the proof of Lemma~W1 (Appendix~\ref{app:t1}, \S2) give\pend{}, at every quota $0\le J\le N$ and ratio at most $\Lambda$ (Lemma~W1(b)),
\[
\pi_i\ge \frac{J}{\Lambda N},\qquad
1-\pi_i\ge\frac{N-J}{\Lambda N}.
\tag{1.1}
\]
The endpoint cases are deterministic and are treated directly, without dividing by zero.

Here is an elementary proof of the weak symmetric-polynomial log-concavity needed for the pairwise covariance formula; no negative-association theorem is used below. If a nonnegative sequence $a$ has contiguous positive support and is log-concave, then $b_j=a_j+w a_{j-1}$, for $w>0$, is log-concave. Indeed,
\[
\begin{aligned}
b_j^2-b_{j-1}b_{j+1}
={}&(a_j^2-a_{j-1}a_{j+1})\\
&+w(a_ja_{j-1}-a_{j-2}a_{j+1})
+w^2(a_{j-1}^2-a_{j-2}a_j)\ge0.
\end{aligned}
\]
The middle inequality follows by multiplying the decreasing successive-ratio inequalities; support endpoints follow directly using zeros. Induction on the number of weights, starting from the one-term sequence $(1)$ for no weights, proves $e_j^2\ge e_{j-1}e_{j+1}$ for any positive weights. Consequently, for distinct $i,j$, with $R=F\setminus\{i,j\}$,
\[
\operatorname{Cov}(X_i,X_j)
=\frac{w_iw_j(e_J(R)e_{J-2}(R)-e_{J-1}(R)^2)}{e_J(F)^2}\le0.
\tag{1.2}
\]
This holds for \textbf{all positive arrays}, not just arrays with bounded ratio, and for every further feasible conditioning.

\paragraph{1.2 Sub-Gaussian linear statistics}\leavevmode\par

\pend{}\begin{rfLemTone}[sub-Gaussian linear statistics]\pend{} For every weighted slice, every real vector $c$, and every real $\lambda$,
\[
\mathbb E\exp\{\lambda c\cdot(X-\mathbb EX)\}
\le \exp\{\lambda^2\|c\|_2^2/4\}.
\tag{1.3}
\]
Thus, for $a>0$ and $c\ne0$,
\[
\Pr(|c\cdot(X-\mathbb EX)|\ge a)
\le2\exp\{-a^2/\|c\|_2^2\}.
\tag{1.4}
\]
For $c=0$ the statistic vanishes identically.\end{rfLemTone}

Lemma~T1 is proved below from Lemma~NCD. Route MG is an alternative proof: it bounds the variance of $c\cdot X$ under every
exponential tilt by (1.5), which \S3.5 also uses, and integrates; the remark after both proofs names two further routes, which
are not used.

\begin{rfLemNCD}[negative cylinder dependence]\pend{} For every positive weighted slice $p_{F,J}$ (any finite $F$, positive weights, $0\le J\le|F|$) with marginals $\pi_i$, and every $T$ contained in $F$,
\[
\Pr(X_i=1\ \forall i\in T)\le\prod_{i\in T}\pi_i,\qquad
\Pr(X_i=0\ \forall i\in T)\le\prod_{i\in T}(1-\pi_i).
\tag{1.3a}
\]
The first left side vanishes unless $|T|\le J$, the second unless $|T|\le|F|-J$.\end{rfLemNCD}

\begin{proof}Induct on $|T|$, simultaneously over all positive weighted slices; $|T|\le1$ is trivial. Pick $t$ in $T$. If $\pi_t=0$, the first left side is zero. Otherwise conditioning on $X_t=1$ leaves the slice $p_{F\setminus t,J-1}$ with the same weights (\S{}1.1), so
\[
\Pr(X_T\equiv1)=\pi_t\Pr(X_{T\setminus t}\equiv1\mid X_t=1)
\le\pi_t\prod_{i\in T\setminus t}\Pr(X_i=1\mid X_t=1)
\le\prod_{i\in T}\pi_i .
\]
The first inequality is the induction hypothesis for the conditioned slice. The second is $\Pr(X_i=1\mid X_t=1)-\pi_i=\operatorname{Cov}(X_i,X_t)/\pi_t\le0$, by (1.2) applied to $p_{F,J}$. For the second family apply the first to $1-X$, which has the weighted-slice law with weights $1/w_i$ and quota $|F|-J$. (1.2) is used at every slice reached by conditioning, including deterministic boundary quotas, where the covariance is zero or the bracket in (1.2) is $-e_{J-1}(R)^2\le0$.\end{proof}

\pend{}

\begin{rfproof}{Proof of Lemma T1 from NCD.}\pend{}Write $Y=c\cdot(X-\E X)$. Split the coordinates into $I_+=\{i: \lambda c_i>0\}$ and $I_-=\{i: \lambda c_i<0\}$. Cauchy--Schwarz gives $\E e^{\lambda Y}\le(\E e^{2 \lambda Y_+})^{1/2}(\E e^{2 \lambda Y_-})^{1/2}$.

\begin{itemize}
  \item On $I_+$, write $e^{2 \lambda c_i(X_i-\pi_i)}=e^{-2 \lambda c_i \pi_i}(1+a_iX_i)$ with $a_i=e^{2 \lambda c_i}-1>0$. Expanding the product, $\E \prod_{I_+}(1+a_iX_i)=\sum_{T \subseteq I_+} (\prod_T a_i) \Pr(X_T\equiv1)\le\prod_{I_+}(1+a_i \pi_i)$, by the first inequality of (1.3a).
  \item On $I_-$, write the factors as $e^{2 \lambda c_i(1-\pi_i)}(1+b_i(1-X_i))$ with $b_i=e^{-2 \lambda c_i}-1>0$, and use the second inequality.
\end{itemize}

So each half is at most the product of the independent-coordinate moment generating functions. Hoeffding's lemma for one Bernoulli (\S{}1.3) gives $\E e^{2 \lambda c_i(X_i-\pi_i)}\le e^{\lambda^2 c_i^2/2}$. Taking square roots proves (1.3), and Chernoff optimization proves (1.4).\end{rfproof}

\pend{}

\pend{}

\begin{rfproof}{Alternative proof of Lemma T1 (route MG).}\pend{}Let $\Sigma$ be the covariance matrix under any positive external tilt of the weights. Fixed total implies $\Sigma\mathbf1=0$; (1.2) implies $\Sigma_{ij}\le0$ for $i\ne j$. Hence
\[
\begin{aligned}
c^\top\Sigma c
&=\sum_{i<j}(-\Sigma_{ij})(c_i-c_j)^2\\
&\le2\sum_i c_i^2\sum_{j\ne i}(-\Sigma_{ij})
=2\sum_i c_i^2\operatorname{Var}(X_i)
\le\frac12\|c\|_2^2.
\end{aligned}
\tag{1.5}
\]
For $g(t)=\log\E\exp(t c\cdot X)$, the tilted law has weights $w_i \exp(t c_i)$, so $g''(t)$ is this variance for every real $t$. Integrating the bound twice, in either direction from zero, proves (1.3). Chernoff optimization at $\lambda=2a/\|c\|_2^2$, and then the negative tail, proves (1.4). This argument needs pairwise negativity \textbf{at all external tilts}, plus the fixed-sum identity; pairwise negativity at one distribution alone would not suffice. No ratio bound is imposed on these auxiliary external tilts.\end{rfproof}

\pend{} Either route, NCD or MG (alternative), proves Lemma T1 with the same constant.\pend{} No full negative-association theorem is used anywhere in the dependency graph.\pend{} Both routes need only (1.2) for all positive weights: NCD at every slice reached by conditioning, MG at every slice reached by an external tilt.

\textbf{Remark (an alternative completion via negative association, not used).}\pend{} \citet[\S4.2, Theorem~4.8, in the arXiv:0707.2340v2 numbering]{borcea2009negative} gives conditional negative association for a conditioning-closed class of finite, fixed-degree laws with pairwise negative correlations. The class of all positive weighted slices, including deterministic boundaries, satisfies those hypotheses by (1.2).\pend{} Theorem 4.9 gives the strongly-Rayleigh route, but is unnecessary here.\pend{}\pend{} A second optional route, also not used, is \citet[Thm.~2.8, p.~290, with Property~\(P_2\), p.~288]{joagdev1983negative}.\pend{}\pend{}

\paragraph{1.3 Independent bounded variables and kept-set counts}

For a variable $Y$ in $[a,b]$, every exponential tilt still has support in that interval and therefore variance at most $(b-a)^2/4$: compare with the squared distance to its midpoint. Twice integrating its centered log-MGF proves Hoeffding's lemma,
\[
\log\mathbb E e^{\lambda(Y-\mathbb EY)}\le\lambda^2(b-a)^2/8.
\]
Independence gives the sum version and optimization gives
\[
\Pr\left(\left|\sum(Y_i-\mathbb EY_i)\right|\ge t\right)
\le2\exp\left(-\frac{2t^2}{\sum(b_i-a_i)^2}\right).
\tag{1.6}
\]
All later conditional uses first fix the kept set or the first-block outcomes, after which the asserted product variables are independent.

For a sum $Z$ of independent, not necessarily identical Bernoulli variables with mean $\mu$,
\[
\mathbb E e^{\lambda Z}\le\exp\{\mu(e^\lambda-1)\}.
\]
Chernoff optimization gives the two usual relative-deviation exponents. The elementary inequalities $(1+d)\log(1+d)-d\ge d^2/3$ and $d+(1-d)\log(1-d)\ge d^2/2$, for $0\le d\le1$, give
\[
\Pr(|Z-\mu|\ge d\mu)\le2e^{-\mu d^2/3}.
\tag{1.7}
\]
These inequalities also follow by differentiating their differences, starting at zero. This is the precise count bound used below; an independently forced one-coordinate atom changes counts by at most one.

\paragraph{1.4 KL facts and sub-probability conventions}

Throughout, $D(p\|q)$ is $\KL(p\|q)$. Whenever $q$ is a sub-probability on the support of $p$, extend it to a probability by adding one cemetery atom, with zero $p$ mass there. Thus the usual KL, log-sum, and Pinsker arguments apply without normalizing away survival loss. All laws here
are on finite sets, and (1.8)--(1.10) are stated for a probability $p$ and a sub-probability $q$ with $q>0$ on the
support of $p$. Where \S4.5 uses them to bound expectations, in (4.18), $q$ is the continuation law at a good child,
which is positive on the target slice by the argument after (4.13).

For completeness, log-sum follows from Jensen under normalized $p$ on each part of a partition. Apply it to the set on which $p>q$ and its complement. The binary KL, as a function of its first parameter, has second derivative $1/[u(1-u)]\ge4$; integrating from its minimizer gives binary KL at least twice the squared parameter difference. Hence
\[
\operatorname{TV}(p,q)\le\sqrt{D(p\|q)/2}.
\tag{1.8}
\]
For $W=\log(p/q)$ under $p$,
\[
\mathbb E W^-\le\operatorname{TV}(p,q),\qquad
\mathbb E(W^-)^2\le1,
\tag{1.9}
\]
because $\log y\le y-1$ and $(\log y)^2\le y-1$ for $y\ge1$. For the latter scalar inequality, the derivative of $y-1-(\log y)^2$ is $1-2 \log(y)/y\ge1-2/e>0$. In particular,
\[
\mathbb E W^+\le D(p\|q)+\sqrt{D(p\|q)/2}.
\tag{1.10}
\]
These estimates will control signed continuation log ratios, not just their positive parts.

\subsection*{2. Logistic tilt, lattice approximation, and the integrated normalizer}

\paragraph{2.1 Tilt parameter and uniform interiority}

For a set of $m$ coordinates define
\[
K(s)=\sum_{i=1}^m\log(1+w_i e^s),\quad
\mu(s)=K'(s),\quad v(s)=K''(s).
\]
For $m\ge1$ the derivative $v$ is positive, and the mean increases continuously from zero to $m$, so every mean in $(0,m)$ has a unique tilt. Independent Bernoulli probabilities are $p_i(s)=w_i e^s/(1+w_i e^s)$. Conditioning their product law on total $U$ gives the weighted slice, independently of $s$; this is the conditional Poisson (rejective) representation of \citet{hajek1964asymptotic} (cf.\ \citealp{bertail2016sharp} for the invariance under a common odds rescaling). For a coordinate set $C$, write $\widetilde P_s$ for this product law at tilt $s$, $S_C=\sum_{i\in C}X_i$, and $s_C(U)$ for the tilt at which $S_C$ has mean $U$; at a parent $(F,J)$, $s_0=s_F(J)$, so the $q_i$ of the notation paragraph are the $p_i(s_0)$.

If the mean density is in $[\theta,1-\theta]$ and the weight ratio is at most $\Lambda$, probability and complementary-probability comparisons give
\[
p_i(s)\in[\theta/\Lambda,1-\theta/\Lambda],\qquad
c_{\theta,\Lambda}m\le v(s)\le m/4.
\tag{2.1}
\]
The same interiority for exact parent marginals follows from (1.1).

\paragraph{2.2 Uniform lattice Edgeworth estimate}

Let $S=\sum_{i=1}^m \operatorname{Bern}(p_i)$ independently, with $p_i \in [a,1-a]$, where $0<a\le1/2$. Write $\mu=\E S$, $s^2=\operatorname{Var}(S)$, $\gamma_3=\sum p_i(1-p_i)(1-2p_i)$, and $y=(k-\mu)/s$. Then there is a finite $K_a$, independent of $m$, $k$, and the array, such that
\[
\sup_{k\in\mathbb Z}\left|
\Pr(S=k)-\frac{\varphi(y)}s
\left(1+\frac{\gamma_3}{6s^3}(y^3-3y)\right)
\right|\le K_a s^{-3},
\tag{2.2}
\]
where $\varphi$ is the standard normal density. Edgeworth expansions of such atoms are classical in rejective sampling; for instance, \citet[Lemma~1]{boistard2012approximation} expand the probability that such a sum equals its integer mean, with relative error $O(s^{-4})$. We prove (2.2) in the form used here, with a constant uniform over arrays with $p_i\in[a,1-a]$.

\textbf{Fourier derivation.} For the uncentered characteristic function $\Phi$, orthogonality of integer exponentials gives inversion on $[-\pi,\pi]$. Also
\[
|1-p+pe^{it}|^2=1-2p(1-p)(1-\cos t),
\]
so
\[
|\Phi(t)|\le e^{-2s^2t^2/\pi^2}\quad(|t|\le\pi).
\tag{2.3}
\]
On a fixed sufficiently small interval about zero, every factor is uniformly separated from zero. Taylor expansion of its logarithm, with the fourth derivative bounded uniformly for $p \in [a,1-a]$, gives
\[
\log(e^{-i\mu t}\Phi(t))
=-s^2t^2/2-i\gamma_3t^3/6+R(t),\qquad |R(t)|\le C m t^4.
\tag{2.4}
\]
The derivative bound can also be obtained directly from the rational derivatives of $\log(1-p+p e^{it})$; the denominator stays separated from zero on the chosen interval. Since $s^2\ge a(1-a)m$, choose a smaller $t_1=t_1(a)>0$ so that $Cm t^4\le s^2t^2/4$ for $|t|\le t_1$. Using $|e^{ib}-1-ib|\le b^2/2$ for real $b$ yields
\[
\left|e^{-i\mu t}\Phi(t)
-e^{-s^2t^2/2}(1-i\gamma_3t^3/6)\right|
\le C e^{-s^2t^2/4}(m t^4+m^2t^6).
\tag{2.5}
\]
The integral of the right side on the real line is at most
\[
C(ms^{-5}+m^2s^{-7})\le C_a s^{-3}.
\]
Outside $[-t_1,t_1]$, (2.3), together with the Gaussian-polynomial tail of the approximant, bounds both the remaining inversion interval and its extension to the real line by $C_a(1+m)e^{-c_a s^2}$: on
$t_1\le|t|\le\pi$, (2.3) gives $|\Phi(t)|\le e^{-2s^2t_1^2/\pi^2}$, so that part of the inversion integral is at most
$2\pi e^{-2s^2t_1^2/\pi^2}$; for the approximant, with $|\gamma_3|\le m$,
$\int_{|t|\ge t_1}e^{-s^2t^2/2}\big(1+\tfrac{|\gamma_3||t|^3}6\big)dt\le2e^{-s^2t_1^2/2}\big(\frac1{s^2t_1}+\frac{m(s^2t_1^2+2)}{6s^4}\big)$;
since $s^2\ge a(1-a)m$, both pieces are at most $C_a(1+m)e^{-c_as^2}$ with $c_a=2t_1^2/\pi^2$. This too is at most $K_a s^{-3}$; small $m$ are covered by increasing $K_a$. The inverse Gaussian transform and its third derivative give (2.2), with the displayed sign: each $y$-derivative of
$\varphi(y)=\frac1{2\pi}\int e^{-i\tau y}e^{-\tau^2/2}d\tau$ brings down $-i\tau$, so
$\frac1{2\pi}\int e^{-i\tau y}(i\tau)^3e^{-\tau^2/2}d\tau=-\varphi'''(y)=(y^3-3y)\varphi(y)$. The Gaussian transform itself follows by integration by parts from its differential equation and its value at zero.

This proves a lattice, triangular-array-uniform estimate. No derivative of its error term is used later.

\pend{}

\paragraph{2.3 Parent mass, exact marginals, and conditional central atoms}

Apply (2.2) at a mean that is an integer. The cubic correction vanishes at zero, giving
\[
P_0:=\widetilde P_{s_0}(S_F=J)
=(2\pi V)^{-1/2}(1+O(V^{-1})).
\tag{2.6}
\]
For one coordinate, write
\[
\pi_i=\frac{q_i R_i}{q_iR_i+1-q_i},\qquad
R_i=\frac{\Pr(S_{-i}=J-1)}{\Pr(S_{-i}=J)}.
\]
Under the tilt $s_0$ the leave-one-out sum $S_{-i}$ has mean $J-q_i$ and variance $s_{-i}^2=V-q_i(1-q_i)\ge V-\frac14$. The
two atoms in $R_i$ sit at $y=(q_i-1)/s_{-i}$ and $y=q_i/s_{-i}$, so $|y|\le1/s_{-i}$. At both atoms, (2.2) gives
$\Pr(S_{-i}=k)=(2\pi s_{-i}^2)^{-1/2}(1+O(V^{-1}))$: the Gaussian factor is $e^{-y^2/2}=1+O(V^{-1})$; since
$|\gamma_3|\le s_{-i}^2$, the cubic term is at most $s_{-i}^2(3|y|+|y|^3)/(6s_{-i}^3)=O(V^{-1})$; and the remainder
$K_as_{-i}^{-3}$ is $O(V^{-1})$ relative to the main term $\ge c/s_{-i}$. Hence $R_i=1+O(V^{-1})$, and
$\pi_i-q_i=q_i(1-q_i)(R_i-1)/(q_iR_i+1-q_i)=O(V^{-1})$. Both $\sum_i\pi_i$ and $\sum_iq_i$ equal $J$, and $V\ge c_{\theta,\Lambda}N$
by (2.1). Therefore
\[
|\pi_i-q_i|\le C/V\le C/N,\qquad
\sum_i(\pi_i-q_i)=0.
\tag{2.7}
\]
This is the known approximation of rejective (conditional Poisson) inclusion probabilities by the canonical Poisson probabilities, here the logistic $q_i$ with $\sum_iq_i=J$ \citep{hajek1964asymptotic}; see also \citet[eq.~(2.11), arXiv v1 numbering]{boistard2012approximation} for the leading term and \citet[Lemma~6]{bertail2016sharp} for a nonasymptotic version.
For an interior child of size $m$, the same central estimate gives
\[
\widetilde P_{s_C(U)}(S_C=U)\ge c/\sqrt m
\tag{2.8}
\]
above one uniform threshold.\pend{}

\paragraph{2.4 The normalizer and its full change of variables}

In \S3.1 the reciprocal of the quota probability $\widetilde P_{s_0}(S_C=U)$ of a coordinate set $C$ of size $m$ (there, a kept set) is written as an integral over tilts divided by the normalizer $I_C(U)$ below. The estimate (2.17) shows that this normalizer is within $O(\sqrt{\log m}/m)$ of one at interior quotas, so that for interior children the normalizers cost only that factor in (3.17) and \S3.8.

Let $U/m \in [\theta,1-\theta]$ and
\[
I_C(U)=\int_{\mathbb R}\widetilde P_s(S_C=U)K''(s)\,ds.
\]
The substitution $\mu=K'(s)$ gives the exact identity
\[
I_C(U)=\int_0^m\Pr_{s(\mu)}(S_C=U)\,d\mu.
\tag{2.9}
\]
In particular, the integral is finite and positive. For nonempty $C$, this remains true at quotas zero and $m$, although the following estimate is asserted only at interior quotas.

Choose $w=\sqrt{m \log m}$ and then an analytic threshold $m_{\rm an}(\theta,\Lambda)$ so that $w\le\theta m/2$. On the complement of $|\mu-U|\le w$, (1.6) gives the deliberately coarse bound
\[
\int_{|\mu-U|>w}\Pr_{s(\mu)}(S_C=U)\,d\mu
\le 2m e^{-2w^2/m}=2/m.
\tag{2.10}
\]
Inside the window all tilt probabilities are uniformly interior by (2.1), so (2.2) applies. Its integrated absolute remainder is at most
\[
2w K m^{-3/2}=C\sqrt{\log m}/m.
\tag{2.11}
\]
It remains to integrate its two explicit terms.

Write $v=v(\mu)$, $s_\mu=\sqrt{v(\mu)}$,
\[
y=\frac{U-\mu}{s_\mu},\qquad
a(\mu)=\frac{\gamma_3(\mu)}{v(\mu)^{3/2}}.
\]
The Bernoulli cumulant identities give
\[
\frac{dv}{d\mu}=\frac{\gamma_3}{v},\quad
\frac{d\gamma_3}{d\mu}=\frac{\gamma_4}{v},\quad
|\gamma_3|\le v,\quad |\gamma_4|\le v.
\]
Here $\gamma_4=\sum p_i(1-p_i)(1-6p_i(1-p_i))$, which proves its absolute bound directly. Consequently, throughout the window,
\[
|v'|\le1,\quad |v''|\le C/m,\quad
|a|\le C m^{-1/2},\quad
|a'|\le C m^{-3/2}.
\tag{2.12}
\]
The Jacobian is \textbf{exactly}
\[
\frac1{s_\mu}\left|\frac{d\mu}{dy}\right|
=\frac1{1+\tfrac12 y a(\mu)}.
\tag{2.13}
\]
Indeed, $dy/d\mu=-(1/s_\mu)(1+y a/2)$. Since $|y|\le C \sqrt{\log m}$, its parenthesis is at least one half above a uniform threshold, so this change is monotone. Let $a_*=a(U)$. Since $|\mu-U|=s_\mu|y|$, (2.12) gives
\[
|a(\mu)-a_*|\le C|y|/m.
\tag{2.14}
\]
The Gaussian plus Edgeworth integrand, including the Jacobian, is therefore
\[
\begin{aligned}
\varphi(y)\frac{1+a(\mu)(y^3-3y)/6}{1+y a(\mu)/2}
=\varphi(y)\left[
1+a_*\left(\frac{y^3}{6}-y\right)+R_m(y)\right],\\
|R_m(y)|\le\frac C m(y^2+y^4).
\end{aligned}
\tag{2.15}
\]
For example, subtract $1+a_*(\mathrm{He}_3/6-y/2)$ from the fraction, use (2.14), and bound the remaining denominator by two. This yields the displayed polynomial bound directly, without differentiating a CLT remainder. The odd polynomial now has a \textbf{fixed} coefficient.

The transformed interval is $[-y_-,y_+]$. Because $v\le m/4$, both endpoint magnitudes are at least $2 \sqrt{\log m}$. Furthermore, $|v'|\le1$ implies
\[
|y_+-y_-|\le Cw^2/m^{3/2}=C\log m/\sqrt m.
\tag{2.16}
\]
There is no need to set these endpoints equal. Extend each explicit integral in (2.15) to the full real line. The omitted Gaussian mass is at most $C m^{-2}$; the absolute omitted odd-polynomial contribution is at most $C m^{-5/2}(1+\log m)$. On the full line the odd term integrates to zero. Finally, the integrated absolute $R_m$ term is at most $C/m$, by finite Gaussian moments.

Combining (2.10), (2.11), and (2.15) proves
\[
\boxed{|I_C(U)-1|\le K_{\Lambda,\theta}\frac{\sqrt{\log m}}m.}
\tag{2.17}
\]
In the proof of Theorem A, take $\theta=\kappa/2$.\pend{} All thresholds and constants above depend only on that fixed interiority and on $\Lambda$. There is no dependence on a limiting empirical weight distribution.

\pend{}

\subsection*{3. Exact identities, one-step error, and target-law and execution-law tails}

This section compares one round of the sampler with the reference law $\mathcal R$ of \S3.1 (as in the main text's
Section~\ref{sec:tilt}). At a state with free set $F$, a round of the sampler (\S4.2) puts each coordinate of $F$ into the kept set $C$ independently with probability $f$, by coins drawn before any value, and commits the block $B=F\setminus C$ by independent draws at the exact conditional marginals frozen at the round's start; $C$ is left to the later levels (in the notation of \S4.2, the kept set of round $t$ is $C=F_{t+1}$). $\mathcal R$ makes the same first step, so $(C,x_B)$ has the same law under both, but then completes the kept set
exactly (fair bits if the quota is infeasible). It is used only in the analysis (\S4.2). Below, target-law bounds are under $p$ and execution-law bounds are under $\mathcal R$, or under $\mathcal R^{\rm joint}$ for events that involve $C$; composition needs both (\S4.5). The section's outputs are the one-step bound
(3.18), the second moment (3.21) of $\psi=p/\mathcal R$, and the two-sided ratio bound (3.34) with the exceptional sets (3.35),
which supply the analytic inputs to Lemma~OS of Section~\ref{sec:tilt}; (4.3), (4.12), (4.16) and (4.22)--(4.24) instantiate them level-wise, and the composition (4.15)--(4.20) uses all of them.

\paragraph{3.1 Reference law and exact identities, including the empty child}

Throughout \S3, $\ell=f\sqrt N\ge\ell_0$ and $N\ge N_0$, for thresholds $\ell_0$ and $N_0$ that depend only on $(\Lambda,\kappa)$; individual steps enlarge them finitely often, and \S5 records the final choice. At an interior parent $(F,J)$, put each coordinate of $F$ into the kept set $C$ independently with probability $0<f\le1/4$, so that $C$ has law $\nu(C)=f^{|C|}(1-f)^{N-|C|}$; let $B=F\setminus C$. Draw independent exact parent marginals on $B$. If the residual quota is feasible, complete $C$ with the exact weighted slice; otherwise complete with fair independent bits. This defines a joint law $\mathcal R^{\rm joint}$ of $(C,x_B,x_C)$ and its law $\mathcal R$ of the string $x=(x_B,x_C)$. Probabilities and expectations of quantities that involve $C$, such as the child $(C,j)$, the good event $G$ of \S4.4 and the log ratio $W$ of \S4.5, are taken under $\mathcal R^{\rm joint}$. The arbitrary-on-failure completion has no contribution at any point of the parent slice. At a slice point $x$ the completion of $C$ is the target's own conditional law given $x_B$, so $\mathcal R(x)/p(x)=\sum_C\nu(C)\,\Pi_B(x_B)/p(x_B)$, where $\Pi_B=\prod_{i\in B}\operatorname{Ber}(\pi_i)$ is the product of the exact marginals on the block and $p(x_B)$ is the target's marginal law of the block. The definitions below evaluate this sum. Writing $p(x_B)$ through the logistic product law at tilt $s_0$, which conditioned on $S_F=J$ is $p$, gives
\[
p(x_B)=\widetilde P_{s_0}(x_B)\,\widetilde P_{s_0}(S_C=j)/P_0,\qquad j=|x_C|;
\]
this brings in the quota probability $\widetilde P_{s_0}(S_C=j)$, whose reciprocal is then written as an integral over the tilts $s_0+u$ divided by the normalizer $I_C(j)$ of \S2.4. The result, (3.1)--(3.2), writes $\mathcal R/p$ as an empty-child term plus, up to these normalizers, $P_0\int e^{A(u)+D_x(u)}\Gamma_x(u)\,du$, in which $x$ enters only through the linear statistic $D_x(u)$ and the factor $\Gamma_x(u)$.

Set $t_i(u)=\sigma(\operatorname{logit}(q_i)+u)$, the marginal $p_i(s_0+u)$ of the logistic product law at tilt $s_0+u$, and, for $x=0,1$,
\[
\rho_i(x)=\frac{\operatorname{Ber}(\pi_i)(x)}{\operatorname{Ber}(q_i)(x)},\quad
\phi_i(u,x)=\frac{\operatorname{Ber}(t_i(u))(x)}{\operatorname{Ber}(q_i)(x)},\quad
d_i=(1-f)\rho_i+f\phi_i.
\]
Define $\alpha_i=\log d_i(u,1)-\log d_i(u,0)$,
\[
A(u)=\sum_i\mathbb E_{\operatorname{Ber}(\pi_i)}\log d_i(u,X_i),\quad
D_x(u)=\sum_i(x_i-\pi_i)\alpha_i(u),
\]
and
\[
\Gamma_x(u)=\sum_i f t_i'(u)\frac{\phi_i(u,x_i)}{d_i(u,x_i)},\qquad
\bar\Gamma(u)=\mathbb E_p\Gamma_X(u).
\]
Then $\sum \log d_i=A+D_x$ and \textbf{$\E_p D_X(u)=0$ exactly}, because centering uses the exact parent marginals.

For $x$ on the parent slice and $j=|x_C|$, conditioning the logistic product law gives
\[
\frac{\mathcal R(x)}{p(x)}
=P_0\left[
\nu(\varnothing)\prod_F\rho_i(x_i)
+\sum_{C\ne\varnothing}\nu(C)\prod_B\rho_i(x_i)
\frac{\int K_C''(s_0+u)\prod_C\phi_i(u,x_i)\,du}{I_C(j)}
\right].
\tag{3.1}
\]
Here $K_C$ is the function $K$ of \S2.1 for the coordinates of $C$. To verify the integral step, for any fixed child of nonzero size the likelihood ratio between two common tilts is constant on its quota slice; its value equals the ratio of the two quota probabilities. Integrating that identity against $K_C''$ gives (3.1). The empty-child contribution is the first term, not a quotient of zero integrals. Its total execution probability is $(1-f)^N\le e^{-fN}$.

Expanding $K_C''=\sum_{l \in C}t_l'$, then summing the other inclusion indicators independently, gives
\[
\sum_C\nu(C)K_C''(s_0+u)\prod_B\rho_i\prod_C\phi_i
=e^{A(u)+D_x(u)}\Gamma_x(u).
\tag{3.2}
\]
The empty-child contribution to this second identity is zero.

\paragraph{3.2 Uniform analytic estimates and the bounds on \(A\)}

Put $a_i(u)=(1-f)\pi_i+f t_i(u)$. A mixture of two Bernoulli masses is Bernoulli, so
\[
d_i(u,x)=\frac{\operatorname{Ber}(a_i(u))(x)}{\operatorname{Ber}(q_i)(x)},\quad
\alpha_i(u)=\operatorname{logit}(a_i(u))-\operatorname{logit}(q_i),
\tag{3.3}
\]
and exactly
\[
A(u)=\sum_i\{D(\pi_i\|q_i)-D(\pi_i\|a_i(u))\}.
\tag{3.4}
\]
Here binary KL is abbreviated by its two parameters.

\textbf{Choosing radii without circular constants.} Fix a radius $R_0>0$ using only the common interiority lower bound. The expression
\[
t_i(u)=\frac{q_i e^u}{1+q_i(e^u-1)}
\]
has an absolutely convergent power series on $|u|\le R_0$, with uniform bounds. The proof is complex-analytic. Write $p_*$ for the common interiority lower bound, so that $\pi_i,q_i\in[p_*,1-p_*]$, and extend $u$ to complex $z$. For $|z|\le1/4$ one has $|q_i(e^z-1)|\le1/2$, so $t_i$ is holomorphic there. For $|z|\le p_*/2$ the arguments $a_i$ and $1-a_i$ of the logarithms in (3.3) have real parts above $p_*/2$, so their principal logarithms are holomorphic. For $|z|\le p_*^5/64$ the real part of $\bar\Gamma$ below is at least $Nfp_*^3/2$, so the summands of $\bar\Gamma/f$ and the coefficients $b_i$ below are holomorphic as well. (Indeed, write
$\bar\Gamma/f=\sum_is_i$ with $s_i=t_i'[\pi_it_i/a_i+(1-\pi_i)(1-t_i)/(1-a_i)]$, by (3.10a) below, since $\Gamma_x$ is affine in
each coordinate. At $z=0$, $t_i=q_i$, and Cauchy--Schwarz with $a_i+(1-a_i)=1$ gives
$\pi_iq_i/a_i+(1-\pi_i)(1-q_i)/(1-a_i)\ge(\sqrt{\pi_iq_i}+\sqrt{(1-\pi_i)(1-q_i)})^2\ge4p_*(1-p_*)$, so
$s_i(0)\ge4p_*^2(1-p_*)^2\ge p_*^2$. On $|z|\le p_*/2$, $\operatorname{Re}a_i,\operatorname{Re}(1-a_i)\ge p_*/2$, as
$|t_i-q_i|\le1.6\,q_i|z|$ there; and $|s_i|\le3/p_*$, as there $|t_i(1-t_i)|\le2.6\,q_i(1-q_i)\le0.65$ and the bracket is at
most $3.6/p_*$, from $|t_i|\le1.8q_i$, $|1-t_i|\le1.4(1-q_i)$ and $|a_i|,|1-a_i|\ge p_*/2$. Cauchy's estimate on circles of
radius $p_*/4$ then gives $|s_i'|\le12/p_*^2$ on $|z|\le p_*/4$, so on $|z|\le p_*^5/64$,
$\operatorname{Re}s_i\ge p_*^2-\frac{12}{64}p_*^3\ge p_*^3/2$. The radius has ample room, and its value is inessential: any smaller positive radius depending only on $p_*$
would also do.) Cauchy's estimates on these circles bound the power-series coefficients. The single radius $R_0=p_*^5/64$ serves all three expansions and depends only on $p_*$; division by $f$ keeps the bounds uniform as $f$ tends to zero. First obtain derivative/coefficient constants on this fixed disk; \textbf{only afterward} choose the smaller bulk radius $r_0<R_0/4$ needed for Gaussian absorption.

Let $\delta_i=\pi_i-q_i=O(N^{-1})$. Formula (3.3) gives
\[
\alpha_i(0)=O(N^{-1}),\qquad
\alpha_i'(0)=f+O(f/N),\qquad
\alpha_i(u)=\alpha_i(0)+\alpha_i'(0)u+O(fu^2).
\tag{3.5}
\]
For its power-series coefficients $\alpha_{k,i}$, the fixed-radius argument gives
\[
|\alpha_{0,i}|\le C/N,\quad |\alpha_{1,i}-f|\le Cf/N,\quad
|\alpha_{k,i}|\le Cf R_0^{-k}\ (k\ge2).
\tag{3.6}
\]
For all real $u$, the quantities $a_i(u)$ remain uniformly interior because $1-f\ge3/4$ and $\pi_i$ are uniformly interior. Thus
\[
|\alpha_i(u)|\le C(f+N^{-1})\le Cf,\quad
|\alpha_i'(u)|\le Cf,
\tag{3.7}
\]
where the last comparison uses $f \sqrt{N}\ge\ell_0$ and a sufficiently large threshold. Also $|\alpha_i(u)-\alpha_i(\pm\infty)|\le Cf e^{-|u|+C}$ on the corresponding tails.

For the global upper bound on $A$, (3.4), (2.7), and binary Pinsker give
\[
A(u)\le C/N-c f^2N\min(u^2,1).
\tag{3.8}
\]
Indeed, $|q_i-t_i(u)|\ge c \min(|u|,1)$ uniformly, and $(\pi_i-t_i(u))^2\ge|q_i-t_i(u)|^2/2-\delta_i^2$.

For a local lower bound, discard the nonnegative first terms in (3.4), and Taylor-expand the binary divergence about its second argument $\pi_i$. With $h_i=f(t_i(u)-\pi_i)$,
\[
D(\pi_i\|\pi_i+h_i)=\frac{h_i^2}{2\pi_i(1-\pi_i)}+O(|h_i|^3).
\]
Use $t_i(u)-\pi_i=q_i(1-q_i)u-\delta_i+O(u^2)$. Since all denominators are uniformly interior, summing yields
\[
A(u)\ge-\frac12 f^2V u^2
-C\{f^2N|u|^3+f^2|u|+f^2u^2+f^2/N\}.
\tag{3.9}
\]
\pend{}

\textbf{Remark (the first-order sum is not small).}\pend{}\pend{} Although $\sum_i(\pi_i-q_i)=0$, the first-order sum $S_1(u)=\sum_i \E_{\operatorname{Ber}(\pi_i)}(d_i(u,X_i)-1)$ is not $O(fNu^2/V)$.

\begin{itemize}
  \item Exactly, $S_1(u)=(1-f)\sum_i \chi^2(\pi_i\|q_i)+f \sum_i(\pi_i-q_i)\frac{t_i(u)-q_i}{q_i(1-q_i)}$.
  \item At $u=0$ it equals $(1-f)\sum_i (\pi_i-q_i)^2/[q_i(1-q_i)]$. This is strictly positive in general, while $fNu^2/V$ vanishes at $u=0$. One example: $N$ even, $J=N/2$, one weight $\lambda>1$, the others one. If all exact marginals equaled the logistic ones, the logistic total mean would exceed $N/2$.\pend{}
\end{itemize}

\textbf{What remains true.}\pend{}

\begin{itemize}
  \item The chi-square term is nonnegative and drops out of a lower bound. The remaining part is $\ge-C f N u^2/V$. So $S_1(u)\ge-CfNu^2/V$, and $S_1(u)\le CN/V^2+CfNu^2/V$.\pend{}
  \item The proof of (3.9) does not use $S_1$: it uses the exact identity (3.4) and drops the nonnegative $D(\pi_i\|q_i)$.\pend{}
\end{itemize}

\pend{}

Write $g_i=\phi_i/d_i$. Since $\Gamma_x$ is affine in each binary coordinate,
\[
\Gamma_x/\bar\Gamma=1+Y_x,
\qquad Y_x=\sum_i b_i(u)(x_i-\pi_i),
\]
where
\[
b_i(u)=\frac{f t_i'(u)[g_i(u,1)-g_i(u,0)]}{\bar\Gamma(u)}.
\tag{3.10}
\]
The relevant differences have the explicit formulas
\[
\begin{gathered}
g_i(u,1)=t_i(u)/a_i(u),\qquad
g_i(u,0)=(1-t_i(u))/(1-a_i(u)),\\
g_i(u,1)-g_i(u,0)
=\frac{(1-f)(t_i(u)-\pi_i)}{a_i(u)(1-a_i(u))}.
\end{gathered}
\tag{3.10a}
\]
Thus the nonconstant numerator is $O(|u|+N^{-1})$, including its constant term $O(N^{-1})$; these bounds follow from the displayed separated denominators.

On the fixed analytic neighborhood, $\bar\Gamma/(fN)$ is bounded away from zero: by the radius argument above, its real part is at least $p_*^3/2$ there. With $b_{k,i}$ the power-series coefficients of $b_i(u)$ about $u=0$, the power-series argument above consequently gives
\[
\begin{gathered}
\bar\Gamma(u)=fV(1+O(|u|+N^{-1})),\qquad
|b_i(u)|\le C(|u|+N^{-1})/N,\\
|b_{0,i}|\le C/N^2,\qquad
|b_{k,i}|\le C R_0^{-k}/N\ (k\ge1).
\end{gathered}
\tag{3.11}
\]
For clarity about the denominator: on $|z|\le p_*^5/64$ the real part of $\bar\Gamma/(fN)$ is at least $p_*^3/2$, so its reciprocal is holomorphic there with modulus at most $2/p_*^3$, and Cauchy's estimates apply. No lower bound on $f$ independent of dimension is used. On the bulk, choose $r_0$ and the dimension threshold so that $|Y_x|\le1/2$ for every slice point. Globally,
\[
0\le\Gamma_x(u)\le CfN e^{-|u|+C}.
\tag{3.12}
\]

\paragraph{3.3 Bulk normalization}

Define
\[
N_0^b=P_0\int_{-r_0}^{r_0}e^{A(u)}\bar\Gamma(u)\,du.
\]
Let $\gamma$ be the centred Gaussian law with variance $1/(f^2V)$. On the bulk, (3.9) and (3.11) give
$e^{A}\bar\Gamma\ge fV\,e^{-f^2Vu^2/2-E(u)}(1-C|u|-C/N)$, with $E(u)$ the braces of (3.9). Since
$P_0fV\int_{\mathbb R}e^{-f^2Vu^2/2}du=P_0\sqrt{2\pi V}=1+O(V^{-1})$ by (2.6), this gives
$N_0^b\ge(1-C/V)\,\mathbb E_\gamma\big[(1-E(u))(1-C|u|-C/N);|u|\le r_0\big]\ge(1-C/V)\big(1-\mathbb E_\gamma E-C\mathbb E_\gamma|u|-C/N-\gamma(|u|>r_0)\big)$.
With $\mathbb E_\gamma|u|=\sqrt{2/\pi}/(f\sqrt V)$, $\mathbb E_\gamma|u|^3=2\sqrt{2/\pi}(f^2V)^{-3/2}$, $\mathbb E_\gamma u^2=1/(f^2V)$
and $V\ge cN$, the expectations of $|u|$ and of the four error terms in (3.9) satisfy
\[
\begin{aligned}
\mathbb E_\gamma|u|&\le C/\ell,\qquad
\mathbb E_\gamma[f^2N|u|^3]\le C/\ell,\\
\mathbb E_\gamma[f^2|u|]&\le C/\ell,\qquad
\mathbb E_\gamma[f^2u^2]+f^2/N\le C/N,\\
\gamma(|u|>r_0)&\le2e^{-c\ell^2}.
\end{aligned}
\]
Use $\exp(-e)\ge1-e$ for the nonnegative error bound, and choose the lower factor from (3.11) positive on the bulk. This proves
\[
N_0^b\ge1-C/\ell.
\tag{3.13}
\]
The global upper bound (3.8) and $\bar\Gamma\le CfN$ also give $N_0^b\le C$ and, for the probability density $\rho^b\propto e^A\bar\Gamma$ on the bulk,
\[
\rho^b(u)\le C\ell e^{-c\ell^2u^2}.
\tag{3.14}
\]
Choose $\ell_0$ so that (3.13) is at least one half.

\paragraph{3.4 The tilted subset law and the all-point lower bound}

For fixed slice point $x$ and fixed bulk $u$, normalize the left side of (3.2) as a law on $C$. It can be generated by choosing a distinguished coordinate $l$ with probabilities proportional to $f t_l'(u) g_l(u,x_l)$, forcing it into $C$, and independently retaining every other coordinate with probability
\[
q_i^{\rm ret}(u,x)=\frac{f\phi_i(u,x_i)}{d_i(u,x_i)}.
\tag{3.15}
\]
Since $K_C''=\sum_{l\in C}t_l'$, the summands with a given $l$ factor as $ft_l'\phi_l(u,x_l)\prod_{i\ne l}d_i(u,x_i)$ after summing over the other inclusions. Uniformly in $x$,
\[
q_i^{\rm ret}/f=1+O(r_0+N^{-1}).
\]
Choose $r_0$ and then $N_0$ so that the relative error is sufficiently small. The parent contains at least $\kappa N$ ones and $\kappa N$ zeros. Apply (1.7) separately to the counts of kept ones and kept zeros and to total size. The forced coordinate changes these counts by at most one, which is absorbed since $fN$ is above a fixed threshold. It follows that
\[
\Pr_{\rm tilted}\left(|C|\notin[fN/2,2fN]\ 
\text{or } |x_C|/|C|\notin[\kappa/2,1-\kappa/2]\right)
\le4e^{-cfN}.
\tag{3.16}
\]
One explicit way to select the good concentration events is to require total size at most $5fN/4$, and each of the one and zero counts at least $3\kappa fN/4$, together with total size at least $fN/2$. For a sufficiently small inclusion perturbation their failures have the claimed exponential bounds, and their intersection gives the required density interval.

Fix a parent-slice point $x$. For $|u|\le r_0$, let $\mu_{u,x}$ be the law on $C$ obtained by normalizing the summands of
(3.2): it is the tilted subset law of (3.15), with total mass $e^{A(u)+D_x(u)}\Gamma_x(u)$. By Fubini, the nonempty-child part
of (3.1) equals $P_0\int e^{A+D_x}\Gamma_x\,\mathbb E_{\mu_{u,x}}[1/I_C(|x_C|)]\,du$. On the event of (3.16), the child is
interior with $|C|\ge fN/2$, above the threshold of (2.17), so there
$1/I_C\ge1-K\sqrt{\log|C|}/|C|\ge1-2K\sqrt{\log(fN)}/(fN)$, as $\sqrt{\log m}/m$ decreases for $m\ge2$; that event has
$\mu_{u,x}$-mass at least $1-4e^{-cfN}$. Dropping the empty child, the other children and $|u|>r_0$ (all terms are
nonnegative) gives, for every parent-slice point,
\[
\frac{\mathcal R(x)}{p(x)}
\ge(1-\epsilon_1)P_0\int_{-r_0}^{r_0}e^{A+D_x}\Gamma_x\,du,
\quad
\epsilon_1=C\frac{\sqrt{\log(fN)}}{fN}+4e^{-cfN}.
\tag{3.17}
\]
Here $\epsilon_1\le C/\ell$; all children to which (2.17) is applied have size above its threshold. The empty-child atom has been discarded, not divided by zero.

\paragraph{3.5 One-step KL and the likelihood-ratio second moment}

Apply logarithmic Jensen to (3.17) using $\rho^b$. Exact centering gives $\E_p D_X(u)=0$. Also $\E_pY=0$, $|Y|\le1/2$, and (1.5), (3.11) give $\E_pY^2\le C/N$. Since $\log(1+Y)\ge Y-2Y^2$,
\[
D(p\|\mathcal R)
\le-\log(1-\epsilon_1)-\log N_0^b+C/N
\le C/\ell.
\tag{3.18}
\]
For $\psi(x)=p(x)/\mathcal R(x)$ on the slice and zero elsewhere, the all-point lower bound and $\Gamma_x\ge\bar\Gamma/2$ give
\[
\psi(x)\le C\,\mathbb E_{\rho^b}e^{-D_x(u)}.
\tag{3.19}
\]
The reciprocal inequality used here is $1/\E Z\le\E(1/Z)$ for positive $Z$.

Since $\sum_i(x_i-\pi_i)=0$ on the slice, subtracting a constant vector from the coefficients of $D$ leaves $D_x(u)$ unchanged there; let $\overline\alpha(u)$ be the coordinate mean of the vector $\alpha(u)$. This centering is what makes the bound below small: by (3.5) the first-order part $\alpha_i'(0)u=fu+O(f|u|/N)$ is, to leading order, the same for every coordinate, and the common vector $fu\mathbf1$ alone has squared norm $\ell^2u^2$, which is not small on the bulk. By (3.5),
\[
\|\alpha(u)-\overline\alpha(u)\mathbf1\|_2^2
\le C/N+C\ell^2u^4.
\tag{3.20}
\]
Equation (1.3) gives $\E_p \exp(-D)\le\exp(C/N+C \ell^2 u^4)$. The constants in this fourth-power term were fixed on $R_0$ before choosing $r_0$. Now choose $r_0$ still smaller so that $C u^4\le c u^2/2$ on the bulk, where $c$ is from (3.14). Integrating (3.19) under $p$ and using (3.14) proves
\[
\boxed{\mathbb E_{\mathcal R}\psi^2
=\mathbb E_p \psi\le C_3.}
\tag{3.21}
\]
The constant is dimension-, array-, and $f$-independent in the stated regime. A target-law tail bound alone would not prove this moment estimate; the integration above is essential.

\paragraph{3.6 Execution-law non-interiority and the exact kept-set mean cancellation}

Let $\rho=J/N$. Under actual execution of the first block,
\[
j-\rho|C|
=\sum_{i\in C}(\pi_i-\rho)-\sum_{i\in B}(x_i-\pi_i).
\tag{3.22}
\]
The first sum is a centered independent-selection statistic: its mean is $f \sum(\pi_i-\rho)=0$. The second sum is centered conditional on $C$, because the block draws are independent $\operatorname{Ber}(\pi_i)$. In (3.23), (3.26) and \S3.8 a child $(C,j)$ is called \emph{non-interior} unless $C\ne\varnothing$,
$\bigl||C|-fN\bigr|<fN/2$ and $\kappa/2\le j/|C|\le1-\kappa/2$; this execution-law event includes a size window.
Equation (1.6), applied to these two sums and to the size $|C|-fN$, therefore gives
\[
\mathcal R^{\rm joint}(\text{non-interior child})\le6e^{-c\ell^2}.
\tag{3.23}
\]
This includes infeasible and empty children, and one may take $c=\kappa^2/32$. The size deviation needs no absorption step: (1.6) gives $\Pr(||C|-fN|\ge fN/2)\le2e^{-f^2N/2}=2e^{-\ell^2/2}$.

For the subtler execution-law tails, let $Y_c=\sum_i c_i(x_i-\pi_i)$ and fix $C$ and the generated $x_B$, with interior residual quota $j$. Under the product law tilted so that the child mean is $j$, write its marginals as $t_i^C$. Since a common logistic tilt moves every coordinate in the same direction,
\[
\sum_{i\in C}|t_i^C-q_i|
=\left|j-\sum_{i\in C}q_i\right|.
\tag{3.24}
\]
Let $D_B=\sum_B(x_i-\pi_i)$. Exact centering gives
\[
j-\sum_Cq_i=-D_B+\sum_C(\pi_i-q_i).
\]
Hence the mean of the child part of $Y_c$ under that product law obeys
\[
\left|\sum_Cc_i(t_i^C-\pi_i)\right|
\le\|c\|_\infty\left(|D_B|+2\sum_C|\pi_i-q_i|\right)
\le\|c\|_\infty(|D_B|+C).
\tag{3.25}
\]

This bound is uniform over the kept set, with no omitted fluctuation of its composition. Replacing $\sum_C \pi_i$ by $\rho |C|$ in this calculation would destroy the exact cancellation and require another error term.

Write $Y_B=\sum_{i\in B}c_i(x_i-\pi_i)$ and $Y_C=Y_c-Y_B$. Conditional on $C$, the coordinates of $B$ are independent, so (1.6) gives $\Pr(|Y_B|>z\sqrt N\|c\|_\infty/3)\le2e^{-2z^2/9}$ and $\Pr(|D_B|>z\sqrt N/3)\le2e^{-2z^2/9}$. Off these two events, (3.25) bounds the mean of $Y_C$ under the tilted child product law by $\|c\|_\infty(z\sqrt N/3+C)\le z\sqrt N\|c\|_\infty/2$, above a dimension threshold. Conditional on $C$, $x_B$ and an interior child, $x_C$ has that product law conditioned on $S_C=j$. Under the product law, (1.6) bounds the probability that $Y_C$ deviates from its mean by more than $z\sqrt N\|c\|_\infty/6$ by $2e^{-z^2/18}$, since $|C|\le N$; conditioning divides by the central atom (2.8), which is at least $c/\sqrt{|C|}$, so under the slice this probability is at most $C\sqrt N e^{-z^2/18}$. Adding the three bounds proves, with fixed sufficiently large threshold constants, for $z\ge1$,
\[
\mathcal R(|Y_c|>z\sqrt N\|c\|_\infty)
\le C\sqrt N e^{-cz^2}+\mathcal R^{\rm joint}(\text{non-interior}),
\tag{3.26}
\]
where the non-interior event is counted just once when taking a union over statistics. More explicitly, apply the union bound only on the common interior event and then add its complement once. For $z$ in a bounded interval enlarge the leading constant. This yields the claimed form for every $z\ge1$.

\paragraph{3.7 Finite typical family, exact coefficient centering, and bulk/tail control}

\S\S3.7--3.8 locate where $\psi=p/\mathcal R$ is close to one from both sides. By (3.17) and the interior part of (3.1), this holds wherever $D_x(u)$ and $Y_x(u)$ are small on the bulk, $D_x(u)$ is small compared with $-A(u)$ on the tails, and non-interior children carry little execution mass at $x$. On the bulk these two functions are power series in $u$ whose coefficients are linear statistics of $x$ ((3.6), (3.11)); on the tails $D_x$ is controlled through its values on a grid and at $\pm\infty$, using (3.7). So it suffices that finitely many linear statistics be typical, which (1.4) gives under the target and (3.26) under the reference law; this defines $T_z$ below. The set $T'$ of \S3.8 also removes outputs that receive too much execution mass from non-interior children, which the interior upper bound through (2.17) does not cover, and (3.35) bounds the target and the execution mass of its complement separately.

For a parent state choose a finite family $\mathcal S$ of statistics consisting of the centered $\alpha$ and $b$ coefficient vectors through
\[
K_0=\lceil2\log_2 N\rceil+C_0,
\]
the coefficient vectors defining $D_x(u)$ on a grid of mesh $1/N$ between each bulk endpoint and $\pm(\log N+C)$, and the two limiting coefficient vectors at infinity. Thus $|\mathcal S|\le C N \log N\le C N^2$. All are real linear statistics.

For a Taylor coefficient vector, subtract its \textbf{arithmetic coordinate mean}. This preserves its statistic on the slice, because $\sum(x_i-\pi_i)=0$, and changes its infinity norm by at most a factor of two. In particular, subtract the constant $f$ from the first-order $\alpha$ vector before bounding it. It is not permissible to discard this common term on outputs off the slice; those outputs are covered by the execution failure event in (3.23).

Define $T_z$ as the subset of the parent slice on which every chosen statistic is at most $z \sqrt{N}\|c\|_\infty$ in absolute value. By (1.4) and (3.26),
\[
p(T_z^c)\le2|\mathcal S|e^{-z^2},\qquad
\mathcal R(T_z^c)\le C|\mathcal S|\sqrt N e^{-cz^2}+6e^{-c\ell^2}.
\tag{3.27}
\]
Complements in the second inequality are in the whole output cube; thus off-slice outputs are included.

On $T_z$, equations (3.6) and (3.11), together with their exact centering, give for $|u|\le r_0$
\[
|D_x(u)|\le C z(N^{-1/2}+\ell u^2),\qquad
|Y_x(u)|\le C z(|u|+N^{-1})/\sqrt N.
\tag{3.28}
\]
The omitted $\alpha$ series tail is at most $CNf 2^{-K_0}\le C/N$, and the omitted $b$ tail is at most $C2^{-K_0}$. They are absorbed by the displayed bounds for $z\ge1$. The refined bound on the constant $b$ coefficient in (3.11), not merely its coarse $O(1/N)$ bound, is what gives the $N^{-1}$ inside the second parenthesis.

For real tails, (3.7) gives $|D_x(u_g)|\le Cz \ell$ at grid points. Interpolation adds at most $Cf$, because the mesh is $1/N$; the comparison to each infinite endpoint adds at most $Cf$ beyond the final grid point. Thus
\[
|D_x(u)|\le Cz\ell\quad (u\in\mathbb R).
\tag{3.29}
\]
Choose a fixed $c_2>0$ small enough that $1\le z\le c_2 \ell$ implies both the Gaussian absorption in the bulk using (3.28) and
\[
A(u)+D_x(u)\le-c'\ell^2\quad (|u|\ge r_0).
\tag{3.30}
\]
The latter follows from (3.8) and (3.29); $c_2$ is chosen after $r_0$. Equations (3.12) and (3.30) bound the entire integral outside the bulk by $Ce^{-c'' \ell^2}$ after the prefactor $P_0$ is included. Any polynomial prefactor is absorbed by the condition $\ell^2\ge C\log N$, which (3.32) below imposes.\pend{}

\paragraph{3.8 The non-interior contribution and two-sided density ratios}

Let $h_{\rm bad}(x)$ be the joint execution mass of output $x$ and a non-interior child. Fix $c_{\rm bad}>0$ once so that (3.23) holds with $6 \exp(-c_{\rm bad} \ell^2)$, and define
\[
T'=T_z\cap\{x:h_{\rm bad}(x)\le e^{-c_{\rm bad}\ell^2/2}p(x)\}.
\]
From (3.23) and Markov,
\[
p(T_z\setminus T')\le 6 e^{-c_{\rm bad}\ell^2/2}.
\tag{3.31}
\]
In the following estimates choose a fixed common tail exponent $c$ no larger than $c_{\rm bad}$ and the other previously obtained exponents. This choice weakens bounds but does not redefine the event $T'$.

Assume
\[
1\le z\le c_2\ell,\qquad \ell^2\ge C\log N,\qquad
N^3e^{-cz^2}\le z/\ell.
\tag{3.32}
\]
Let $\eta_z=z/\ell+z/\sqrt{N}$. In the bulk, (3.14) and (3.28) show
\[
\mathbb E_{\rho^b}e^{\pm |D_x|}=1+O(\eta_z)
\]
in the following precise sense: the plus expectation is at most $1+C \eta_z$, and the minus expectation is at least $1-C \eta_z$. For the plus bound, integrate $\ell u^2 \exp(Cz \ell u^2)$ against (3.14), using $Cz \ell\le c \ell^2/2$. The analogous factors involving $Y_x$ have the same or smaller error.

For every $x \in T_z$, (3.17) therefore gives a lower bound $N_0^b(1-C \eta_z)$. For the upper bound, split $\mathcal R(x)=\mathcal R_{\rm int}(x)+h_{\rm bad}(x)$ on the parent slice, where $h_{\rm bad}$
(defined before (3.31)) collects the joint mass of $x$ with every non-interior child in the sense of (3.23): the empty
child, the size-window failures and the density failures. (Off the slice, a feasible child gives $x$ zero mass, since its
completion lands on the slice; so every child that gives $x$ positive mass is infeasible, hence non-interior, and there
$\mathcal R=h_{\rm bad}$.) For an interior child, $|C|>fN/2$ exceeds the threshold
of (2.17) and $j/|C|\in[\kappa/2,1-\kappa/2]$, so $I_C(j)\ge1-K\sqrt{\log|C|}/|C|\ge\frac12$ and
$1/I_C(j)\le1+2K\sqrt{\log|C|}/|C|\le1+4K\sqrt{\log(fN)}/(fN)=:1+\epsilon_2$, with $\epsilon_2\le C/\ell$. Bounding
$1/I_C$ by $1+\epsilon_2$ on interior children and adding back the non-interior nonempty children (nonnegative terms),
(3.2) gives $\mathcal R_{\rm int}(x)/p(x)\le(1+\epsilon_2)P_0\int e^{A+D_x}\Gamma_x\,du$. On the bulk,
$P_0\int_{|u|\le r_0}e^{A}e^{D_x}\bar\Gamma(1+Y_x)du=N_0^b\,\mathbb E_{\rho^b}[e^{D_x}(1+Y_x)]\le N_0^b(1+C\eta_z)$ on $T_z$;
off the bulk the integral is at most $Ce^{-c''\ell^2}$, by (3.12) and (3.30). So the \textbf{interior} contribution is at
most $(1+\epsilon_2)(N_0^b(1+C\eta_z)+Ce^{-c''\ell^2})$; on $T'$ add $h_{\rm bad}/p\le e^{-c_{\rm bad}\ell^2/2}$, and off $T'$
(3.35) bounds both masses. The errors $\epsilon_1$, the Gaussian tail, and the probabilities in (3.27) are absorbed by (3.32). In particular $\epsilon_1\le C/\ell\le Cz/\ell$.

We already have $N_0^b\le C$ from (3.14). Moreover, integrate the all-point lower estimate on $T_z$ and use $\sum_x \mathcal R(x)\le1$. By (3.27),(3.32),
\[
N_0^b\le\frac1{(1-C\eta_z)p(T_z)}\le1+C\eta_z.
\tag{3.33}
\]
Together with (3.13), this places the baseline near one. On $T'$ these give $N_0^b(1-C\eta_z)\le\mathcal R(x)/p(x)\le(1+\epsilon_2)(N_0^b(1+C\eta_z)+Ce^{-c''\ell^2})+e^{-c_{\rm bad}\ell^2/2}$. Since $N_0^b\ge\frac12$ and $\epsilon_2,e^{-c\ell^2}\le C/\ell\le C\eta_z$ (as $z\ge1$),
dividing by $N_0^b$ gives $1-C\eta_z\le\mathcal R/(pN_0^b)\le1+C'\eta_z$. Choose $c_2$ so that $C'\eta_z\le\frac12$; then
$\log(1+a)\le a$ and $\log(1-a)\ge-2a$ on $[0,\frac12]$ give the first bound below. By (3.13) and (3.33),
$|\log N_0^b|\le2C\eta_z$ (using $1/\ell\le\eta_z$), so $|\log \psi|\le C''\eta_z$, and $e^{-x}\ge1-x$ and $e^x\le1+2x$ on
$[0,1]$ give the second, after renaming the constant:
\[
\left|\log\frac{\mathcal R(x)}{p(x)}-\log N_0^b\right|
\le C\eta_z,
\qquad
1-C\eta_z\le \psi(x)\le1+C\eta_z.
\tag{3.34}
\]
Choose $c_2$ smaller if necessary so that the latter lower bound is positive.

Finally, Markov's inequality under $p$ bounds only target masses; for the execution mass, on $T_z\setminus T'$ the already bounded interior density is at most a fixed constant times $p$; the bad component has total mass bounded by (3.23). Thus
\[
\begin{aligned}
p(T'^c)&\le2|\mathcal S|e^{-z^2}+Ce^{-c\ell^2/2},\\
\mathcal R(T'^c)&\le C|\mathcal S|\sqrt N e^{-cz^2}+Ce^{-c\ell^2/2}.
\end{aligned}
\tag{3.35}
\]
These are the distinct target-law and execution-law bounds needed by composition.

\subsection*{4. The kept-set ladder, signed composition, and a legal sampler}

\paragraph{4.1 Integer sizes and their two-sided growth}

The sizes $m_j$ below fix the nominal free counts $M_t$ of the levels (4.4); they depend only on $(n,A)$ and are chosen before any value is drawn. They make the nominal one-step scale $L_t=f_t\sqrt{M_t}$ of the level whose kept set has nominal size $m_d$ at least $A2^{3d/4}/(2\sqrt2)$ (4.5), so it grows geometrically in $d$, faster than the typicality level $z_t$ of (4.22), because $\log M_t\le C(2^d+\log A)$ (4.6). Hence the per-level errors of \S4.5 are summable ((4.23), (4.27), (4.28)), and there are $R\le\log_2\log n+3$ levels with a kept set (4.6), followed by the terminal level $R+1$.

Assume first $A\ge1$ and set
\[
m_0=\lceil4A^2\rceil,\qquad
m_{j+1}=\left\lceil\frac{m_j^2}{A^2 2^{3j/2}}\right\rceil.
\tag{4.1}
\]
$m_0$ is a schedule size; the choice of $A_0$ in \S4.6 makes it exceed every analytic threshold used below.

Put $y_j=\log_2(m_j/A^2)$. Because the pre-ceiling quantity is at least one,
\[
2y_j-\tfrac32j\le y_{j+1}\le2y_j-\tfrac32j+1,
\qquad 2\le y_0\le\log_2 5.
\]
Solving the lower affine recursion and bounding the upper one gives
\[
\tfrac12 2^j+\tfrac32j+\tfrac32\le y_j\le C2^j.
\tag{4.2}
\]
In particular,
\[
m_{j+1}\ge4m_j,\qquad
m_j\ge4A^2 4^j,\qquad
\log m_j\le C(2^j+\log A).
\tag{4.3}
\]
If $n<4m_0$, execute the $n$ singletons in label order and stop this analysis: that schedule is exact and its cap is $n<16A^2+4$.

Otherwise define
\[
R=\max\{r\ge1:4m_{r-1}\le n\},\quad
M_1=n,\quad M_t=m_{R+1-t}\ (2\le t\le R+1),\quad
f_t=M_{t+1}/M_t.
\tag{4.4}
\]
The maximum exists because of the factor-four growth. All $f_t\le1/4$. With $d=R-t$, ceiling bounds give
\[
L_t=f_t\sqrt{M_t}\ge\frac{A2^{3d/4}}{2\sqrt2}\quad(1\le t\le R),
\tag{4.5}
\]
with the stronger denominator $\sqrt{2}$ when $t\ge2$. At the top, use $n<4m_R$ before applying the ceiling bound. Also
\[
\log M_t\le C(2^d+\log A),\qquad
R\le\log_2\log n+3.
\tag{4.6}
\]
For the last bound, $n\ge4m_{R-1}$ and (4.2) imply $\log n\ge(\log 2)2^{R-2}$, and $2-\log_2(\log 2)<3$.

A useful stronger comparison for remainders is, with $m_d=M_{t+1}$,
\[
M_t\le8m_d^2\quad(A\ge1).
\tag{4.7}
\]
It includes the truncated top level and avoids trying to absorb a huge $M_t$ prefactor using only a coarse geometric lower bound on $m_d$.

\paragraph{4.2 A finite law on full labelled schedules, sampled before any bits}

Draw all $nR$ independent Bernoulli variables $\zeta_{t,i}$ with parameters $f_t$ in advance, including variables never subsequently used. Put
\[
F_1=[n],\qquad F_{t+1}=\{i\in F_t:\zeta_{t,i}=1\},\quad
B_t=F_t\setminus F_{t+1}.
\]
Delete empty $B_t$. At the end, append the singletons of $F_{R+1}$ in label order when $|F_{R+1}|\le2m_0$; otherwise append the one block $F_{R+1}$. If the final set is empty, append nothing.

Every realization is an ordered partition of $[n]$. Its randomness is a pushforward of a finite Boolean product space, hence gives a finite schedule distribution. It depends only on $(n,A)$. The terminal size branch is determined by these predrawn coins, not generated values. Empty-block deletion is operationally equivalent to doing nothing at those levels.

Now execute each nonempty block using fresh independent bits at the exact conditional marginals frozen at its start; after an infeasible prefix, execute every remaining scheduled bit fairly. No early stopping, repair, resampling, or value-dependent schedule choice occurs. The reference law's exact completion in \S{}3 is an analysis device, not an extra operation in this sampler. Only the terminal singleton branch samples jointly exactly, by the ordinary sequential chain rule.

For every realization,
\[
\tau=|U|\le R+2m_0\le\log_2\log n+8A^2+5.
\tag{4.8}
\]
In the fallback, the appended block costs one, which is also at most $2m_0$. Every failed run is fully charged. Sampling a schedule needs only the $R+1\le\log_2\log n+4$ sizes from (4.1) and at most $nR$ coins with rational biases $f_t$; the support, of at most $(R+1)^n$ schedules, is never listed. The guarantee needs $A\ge A_0$, and $A_0$ and $C^*$ are existence-only, so the parameter attaining a given $\varepsilon$ is not explicit; the cost of computing exact conditional marginals is the oracle's and is not analysed.

\paragraph{4.3 Suffix kernels and good states: actual sizes versus nominal sizes}

The retention probabilities $f_t$ are fixed by the nominal sizes before any value is drawn, while the actual free count $N$ and quota $J$ at level $t$ are random. The one-step results of \S3 need an interior parent, $f\le1/4$, and $N$ and $\ell_t=f_t\sqrt N$ above fixed thresholds, so the recursion runs over good states (4.10): their actual size is within a factor $1\pm\delta_t$ of nominal, and their density stays at least $\kappa_0/2$ from either endpoint (4.11). The quantity $\mathcal K_t$ of (4.13) is the largest $D(p_{F,J}\|Q_t)$, target first, over good states $(F,J)$ at level $t$; \S\S4.4--4.7 bound it backward from $\mathcal K_{R+1}=0$, and the start state is good.

For an arbitrary residual state at level $t$, define $Q_t$ by the remaining independent retention levels and that same terminal rule, with the original nominal parameters. Two exact facts about these kernels are used. For $n\ge4m_0$, $Q_1$ is the law of the sampler of \S{}4.2; and for every level $t\le R$, $Q_t$ obeys the first-block recursion
\[
Q_t(x\mid h)=\sum_{C\subseteq F}\nu_t(C)\sum_{S\subseteq F\setminus C}P_h(F\setminus C,S)\,Q_{t+1}\bigl(x\mid(F\setminus C,S),h\bigr),
\]
where $F$ is the residual set of the state $h$, $\nu_t$ retains each coordinate of $F$ independently with probability $f_t$, and $P_h(B,S)$ is the probability that executing the block $B$ at the conditional marginals frozen at $h$ returns the ones $S$ (fair bits after an infeasible prefix). The recursion reflects that past bit draws depend only on past blocks and values, not on unrevealed future coins; no independence after conditioning on the entire predrawn schedule is claimed.

Set
\[
\delta_t=\sum_{s<t}M_{s+1}^{-1/3},\qquad
\Delta_t=\frac{\kappa_0}{16}2^{-(R-t)/4},\qquad
\kappa^{(t)}=\kappa_0-\sum_{s<t}\Delta_s.
\tag{4.9}
\]
A state is good at level $t$ when
\[
N/M_t\in[1-\delta_t,1+\delta_t],\qquad
J/N\in[\kappa^{(t)},1-\kappa^{(t)}].
\tag{4.10}
\]
The geometric bounds in (4.3) imply
\[
\delta_t\le C A^{-2/3}\le1/4
\]
after fixing $A_0$. Also
\[
\sum_{s=1}^R\Delta_s
\le\frac{\kappa_0}{16(1-2^{-1/4})}<\kappa_0/2.
\tag{4.11}
\]
Thus every good state has density at least $\kappa_0/2$ away from either endpoint. At a good state,
\[
\sqrt{3/4}\,L_t\le\ell_t=f_t\sqrt N\le\sqrt{5/4}\,L_t.
\tag{4.12}
\]
All one-step results may therefore be applied with a fixed interiority lower bound and these explicit scale comparisons. At the last level a good state has $N\le5m_0/4<2m_0$, so the suffix sampler is exact.

Define
\[
\mathcal K_t=\sup_{\text{good states at }t}D(p_{F,J}\|Q_t),\qquad
\mathcal K_{R+1}=0.
\tag{4.13}
\]
This supremum is finite before any recursion is solved. Along a prescribed feasible target output, every non-deterministic bit in every legal suffix schedule has its matching value with probability at least $1/(\Lambda N)$, by (1.1); deterministic boundary bits match with probability one. Thus $Q_t(x)\ge(\Lambda N)^{-N}$ on the target slice, and the KL is at most $N \log(\Lambda N)$. If a class of good states is empty, take its nonnegative supremum as zero; none of the argument relies on existence of an unattainable state.

\paragraph{4.4 Conditional probability of leaving the good set}

At a good parent, $|C|$ is binomial with mean $f_t N$, lying in $[(1-\delta_t)M_{t+1},(1+\delta_t)M_{t+1}]$. A size failure at the next level entails a deviation at least $M_{t+1}^{2/3}$ from this mean. Apply (1.7); for large enough $A_0$ the relative deviation is at most one and the exponent is at least $M_{t+1}^{1/3}/6$.

On the size-good event, $|C|\ge c f_t N$. Apply the two centered bounds from (3.22), now at threshold a constant times $\Delta_t f_t N$. The independent-selection statistic and, conditional on $C$, the independent draw statistic each satisfy (1.6). Since exiting the density interval requires a drift greater than $\Delta_t$,
\[
\mathcal R^{\rm joint}(G^c)
\le 2e^{-M_{t+1}^{1/3}/6}
+4e^{-c\Delta_t^2\ell_t^2},
\tag{4.14}
\]
where $G$ is the event that the child state $(C,j)$, with $j=J-|x_B|$, is good at level $t+1$.\pend{} By (4.10) this requires both the \emph{size condition} $|C|/M_{t+1} \in [1-\delta_{t+1},1+\delta_{t+1}]$ and the \emph{density condition} $j/|C| \in [\kappa^{(t+1)},1-\kappa^{(t+1)}]$. So $G^c$ is the union of a size failure, bounded by the first term of (4.14), and a density failure on the size-good event, bounded by the second. An infeasible residual quota fails the density condition. This bound concerns the \textbf{pre-completion state $(C,x_B)$}. It consequently holds for $Q_t$ as well as the reference law, because their first-block operations agree.

\paragraph{4.5 Jensen composition with signed continuation log ratios}

This subsection turns the one-step comparison of \S3 into the recursion (4.20) for the good-state suprema $\mathcal K_t$ of (4.13), which \S4.7 solves. Because $Q_t$ and $\mathcal R$ make the same first step (\S4.3), Jensen's inequality (4.15) bounds $D(p\|Q_t)$ by $D(p\|\mathcal R)$, which (3.18) controls, plus the mean of $\psi W$ under the joint reference law of $(C,x_B,x_C)$ ($\mathcal R^{\rm joint}$ of \S3.1), where $W$ compares the exact completion of the kept set with the sampler's continuation. Given $(C,x_B)$ at a good child, $W$ has conditional mean equal to that child's divergence, at most $\mathcal K_{t+1}$; but $W$ can be negative pointwise and $\psi$ varies with $x_C$, so an upper bound $\psi\le1+\eta_t$ alone bounds $\psi W$ only by $(1+\eta_t)W^+$ (here $1-\eta_t\le\psi\le1+\eta_t$ on $T'$, by (3.34) and (4.16)), whose mean can exceed $\mathcal K_{t+1}$ by $\sqrt{\mathcal K_{t+1}/2}$ (1.10); (4.17) therefore uses both sides of (3.34) on $T'$, so that this excess enters (4.20) only with the small factor $2\eta_t$. Off $T'$ and off the good event $G$ of (4.14), the remaining terms, $\sqrt{\mathcal R(T'^c)}$ in (4.18) and the two terms of (4.19), are controlled by the separate execution-law and target-law masses (3.35) of $T'^c$, the bound (4.14) on $\mathcal R^{\rm joint}(G^c)$, the second moment (3.21) and the atom bound of \S4.3.

For a feasible child define
\[
W=\log\frac{p_{C,j}(x_C)}{Q_{t+1}(x_C\mid C,j)}.
\]
For an infeasible child, both conditional completions are fair; set $W=0$ there. This makes the joint likelihood identity unambiguous, though the off-slice terms have zero weight under the parent target in any event.

Let $\pi_{\mathcal R^{\rm joint}}(C\mid x)$ denote the reference law's posterior over kept sets, and let $\psi=p/\mathcal R$. At each parent-slice point,
\[
\frac{Q_t(x)}{\mathcal R(x)}
=\mathbb E_{\pi_{\mathcal R^{\rm joint}}(\cdot\mid x)}e^{-W}.
\]
Logarithmic Jensen, integrated under $p$, gives
\[
D(p\|Q_t)\le D(p\|\mathcal R)+
\mathbb E_{\mathcal R^{\rm joint}}[\psi W].
\tag{4.15}
\]
Also $\E_{\mathcal R} \psi=\sum_xp(x)=1$, since $\psi=0$ off the slice and, on it, $\mathcal R(x)\ge f^Np(x)>0$ (the term $C=F$ of \S3.1). On a feasible child, conditional on $(C,x_B)$, the reference law's completion is exactly $p_{C,j}$, so (1.9) applies to $W$.

Choose $T'$ from \S{}3.8 with $z=z_t$, the typicality level fixed in (4.22) below, and let
\[
\eta_t=C z_t/L_t<1/2,
\tag{4.16}
\]
enlarging the fixed coefficient to cover (4.12) and $z/\sqrt{N}\le z/(4\ell)$. On $T'$, $1-\eta_t\le\psi\le1+\eta_t$, hence pointwise for either sign of $W$,
\[
\psi W\le(1-\eta_t)W+2\eta_tW^+.
\tag{4.17}
\]
Now retain the event $G$ until taking the appropriate conditional expectations. Equations (1.9),(1.10),(4.13) give
\[
\begin{aligned}
\mathbb E_{\mathcal R^{\rm joint}}[W\mathbf1_{G\cap T'}]
&\le\mathcal K_{t+1}+\sqrt{\mathcal R(T'^c)},\\
\mathbb E_{\mathcal R^{\rm joint}}[W^+\mathbf1_G]
&\le\mathcal K_{t+1}+\sqrt{\mathcal K_{t+1}/2}.
\end{aligned}
\tag{4.18}
\]
For the first inequality, subtracting $G \cap T'^c$ can increase the expectation only by its negative part; apply Cauchy--Schwarz and $\E_{\mathcal R^{\rm joint}}[(W^-)^2\mathbf1_G]\le1$. This is the signed step that would be lost by simply assuming $W\ge0$.

Let $B_N=N \log(\Lambda N)$. The feasible-output atom bound in \S{}4.3 gives $W^+\le B_N$. On the discarded events, upper-bound by nonnegative quantities, so overlapping events cause no sign error:
\[
\begin{aligned}
\mathbb E_{\mathcal R^{\rm joint}}[\psi W\mathbf1_{G\cap T'^c}]&\le B_N p(T'^c),\\
\mathbb E_{\mathcal R^{\rm joint}}[\psi W\mathbf1_{G^c}]&\le B_N\sqrt{C_3\mathcal R^{\rm joint}(G^c)}.
\end{aligned}
\tag{4.19}
\]
Combining (3.18), (4.15)--(4.19), and taking the supremum over good parents yields
\[
\mathcal K_t\le C/L_t+(1+\eta_t)\mathcal K_{t+1}
+2\eta_t\sqrt{\mathcal K_{t+1}/2}+\xi_t,
\tag{4.20}
\]
where a valid nonnegative remainder is
\[
\xi_t=\sqrt{\mathcal R(T'^c)}+
B_N\{p(T'^c)+\sqrt{C_3\mathcal R^{\rm joint}(G^c)}\},
\tag{4.21}
\]
with the right sides replaced by their uniform good-state bounds. Because $1-\eta_t\le1$, no omitted multiplicative constant is required in front of its first term. Sub-probability continuation laws are handled by the cemetery extension of \S{}1.4, not by changing $W$ to a conditional-on-survival ratio.

\paragraph{4.6 One simultaneous choice of parameters and uniform remainder bounds}

Choose, with a sufficiently large fixed coefficient $D$,\pend{}
\[
z_t^2=D(\log M_t+\log A+d),\qquad d=R-t.
\tag{4.22}
\]
The coefficient $D$ depends only on the previously fixed one-step tail constants. Since $M_t\ge m_d\ge4A^24^d$ by (4.3) (for $t=1$, $M_1=n\ge4m_{R-1}$), $\log A+d\le\log M_t$, so $z_t\le\sqrt{2D\log M_t}$ and $\eta_t=Cz_t/L_t=O(\sqrt{\log M_t}/L_t)$, the form quoted in the main text's Section~\ref{sec:tilt}. Equations (4.5),(4.6),(4.12) imply, uniformly over good states,
\[
\frac{z_t}{\ell_t}
\le\frac C A\left(2^{-d/4}+\sqrt{\log A}\,2^{-3d/4}\right).
\tag{4.23}
\]
Its supremum tends to zero as $A$ tends to infinity. Also
\[
\frac{\ell_t^2}{\log N}\ge c\frac{A^2}{1+\log A}\longrightarrow\infty
\]
uniformly in $d$. Choose $D$ sufficiently large that the Gaussian-family tails in (3.35), including their at-most-$CN^{5/2}$ prefactor, are bounded, for example, by
\[
C(1+N)^{-10}A^{-12}2^{-12d}.
\tag{4.24}
\]
For instance, let $c$ be the common tail exponent of (3.35) and take $cD\ge13$. By (4.22), $e^{-cz_t^2}=(M_tAe^d)^{-cD}$, and on a good state $N$ is comparable with $M_t$, so
\[
N^{5/2}e^{-cz_t^2}\le C\,N^{5/2-cD}A^{-cD}e^{-cDd}\le C(1+N)^{-10}A^{-12}2^{-12d},
\]
because $cD-5/2\ge21/2$, $A\ge1$ and $e^{-cD}\le e^{-13}<2^{-12}$; the comparability constants, raised to the power $cD$, are absorbed into $C$. The same bound gives the last condition in (3.32): since $\ell_t\le\sqrt{N}/4$ and $z_t\ge1$, it suffices that $N^3e^{-cz_t^2}\le4/\sqrt N$, and $N^3e^{-cz_t^2}\le C\,N^{3-cD}A^{-cD}\le CN^{-10}A^{-13}$, which is at most $4/\sqrt N$ once $A_0$ is large.

It remains to account for non-Gaussian tails multiplied by $B_N$; their prefactors cannot just be ignored.

\textbf{Size deviations.} Let $m=M_{t+1}=m_d$. By (4.7) and good-state comparability, $N\le10m^2$. Therefore
\[
B_N e^{-c m^{1/3}}
\le C m^2(1+\log m)e^{-c m^{1/3}}
\le C e^{-c' m^{1/3}}
\tag{4.25}
\]
above one fixed threshold. The same absorption works with the halved exponent from the square root in (4.21). Since $m_d\ge4A^2 4^d$, this is bounded by $C \exp(-c A^{2/3}2^{2d/3})$. Increasing $A_0$ gives the bound $CA^{-2}2^{-d}$ that (4.27) uses. One direct verification uses $2^{2d/3}\ge1+(2 \log 2/3)d$ to separate the $A$ and $d$ decay.

\textbf{Density deviations.} Equations (4.5),(4.9),(4.12) give
\[
\Delta_t^2\ell_t^2\ge c A^2 2^d.
\tag{4.26}
\]
On the other hand, (4.6) bounds $B_N$ by $\exp(C(2^d+\log A))$. Choose $A_0$ so that the negative $A^2 2^d$ coefficient dominates this positive coefficient, even after the square root in (4.21). The result is again at most $CA^{-2}2^{-d}$. The exceptional-set terms $\exp(-c \ell_t^2)$ have still faster decay and are handled in exactly this comparison, replacing $2^d$ by $2^{3d/2}$ in the negative exponent.

Combining (4.24)--(4.26), taking square roots where (4.21) requires them, proves
\[
\boxed{\xi_t\le C A^{-2}2^{-d}.}
\tag{4.27}
\]
All constants were chosen before $n$, the realized array, or the level was specified.

\paragraph{4.7 Summation and a fully constant-tracked scalar recursion}

By (4.5),(4.23),(4.27),
\[
\sum_t L_t^{-1}\le C/A,\quad
\sum_t\eta_t\le C\sqrt{\log A}/A,\quad
\sum_t\xi_t\le C/A^2.
\tag{4.28}
\]
To keep all fixed coefficients explicit, define\pend{}
\[
E=\sum_t(C/L_t+\xi_t),\quad
B=\sqrt2\sum_t\eta_t,\quad
P=\prod_t(1+\eta_t),\quad
\mathcal K_*=\max_t\mathcal K_t.
\]
Increase $A_0$ so that $P\le\exp(\sum \eta_t)\le2$. Unroll (4.20), using $\mathcal K_{R+1}=0$ and $\sqrt{\mathcal K_{t+1}}\le\sqrt{\mathcal K_*}$. Then
\[
\mathcal K_*\le P(E+B\sqrt{\mathcal K_*})
\le2E+2B\sqrt{\mathcal K_*}.
\]
Young's inequality $2Bx\le x^2/2+2B^2$ yields
\[
\mathcal K_*\le4E+4B^2
\le C/A+C\log A/A^2\le C^*/A.
\tag{4.29}
\]
The start state is good, so
\[
\boxed{D(p_w\|Q_{U(A)})\le\mathcal K_1\le C^*/A.}
\tag{4.30}
\]
This establishes item~3 of Theorem~A, with this $C^*$, uniformly for all $A\ge A_0$, all admissible dimensions, quotas, and bounded-ratio arrays. The exact singleton branch already covers all small dimensions.

\paragraph{4.8 Survival and within-slice shape}

If a run never exits the good-state sets, its final suffix is exact, so its output lies on the target slice. The event of any failure is therefore contained in the union of the first exits. The conditional first-exit bounds are (4.14), under the actual sampler by the pre-completion-state equality.

\pend{} In (4.14) the scale $\ell_t=f_t \sqrt{N}$ belongs to the actual parent state. Every good state at level $t$ has $\ell_t\ge\sqrt{3/4} L_t$, by (4.12). So each first-exit probability at a good parent is at most $2e^{-M_{t+1}^{1/3}/6}+4e^{-(3c/4)\Delta_t^2L_t^2}$, a bound that is uniform over good parents and depends on nothing but the level. This is the supremum over good states. Hence
\[
1-Q(S_k)
\le\sum_{t=1}^R\left(2e^{-M_{t+1}^{1/3}/6}+4e^{-(3c/4)\Delta_t^2L_t^2}\right)
\le C e^{-c A^{2/3}}.
\tag{4.31}
\]
The sum is uniformly bounded using (4.3),(4.26); the smallest size is $m_0=\lceil4A^2\rceil$. Increase $A_0$ so that the right side is at most one half. Then $-\log(1-u)\le2u$ gives
\[
\boxed{-\log Q(S_k)\le C e^{-c A^{2/3}}.}
\tag{4.32}
\]
For a target supported on the slice the exact decomposition is
\[
D(p\|Q)=D(p\|Q(\cdot\mid S_k))-\log Q(S_k).
\tag{4.33}
\]
Thus the shape term is $F+\log Q(S_k)\le F$. (4.30) bounds the shape term as well as survival.

\paragraph{4.9 Fixed error, small \(n\), and the diagonal scheduled-mean coefficient}

For any fixed $\varepsilon>0$, set $A=\max(A_0,C^*/\varepsilon)$. Equations (4.8),(4.30) and the exact branch give
\[
F\le\varepsilon,\qquad
\operatorname{cap}\le\log_2\log n+O_{\Lambda,\kappa_0}(1+\varepsilon^{-2}).
\tag{4.34}
\]
Every admissible dimension has $n\ge2$: the quota condition at $n=1$ admits no integer quota. Consequently the lower bound $\log_2 \log n\ge\log_2 \log 2$ lets the exact small-$n$ branch be included in the same additive-constant expression. No expression $\log \log 1$ is used for an actual admissible instance.

For the diagonal claim, take, for sufficiently large $n$,
\[
A_n=\max(A_0,(\log\log n)^{1/4}),\qquad
\varepsilon_n=C^*/A_n\to0.
\]
The all-$A$ uniform statement (4.30) applies immediately. Eventually $n\ge4\lceil4A_n^2\rceil$ and $\varepsilon_n<\varepsilon$. Put
\[
\alpha_n=1-\beta=e^{-\varepsilon+\varepsilon_n}\in(0,1),\qquad
\mu_n'=\alpha_n U(A_n)+(1-\alpha_n)\{\text{one full block}\}.
\]
Since its output law is at least $\alpha_n Q_{U(A_n)}$ pointwise,
\[
F(\mu_n')\le-\log\alpha_n+\varepsilon_n=\varepsilon.
\tag{4.35}
\]
Every component's full schedule is charged, so the mean is affine and
\[
C(\mu_n')\le\alpha_n(\log_2\log n+8A_n^2+5)+(1-\alpha_n)
=(e^{-\varepsilon}+o(1))\log_2\log n.
\tag{4.36}
\]
The hard cap is $\log_2 \log n+O(\sqrt{\log \log n})$.

For precision, the upper bound (4.36) alone proves the coefficient as an achievable upper coefficient. The construction also gives the corresponding asymptotic equality for its own mean, without importing a lower-bound theorem. Along this diagonal, (4.2),(4.4) imply $R=\log_2 \log n+O(1)$, since $\log A_n=o(\log n)$. (The upper direction is (4.6). For the lower one,
$R$ is maximal, so $4m_R>n$; from (4.1), $y_{j+1}\le2y_j+1$ and $y_0\le\log_25$, so $y_R\le(1+\log_25)2^R$. Hence
$\log n<\log4+2\log A+(\log10)\,2^R$, and $2^R>(\log n-\log4-2\log A)/\log10$, which gives $R\ge\log_2\log n-O(1)$
once $\log A=o(\log n)$. When $n\ge4A^4$, $2\log A\le(\log n-\log4)/2$, so $R>\log_2(\log n-\log4)-\log_2(2\log10)$,
which exceeds $\log_2\log n-3$ for $n>26$; for smaller $n$, $R\ge1>\log_2\log n-3$ directly.) Each coordinate lies in the large block $B_t$ with probability $(M_t-M_{t+1})/n$, because the retention products telescope, so $\E|B_t|=M_t-M_{t+1}$ and $\Pr(B_t=\varnothing)=(1-(M_t-M_{t+1})/n)^n$. Therefore
\[
\Pr(B_t=\varnothing)\le e^{-(M_t-M_{t+1})}\le e^{-3M_{t+1}},
\]
and the sum of these empty-block probabilities is bounded by $Ce^{-cA_n^2}$ using (4.3). The expected number of nonempty large blocks is $R-O(e^{-cA_n^2})$, while the terminal contribution is between zero and $2m_0=O(A_n^2)$. Thus
\[
\mathbb E\tau_{U(A_n)}=\log_2\log n+o(\log\log n),
\quad
\mathbb E\tau_{\mu_n'}=(e^{-\varepsilon}+o(1))\log_2\log n.
\tag{4.37}
\]
This equality concerns the constructed family. The matching lower bound is Appendix~\ref{app:t1}, \S10(c): under the hypotheses of \S0 and for $n\ge\kappa_0^{-2}$, every causal policy with fresh exact-marginal draws and output KL at most $\varepsilon$ has $\E\tau\ge e^{-\varepsilon}\log_2\log n-e^{-\varepsilon}-c(K_\Lambda)$, and a predrawn mixture executes $\tau=|U|$ rounds. So $e^{-\varepsilon}$ is the least leading coefficient both for the scheduled mean of predrawn mixtures and for the expected executed rounds of causal policies.\pend{}\pend{}\pend{} The cap (4.8) is a uniform bound on component lengths, not a common length or a fixed block profile.\pend{}

\subsection*{5. Constants and a noncircular choice order}

\begin{center}
\small
\begin{tabular}{@{}p{\dimexpr(\linewidth-6\tabcolsep)*24/100\relax}p{\dimexpr(\linewidth-6\tabcolsep)*22/100\relax}p{\dimexpr(\linewidth-6\tabcolsep)*22/100\relax}p{\dimexpr(\linewidth-6\tabcolsep)*32/100\relax}@{}}
\toprule
\raggedright Object & \raggedright Depends on & \raggedright Must not depend on & \raggedright How fixed \tabularnewline
\midrule
\raggedright Logistic interiority lower bound & \raggedright $\Lambda,\kappa_0$ & \raggedright $n,w,A,t$ & \raggedright Use parent density $\ge\kappa_0/2$ and child density half that. \tabularnewline
\raggedright $K_a$ in (2.2) & \raggedright the probability lower bound $a$ & \raggedright dimension, lattice point, Bernoulli array & \raggedright Fourier estimates (2.3)--(2.5). \tabularnewline
\raggedright $m_{\rm an},N_{\rm an}$ & \raggedright $\Lambda,\kappa_0$ & \raggedright dimension, array, level & \raggedright Window interiority, central atoms, marginal approximation. \tabularnewline
\raggedright Fixed power-series radius $R_0$ and coefficient constants & \raggedright $\Lambda,\kappa_0$ & \raggedright $f,N,\text{array}$ & \raggedright Cauchy estimates on fixed complex discs where the logarithm arguments and the denominators have real parts bounded below. \tabularnewline
\raggedright Bulk radius $r_0$ & \raggedright previously fixed analytic and Gaussian constants & \raggedright $A,n,t$ & \raggedright Choose after derivative constants, so the fourth-power moment term is absorbed. \tabularnewline
\raggedright $\ell_0,N_0,C_3$ & \raggedright preceding constants & \raggedright all instance parameters & \raggedright Normalization lower bound, interior-child threshold, inverse moment. \tabularnewline
\raggedright $c_2$ and a common smaller tail exponent $c$ & \raggedright preceding constants & \raggedright $A,n,t$ & \raggedright Bulk quadratic absorption, real tails, positive ratio lower endpoint. \tabularnewline
\raggedright Typicality coefficient $D$ & \raggedright common tail exponents and polynomial prefactors & \raggedright $A,n,t$ & \raggedright Make (4.24) stronger than all needed square-root and atom-cost bounds. \tabularnewline
\raggedright $A_0$ & \raggedright all preceding constants, $\Lambda,\kappa_0$ & \raggedright dimension, realized array, level & \raggedright Simultaneous conditions listed below. \tabularnewline
\raggedright $C^*$ & \raggedright fixed recursion constants & \raggedright $A,n,w,k$ & \raggedright (4.28)--(4.30). \tabularnewline
\bottomrule
\end{tabular}
\end{center}

A single sufficiently large finite $A_0$ satisfies all of the following: $A_0\ge1$; smallest good parent and interior kept-child sizes exceed the analytic thresholds; smallest actual $\ell$ exceeds $\ell_0$; $\delta_t\le1/4$; the uniform ratios in (4.23) are below $c_2$ and give $\eta_t<1/2$; (3.32) holds; the polynomial/exponential absorptions in (4.25),(4.26) hold; $\sum \eta_t\le\log 2$; and the survival bound is at most one half. Each requirement either holds above a fixed threshold or follows from a displayed scalar function tending to zero/infinity as $A$ increases. The bounds were uniform in $d\ge0$; taking their finite maximum is therefore independent of $n$.

The constants are existence-only: no numerical value of $A_0$ or finite-size constant is derived.

\subsection*{6. Dependency graph}\pend{}

\begin{itemize}
\item weighted-slice algebra + elementary coefficient log-concavity
  \begin{itemize}
  \item marginal floors and feasible-output atoms
  \item conditioning closure + (1.2) \(\longrightarrow\) Lemma NCD (1.3a) \(\longrightarrow\) Lemma T1\pend{}
  \item covariance Laplacian at every external tilt \(\longrightarrow\) Lemma T1 (route MG)
  \end{itemize}
\item logistic cumulants + finite Fourier inversion + Hoeffding
  \begin{itemize}
  \item uniform lattice estimate (2.2)
    \begin{itemize}
    \item central atom + exact-marginal approximation
    \item explicit mean/Jacobian expansion \(\longrightarrow\) \(I_C(U)=1+O(\sqrt{\log m}/m)\)
    \end{itemize}
  \end{itemize}
\item exact identities (3.1)--(3.2) + uniform analytic estimates\pend{}
  \begin{itemize}
  \item size-biased tilted subset concentration
  \item bulk normalization + exact target centering \(\longrightarrow\) one-step KL
  \item Lemma T1 + Gaussian domination \(\longrightarrow\) inverse-ratio moment
  \item kept-set cancellation + central child atom
    \begin{itemize}
    \item target/execution typicality + exceptional-component control
      \begin{itemize}
      \item two-sided ratio interval
      \end{itemize}
    \end{itemize}
  \end{itemize}
\item predrawn integer ladder + actual/nominal comparisons + drift bounds
  \begin{itemize}
  \item signed Jensen composition + sub-probability KL inequalities
    \begin{itemize}
    \item simultaneous constants + summable recursion \(\longrightarrow\) Theorem A
    \end{itemize}
  \end{itemize}
\item same predrawn construction \(\longrightarrow\) cap + fully charged cost
\item first-exit argument \(\longrightarrow\) survival; exact KL decomposition \(\longrightarrow\) shape
\item uniform all-\(A\) theorem + pointwise mixture bound \(\longrightarrow\) A\('\) and A\(''\)
\end{itemize}

\pend{}

\pend{}\pend{}

\pend{}
\pend{}

\pend{}

\pend{}

\section{Two-group reductions behind the finite exhibits}\label{app:tgred}
\pend{}

This appendix proves the reductions behind the two-group finite computations reported in Appendix~\ref{app:exhibits} and summarized in Appendix~\ref{app:alt}. Lemma~TGR1 shows that a deterministic policy's output KL equals its path cost and computes the least such cost by a count-state recursion; Lemmas~TGR2 and~TGR3 reduce the output KL of any causal randomized policy to a convex function of its terminal count vector, minimized over the convex hull of the vectors of count policies; Lemma~TGR4 gives dual lower certificates; TGR5 links that hull to the survival optimum of Appendix~\ref{app:tgenv}; Lemma~TGR6 treats predrawn schedules; and Lemma~TGR7 brackets the optimal law's leakage--shape split. After the setting and some additional definitions, a list maps each reported quantity to the lemmas behind it; the statements TGR0--TGR7 follow, then their proofs and a description of the computations.

\emph{Setting.} We use the Setting, Interface and Policies of Appendix~\ref{app:tgenv} unchanged: the target
$p_\lambda$, with $0<k<n$ and $\lambda>0$; prefixes, count states $s=(m_A,m_B,j)$, $M=m_A+m_B$, and legality; the
exact-marginal interface, with empty batches allowed and arbitrary values at an illegal prefix; causal randomized
policies, their seed $\omega$ and the $\sigma$-fields $\mathcal F_t$; the every-path hard cap $r$ and the class
$\mathrm{RAND}_r$; full labelled schedules $U$ and predrawn mixtures $\mu$; the output laws $Q_\pi$, $q_U$, $Q_\mu$; the
count kernel; the recursion $G^\lambda_r$; and $\mathrm{root}=(n_A,n_B,k)$.

\emph{Additional definitions.} Throughout, $r\ge1$: at $r=0$ no policy commits the $n\ge1$ coordinates, so the classes
below are empty.
\begin{itemize}
\item $\mathrm{DET}_r\subset\mathrm{RAND}_r$: the members of $\mathrm{RAND}_r$ that draw no seed, so that $\beta_{t+1}$ is
a function of the values committed through round $t$ alone (deterministic value-adaptive policies).
\item $\mathrm{NA}_r\subset\mathrm{RAND}_r$: predrawn mixtures $\mu=\sum_iw_i\delta_{U_i}$ of full labelled schedules with at
most $r$ blocks. A schedule with at most $r$ nonempty blocks commits every coordinate within $r$ rounds.
\item \emph{No latent peeking.} The batch depends on $\omega$ and the committed values only. A controller that could see an
uncommitted coordinate's auxiliary variable (such as the uniform $\zeta_i$ of the interface in Appendix~\ref{app:tgenv}) is outside the class, and the count reduction fails for it. Take
$n_A=n_B=k=1$ and $\lambda=1$, so that each exact marginal is $\frac12$, and suppose the controller commits A when
$\zeta_A\le\frac12$ (A's value would be $1$) and B otherwise. Then the committed value is $1$ with probability
$\frac12+\frac12\cdot\frac12=\frac34$, not $\frac12$.
\item \emph{Weight classes.} $\mathcal W=\{w:\max(0,k-n_B)\le w\le\min(k,n_A)\}$ and $N_w=\binom{n_A}{w}\binom{n_B}{k-w}$.
Then $P_w=p_\lambda(|x_A|=w)=N_w\lambda^w/\sum_{v\in\mathcal W}N_v\lambda^v$, and every $P_w>0$.
\item \emph{Terminal vector.} For a sub-probability $Q$ on $\{0,1\}^n$, let $q_w(Q)=Q(x\in S_k,\ |x_A|=w)$ and
$q(Q)=(q_w)_{w\in\mathcal W}$, so that $s(q)=\sum_wq_w=Q(S_k)$. Write $q(\pi)=q(Q_\pi)$, $q(U)=q(q_U)$ and
$q(\mu)=q(Q_\mu)$. With an argument, $q(\cdot)$ is a terminal vector; $q_U$ is still the output law of $U$.
\item $\Phi(q)=\sum_{w\in\mathcal W}P_w\ln(P_w/q_w)\in(-\infty,+\infty]$ for $q\in[0,\infty)^{\mathcal W}$, which is
$+\infty$ if some $q_w=0$. Since $\Phi(q)=-\ln s(q)+\KL(P\|q/s(q))$ when $s(q)>0$, and $\Phi(q)=+\infty=-\ln s(q)$ when $s(q)=0$, $\Phi(q)\ge-\ln s(q)\ge0$ whenever $s(q)\le1$, in
particular on sub-probability vectors; TGR4(a) also evaluates $\Phi$ at arbitrary positive vectors, where it can be negative.
\item \emph{The three frontier values.} They are
\begin{gather*}
F^*_r=\inf_{\pi\in\mathrm{RAND}_r}\KL(p_\lambda\|Q_\pi),\qquad D_r=\inf_{\pi\in\mathrm{DET}_r}\KL(p_\lambda\|Q_\pi),\\
F^{\mathrm{NA}}_r=\inf_{\mu\in\mathrm{NA}_r}\KL(p_\lambda\|Q_\mu).
\end{gather*}
Since $\mathrm{DET}_r,\mathrm{NA}_r\subset\mathrm{RAND}_r$, we have $F^*_r\le\min(D_r,F^{\mathrm{NA}}_r)$.
\item \emph{Path cost (the path-certificate criterion).} For $\pi\in\mathrm{DET}_r$,
$H(\pi)=\E_{x\sim p_\lambda}\sum_t\mathrm{TC}_t(x)$. Here $\mathrm{TC}_t(x)$ is the total correlation of the values of $\beta_t$ under the target, given the prefix before
round $t$: the divergence of their conditional joint law from the product of their exact marginals. For $\pi\in\mathrm{RAND}_r$, let $\pi_\omega$ be $\pi$ with its seed
fixed. Horizons, histories and action sets are finite, so the seed's law pushes forward onto finitely many seedless labelled
policies $\pi_\omega$. Every one of positive mass is in $\mathrm{DET}_r$, because its execution paths are execution paths of
$\pi$ and so complete within $r$ rounds; a seed value of probability zero may give a $\pi_\omega$ outside $\mathrm{DET}_r$,
but it carries no mass. $H(\pi)$ is the average of $H(\pi_\omega)$ under this pushforward. This extends the
$H(\mu)$ of Section~\ref{sec:setup} from predrawn mixtures to randomized policies: an average of the output KLs of
seed-fixed policies (Lemma~TGR1(a)), which by convexity is at least $\KL(p_\lambda\|Q_\pi)$.
\item \emph{Target count kernel.} For legal $s$ and a composition $b=(b_A,b_B)$ with $0\le b_g\le m_g$, let
\[
T^{\mathrm{tgt}}_b(s)(u,v)=\binom{b_A}{u}\binom{b_B}{v}\lambda^u\,\frac{Z(m_A-b_A,\,m_B-b_B,\,j-u-v)}{Z(m_A,m_B,j)},
\]
where
\[
Z(m_A,m_B,j)=\sum_u\binom{m_A}{u}\binom{m_B}{j-u}\lambda^u,
\]
with $Z=0$ unless $0\le j\le m_A+m_B$. By Lemma~TG0(i), $T^{\mathrm{tgt}}_b(s)$ is the law under $p_\lambda$, given a prefix
with state $s$, of the numbers of ones $(h_A,h_B)$ in a batch of composition $b$; under the target, the child is always
legal. The sampler's law is $\Bin(b_A,a_A(s))\otimes\Bin(b_B,a_B(s))$ (Lemma~TG0(iv)). Define
$c(s,b)=\KL\big(T^{\mathrm{tgt}}_b(s)\,\|\,\Bin(b_A,a_A(s))\otimes\Bin(b_B,a_B(s))\big)$, so $c(s,0)=0$. Then
$c(s,b)<\infty$: if $0<j<M$, every relevant marginal lies in $(0,1)$ and the sampler's law gives positive probability to every count; if
$j\in\{0,M\}$, both laws are the same point mass.
\item \emph{Extended count state.} After round $t$, let $\bar s_t=(m_A,m_B,j,u_A)$ at a legal prefix, where $u_A$ is the
number of committed ones in group A, and $\bar s_t=(\bot,m_A,m_B)$ at an illegal prefix. The count $u_A$ is carried because the A-count $|x_A|$ of a valid output, which indexes the terminal vector, is the final value of $u_A$ (proof of Lemma~TGR3, step~(1)). A \emph{deterministic Markov count
policy} $\delta=(\delta_t)_{0\le t<r}$ maps $\bar s_t$ to a composition $b\le(m_A,m_B)$; it is realised by committing the
lowest-indexed free coordinates of each group. $\mathcal D_r$ is the finite set of such $\delta$ whose realisation satisfies
the every-path hard cap $r$, and $\mathcal V_r=\{q(\delta):\delta\in\mathcal D_r\}$; by Lemma~TGR3, $\mathrm{conv}\,\mathcal V_r$ is the set of terminal vectors of the policies in $\mathrm{RAND}_r$. For $c\in(0,\infty)^{\mathcal W}$,
$M^*(c)=\max_{\delta\in\mathcal D_r}\sum_wP_wc_wq_w(\delta)$, the largest value over $\mathcal V_r$ of a linear function of the terminal vector; it is the oracle of Lemma~TGR4's lower bound, and TGR4(b) computes it by a recursion.
\item \emph{Controlled count kernel.} For an extended count state $\bar s$ and a composition $b=(b_A,b_B)$ with
$b\le(m_A,m_B)$, $K(\cdot\mid\bar s,b)$ is the following law of the next extended state. If $\bar s=(m_A,m_B,j,u_A)$ is
legal, draw $h_A\sim\Bin(b_A,a_A(s))$ and $h_B\sim\Bin(b_B,a_B(s))$ independently, and move to
$(m_A-b_A,m_B-b_B,j-h_A-h_B,u_A+h_A)$ if that child is legal, and to $(\bot,m_A-b_A,m_B-b_B)$ otherwise. If
$\bar s=(\bot,m_A,m_B)$, move to $(\bot,m_A-b_A,m_B-b_B)$. Step~(1) of the proof of Lemma~TGR3 shows that, under every policy in $\mathrm{RAND}_r$,
$K(\cdot\mid\bar s_t,b_{t+1})$ is the conditional law of $\bar s_{t+1}$ given $\mathcal F_t$; it does not depend on the policy.
\end{itemize}

\emph{Notation.} Some letters have two roles here. $D_r$ is a frontier value and $\mathcal D_r$ a set of count policies. $M_t$ is the number of free coordinates after round $t$, while $M^*(c)$ is an oracle value and $\mathcal M_t$ its recursion (TGR4(b)). A count state is $s$, and $s(q)=\sum_wq_w$ is the success mass of a terminal vector. $c(s,b)$ is a one-round cost, while $c\in(0,\infty)^{\mathcal W}$ is the argument of $M^*$ in TGR4 and of the analogous maximum over $\Sigma_r$ in TGR6(c). $G$ is the group of TGR2 and $g$ its elements, while $G^\lambda_r$ is the survival optimum, $\mathcal G_t$ is a $\sigma$-field in the proof of TGR1, a subscript $g\in\{A,B\}$ names a group, and $g$ in TGR7 bounds a gap. $w\in\mathcal W$ indexes weight classes, while $w_i$ are mixture weights.

\emph{Reported quantities and the lemmas behind them.} Each item names quantities reported in Appendix~\ref{app:exhibits} (and summarized in Appendix~\ref{app:alt}), the object they are values of, how it is computed and the lemmas it rests on. The lemmas identify each object, which is defined through policies, with the value of a finite program (a recursion over count states, or the minimum of $\Phi$ over the convex hull of finitely many terminal vectors), or bracket it by such values; ``computed by'' names that program, and no proof depends on how the program was evaluated. The arithmetic of each evaluation (exact rationals, $60$-digit, float64 or ball arithmetic) is described after the proofs, and the evidence class of each printed number is given in Appendix~\ref{app:exhibits}.
\begin{itemize}
\item $D_r$, the ratio $D_r/F^*_r$ (when $F^*_r>0$) and the least caps through the path certificate ($\min\{r:D_r\le\varepsilon\}$): the object is
$D_r=\min_{\mathrm{DET}_r}\KL=\min_{\mathrm{RAND}_r}H$, computed by the recursion $\mathcal C_r(\mathrm{root})$; it rests on
Lemma~TGR1 (with Lemma~TG0).
\item $F^*_r$, including its certified brackets at $r\le4$, and the least caps for the output law ($\min\{r:F^*_r\le\varepsilon\}$): the object is
$\inf_{\mathrm{RAND}_r}\KL$; a bracket's lower end is TGR4(a) with $c=1/\hat q$, and its upper end is $\Phi(\hat q)$ of an explicitly
realised mixture (TGR4(c)); it rests on Lemmas~TGR2, TGR3 and~TGR4.
\item $G^\lambda_r$; the sandwich $-\ln G^\lambda_r\le F^*_r\le F(\text{survival planner})$, with the planner of TGR5 below; the comparison of $-\ln G^\lambda_r$ with $F^*_r$ reported in Appendix~\ref{app:exhibits}; the
case $G^\lambda_3=1$, $F^*_3\approx5.0\cdot10^{-4}$ at $(n_A,n_B,k)=(2,2,2)$, $\lambda=2$, $r=3$: the object is the survival optimum, computed by the recursion of
Appendix~\ref{app:tgenv}; it rests on TGR5 (Lemmas~TG9 and~TG11), TGR2 and~TGR4.
\item $F^{\mathrm{NA}}_r$; the witness factor $\ge2.35$ at $(n_A,n_B,k)=(3,3,3)$, $\lambda=4$, $r=3$; the $2$--$9\%$ losses of predrawn mixtures at $\lambda=1$ (equal splits $n_A=n_B=k$ with $n\ge16$, $r=3,4$; ratios to $F^*_r$, when
$F^*_r>0$): the object is
$\min_{\mathrm{NA}_r}\KL$, computed over the hull of $\Sigma_r$ (TGR6); it rests on Lemmas~TGR6 and~TGR2, and on TGR4 for the $F^*_r$
in the ratio.
\item the survival given up by the optimal law; the $29$--$65\%$ ($n=12,16$, $r=5,6$) and $16\%$ ($n=20$, $r=5$) shape shares, at the equal splits $n_A=n_B=k$ under Grid~Z's strongest tilt $z=2$ (Appendix~\ref{app:exhibits}): the object is the leakage--shape split of the unique minimiser $q^*$ of TGR3,
computed by the TGR7 brackets at a Frank--Wolfe iterate; it rests on TGR3(6), TGR7 and~TGR5.
\item the simple planners, namely the survival planner of TGR5 and the tilt-blind planner (the balanced slice's survival-optimal rule, relabelled over all coordinates): the object is the KL of a relabelled count policy, computed as
$\Phi(q(\pi))$; it rests on Lemma~TGR2.
\end{itemize}
Ratios with $F^*_r$ in the denominator are stated only when $F^*_r>0$. At $r=n$ all three frontier values are $0$:
committing one coordinate per round samples $p_\lambda$ exactly.

\emph{TGR0 (conditional law).} This is Lemma~TG0, (i)--(iv), of Appendix~\ref{app:tgenv}, used unchanged.

\emph{Lemma TGR1 (deterministic policies).} (a) For every $\pi\in\mathrm{DET}_r$,
$\KL(p_\lambda\|Q_\pi)=H(\pi)=\E_{p_\lambda}\sum_{t=0}^{r-1}c(s_t,b_{t+1})$. Here $b_{t+1}$ is the composition of
$\beta_{t+1}$, and under $p_\lambda$ the state moves from $s_t$ to $(m_A-b_A,\,m_B-b_B,\,j-u-v)$ with
$(u,v)\sim T^{\mathrm{tgt}}_{b_{t+1}}(s_t)$. (b) $D_r=\mathcal C_r(\mathrm{root})$ for the following recursion on legal
states: $\mathcal C_0(s)=0$ if $M=0$, and $\mathcal C_0(s)=+\infty$ if $M\ge1$; for $t\ge1$, $\mathcal C_t(s)=0$ if $M=0$,
and otherwise
\[
\mathcal C_t(s)=\min_{0\le b\le(m_A,m_B)}\Big\{c(s,b)+\E_{T^{\mathrm{tgt}}_b(s)}\,\mathcal C_{t-1}(s')\Big\}.
\]
The infimum defining $D_r$ is attained in $\mathrm{DET}_r$. (c) $H(\pi)\ge D_r$ for every $\pi\in\mathrm{RAND}_r$, with
equality at a minimiser of (b). So $D_r$ is also the least path cost over $\mathrm{RAND}_r$, and, when $F^*_r>0$,
$D_r/F^*_r$ is the path certificate's overstatement at a matched hard cap.

\emph{Lemma TGR2 (orbit reduction).} Let $G=S_{n_A}\times S_{n_B}$ act by permuting coordinates within groups. (a) For every
probability law $Q$ on $\{0,1\}^n$, $\KL(p_\lambda\|Q)\ge\Phi(q(Q))$, with equality when $Q$ restricted to $S_k$ is
$G$-invariant. The output law of every policy in $\mathrm{RAND}_r$ is a probability law, because every execution path
commits all $n$ coordinates. (For a sub-probability $Q$, extend it to $\{0,1\}^n\cup\{\dagger\}$ with
$Q(\dagger)=1-Q(\{0,1\}^n)$ and $p_\lambda(\dagger)=0$; the proof of (a) runs verbatim with $\dagger\mapsto\bot$, and neither side
changes.) (b) For $\pi\in\mathrm{RAND}_r$,
let $\bar\pi$ draw $g\sim\mathrm{Unif}(G)$ together with its seed and run $\pi$
on the coordinates relabelled by $g$. Then $\bar\pi\in\mathrm{RAND}_r$, the law $Q_{\bar\pi}=|G|^{-1}\sum_gg_\#Q_\pi$ is
$G$-invariant, and $q(\bar\pi)=q(\pi)$. If $\pi\in\mathrm{NA}_r$, then $\bar\pi\in\mathrm{NA}_r$. (c) Hence
$F^*_r=\inf_{\pi\in\mathrm{RAND}_r}\Phi(q(\pi))$ and $F^{\mathrm{NA}}_r=\inf_{\mu\in\mathrm{NA}_r}\Phi(q(\mu))$. Moreover
$\Phi(q)=-\ln s(q)+\KL(P\|q/s(q))$ when $s(q)>0$: leakage plus between-orbit shape.

\emph{Lemma TGR3 (hull of count policies).} $\{q(\pi):\pi\in\mathrm{RAND}_r\}=\mathrm{conv}\,\mathcal V_r$. Consequently
\[
F^*_r=\min_{q\in\mathrm{conv}\,\mathcal V_r}\Phi(q),
\]
and the minimum is attained at a unique $q^*$, with $q^*>0$.

\emph{Lemma TGR4 (certified lower bound).} (a) For every $c\in(0,\infty)^{\mathcal W}$,
\[
F^*_r\ge\sum_wP_w\ln(P_wc_w)+1-M^*(c).
\]
In particular, for any $\hat q\in(0,\infty)^{\mathcal W}$, which need not be feasible,
$F^*_r\ge L(\hat q):=\Phi(\hat q)+1-M^*(1/\hat q)$. (b) $M^*(c)=\mathcal M_r(\bar s_0)$, with $\bar s_0=(n_A,n_B,k,0)$, for
the backward recursion over the extended count states: $\mathcal M_0(y)=P_wc_w$ if $y=(0,0,0,w)$, and $\mathcal M_0(y)=0$
otherwise; $\mathcal M_t(y)=\max_b\sum_{y'}K(y'\mid y,b)\,\mathcal M_{t-1}(y')$ for $t\ge1$, where $K$ is the controlled count
kernel defined above. (c) $F^*_r\le\Phi(q)$ for every $q\in\mathrm{conv}\,\mathcal V_r$.

\emph{TGR5 (survival; Appendix~\ref{app:tgenv}).} For every $\pi\in\mathrm{RAND}_r$,
$s(q(\pi))=Q_\pi(S_k)\le G^\lambda_r(\mathrm{root})$ (Lemma~TG9), and equality holds for a seedless count policy $\pi^\circ$
(Lemma~TG11). Hence $-\ln G^\lambda_r(\mathrm{root})\le F^*_r\le\Phi(q(\pi^\circ))$. The right-hand side is the KL of
$\pi^\circ$ followed by a uniform within-group relabelling (TGR2), and the ``survival planner'' of the tables is such a policy. Where several block compositions attain the maximum in the recursion for $G^\lambda_r$, the planner takes the
lexicographically first $(b_A,b_B)$, comparing exact values.

\emph{Lemma TGR6 (predrawn frontier).} For a full labelled schedule $U=(\beta_1,\dots,\beta_L)$, let
$\mathbf b(U)=\big((|\beta_t\cap A|,|\beta_t\cap B|)\big)_{t\le L}$. Let $\Sigma_r$ be the finite set of sequences of at most
$r$ nonzero compositions that sum to $(n_A,n_B)$, and let $q(\mathbf b)$ be the terminal vector of the controlled count kernel $K$
run with the fixed actions $\mathbf b$. Then: (a) $q(U)=q(\mathbf b(U))$. (b)
$F^{\mathrm{NA}}_r=\min\{\Phi(q):q\in\mathrm{conv}\{q(\mathbf b):\mathbf b\in\Sigma_r\}\}$, and the minimum is attained. (c) For
every $c\in(0,\infty)^{\mathcal W}$,
$F^{\mathrm{NA}}_r\ge\sum_wP_w\ln(P_wc_w)+1-\max_{\mathbf b\in\Sigma_r}\sum_wP_wc_wq_w(\mathbf b)$; and $F^{\mathrm{NA}}_r\le\Phi(q)$
for every $q$ in that hull. (d) $F^*_r\le F^{\mathrm{NA}}_r$.

\emph{Lemma TGR7 (the optimal law's leakage--shape split).} Let $q^*$ be the minimiser of TGR3 and $s^*=s(q^*)$, so that
$F^*_r=-\ln s^*+\KL(P\|q^*/s^*)$. For every $\hat q\in\mathrm{conv}\,\mathcal V_r$ with $\hat q>0$, and every
$g\ge\Phi(\hat q)-F^*_r$,
\[
\sum_wP_w\,\varphi(\hat q_w/q^*_w)\le g,\qquad\varphi(y)=y-1-\ln y\ge0.
\]
Hence each ratio $\hat q_w/q^*_w$ lies between the two roots $y^-_w\le1\le y^+_w$ of $\varphi(y)=g/P_w$, and
\[
s^*\in\big[s_{\mathrm{lo}},\,s_{\mathrm{hi}}\big]=\Big[\sum_w\hat q_w/y^+_w,\ \min\Big\{G^\lambda_r(\mathrm{root}),\,1,\,
\sum_w\min(1,\hat q_w/y^-_w)\Big\}\Big].
\]
With $F^*_r\in[\Phi(\hat q)-g,\,\Phi(\hat q)]$ and $\Phi(\hat q)-g>0$, the shape share $1+\ln s^*/F^*_r$ lies in
$\big[1+\ln s_{\mathrm{lo}}/(\Phi(\hat q)-g),\ 1+\ln s_{\mathrm{hi}}/\Phi(\hat q)\big]$.
\emph{Proof of TGR1.} (a) Fix $x$ with $p_\lambda(x)>0$. \emph{Batches.} Because $\pi$ has no seed, the batches
$\beta_1(x),\beta_2(x),\dots$ that it uses when the committed values agree with $x$ are functions of $x$, and every prefix
along $x$ is legal. \emph{Sampler probability.} $Q_\pi(x)=\prod_t\prod_{i\in\beta_t(x)}a_i^{x_i}(1-a_i)^{1-x_i}$, with the
marginals taken at the prefix before round $t$. This is positive: at $0<j<M$ every relevant $a_i\in(0,1)$, and at
$j\in\{0,M\}$ the marginals are $0$ or $1$ and agree with $x$. \emph{Target probability.}
$p_\lambda(x)=\prod_tp_\lambda(x_{\beta_t(x)}\mid\text{prefix})$. \emph{Expectation.} Take logarithms, then $\E_{p_\lambda}$,
conditioning on the prefix before each round. Given the prefix, $\beta_t$ is fixed and $x_{\beta_t}$ has the target's
conditional law, so the $t$-th term contributes $\E\,\mathrm{TC}_t$. \emph{Reduction to counts.} By Lemma~TG0(i), the
target's conditional joint law of the batch values is $\lambda^{u}Z(m_A-b_A,m_B-b_B,j-u-v)/Z(s)$ at every configuration
with counts $(u,v)$. The product of the marginals is $a_A^u(1-a_A)^{b_A-u}a_B^v(1-a_B)^{b_B-v}$. Both laws are constant on
each fiber $\{(h_A,h_B)=(u,v)\}$, so the likelihood ratio is a function of the counts, and the divergence equals that of
the count laws, $c(s_t,b_{t+1})$. \emph{Transition.} The target transition of the count state is Lemma~TG0(i),(iii).

(b) \emph{Lower bound.} Let $\pi\in\mathrm{DET}_r$ and work under $p_\lambda$. Let $\mathcal G_t$ be the $\sigma$-field of
the values committed through round $t$, and $W_t=\sum_{i=1}^{t}c(s_{i-1},b_i)+\mathcal C_{r-t}(s_t)$. The composition
$b_{t+1}$ is $\mathcal G_t$-measurable, and by Lemma~TG0(i),(iii) the next counts have conditional law
$T^{\mathrm{tgt}}_{b_{t+1}}(s_t)$ given $\mathcal G_t$. So
\[
\E[W_{t+1}\mid\mathcal G_t]=W_t-\mathcal C_{r-t}(s_t)+c(s_t,b_{t+1})+\E_{T^{\mathrm{tgt}}_{b_{t+1}}(s_t)}\mathcal C_{r-t-1}(s')\ge W_t,
\]
by the definition of $\mathcal C$ as a minimum. For $M_t=0$ the batch is empty and the added terms vanish. Every term is
finite for $t<r$: for $t\ge1$, $\mathcal C_t(s)\le c(s,\text{full batch})<\infty$, because the full batch leads to a state
with $M=0$. Every factor of the sampler probability of $x\sim p_\lambda$ is positive (proof of (a)),
so every truncation of $x$'s path is an execution path of $\pi$, and the every-path hard cap applies to it: $M_r=0$, and
$\mathcal C_0(s_r)=0$. Hence $H(\pi)=\E\,W_r\ge W_0=\mathcal C_r(\mathrm{root})$.
\emph{Attainment.} For $1\le\varrho\le r$ and legal $s$ with $M\ge1$, fix a minimiser $b^\circ(\varrho,s)$. Consider the seedless
policy that acts at round $t+1$ as follows: it commits the lowest-indexed $b^\circ(r-t,s_t)$ free coordinates when $s_t$ is legal
with $M_t\ge1$; it commits all remaining coordinates when $s_t$ is illegal; and it does nothing when $M_t=0$. It lies in
$\mathrm{DET}_r$: for every non-full batch $b$, $\mathcal C_1(s)<\infty=\E\,\mathcal C_0(s')$, which forces $b^\circ(1,s)$
to be the full batch, so every legal path completes by round $r$, and a path that first becomes illegal after round $t<r$
completes in round $t+1$. Along this policy $W_t$ is a martingale, so $H=\mathcal C_r(\mathrm{root})$, and by (a) its KL
equals $H$.

(c) $H(\pi)$ is an average of $H(\pi_\omega)$ over seedless policies of positive mass, each in $\mathrm{DET}_r$ (see the
definitions), so $H(\pi)\ge D_r$ by (a) and (b). \qed

\emph{Proof of TGR2.} (a) Apply data processing under the map $x\mapsto|x_A|$ on $S_k$ and $x\mapsto\bot$ off $S_k$. The image
of $p_\lambda$ is $P$, because $p_\lambda(S_k)=1$; the image of $Q$ is $(q,\,1-s(q))$, because $Q$ is a probability law; and
the divergence of the images is $\Phi(q)$, since $P$ puts no mass on $\bot$. For equality: $G$ acts transitively on each class $\{x\in S_k:|x_A|=w\}$, which has $N_w$ points. There
$p_\lambda(x)=P_w/N_w$. A $G$-invariant $Q$ is constant on each $G$-orbit, and this class is one orbit with $Q$-mass $q_w$,
so $Q(x)=q_w/N_w$ there. Hence $\KL(p_\lambda\|Q)=\sum_w\sum_{|x_A|=w}\frac{P_w}{N_w}\ln\frac{P_w/N_w}{q_w/N_w}=\sum_wP_w\ln\frac{P_w}{q_w}=\Phi(q)$, where both sides are $+\infty$ if some $q_w=0<P_w$.

(b) \emph{Relabelling commutes with the interface.} $p_\lambda$ is $G$-invariant, so the exact marginal of a coordinate at a
prefix depends only on its group and the count state (Lemma~TG0(ii)), and $g$ preserves both. Relabel the auxiliary
variables together with the coordinates. This couples each run of $\pi$ that produces $x$ with a run of the relabelled
policy that produces $gx$, so the relabelled policy's output law is $g_\#Q_\pi$, and its execution paths are the relabelled
paths of $\pi$, with the same number of rounds. The draw of $g$ is made before any bit, as part of the seed, which is independent of all
auxiliary randomness. The terminal vector $q$ is unchanged because $|x_A|$ and $|x|$ are $G$-invariant. Relabelling a full labelled schedule
gives a full labelled schedule with the same number of blocks.

(c) By (a) and (b), $\KL(p_\lambda\|Q_\pi)\ge\Phi(q(\pi))=\KL(p_\lambda\|Q_{\bar\pi})$, with $\bar\pi$ in the same class as
$\pi$. The decomposition is algebra, for $s=s(q)>0$: $\sum_wP_w\ln(P_w/q_w)=-\ln s+\sum_wP_w\ln\big(P_w/(q_w/s)\big)$. \qed

\emph{Remarks.} A relabelling across the groups may also be drawn as part of the seed, but for $\lambda\ne1$ it is not a symmetry of $p_\lambda$,
and it gives nothing beyond (a). For a policy whose output law is not $G$-invariant, such as a fixed labelled schedule, the
inequality in (a) is in general strict.

\emph{Proof of TGR3.} Step~(1) shows that the extended count state is a controlled Markov chain whose kernel $K$ is the same under every policy; this uses the exact-marginal interface and the no-latent-peeking condition. Steps~(2)--(4) prove the inclusion $\subseteq$: replacing the policy's batch choice by its conditional law given $\bar s_t$ preserves the law of each $\bar s_t$, hence $q(\pi)$ (2); the resulting Markov rule is a mixture, drawn with the seed, of deterministic Markov count policies (3); and each component of positive weight satisfies the every-path hard cap, so lies in $\mathcal D_r$ (4). Step~(5) proves the reverse inclusion by realising each $\delta\in\mathcal D_r$ and each convex combination, and step~(6) proves attainment and uniqueness. (1) \emph{Controlled kernel.} Let $b_{t+1}$ be the composition of $\beta_{t+1}$; it is
$\mathcal F_t$-measurable. If $\bar s_t$ is legal, then by Lemma~TG0(iv) the counts, conditionally on $\mathcal F_t$, are
$\Bin(b_A,a_A(s_t))\otimes\Bin(b_B,a_B(s_t))$, and $\bar s_{t+1}$ is a function of $(\bar s_t,b_{t+1},h_A,h_B)$: either a
legal child, or $(\bot,m')$ if the child is illegal. If $\bar s_t=(\bot,m)$, then $\bar s_{t+1}=(\bot,m-b_{t+1})$, whatever
values are drawn (illegality is absorbing; see the proof of Lemma~TG9). So
$\Pr(\bar s_{t+1}=y'\mid\mathcal F_t)=K(y'\mid\bar s_t,b_{t+1})$ for the controlled count kernel $K$ defined before Lemma~TGR1, which does not depend on the policy. This uses
the no-latent-peeking condition. Under the every-path hard cap, the output lies in $S_k$ with $|x_A|=w$ iff
$\bar s_r=(0,0,0,w)$. Hence $q_w(\pi)=\Pr_\pi(\bar s_r=(0,0,0,w))$.

(2) \emph{Markovization.} For $t<r$, let $d_t(b\mid y)=\Pr_\pi(b_{t+1}=b\mid\bar s_t=y)$ when $\Pr_\pi(\bar s_t=y)>0$, and any
law otherwise. Let $\pi^d$ draw $b_{t+1}\sim d_t(\cdot\mid\bar s_t)$, using independent uniforms included in the seed, and
commit the lowest-indexed free coordinates of that composition; so $\pi^d$ is a causal randomized policy. \emph{Claim:} for
every $t\le r$, $\bar s_t$ has the same law under $\pi$ and $\pi^d$, and for $t<r$ so does the pair $(\bar s_t,b_{t+1})$.
\emph{Induction.} $\bar s_0=(n_A,n_B,k,0)$ under both. If $\bar s_t$ has the same law under both, then
$\Pr_\pi(\bar s_t=y,\,b_{t+1}=b)=\Pr_\pi(\bar s_t=y)\,d_t(b\mid y)=\Pr_{\pi^d}(\bar s_t=y,\,b_{t+1}=b)$. By (1) and the tower
property, $\Pr(\bar s_{t+1}=y')=\sum_{y,b}\Pr(\bar s_t=y,\,b_{t+1}=b)\,K(y'\mid y,b)$ under both. In particular
$q(\pi)=q(\pi^d)$.

(3) \emph{Derandomization.} Draw, as part of the seed, independent actions $\delta_t(y)\sim d_t(\cdot\mid y)$ for all of the
finitely many pairs $(t,y)$. This gives a random deterministic Markov count policy $\delta$. Along any trajectory each pair
$(t,\bar s_t)$ occurs once, so a trajectory $(y_0,b_1,y_1,\dots,b_r,y_r)$ has probability
$\prod_td_t(b_{t+1}\mid y_t)\,K(y_{t+1}\mid y_t,b_{t+1})$ under both $\pi^d$ and the mixture. Hence
$q(\pi)=\sum_\delta\Pr(\delta)\,q(\delta)$.

(4) \emph{The components respect the hard cap.} An
\emph{execution path} is a path in the support of the seed pushforward and the kernels; a word of probability zero is not a
path, and components of zero weight are omitted. A continuous seed causes no difficulty: horizons, histories and action
sets are finite, so the pushforward onto deterministic tables is finite. Let $\delta$ have positive weight, and take an execution path of $\delta$. At each round the path uses an action $b$ with
$d_t(b\mid y)>0$ at its current state, and a transition of positive $K$-probability; by induction the path has positive
probability under $\pi^d$. Under $\pi$ every execution path is complete after round $r$, so $\Pr_\pi(M_r=0)=1$, and by (2)
$\Pr_{\pi^d}(M_r=0)=1$. A positive-probability path of $\pi^d$, and in particular this path of $\delta$, is therefore complete
after round $r$, so $\delta\in\mathcal D_r$.

(5) \emph{Realisation.} Each $\delta\in\mathcal D_r$ is realised by a seedless policy in $\mathrm{DET}_r$: commit the
lowest-indexed free coordinates of each group, noting that $\bar s_t$ is a function of the committed values. This policy is causal and seedless. Its batch at round
$t+1$ is $\delta_t(\bar s_t)$, so by step (1) the extended count state is the controlled chain with kernel
$K(\cdot\mid y,\delta_t(y))$, the same chain that defines $q(\delta)$; its terminal vector is therefore $q(\delta)$. Since
$\delta\in\mathcal D_r$, every positive-probability path of that chain completes after round $r$, so the policy has the
every-path hard cap $r$ and lies in $\mathrm{DET}_r$. A predrawn choice of $\delta$ realises any convex combination. With (3) and (4), this gives
$\{q(\pi)\}=\mathrm{conv}\,\mathcal V_r$.

(6) \emph{Attainment and uniqueness.} $\mathrm{conv}\,\mathcal V_r$ is compact, and $\Phi$ is lower semicontinuous with values
in $[0,\infty]$. $\Phi$ is finite at the terminal vector of the one-round policy, which is positive because $0<k<n$ puts every
marginal in $(0,1)$. So a minimiser exists, and it is positive, because $\Phi$ is $+\infty$ wherever some $q_w=0$. $\Phi$ is strictly convex on $(0,\infty)^{\mathcal W}$ (its
Hessian is $\mathrm{diag}(P_w/q_w^2)$), so the minimiser is unique. With TGR2(c), $F^*_r=\min_{\mathrm{conv}\,\mathcal V_r}\Phi$.
\qed

\emph{Remark.} The count policies may depend on $u_A$. The target ignores $u_A$, but $\Phi$ does not, so an optimal policy may
need to be value-adaptive in the committed A-count. In an exact check this dependence is needed at $n_A=n_B=k=3$, $r=3$,
$\lambda\in\{2,4\}$ (the optimal value over predrawn randomized mixtures of $u_A$-independent deterministic count tables is worse by
$3.5\cdot10^{-5}$ and $2.3\cdot10^{-4}$ nats); in the
other instances tested, all at $r=3$ ($n\le5$, $\lambda=1$, and $(n_A,n_B,k)=(2,4,3)$ with $\lambda=2$), $u_A$-independent
policies attain $F^*_r$.

\emph{Proof of TGR4.} (a) Let $\pi\in\mathrm{RAND}_r$ and $q=q(\pi)$. By TGR2(a), $\KL(p_\lambda\|Q_\pi)\ge\Phi(q)$; if some
$q_w=0$ there is nothing to prove. Otherwise, $\ln y\le y-1$ at $y=c_wq_w$ gives $-\ln q_w\ge\ln c_w+1-c_wq_w$. Multiply by
$P_w$, add $P_w\ln P_w$, sum, and use $\sum_wP_w=1$:
\[
\Phi(q)\ge\sum_wP_w\ln(P_wc_w)+1-\sum_wP_wc_wq_w.
\]
By TGR3, $q\in\mathrm{conv}\,\mathcal V_r$, and the linear function $q\mapsto\sum_wP_wc_wq_w$ is at most its maximum over the
vertices $\mathcal V_r$, which is $M^*(c)$. Take the infimum over $\pi$. With $c=1/\hat q$ the right-hand side is
$\Phi(\hat q)+1-M^*(1/\hat q)$.

(b) \emph{The recursion computes $M^*$.} For deterministic Markov policies of the finite controlled chain $K$, the expected
terminal reward from $(t,y)$ follows the recursion along the policy, and the maximum over policies follows the recursion with
a maximum (induction on the number of remaining rounds). Maximising over every $\delta$, not only over $\mathcal D_r$, does
not change the value: the reward is nonnegative, a final state with $M>0$ earns $0$, and replacing every last-round action by
the full batch completes every path without lowering the value.

(c) TGR3(5) realises every $q\in\mathrm{conv}\,\mathcal V_r$ by some $\pi\in\mathrm{RAND}_r$; TGR2(b) makes it $G$-invariant
without changing $q$, and TGR2(a) then gives $\KL=\Phi(q)$. \qed

\emph{Proof of TGR5.} It follows from Lemmas~TG9 and~TG11 of Appendix~\ref{app:tgenv}, together with TGR2(c): $\Phi(q)\ge-\ln s(q)$
because the shape term is a KL. \qed

\emph{Proof of TGR6.} (a) $U$ has no seed, and its batches do not depend on values; by step~(1) of the proof of TGR3, its extended count
state evolves by the controlled count kernel $K$ with the actions $\mathbf b(U)$. (b) By TGR2(c), $F^{\mathrm{NA}}_r=\inf_\mu\Phi(q(\mu))$, where
$q(\mu)=\sum_iw_iq(\mathbf b(U_i))$. Every $\mathbf b\in\Sigma_r$ equals $\mathbf b(U)$ for some schedule (take the lowest indices
first), so $q(\mu)$ ranges over the whole hull. The minimum is attained by compactness and lower semicontinuity, since $\Phi$
is finite at the one-block schedule. (c) As in TGR4(a),(c), with the finite set $\{q(\mathbf b)\}$ in place of $\mathcal V_r$.
The lower bound needs only TGR2(a) and $q(\mu)$ in the hull; the upper bound also uses that $\mathrm{NA}_r$ is closed under
relabelling (TGR2(b)). (d) $\mathrm{NA}_r\subset\mathrm{RAND}_r$. \qed

\emph{Proof of TGR7.} On $(0,\infty)^{\mathcal W}$, $\partial_w\Phi(q)=-P_w/q_w$, and $q^*>0$ (TGR3(6)). Since $q^*$ minimises the
convex $\Phi$ over the convex set $\mathrm{conv}\,\mathcal V_r$, $\nabla\Phi(q^*)\cdot(\hat q-q^*)\ge0$. The Bregman identity
\[
\Phi(\hat q)-\Phi(q^*)-\nabla\Phi(q^*)\cdot(\hat q-q^*)=\sum_wP_w\big[\ln(q^*_w/\hat q_w)+\hat q_w/q^*_w-1\big]
=\sum_wP_w\,\varphi(\hat q_w/q^*_w)
\]
then gives $\sum_wP_w\varphi(\hat q_w/q^*_w)\le\Phi(\hat q)-F^*_r\le g$; each term is nonnegative, so each is at most $g$.
Moreover $q^*_w\le1$, and $s^*\le G^\lambda_r(\mathrm{root})$ by TGR5. The share bracket uses that $1+\ln s/F$ increases in $s$
and, because $\ln s\le0$, in $F$. \qed

\emph{Remark (order).} As $g\to0$, $\varphi(1+u)=u^2/2+O(u^3)$ gives $y^\pm_w=1\pm\sqrt{2g/P_w}+O(g/P_w)$, so
$1/y^-_w-1/y^+_w=2\sqrt{2g/P_w}+O(g/P_w)$. Hence, before the caps at $G^\lambda_r(\mathrm{root})$ and $1$ (which bind in some tested
rows), the half-width of the $s^*$ bracket is $\sqrt{2g}\sum_w\hat q_w/\sqrt{P_w}+o(\sqrt g)$. The order $\sqrt g$ cannot be
improved for brackets of this form, as a function of $(P,\hat q,g)$ on a convex set, as the following example shows; it is an abstract segment, not
a realised $\mathrm{conv}\,\mathcal V_r$. Take $P=(\frac12,\frac12)$ and a segment through $q^*=(\frac15,\frac25)$ in the
direction $(1,-2)$, along which $q^*$ is the minimiser (the directional derivative is $0$). Along it $\Phi$ rises by
$-\frac12\ln(1-25t^2)=\Theta(t^2)$, while $s$ moves by $-t$.
\emph{Computation behind Appendix~\ref{app:exhibits} (not part of the proofs).} The Frank--Wolfe method is fully corrective over the columns $q(\delta)$: the
master problem $\max_\theta\sum_wP_w\ln(\mathsf Q\theta)_w$, where the columns of $\mathsf Q$ are the $q(\delta)$ generated so far, is solved by multiplicative updates polished by SLSQP, and the weights
stay on the simplex, so the iterate lies in the hull. The linear oracle is TGR4(b), run in float64; an independent recursion
agrees with it to within $2^{-52}\approx2.22\cdot10^{-16}$ on all rows, and the float64 value differs from a $60$-digit evaluation by
$1.7$--$3.1\cdot10^{-16}$ on the four checked rows. For the $F^*_r$
certificates, when $\lambda$'s denominator is below $100$ the Frank--Wolfe weights are each rationalised to a denominator at most
$10^{15}$ and then normalized exactly to sum to one, $q$ and $M^*$ are exact rationals, and only the final log-sums are evaluated at $60$ digits, with a $10^{-40}$
outward margin; otherwise the computation is $60$-digit throughout, non-interval, with an a-priori margin. The certificates
store their endpoints rounded to nearest at $20$ significant digits (widen them by $10^{-19}$). $D_r$ (TGR1(b)), $G^\lambda_r$ and
$F^{\mathrm{NA}}_r$ are computed in float64, $F^{\mathrm{NA}}_r$ by Frank--Wolfe over the hull of $\Sigma_r$, enumerated exhaustively
with prefix sharing. The exceptions are the $27$ validation cells with $n\le8$ (rational arithmetic, logs at $60$ digits,
checked against a labelled brute force) and the seven $F^{\mathrm{NA}}_r$ certificates, whose exact rational ingredients are evaluated in Arb ball arithmetic (rigorous enclosures). The TGR7 brackets are evaluated
in float64 with $g=\max(\text{Frank--Wolfe gap},0)+10^{-13}$, so they are estimates, not certified bounds. Grid~Z, the grid of normalized tilts used in some exhibit rows, is
defined in Appendix~\ref{app:exhibits}.

\section{Finite exhibits: details and evidence classes}\label{app:exhibits}
This appendix gives the details and evidence classes of the finite exhibits of Section~\ref{sec:finite} and Appendix~\ref{app:alt}: two groups, the balanced slice (Table~\ref{tab:finite}) and full support (Table~\ref{tab:fs}).
\emph{Terms.} The \emph{owner} implementation of a number is the one that first produced it; a \emph{blind}
implementation was written from a prose specification without access to the owner's code; \emph{owner-only} means that
no second rigorous implementation reproduced the value; a kernel-checked Lean bracket, where one exists, is stated separately and does not count as such an implementation.
\paragraph{Two groups.} The target is $p_\lambda(x)\propto\lambda^{|x_A|}$ on $\{x:|x|=k\}$, with $n_A$ coordinates in group~A, $n_B=n-n_A$ in group~B and weight $\lambda>0$ on group~A (Appendix~\ref{app:tgenv}); a target is written $(n_A,n_B,k,\lambda)$, and an \emph{equal split} has $n_A=n_B=k=n/2$. The two-group numbers use the notation of Appendix~\ref{app:alt}: $F^*_r$ is the least output KL of causal randomized policies with hard cap $r$ (the class $\mathrm{RAND}_r$ of Appendix~\ref{app:tgenv}, which is the class $\mathcal A_r$ of Section~\ref{sec:setup} for this target), $D_r$ the least output KL of the deterministic ones, and $G^\lambda_r$ the largest probability with which such a policy meets the quota within $r$ rounds. The finite programs behind them are justified in Appendix~\ref{app:tgred}, whose list of reported quantities names the lemmas behind each number. The balanced identity of Theorem~\ref{thm:conv}(a) becomes a sandwich on two groups, $-\ln G^\lambda_r\le F^*_r\le F(\text{survival planner})$ (TGR5 in Appendix~\ref{app:tgred}); at deeper caps the optimal law gives up some survival to reduce between-orbit shape. A (possibly randomized) rule that chooses each block's group composition from the committed counts beats every predrawn mixture by a factor of at least $2.35$ in a six-coordinate witness ($(n_A,n_B,k,\lambda)=(3,3,3,4)$, $r=3$; kernel-checked: $F^{\mathrm{NA}}_3\ge2.35\,F^*_3$, and also with $2.358$, Appendix~\ref{app:lean}). At $r=4$ and $\lambda\ne1$ this adaptive gain, the ratio $F^{\mathrm{NA}}_r/F^*_r$ defined below, shrinks with $n$ in the equal splits computed ($n=12$ to $24$); at $r=3$ it does not, except under Grid~Z's $z=2$. Grid~Z denotes the normalized tilts $\lambda_z$, rational approximations (denominators at most
$10^9$) of $\exp(z/\sigma_n)$ with $\sigma_n^2=n^2/(16(n-1))$ and $z\in\{\frac12,1,2\}$; the strongest tilt is $z=2$.
$F^{\mathrm{NA}}_r$ denotes the predrawn optimum, the least output KL of predrawn mixtures of schedules with at most $r$
blocks. Evidence classes at $r\le4$, by object, follow. \emph{$F^*_r$:} rigorous enclosures (ball arithmetic) by code written
blind from the owner's specification (code-independent, not specification-independent), on the specification grids
(equal splits only: the $27$ validation cells with $n\in\{4,6,8\}$, $\lambda\in\{1,2,4\}$ and $r\le3$, and the $30$ cells with $n\in\{12,16\}$, $\lambda\in\{1,2\}$ or a Grid~Z tilt, and $r=2,3,4$) and on the $24$
held-out equal-split cells at $n=20,24$ ($\lambda=2$ and Grid~Z), the latter by an owner-run replay of the blind code. On
the remaining $18$ rows, which are owner-only (the unequal-split and off-centre targets $(n_A,n_B,k,\lambda)=(4,8,6,3)$,
$(5,11,8,2)$, $(6,10,4,4)$ and $(8,8,5,2)$, and $\lambda=1$ at $n=20,24$, each at $r=2,3,4$), the owner's certificate has
exact rational ingredients (the terminal vector of an exactly realised mixture, whose Frank--Wolfe weights are each
rationalised to a denominator at most $10^{15}$ and then normalized exactly, and the value $M^*(c)$ of Lemma~TGR4's linear oracle, which gives the lower end) and evaluates the final logarithms at $60$ digits with a
$10^{-40}$ outward margin (upper end rounded up, lower end down), not in interval arithmetic. An owner-run re-evaluation of
those logarithms in Arb ball arithmetic \citep{johansson2017arb} ($256$ bits) encloses $F^*_r$ rigorously on these rows too, to width below
$10^{-12}$. Each Arb bracket intersects the stored non-interval bracket once the latter's printed endpoints are widened by
$10^{-19}$ (a consistency check, not a proof that the stored bracket contains $F^*_r$). On all $18$ rows a kernel-checked Lean~4 theorem also brackets $F^*_r$ rationally, to
width below $10^{-12}$, for the class $\mathrm{RAND}_r$ with the almost-sure cap and every placement of group~A
(Appendix~\ref{app:lean}). \emph{$D_r$ and
$G^\lambda_r$:} the owner computes them in rational arithmetic with $60$-digit logarithms on the $27$ validation cells
(matched to a labelled brute force) and as float64 evaluations of exact dynamic programs elsewhere; the blind code
encloses them rigorously on the specification grids and the $24$ replayed cells, agreeing with the owner's values to within
$6.7\cdot10^{-16}$, and they are owner-only elsewhere. An independent float64 planner also reproduces $D_r$, $F^*_r$ and
$F^{\mathrm{NA}}_r$ at $n\in\{12,16\}$, to float64 precision. \emph{$F^{\mathrm{NA}}_r$:} owner float64 estimates, except at
seven cells, $(n_A,n_B,k,\lambda,r)=(3,3,3,1,3)$,
$(3,3,3,2,3)$, $(3,3,3,4,3)$ (the six-coordinate witness), $(4,8,6,3,3)$, $(4,8,6,3,4)$, $(6,6,6,2,3)$ and $(6,6,6,2,4)$, where
owner certificates for the two bounds of Lemma~TGR6(c) (Appendix~\ref{app:tgred}), with exact rational ingredients and
evaluated in Arb ball arithmetic ($256$ bits), enclose $F^{\mathrm{NA}}_r$ rigorously, and a second program, which imports none of the generator's code, re-checks all seven certificates. At all seven cells a kernel-checked Lean~4 theorem also brackets
$F^{\mathrm{NA}}_r$ rationally, on predrawn mixtures with at most $r$ blocks and for every placement of group~A
(Appendix~\ref{app:lean}). \emph{Depth runs
($r\le6$) and the deep-cap split ($r=5,6$; the optimal law's leakage--shape split of Lemma~TGR7):} owner-only float64 estimates; the Frank--Wolfe convergence diagnostic is at
most $5.4\cdot10^{-10}$ at every $r$ for the $F^*_r$ depth runs and the split and at most $1.4\cdot10^{-9}$ for the $F^{\mathrm{NA}}_r$ runs at $r=5,6$, made after the predictions below were frozen; for the owner's $F^{\mathrm{NA}}_r$ estimates at $r\le4$, on which rows~3--6 below are scored, the Frank--Wolfe value exceeds its dual lower value by at most $6.6\cdot10^{-9}$ (float64 diagnostics, not certified bounds), and the split's float64 interval
estimates are not certified. On the tested grid, the ratio $D_r/F^*_r$ increases with $\lambda$ at $r=2$ and $4$ but not at $r=3$. At
$\varepsilon=0.1$ the least cap is $5$--$8$ rounds through the path certificate (the least $r$ with $D_r\le\varepsilon$; Appendix~\ref{app:tgred}) against $4$--$5$ for the output law at
$n\le16$, and $10$ versus $5$ at $n=24$, $\lambda=1$. The survival bound $-\ln G^\lambda_r$ is at least $87.9\%$ of $F^*_r$ in $71$ of the $72$ cells at $r\le4$, $n\ge12$;
the other, the discovery cell $n=12$, Grid~Z with $z=2$, $r=4$, has $71\%$. Over the $36$ held-out cells of prediction~1 below, the minimum is $92.57\%$. At $\lambda=2$, $r=4$ the predrawn-to-optimal factor falls from $1.49$ to $1.26$ between $n=12$ and
$24$, with additive difference $0.027$--$0.036$ nats, and in the $48$ of the $63$ (target, $\varepsilon$) pairs of the depth runs ($20$ equal splits with $12\le n\le24$ and the $4+8$ target $(4,8,6,3)$, with $\varepsilon\in\{\ln2,0.1,0.02\}$) where both least caps are determined, the least cap of predrawn mixtures equals that of $F^*_r$ or exceeds it by one round (by one in $6$ pairs; computed after the predictions were
frozen). At $\lambda=1$ ($n\ge16$, $r=3,4$) predrawn mixtures lose only $2$--$9\%$ to adaptive policies. The ordinary survival planner comes within $3.8\%$ of $F^*_r$ for $\lambda\ne1$ at $r\le4$. At $r=5,6$ survival-optimal plans can tie, and the planner's $F$ then depends on the tie-break: with the lexicographic tie-break it is $2.2$ times off at $r=5$ on the $4+8$ target ($\lambda\ne1$). At these caps ($r=5,6$),
the optimal law's own between-orbit shape share is $29$--$65\%$ at $r=5,6$ under the strongest tilt at $n=12,16$, and $16\%$ at $n=20$, $r=5$ (owner-only float64
estimates at a Frank--Wolfe iterate, with float64 interval estimates for the optimal law's split), and its
survival is below $G^\lambda_r$.

\emph{Frozen predictions and phases of the cells.} The eight two-group predictions below were frozen, with their pass criteria, before any held-out cell was computed; the
held-out cells are the equal splits at $n\in\{20,24\}$ ($\lambda=2$ and Grid~Z) and the four unequal or off-centre targets above,
each at $r=2,3,4$. The \emph{discovery} cells are those computed before the predictions were frozen: every equal-split cell at $n=12$ ($\lambda\in\{1,2\}$ and the Grid~Z tilts $z\in\{\frac12,1,2\}$) at $r=2,3,4$, part of the validation cells with $n\le8$, and part of $n=16$ (all $r=2$; $r=3$ at $z\in\{0,\frac12,1,2\}$, where $z=0$ is $\lambda=1$; $r=4$ at $z=2$). At those $n=16$ cells only $F^*_r$, $-\ln G^\lambda_r$, $D_r$ and the best simple planner had been computed; $F^{\mathrm{NA}}_r$ had been computed only at $n=12$. Of the $72$ cells with $12\le n\le24$ and $r=2,3,4$, $25$ are discovery cells, $36$ are held-out, and $11$ were computed after the freeze without being held out ($\lambda=1$ at $n=20,24$; and at $n=16$, $\lambda=2$ at $r=3$, and $\lambda\in\{1,2\}$ and $z\in\{\frac12,1\}$ at $r=4$). Five passed and three failed; the last column gives the statement each failure
narrowed. Observed values are owner float64 values as scored after the freeze, at the scoring record's precision; the evidence
class of each object is given above. In the list, $F^*_r$, $D_r$, $G^\lambda_r$ and $F^{\mathrm{NA}}_r$ are as above; the survival planner is the survival-optimal
count rule with relabelling within groups, and the tilt-blind planner is the balanced slice's survival-optimal rule with
relabelling over all coordinates.

\emph{Scoring scopes.} Rows~1 and~7 are scored on all $36$ held-out cells, row~2 on the $24$ held-out cells with $r\ge3$, rows~3 and~8 on the six held-out equal splits at $r=4$ with $\lambda=2$ or $z\in\{1,2\}$, and rows~5 and~6 on the held-out cells they name. Row~4 is scored on the eight equal splits with $\lambda=1$, $n\in\{12,16,20,24\}$ and $r=3,4$, none of them held-out; its two passes are the $n=12$ cells, the only ones whose $F^{\mathrm{NA}}_r$ had been computed before the freeze.
\begin{center}\small\setlength{\tabcolsep}{5pt}
\begin{tabular}{@{}l>{\raggedright\arraybackslash}p{0.16\textwidth}>{\raggedright\arraybackslash}p{0.2\textwidth}>{\raggedright\arraybackslash}p{0.19\textwidth}l>{\raggedright\arraybackslash}p{0.2\textwidth}@{}}
\toprule
& quantity & frozen criterion & observed & verdict & consequence\\
\midrule
1 & $-\ln G^\lambda_r/F^*_r$ & $\ge0.8$ where $F^*_r\ge0.01$ & min $0.9257$ (36 held-out cells) & pass, 36/36 & ---\\
2 & $D_r/F^*_r$ & $\ge2$ at $r\ge3$ & $1.846$--$1.991$ in four $r=3$ cells & fail, 20/24 & Appendix~\ref{app:alt} states measured ranges instead\\
3 & $F^{\mathrm{NA}}_r/F^*_r$ & $\ge1.15$ at $r=4$, equal splits, $\lambda\ge2$ & $1.255$--$1.956$ (6 cells) & pass, 6/6 & ---\\
4 & $F^{\mathrm{NA}}_r/F^*_r$, $\lambda=1$ & $\le1.01$ wherever computed & $1.0$ at $n=12$ (discovery); $1.0199$--$1.0908$ at $n=16,20,24$ & fail, 2/8 & no claim that predrawn mixtures attain $F^*_r$ at $\lambda=1$\\
5 & $F^{\mathrm{NA}}_4/F^*_4$, $\lambda=2$ & $\ge1.34$ at $n=20,24$ ($n=12$: $1.49$) & $1.311$, $1.255$ & fail, 0/2 & stated above: at $r=4$ and $\lambda\ne1$ the adaptive gain shrinks with $n$ in the cells computed\\
6 & $F^{\mathrm{NA}}_4/F^*_4$, Grid~Z, $z=1$ & below its $n=12$ value $1.96$ at $n=24$ & $1.302$ & pass & ---\\
7 & survival planner$/F^*_r$ & $\le1.05$ in every cell & max $1.0326$ (36 cells) & pass, 36/36 & ---\\
8 & tilt-blind planner$/F^*_r$ & $\ge1.2$ at $r=4$, equal splits, $\lambda\ge2$ & $1.269$--$2.431$ (6 cells) & pass, 6/6 & ---\\
\bottomrule
\end{tabular}
\end{center}
\noindent The two full-support predictions were frozen on 2026-09-29, before the runs they scored; both concern a fitted
growth exponent of best-found profiles for other Beta priors, Beta$(\alpha,\alpha)$, not a proved quantity, and no rate is claimed
for priors other than Beta$(1,1)$, the only case that Appendix~\ref{app:thmm} proves.
\begin{center}\small
\begin{tabular}{@{}l>{\raggedright\arraybackslash}p{0.2\textwidth}>{\raggedright\arraybackslash}p{0.2\textwidth}>{\raggedright\arraybackslash}p{0.17\textwidth}>{\raggedright\arraybackslash}p{0.11\textwidth}>{\raggedright\arraybackslash}p{0.12\textwidth}@{}}
\toprule
& quantity & frozen criterion & observed & verdict & consequence\\
\midrule
9 & local slope of $\log m_{R-1}$ against $\log n$ between the two largest $n$; $m_{R-1}$: coordinates committed before the last round of the best-found single relabelled profile with $R$ rounds (least $F$; not certified); Beta$(\alpha,\alpha)$, $R\in\{3,4\}$, $n\in\{2^8,\dots,2^{14}\}$ & $\alpha=\frac12$: $\ge0.58$ at both $R$; $\alpha=2$: $\le0.56$ at both $R$; difference $\ge0.06$ at each $R$ & $\alpha=\frac12$: $0.521$, $0.555$; $\alpha=2$: $0.431$, $0.384$ ($R=3$, $4$) & fail ($\alpha=\frac12$ below $0.58$) & no rate for other Beta priors\\
10 & mean of the two local slopes of $\log m_{R-1}$ against $\log n$, $n\in\{2^{10},2^{12},2^{14}\}$, $R\in\{3,4\}$ & $\alpha=3$: $\le0.36$ ($R=3$), $\le0.40$ ($R=4$); $\alpha=0.25$: in $[0.50,0.68]$ ($R=3$), $[0.57,0.75]$ ($R=4$) & $\alpha=3$: $0.364$, $0.327$; $\alpha=0.25$: $0.576$, $0.635$ ($R=3$, $4$) & fail ($\alpha=3$, $R=3$: $0.364>0.36$) & as row 9\\
\bottomrule
\end{tabular}
\end{center}

\begin{table}[t]
\caption{Finite comparison at $\varepsilon=\ln2$ (finite computation; details in Appendix~\ref{app:exhibits}). $\LH,\LF$: mean-cost optima of the same class
(two-length linear programs; $\LF$ bracketed below by the lower program over the computed $G_r$ and above by best-found relabelled profiles; at $2^{16}$, where the program's $d^*_r$ were not computed, the $\LH$ entry brackets the optimum between the kernel-checked lower bound and the length $A$ of one schedule);
$A$: least length of one fixed schedule with KL $\le\ln2$ (computed optimum: float64 evaluation of an exact dynamic program) and the shape share of its
divergence; $R_F^{\mathrm{PROF}}$: least length of one relabelled profile with $F\le\ln2$ (exact at $2^4$; at $2^8$ and $2^{10}$ a computed optimum, float64 with margins of at least $0.02$ from $\frac12$; best-found upper
bound at $2^{12},2^{16}$; $\ge4$ at $2^{11}$, since $G_3(2048,1024)=0.3145<\frac12$ in float64 bounds the success of every relabelled profile with at most three rounds); TV: total-variation distance between the target and the output law of the $\delta=\frac12$ ladder of Theorem~\ref{thm:sep}(b)'s proof with the Hoeffding--Ville choice $K=\lceil2\ln(2/\delta)\rceil=3$ (the stated bound takes the smaller $K_\delta=2$ of the Kolmogorov branch; both ladders fail with probability at most $\delta$), whose bulk blocks are $(160,72)$, $(256,672,72)$, $(3328,672,72)$ and $(40960,23808,672,72)$ at $2^8$, $2^{10}$, $2^{12}$ and $2^{16}$, each followed by $24$ singletons; fixed labelling versus relabelled ($=1-s$, with $s$ the ladder's success probability; exact at $2^8$--$2^{12}$, within $2\cdot10^{-57}$ at $2^{16}$); the fixed-labelling TV is a $30$-digit enumeration at $2^8$--$2^{12}$ (pruned mass below $10^{-31}$) and at $2^{16}$ a seeded Monte Carlo estimate of $\E_P[(1-Q/P)_+]$ over $400{,}000$ target draws (standard error $0.0004$). Optima, shares and TV values are rounded to nearest, bounds outward; --: not computed.}\label{tab:finite}
\vspace{4pt}\centering\footnotesize
\setlength{\tabcolsep}{3.5pt}
\begin{tabular}{@{}r r r r r r r@{}}
\toprule
$n$ & $\LH$ & $\LF$ & $A$ & shape & $R_F^{\mathrm{PROF}}$ & TV fixed / mixed\\
\midrule
$2^4$ & 2.66 & 1.999 & -- & -- & 2 & --\\
$2^8$ & 8.19 & 3.28--3.39 & 9 & 89\% & 4 & 0.412 / 0.005\\
$2^{10}$ & 12.56 & 3.71--3.86 & 13 & 95\% & 4 & 0.574 / 0.014\\
$2^{11}$ & 15.14 & 3.92--4.04 & -- & -- & $\ge4$ & --\\
$2^{12}$ & 18.06 & -- & 19 & 98\% & $\le5$ & 0.673 / 0.015\\
$2^{16}$ & 26.20--33 & 1.99--4.69 & 33 & 99\% & $\le5$ & 0.831 / 0.022\\
\bottomrule
\end{tabular}
\end{table}

\paragraph{Mean costs as two-length linear programs.} For a finite predrawn mixture $\mu$ of labelled schedules,
$H(\mu)=\E_\mu\KL(p\|q_U)$ and the scheduled mean $C(\mu)=\E_\mu|U|$ are averages over the components, and $\KL(p\|q_U)$
depends only on the profile of $U$ (Lemma~\ref{lem:tools}(d)). Replacing each component by a least-divergence profile with
the same number of blocks keeps $C(\mu)$ and does not increase $H(\mu)$, so
\[
\LH(n,\varepsilon)=\min\Bigl\{\textstyle\sum_r w_r\,r:\ \sum_r w_r\,d^*_r\le\varepsilon,\ \sum_r w_r=1,\ w\ge0\Bigr\},
\]
with $d^*_r$ the least $\KL(p\|q_U)$ over schedules with $r$ blocks; with one equality and one inequality
constraint, an optimal $w$ uses at most two lengths. For $F$, Lemma~\ref{lem:tools}(a) gives $F(\mu)\ge-\ln Q_\mu(V)$, and
$Q_\mu(V)=\E_\mu q_U(V)\le\E_\mu G_{|U|}(n,k)$ (Lemma~\ref{lem:surv}(a)), so $F(\mu)\le\varepsilon$ forces
$\E_\mu G_{|U|}(n,k)\ge e^{-\varepsilon}$ and
\[
\LF(n,\varepsilon)\ge\min\Bigl\{\textstyle\sum_r w_r\,r:\ \sum_r w_r\,G_r(n,k)\ge e^{-\varepsilon},\ \sum_r w_r=1,\ w\ge0\Bigr\},
\]
with $G_r\le1$ used where $G_r$ is not computed. Conversely, a mixture of uniformly relabelled profiles is
relabelling-invariant, so its $F$ is $-\ln\sum_r w_r s_r$, with $s_r$ the profiles' success probabilities
(Lemma~\ref{lem:tools}(e)); the same program with the successes of explicit profiles in place of $G_r$ is attained by
such a mixture and bounds $\LF$ from above. At $n=2^{16}$ the lower program mixes one round with the bound $G_3\le1$ and
gives $1.996$; the upper one mixes the one-round schedule with the best-found six-block profile $(64717,749,56,10,3,1)$, of success $0.678$, and gives $4.681$. The upper ends at $2^8$, $2^{10}$ and $2^{11}$ mix, in the same way, the best-found profiles $(233,20,3)$ and $(227,23,5,1)$, the one-round schedule and $(961,53,8,2)$, and the one-round schedule and $(1937,92,14,4,1)$. Both optima
have the shape of the dilution in Corollary~\ref{cor:t2}(c): a cheap one-round branch that usually fails, mixed with a longer one.

\paragraph{Balanced slice.} In Table~\ref{tab:finite} the lower end of the $\LF$ bracket is the value of the lower program above over the computed adaptive success probabilities $G_r$, using $G_r\le1$ where it is not computed (at $2^{16}$ only $G_1$ and $G_2$ are computed). At $n=2^{16}$ the best-found five-round relabelled profile, $(64881,608,40,6,1)$, has success $0.522$ and $F=0.650<\ln2$, which is the $R_F^{\mathrm{PROF}}\le5$ entry: a non-exhaustive float64 search found it, the implementation that found it computed its success in exact rational arithmetic, and a second implementation, written from the definitions, reproduces the success in float64 with a bounded truncation error; it is not kernel-checked, and its minimality is not shown. Computed optima and brackets in the same class and at the same cost give $\LH/\LF\approx1.3$ at $n=16$ and about $3.8$ at $n=2048$ (finite values, not rates or a monotone trend).

\paragraph{Full support.} At $\varepsilon=0.1$ the shortest best-found single profiles with $F\le\varepsilon$ have $5$--$9$ rounds (the $\LF$ bounds of Table~\ref{tab:fs}, rounded up to integers); these lengths do not by themselves separate $\log\log n$ from $\sqrt{\log n}$ or a small power, and the good profiles are \emph{increasing}, the reverse of the quota ladder. For $\alpha=\beta=1$, the single-profile rate $\log_2\log_2n+O_\varepsilon(1)$ and the mixture rate $\LF=(e^{-\varepsilon}/2+o(1))\log_2\log_2n$ are proved in Appendix~\ref{app:thmm}, with astronomical constants; the finite mixing gains below are at fixed round caps. For $n\in\{8,16,32,64\}$ and $r\le6$, every exhaustively optimal single relabelled profile is
non-decreasing, and for $2\le r\le6$ the reversed optimal profile has $4.2$--$148$ times its $F$ (extremes at $n=8$, $r=2$
and $n=64$, $r=6$) and the path-optimal profile $1.00$--$1.76$ times, and mixing profiles or adapting block sizes gains at most $3\%$ (floating-point conditional-gradient duality gaps
$\le3.1\cdot10^{-6}$); for the symmetric priors Beta$(\alpha,\alpha)$ with $\alpha\in\{0.02,0.05,0.1,0.5,1\}$ at
$(n,r)\in\{(10,2),(32,2),(32,3),(32,4),(64,2),(64,3)\}$ the gain reaches $50\%$, at the U-shaped $\alpha=0.02$, $n=10$, $r=2$ (floating-point duality brackets with
gaps $\le3.1\cdot10^{-5}$; a second implementation agrees on the $14$ cells it covers, including the $50\%$ cell). Near $\varepsilon=0.1$, mixing profiles or adapting block sizes gains at most $5.2\%$ at $n=256$ and $8.5\%$ at $n=512$ (caps $5$ and $6$, where the best-found single profiles are
$(1,3,8,31,213)$ and $(1,2,5,12,38,198)$ at $n=256$ and $(2,4,12,51,443)$ and $(1,2,5,15,57,432)$ at $n=512$) over the best
single profile; these near-$\varepsilon$ bounds are floating-point duality bounds from one implementation, with duality interval width $\le10^{-5}$. At a fixed small number of rounds the gain can be larger: mixing profiles gains $27.9\%$ at $n=1024$ with two rounds
(float64 duality bounds). Close agreement of the first four moments need not make $F$ close: at $\alpha=\beta=1$ and $n=4096$, four post-selected
pairs of profiles with equal round counts differ in $F$ by factors $5.5$--$33.4$ (float64): for $m\in\{16,32,49,64\}$, the profile
of $m$ unit rounds and one block of $n-m$ against, respectively, $(1^{15},287,3794)$, $(1^{31},1014,3051)$,
$(1^{42},393,422,451,484,518,555,594,637)$ and $(1^{47},218,220,220,220,222,222,223,224,225,225,226,227,227,229,229,230,230,232)$,
where $1^k$ stands for $k$ unit rounds. In each pair the means and third central moments coincide by the $0\leftrightarrow1$
symmetry of the target, and the variances and fourth central moments agree only approximately (relative gaps
$\le7.7\cdot10^{-6}$ and $\le1.2\cdot10^{-3}$); the pairs were chosen by moment agreement within the same search that
computes $F$. Over all relabelled profiles with $2$--$5$ rounds at $n=32$ and $2$--$4$ at $n=48$, profiles whose output count laws have
log-variances within $10^{-3}$ differ in $F$ by at most $5.5\%$ for $\alpha=\beta\in\{16,64\}$ and by at most $17\%$ for
$(\alpha,\beta)=(2,30)$; for $\alpha=\beta\in\{0.2,1,4\}$ the largest such ratio at each prior and size is between
$1.335$ and $16.77$ (float64). Two frozen predictions of fitted growth exponents for other Beta
priors failed. Table~\ref{tab:fs} is a float64 computation, not interval-certified, except its $\LF$ bounds at $n\le2^{10}$, which a separate exact evaluation with interval arithmetic certifies as upper bounds.

\begin{table}[t]
\caption{Full-support target Beta$(1,1)$--Bernoulli: computed $\LH$ optima (float64 evaluation of the exact optimization) and best-found upper bounds on $\LF$ (float64; interval-certified only where stated). Each $\LF$ bound mixes the best-found profiles with $R$ and $R+1$ rounds whose $F$ values bracket $\varepsilon$, weighted so that the average $F$ is $\varepsilon$; as $\KL(p\|\cdot)$ is convex, the mixture has $F\le\varepsilon$, and the $(R+1)$-round profile alone has $F\le\varepsilon$, so each bound, rounded up to an integer, is the length of the shortest best-found single profile with $F\le\varepsilon$. All $\LF$ entries come from one evaluation pipeline applied to the candidate profiles of two independent optimizers; at $n\le2^{10}$ a separate implementation, which recomputes each profile's output count law exactly and encloses $F$ in interval arithmetic, certifies these $\LF$ entries as upper bounds; a second implementation reproduces $\LH$ at $n=2^8$--$2^{12}$ ($\varepsilon=0.1$). $\LH$ rounded down, upper bounds up, ratios (computed from the unrounded values) down.}\label{tab:fs}
\vspace{4pt}\centering\footnotesize
\begin{tabular}{@{}r r r r r r r@{}}
\toprule
 & \multicolumn{3}{c}{$\varepsilon=0.1$} & \multicolumn{3}{c}{$\varepsilon=\ln2$}\\
\cmidrule(lr){2-4}\cmidrule(l){5-7}
$n$ & $\LH$ & $\LF\le$ & ratio$\ge$ & $\LH$ & $\LF\le$ & ratio$\ge$\\
\midrule
$2^6$ & 14.55 & 4.36 & 3.3 & 4.35 & 2.29 & 1.9\\
$2^8$ & 29.48 & 5.59 & 5.2 & 8.03 & 2.91 & 2.7\\
$2^{10}$ & 51.61 & 6.50 & 7.9 & 13.10 & 3.43 & 3.8\\
$2^{12}$ & 81.45 & 7.12 & 11.4 & 19.29 & 3.82 & 5.0\\
$2^{14}$ & 119.34 & 7.68 & 15.5 & 26.75 & 4.04 & 6.6\\
$2^{15}$ & 141.37 & 7.86 & 17.9 & 30.93 & 4.25 & 7.2\\
$2^{16}$ & 165.50 & 8.02 & 20.6 & 35.44 & 4.42 & 8.0\\
\bottomrule
\end{tabular}
\end{table}
\pend{}

\begin{table}[h]
\caption{Balanced slice, $n=65536$, $\varepsilon=\ln 2$. K: kernel-checked; Fin: finite computation, of two kinds: $\LH\le33$ is one schedule found by a float64 dynamic program, its KL recomputed at $60$ digits; the $\LF$ bracket comes from exact-rational linear programs over inputs selected by a float64 search. The lower bound on $\LH$ uses the three-term form of Theorem~\ref{thm:sep}(a) (the relaxed form gives $23.02$); the $\LF\le11$ certificate relabels one explicit profile (numbers of unrevealed coordinates $65536\to1735\to201\to50\to19\to9\to5$, then five singletons). The kernel check bounds that profile's failure probability by the second-moment budget $\sum_tb_t/m_t^2$ over its shrinking blocks ($m_t$ the coordinates left after block $t$; the singleton tail adds nothing), here $0.4889<\frac12$, so its success exceeds $\frac12$ and $F<\ln2$. The five-round profile behind the finite computation $F=0.650$ has a budget above $7$, which is why the certificate uses eleven rounds.}\label{tab:inst}
\vspace{4pt}\centering\footnotesize
\begin{tabular}{@{}l l l@{}}
\toprule
quantity & value & \\
\midrule
$\LH$ & $>26.20$ & K\\
$\LH$ & $\le33$ (one schedule) & Fin\\
$\LF$ & $\le11$ (one relabelled profile) & K\\
$\LH/\LF$ & $>2$ & K\\
$\LF$ (finite brackets) & $[1.996,\,4.681]$ & Fin\\
\bottomrule
\end{tabular}
\end{table}
\begin{figure}[h]
\centering\includegraphics{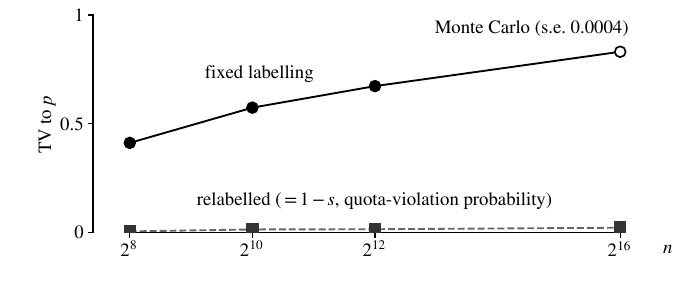}
\caption{The mechanism at finite $n$ ($\varepsilon=\ln2$; finite computation, Table~\ref{tab:finite}): total variation to $p$ of the
$K=\lceil2\ln(2/\delta)\rceil=3$ ladder of Theorem~\ref{thm:sep}(b)'s proof ($\delta=\frac12$; the Hoeffding--Ville choice, where the stated bound takes $K_\delta=2$; Table~\ref{tab:finite} lists its blocks) under the fixed
labelling (circles) and under its relabelling (squares; $=1-s$, the quota-violation probability). Filled: computed; open: a seeded Monte Carlo
estimate (standard error $0.0004$). The least lengths of the same table are plotted in Figure~\ref{fig:finite}.}\label{fig:mech}
\end{figure}

\section{Reproducibility}\label{app:repro}
\emph{Software, licenses and compute.} The formal proofs use Lean~4 \citep{moura2021lean4} and Mathlib \citep{mathlib2020}
(Apache License~2.0); the learned heads use PyTorch \citep{paszke2019pytorch}; the numerical scripts also use Arb
\citep{johansson2017arb} through python-flint, and NumPy, SciPy, SymPy, mpmath and Matplotlib. PyTorch, NumPy, SciPy, SymPy and
mpmath are released under the BSD 3-Clause license, Matplotlib under its PSF-based license, FLINT, which contains Arb, under the
LGPL (version 3 or later), and python-flint under the MIT license. No external data or pretrained models are used; all data are
synthetic draws from the stated targets. The learned heads used NVIDIA H100 80GB GPUs, with several jobs sharing a GPU, about
25 job-hours in total (Appendix~\ref{app:controlled}); the kernel checks and the other finite computations use CPUs only, and
their run times are not reported.

\begin{figure}[h]
\centering\includegraphics{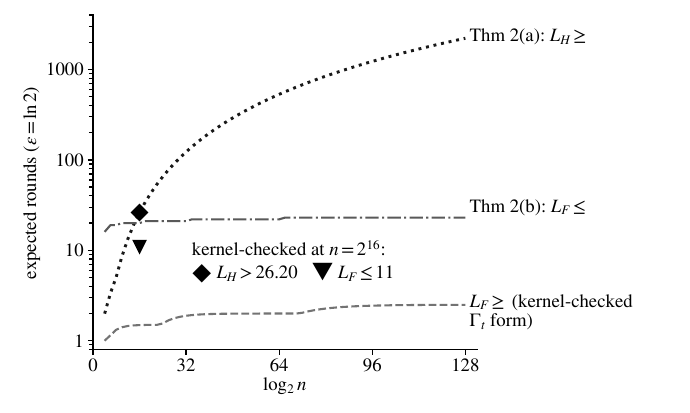}
\caption{Proved bounds as functions of $n$ at $\varepsilon=\ln2$ and the kernel-checked finite certificates at
$n=2^{16}$. Curves are formula evaluations, not measurements.}\label{fig:bounds}
\end{figure}

This appendix evaluates the general balanced-slice bounds at finite sizes, as cited in Section~\ref{sec:finite}, and names the scripts behind Table~\ref{tab:bounds} and Figures~\ref{fig:hf}, \ref{fig:finite}, \ref{fig:grids}, \ref{fig:sampler}, \ref{fig:mech}, \ref{fig:bounds} and~\ref{fig:controlled}.
\begin{table}[h]
\caption{Proved bounds versus $n$ ($\varepsilon=\ln2$). Formula evaluations; lower bounds rounded down, upper bounds up (script hash below). The second and third columns evaluate Theorem~\ref{thm:sep}(a) and~(b), and the two right-hand columns the kernel-checked envelope forms (Appendix~\ref{app:lean}), with $\Gamma_t(n)$ of Appendix~\ref{app:env}.}\label{tab:bounds}
\vspace{4pt}\centering\footnotesize
\begin{tabular}{@{}rrrrr@{}}
\toprule
$\log_2n$ & $\LH\ge$ & $\LF\le$ & $\LF\ge$ & $R_F^{\mathrm{adapt}}\ge$\\
\midrule
8 & 5.09 & 19 & 1.32 & 2\\
16 & 26.20 & 20 & 1.48 & 2\\
32 & 122.43 & 21 & 1.88 & 3\\
64 & 530.82 & 22 & 1.99 & 4\\
128 & 2211.37 & 23 & 2.49 & 4\\
1024 & $1.46\cdot10^5$ & 26 & 3.49 & 7\\
\bottomrule
\end{tabular}
\end{table}

With $\Gamma_t(n)=\min\{1,\,2\cdot6^{t-1}(k+1)^{-2^{-t}}\}$, the $\kappa_r$ form of the survival envelope of Appendix~\ref{app:env} (prefactor $\kappa_t=2\cdot6^{t-1}$ in place of the bounded $A_t<36$ of \eqref{eq:env}), $G_t(n,k)\le\Gamma_t(n)$. Since $R_F^{\mathrm{adapt}}(n,\varepsilon)$ is the least $t$ with $G_t(n,k)\ge e^{-\varepsilon}$ (Corollary~\ref{cor:t2}(a)),
$R_F^{\mathrm{adapt}}(n,\varepsilon)\ge\min\{t:\Gamma_t(n)\ge e^{-\varepsilon}\}$; and since $\LF(n,\varepsilon)\ge1+t\,(e^{-\varepsilon}-G_t(n,k))$ for every $t\ge1$ (Appendix~\ref{app:conv}, the single-horizon bound),
$\LF(n,\varepsilon)\ge\max_{t\ge1}\bigl\{1+t\,[e^{-\varepsilon}-\Gamma_t(n)]_+\bigr\}$, the bound being kernel-checked for each $t$;
these are the two right-hand columns of Table~\ref{tab:bounds}.

Table~\ref{tab:bounds} and Figure~\ref{fig:bounds} show where the general bounds are informative: the proved upper bound on $\LF$ is below the proved lower bound on $\LH$ for every even $n\ge2^{15}$ at $\varepsilon=\ln2$ and every even $n\ge2^{13}$ at $\varepsilon=0.1$ (kernel-checked); the two formulas first cross at even $n=19{,}286$ and $n=4{,}788$, so $2^{15}$ and $2^{13}$ are the first powers of two past the crossing. The lower bounds on $\LF$ and $R_F^{\mathrm{adapt}}$ (the terminal lower bounds) are of the right order but weak at every size shown, because the prefactor
of the survival envelope dominates: the weaker $\kappa_r$ form $\Gamma_r(n)<1$ of Appendix~\ref{app:env} requires $n>4.1\cdot10^4$ for $r=2$, $n>1.4\cdot10^{15}$
for $r=3$ and $n>2.9\cdot10^{42}$ for $r=4$, and even the bounded prefactor of \eqref{eq:env} needs $n>2590$, $n>4.3\cdot10^9$ and $n>1.2\cdot10^{22}$.

Scripts: Table~\ref{tab:bounds} and Figure~\ref{fig:bounds} come from \texttt{bounds\_v2.py} (sha256 c00a4cae\ldots; output 11065153\ldots); Figure~\ref{fig:bounds} is drawn by \texttt{fig1\_bounds\_alt.py} (64ca16b9\ldots). Figure~\ref{fig:mech} plots the total-variation values of Table~\ref{tab:finite}, listed in \texttt{make\_fig\_mech.py} (4c0b80d7\ldots) and drawn by \texttt{make\_fig\_mech\_alt.py} (76798c75\ldots); the least lengths of that table are re-checked by \texttt{fig\_lengths\_compute.py} (e79357cf\ldots). Figures~\ref{fig:hf} and~\ref{fig:grids} come from \texttt{fig\_a\_compute.py} (3c725a52\ldots) and \texttt{fig\_b\_compute.py} (98a0ef67\ldots), in exact rational arithmetic, read by \texttt{fig\_a\_plot.py} (9a173461\ldots) and \texttt{fig\_b\_plot.py} (2be95ef0\ldots) and drawn by \texttt{fig\_a\_plot\_alt.py} (9e00ebe8\ldots) and \texttt{fig\_b\_plot\_alt.py} (18cc56f0\ldots); Figure~\ref{fig:sampler} is a schematic with no data, drawn by \texttt{fig\_sampler\_ref\_alt.py} (f1236ad8\ldots); Figure~\ref{fig:finite} comes from \texttt{fig\_finite\_sep.py} (aad2b2dc\ldots), which evaluates the two bounds of Theorem~\ref{thm:sep} and reads the re-checked least lengths of Table~\ref{tab:finite} and the kernel-checked bounds of Table~\ref{tab:inst}, and is drawn by \texttt{fig\_finite\_sep\_alt.py} (ce2607a5\ldots); and Figure~\ref{fig:controlled} is drawn from the two exported tables of the first frozen test by \texttt{make\_exhibit\_alt.py} (6634f972\ldots).\pend{}

\section{Controlled learned-conditional test}\label{app:controlled}
Do exact-oracle rankings of schedule profiles survive learned conditionals? On small masked transformers trained on
known targets ($n\le16$), not language models, we tested exact-oracle rankings fixed in advance against the
profiles' complete-law output KL (float64 dynamic program) under fresh heads: eight targets in the main test and four held-out Beta--Bernoulli targets
in a second test. The main test's menu entries (exact-oracle optima over size profiles, other relabelled profiles, fixed orders, and for quota targets a value-adaptive policy) are not the asymptotic ladders, and its predictions held
under their frozen eligibility screens; the second test's held in every tested cell, and a later two-group test passed
three items and, as written, failed a fourth, a control that named the wrong comparator. All passes are sample-level: with
six independent clusters per target, even a $100\%$ pass rate bounds a population pass probability from below only by
$0.61$ (one-sided $95\%$). On the main test's extended menu the terminal-optimal profile's regret exceeded $0.005$ nats in
$14$ of $486$ pairs. Frozen predictions of exact-oracle finite-size quantities, which involve no learned heads, fared worse: of eight two-group predictions five passed and three failed, and both of the two
frozen predictions of fitted growth exponents for Beta priors other than Beta$(1,1)$ failed; Appendix~\ref{app:exhibits} lists all ten, and the statements that the two-group failures narrowed. Exploratory target-aware plug-in rules using the
heads' own conditionals chose the exact-oracle rule's profile in $5{,}559$ of $5{,}760$ replicate rows (not independent)
on the three Beta--Bernoulli targets and the two-group target, so we show no decision value over them, nor, on all eight targets, over direct validation (below), and no rate
or robustness theorem under learned error. On the four uniform quota targets, restarting until a valid string appears, a step outside
the interface, gives much lower output KL at comparable expected rounds (measured, not scored), so nothing here
recommends the restricted interface for practice. This appendix gives the protocol and\ the reproduction checks that were run.

\begin{figure}[t]
\centering\includegraphics{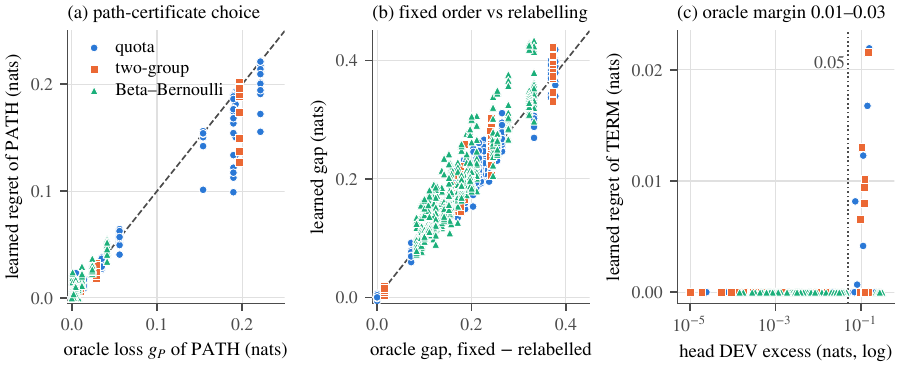}
\caption{Controlled test with learned conditionals (144 fresh trajectories, 48 independent clusters, six held-out
regimes and fresh seeds on two discovery targets; output KL of the complete output law, computed in float64 by dynamic programming over revealed sets (no sampling), $n\le16$; predictions frozen before evaluation).
(a) Learned regret of the path-optimal profile PATH versus its oracle loss $g_P$ (identity dashed).
(b) Learned versus oracle gap between a fixed labelled order and its relabelling. (c) Learned regret of the exact-oracle
terminal-optimal profile TERM against the head's development-loss excess (``DEV excess''), for cells with oracle margin $0.01$--$0.03$ and the
EARLY, ES and END checkpoints of each run (descriptive, not a frozen pass/fail item; dotted line at $0.05$, the top boundary of the freeze's descriptive head-error bands $\le0.02$, $0.02$--$0.05$ and $>0.05$; DEV excess, which can be negative, is drawn clipped below at $10^{-5}$). Controlled prediction only.}
\label{fig:controlled}
\end{figure}

Heads: masked transformers (PyTorch, \citealp{paszke2019pytorch}; two pre-LN layers, width 64, four attention heads,
feed-forward width 256, GELU, no dropout) trained on $N\in\{256,1024,8192\}$ draws with AdamW (learning rate
$2\cdot10^{-3}$, weight decay $10^{-4}$, cosine decay to $5\%$), batch 512 and 4000 steps. Each training example reveals
a uniformly random set whose size is uniform on $\{0,\dots,n-1\}$, and the loss is the mean binary cross-entropy over
the unrevealed coordinates. The checkpoint used for the frozen predictions, called ES, is the one with the lowest masked cross-entropy on 2048 separate development
draws, each with four fixed masks, among the evaluations at steps $0,50,100,\dots,4000$; it reads
no evaluator quantity (target, KL or law). Each run also saved the checkpoints at step 300 (EARLY) and step 4000 (END); these two enter only Figure~\ref{fig:controlled}(c), whose EARLY and END points come from the evaluation rows of those checkpoints. The same settings were used for every head. Heads are evaluated in float64
with PyTorch's fused-attention fast path disabled. Compute: NVIDIA H100 80GB GPUs, with several jobs sharing a GPU; about 25 job-hours in total
for training and evaluation (job logs), 9.5 of them in the rows' law-evaluation timing fields (complete-law DP, plus the path criterion where computed); menu construction and the two-group dynamic programs ran on CPUs.
Targets: uniform quota (uniform on the strings with a given number of ones; ``quota'' below), given as (length, number of ones), $(14,7),(16,5),(10,5),(12,6)$; Beta--Bernoulli, given as (length; Beta parameters), $(12;\tfrac12,2),(14;1,1),(12;1,1)$; two groups ($4+8$
coordinates, $k=6$, tilt $3$: on the strings with $k$ ones, probability proportional to $3^{|x_A|}$, where $|x_A|$ is the number of ones among the four coordinates of group~A; Appendix~\ref{app:tgenv}); quota $(12,6)$ and Beta--Bernoulli $(12;1,1)$ are discovery targets, on which the thresholds were calibrated with other seeds, and the other six are held out. Menu per target and $r$ ($r=2,3,4$ for every target, and also $r=5$ for quota $(14,7)$ and $(16,5)$ and Beta--Bernoulli $(14;1,1)$): TERM (exact-oracle terminal optimum: the size profile of least exact-oracle output KL), PATH (path-certificate optimum, minimizing the exact-oracle schedule-averaged KL $H$, an upper bound on the exact-oracle output KL), uniform (the most even profile with $r$ blocks), TERM reversed, a halving ladder (each block half of the coordinates still to commit, rounded up but leaving at least one for each later block; the last block takes the rest), the other family's optimum (a misspecified-family control: the exact-oracle terminal optimum, at the same $n$ and $r$, of the quota target with $n/2$ ones for Beta--Bernoulli targets, and of Beta$(1,1)$--Bernoulli for the quota and two-group targets), fixed-order versions of TERM, PATH and uniform, and for quota targets the value-adaptive optimum (the maximizing count-state rule of Theorem~\ref{thm:conv}(a), run in a uniformly random order; under the exact oracle its output KL is $-\ln G_r(n,k)$, the least over $\mathcal A_r$). Endpoint: the complete-law $\KL(p\|Q_{\theta,\mathrm{law}})$ (float64 dynamic program), the output KL of a design under the learned head. Regret is measured against the best, under the learned head, of the five relabelled profiles TERM, PATH, uniform, TERM reversed and the halving ladder (the frozen scorer's predrawn menu). The \emph{oracle margin} of a target and $r$ is the exact-oracle output KL of the best other menu entry with a different size profile minus that of TERM. The \emph{development-loss excess} is the head's masked cross-entropy on its $2048$ development draws (four fixed masks each) minus that of the exact conditionals on the same draws and masks. The first frozen test has the items P1--P5 of the table below. P1's units are the (trajectory, $r$) pairs that pass the \emph{screen}: margin $\ge0.01$ and excess $\le0.02$. PATH's \emph{oracle loss} $g_P$ at a target and $r$ is the exact-oracle output KL of PATH minus that of TERM; the \emph{relabelling gain} $G_o$ of a profile (the oracle gap of Figure~\ref{fig:controlled}(b)) is the exact-oracle output KL of its fixed labelled order minus that of its uniform relabelling. The head's \emph{evidence-bound choice} (P2b) is the menu profile minimizing the learned head's own schedule-averaged KL $H$, computed in float64 without sampling. TERM and PATH are exact-oracle optima over size profiles, the value-adaptive optimum is a policy, the fixed-order versions are fixed labelled schedules, and the other entries are relabelled profiles; none is an asymptotic ladder of Section~\ref{sec:main}. Selectors were given the family and its parameters, which determine the target, so these are not target-free selections.
\begin{center}\footnotesize\setlength{\tabcolsep}{4pt}
\begin{tabular}{@{}l l r@{}}
\toprule
item & frozen prediction (units; threshold) & result\\
\midrule
P1 & TERM regret $\le0.002$ (oracle margin $\ge0.01$, dev.\ excess $\le0.02$; $\ge95\%$) & 342/342\\
P2a & PATH regret $\ge\frac12g_P$ (cells $g_P\ge0.02$; $\ge90\%$) & 180/180\\
P2b & head's evidence-bound choice regret $\ge\frac12g_P$ ($\ge90\%$) & 180/180\\
P3a & fixed order loses $\ge\frac12G_o$ (triples over TERM, PATH, uniform; $G_o\ge0.05$; $\ge95\%$) & 1386/1386\\
P3b & quota: $|\Delta\text{validity}|,|\Delta H|\le0.1G_o$ ($\ge90\%$); Beta--Bernoulli: $|\Delta H|\le0.3G_o$ & 689/702, 684/702; 540/540\\
P3c & quota: shape $\ge80\%$ of the learned gap ($\ge90\%$) & 701/702\\
P4 & reversed / other-family profile loses $\ge0.1$ ($\ge90\%$ each) & 486/486 each\\
P5 & quota: value-adaptive optimum $\le$ TERM $+0.005$ ($\ge90\%$) & 242/252\\
\bottomrule
\end{tabular}
\end{center}
The table summarises the first frozen test only. Units, nested in clusters: P1, P2, P4, P5 count (trajectory, $r$) pairs; P3 counts (trajectory, $r$, profile) triples. There are 144 trajectories (training runs, one per target, seed and training size); the three training sizes of one target--seed pair share seed streams, so there are 48 independent clusters. Clusters in which every unit passes: P1 48/48, P2 42/42, P3a 48/48, P4 48/48, P3b 17/24 and 18/24, P3c 23/24, P5 17/24. For the items below 100\%, the pooled unit pass proportion, resampled over the (target, seed) clusters ($20{,}000$ cluster-bootstrap draws), has its 5\% quantile above the item's frozen threshold (P3b $0.970/0.955$, P3c $0.996$, P5 $0.935$). In words: the path-optimal profile and the profile with the
smallest learned-head $H$ kept a learned regret of at least half of PATH's oracle loss $g_P$ in every eligible pair, and a fixed labelled order
lost at least half of the relabelling gain in every eligible triple, while, on quota targets, the log validity rate and
the path divergence moved by at most a tenth of that gain in $97$--$98\%$ of triples. A second frozen test (plan written before training, scorer fixed before any row;
72 fresh trajectories from four held-out Beta--Bernoulli targets, Beta$(\frac12,\frac12)$ and Beta$(4,4)$ at $n=14$ and $16$; 24 independent clusters) scored the variance-matching
rule (which picks the composition whose complete output law under the head (float64 dynamic program) has the variance of the number of ones closest to the target's Beta--Binomial variance (closed form)) against all relabelled compositions with exactly $r$ blocks: on Beta$(\frac12,\frac12)$ it kept at least half of its
oracle regret in $18/18$, $18/18$, $17/18$ heads ($n=14$, $r=2,3,4$) and $18/18$ ($n=16$, $r=2$); on Beta$(4,4)$ its regret
was at most $0.005$ in $18/18$ heads in every tested cell; $n=16$ with $r=3,4$ was not tested ($105$ and $455$ compositions per head, estimated at over $50$ GPU-hours).
A third frozen test (the predictions were frozen before the execution plan was written, and the plan before any value on a
learned head; two groups of $4+8$ coordinates, $k=6$, tilt $3$; the main test's $18$ two-group heads, $6$ independent clusters) scored a
group-aware, value-adaptive survival rule: the two-group survival planner of Appendix~\ref{app:tgred} (TGR5), which, from the rounds left, the free counts of the two groups and the residual quota, commits the block composition $(b_A,b_B)$ that maximizes the exact-oracle probability of meeting the quota within the remaining rounds (ties to the lexicographically first, comparing exact values), commits everything in the last round, and places each group's share uniformly within the group by a predrawn seed. Against a comparator $X$ its \emph{oracle gain} is $F_{\mathrm{exact}}(X)-F_{\mathrm{exact}}(\text{rule})$, and ``by at least half the oracle gain'' means $F_{\mathrm{head}}(X)-F_{\mathrm{head}}(\text{rule})\ge\frac12[F_{\mathrm{exact}}(X)-F_{\mathrm{exact}}(\text{rule})]$. It beat the group-blind terminal-optimal profile by at least half the oracle
gain in $36/36$ (head, $r$) pairs at $r=3,4$ (median gain $0.045$ nats, oracle $0.045$), was the best law of the menu in
$36/36$, and beat the tilt-blind balanced adaptive rule by at least half the oracle gain at $r=4$ in $18/18$ (under the exact oracle that rule has TERM's law at $r=4$: it commits $7$, $3$, $1$ and $1$ coordinates whenever the quota is not already forced, so this item repeats the first item's $r=4$ half, and its head values differ from TERM's by at most $0.0025$). The fourth item, a control at $r=2$, where no adaptivity is possible, named the wrong comparator. As frozen, it required the
rule's output KL under each head to be within $0.005$ of that of TERM. At $r=2$, TERM, the exact-oracle terminal optimum over
size profiles, is group-blind (blocks of $9$ and $3$ coordinates), and the rule beats it by $0.0147$ nats on exact-oracle values
alone, so the item fails by construction (within $0.005$ in $2$ of $18$ heads). The intended comparator was the group-aware
profile that commits $3$ of group~A's and $7$ of group~B's coordinates and then the remaining two; at $r=2$ the rule is exactly
this profile, so the intended comparison is an identity ($18/18$, difference $0$) and checks only that the two evaluations agree. Frozen predictions about exact-oracle finite-size quantities fared worse: of eight two-group predictions five passed and three failed, and both of the
two frozen predictions of fitted growth exponents for Beta priors other than Beta$(1,1)$ failed (Appendix~\ref{app:exhibits}).

Passes are sample-level: with six independent clusters per target, even a $100\%$ pass rate bounds a population success
probability from below only by $0.05^{1/6}\approx0.61$ (one-sided $95\%$), so the pass-rate thresholds are not certified at the
population level. Over all $48$ clusters of the main test the corresponding bound for P1 is $0.05^{1/48}\approx0.94$, and even that assumes a common cluster-success probability; it says nothing about unseen target families. None of these tests is a rate or robustness theorem under learned error;
Appendix~\ref{app:imperfect} gives the exact statements that are proved. In the quota comparisons, which the freeze named as contextual (measured, not scored; not tabulated),
restarting until a valid string appears, a step outside the interface, gives much lower output KL at comparable
expected rounds (medians over the $18$ heads of each target: $0.002$--$0.005$ nats at $3.5$--$4.0$ expected rounds, against $0.068$--$0.177$ nats for TERM with $4$ rounds).

\emph{Plug-in comparison (after the freeze; not a frozen test).} The comparison covers the Beta--Bernoulli and two-group targets: the three Beta--Bernoulli targets and the two-group target. It uses their $72$ fresh ES heads from the main test, in $24$
clusters, with every head and every $r$ included (no screen).
\begin{itemize}
\item \emph{What the rules use.} Like the frozen rule, each rule knows the target: its family, its parameters and, for
  two groups, the group split.
  \begin{itemize}
  \item Each rule averaged the head's conditionals onto count states (group-count states for two groups), over $16$
    random contexts per state. That takes $1{,}248$--$6{,}880$ head queries per estimate; the frozen rule makes none.
  \item It inserted these conditionals into the exact count-state recursion, giving each of the exact oracle's six
    relabelled profiles (P1's five plus the other family's optimum) an estimated output KL to the known target.
  \item The first rule chose the smallest estimated output KL. On the two-group target, a second rule chose the
    largest estimated valid mass from the same estimates.
  \end{itemize}
\item \emph{The rows.} Each head and $r$ was run in $20$ replicates. That gives $5{,}760$ replicate rows: $16$
  (target, $r$, rule) cells, $18$ heads each, $20$ replicates. Because the two-group rules share their estimates, these
  rows hold $4{,}680$ distinct estimates. The rows are not independent.
\item \emph{Agreement.} The rules chose the frozen rule's profile in $5{,}559$ rows: $4{,}479$ of $4{,}680$ for the first
  rule and all $1{,}080$ for the second.
\item \emph{Disagreement.} All $201$ differing rows are Beta--Bernoulli, and in every one of them the plug-in chose the
  path-optimal profile. Under the learned head, that choice had lower output KL than the frozen rule's in $89$ rows,
  by at most $0.0036$ nats, and higher output KL in $112$, by at most $0.022$ nats.
\item \emph{Earlier rivals.} The freeze also named zero-cost comparators and sampling-based validation rivals, as
  quantities to be measured rather than scored. They are reported descriptively and support no decision-value or cost
  claim.
\end{itemize}
The plug-in thus mostly repeats the frozen rule's choice, TERM. Direct validation, which computes each of the six relabelled profiles' complete-law output KL under the learned head and keeps the lowest, improves on TERM by more than $0.005$ nats in only $3$ of the main test's $486$ (trajectory, $r$) pairs, over all eight targets (extended-menu note below). We claim no decision
value for the frozen rule over the plug-in (on its four targets) or over direct validation (on all eight). Neither comparison shows that selection adds
no value on other targets.

\begin{center}\footnotesize
\begin{tabular}{@{}p{0.24\linewidth}p{0.40\linewidth}p{0.30\linewidth}@{}}
\toprule
check (units) & result on these targets & limits and failures\\
\midrule
Main frozen test: $144$ trajectories, $48$ clusters; (trajectory, $r$) pairs, or triples for P3 &
\begin{tabular}[t]{@{}p{\linewidth}@{}}
under the screen, the exact-oracle terminal choice transfers (P1 $342/342$);\\
the path certificate and the head's own $H$ misselect as in the oracle (P2);\\
the relabelling gain survives and is nearly invisible to validity rate and $H$ on quota: at most a tenth of it in
$97$--$98\%$ of triples (P3);\\
TERM reversed and the other family's optimum lose $\ge0.1$ nats (P4);\\
the value-adaptive quota optimum is at most $0.005$ above TERM (P5 $242/252$)
\end{tabular} &
\begin{tabular}[t]{@{}p{\linewidth}@{}}
no population pass rates: the one-sided $95\%$ bound is $0.61$ per target;\\
nothing outside the screen: Figure~\ref{fig:controlled}(c) is descriptive;\\
no robustness theorem under learned error
\end{tabular}\\
Second test: $72$ trajectories, $24$ clusters &
the frozen variance-matching predictions pass in every tested cell (lowest $17/18$; frozen rates $80\%$ for Beta$(\frac12,\frac12)$ and $90\%$ for Beta$(4,4)$) &
$n=16$ with $r=3,4$ not tested\\
Third test: the main test's $18$ two-group heads, $6$ clusters &
the group-aware rule's exact-oracle advantage persists ($36/36$, $36/36$, $18/18$; median gain $0.045$ nats) &
the fourth item, as written, fails: its comparator was wrong\\
Exact-oracle finite-size predictions (no learned heads) &
five of eight two-group predictions passed &
three two-group predictions failed, narrowing the claims, and both frozen growth-exponent predictions failed\\
Plug-in, after the freeze: the main test's $72$ Beta--Bernoulli and two-group heads, $5{,}760$ replicate rows &
the plug-in and the frozen rule, both using the known target, agree in $5{,}559$ rows &
no decision value over the plug-in or, on all eight targets, direct validation; the plug-in excludes uniform quota targets; no cost comparison\\
Restart until valid (outside the interface; named in the freeze, measured but not scored; uniform quota targets) &
lower output KL than the restricted interface at comparable expected rounds &
no recommendation of the restricted interface for practice\\
Overall & --- & no native language-model or deployment benefit (not tested)\\
\bottomrule
\end{tabular}
\end{center}

\emph{Reproduction checks.}
\begin{enumerate}
\item[(i)] \emph{Cached results.} On a CPU, from the cached{} per-head results, the $82$ items of the{} checker
  regenerate with $0$ mismatches, in clean rooms on macOS and Linux. The items are: $61$ printed values, $4$ checks of
  Figure~\ref{fig:controlled}, $9$ source-note values and $8$ consistency checks. The plug-in counts in this appendix
  that are not among them (heads, clusters, cells, replicates, the disagreement split) were recounted from the same
  rows.
\item[(ii)] \emph{Checkpoint inference.} One (ES) checkpoint was kept per head, $216$ in all.{} A subset of $29$ of the $216$
  checkpoints was fixed before any recomputation, and it covers every family and scorer.
  \begin{itemize}
  \item An executor not involved in producing the results re-derived that subset from the weights, on a CPU, with no
    cached result in its isolated input tree.
  \item The re-derived results agree with the cached{} ones to within $1.5\cdot10^{-14}$, and they regenerate the third
    test's printed values, subject to that executor's stated limits: coverage ($25$ of the $144$ fresh and $4$ of the $72$ full-support heads), run observability (no pre-run proof of an empty output folder, no exclusion of a concurrent writer, file times not attested), timing fields untested, and only the recorded Monte Carlo draws checked.
  \item The other $187$ checkpoints are tied to their results by file hashes only.
  \end{itemize}
\item[(iii)] \emph{Training.} Retraining{} needs a CUDA GPU, as the original training did. It was not
  rerun, and a rerun would agree only up to training variation.
\item[(iv)] \emph{Portability.} Regeneration of (i) was tested only on CPU, on macOS and Linux with the pinned Python packages, where all $82$ checked
  items match; nothing was rerun on a GPU, and the original evaluations ran in float64, on H100 GPUs except menu construction and the two-group dynamic programs (CPUs).
\end{enumerate}

\emph{Extended menu (main test).} On the extended menu, which adds the other family's optimum and, for quota targets, the value-adaptive optimum, the
terminal-optimal profile's regret exceeded $0.005$ nats in $14$ of the $486$ (trajectory, $r$) pairs, screened or not;
the largest was $0.0080$.
\begin{itemize}
\item In $11$ of these $14$, the value-adaptive optimum (a different policy class) beat it.
\item On the six relabelled profiles alone (without the value-adaptive optimum) it exceeded $0.005$ in $3$ of $486$.
\item Every cell mean stayed below $0.005$ (largest $0.0048$), so statements about cell means are not statements about
  every pair.
\end{itemize}

\pend{} \pend{}

\section{Imperfect conditionals}\label{app:imperfect}
This appendix drops the exact-oracle assumption for two schedule families, with $p$ uniform on $V=\{|x|=k\}$, $n\ge2$ and
$1\le k\le n-1$; unlike the balanced-slice results of Section~\ref{sec:bal} and Appendices~\ref{app:hlow}, \ref{app:lhup}
and~\ref{app:conv}, this appendix does not assume $n=2k$, except in Theorem~IC2. Proposition~IC1 treats relabelled singletons with a general head, and Theorem~IC2 treats the uninformed head on fixed relabelled profiles and on a count-adaptive policy; Appendix~\ref{app:alt} summarizes both results. A \emph{head}
$\hat h$ gives each masked coordinate $i$, given the revealed coordinates $R$ and their values, a probability
$\hat h_i(x_R)$ of a one; it equals the exact marginal $a_i(x_R)$ wherever that marginal is $0$ or $1$ (forced states).
Its \emph{population masked-prediction excess} is
\[
A(\hat h)=\E\,\mathrm{kl}\bigl(a_i(x_R)\,\big\|\,\hat h_i(x_R)\bigr),
\]
where $\mathrm{kl}$ is the Bernoulli KL, $x\sim p$, the number $m$ of masked coordinates is uniform on $\{1,\dots,n\}$, the masked set
is uniform of size $m$, and $i$ is uniform in it, so that the loss is averaged over the masked coordinates. This quantity is the excess, over the exact head, of
the masked training objective with mask size uniform on $\{1,\dots,n\}$ and a per-context mean loss. The development-loss
excess of Appendix~\ref{app:controlled} is a sample estimate of it. Let $\hat Q$ be the output law of \emph{relabelled singletons} run with $\hat h$: a uniformly random order, one
coordinate per round, each drawn from $\hat h$. At an interior (non-forced) state both values are legal, and at a forced state the head is exact, so
$\hat Q(V)=1$. For an output law $Q$ we write $F(Q)=\KL(p\|Q)$, the output-law divergence of Section~\ref{sec:setup}.

\emph{Proposition IC1 (imperfect conditionals on relabelled singletons).}
(a) $F(\hat Q)\le nA(\hat h)$, the any-order likelihood bound in KL form \citep{uria2014deep,shih2022training,hoogeboom2022autoregressive}, with equality whenever $\hat h$ is \emph{coherent}, that is, its conditionals are those of a
joint law $\nu$ on $V$ with $\nu>0$; then $\hat Q=\nu$. Hence, over heads exact at forced states, $\sup\{F(\hat Q):A(\hat h)\le a\}=na$ for every finite $a\ge0$.
(b) If $\hat h$ is a \emph{label-invariant count-state head}, that is, $\hat h_i(x_R)=g(M,j)$ at the count state $(M,j)$ of Section~\ref{sec:setup}, for every masked $i$, with one
function $g$ independent of $i$, of all coordinate labels and of $R$, then $F(\hat Q)=0$. A dependence on labels is not allowed: at $n=2$, $k=1$, the head with
$\hat h_1=0.9$, $\hat h_2=0.1$ at the root and exact forced continuations gives $F=\ln(5/3)>0$.
(c) If $\operatorname{logit}\hat h_i=\operatorname{logit}(j/M)+s\,c_i$ at interior states, with $\sum_ic_i=0$ and $c\ne0$, then
$F(\hat Q)/(nA(\hat h))\to\rho_n:=n(n-\Hc_n)/(n-1)^2$ as $s\to0$, where $\Hc_n=\sum_{m\le n}1/m$. (A constant field $c$ gives a
label-invariant count-state head, which by (b) costs nothing in $F$ but is charged in $A$; the constraint removes it.) The constant $\rho_n$ is below $1$ for $n\ge3$, tends to $1$,
and its minimum is $\rho_5=163/192\approx0.84896$.

\emph{Proof.} (a) Let $q_\sigma$ be the law of the singleton schedule in order $\sigma$ run with $\hat h$. By convexity,
$F(\hat Q)=\KL(p\|\E_\sigma q_\sigma)\le\E_\sigma\KL(p\|q_\sigma)$, and by the chain rule along $\sigma$,
$\KL(p\|q_\sigma)=\sum_{t=1}^n\E_p\,\mathrm{kl}(a_{\sigma_t}\|\hat h_{\sigma_t})$ at the revealed set $\{\sigma_1,\dots,\sigma_{t-1}\}$.
Under a uniform $\sigma$, step $t$ queries a uniform coordinate of a uniform masked set of size $n-t+1$, and the steps
$t=1,\dots,n$ visit each mask size once, so $\E_\sigma\KL(p\|q_\sigma)=nA(\hat h)$. If $\hat h$ is coherent with joint law
$\nu$, the product of $\nu$'s conditionals along any order is $\nu$, so $q_\sigma=\nu$ for every $\sigma$, $\hat Q=\nu$ and
$F(\hat Q)=\KL(p\|\nu)=nA(\hat h)$. Taking $\nu$ with $\KL(p\|\nu)=na$ attains the supremum; such $\nu$ exists for every $a\ge0$, since $\KL(p\|\cdot)$ is
continuous on full-support laws on $V$, equals $0$ at $p$, and is unbounded there because $|V|\ge2$. The gap $nA-F$ is the ambiguity
term of the cross-entropy decomposition for arithmetic-mean ensembles \citep[Prop.~15]{wood2023unified}.
(b) Such a head ignores labels, so $\hat Q(\pi x)=\hat Q(x)$ for every permutation $\pi$
of the coordinates. Since $\hat Q(V)=1$ and $V$ is one orbit, $\hat Q=p$, as in Lemma~\ref{lem:tools}(e).
(c) Both $F(\hat Q)$ and $nA(\hat h)$ are of order $s^2$, so it suffices to compare their $s^2$ coefficients. When
coordinate $i$ is queried, $a_i$ is its exact marginal, and $\hat h(x_i)$ and $a(x_i)$ are the probabilities that the head
and the exact conditional give to its value $x_i$. At an interior state, $\ln\hat h(x_i)-\ln a(x_i)=s\,c_i(x_i-a_i)-\tfrac12s^2c_i^2a_i(1-a_i)+O(s^3)$, and the exact
conditionals along any order multiply to $p$. So $Y_\sigma:=\ln(q_\sigma/p)=sS_\sigma-\tfrac12s^2B_\sigma+O(s^3)$, where
$S_\sigma=\sum_tc_{\sigma_t}(x_{\sigma_t}-a_{\sigma_t})$ is a sum of martingale differences under $p$ and
$B_\sigma=\sum_tc_{\sigma_t}^2a_{\sigma_t}(1-a_{\sigma_t})$, so that $\E_pS_\sigma=0$ and $\E_pS_\sigma^2=\E_pB_\sigma$. Since
$\hat Q/p=\E_\sigma e^{Y_\sigma}$ on $V$, $F(\hat Q)=-\E_p\ln\E_\sigma e^{Y_\sigma}$, and by (a) $nA(\hat h)=\E_\sigma\KL(p\|q_\sigma)=-\E_\sigma\E_pY_\sigma$.
Expanding $\ln\E_\sigma e^{Y_\sigma}=\E_\sigma Y_\sigma+\frac12\mathrm{Var}_\sigma Y_\sigma+O(s^3)$, the first-order terms vanish because
$\E_pS_\sigma=0$, and $\E_p\E_\sigma B_\sigma=\E_p\E_\sigma S_\sigma^2$ leaves only $\E_p[(\E_\sigma S_\sigma)^2]$ in $F$; this gives
\[
F(\hat Q)=\tfrac12s^2\,\E_p\bigl[(\E_\sigma S_\sigma)^2\bigr]+O(s^3),\qquad nA(\hat h)=\tfrac12s^2\,\E_\sigma\E_pB_\sigma+O(s^3).
\]
The coordinate $i$ is queried after a uniform number $r\in\{0,\dots,n-1\}$ of reveals, taken uniformly from the other
coordinates, so $\E_\sigma a_i=\sum_{r}\frac1n\bigl(k-r\frac{k-x_i}{n-1}\bigr)/(n-r)$, and, since $\sum_{r=0}^{n-1}r/(n-r)=n\Hc_n-n$, the coefficient of $x_i$ in $\E_\sigma a_i$ is $(\Hc_n-1)/(n-1)$ and
$\E_\sigma(x_i-a_i)=\gamma_nx_i+\mathrm{const}$ with $\gamma_n=(n-\Hc_n)/(n-1)$. With $\sum_ic_i=0$ the constant drops, so $\E_\sigma S_\sigma=\gamma_n\sum_ic_ix_i$, and
the covariances of the uniform slice give $\E_p[(\E_\sigma S_\sigma)^2]=\gamma_n^2\,\frac{k(n-k)}{n(n-1)}\sum_ic_i^2$. At mask size $m$
the hypergeometric second moment is $\E\,a(1-a)=\frac{k(n-k)}{n^2}\cdot\frac{n(m-1)}{m(n-1)}$, and averaging over a uniform mask size gives
$\E\,a(1-a)=\frac{k(n-k)}{n^2}\gamma_n$, so
$\E_\sigma\E_pB_\sigma=\gamma_n\frac{k(n-k)}{n^2}\sum_ic_i^2$. The ratio is $\gamma_nn/(n-1)=\rho_n$. Finally, $1-\rho_n=[n(\Hc_n-1)-(n-1)]/(n-1)^2$, so $\rho_n<1$ for $n\ge3$
because $n\Hc_n>2n-1$, and $\rho_n\to1$. Since $\Hc_n-1\le\ln n$, $\rho_n\ge1-\ln n/(n-1)$, which is increasing in $n$ and exceeds
$0.86>163/192$ for $n\ge25$; exact values for $2\le n\le24$ give the minimum $\rho_5=163/192$. \qed

\emph{Theorem IC2 (the uninformed head; balanced slice, $n=2k$).} Let $\hat h\equiv\frac12$ at interior states, exact at
forced states. Schedules draw a whole block at its starting state: at an interior state its coordinates are independent fair bits, and at a forced state they are exact; there is no stopping inside a block, no repair and no
resampling.
(i) $A(\hat h)\le\ln2\,(\Hc_n-1)/(n-1)\to0$.
(ii) For a fixed profile $b$ with block ends $0=s_0<s_1<\dots<s_r=n$, uniformly relabelled, with output law $\hat Q_b$,
\begin{gather*}
\hat Q_b(V)=\sum_{t=1}^r\ \sum_{s_{t-1}<\ell\le s_t}\varphi(\ell)\,2^{-(s_t-\ell)},\\
\varphi(\ell)=2\binom{\ell-1}{k-1}2^{-\ell}\ \ (k\le\ell\le n-1),\quad\varphi(\ell)=0\ \text{otherwise},
\end{gather*}
where $\varphi(\ell)$ is the probability that, among fair bits read in commitment order, one of the two quotas is first used
up at the $\ell$-th bit (the factor $2$ counts the two quotas, which are symmetric; one of them is used up by the $(n-1)$-th bit, since $n-1=2k-1$ bits contain $k$ ones or $k$ zeros),
and $F=-\ln\hat Q_b(V)$. Hence, for $n\ge4$, $\hat Q_b(V)\le2r\varphi(n-1)\le2r\sqrt{2/(\pi(n-2))}$, and $F\le\ln2$ forces
$r\ge\frac14\sqrt{\pi(n-2)/2}$; at $n=2$ one block of two fair bits already gives $\hat Q_b(V)=\frac12$. For $n\ge4$ the profile $(n-L,1,\dots,1)$ with $L=\lceil\sqrt{2n}\,\rceil<n$ singletons has $\hat Q_b(V)\ge\frac12$.
So the least length of a fixed relabelled profile with $F\le\ln2$ is $\Theta(\sqrt n)$.
(iii) Let $t=\lceil\log_2n\rceil$. The count-adaptive \emph{minority policy} runs up to $t$ minority-sized rounds, followed if necessary by
one deadline round that commits all remaining coordinates; its output law is $\hat Q_{\min}$. A minority-sized round commits, at an interior state, a block whose size is the
current minority count $d$ (the smaller of the numbers of ones and zeros still required, the ones on a tie), uniformly
relabelled. Once $d=0$ the state is forced; the policy then commits all remaining coordinates and stops, and $d$ stays
$0$. Its success probability is at least $1-\frac n22^{-t}\ge\frac12$, so it achieves $F\le\ln2$ with a pathwise hard cap of at most $t+1$ rounds. This is
an achieved cap, not the least one, and not a mean-cost law.

\emph{Proof.} (i) The loss is $0$ at a forced state, where $\hat h$ is exact; at an
interior state it is $\mathrm{kl}(a\|\frac12)=\ln2-h(a)$, with $h$ the binary entropy in nats. In both cases it is at most $4\ln2\,(a-\frac12)^2$, because $h(a)\ge4\ln2\,a(1-a)$ (in bits, the standard bound $h_2(a)\ge4a(1-a)$) and $4\ln2\,(a-\frac12)^2=\ln2$ at
$a\in\{0,1\}$. At mask size $m$ the
exact marginal is $a=J/m$ with $J$ hypergeometric, $\E a=\frac12$ and $\mathrm{Var}\,a=\frac14\frac{n-m}{m(n-1)}$. Averaging
over $m$ gives $\frac{\ln2}{n(n-1)}\sum_{m=1}^n(\frac nm-1)=\ln2\,(\Hc_n-1)/(n-1)$.
(ii) A block that starts at an interior state draws fair bits, and one that starts at a forced state is exact. Read the
bits in commitment order. The first time one of the two quotas is used up is at a position $\ell\in[k,n-1]$, with
probability $\varphi(\ell)$. Nothing fails before $\ell$, and every block whose first bit comes no later than position $\ell$ starts at an interior
state, so all bits up to the end of the block containing $\ell$ are fair. After $\ell$, the output lies in $V$ exactly when the remaining
$s_t-\ell$ fair bits of the current block all take the other value; later blocks start at a forced state. This gives the
identity. The weight $\varphi$ increases on $[k,n-1]$ (the ratio $\varphi(\ell+1)/\varphi(\ell)=\ell/(2(\ell-k+1))$ is at least $1$
up to $\ell=n-2$), and $\varphi(n-1)=\binom{n-2}{k-1}2^{-(n-2)}\le\sqrt{2/(\pi(n-2))}$. Each block contributes at most
$2\varphi(n-1)$. The head is label-blind, so the relabelled output is uniform on $V$ given $V$, and $F=-\ln\hat Q_b(V)$. For the
upper profile, singletons never fail, so the output lies in $V$ exactly when the first block's number of ones $X$, a
$\Bin(n-L,\frac12)$ variable, lies in $[k-L,k]$, that is, within $L/2$ of its mean. Chebyshev's inequality bounds the
failure probability by $(n-L)/L^2\le\frac12$.
(iii) Let $\mathcal F_s$ be the history before round $s+1$. In a minority-sized round the block cannot exceed either quota, so the output can leave $V$ only in the final round.
With $X\sim\Bin(d_s,\frac12)$ the number of minority values drawn in round $s+1$, the next minority count satisfies
$d_{s+1}\le d_s-X$, so $\E[d_{s+1}\mid\mathcal F_s]\le d_s/2$ and $\E d_t\le\frac n22^{-t}$. By Markov's inequality
$\Pr(d_t\ge1)\le\frac n22^{-t}$. The run succeeds whenever the minority count reaches $0$ within the $t$ minority rounds,
since the state is then forced and the remaining coordinates are exact. So the success probability is at least
$1-\frac n22^{-t}$, which is at least $\frac12$ at $t=\lceil\log_2n\rceil$. The policy is label-blind and relabelled, so $F=-\ln\hat Q_{\min}(V)\le\ln2$. \qed

\emph{Scope and prior art.} Proposition~IC1(a) is the any-order likelihood bound \citep{uria2014deep,shih2022training}, written in KL form; its uniform-step form is
\citet[Eq.~3]{shih2022training} and \citet[\S3, derived from their Eq.~2]{hoogeboom2022autoregressive}, who trace that objective to \citet{uria2014deep}. Part (b) and the equality case of (a), which gives the supremum, are ordinary orbit and
chain-rule consequences. The same chain rule covers any head once the output is relabelled as well. Let $Y$ be the output of any sampler run with
the head, with law $Q$, and let the groups be the classes of coordinates that the target treats alike (one class on the
slice; the weight groups on tilts). Outputting $\pi Y$ with $\pi$ uniform on the full within-group permutation group and drawn independently of $Y$ (a post-processing step: it is not a round of the sampler and makes no oracle queries) gives, for a target that is exactly within-group exchangeable, with divergences in $[0,\infty]$ and the conditional term read as $+\infty$ for any $s$ with $Q_{\mathrm{sig}}(s)=0<p_{\mathrm{sig}}(s)$, $\KL(p\|Q)=\KL(p\|\bar Q)+\E_{s\sim p_{\mathrm{sig}}}\KL(p(\cdot\mid s)\|Q(\cdot\mid s))$ and $\KL(p\|\bar Q)=\KL(p_{\mathrm{sig}}\|Q_{\mathrm{sig}})$, where $\bar Q$ is the law of $\pi Y$, $s$ is the vector of within-group counts (here $s$ is not the scale of
Proposition~IC1(c)), and $p_{\mathrm{sig}}$ and $Q_{\mathrm{sig}}$ are its laws under $p$ and $Q$. Both identities are the chain rule for the map $x\mapsto s$:
$\pi$ preserves $s$, so $\bar Q_{\mathrm{sig}}=Q_{\mathrm{sig}}$, and $\bar Q(\cdot\mid s)$, like $p(\cdot\mid s)$, is uniform on the set of strings with signature $s$, so the
conditional term vanishes for $\bar Q$. On the balanced slice this is the quota leakage $-\ln Q(V)$ (Lemma~\ref{lem:tools}(e)); on tilts the between-group split error (the between-orbit term of Appendix~\ref{app:alt}) remains, and averaging over a proper subgroup need not suffice. What is specific to the quota target is the exact constant $\rho_n$ in $F/(nA)\to\rho_n$ for
the coordinate fields of Proposition~IC1(c): as their scale tends to $0$ they cost between $\frac{163}{192}$ and $1$ times $nA$. No such lower bound is
claimed for other symmetry-breaking errors, or uniformly in $n$ at a fixed scale. 

Theorem~IC2 is a hard-cap comparison for one head (fixed-profile length against an adaptive pathwise cap), not a
mean-cost law. It
shows that the aggregate $A$ alone does not control fixed-profile length. It does not show that a worst-context bound
is necessary: additional calibration assumptions, at individual states or weighted by how often the sampler visits them, may rule
out this example. Part (iii) gives
only the displayed $O(\log n)$ hard-cap upper bound; it neither identifies the least adaptive cap nor proves a doubly
logarithmic adaptive bound. Numerical checks:
\begin{itemize}
\item IC1(a)--(c) at $n=4$--$7$, exhaustively over orders, and IC1(a) and $\rho_n$ at $n=10,12$ in a separate float64 implementation;
\item the identity of IC2(ii), exactly on all $680$ profiles (compositions) of $n=4,6,8,10$;
\item IC2(i) at $n\le256$;
\item an exact rational check of the IC2(ii) identity on 320 seeded profile draws at $n\in\{6,8,10,12,16\}$ (for each $n$, $60$ random compositions with at most six cuts, repeats allowed, and four fixed profiles), in a separate implementation;
\item dynamic programs for IC2(iii) in two implementations: at $n=16,64,256,1024$ evaluated in float64 (success probability $0.67$--$0.88$ within the cap $t+1$, the deadline round included), and
at $n=8,16,32,64,128$ in exact rational arithmetic ($0.75$--$0.94$); the values at $n=16,64$ agree.
\end{itemize}
\pend{}

\section{Alternative explanations and boundary cases}\label{app:alt}
Each paragraph below changes one feature of the setting to test whether something other than the survival mechanism of Section~\ref{sec:bal} explains the gap between the two criteria.

\pend{}

\emph{The choice of divergence.} Validity alone does not separate the criteria, since the probability of a valid output is linear in the mixture and depends on each schedule only through its profile (Lemma~\ref{lem:tools}(c)), so relabelling leaves it unchanged. Nor is the shape error an artifact of KL: at $n=4096$ the fixed labelled ladder of Theorem~\ref{thm:sep}(b)'s proof with $\delta=\tfrac12$ and the Hoeffding--Ville choice $K=\lceil2\ln(2/\delta)\rceil=3$ (the stated bound takes the smaller $K_\delta=2$ of the Kolmogorov branch; both ladders fail with probability at most $\delta$) is at total variation $0.673$ from $p$, while its
relabelled mixture is at $1-s=0.015$, with $s$ the ladder's success probability (Lemma~\ref{lem:tools}(c)), since a relabelling-invariant law $Q$ has $\mathrm{TV}(p,Q)=1-Q(V)$ (Appendix~\ref{app:conv}) (Figure~\ref{fig:mech}). The randomized-side results also hold in total variation (Appendix~\ref{app:conv}).

\emph{Symmetry in the leakage-only reduction.} Let $V$ be the support of the target $p$ and $\mathcal G$ a group of coordinate permutations with $p\circ g=p$ for
all $g\in\mathcal G$. The exact oracle is then $\mathcal G$-equivariant, $q_{gU}(gx)=q_U(x)$, and relabelling a schedule by a uniform
$g\in\mathcal G$ is implementable and yields a $\mathcal G$-invariant output law $Q$. If $V$ splits into $\mathcal G$-orbits $V_1,\dots,V_m$ with
masses $p_i=p(V_i)$, the chain rule gives, for every such $Q$, $\KL(p\|Q)=-\ln Q(V)+\KL\big((p_i)_i\,\big\|\,(Q(V_i)/Q(V))_i\big)$, because $p$ and $Q$ are both uniform within each orbit. This orbit chain rule is standard. On the balanced slice with $\mathcal G=S_n$ there is one orbit and the
second term vanishes: this is Lemma~\ref{lem:tools}(e). With several orbits, as for two groups with $\mathcal G=S_{n_A}\times S_{n_B}$, relabelling still removes all within-orbit
shape, but a \emph{between-orbit} term remains, which survival does not control (Theorem~\ref{thm:tilt} controls it
differently). Without symmetry, randomizing the schedule does not symmetrize the output, and conditioning independent
draws on a count distorts the target \citep[Problem~P1]{hajek1964asymptotic}.

\begin{proposition}[permutations: fixed schedules have no shape error]\label{prop:perm}
Let $p$ be uniform on permutations of $[n]$, viewed as strings in $[n]^n$ with distinct entries, with the analogous exact
oracle (every unrevealed coordinate is uniform on the unused values), and let the valid strings $V$ be those with
distinct entries. For every fixed schedule $U$, the restriction of $q_U$ to $V$ is constant, so $\KL(p\|q_U)=-\ln q_U(V)$. Consequently, for $r\ge1$, every mixture $\mu$ supported on schedules of length at most $r$
has $F(\mu)\ge\min_{|U|\le r}\KL(p\|q_U)$: randomization over schedules cannot lower the least hard cap, which
\citet[Thm.~7]{wen2026conditional} characterize exactly over deterministic value-adaptive policies, attained by a fixed schedule (their KL is in bits).
\end{proposition}
\emph{Proof.} A block of size $b$ drawn when $M$ values are unused assigns any particular distinct values with probability
$M^{-b}$, so $q_U(x)=\prod_tM_{t-1}^{-b_t}$ for every valid $x$, with $M_{t-1}$ the number of unused values before round $t$, which
is $n-\sum_{s<t}b_s$ on every valid path and so does not depend on $x$. A mixture of laws uniform on $V$ is uniform on $V$, so
$F(\mu)=-\ln\E_\mu q_U(V)\ge-\ln\max_{|U|\le r}q_U(V)$. \qed

On the balanced slice, by contrast, $q_U(x)$ depends on the block counts of $x$, and it is this dependence that relabelling
averages out. The hard-cap statement extends to causal adaptive policies (the class of Section~\ref{sec:setup} with this oracle; a
committed history is legal when no value repeats): given a legal committed history with $u$ unused values and a chosen block of size $b$, the $b$ independent uniform draws are
distinct (and so keep the history legal) with probability $g(u,b)=\prod_{i<b}(1-i/u)$, which depends on the history only through $u$; and on
legal paths $u$ falls to $u-b$ whatever values are drawn. So $W_0(u)=\mathbf 1\{u=0\}$ and
$W_r(u)=\max\{W_{r-1}(u),\max_{1\le b\le u}g(u,b)W_{r-1}(u-b)\}$ (the first term is an empty round) bound the conditional success
within the remaining rounds, by the tower argument of Lemma~\ref{lem:surv}(a) run on $u$ in place of the count state. The fixed
schedule whose block sizes attain these maxima from $u=n$ attains $W_r(n)$, so a
single fixed schedule attains the largest success probability and, its output being constant on $V$, also the least KL
(Lemma~\ref{lem:tools}(a) gives $\KL\ge-\ln Q(V)$ for every policy); an adaptive policy cannot improve on it and may add shape error. \pend{}

\emph{Two groups: survival is not enough.} For the two-value tilts of Section~\ref{sec:tilt}, with weight $\lambda>0$ on a group~A of coordinates and $1$ on the others,
so that $p_\lambda(x)\propto\lambda^{a(x)}$ on $V$ with $a(x)$ the number of ones in group~A (Appendix~\ref{app:tgenv}), the orbit
chain rule gives, for every within-group relabelled $Q$, $F=-\ln Q(V)$ plus a divergence between the laws of $a$.
Hard-cap and adaptive lower bounds persist (Appendix~\ref{app:tgenv}). Let $D_r$ be the least KL of a deterministic
value-adaptive policy with at most $r$ rounds, $F^*_r$ that of causal randomized policies, and $G^\lambda_r$ the largest
probability of meeting the quota within $r$ rounds ($D_r$: Appendix~\ref{app:tgred}; $F^*_r$, $G^\lambda_r$: Appendix~\ref{app:tgenv}). Finite computations with an exact oracle give
$D_r/F^*_r$ of $1.3$--$1.9$, $1.8$--$3.0$ and $2.6$--$9.2$ at $r=2,3,4$ on $24$ targets with $12\le n\le24$: the equal splits $|A|=k=n/2$ at $n\in\{12,16,20,24\}$, with $\lambda\in\{1,2\}$ or one of the three Grid~Z tilts of Appendix~\ref{app:exhibits}, and $(|A|,n,k,\lambda)=(4,12,6,3)$, $(5,16,8,2)$, $(6,16,4,4)$ and $(8,16,5,2)$. Appendix~\ref{app:tgred} justifies the programs, and Appendix~\ref{app:exhibits} gives the evidence class of each value, that is, how it was computed and checked (ball-arithmetic enclosure, certificate with exact rational ingredients, or float64 evaluation). A kernel-checked case ($|A|=2$, $n=4$, $k=2$, $\lambda=2$,
$r=3$; its Lean~4 proof is checked by the Lean kernel, Appendix~\ref{app:lean}) has $G^\lambda_r=1$ yet
$F^*_r=5.0\cdot10^{-4}$: survival alone does not control the between-orbit term (the divergence between the laws of $a$), which is why
Theorem~\ref{thm:tilt} needs a different construction. Eight two-group predictions were frozen, with their pass criteria, before the held-out cells were computed; five
passed and three failed (Appendices~\ref{app:exhibits} and~\ref{app:controlled}).

\pend{}

\emph{A full-support target: the gap without a quota.} Removing the quota does not remove the gap: on Beta$(1,1)$--Bernoulli every string is valid, and the separation is treated in Section~\ref{sec:fs} and Appendix~\ref{app:thmm}, with finite exhibits in Section~\ref{sec:finite} and Appendix~\ref{app:exhibits}.

\emph{Imperfect conditionals.} For a head exact at forced states, with population masked-prediction excess $A$,
relabelled singletons have $F\le nA$ \citep{uria2014deep,hoogeboom2022autoregressive};
coherent heads, whose conditionals are those of a joint law $\nu$ on $V$ with $\nu>0$, attain equality, and a label-invariant count-state head has $F=0$
(Appendix~\ref{app:imperfect}). Order randomization erases the error of such a head but need not erase label-dependent error: for small label-dependent tilts of the exact head, $F/(nA)\to\rho_n\in[163/192,1]$ (Proposition~IC1(c)). Vanishing $A$ does not bound fixed-profile length uniformly in $n$: on the balanced slice the uninformed head ($\frac12$ at every non-forced state) has
$A\to0$, yet the best fixed relabelled profile needs $\Theta(\sqrt n)$ rounds for $F\le\ln2$, while a count-adaptive policy
achieves it with a pathwise cap of at most $\lceil\log_2n\rceil+1$ rounds (not shown to be least). A learned model's conditionals need not be compatible with any joint law \citep{torrobahennigen2023deriving}.

\emph{What $F$ says about parts of the output.} By data processing $F$ bounds the divergence of every fixed coordinate projection of the
output, such as a fixed prefix, and by the chain rule it bounds, on average over the target's prefix law, the conditional divergence given it; neither $F$
nor $H$ gives a uniform guarantee conditional on every externally specified rare prefix.

\emph{Sampling the mixtures.} Rounds are the only cost: describing or selecting a mixture is not charged, and its support may grow with $n$ (Section~\ref{sec:setup}). A large support does not by itself mean an expensive sampler. On the balanced slice each construction is one uniform permutation of the coordinates and, for the mixtures, one biased coin choosing between the ladder and the one-round schedule (Theorem~\ref{thm:sep}(b), Corollary~\ref{cor:t2}(c)). Given their parameters, both weighted constructions are sampled before any bit by explicit procedures that never list their supports: one uniform permutation of the reservoir (Appendix~\ref{app:t1}, \S6), or $O(n\log\log n)$ independent coins with rational biases (Appendix~\ref{app:rf}, \S4.2), whose parameter for a given $\varepsilon$ is not explicit because the constants are existence-only; computing the oracle's exact marginals is not analysed. For the full-support construction see Appendix~\ref{app:thmm}, after Corollary~\ref{cor:mixexact}. An extra nonnegative charge for describing or sampling a mixture could only raise costs: every lower bound here would stand, but the upper bounds would need new arguments.

\emph{Restart.} Restarting until a valid string appears, a step outside the interface, removes a
relabelled design's quota leakage, which on the balanced slice is its entire divergence (Lemma~\ref{lem:tools}(e)); it does not remove a fixed
schedule's shape error. Restart also undoes dilution: run until a valid output appears, the mixture of Corollary~\ref{cor:t2}(c), which runs the ladder (length $L$, failure probability at most $\delta_n\to0$) with weight $w=e^{-\varepsilon}/(1-\delta_n)$ and the one-round schedule otherwise, costs $(1+o(1))L$ expected rounds per valid output: each attempt costs $wL+1-w$
expected rounds and succeeds with probability between $w(1-\delta_n)=e^{-\varepsilon}$ and $w+G_1(n,k)$, both tending to $e^{-\varepsilon}$, so
by Wald's identity the expected rounds per valid output lie between $(wL+1-w)/(w+G_1(n,k))$ and $(wL+1-w)e^{\varepsilon}=L+e^{\varepsilon}-1+\delta_n(L-1)/(1-\delta_n)$. So the factor $e^{-\varepsilon}$ belongs to the one-output criterion and is not a saving per valid sample.

\begin{remark}[scope of the converse]
The converse forbids repair: resampling the current block until the quota remains satisfiable, with a uniform relabelling, outputs $p$ exactly (Lemma~\ref{lem:tools}(e); the number of resampling steps is not analysed), and restart with state reuse, remasking, lookahead, joint
within-round draws and approximate conditionals are also outside $\mathcal A_r$: at a small instance each reaches, within $r$
rounds, a success probability above the largest one in $\mathcal A_r$, so its output law is that of no member of
$\mathcal A_r$. At $(n,k)=(4,2)$, remasking and lookahead reach $1>3/4=G_2(4,2)$ in two rounds (for instance, by committing two equal bits among three drawn and redrawing the third), joint draws reach
$1>3/8=G_1(4,2)$ in one, and approximate conditionals with root marginals $\frac12+\delta$ on two coordinates and $\frac12-\delta$ on the other two,
$0<|\delta|\le\frac12$, reach $3/8+\delta^2+6\delta^4>3/8$ in one; restart reusing the first round's values reaches
$103/108>11/12=G_3(6,3)$ at $(n,k)=(6,3)$ in three rounds (commit three coordinates; if the quota is then forced, commit the
other three at once; otherwise commit two of them, and if that leaves the state illegal, discard those two values and
commit the last three coordinates at once, keeping the first round's values, and otherwise commit the last coordinate). Conditioning a round's draws on the count, or letting the seed write values, is exact in one round. In the definition of $\mathcal A_r$ (Section~\ref{sec:setup}), a filtration that also contained unused oracle randomness would let a nominally seedless policy randomize.
\end{remark}

\section{Further related work}\label{app:related}
This appendix expands the related-work discussion of the introduction and discusses selected closely related work in detail; the table of
selected related results summarizes it. Theorem numbers of \citet{chen2026optimal}
follow the PMLR version (arXiv:2511.04647v2: Theorems~4, 9, 16, 24, 25 and Definition~15 are 1.4, 1.9, 3.3, 4.9, 4.10 and
3.2).

\begin{table}[p]
\centering\footnotesize
\parbox{\textwidth}{\textbf{Selected related results.} Each row gives what the result bounds, for which samplers, round counts and targets, and what it implies for the least round count under $F\le\varepsilon$. The interfaces differ, so the rows describe and do not rank; the text has the exact statements.}\\[3pt]
\setlength{\tabcolsep}{4pt}
\begin{tabular}{@{}>{\raggedright\arraybackslash}p{0.13\textwidth}>{\raggedright\arraybackslash}p{0.30\textwidth}>{\raggedright\arraybackslash}p{0.22\textwidth}>{\raggedright\arraybackslash}p{0.295\textwidth}@{}}
\toprule
Result & Criterion; samplers or policies & Round count; target & Bearing on the least rounds under $F\le\varepsilon$\\
\midrule
\multicolumn{4}{@{}l}{\emph{Bounds on $H$, which bound $F$ only from above, through $F\le H$}}\\
\citet[Def.~15, Thms.~4, 9, 16]{chen2026optimal} & $H$, the expected per-schedule KL: exact formula, optimal block sizes, a bound via total correlation; $F\le H$. \emph{Samplers:} random unmasking: prescribed block sizes, uniformly random positions & number of blocks; any target & upper bounds on $F$ through $F\le H$\\
\addlinespace[3pt]
\citet[Prop.~4]{secchi2026schedule} & factorization error: the $H$ of the exact-conditional sampler; output KL at most it plus a learning error. \emph{Samplers:} $\tau$-leaping (random block sizes) versus deterministic block sizes, both with random positions & steps; limit at a fixed ratio of length to steps; under all assumptions of their Prop.~4 & on $H$, random block sizes cost at least as much in the limit; $F$ enters only through $F\le H$ (exact conditionals) or $F\le H+\varepsilon_{\mathrm{learn}}$ (learned ones)\\
\addlinespace[2pt]
\multicolumn{4}{@{}l}{\emph{Lower bounds on the output law}}\\
\citet[Thm.~4.4]{feng2025theoretical} & output mass off the support; above $1/2$ it gives $F>\ln2$. \emph{Samplers:} masked-diffusion sampling with any masking schedule, each masked position revealed independently; conditionals exact or with per-token KL strictly below $1/128$ & sampling steps; an explicit hidden Markov target & $\Omega(n)$ steps for $F\le\ln2$, also for mixtures of schedules and for $H$; does not separate $H$ from $F$\\
\addlinespace[3pt]
\citet[Def.~22, Thms.~24--25]{chen2026optimal} & output TV. \emph{Samplers:} randomized adaptive conditional-marginal queries, unrestricted output & at most $T$ queries with probability~1; worst case over a known family, the target in it unknown & family-level; does not constrain a fixed known target\\
\addlinespace[2pt]
\multicolumn{4}{@{}l}{\emph{Deterministic adaptive policies against randomized ones}}\\
\citet[Thm.~11, Rem.~2]{wen2026conditional} & output KL in bits ($F$; the schedule is a function of the output). \emph{Policies:} deterministic value-adaptive policies, exact marginals & target-weighted policy-tree depth; hard cap; balanced slice & $\Theta_\varepsilon(\log^2n)$ for deterministic policies; extension of their lower bounds to randomized policies left open\\
\addlinespace[3pt]
\citet[Thms.~4, 8; Cor.~9]{li2026randomization} & output KL (and TV) of mixtures of fixed schedules, against deterministic policies. \emph{Policies:} deterministic value-adaptive policies; fixed schedules; finite mixtures of target-structured fixed schedules; exact marginals & hard cap; a nine-point full-support binary family; a noisy-matching family & mixtures beat every deterministic adaptive policy in accuracy at a cap of three, and attain in three rounds an accuracy that needs $\Omega(n/\log n)$ rounds of a fixed schedule; uniform random positions leave linear error there\\
\midrule
This paper & $F$ (nats) and $H$. \emph{Policies:} predrawn schedule mixtures (description uncharged); causal randomized adaptive policies, exact marginals & scheduled mean, failed runs charged; executed mean $\E\tau$; hard cap; \emph{targets:} balanced slice; Beta$(1,1)$--Bernoulli; bounded-ratio weighted slices & under $F\le\varepsilon$, order $\log\log n$: scheduled and executed means matched to leading order; hard cap matched up to an additive constant on the two slices, while on full support every mixture's $F$ diverges under a fixed cap; under $H\le\varepsilon$, order $(\log n)^2$, whose upper half prior work gives (on the balanced slice, the whole order)\\
\bottomrule
\end{tabular}
\end{table}

\paragraph{What the schedule analyses bound.} For random schedules, several analyses state their guarantee for an
\emph{average over schedules} of a per-schedule divergence: the expected integration error of
\citet{chen2026optimal}, information-profile and factorization objectives \citep{lavenant2025error,secchi2026schedule}, and
expected KL over random masks or permutations \citep{zhao2026adaptation,li2025breaking,cai2026confidence}; each reaches the
sampler's output law, the \emph{mixture} over schedules, through convexity or marginalization. Others work differently:
\citet{wainwright2026data} bounds the output KL of Bernoulli and fixed-cardinality unmasking samplers through increments of
a path-resolved complexity, \citet{dmitriev2026efficient} bound a path-measure KL on a deterministic time grid and pass to the
output law by data processing, and \citet{park2024jump} optimize a KL upper bound for a specified schedule. The fixed-block
identity $\KL(p\|q_U)=\sum_t\E\,\mathrm{TC}_t$ \citep[Eqs.~15--16]{zhang2026generation} concerns one fixed schedule. On
schedule averages, randomness in the schedule cannot help at a fixed number of rounds, and design
conclusions are read off the average: by Markov's inequality, for every $c>1$, a fixed, randomly generated ordering has factorization
error at most $c$ times the average with probability at least $1-1/c$, and, within their random-order model, deterministic
block sizes suffice to minimize the average \citep[\S4.4 and \S5]{lavenant2025error}. For $\tau$-leaping, where the
randomness is in the block sizes, \citet[Prop.~4]{secchi2026schedule} show, in a joint scaling limit at matched round counts
and under their coefficient-convergence, nondegeneracy and schedule-regularity assumptions, that the randomized sampler's
factorization cost is at least that of the planner with deterministic block sizes (both randomize positions). These are
correct statements about the average, and a random ordering fixed once and reused across outputs gives, conditionally on the
ordering, a fixed-schedule law. The random permutation of \citet{cai2026confidence} is introduced to facilitate the analysis.
\citet[Thm.~2 and the remark after it]{dmitriev2026efficient} prove a lower bound on the path-measure KL of uniform-noise
$\tau$-leaping and note that it is not a bound on the terminal output law, although, to the best of their knowledge, all existing KL
upper-bound analyses proceed through path measures. Order-agnostic autoregressive diffusion trains the Jensen bound over orderings
\citep{hoogeboom2022autoregressive}, while order-ensemble density estimators evaluate the mixture over orderings directly and
report that it improves on the order-averaged log-likelihood \citep{uria2014deep}.

\citet{chen2026optimal} define random unmasking with prescribed block sizes (their Definition~15), give an exact formula for
its expected per-schedule KL through the information curve, and bound the output KL by that average (their Theorem~16). On
the balanced slice their formula equals $\KL(p\|q_U)$ for every labelled $U$ with the given profile, by
Lemma~\ref{lem:tools}(d), and their Theorem~4 characterizes the fixed-schedule optimum of $H$. Their Theorem~9 gives the upper
order of $H$ there and on the weighted slices of Section~\ref{sec:tilt}: such a slice is a Poisson design conditioned on its
size, so its total correlation is at most minus the log-probability of that size, $O(\log n)$ by the local limit of
\citet[Lemma~1]{boistard2012approximation} (cf.\ \citealp{hajek1964asymptotic}); for the full-support target see
Proposition~\ref{prop:hfs}. Our relabelled-profile constructions are their algorithm with particular profiles, or finite
mixtures of them; the kept-set sampler of Theorem~\ref{thm:tilt}(a) agrees with such a mixture in output law, as the
coordinates' commitment levels are independent and identically distributed and its terminal singletons are exact in any
order, while the count-adaptive optimum of Theorem~\ref{thm:conv}(a) and the within-group relabelling of
Theorem~\ref{thm:tilt}(b) are not of this form. The separation of Section~\ref{sec:bal} quantifies their observation (after
their Theorem~16) that re-randomizing the schedule for each sample is, on average, at least as good as fixing it: on the
output law of the balanced slice, randomizing positions changes the order of the least round count
(Theorem~\ref{thm:sep}).

\paragraph{Lower bounds.} That step counts depend on the metric is known \citep{feng2025theoretical}, and so
is an output-law converse over random-position schedules: for an explicit hidden Markov target, every masking schedule of
standard masked-diffusion sampling with fewer than order-$n$ steps leaves probability above $1/2$ outside the target's support
\citep[Thm.~4.4]{feng2025theoretical}. As that probability is affine in the output law, the bound persists for mixtures of
schedules and gives $F>\ln2$; this is the leakage mechanism of Section~\ref{sec:bal}, on a target where it forces linearly
many rounds. The distribution-oblivious lower bound of \citet{anari2024parallel} and the
family-level terminal-output lower bounds of \citet[Thms.~24--25]{chen2026optimal}, which hold for randomized adaptive
algorithms with unrestricted output, quantify over unknown targets and do not constrain a fixed known target. \citet{kou2026lower} prove round lower bounds of the same kind for randomized
algorithms that make polynomially many approximate-score queries per round to sample continuous targets in $\mathbb R^d$,
over families of hidden instances. Neither these
bounds of \citet{chen2026optimal} nor that of \citet{feng2025theoretical} separates $H$ from $F$. \citet{chen2026parallelism} separate diffusion paradigms for random empirical measures unknown to
the sampler. Our converses concern fixed, known targets in this interface. \citet{kawata2026counterfactual} lower-bound the
probe and commit budgets of samplers that use the same commit primitive with counterfactual probes (their Def.~2.6
and Thm.~3.2), but for a risk averaged over the decision-rule seed before the output laws are mixed, and in the worst case over
a hidden-forest target class with an approximate oracle; this gives no round lower bound under the mixed output law $F$ on a
fixed known target.

\paragraph{Deterministic adaptive policies and randomized ones.} \citet{wen2026conditional} study deterministic value-adaptive
policies with exact marginals. For the balanced slice they prove a $\Theta_\varepsilon(\log^2n)$ order (their Theorem~11, KL
in bits) for two costs: the target-weighted policy-tree depth $D_\varepsilon$, the least $\E_pR$ over such policies with KL at
most $\varepsilon$, and the hard cap $\bar D_\varepsilon$, the least $\max_xR(x)$, where $R(x)$ is the number of rounds on
realization $x$. The upper bound is attained by a fixed schedule. The mean $\E_p$ averages over $x\sim p$, not over the
sampler's own outputs as our executed mean $\E\tau$ does, and the two means can differ once a policy adapts to values (their
Remark~1). We cite their lower bound as stated and do not rely on it. Theorem~\ref{thm:sep}(a) is a mixture-level bound with explicit constants; its
rate also follows from their Theorem~11 by a Markov argument (Appendix~\ref{app:hlow}). Their observation that adaptivity
has nothing to exploit on this instance (Remark~8, at the level of rates) concerns deterministic policies, and they leave the
extension of their lower bounds to randomized policies open (Remark~2 and \S IV-H; see \citet{li2026randomization} below). Their exact hard-cap characterization for
uniform permutations (their Theorem~7), attained by a fixed schedule, is the setting of Proposition~\ref{prop:perm}: there
every fixed schedule's output law is already constant on valid strings, so, unlike on the balanced slice, randomizing the
schedule cannot lower the least hard cap.

\citet{li2026randomization} work in the same interface, with exact marginals and a hard cap, and compare deterministic
value-adaptive policies, fixed schedules and finite mixtures of fixed schedules drawn with a hidden seed; their Eq.~(4) is the
identity of Lemma~\ref{lem:tools}(b). On a nine-point full-support binary family of dimension $9r+2$, the least KL of a
deterministic adaptive policy with three rounds is $r\gamma-\log4+o(1)$, while a uniform mixture of four fixed schedules built
from the target's structure has KL at most $3(3r+2)\beta^r$ (their Thm.~4); deterministic adaptive sampling is exact there in
four rounds, so the gap is in accuracy at a fixed depth, and their Theorem~4 answers Remark~2 of \citet{wen2026conditional}
negatively on a constructed family. On a noisy-matching family, fixed schedules need $\Omega(n/\log n)$ rounds for a fixed KL that a
three-round mixture attains, uniform random positions leave linear mixture error, and a deterministic content-adaptive policy
is exact in three rounds (their Thm.~8 and Cor.~9). For their selector families with a fixed finite library of schedules, and under their
Lemma~7 with vanishing error, they characterize the limiting least KL of library mixtures and when it vanishes (their Cor.~10); they list an
expected round budget, instead of a hard cap, as open (their Section~VII). On the balanced slice, Theorem~\ref{thm:conv} and Corollary~\ref{cor:t2} give the randomized optimum itself, for
the hard cap and for the expected executed rounds.

\paragraph{Other interfaces and resources.} \citet{bansal2026understanding} use the same exact-marginal product-commit
primitive on graph-walk state coordinates, compare random and greedy-entropy two-at-a-time policies by coherence on selected
DAGs, and, for order-$k$ Markov walks and bridges, prove exact bisection with depth $O(k(1+\log(L/k)))$; this does not
transfer to our bit coordinates, where two or more unforced increments are conditionally dependent. \citet{yao2026dont} changes
the decoder itself: a learned marker gates a coordination pass over the selected positions before per-position greedy
argmax, which targets a joint mode rather than a joint draw; this lies outside this interface. Samplers with the same conditional-marginal oracle that may reject and redraw a batch reach approximate samples, at fixed total-variation accuracy, of laws on $k$-subsets whose conditional laws are all $1/\alpha$-entropically independent with $\alpha=\Omega(1)$, in $\tilde O(k^{1/2+c})$ parallel time for any constant $c>0$ \citep{anari2023quadratic}, and exact samples of any target on $[q]^n$ in $\tilde O(n^{1/2})$ expected parallel time \citep{anari2025autospeculation}; rejection lies outside this interface. Deterministic unmasking avoids the mixture \citep{turok2026duel}, and learned-order models train the order as a
latent variable \citep{wang2025learning,fox2026learning}. \citet{turok2026duel} and \citet{fox2026learning} compare step or
evaluation budgets empirically, and none of these works proves a converse for the least hard cap of rounds in the
exact-marginal interface. \citet[Prop.~5.1]{zhang2026generation}
lower-bound the expected number of proposals needed to verify a block exactly by accept--reject, for one fixed ordered
partition, not for the output mixture $\KL(p\|\E_Uq_U)$ or its round rate. \citet{zhang2026cost} study mixtures of
pre-selected exact-support plans on the same slice, measured in total variation, with plan count and memory rather than rounds
as the resource; their orbit-averaging lemma is the identity behind Lemma~\ref{lem:tools}(e). Randomized parallel algorithms are long known to be much more powerful than deterministic ones for search with an independence oracle: for an $n$-element independence system, finding a maximal independent set takes, in the worst case, $\Omega(n/\log p)$ parallel steps for deterministic algorithms with $p$ processors but $O(\sqrt n)$ expected steps for randomized ones with $n^{3/2}$ processors \citep{karp1988complexity}. Randomizing a sampler to improve
its output law is also known \citep{shen2019randomized}; here the gap is between the schedule-averaged divergence $H$ and the
output-law divergence $F$ of the same sampler. The square-root ladder of Theorem~\ref{thm:sep}(b) is a short computation on the doubly exponential grid of batched
bandits \citep{perchet2016batched,gao2019batched}, whose round lower bounds concern regret; the substance here is the converse
(Section~\ref{sec:bal}).

\section{Kernel-checked declarations}\label{app:lean}
{\raggedright
All declarations below, except the \texttt{TGCert3} brackets of Appendix~\ref{app:exhibits}, are in the authors'{} Lean
package (Lean 4.33.1 \citep{moura2021lean4}, Mathlib v4.33.1
\citep{mathlib2020}; 369 modules of its own). The \texttt{TGCert3} brackets are in a separate certificate package, whose 1129 modules build on the first.\ Both packages are included in the ancillary formal companion archive.{} Each package rebuilds from source in a clean environment with no \texttt{sorry}. The certificate package was built a second time
from source, in two stages (its first 616 modules in a clean environment, the other 513 on top of that build), with
identical axiom reports. \texttt{\#print axioms} gives exactly propext, Classical.choice
and Quot.sound for every theorem named here. A kernel check certifies a formal statement,
not its agreement with the printed one; the package was built and replayed from source, and its declarations were compared with the printed results group by group. Each item below names the parts of its appendix that are proved by ordinary
arguments only. Where a printed proof of a kernel-checked statement argues differently from its kernel proof, the
printed argument is ordinary unless the item names it as a printed route. Declaration names, Lean comments and the packages' README files keep an earlier numbering (\texttt{thm1b\_*} is Theorem~\ref{thm:sep}(b), \texttt{thm2b\_*} Theorem~\ref{thm:conv}(b)), and an earlier appendix order: \texttt{AppJ} is Appendix~\ref{app:tgenv}, \texttt{AppM} is Appendix~\ref{app:imperfect} (\texttt{AppM.M1*} is Proposition~IC1 and \texttt{AppM.M2*} Theorem~IC2), \texttt{AppGSep} and \texttt{AppGTable} are Appendix~\ref{app:repro}, and \texttt{TGRed.K$i$} is Lemma~TGR$i$. The README files' appendix, section, theorem, table and figure numbers are those of an earlier version of this paper; this appendix gives the current correspondence.
\par\smallskip
The table maps each main result to its ordinary proof, to the declarations whose types state it and to what is not
kernel-checked; the items after it list the supporting declarations. Every declaration in the third column is a theorem
whose type and axiom report the packages record, and \texttt{A36W.TGCert3.*.bracket} stands for the 24 brackets that the
two-group item below names; where several declarations are listed, their types state the result jointly, and the last
column names every clause that they do not state. The conventions are $\delta=1-e^{-\varepsilon}$ for \texttt{Fopt}, and a
hard cap read almost surely, \texttt{CapAS}, with the cap on all paths, including those of probability zero,
\texttt{CapEvery}, also stated where available; some
declarations hold on a wider range, such as every $k\le n$ rather than $n=2k$, or every $\varepsilon\ge0$. The hard-cap
declarations of the rows for Theorem~\ref{thm:conv}(b), Corollary~\ref{cor:t2}(a), Theorem~\ref{thm:tilt}(a) and
Section~\ref{sec:finite} are stated with the output law after $r$ rounds; under \texttt{CapAS}~$r$ that law is the terminal
law $Q_\pi$ (\texttt{A36RW.Converse.outLim\_eq\_outBy\_of\_capAS}, for every oracle whose conditional marginals lie in
$[0,1]$, as exact ones do). For Lemma~\ref{lem:tools}(a) the first listed declaration assumes $q(V)>0$, and \texttt{A36RW.Lemma1aZero.lemma1a\_zero}
states the case $q(V)=0$, with both sides $+\infty$; the identity of Lemma~\ref{lem:tools}(b) is stated, as in its proof, for $F(\mu)<\infty$. For $\mathcal A_r$ and
$\mathcal A_\infty$ the declarations quantify over the formal class \texttt{A36RW.PolicyE}. For every exact oracle,
\texttt{A36RW.W13.classId\_of\_exact} translates it in both directions to and from a formalization of the printed class
with a round clock, preserving the output laws at every horizon and in the limit, the caps, the executed means, the KL
values, the sets of hard-cap KL values and $\LF^{\mathrm{ad}}$. The description of Section~\ref{sec:setup} is also formalized as a probability-space process (\texttt{A36RW.W13x.PrintedProcess}:
a seed, committed transcripts, decisions and a completion measurable with respect to $\mathcal F_{t-1}$, and conditionally
fresh draws). For every exact oracle, every such process has the output law at every horizon and in the limit, the
completion probabilities, the almost-sure cap and the executed mean of a policy of the formal class
(\texttt{A36RW.W13x.printed\_process\_in\_class}); conversely, these quantities of every policy of the formal class
are those of such a process (\texttt{A36RW.W13x.realize\_in\_class}); and a completion driven by explicit randomness
of its own, with the draws fresh given the seed, that randomness and the transcript, folds into the seed
(\texttt{A36RW.W13x.foldRC\_in\_class}). \dag\ marks the rows that rest on this identification. Its one ordinary step
is reading the sentence of Section~\ref{sec:setup} as that process, with a completion given only as a randomized rule
represented by explicit randomness.
\par\smallskip{\footnotesize\renewcommand{\theHtable}{fqfront.\arabic{table}}\setlength{\tabcolsep}{3pt}\setlength{\LTleft}{0pt}\setlength{\LTright}{0pt}\setlength{\LTpre}{\smallskipamount}\setlength{\LTpost}{\smallskipamount}
\begin{longtable}{@{}>{\raggedright\arraybackslash}p{2.2cm}>{\raggedright\arraybackslash}p{2.3cm}>{\raggedright\arraybackslash}p{6.8cm}>{\raggedright\arraybackslash}p{3.3cm}@{}}
\toprule
result & ordinary proof & kernel-checked declarations & not kernel-checked\\
\midrule
\endfirsthead
\toprule
result & ordinary proof & kernel-checked declarations & not kernel-checked\\
\midrule
\endhead
\bottomrule
\endfoot
\bottomrule
\endlastfoot
Lemma~\ref{lem:tools}(a)--(e) & App.~\ref{app:hlow} & (a)~\texttt{A36RW.\allowbreak{}Relabel.\allowbreak{}KLx\_eq\_neg\_log\_add} and, for $q(V)=0$, \texttt{A36RW.\allowbreak{}Lemma1aZero.\allowbreak{}lemma1a\_zero}; (b)~\texttt{A36RW.W11.Fext\_le\_Hext}, \texttt{A36RW.W11.Hext\_eq\_add}; (c)~\texttt{AGF.\allowbreak{}agfv\_succMass\_eq\_snOf}; (d)~\texttt{A36RW.\allowbreak{}Relabel.\allowbreak{}KLx\_schedLawE\_profile}; (e)~\texttt{A36RW.\allowbreak{}Relabel.\allowbreak{}lemma1e\_balanced} & ---\\
\eqref{eq:env} & App.~\ref{app:env} & \texttt{A36Sharp.sharp\_envelope}, \texttt{A36Sharp.sharpKap\_lt\_36} & ---\\
Lemma~\ref{lem:surv} & App.~\ref{app:conv} & (a)~\texttt{A36RW.\allowbreak{}Balanced.\allowbreak{}lemma2a\_printed}, \texttt{A36.GrRec\_mono}; for fixed schedules: \texttt{A36.\allowbreak{}succMass\_qSched\_le\_GrRec}; (b)~\texttt{A36RW.\allowbreak{}ExecMean.\allowbreak{}lemma2b\_tail}, \texttt{A36RW.\allowbreak{}Converse.\allowbreak{}deadline\_le\_GrRec}, \texttt{A36RW.\allowbreak{}Converse.\allowbreak{}margUnit\_balanced}, \texttt{A36RW.\allowbreak{}Converse.\allowbreak{}policy\_outBy\_le\_succ}, \texttt{A36RW.\allowbreak{}Converse.\allowbreak{}policy\_outBy\_nonneg}, \texttt{A36RW.\allowbreak{}Converse.\allowbreak{}policy\_doneBy\_le\_one} & \dag\\
Theorem~\ref{thm:conv}(a) & App.~\ref{app:conv} & \texttt{A36RW.\allowbreak{}Balanced.\allowbreak{}hardcap\_optimum\_outLim}; attained by the randomly ordered rule: \texttt{A36RW.Attain.sigmaStar\_KL}, \texttt{A36RW.\allowbreak{}Attain.\allowbreak{}sigmaStar\_capAS}, \texttt{A36RW.\allowbreak{}Converse.\allowbreak{}outLim\_eq\_outBy\_of\_capAS}, \texttt{A36RW.\allowbreak{}Converse.\allowbreak{}margUnit\_balanced} & \dag\\
Theorem~\ref{thm:conv}(b) & Apps.~\ref{app:env}, \ref{app:conv-window} & \texttt{A36RW.\allowbreak{}Window.\allowbreak{}window\_two\_sided}, \texttt{A36RW.Window.thm2b\_upper} & \dag\\
Theorem~\ref{thm:sep}(a) & App.~\ref{app:hlow} & \texttt{AGFH.\allowbreak{}Hopt\_ge\_three\_unfolded}, \texttt{AGFCert.Hopt\_family\_lb} & ---\\
Theorem~\ref{thm:sep}(b) & App.~\ref{app:hlow} & \texttt{A36RW.HV.thm1b\_Fopt}, \texttt{A36RW.HV.thm1b\_attained} & ---\\
Theorem~\ref{thm:sep}(c) & Apps.~\ref{app:hlow}, \ref{app:lhup}; monotonicity in $\varepsilon$: definition of $\LH$ & \texttt{A36RW.LHUp.ratio\_Theta}, \texttt{A36RW.LHUp.appB\_headline}, \texttt{A36RW.LHUp.LHgen\_Theta}, \texttt{A36RW.LHUp.LHgen\_antitone}, \texttt{A36Sep.\allowbreak{}separation\_unbounded} & ---\\
Corollary~\ref{cor:hard} & App.~\ref{app:conv} & \texttt{A36RW.Cor2a.cor1\_formal}, \texttt{A36RW.LHUp.Afix\_Theta} & ---\\
Corollary~\ref{cor:t2}(a) & App.~\ref{app:conv} & \texttt{A36RW.\allowbreak{}Cor2a.\allowbreak{}cor2a\_characterization}, \texttt{A36RW.\allowbreak{}HV.\allowbreak{}cor2a\_bounds\_Kdelta} & \dag\\
Corollary~\ref{cor:t2}(b) & App.~\ref{app:conv} & \texttt{A36RW.\allowbreak{}Balanced.\allowbreak{}cor2b\_printed}, \texttt{A36RW.Balanced.cor2b\_LF}, \texttt{A36RW.\allowbreak{}TVJoin2.\allowbreak{}cor2b\_LFgen\_inf} & \dag\\
Corollary~\ref{cor:t2}(c), and the least $\E\tau$ after it & App.~\ref{app:conv} & \texttt{A36RW.Cor2c.headline\_cor2c}, \texttt{A36RW.LHUp.ratio\_Theta} & \dag\\
Theorem~\ref{thm:fs}(a)--(d) & (a)~App.~\ref{app:thmm-hfs}; (b)~\ref{app:thmm-one-lower}, \ref{app:thmm-one-upper}; (c)~\ref{app:thmm-mix-lower}, \ref{app:thmm-mix-upper}; (d)~\ref{app:thmm-mix-lower} & (a)~\texttt{A36FS.Prop2.LHgen\_isTheta}; (b)~\texttt{A36FS.\allowbreak{}Misc.\allowbreak{}ProfAllEps.\allowbreak{}thm3b\_F\_all}; (c)~\texttt{A36FS.Thm8.Dilution.cor5}; (d)~\texttt{A36FS.Thm7C.thm3d} & ---\\
Theorem~\ref{thm:fsadapt} & App.~\ref{app:thmm-adapt} & \texttt{A36FS.Thm9.thm9a\_full}, \texttt{A36FS.Thm9.LFad\_ge}, \texttt{A36FS.Thm9.thm9b\_liminf}, \texttt{A36FS.Thm8.Dilution.thm9c} & \dag\\
Theorem~\ref{thm:tilt}(a) & App.~\ref{app:t1} \S\S3--4 ($H$), \S10 ($F$ lower); App.~\ref{app:rf} ($F$ upper) & $\LH^w$: \texttt{A36W.HL.LH\_theta}; $R_F^{\mathrm{adapt}}$: \texttt{A36W.\allowbreak{}RFAttain.\allowbreak{}rf\_least\_adaptive\_hardcap}; $\LF^w$: \texttt{A36W.\allowbreak{}RFAttain.\allowbreak{}rf\_sharp\_mean\_two\_sided}; $\E\tau$: \texttt{A36W.\allowbreak{}RFAttain.\allowbreak{}rf\_sharp\_mean\_two\_sided\_ad} & \dag\\
Theorem~\ref{thm:tilt}(b) & App.~\ref{app:t1} \S\S6--7 & \texttt{A36W.FU.R.theoremR}, \texttt{A36W.\allowbreak{}Attain.\allowbreak{}common\_length\_asymptotic\_R} & ---\\
Section~\ref{sec:finite} & evidence: Apps.~\ref{app:repro}, \ref{app:exhibits}, \ref{app:controlled}; finite reductions: App.~\ref{app:tgred} & $n=65536$: \texttt{AGFCert.Hopt\_65536\_gt}, \texttt{AGFPot.Fopt\_le\_eleven}; $\LH>\LF$ from $n=2^{15}$: \texttt{A36RW.\allowbreak{}AppGSep.\allowbreak{}LF\_lt\_LH\_log\_two}; the 25 certified two-group optima of App.~\ref{app:exhibits} (18 rows of $F^*_r$, 7 cells of $F^{\mathrm{NA}}_r$): the 24 \texttt{A36W.TGCert3.*.bracket} and \texttt{A36W.\allowbreak{}TGCert.\allowbreak{}W49\_FNA\_bracket}; the six-coordinate witness's factor $2.35$ and its every-predrawn bound: \texttt{A36W.TGCert.W49}, \texttt{A36W.\allowbreak{}TGCert.\allowbreak{}W49\_beats\_every\_predrawn}, attained by a (possibly randomized) rule choosing each block's group composition from the committed counts: \texttt{A36W.\allowbreak{}TGCertRule.\allowbreak{}W49\_rule\_beats\_every\_predrawn} & every other finite value or bound, in the evidence class that Apps.~\ref{app:reference}, \ref{app:exhibits} and \ref{app:controlled} state for it (ball-arithmetic enclosures, certificates with exact rational ingredients, float64 evaluations and estimates, best-found bounds); the frozen-prediction checks; the learned-head tests; \dag\\
\end{longtable}
\addtocounter{table}{-1}}\par
\begin{itemize}
\item Canonical modules: \texttt{AGFCert.Hopt\_family\_lb}, \texttt{AGFH.Hopt\_ge\_three\_unfolded}, \texttt{AGFH.klSum\_chain},
\texttt{AGFCert.Hopt\_65536\_gt}, \texttt{AGFPot.Fopt\_le\_eleven}, \texttt{AGFCert.Hopt\_65536\_gt\_two\_Fopt},
\texttt{AGF.agfv\_c08s\_claim\_C08\_support}, \texttt{AGF.agfv\_succMass\_eq\_snOf}, \texttt{agfv\_Fopt\_le\_of\_leakage},
\texttt{AGFPot.lossSched\_rcProfile\_le\_rcBudget}. Here \texttt{AGF.agfv\_c08s\_claim\_C08\_support} is the envelope
$G_r(M,j)\le\min\{1,2\cdot6^{r-1}(d+1)^{-2^{-r}}\}$ of Appendix~\ref{app:env}, so $G_r(n,k)\le\Gamma_r(n)$.
\item Balanced slice, for fixed schedules, relabelled profiles and their mixtures: \texttt{A36RC.Fopt\_le\_rc\_tower} (Theorem~\ref{thm:sep}(b), $K=\lceil1/\delta\rceil$),
\texttt{A36.succMass\_qSched\_le\_GrRec}, \texttt{A36.GrRec\_mono} (Lemma~\ref{lem:surv}(a) for fixed schedules),
\texttt{A36.Fopt\_ge\_mean\_bridge} (the single-horizon form of Lemma~\ref{lem:surv}(b) for mixtures with $\theta=e^{-\varepsilon}$:
$\LF(n,\varepsilon)\ge1+t(e^{-\varepsilon}-G_t(n,k))$ for every $t$), \texttt{A36.Fopt\_ge\_envelope} (the same with
$\Gamma_t(n)$ in place of $G_t(n,k)$, $t\ge1$), \texttt{A36Det.det\_schedule\_length\_ge},
\texttt{A36RC.relabelled\_rc\_hard} (Corollary~\ref{cor:hard}), \texttt{A36Sep.separation\_unbounded}
(the unboundedness part of Theorem~\ref{thm:sep}(c); its rate, in the two-sided form of Corollary~\ref{cor:t2}(c), is
\texttt{A36RW.LHUp.ratio\_Theta} below).
\item The bounded-prefactor envelope: \texttt{A36Sharp.sharp\_envelope} and
\texttt{A36Sharp.sharpKap\_lt\_36}, the bounded-prefactor envelope \eqref{eq:env} of Appendix~\ref{app:env},
$G_r(M,j)\le\min\{1,A_r(d+1)^{-2^{-r}}\}$ at every legal state $(M,j)$, $d=\min(j,M-j)$, with $A_r<36$.
\item Behavioural and seeded policies: \texttt{AGF.CapV2.agfv\_apb\_claim\_C08\_all\_policy} and \texttt{AGF.CapV2.agfv\_apb\_seed\_succWithinOf\_le\_GrRec} (Lemma~\ref{lem:surv}(a) at horizon $r$ for behavioural and seeded policies: success within $r$ rounds is at most
$G_r(n,k)$; for these classes the passage to output KL in Theorem~\ref{thm:conv}(a) is ordinary, and
Theorem~\ref{thm:conv}(a) itself is kernel-checked for $\mathcal A_r$ below), \texttt{AGF.CapV2.agfv\_apb\_seed\_envelope} (the seeded envelope bound), \texttt{AGF.MB.agfv\_mb\_claim\_C08\_mean\_bridge\_closed} (the single-horizon form of Lemma~\ref{lem:surv}(b), $1+t(\theta-G_t(n,k))\le\E\tau$ for $t\ge1$, behavioural reading; it is stated for a success threshold
$\theta$, and $\theta=e^{-\varepsilon}$ by Lemma~\ref{lem:tools}(a)). The policy classes of these branch files have no empty rounds; when every run completes ($\tau<\infty$ almost surely), compressing out empty
rounds, with the seed and a simulated round clock kept, preserves the output law and does not increase the round count, an ordinary step; if $\Pr(\tau=\infty)>0$ the expected-round bounds hold trivially.
\item Balanced slice, for the class $\mathcal A_r$ of Section~\ref{sec:setup}. The formal class
\texttt{A36RW.PolicyE} has a seed with an arbitrary probability law, independent of the oracle; each batch a function of
the seed and of the executed labelled history, empty rounds included, measurable in the seed; fresh product draws at
the exact conditional marginals at legal prefixes and an arbitrary completion at illegal ones; and every executed
round charged, empty or not. The hard cap is almost sure ($\Pr(M_r=0)=1$, \texttt{CapAS}); Theorem~\ref{thm:conv}(a)
also holds for the stronger reading on all paths, including those of probability zero (\texttt{CapEvery}). Declarations:
\texttt{A36RW.Balanced.hardcap\_optimum} and \texttt{A36RW.Attain.sigmaStar\_KL} (Theorem~\ref{thm:conv}(a), for every
$k\le n$, attained by the randomly ordered count-state rule, stated with the output law after $r$ rounds);
\texttt{A36RW.Balanced.hardcap\_optimum\_outLim}, \texttt{A36RW.Attain.sigmaStar\_capAS} and
\texttt{A36RW.Converse.outLim\_eq\_outBy\_of\_capAS} (the same for the terminal output law $Q_\pi$, which under an
almost-sure cap $r$ is the law after $r$ rounds; the rule obeys that cap); \texttt{A36RW.Window.window\_two\_sided} and \texttt{A36RW.Window.thm2b\_upper}
(Theorem~\ref{thm:conv}(b), with the relabelled witness and the factor $1728$); \texttt{A36RW.Balanced.lemma2a\_printed}
and \texttt{A36RW.ExecMean.lemma2b\_tail} (Lemma~\ref{lem:surv}; (a) for every $k\le n$; (b) at every threshold
$\theta$, its law hypotheses holding for every policy by \texttt{A36RW.Converse.margUnit\_balanced},
\texttt{A36RW.Converse.policy\_outBy\_le\_succ}, \texttt{A36RW.Converse.policy\_outBy\_nonneg},
\texttt{A36RW.Converse.policy\_doneBy\_le\_one} and \texttt{A36RW.Converse.deadline\_le\_GrRec});
\texttt{A36RW.Cor2a.cor2a\_characterization} and \texttt{A36RW.HV.cor2a\_bounds\_Kdelta} (Corollary~\ref{cor:t2}(a));
\texttt{A36RW.Balanced.cor2b\_printed}, \texttt{A36RW.Balanced.cor2b\_LF} and \texttt{A36RW.TVJoin2.cor2b\_LFgen\_inf}
(Corollary~\ref{cor:t2}(b), also for $\LF$); \texttt{A36RW.Cor2c.headline\_cor2c} and \texttt{A36RW.LHUp.ratio\_Theta}
(Corollary~\ref{cor:t2}(c) and the expected-rounds infimum after it, with the rate $30/\sqrt{\log_2\ln n}$ of Appendix~\ref{app:conv}); \texttt{A36RW.HV.thm1b\_Fopt} and
\texttt{A36RW.HV.thm1b\_attained} (Theorem~\ref{thm:sep}(b), with $K_\delta$ as printed, attained);
\texttt{A36RW.LHUp.appB\_headline} and \texttt{A36RW.LHUp.LHgen\_Theta} (the bounds on $\LH$ in
Theorem~\ref{thm:sep}(c), Appendix~\ref{app:lhup}); \texttt{A36RW.LHUp.LHgen\_antitone} (its monotonicity in
$\varepsilon$); \texttt{A36RW.LHUp.KL\_le\_Phi} (Step~2 of Appendix~\ref{app:lhup}: $\KL(p\|q_U)\le\Phi(b)$ for every fixed
schedule $U$ with size profile $b$); \texttt{A36RW.CW1b.half\_budget\_headline} (the first sentence of the remark on the asymptotic constant in
Appendix~\ref{app:lhup}, for fixed $\varepsilon>0$ as $n\to\infty$: the chain's bound on $\Phi$ eventually bounds $\Phi$ and the KL
of every full schedule with that profile, and its ratio to $\varepsilon/2$ tends to $1$; the length bound of this route has
ratio to $(\ln n)^2/\varepsilon$ tending to $1$); \texttt{A36RW.LHUpR.remark\_headline} and
\texttt{A36RW.LHUpR.remark\_asymptotic} (the rest of that remark,
with $n\gtrsim10^3$ at $\varepsilon=\ln2$ as the sufficient condition $n\ge1000$); \texttt{A36RW.Cor2a.cor1\_formal} and
\texttt{A36RW.LHUp.Afix\_Theta} (Corollary~\ref{cor:hard}); \texttt{A36RW.W11.Fext\_le\_Hext},
\texttt{A36RW.W11.Hext\_eq\_add} and \texttt{A36RW.W11.Hext\_eq\_top\_and} (Lemma~\ref{lem:tools}(b), every target, the
case of two infinite sides included); \texttt{A36RW.Relabel.KLx\_eq\_neg\_log\_add}, \texttt{A36RW.Relabel.KLx\_schedLawE\_profile} and
\texttt{A36RW.Relabel.lemma1e\_balanced} (Lemma~\ref{lem:tools}(a) as an equality, (d), and (e) for every law
invariant on $V$); \texttt{A36RW.Relabel.orbit\_chain\_rule} (the orbit chain rule of Appendix~\ref{app:alt}, for
every group of coordinate permutations fixing $p$); \texttt{A36RW.AltPerm.prop\_perm} and
\texttt{A36RW.AltPerm.perm\_adaptive\_hardcap} (Proposition~\ref{prop:perm} of Appendix~\ref{app:alt} and its
extension to causal adaptive policies, under both cap readings); \texttt{A36RW.W11.each\_restriction\_needed}
(the $n=4$, $r=2$ facts of the remark on the scope of the converse in Appendix~\ref{app:alt}, as explicit output
laws) and
\texttt{A36RW.W11R.each\_restriction\_needed\_classes} (the three relaxed models of that remark as policy
classes that contain the formal class, with the $n=4$ witnesses as members, under the cap on all paths, including those
of probability zero) and
\texttt{A36RW.CW1b.each\_restriction\_needed\_AS} with \texttt{A36RW.CW1b.best\_success\_AS\_4\_2} (the same under the
almost-sure cap of $\mathcal A_r$, where the class's best success at $n=4$, $r=2$ is again $3/4$);
and, in total variation (Appendix~\ref{app:conv}), \texttt{A36RW.W11.hardcap\_TV\_optimum},
\texttt{A36RW.TVJoin2.thm1b\_TV\_attained\_Kdelta}, \texttt{A36RW.TVJoin2.cor2a\_TV\_chain\_Kdelta},
\texttt{A36RW.W11.cor2b\_TV}, \texttt{A36RW.TVJoin2.cor2b\_LFtv\_inf} and \texttt{A36RW.W11.cor2c\_TV} (Theorems~\ref{thm:conv}(a) and~\ref{thm:sep}(b),
Corollary~\ref{cor:t2}(a), (b) and the $\LF$ clause of (c), with $\varepsilon=-\ln(1-\delta)$). The identification of the formal class, whose histories record empty rounds, with the description of
Section~\ref{sec:setup} by seed, committed transcript and round clock is kernel-checked as well
(\texttt{A36RW.W13.classId\_of\_exact}: under an exact oracle the two classes have the same output law at every
horizon, seed by seed, hence the same costs, caps and optima; the printed-class forms include
\texttt{A36RW.W13.hardcap\_optimum\_P}, \texttt{A36RW.W13.lemma2a\_printed\_P}, \texttt{A36RW.W13.cor2b\_printed\_P}
and \texttt{A36RW.W13.headline\_cor2c\_P}); for every exact oracle, the printed description, as the process \texttt{A36RW.W13x.PrintedProcess}, has the same
output laws, completion probabilities, almost-sure caps, executed means and limiting laws as the formal class, in both
directions (\texttt{A36RW.W13x.printed\_process\_in\_class}, \texttt{A36RW.W13x.realize\_in\_class}); only reading the
description as that process is ordinary, as in the preamble of this appendix. The repair clause of the remark on the scope of the converse is
kernel-checked as well: \texttt{A36RW.W22.repaired\_relabelled} (resampling a block until the quota remains
satisfiable, with a uniform relabelling, outputs $p$ exactly, by Lemma~\ref{lem:tools}(e); the resampling loop enters
through its law, the product law conditioned on a legal child) and \texttt{A36RW.W22.repaired\_not\_policy} (no policy
of the class has that law within $|U|$ rounds when $G_{|U|}(n,k)<1$, a hypothesis discharged at $(n,k,|U|)=(2,1,1)$).
The rest of that remark is kernel-checked: the law separations as \texttt{A36RW.BRem.scope\_outside}, at the instances
the remark names (each listed mechanism produces within $r$ rounds a law that no member of $\mathcal A_r$ produces within
$r$ rounds), the one-round exactness of conditioning a round's draws on the count (on every slice) and of letting the seed
write values (for every target) as \texttt{A36RW.W11R.CondW.cond\_witness} and \texttt{A36RW.W11R.WriteW.write\_witness},
and the success values and class bounds as \texttt{A36RW.BRem.RemaskW.remask\_witness},
\texttt{A36RW.W11R.LookW.look\_witness}, \texttt{A36RW.BRem.JointW.joint\_witness}, \texttt{A36RW.BRem.ApproxW.approx\_succ},
\texttt{A36RW.BRem.RestartW6.restart6\_witness}, \texttt{A36RW.BRem.class\_succ\_le\_one}, \texttt{A36RW.BRem.class\_succ\_le\_two}
and \texttt{A36RW.BRem.GrRec6.G3\_6\_3}, with \texttt{A36RW.W11.GrRec\_1\_4\_2} and \texttt{A36RW.W11.GrRec\_2\_4\_2} for the
values $G_1(4,2)=3/8$ and $G_2(4,2)=3/4$. Ordinary only: the estimate of Appendix~\ref{app:alt} that restarting the
mixture of Corollary~\ref{cor:t2}(c) until a valid output appears costs about $L+e^{\varepsilon}-1$ expected rounds per
valid output, where $L$ is the ladder's length.
\item Reproducibility (Appendix~\ref{app:repro}). Table~\ref{tab:bounds}, row by row:
\texttt{A36RW.AppGTable.table2\_row8}, \texttt{A36RW.AppGTable.table2\_row16}, \texttt{A36RW.AppGTable.table2\_row32},
\texttt{A36RW.AppGTable.table2\_row64}, \texttt{A36RW.AppGTable.table2\_row128} and \texttt{A36RW.AppGTable.table2\_row1024}
(at $n=2^m$ and $\varepsilon=\ln2$, each printed entry as a bound on its quantity: $\LH$, $\LF$ from both sides, and
$R_F^{\mathrm{adapt}}$ under both cap readings). The paragraph after Figure~\ref{fig:bounds}:
\texttt{A36RW.AppGSep.sep\_log\_two} and \texttt{A36RW.AppGSep.sep\_tenth} (the upper bound of
Theorem~\ref{thm:sep}(b) is strictly below the lower bound of Theorem~\ref{thm:sep}(a) for every even $n\ge2^{15}$ at
$\varepsilon=\ln2$ and every even $n\ge2^{13}$ at $\varepsilon=0.1$), with \texttt{A36RW.AppGSep.LF\_lt\_LH\_log\_two} and
\texttt{A36RW.AppGSep.LF\_lt\_LH\_tenth} (hence $\LF<\LH$ there); \texttt{A36RW.AppGSep.gamma\_thresholds} and
\texttt{A36RW.AppGSep.env\_thresholds} (the thresholds on $n$ for $\Gamma_r(n)<1$, and for the bounded prefactor of
\eqref{eq:env} to give a bound below~$1$, at $d=k$ and $r=2,3,4$). Ordinary only: the rest of that paragraph (the order
and weakness of the terminal lower bound, the dominance of the prefactor, and the comparison of the two prefactors),
that each entry of Table~\ref{tab:bounds} is its formula rounded to the printed precision, and the curves of
Figure~\ref{fig:bounds}, which are formula evaluations. At $n=2^{16}$, Table~\ref{tab:bounds}'s entries $\LH\ge26.20$ and $\LF\le20$ also follow from
\texttt{AGFCert.Hopt\_65536\_gt} and \texttt{AGFPot.Fopt\_le\_eleven} above, respectively.
\item Two-group survival envelope (Appendix~\ref{app:tgenv}): \texttt{A36RW.AppJ.thm\_tgenv\_sharp} with \texttt{A36RW.AppJ.Csharp\_lt} (Theorem~\ref{thm:tgenv}, for every $\lambda>0$, all
group sizes and every legal state; its geometric form is \texttt{A36RW.AppJ.thm\_tgenv}); Lemmas~TG0--TG10, among them \texttt{A36RW.AppJ.TG1}, \texttt{A36RW.AppJ.W16.TG4\_ii},
\texttt{A36RW.AppJ.W16.TG5}, \texttt{A36RW.AppJ.W16.TG6}, \texttt{A36RW.AppJ.TG9} (every causal randomized policy) and
\texttt{A36RW.AppJ.TG10}; Corollary~(a) (\texttt{A36RW.AppJ.corA\_capKLSetAS\_full}, almost-sure cap), (b)
(\texttt{A36RW.AppJ.markov\_split} and \texttt{A36RW.AppJ.corB\_asymptotic\_full}, for $\varepsilon\ge0$) and (d)
(\texttt{A36RW.AppJ.corD\_capKLSetAS\_full}, \texttt{A36RW.AppJ.corD\_rounds\_full}).
Also kernel-checked along the printed routes: the case $v=0$ of the proof of Lemma~TG1 (\texttt{A36W.PV.RouteJ.TG1\_v0}); the TP2 proof of Lemma~TG2, for $a_A$ and, through $W=x-U_A$, for $a_B$, with its displayed consequence (\texttt{A36W.PV.RouteJ2.aA\_mono\_route}, \texttt{A36W.PV.RouteJ2.aB\_mono\_route}, \texttt{A36W.PV.RouteJ2.TG2\_route}); step~(1) of the proofs of Lemmas~TG5 (\texttt{A36W.PV.RouteJ.TG5\_step1}) and~TG8 (\texttt{A36W.PV.RouteJ.TG8\_step1\_law}, \texttt{A36W.PV.RouteJ.TG8\_step1\_means}); and the supermartingale proof of Lemma~TG9, with its empty-batch case through Lemma~TG10 (\texttt{A36W.PV.RouteJ2.Y\_isSupermartingale}, \texttt{A36W.PV.RouteJ2.TG9\_route\_prob}; its consequence for $Q_\pi$, with the cap read almost surely, is \texttt{A36W.PV.RouteJ2.TG9\_route\_capAS}).
The declarations state Lemma~TG4 for the law of the batch values (\texttt{A36RW.AppJ.CW3.TG4v\_i\_A}, \texttt{A36RW.AppJ.CW3.TG4v\_i\_B},
\texttt{A36RW.AppJ.CW3.TG4v\_i\_mean} and \texttt{A36RW.AppJ.CW3.TG4v\_ii}, through \texttt{A36RW.AppJ.CW3.batchValues\_pushforward}),
and the rounds forms also hold under the printed hypothesis $F^*_r\le\varepsilon$, with $F^*_r$ the infimum
(\texttt{A36RW.AppJ.W20.corD\_rounds\_Fstar}, \texttt{A36RW.AppJ.W20.corD\_RF\_gt}, \texttt{A36RW.AppJ.W20.corA\_RF\_ge}).
Also kernel-checked: Lemma~TG11 (\texttt{A36RW.AppJ.W21.TG11}: a deterministic seedless policy in $\mathrm{RAND}_r$
attains $G^\lambda_r(\mathrm{root})$, which is therefore the maximum over $\mathrm{RAND}_r$ and the supremum, over every
causal randomized policy, of the probability of success within $r$ rounds); in (d), $c_{\mathrm{TG}}<3.154$
(\texttt{A36RW.AppJ.W20.cTG\_lt}), the executed mean bound (\texttt{A36RW.AppJ.W20.corD\_execMean\_ge}) and the
mixture clause (\texttt{A36RW.AppJ.W20.corD\_LFgen\_ge}, for $\varepsilon\ge0$); Corollary~(c)1
(\texttt{A36RW.AppJ.W21.corC1\_two\_sided}, \texttt{A36RW.AppJ.W21.corC1\_no\_uniform}); and the second remark after (c):
for $\lambda=N^4$ (\texttt{A36RW.AppJ.W21.Remark.remark\_RF\_eq\_one}, \texttt{A36RW.AppJ.W21.Remark.remark\_LF\_eq\_one}) and, by the $A\leftrightarrow B$ symmetry, for $\lambda=N^{-4}$ (\texttt{A36RW.AppJ.CW4.famTinv\_eq}, \texttt{A36RW.AppJ.CW4.famOinv\_eq}, \texttt{A36RW.AppJ.CW4.oneRound\_KL\_le\_inv}, \texttt{A36RW.AppJ.CW4.remark\_oneRound\_inv}, \texttt{A36RW.AppJ.CW4.remark\_RF\_eq\_one\_inv}, \texttt{A36RW.AppJ.CW4.remark\_LF\_eq\_one\_inv}); that on this family, which violates condition~(1), no $\lambda$-uniform lower bound tending to infinity holds (\texttt{A36RW.AppJ.CW4.fam\_not\_cond1}, \texttt{A36RW.AppJ.CW4.no\_uniform\_lower\_bound\_RF}, \texttt{A36RW.AppJ.CW4.no\_uniform\_lower\_bound\_LF}, \texttt{A36RW.AppJ.CW4.no\_uniform\_lower\_bound\_RF\_inv}, \texttt{A36RW.AppJ.CW4.no\_uniform\_lower\_bound\_LF\_inv}), in particular neither $R^\lambda_F(\varepsilon)\ge c\log_2\log_2n-C$ nor $\LF^\lambda(n,\varepsilon)\ge c\log_2\ln n-C$ for any $c>0$ (\texttt{A36RW.AppJ.CW4.no\_doublyLog\_RF}, \texttt{A36RW.AppJ.CW4.no\_doublyLog\_LF}, \texttt{A36RW.AppJ.CW4.no\_doublyLog\_RF\_inv}, \texttt{A36RW.AppJ.CW4.no\_doublyLog\_LF\_inv}); Steps~1--3 (\texttt{A36RW.AppJ.CW4.step1\_ratio}, \texttt{A36RW.AppJ.CW4.expZ\_famT}, \texttt{A36RW.AppJ.CW4.step1\_expZ\_le}, \texttt{A36RW.AppJ.CW4.nu\_eq\_min}, \texttt{A36RW.AppJ.CW4.KLx\_prodMarg\_eq}, \texttt{A36RW.AppJ.CW4.oneRound\_KL\_eq\_fam}); and the identity for $\KL(p_\lambda\|Q)$ with $\lambda'$ as \texttt{A36RW.AppJ.CW4.lamPrime}, whose two terms Step~3 bounds at once (\texttt{A36RW.AppJ.CW4.KLx\_split}, \texttt{A36RW.AppJ.CW4.KLx\_oneRound\_split\_fam}, \texttt{A36RW.AppJ.CW4.KLx\_split\_terms\_fam}).
Further kernel-checked: the mathematical content of the remarks after Theorem~\ref{thm:tgenv} (\texttt{A36RW.AppJ.CW3.const\_depends\_on\_r\_only}, \texttt{A36RW.AppJ.CW3.Cr\_values}, \texttt{A36RW.AppJ.CW3.Cr\_ratio\_le}, \texttt{A36RW.AppJ.CW3.Cr\_le\_geo}, \texttt{A36RW.AppJ.CW3.V\_oneSided\_asymp\_inv}, \texttt{A36RW.AppJ.CW3.V\_NNN\_tendsto\_N}, \texttt{A36RW.AppJ.CW3.V\_one\_balanced\_root}, \texttt{A36RW.AppJ.CW3.tgenv\_balanced\_shape}, \texttt{A36RW.AppJ.CW3.Cr\_div\_Ar\_mem}, \texttt{A36RW.AppJ.CW3.base\_factor\_le}); Corollary~(c)2 (\texttt{A36RW.AppJ.CW3.corC2\_conclusion}, \texttt{A36RW.AppJ.CW3.example\_theta}, \texttt{A36RW.AppJ.CW3.example\_root}), with the paragraph on where $c$ is small (\texttt{A36RW.AppJ.CW3.sig2\_floor}, \texttt{A36RW.AppJ.CW3.cLim\_ge\_floor}, \texttt{A36RW.AppJ.CW3.necessity}, \texttt{A36RW.AppJ.CW3.limits\_atTop\_gt}, \texttt{A36RW.AppJ.CW3.limits\_atTop\_lt}, \texttt{A36RW.AppJ.CW3.limits\_atTop\_eq}, \texttt{A36RW.AppJ.CW3.cLim\_tendsto\_zero}, \texttt{A36RW.AppJ.CW3.tendsto\_atTop\_zero\_iff}, \texttt{A36RW.AppJ.CW3.tendsto\_zero\_zero\_iff}, \texttt{A36RW.AppJ.CW3.both\_zero\_iff}, \texttt{A36RW.AppJ.CW3.diag\_no\_uniform}, \texttt{A36RW.AppJ.CW3.offDiag\_uniform}, \texttt{A36RW.AppJ.CW3.example\_half\_quarter}, \texttt{A36RW.AppJ.CW3.example\_03\_05}) and the one on one-sided bounds (\texttt{A36RW.AppJ.CW3.V\_oneSided\_asymp}, \texttt{A36RW.AppJ.CW3.V\_oneSided\_asymp\_inv}, \texttt{A36RW.AppJ.CW3.V\_oneSided\_tendsto}, \texttt{A36RW.AppJ.CW3.V\_oneSided\_symm}, \texttt{A36RW.AppJ.CW3.V\_oneSided\_printed}, \texttt{A36RW.AppJ.CW3.V\_oneSided\_printed\_inv}); Corollary~(c)3 (\texttt{A36RW.AppJ.CW3.consequence1\_loglog}, \texttt{A36RW.AppJ.CW3.consequence2\_loglog}, \texttt{A36RW.AppJ.CW3.consequence1\_RF}, \texttt{A36RW.AppJ.CW3.consequence2\_RF}, \texttt{A36RW.AppJ.CW3.consequence1\_LF}, \texttt{A36RW.AppJ.CW3.consequence2\_LF}, and with (d) \texttt{A36RW.AppJ.CW3.consequence1\_RF\_d}, \texttt{A36RW.AppJ.CW3.consequence2\_RF\_d}, \texttt{A36RW.AppJ.CW3.consequence1\_execMean\_d}, \texttt{A36RW.AppJ.CW3.consequence2\_execMean\_d}); the exact facts that interpret the numerical checks, whose outcomes are finite falsification evidence and are not
kernel-checked: which checks are vacuous at the tested sizes, which tested states are informative, the constants used there
and the reduction to deterministic policies (\texttt{A36RW.AppJ.CW3.env\_two\_gt\_one\_iff}, \texttt{A36RW.AppJ.CW3.env\_three\_gt\_one\_of\_lt}, \texttt{A36RW.AppJ.CW3.envGeo\_three\_gt\_one\_of\_lt}, \texttt{A36RW.AppJ.CW3.tg6\_bound\_ge\_one\_iff}, \texttt{A36RW.AppJ.CW3.tg6\_bound\_eq\_one\_iff}, \texttt{A36RW.AppJ.CW3.checks\_vacuous\_n16}, \texttt{A36RW.AppJ.CW3.V\_le\_four}, \texttt{A36RW.AppJ.CW3.caseF\_vacuous\_n200}, \texttt{A36RW.AppJ.CW3.dpEnv\_vacuous\_n400}, \texttt{A36RW.AppJ.CW3.tg6\_informative\_states}, \texttt{A36RW.AppJ.CW3.caseS\_bound\_lt\_one}, \texttt{A36RW.AppJ.CW3.g32\_digits}, \texttt{A36RW.AppJ.CW3.gS\_32\_lt}, \texttt{A36RW.AppJ.CW3.gS\_isGreatest}, \texttt{A36RW.AppJ.CW3.det\_suffice}) and of the remark on loose constants (\texttt{A36RW.AppJ.CW3.env\_lt\_one\_imp}, \texttt{A36RW.AppJ.CW3.env\_two\_lt\_one\_imp}, \texttt{A36RW.AppJ.CW3.env\_lt\_one\_imp\_r3}, \texttt{A36RW.AppJ.CW3.Cr\_pow\_mono}, \texttt{A36RW.AppJ.CW3.V\_one\_balanced\_root}, \texttt{A36RW.AppJ.CW3.balanced\_need\_r2}, \texttt{A36RW.AppJ.CW3.balanced\_need\_r3}, \texttt{A36RW.AppJ.CW3.balancedEnv\_two\_lt\_one\_iff}). The first remark after (c), that (a) and (b) hold in terms of
$V_{\mathrm{root}}$ whenever $V_{\mathrm{root}}\to\infty$, is the form in which \texttt{A36RW.AppJ.corA\_capKLSetAS\_full} and
\texttt{A36RW.AppJ.corB\_asymptotic\_full} are stated, with no condition on $n_A/n$ or $k/n$ and none on $\lambda$ beyond $\lambda>0$. That its two hypotheses are not necessary is \texttt{A36RW.AppJ.CW5.not\_necessary}: the family $n_A=n_B=k=N$ with
$\lambda_N=(\ln N)^2$ violates (1) for every $\delta>0$ and (2), and has $V_{\mathrm{root}}\to\infty$ and
$\log_2\ln V_{\mathrm{root}}=\log_2\ln n+o(1)$. Ordinary only: the closing
numerical sentence of the second remark.
\item Two-group reductions (Appendix~\ref{app:tgred}), for every $\lambda>0$, all group sizes and every $k\le n$ ($\mathrm{RAND}_r$ with the almost-sure cap,
$\mathrm{NA}_r$ the predrawn mixtures of full labelled schedules with at most $r$ blocks; where a lemma needs the Setting's
$0<k<n$ or $r\ge1$, it is assumed): \texttt{A36W.TGRed.K1a}, \texttt{A36W.TGRed.K1b} and \texttt{A36W.TGRed.K1c}
(Lemma~TGR1, with $\mathcal C_r$ as \texttt{A36W.TGRed.Crec}); \texttt{A36W.TGRed.K2a} (Lemma~TGR2(a);
\texttt{A36W.TGRed.K2a\_le} and \texttt{A36W.TGRed.K2a\_eq} state its two parts without normalisation, which covers
sub-probabilities), \texttt{A36W.TGRed.K2b\_RAND} and \texttt{A36W.TGRed.K2b\_NA} (TGR2(b), with the relabelled mixture for
$\mathrm{NA}_r$), \texttt{A36W.TGRed.K2c\_Fstar}, \texttt{A36W.TGRed.K2c\_NA} and \texttt{A36W.TGRed.Phi\_decomp} (TGR2(c),
the decomposition for $q>0$), and \texttt{A36W.TGRed.K2a\_lt} with \texttt{A36W.TGRed.nonvac\_K2a\_strict} (the
strictness remark after TGR2, at an explicit fixed schedule); \texttt{A36W.TGRed.K3} (Lemma~TGR3;
\texttt{A36W.TGRed.K3\_hull} is the hull identity for every $r$); \texttt{A36W.TGRed.K4a\_Mstar},
\texttt{A36W.TGRed.K4a\_hat\_Mstar}, \texttt{A36W.TGRed.K4b} and \texttt{A36W.TGRed.K4c\_hull} (Lemma~TGR4, (b) for every
$c\ge0$); \texttt{A36W.TGRed.K5\_success}, \texttt{A36W.TGRed.K5\_lower} and \texttt{A36W.TGRed.K5\_sandwich} (TGR5, the
sandwich for $r\ge1$ with the policy of Lemma~TG11 as $\pi^\circ$); \texttt{A36W.TGRed.K6a\_printed},
\texttt{A36W.TGRed.K6b}, \texttt{A36W.TGRed.K6c} and \texttt{A36W.TGRed.K6d} (Lemma~TGR6); \texttt{A36W.TGRed.K7}
(Lemma~TGR7, for every minimiser); \texttt{A36W.TGRed.frontier\_le} ($F^*_r\le\min(D_r,F^{\mathrm{NA}}_r)$) and
\texttt{A36W.TGRed.no\_latent\_peeking} (the no-latent-peeking example, as a computation outside the formal class).
The remark on $u_A$ at $n_A=n_B=k=3$, $r=3$, $\lambda\in\{2,4\}$: \texttt{A36W.TGCert2.W43\_no\_uA\_indep\_attains\_lam2}
and \texttt{A36W.TGCert2.W43\_no\_uA\_indep\_attains\_lam4} (every element of \texttt{A36W.TGCert2.subKLSet} exceeds
$F^*_3$; that set contains the output KL of every predrawn randomized mixture of $u_A$-independent deterministic count
tables, by \texttt{A36W.TGCert2.mix\_sub\_mem}), and \texttt{A36W.TGCert2.W43\_gap\_lam2} and
\texttt{A36W.TGCert2.W43\_gap\_lam4} (the gaps $3.5\cdot10^{-5}$ and $2.3\cdot10^{-4}$ nats, as brackets of
$\texttt{Fsub}-F^*_3$; \texttt{A36W.TGCert2.Fsub} is the infimum of $\Phi$ over the hull of those tables' terminal
vectors, and equals the infimum of \texttt{subKLSet} by \texttt{A36W.TGCert2.subKLSet\_sInf}).
Exact finite certificates from these reductions, for every placement of group~A. Appendix~\ref{app:exhibits}'s
six-coordinate witness, at $(3,3,3)$, $\lambda=4$, $r=3$: \texttt{A36W.TGCert.W49} ($F^*_3$ and $F^{\mathrm{NA}}_3$ are
finite, $F^*_3>0$ and $F^{\mathrm{NA}}_3\ge2.35\,F^*_3$), \texttt{A36W.TGCert.W49\_factor\_2358} (the same with $2.358$ in
place of $2.35$) and \texttt{A36W.TGCert.W49\_beats\_every\_predrawn} (a policy in $\mathrm{RAND}_3$ has finite output KL
at most $1/2.35$ times that of every predrawn mixture with at most three blocks), and \texttt{A36W.TGCertRule.W49\_rule\_beats\_every\_predrawn}, which exhibits such a policy: a seeded mixture of two count tables, relabelled uniformly within the groups, in which the composition of every block is its table's value at the round index and the committed counts; its output KL $a$ is positive and at most $1/2.35$ times that of every such predrawn mixture. Appendix~\ref{app:alt}'s certified case
$n=4$: \texttt{A36W.TGCert.W31a} (at $(2,2,2)$, $\lambda=2$, $r=3$: $G^\lambda_3=1$, and $F^*_3$ lies in the rounding
interval of $5.01857\cdot10^{-4}$, so $F^*_3>0$). Appendix~\ref{app:exhibits}'s certified brackets, from the separate certificate package: for $F^*_r$ on
$\mathrm{RAND}_r$ with the almost-sure cap, at the $18$ owner-only rows, \texttt{A36W.TGCert3.R4\_8\_6\_l3\_r2.bracket}, \texttt{A36W.TGCert3.R4\_8\_6\_l3\_r3.bracket}, \texttt{A36W.TGCert3.I486.bracket} ($(4,8,6)$, $\lambda=3$, $r=4$), \texttt{A36W.TGCert3.R5\_11\_8\_l2\_r2.bracket}, \texttt{A36W.TGCert3.R5\_11\_8\_l2\_r3.bracket}, \texttt{A36W.TGCert3.R5\_11\_8\_l2\_r4.bracket}, \texttt{A36W.TGCert3.R6\_10\_4\_l4\_r2.bracket}, \texttt{A36W.TGCert3.R6\_10\_4\_l4\_r3.bracket}, \texttt{A36W.TGCert3.R6\_10\_4\_l4\_r4.bracket}, \texttt{A36W.TGCert3.R8\_8\_5\_l2\_r2.bracket}, \texttt{A36W.TGCert3.R8\_8\_5\_l2\_r3.bracket}, \texttt{A36W.TGCert3.R8\_8\_5\_l2\_r4.bracket}, \texttt{A36W.TGCert3.R10\_10\_10\_l1\_r2.bracket}, \texttt{A36W.TGCert3.R10\_10\_10\_l1\_r3.bracket}, \texttt{A36W.TGCert3.R10\_10\_10\_l1\_r4.bracket}, \texttt{A36W.TGCert3.R12\_12\_12\_l1\_r2.bracket}, \texttt{A36W.TGCert3.R12\_12\_12\_l1\_r3.bracket} and \texttt{A36W.TGCert3.R12\_12\_12\_l1\_r4.bracket}; for $F^{\mathrm{NA}}_r$ on predrawn mixtures with at most $r$ blocks, at
the seven certified cells, \texttt{A36W.TGCert3.N3\_3\_3\_l1\_r3.bracket}, \texttt{A36W.TGCert3.N3\_3\_3\_l2\_r3.bracket}, \texttt{A36W.TGCert3.N4\_8\_6\_l3\_r3.bracket}, \texttt{A36W.TGCert3.N4\_8\_6\_l3\_r4.bracket}, \texttt{A36W.TGCert3.N666.bracket} ($(6,6,6)$, $\lambda=2$, $r=3$), \texttt{A36W.TGCert3.N6\_6\_6\_l2\_r4.bracket} and \texttt{A36W.TGCert.W49\_FNA\_bracket} (the witness, in the main package). Each states $L\le F\le U$ for the optimum $F$ it names, with rational $L$ and $U$ and for every placement of
group~A; a tag such as \texttt{R5\_11\_8\_l2\_r4} gives $(n_A,n_B,k)=(5,11,8)$, $\lambda=2$ and $r=4$.
Further kernel-checked, for the two conclusions of the remark on the order of the TGR7 bracket:
\texttt{A36W.TGRem.halfWidth\_expansion} (for every finite $W$, $P>0$ on $W$ and every $\hat q$, as $g\to0^+$ the half-width
before the caps is $\sqrt{2g}\sum_w\hat q_w/\sqrt{P_w}+o(\sqrt g)$), and, on the remark's segment example at
$P=(\frac12,\frac12)$, \texttt{A36W.TGRem.bracket\_width\_ge} (every bracket that is a function of $(\hat q,g)$ and valid on convex
sets has width at least $|t|$ at $(q(t),g(t))$, $|t|<\frac15$) and \texttt{A36W.TGRem.halfwidth\_not\_isLittleO} (so its half-width
is not $o(\sqrt{g(t)})$ as $t\to0$), with \texttt{A36W.TGRem.gex\_isTheta} ($g(t)=\Theta(t^2)$) and
\texttt{A36W.TGRem.sOf\_shift\_isTheta} ($s(q(t))-s(q^*)=\Theta(\sqrt{g(t)})$). The remarks that at $r=0$ the classes are empty, that $H(\pi)\ge\KL(p_\lambda\|Q_\pi)$ by convexity, and that at $r=n$ all three frontier values are $0$: \texttt{A36W.PV.classes\_r0\_empty} (for every $n\ge1$, target and oracle), \texttt{A36W.PV.KLx\_le\_H} (for every $\pi\in\mathrm{RAND}_r$; for every exact pair, \texttt{A36W.PV.KLx\_le\_pathH\_avg\_of\_exact}) and \texttt{A36W.PV.frontier\_values\_at\_n}.
Ordinary only: the rest of the remark on $u_A$ (including the other instances tested), the rest of the remark on the order
of the TGR7 bracket, the identification of the tables' survival planner with Lemma~TG11's policy, and the rest of
Appendix~\ref{app:tgred}.
\item Imperfect conditionals (Appendix~\ref{app:imperfect}): \texttt{A36RW.AppM.Qhat\_vmass} (the setting: relabelled singletons never leave $V$);
\texttt{A36RW.AppM.M1a.M1a\_le}, \texttt{A36RW.AppM.M1a.M1a\_coherent}, \texttt{A36RW.AppM.M1a.M1a\_sup} and
\texttt{A36RW.AppM.M1a.Hext\_eq\_nAext} (Proposition~IC1(a)); \texttt{A36RW.AppM.M1b.M1b\_F\_zero} and
\texttt{A36RW.AppM.M1b.M1b\_cex\_F} ((b), with its counterexample); \texttt{A36RW.AppM.M1cMain.M1c\_limit\_family} and
\texttt{A36RW.AppM.M1cMain.M1c\_rho} ((c), proved through $o(s^2)$ limits in place of the $O(s^3)$ expansions of its proof,
whose remainders stay ordinary); \texttt{A36RW.AppM.M2i.M2i\_bound} and \texttt{A36RW.AppM.M2i.M2i\_tendsto} (Theorem~IC2(i));
\texttt{A36RW.AppM.M2ii.M2ii\_formula}, \texttt{A36RW.AppM.M2ii.M2ii\_F}, \texttt{A36RW.AppM.M2ii.M2ii\_upper},
\texttt{A36RW.AppM.M2ii.M2ii\_n2}, \texttt{A36RW.AppM.M2ii.M2ii\_profile\_lower} and \texttt{A36RW.AppM.M2ii.M2ii\_Theta} ((ii));
\texttt{A36RW.AppM.M2iii.M2iii\_statement} ((iii)). The relabelled-output chain rule of its scope and prior-art
remarks: \texttt{A36W.SigChain.relabel\_chain\_rule\_head\_ext} (for the output law of any head and policy, both
identities in $[0,\infty]$), with \texttt{A36W.SigChain.relabel\_chain\_rule\_ext} (the first identity for every output
law), \texttt{A36W.SigChain.KLx\_Qbar\_eq\_sigKLx} (the second identity for every output law, with
$\KL(p_{\mathrm{sig}}\|Q_{\mathrm{sig}})=\infty$ when $Q_{\mathrm{sig}}(s)=0<p_{\mathrm{sig}}(s)$ for some $s$) and
\texttt{A36W.SigChain.KLx\_Qbar\_eq\_top\_iff} ($\KL(p\|\bar Q)=\infty$ exactly when $Q$ gives mass $0$ to an orbit
of the support); \texttt{A36W.SigChain.remark\_proper\_subgroup}, \texttt{A36W.SigChain.remark\_one\_group\_subgroup}
and \texttt{A36W.SigChain.remark\_one\_group\_subgroup\_tilt} (averaging over a proper subgroup need not suffice).
Ordinary only: the rest of the scope and prior-art remarks, including the description of relabelling as
post-processing, and the numerical checks of Appendix~\ref{app:imperfect}.
\item Weighted adaptive converse (Appendix~\ref{app:t1}, \S10): \texttt{A36W.WAFinal.WA\_policy} (the Theorem with its policy
paragraph: for every causal randomized policy with fresh exact-marginal draws, success within $r$ rounds is at most
$\min\{1,A_r(d_0+1)^{-2^{-r}}\}$, with that section's $\Lambda$-dependent $A_r=2\Lambda(12\Lambda)^{2-2^{2-r}}$ (not the $A_r$ of~\eqref{eq:env}), and at most $\min\{1,288\Lambda^3(d_0+1)^{-2^{-r}}\}$), \texttt{A36W.WAInst.WA\_envelope}
(the envelope at every legal state), \texttt{A36W.WAFinal.corA\_KLx\_ge\_outLim\_capAS} and
\texttt{A36W.WAFinal.corA\_rounds\_outLim\_capAS} (Corollary~(a) for the almost-sure cap; \texttt{CapEvery} forms are
also proved), \texttt{A36W.WAFinal.corB\_execMean} and \texttt{A36W.WAFinal.cK\_288\_le} (Corollary~(b)), and
\texttt{A36W.WAFinal.corC\_rounds\_outLim\_capAS} and \texttt{A36W.WAFinal.corC\_execMean} (Corollary~(c)). For $d_0=0$ (the end of (b)): \texttt{A36W.PV.bdry\_target} (the target is the point mass at the unique word with $k$ ones), \texttt{A36W.PV.bdry\_oneRound\_KL} (one round committing every coordinate has output KL $0$) and \texttt{A36W.PV.bdry\_policy} (as a policy: a cap of one round on every path and almost surely, output KL $0$ and $\E\tau\le1$).
\texttt{A36W.WAMixMean.LFgen\_ge} gives Corollary~(c)'s mean bound for finite predrawn mixtures,
$\LF^w(n,\varepsilon)\ge e^{-\varepsilon}\log_2\ln n-e^{-\varepsilon}-c(K_\Lambda)$ for $\varepsilon\ge0$ under the
hypotheses of~(c), uniformly in the weights and $k$; this is an explicit form of the $F$-lower clause of
Theorem~\ref{thm:tilt} and of the Remark in \S7. The
attainment paragraph of \S10, under a homogeneous group of size at least $\rho n$ (the hypotheses of the Remark after the
\S7 table): \texttt{A36W.Attain.theoremR\_policy\_floor} (for $n\ge\lfloor(M^*-1)/\rho\rfloor+1$, or $M\le8K_4$, the profile is in
$\mathrm{RAND}_{\lfloor\log_2\ln n+C_R\rfloor}$ with $F\le\varepsilon$), \texttt{A36W.Attain.least\_adaptive\_hardcap\_R}
(the least adaptive hard cap is $\log_2\ln n+O_{\varepsilon,\Lambda,\kappa_0,\rho}(1)$) and
\texttt{A36W.Attain.sharp\_mean\_two\_sided\_ad\_R} (the least expected executed rounds over all causal policies are
$(e^{-\varepsilon}+o(1))\log_2\ln n$).
\item Two-value tilts: the $F$ upper bound and the least hard cap and mean of \S9 (Appendix~\ref{app:t1}, \S\S6
and~9): \texttt{A36W.FU.FU\_theorem} and
\texttt{A36W.FU.FU\_LF\_le} (item (FU) of the Result: for $n\ge n_0^{\mathrm{ff}}$, or $M\le8K_4$, the profile
$\mu_{\mathrm{ff}}$ of \S6 has $F\le\varepsilon$, and every component has length at most
$\max\{1,\lceil2\Lambda^2/\varepsilon\rceil\}+8K_4+\lceil\log_2\log_2n\rceil$), with
\texttt{A36W.FU.length\_bound\_le\_C\_FU} for the last step of that chain ($\le\log_2\ln n+C_{\mathrm{FU}}$, $n\ge2$);
\texttt{A36W.FU.FB\_KL} (the identity (FB)); \texttt{A36W.FU.stageA\_le\_eps4\_of\_isSchedule} (the Stage~A bound (FA) of
\S6.4, $D(p_A\|Q_A)\le\varepsilon/4$); \texttt{A36W.FU.RwF\_two\_sided\_printed} (\S9 (i), and (ii) with
\texttt{A36W.FU.length\_bound\_le\_C\_FU} for its last step) and \texttt{A36W.FU.RwF\_asymptotic}
($R^w_F=\log_2\ln n+O_{\varepsilon,\Lambda,\kappa_0}(1)$, with $\mathrm{RAND}_r$ read as the almost-sure cap); and
\texttt{A36W.FU.sharp\_mean\_two\_sided} and \texttt{A36W.FU.sharp\_mean\_two\_sided\_ad} (\S9, ``Sharp mean
coefficient'': $\LF^w$ and its adaptive analogue are $(e^{-\varepsilon}+o(1))\log_2\ln n$), all uniformly in
$\lambda$, the split and $k$. For the two class facts of \S6.2, which \S6.2 marks as not used below: \texttt{A36W.WC.valTree\_outside\_predrawn\_weighted} (an explicit rule that chooses its second batch from a committed value completes within $3$ rounds, yet no predrawn mixture has its law of the first two rounds), \texttt{A36W.WC.late\_draw\_eq\_mixLawE} (drawing the rest of the schedule later, from a coin that does not see the committed values, has the output law of the mixture that draws it first), and \texttt{A36W.WC.posterior\_weights} with \texttt{A36W.WC.posterior\_ne\_prior\_instance} (when the components have different predetermined prefixes, their conditional weights given an $x_A$ of positive probability are the posterior weights, which can differ from the prior ones). For more weight values with a homogeneous group (the Remark after the \S7 table):
\texttt{A36W.FU.R.theoremR} and \texttt{A36W.FU.R.theoremR\_LF\_le} (for any weights with ratio at most $\Lambda$, a
set of $M\ge\rho n$ equal-weight coordinates, an interior quota and $n\ge\lfloor(M^*-1)/\rho\rfloor+1$, or $M\le8K_4$,
the profile has $F\le\varepsilon$ and every component has length at most $\log_2\ln n+C_R$); the geometric Stage~A
sentence is formal: its round bound \texttt{A36W.FU.R.theoremR\_geo}, and its conclusion under the explicit size condition
(\texttt{A36W.FU.R.theoremR\_geo\_loglog}) and, for a fixed $C$ and large $n$, as \texttt{A36W.Attain.geo\_loglog\_eventually}; \texttt{A36W.Attain.theoremR\_q},
\texttt{A36W.Attain.least\_adaptive\_hardcap\_q}, \texttt{A36W.Attain.sharp\_mean\_two\_sided\_q} and
\texttt{A36W.Attain.common\_length\_asymptotic\_q} give the Remark's extension to any fixed number $q$ of weight values,
and \texttt{A36W.Attain.sharp\_mean\_two\_sided\_R} and \texttt{A36W.Attain.common\_length\_asymptotic\_R} its mean
$\LF^w=(e^{-\varepsilon}+o(1))\log_2\ln n$ and least common length $\log_2\ln n+O_{\varepsilon,\Lambda,\kappa_0,\rho}(1)$
under a homogeneous group, with \texttt{A36W.Attain.common\_length\_lower\_any} for the common-length lower bound for any
weights. For any weights
with ratio at most $\Lambda$ (\S\S3--4): \texttt{A36W.HL.HL\_bound} and \texttt{A36W.HL.HL\_bound\_simple} ((HL), for an
interior quota), \texttt{A36W.HU.HU\_schedule} and \texttt{A36W.HU.HU\_bound} ((HU): one fixed schedule with
$\KL\le2\varepsilon/9$ and at most $1+\max\{8(\ln n+1),16\Lambda^3(\ln n+1)^2/\varepsilon\}$ rounds, for any $k$), and
\texttt{A36W.HL.LH\_theta} ($\LH^w=\Theta_{\varepsilon,\Lambda,\kappa_0}((\log n)^2)$, uniformly, for an interior quota);
\texttt{A36W.HL.HL\_hardLength}, \texttt{A36W.HL.HL\_hardLength\_asymptotic} and \texttt{A36W.HU.HU\_hardLength} give
the fixed-schedule half of the hard-length analogue of \S9, which \texttt{A36W.Attain.fixed\_schedule\_length\_theta}
states as one $\Theta_{\varepsilon,\Lambda,\kappa_0}((\ln n)^2)$ bound; \texttt{A36W.Attain.common\_length\_two\_sided}
gives its relabelled-profile half (the least common component length is $\log_2\ln n+O_{\varepsilon,\Lambda,\kappa_0}(1)$
for every $n\ge2$), with \texttt{A36W.Attain.LcomRel\_le\_lenFF} and \texttt{A36W.FU.FU\_theorem} for the sharp
construction at $n\ge n_0^{\mathrm{ff}}$. Item (FL) of the Result (\S5): \texttt{A36W.FL.FL\_r\_form} (the bound for every
integer $r\ge0$) and \texttt{A36W.FL.FL\_asymptotic} (its consequence $\LF^w\ge(e^{-\varepsilon}-o(1))\log_2\ln n$), with
\texttt{A36W.FL.FL\_r\_choice} for the choice of $r$ and \texttt{A36W.FL.V\_root\_ge\_attained} for the equality case; the
Remark after it: \texttt{A36W.FL.FL\_remark\_display\_LFgen} (its display) and \texttt{A36W.FL.FL\_remark\_n}
($\LF^w\ge e^{-\varepsilon}\log_2\ln n-4.154$ for $n\ge(2\Lambda/\kappa_0)^2$), with \texttt{A36W.FL.FL\_remark\_execMean}
and \texttt{A36W.FL.FL\_remark\_LFad} for $\E\tau$ and $\LF^{\mathrm{ad}}$. The \S6.9 table:
\texttt{A36W.FU.Table.table\_6\_9} (its fifteen entries, as exact values of the formal constants). The first failing
variant of \S6.9 (without uniform B-relabelling), as a joint mapping: \texttt{A36W.WN.orbit\_xa} (the residual slice
$\{1100,1010,1001\}$ is one orbit), \texttt{A36W.WN.FB\_fails} ($D(p_{B|A}\|Q_{B|A})>-\ln s_J$, with slice mass
$s_J=8/9$), with \texttt{A36W.WN.massOn\_QN}, \texttt{A36W.WN.c\_pos}, \texttt{A36W.WN.c\_eq\_QA} and
\texttt{A36W.FU.ExampleFB.succCount\_ex} (the count-process success probability from $(M,J)=(3,1)$ is $8/9$);
\texttt{A36W.WN.QN\_xa}, \texttt{A36W.WN.QN\_xb}, \texttt{A36W.WN.QN\_xc} and \texttt{A36W.WN.unrelabelled\_not\_uniform}
(the normalized restriction $(1/4,1/4,1/2)$, not uniform); \texttt{A36W.WN.condKL\_eq},
\texttt{A36W.WN.unrelabelled\_chain} and \texttt{A36W.WN.shape\_term\_pos} (the within-slice shape term); and
\texttt{A36W.WN.F\_strict} ($F$ strictly larger than for the uniformly B-relabelled counterpart of the displayed
schedule, as in \S6.1). The \S7 table of edge
cases, one lemma per row (three for the small-$\varepsilon$ row): \texttt{A36W.FU.Table.edge\_N0}, \texttt{A36W.FU.Table.edge\_N1},
\texttt{A36W.FU.Table.edge\_lambda\_one}, \texttt{A36W.FU.Table.edge\_N\_ll\_M}, \texttt{A36W.FU.Table.edge\_N\_eq\_M},
\texttt{A36W.FU.Table.edge\_J\_extreme}, \texttt{A36W.FU.Table.edge\_J\_infeasible}, \texttt{A36W.FU.Table.edge\_M\_small},
\texttt{A36W.FU.Table.edge\_small\_eps\_LA} ($L_A\le\lceil2\Lambda^2/\varepsilon\rceil$),
\texttt{A36W.FU.Table.edge\_small\_eps\_K4} and \texttt{A36W.FU.Table.edge\_small\_eps\_n0ff} (the two approximations, as
ratios tending to $1$), \texttt{A36W.FU.Table.edge\_small\_n} and \texttt{A36W.FU.Table.edge\_H\_side}. \S8's warning example, at $r=3$ and for every placement of group~A: \texttt{A36W.TGCert.W31a} (at $(2,2,2)$, $\lambda=2$:
$G^\lambda_3=1$, and $F^*_3$ lies in the rounding interval of $5.01857\cdot10^{-4}$) and \texttt{A36W.TGCert.W31b} (at
$(3,3,3)$, $\lambda=2$: $-\ln G^\lambda_3$ and $F^*_3$ lie in the rounding intervals of $0.05518$ and $0.05881$, and
$-\ln G^\lambda_3<F^*_3$). The uniform-in-$\varepsilon$ reading at the end of \S4: \texttt{A36W.PV.HU\_uniform\_eps} (one absolute constant $C$ with $1+\max\{8(\ln n+1),16\Lambda^3(\ln n+1)^2/\varepsilon\}\le C(\ln n+\Lambda^3(\ln n)^2/\varepsilon)$ for all $\Lambda\ge1$, $\varepsilon>0$ and $n\ge3$). The estimate $r=\log_2\ln n-O(\log\log\log n)$ of \S5, for the $r$ chosen there: \texttt{A36W.PV.rFL\_isBigO}, with the explicit bounds \texttt{A36W.PV.rFL\_two\_sided\_eventually} (eventually $0\le\log_2\ln n-r\le\frac{15}{2}\ln\ln\ln n$) and \texttt{A36W.PV.rFL\_bounds\_paper} ($\ln\ln\ln n\le\log_2\ln n-r\le\frac32\ln\ln\ln n+6$ for $n\ge3000$ and $n\ge(2\Lambda/\kappa_0)^2$), and \texttt{A36W.PV.rFL\_isTheta} (the $O$ is sharp). The second failing variant of \S6.9 (Stage~A in one block, at $\lambda=1$ and $k=n/2$): \texttt{A36W.PV.variant\_fails\_cell} (for every even $n\ge2000$ and $|A|=n/2$, the variant's law of $x_A$ is the product of $\mathrm{Ber}(\frac12)$ and $F>0.01$) and \texttt{A36W.PV.variant\_not\_eventually\_le} (so it cannot meet $\varepsilon=0.01$ uniformly), through \texttt{A36W.PV.cell\_DA\_le\_F} and \texttt{A36W.PV.chainRule\_route} ($F\ge D_A$, by data processing and by the chain rule of \S6.3), \texttt{A36W.PV.DAcell\_ge} (the displayed lower bound on $D_A$), with the steps of its printed derivation: \texttt{A36W.PV.DAcell\_eq\_count} ($D_A$ is the divergence between the hypergeometric and $\Bin(N,\frac12)$ laws of $w=|x_A|$), \texttt{A36W.PV.dv\_KLW} (the Donsker--Varadhan inequality), \texttt{A36W.PV.Epf\_eq} ($\E_pf=-n/(4(n-1))$), \texttt{A36W.PV.integral\_cos\_mul\_gaussian} and \texttt{A36W.PV.sum\_cos\_ones} (the Gaussian representation and the binomial cosine identity), \texttt{A36W.PV.sum\_exp\_fTest\_eq\_integral} ($\E_{\Bin}e^f=\E_g[\cos(g/\sqrt N)^N]$), \texttt{A36W.PV.cos\_le\_exp\_neg\_sq\_half} ($\cos x\le e^{-x^2/2}$ for $|x|\le\pi/2$) and \texttt{A36W.PV.binom\_exp\_le} ($\E_{\Bin}e^f\le2^{-1/2}+\Pr(|g|\ge\frac\pi2\sqrt N)$); and \texttt{A36W.PV.DAcell\_gt} ($D_A>0.096$ for every even $n\ge2000$), with \texttt{A36W.PV.Vb\_strictMonoOn}, \texttt{A36W.PV.Vb\_2000} and \texttt{A36W.PV.Vb\_tendsto} (the bound increases in $n$, lies in $[0.09644,0.09645]$ at $n=2000$ and tends to $\frac12\ln2-\frac14$, which \texttt{A36W.PV.limit\_value} places in $(0.0965,0.0966)$) and \texttt{A36W.PV.cell\_admissible} (the cell is admissible for every row of the table). Ordinary only: the
Informativeness paragraph of \S3, the computations of \S6.9 (the independent evaluation of the exact law, and the float64 grid of the second failing variant) and its remark on the size of the caps, the conclusion drawn from \S8's warning example, and the finite-$n$ figure of \S9 on the sharp mean coefficient.
\item Bounded weights without a reservoir (Appendix~\ref{app:rf}): \texttt{A36W.RF.Cut3.rf\_theoremA} (Theorem~A: items 1 and 3--5, item~2's hard-cap bounds and the schedule law's independence of $w$, for every $\Lambda\ge1$ and
$0<\kappa_0\le\frac12$, uniformly in the weights and $k$; the weights need not be normalized),
\texttt{A36W.RF.Final.rf\_small\_n\_branch} (item~2's exact sampler when $n<4m_0$) and
\texttt{A36W.RF.Cut3.rf\_corollaries} (Corollaries~A$'$ and~A$''$; \texttt{A36W.RF.Final.rf\_corA''\_uncond} gives
A$''$'s $\varepsilon_n\to0$, the range of its coefficient $1-\beta=e^{-(\varepsilon-\varepsilon_n)}$ for large $n$ and that
coefficient in $\mu'$), with \texttt{A36W.RF.Final.rf\_not\_common\_length}
(at $A=1$ and $n=20$, two components of positive probability have lengths $9$ and $1$, so the cap is not a common length).
Along the proof, among others: \texttt{A36W.RF.T3} ((2.2)), \texttt{A36W.RF.cutI\_2\_17} ((2.17)),
\texttt{A36W.RF.Cut2.powerSeries\_package} (the power series of \S3.2, on one radius), \texttt{A36W.RF.Cut2.eq\_3\_18\_19\_21\_chain},
\texttt{A36W.RF.Cut2.eq\_3\_20} and \texttt{A36W.RF.Cut3.eq\_3\_19\_all} ((3.18)--(3.21)), \texttt{A36W.RF.Cut2.rf\_3\_7} and \texttt{A36W.RF.Cut2.rf\_3\_8}
(\S\S3.7 and~3.8), \texttt{A36W.RFCandidate.rf\_ladder} (the integer ladder of \S4.1), \texttt{A36W.PV.rf\_sizes\_le}, \texttt{A36W.PV.rf\_coin\_count}, \texttt{A36W.PV.rf\_frac\_rat}, \texttt{A36W.PV.rf\_coinP\_eq} and \texttt{A36W.PV.rf\_support\_card\_le} (the sampling cost at the end of \S4.2: $R+1\le\log_2\log n+4$ sizes, exactly $nR$ coins with rational biases, independent with the product law, and at most $(R+1)^n$ schedules; the $O(n\log\log n)$ independent coins of Appendix~\ref{app:alt} follow by an ordinary step, and the remark that the support is never listed is ordinary only), \texttt{A36W.RF.Cut3.rf\_4\_5} (the
signed composition of \S4.5, for one parent and abstract continuation kernels), \texttt{A36W.RF.Cut3.oneParent\_holds} and
\texttt{A36W.RF.Cut3.sig420\_holds} ((4.20) at every good parent, for the actual suffix laws), \texttt{A36W.RF.Cut3.level\_420}
((4.20) at the level, after the supremum over good parents, from that per-parent bound), \texttt{A36W.RF.Cut3.rf\_constants} (the simultaneous choice of \S4.6) and \texttt{A36W.RF.Cut3.rf\_survival}
((4.31)--(4.32)). Lemma~T1: \texttt{A36W.sliceNCD\_up} and \texttt{A36W.sliceNCD\_down} (Lemma~NCD's inequalities (1.3a)),
\texttt{A36W.sliceNCD\_up\_supp} and \texttt{A36W.sliceNCD\_down\_supp} (its two support conditions), and
\texttt{A36W.slice\_subGaussian} and \texttt{A36W.slice\_tail} ((1.3)--(1.4), the latter for $c\ne0$), and \texttt{A36W.slice\_linStat\_eq\_zero\_of\_coeff\_zero} (for $c=0$ the statistic vanishes identically); \texttt{A36W.RF.T1Alt.t1\_route\_mg} and \texttt{A36W.RF.T1Alt.t1\_route\_mg\_tail} (the same bounds by route MG, Appendix~\ref{app:rf}'s alternative proof that integrates the variance bound (1.5) along exponential tilts); and \texttt{A36W.RF.T1Alt.lemmaT1\_both\_routes} and \texttt{A36W.RF.T1Alt.t1\_route\_mg\_same\_statement} (the two routes prove the same statements, with the same constant). For Appendix~\ref{app:t1}, \S10 without a
homogeneous group (the bounded-weights $F$ rows of Table~\ref{tab:classes}), uniformly in the weights and $k$:
\texttt{A36W.RFAttain.rf\_least\_adaptive\_hardcap} (the least adaptive hard cap is
$\log_2\ln n+O_{\varepsilon,\Lambda,\kappa_0}(1)$), \texttt{A36W.RFAttain.rf\_RwF\_two\_sided} (the explicit lower
constant of (c) with an existence-only upper constant $O_{\Lambda,\kappa_0}(1+\varepsilon^{-2})$),
\texttt{A36W.RFAttain.rf\_sharp\_mean\_two\_sided} (the least mean over predrawn mixtures is
$(e^{-\varepsilon}+o(1))\log_2\ln n$) and \texttt{A36W.RFAttain.rf\_sharp\_mean\_two\_sided\_ad} (the same for the
least expected executed rounds over causal policies); the mixtures of Corollaries~A$'$ and~A$''$ attain these
orders by \texttt{A36W.RFAttain.rf\_policy\_floor} and \texttt{A36W.RFAttain.rf\_muP\_bridge}. Further kernel-checked, for the remark after (3.9) on the first-order sum $S_1$: \texttt{A36W.RF.Cut2.firstOrder\_eq} and
\texttt{A36W.RF.Cut2.firstOrder\_zero} (the exact form of $S_1(u)$, and its value at $u=0$, for every $N$, $\pi$, $q$, $f$ and $u$ with
$0<q_i<1$); \texttt{A36W.RF.Cut2.firstOrder\_example\_pos} (on the remark's example, $N=2K$ with $K>0$, $J=K$, one weight $\lambda>1$
and the others $1$, with the slice's exact marginals as $\pi$, the logistic probabilities as $q$, and $0<f\le\frac14$: $S_1(0)>0$) and
\texttt{A36W.RF.Cut2.firstOrder\_not\_small} (no single constant $C$ gives $|S_1(u)|\le CfNu^2/V$ for all such $K$, $\lambda$, $f$ and all
$u$); \texttt{A36W.RF.Cut2.firstOrder\_lower} and \texttt{A36W.RF.Cut2.firstOrder\_upper} (for $0<a\le\frac12$ and $C_1,C_2\ge0$, each gives
one constant $C\ge0$ such that whenever $N>0$, all $\pi_i,q_i\in[a,1-a]$, $|\pi_i-q_i|\le C_1/V\le C_2/N$, $\sum_i(\pi_i-q_i)=0$
and $0<f\le\frac14$, every $u$ has $S_1(u)\ge-CfNu^2/V$, respectively $S_1(u)\le CN/V^2+CfNu^2/V$). For the conditions
listed after the table of \S5: \texttt{A36W.RF.Cut3.A0c} (a formula for $A_0$ from the one-step constants and $\kappa_0$) and
\texttt{A36W.RF.Cut3.A0c\_spec} (eight threshold inequalities that follow from $A\ge A_0$ when $\kappa_0,C,c,c_2>0$), with \texttt{A36W.RF.Cut3.rf\_constants}
above. Also kernel-checked, for Appendix~\ref{app:related}'s statement that the kept-set sampler of
Theorem~\ref{thm:tilt}(a) agrees in output law with a finite mixture of runs of random unmasking with prescribed block
sizes: \texttt{A36W.RFMixLaw.rfMix\_law\_eq\_relabel} (for every exact oracle and every word, the output law of the
mixture of \S4.1--4.2 equals that of its uniform relabelling, the mixture over its coin outcomes of runs with their block
sizes). Both identities of the construction's own-mean equality (4.37) are part of the conclusion of \texttt{A36W.RF.Cut3.rf\_corollaries}, which proves them by the direct block count after (4.36), without the lower-bound theorem. The steps of that count are stated by \texttt{A36W.RF.Final.rf\_mean\_ingredients} (for $A\ge1$ and $n\ge4m_0$: $\E|B_t|=M_t-M_{t+1}$, $\Pr(B_t=\varnothing)\le e^{-3M_{t+1}}$ with the bound $2e^{-12A^2}$ on the sum of these probabilities, the terminal contribution between $0$ and $2m_0$, and $R\le\log_2\log n+3$, with $R>\log_2\log n-3$ when $4A^4\le n$), and (4.37) itself, with the eventual bound $|\E\tau_{U(A_n)}-\log_2\log n|\le8A_n^2+5$, by \texttt{A36W.RF.Final.rf\_mean\_437}, for every $A_0\ge1$, $C_*$ and $\varepsilon>0$. Ordinary only: the remark of \S1.2, the rest of the remark of \S3.2, the strategy and dependency discussion of the two
routes to Lemma~T1 at the end of \S1.2, the rest of \S5 (its constants table, the conditions listed after it that the
declarations above do not state, and its closing remarks), and the dependency graph of \S6. Also kernel-checked along the printed routes: Lemma~NCD's reduction of its second family to the first through $1-X$, which has the weighted-slice law with weights $1/w_i$ and quota $|F|-J$ (\texttt{A36W.PV.RouteQ.compl\_law\_eq}, \texttt{A36W.PV.RouteQ.sliceNCD\_down\_printed}; \texttt{A36W.sliceNCD\_down} proves that family by its own induction); the proof of Lemma~T1 from NCD through the Cauchy--Schwarz display, which combines the two halves (\texttt{A36W.PV.RouteQ.cs\_display}; (1.3) and (1.4) as \texttt{A36W.PV.RouteQ.slice\_subGaussian\_printed} and \texttt{A36W.PV.RouteQ.slice\_tail\_printed}; \texttt{A36W.slice\_subGaussian} uses the pointwise bound \texttt{A36W.exp\_add\_le\_half} instead); and the explicit choice of concentration events after (3.16), through which (1.7) gives (3.16) from (3.15), the generative form of the tilted law (\texttt{A36W.PV.RouteQ2.printed\_events\_interior}, \texttt{A36W.PV.RouteQ2.eq\_3\_16\_printed}; \texttt{A36W.RF.Cut2.eq\_3\_16} uses relative-deviation events, through \texttt{A36W.RF.Cut2.ss\_bad\_le}). Ordinary only, as a route (its statement is kernel-checked by another route): the assembly of (2.17) from (2.10), (2.11) and (2.15) (\texttt{A36W.RF.cutI\_2\_17} goes through the tilt variable, by \texttt{A36W.RF.main\_term\_eq}; (2.15) is kernel-checked separately, as \texttt{A36W.RF.eq\_2\_15}).
\item Full support (Appendix~\ref{app:thmm}): \texttt{A36FS.Thm7C.thm3d}
(Theorem~\ref{thm:fs}(d)) and \texttt{A36FS.Thm7C.table1\_rate} (its $n^{3/(4^R-1)-o(1)}$ form, for all large $n$);
\texttt{A36FS.Thm7C.thm7c} (the displayed bound of Theorem~\ref{thm:mixf}(c)) and \texttt{A36FS.Thm7C.aNc\_le\_085}
(the bound $a_n\le0.85$ of Theorem~\ref{thm:mixf}(c) for all $n\ge14$ in range, at $\bar\omega=1/(3Z(\vartheta))$); \texttt{A36FS.Thm7.thm7b\_lower\_liminf}
(the lower half of Theorem~\ref{thm:mixf}(b)); \texttt{A36FS.Thm9.thm9b\_liminf} (Theorem~\ref{thm:fsadapt}(b));
\texttt{A36FS.Thm7.LFad\_le\_LFgen} ($\LF^{\mathrm{ad}}\le\LF$ for $\varepsilon\ge0$); \texttt{A36FS.Thm7.thm7a} and
\texttt{A36FS.Thm9.thm9a\_full} (Theorems~\ref{thm:mixf}(a) and~\ref{thm:fsadapt}(a)), with \texttt{A36FS.Thm7.LFgen\_ge} and \texttt{A36FS.Thm9.LFad\_ge} (the bounds on $\LF$ and $\LF^{\mathrm{ad}}$ in their last sentences); \texttt{U7Kol.U7\_printed}
(Lemma~U7); and, for Proposition~\ref{prop:hfs}, \texttt{A36FS.Prop2.prop2\_main} (its inequality),
\texttt{A36FS.Prop2.LHgen\_ge} and \texttt{A36FS.Prop2.LHgen\_le} (its two bounds on $\LH$),
\texttt{A36FS.Prop2.LHgen\_zero} ($\LH(n,0)=n$) and \texttt{A36FS.Prop2.LHgen\_isTheta} (Theorem~\ref{thm:fs}(a));
and, for the upper bounds, \texttt{A36FS.Thm8.Main.thm8} (Theorem~\ref{thm:mixu}; Lemma~Z5 is
\texttt{A36FS.Thm8.Transfer.transfer\_TV}), \texttt{A36FS.Thm8.Dilution.cor5} (Corollary~\ref{cor:mixexact}, which is
Theorem~\ref{thm:fs}(c)) and \texttt{A36FS.Thm8.Dilution.thm9c} (Theorem~\ref{thm:fsadapt}(c)); and, for single relabelled profiles, \texttt{A36FS.T1.thm5\_F\_exists}
(Theorem~\ref{thm:thmm}), \texttt{A36FS.T1.cor3\_F} (Corollary~\ref{cor:thmm}), \texttt{A36FS.T1.thm6\_F} (Theorem~\ref{thm:cf1}) and
\texttt{A36FS.T1.cor4\_F} (Corollary~\ref{cor:cf1} for $\varepsilon\le1$), with \texttt{A36FS.Misc.ProfAllEps.cor4\_F\_all} and
\texttt{A36FS.Misc.ProfAllEps.thm3b\_F\_all} (Corollary~\ref{cor:cf1} and Theorem~\ref{thm:fs}(b), respectively, for every $\varepsilon>0$), through
\texttt{A36FS.T1.F\_relabel\_eq\_Fcount} (T1); and \texttt{A36FS.HFS.thm7b\_upper} (the upper half of
Theorem~\ref{thm:mixf}(b)) and \texttt{A36FS.HFS.cor6} (Corollary~\ref{cor:hfs}); and \texttt{FSCore.RemMixf.rem\_mixf\_incl\_ge\_two}, \texttt{FSCore.RemMixf.rem\_mixf\_incl\_lt\_two} and
\texttt{FSCore.RemMixf.rem\_mixf\_limit\_four} (the two inclusions and the limit of Remark~\ref{rem:mixf}), with
\texttt{FSCore.RemMixf.not\_eventually\_mass\_le\_of\_le\_four}, \texttt{FSCore.RemMixf.const\_lt\_half} and
\texttt{FSCore.RemMixf.etaInf\_le\_one}, on which the remark's closing sentence, on the limit of the method, rests (for
$1<\vartheta\le4$ the test-set premise fails; for $\vartheta>4$ the constant $(\ln2/\ln\vartheta)\,\eta_\infty$ of
Theorem~\ref{thm:mixf}(b)'s proof is below $\frac12$); and, for lemmas and remarks of
Appendix~\ref{app:thmm}, \texttt{A36FS.PropL.propL} (Proposition~L), \texttt{A36FS.XP.X1}, \texttt{A36FS.XP.X3i}
and \texttt{A36FS.XP.X3ii} (Lemmas~X1 and X3), \texttt{A36FS.X2Paper.X2\_paper} and \texttt{A36FS.X2Paper.X2\_R0}
(Lemma~X2: the forced jump, and the formula for $R_0(n)$), \texttt{A36FS.XP.X1\textquotesingle} and
\texttt{A36FS.XP.X3\textquotesingle} (Lemmas~X1$'$ and X3$'$, counting completed runs, which is their use in the proof
of Theorem~\ref{thm:fsadapt}(a)), \texttt{A36FS.K2.main\_text\_scope} (the floors, top floor, minimum prefix and
hard-cap scope of the mixture of Theorem~\ref{thm:mixu}), \texttt{A36FS.Misc.MixCap.mix\_hardcap\_lower} and
\texttt{A36FS.Misc.MixCap.mix\_hardcap\_lower\_cap} (the least hard cap of finite predrawn mixtures with $F\le\varepsilon$ is at least $(\frac12-o(1))\log_2\log_2n$, from Theorem~\ref{thm:mixf}(a); with \texttt{A36FS.Thm9.thm9a\_full}, since a policy in $\mathcal A_r$ completes, and has $\tau_+\le\tau\le r$, almost surely, the same lower bound $(\frac12-o(1))\log_2\log_2n$ holds for the least hard cap of causal policies with $\KL(p\|Q_\pi)\le\varepsilon$, a derived consequence that no single declaration states), \texttt{A36FS.Misc.DepthBridge.callComps\_chain} and \texttt{A36FS.Misc.DepthBound.depth\_bound} (on every branch of the
construction's recursion the nonbase calls form a chain of nested calls, empty at a base size, and every such chain
from size $n$ has fewer than $1+\log_4\ln(n+2)$ calls), with \texttt{A36FS.Misc.DepthBridge.depth\_every\_branch} (every component of the
recursion's law comes from such a branch), \texttt{A36FS.Misc.ChargeMono.LFgen\_le\_charged}
and \texttt{A36FS.Misc.ChargeMono.LFad\_le\_charged} (an extra nonnegative charge can only raise $\LF$ and
$\LF^{\mathrm{ad}}$), with \texttt{A36FS.Misc.ChargeMono.thm7a\_bound\_charged} and
\texttt{A36FS.Misc.ChargeMono.thm9b\_charged} (Theorem~\ref{thm:mixf}(a)'s bound and Theorem~\ref{thm:fsadapt}(b)
under such a charge), \texttt{A36FS.Misc.ExactSamplerPS.exactSampler\_exists} and
\texttt{A36FS.Misc.ExactSamplerNonProduct.not\_first\_round} (the exact sampler and, for $n\ge2$, its contrast with
one interface round), and \texttt{A36FS.Misc.ScopeL3.W1\_counterexample},
\texttt{A36FS.Misc.Round1Alpha.alpha1\_exact\_min} and
\texttt{A36FS.Misc.Thm9aModerate.thm9a\_bound\_le\_two} (the Scope paragraph's counterexample to Lemma~L3, the
remark in round~1 of the proof of Theorem~\ref{thm:thmm}, and the moderate-$n$ remark after
Theorem~\ref{thm:fsadapt}).
The declarations above prove Theorem~\ref{thm:mixf} by other routes: parts~(a) and~(c) and the lower half of~(b)
through the embedding of mixtures as causal policies, and the upper half of~(b) through Corollary~\ref{cor:mixexact}.
The printed arguments of its parts are kernel-checked as well, with the same statements: \texttt{A36FS.PrintedRoute.thm7a\_printed\_route}
(part~(a)), \texttt{A36FS.PrintedRoute.thm7b\_lower\_liminf\_printed\_route} and
\texttt{A36FS.PrintedRoute.thm7b\_upper\_printed\_route} (part~(b)), and \texttt{A36FS.PrintedRoute.thm7c\_printed\_route}
and \texttt{A36FS.PrintedRoute.thm3d\_printed\_route} (part~(c) and Theorem~\ref{thm:fs}(d)). Its count reduction X0 is
\texttt{A36FS.X0.X0\_mixture\_le} for every finite predrawn mixture (in count-marginal form,
\texttt{A36FS.X0.KLx\_symW\_eq\_KLW}), with equality under the uniform relabelling (\texttt{A36FS.X0.X0\_relabel\_eq}). In the proof of
Theorem~\ref{thm:thmm}, the printed derivation of both Case-1 conditions on $r$ from $r\ge K_1\sqrt{1+\ln c}$ is
kernel-checked as well: \texttt{A36FS.Misc.Thm5Case1Route.printed\_chain} and
\texttt{A36FS.Misc.Thm5Case1Route.case1\_conditions} (for $K_0\ge80e^2$ and $c\ge1$), and
\texttt{A36FS.Misc.Thm5Case1Route.case1\_conditions\_paper} with \texttt{A36FS.Misc.Thm5Case1Route.Cc\_side} (at
$K_0=4608e^2$ and $K_1=2K_0$).
\end{itemize}\par}
\pend{}

\end{document}